\documentclass[a4paper,11pt]{book}
\usepackage[T1]{fontenc}
\usepackage[utf8]{inputenc}
\usepackage{imakeidx}
\makeindex

\usepackage{lmodern}

\usepackage{pgfplots}
\pgfplotsset{compat=1.18}

\usepackage{graphicx}
\usepackage{subcaption}
\usepackage{tikz-network}
\usepackage{tikz}
\usepackage{tikz-3dplot}
\usetikzlibrary{decorations.pathmorphing}

\usepackage[section]{algorithm}
\usepackage{algorithmic}

\usepackage{amsmath,amsthm,amssymb}
\usepackage{booktabs}

\usepackage{cases}
\usepackage{mathfont}

\usepackage[sort]{cite}

\newtheorem{theorem}{Theorem}[section]
\newtheorem{corollary}[theorem]{Corollary}
\newtheorem{proposition}[theorem]{Proposition}
\newtheorem{lemma}[theorem]{Lemma}

\theoremstyle{definition}
\newtheorem{definition}[theorem]{Definition}
\newtheorem{assumption}[theorem]{Assumption}
\newtheorem*{remark}{Remarks}
\newtheorem{example}[theorem]{Example}

\DeclareMathOperator*{\argmax}{arg\,max}
\DeclareMathOperator*{\argmin}{arg\,min}
\DeclareMathOperator{\var}{Var}
\DeclareMathOperator{\cov}{Cov}

\newcommand{\sumid}{\sum_{i=1}^{d}}
\newcommand{\dcube}{[0,1]^{d}}
\newcommand{\pem}{p_{\epsilon,M}}
\newcommand{\pdm}{p_{\delta,M}}
\newcommand{\qd}{q_{\delta}}
\newcommand{\supp}{\mathrm{supp}}
\newcommand{\phim}{\phi_{\mbf}}
\newcommand{\phimx}{\phi_{\mbf}(\xbf)}
\newcommand{\psii}{\psi\Big(3N \big(x_i - \frac{m_i}{N}\big) \Big)}
\newcommand{\nximi}{3N \big(x_{i} - \frac{m_i}{N} \big)}
\newcommand{\pmx}{P_{\mbf}(\xbf)}
\newcommand{\amkf}{a_{\mbf,\kbf}^{f}}
\newcommand{\xmk}{\Big(\xbf - \frac{\mbf}{N} \Big)^{\kbf}}
\newcommand{\hmkx}{h_{\mbf,\kbf}(\xbf)}

\newcommand{\vkN}{v_{k}^{N}}

\newcommand{\xdot}{\dot{x}}
\newcommand{\ydot}{\dot{y}}
\newcommand{\pdot}{\dot{p}}
\newcommand{\zdot}{\dot{z}}
\newcommand{\xeps}{x^{\epsilon}}

\newcommand{\ueps}{u^{\epsilon}}
\newcommand{\zeps}{z^{\epsilon}}
\newcommand{\xepsdot}{\dot{x}^{\epsilon}}

\newcommand{\fbar}{\bar{f}}
\newcommand{\gbar}{\bar{g}}
\newcommand{\pbar}{\bar{p}}
\newcommand{\xbar}{\bar{x}}
\newcommand{\ybar}{\bar{y}}

\newcommand{\rhoeps}{\rho^{\epsilon}}
\newcommand{\rhostar}{\rho^{*}}

\newcommand{\rhoteps}{\rho_{t}^{\epsilon}}
\newcommand{\rhotstar}{\rho_{t}^{*}}

\newcommand{\sigmat}{\sigma_{t}}
\newcommand{\sigmatau}{\sigma_{\tau}}

\newcommand{\rhoseps}{\rho_{s}^{\epsilon}}
\newcommand{\rhosstar}{\rho_{s}^{*}}

\newcommand{\rhotaueps}{\rho_{\tau}^{\epsilon}}
\newcommand{\rhotaustar}{\rho_{\tau}^{*}}
\newcommand{\rhotaueeps}{\rho_{\tau-\epsilon}^{\epsilon}}
\newcommand{\rhotauestar}{\rho_{\tau-\epsilon}^{*}}

\newcommand{\uteps}{u_{t}^{\epsilon}}
\newcommand{\utstar}{u_{t}^{*}}
\newcommand{\ustar}{u^{*}}

\newcommand{\utt}{u_{\theta(t)}}
\newcommand{\ue}{u^{\epsilon}}
\newcommand{\utaustar}{u_{\tau}^{*}}

\newcommand{\pis}{\pi^{*}}
\newcommand{\vs}{v^{*}}
\newcommand{\xs}{x^{*}}
\newcommand{\qs}{q^{*}}
\newcommand{\Ts}{T^{*}}

\newcommand{\vpi}{v^{\pi}}
\newcommand{\qpi}{q^{\pi}}
\newcommand{\Tpi}{T^{\pi}}

\newcommand{\kl}{\text{KL}}
\newcommand{\js}{\text{JS}}
\newcommand{\elbo}{\text{ELBO}}
\newcommand{\pdata}{p_{\text{data}}}
\newcommand{\pinit}{p_{\text{init}}}

\newcommand{\ftheta}{f_{\theta}}
\newcommand{\geta}{g_{\eta}}
\newcommand{\ptheta}{p_{\theta}}
\newcommand{\qeta}{q_{\eta}}
\newcommand{\pet}{p_{\eta,\theta}}

\newcommand\blfootnote[1]{%
  \begingroup
  \renewcommand\thefootnote{}\footnote{#1}%
  \addtocounter{footnote}{-1}%
  \endgroup
}

\usepackage{hyperref}
\usepackage{graphicx}
\usepackage[english]{babel}

\newenvironment{dedication}
{
   \cleardoublepage
   \thispagestyle{empty}
   \vspace*{\stretch{1}}
   \hfill\begin{minipage}[t]{0.66\textwidth}
   \raggedright
}
{
   \end{minipage}
   \vspace*{\stretch{3}}
   \clearpage
}

\makeatletter
\renewcommand{\@chapapp}{}% Not necessary...

\makeatother

\title{\Huge \textbf{Mathematical Foundations of Deep Learning}\blfootnote{Draft version. Final version is published in ``Chapman \& Hall/CRC Mathematics and Artificial Intelligence Series'' by Taylor \& Francis in 2026.}\\ \bigskip {\LARGE Theory and Algorithms}}
\author{\textsc{Xiaojing Ye}}

\begin{document}

\frontmatter
\maketitle

\begin{verbatim}
@book{ye2026mathematical,
  title = {Mathematical Foundations of Deep Learning},
  author = {Ye, Xiaojing},
  year = {2026},
  publisher = {Chapman and Hall/CRC}
}
\end{verbatim}

%%%%%%%%%%%%%%%%%%%%%%%%%%%%%%%%%%%%%%%%%%%%%%%%%%%%%%%%%%%%%%%
% Add a dedication paragraph to dedicate your book to someone %
%%%%%%%%%%%%%%%%%%%%%%%%%%%%%%%%%%%%%%%%%%%%%%%%%%%%%%%%%%%%%%%
\begin{dedication}
To my family
\end{dedication}

%%%%%%%%%%%%%%%%%%%%%%%%%%%%%%%%%%%%%%%%%%%%%%%%%%%%%%%%%%%%%%%%%%%%%%%%
% Auto-generated table of contents, list of figures and list of tables %
%%%%%%%%%%%%%%%%%%%%%%%%%%%%%%%%%%%%%%%%%%%%%%%%%%%%%%%%%%%%%%%%%%%%%%%%
\tableofcontents

% \listoffigures
% \listoftables

\mainmatter

%%%%%%%%%%%
% Preface %
%%%%%%%%%%%
\chapter*{Preface}

Over the past fifteen years, deep learning has rapidly evolved into one of the central pillars of modern science and engineering, and it is now widely regarded as the most promising approach toward general artificial intelligence. Despite its remarkable empirical success, theoretical understanding has lagged far behind practical advances for much of this period. This book is based on a collection of literature and notes I took in the past years. It is written in response to a growing need, particularly from people interested in learning and understanding deep learning from mathematical perspective: A relatively coherent, principled, and mathematically grounded account of deep learning that focuses on the underlying structures, assumptions, and theoretical guarantees of deep learning models.

At its core, deep learning is a mathematical enterprise. Neural networks are function approximators; training corresponds to solving large-scale non-convex optimization problems; generalization is fundamentally a question of probability and statistics; and expressivity, stability, and robustness are governed by ideas from approximation theory, functional analysis, and dynamical systems. The goal of this book is to provide some mathematical foundations needed to understand why deep neural networks work, how to train them effectively, and how to leverage deep learning to solve problems arising from machine learning, scientific computing, automatic control, etc.

I hope this book can be beneficial to people in several fields. For readers with a background in mathematics, it provides an entry point into deep learning that respects rigor and abstraction while remaining closely connected to modern practice. For readers trained primarily in machine learning or engineering, it offers a deeper theoretical perspective, unifying familiar concepts under precise mathematical formulations. Throughout the book, I tend to emphasize clarity of assumptions, careful definitions, and explicit connections between theory and algorithms.

The materials in this book are organized around several foundational themes. This book begins by viewing neural networks as parametric function classes and studying their expressive power through the lens of approximation theory. Then the discussion focuses on optimization theory and numerical algorithms, which underpin the training of deep neural networks. Subsequent chapters explore the integration of deep learning with optimal control, reinforcement learning, and generative modeling, highlighting the mathematical principles that govern their design and behavior. Wherever possible, connections are drawn between classical results and contemporary developments.

This book is definitely not exhaustive nor intended to be a comprehensive survey. Instead, it should be considered as an introduction to deep learning through a mathematics-oriented approach. It emphasizes ideas, concepts, and results that are likely to remain relevant as the field continues to evolve extremely fast. Proofs are provided for central results, and examples and remarks are included to build intuition and to illustrate how abstract theory informs practical modeling choices. 

The intended audience includes graduate students (and advanced undergraduates) and researchers in mathematics, statistics, computer science, electrical engineering, and related fields. A solid background in the standard sequence of calculus courses, linear algebra, and probability is required, while familiarity with advanced calculus and real analysis will be very helpful. 

Deep learning continues to reshape both applied sciences and theoretical inquiry. By presenting from a mathematical perspective, I hope this book will help readers develop a deeper understanding of this subject, critically assess emerging ideas, and contribute to the ongoing effort to place deep learning on firm theoretical ground.

I wish to thank Yunmei Chen, Hongyuan Zha, and Haomin Zhou for many years of fruitful collaboration and insightful discussions in the field of mathematics and deep learning, and also acknowledges my excellent students and collaborators, including Nathan Gaby, Shu Liu, Hao Wu, Yaohua Zang, and many others, for the valuable discussions and joint work these years. I am grateful to National Science Foundation for the support of my research.

\mbox{}\\
\mbox{}\\
\noindent Xiaojing Ye \\
\noindent Atlanta, Georgia, USA \\

%\section*{Another sample section}
%Lorem ipsum dolor sit amet, consectetur adipiscing elit. Duis risus ante, auctor et pulvinar non, posuere ac lacus. Praesent egestas nisi id metus rhoncus ac lobortis sem hendrerit. Etiam et sapien eget lectus interdum posuere sit amet ac urna. Aliquam pellentesque imperdiet erat, eget consectetur felis malesuada quis. Pellentesque sollicitudin, odio sed dapibus eleifend, magna sem luctus turpis, id aliquam felis dolor eu diam. Etiam ullamcorper, nunc a accumsan adipiscing, turpis odio bibendum erat, id convallis magna eros nec metus.
%
%\section*{Structure of book}
%% You might want to add short description about each chapter in this book.
%Each unit will focus on <SOMETHING>.
%
%\section*{About the companion website}
%The website\footnote{\url{https://github.com/amberj/latex-book-template}} for this file contains:
%\begin{itemize}
%  \item A link to (freely downlodable) latest version of this document.
%  \item Link to download LaTeX source for this document.
%  \item Miscellaneous material (e.g. suggested readings etc).
%\end{itemize}
%
\newpage

\section*{List of Common Notations}

\begin{center}
\begin{tabular}{ll}
\toprule
Notation & Meaning \\
\midrule
$A := B$ & $A$ is defined to be $B$ (or $B =: A$) \\
s.t. & subject to \\
$\mathbb{N}$ & the set of natural numbers $\{1,2,\dots\}$ \\
$\mathbb{R}$ & the set of real numbers \\
$\mathbb{R}^{n}$ & the $n$-dimensional real Euclidean space \\
$[N]$ & $\{1,2,\dots,N\}$ for any $N \in \mathbb{N}$\\
$I_{d}$ & the $d$-by-$d$ identity matrix \\
$A^{\top}$ & (conjugate) transpose of (complex) matrix $A$ \\
$\mathrm{tr}(A)$ & trace of square matrix $A$ \\
$A \succeq B$ & $A-B$ is positive semi-definite (positive definite if $\succ$) \\
$| \cdot |$ & absolute value, or vector and matrix 2-norm \\
$\| \cdot \|_{*}$ & function norm specified by subscript $*$ \\
$x \odot y$ & componentwise product of $x, y \in \mathbb{R}^{n}$ \\
$\langle x, y \rangle$ & inner product of $x$ and $y$ in $\mathbb{R}^{n}$ or a Hilbert space \\
$x'(t)$ or $\dot{x}(t)$ & derivative of a function $x(t)$ with respect to $t$ \\
$\nabla_{\theta}$ & gradient with respect to $\theta$ ($\nabla = \nabla_{x}$ for short) \\
$\nabla \cdot$ & divergence with respect to $x$ \\
$\nabla^{2}$ & Hessian with respect to $x$ \\
$\Delta$ & Laplacian with respect to $x$ \\
$C^{k}$ & the set of $k$ times continuously differentiable functions \\
$\argmin$ ($\argmax$) & the set of minimizers (maximizers) \\
$X \sim p$ & $X$ is a random variable following probability distribution $p$ \\
$x \sim p$ & $x$ is a sample drawn from the probability distribution $p$ \\
i.i.d. & independent and identically distributed \\
$\Ncal(\mu, \Sigma)$ & Gaussian distribution with mean $\mu$ and variance $\Sigma$ \\
$\Ncal(\cdot; \mu, \Sigma)$ & probability density function of $\Ncal(\mu; \Sigma)$ \\
$\mathbb{E}[f(X)]$ & expectation of $f(X)$ \\
$\mathbb{E}_{X \sim p(\cdot|Y)}[f(X,Y)]$ & conditional expectation of $f(X,Y)$ given $Y$ \\
$\var[f(X)]$ & variance of $f(X)$ \\
\bottomrule
\end{tabular}
\end{center}

%%%%%%%%%%%%%%%%%%%%%%%%%%%%%%%%%%%%%
%% Give credit where credit is due. %
%% Say thanks!                      %
%%%%%%%%%%%%%%%%%%%%%%%%%%%%%%%%%%%%%
%\section*{Acknowledgements}
%\begin{itemize}
%\item A special word of thanks goes to Professor Don Knuth\footnote{\url{http://www-cs-faculty.stanford.edu/~uno/}} (for \TeX{}) and Leslie Lamport\footnote{\url{http://www.lamport.org/}} (for \LaTeX{}).
%\item I'll also like to thank Gummi\footnote{\url{http://gummi.midnightcoding.org/}} developers and LaTeXila\footnote{\url{http://projects.gnome.org/latexila/}} development team for their awesome \LaTeX{} edIt\^{o}rs.
%\item I'm deeply indebted my parents, colleagues and friends for their support and encouragement.
%\end{itemize}
%\mbox{}\\
%%\mbox{}\\
%\noindent Amber Jain \\
%\noindent \url{http://amberj.devio.us/}

%%%%%%%%%%%%%%%%
% NEW CHAPTER! %
%%%%%%%%%%%%%%%%

\chapter{Deep Neural Networks}
\label{chpt:dnn}

Deep neural networks are the central building blocks of modern deep learning systems. From a mathematical perspective, they serve as highly expressive function approximators whose architectures are inspired by the layered organization of biological neural systems and motivated by the need to learn hierarchical representations directly from data. 

This chapter begins by introducing the motivations and foundational concepts of deep neural networks. We then focus on one of the most fundamental results in neural network theory, known as the Universal Approximation Theorem, and provide a rigorous proof with quantitative error estimate when approximating a general class of continuous functions defined on compact subsets of Euclidean spaces.

Then we review commonly used activation functions and network blocks that form the core components of modern deep neural network architectures. This is intended to give readers without extensive practical experience an intuitive understanding of how deep networks are constructed. More importantly, we introduce a number of illustrative techniques for designing networks that satisfy prescribed function properties. Such constructions often lead to significant improvements in network training efficiency and empirical performance. Finally, we briefly discuss objective functions used for training neural networks, and provide some inspiring examples that formulate objective functions based on laws of physics without any given data.

\section{Function Approximations}

Function approximators have been extensively used in statistics and data science.
We first recall two classical examples of function approximators for linear regression and logistic classification in statistics.

\begin{example}
[Linear regression]
\label{ex:regression}
Suppose there are $d$ variables $x_1, \dots, x_d \in \mathbb{R}$ called \emph{features}, which combined form a \emph{feature vector} $x = (x_{1},\dots, x_{d})^{\top} \in \mathbb{R}^{d}$. 
In addition, there is an observation $y \in \mathbb{R}$ known to be an affine function\index{Affine function} of these variables plus some noise:
\begin{equation}
\label{eq:linear-regression-model}
    y = \sumid w_i x_i + b + e ,
\end{equation}
where $w = (w_1, \dots, w_d) \in \mathbb{R}^{d}$ and $b$ are called the \emph{weights} and \emph{bias}, respectively, which are unknown.
In \eqref{eq:linear-regression-model}, we assume that the noise $e$ follows the Gaussian distribution $\mathcal{N} ( 0, s^2)$ with mean 0 and some unknown standard deviation $s$. 
Namely, the probability distribution of $e$ is
\begin{equation*}
\Ncal(e; 0, s^{2}) := \frac{1}{\sqrt{2 \pi s^{2}} } e^{- \frac{|e|^{2}}{2 s^{2}}} .
\end{equation*}
The question is: How do we estimate $(w, b)$ using this sample $(x,y)$ and their relationship given in \eqref{eq:linear-regression-model}?

Clearly, we cannot estimate $(w,b)$ from a single sample $(x,y)$. Instead, we need a dataset $\Dcal$ of multiple (say $M \in \mathbb{N}$) samples:
\begin{equation*}
\Dcal = \Big\{ (x^{(j)}, y^{(j)}) \in \mathbb{R}^{d} \times \mathbb{R}: \ j \in [M] \Big\} , 
\end{equation*}
where each sample $(x^{(j)}, y^{(j)})$ follows the same affine relation in \eqref{eq:linear-regression-model} with independent and identically distributed (i.i.d.) Gaussian noises $ e^{(j)} \sim \mathcal{N}(0, s^2)$ for $j \in [M]:=\{1,2,\dots,M\}$.

For notation simplicity, we denote by $\theta$ the concatenation of $w$ and $b$ as a $(d+1)$-dimensional column vector:
\begin{equation*}
\theta := \begin{pmatrix}
w \\ b
\end{pmatrix} \in \mathbb{R}^{d+1} .
\end{equation*}
Then, with the dataset $\Dcal$ and 
\begin{equation*}
y^{(j)} - \theta^{\top} \begin{pmatrix}
x^{(j)} \\ 1
\end{pmatrix}
= 
y^{(j)} - (w^{\top} x^{(j)} + b)
= e^{(j)} \sim \Ncal (0, s^{2})
\end{equation*}
for all $j \in [M]$, we have the \emph{likelihood function}\index{Likelihood function} $L$ of $\theta$ defined by
\begin{align*}
L( \theta ) 
& := \prod_{j=1}^{M} \Ncal (e^{j}; 0, s^{2}) \\
& \;= \prod_{j=1}^{M} \Ncal (y^{(j)} - w^{\top} x^{(j)} - b; 0, s^{2}) \\
& \;= (2 \pi s^2 )^{-\frac{M}{2}} e^{-\frac{1}{2 s^2} \sum_{j=1}^{M} |y^{(j)} - w^{\top} x^{(j)} - b|^2} .
\end{align*}
The maximum likelihood estimate method commonly used in statistics suggests that the optimal $\theta^{*}$ is the maximizer of $L(\theta)$.
Since the negative of logarithm is strictly decreasing, $\theta^{*}$ is also the minimizer of the \emph{negative log-likelihood function}\index{Likelihood function!log-likelihood function} $\ell$:
\begin{align*}
\ell (\theta) 
& := - \log L(\theta) \\
& \;= \frac{1}{2 s^2} \sum_{j=1}^{M} |y^{(j)} - w^{\top} x^{(j)} - b|^2 + \frac{M}{2} \log (2 \pi s^2) .
\end{align*}
Then we obtain the \emph{maximum likelihood estimate}\index{Maximum likelihood estimate} $\theta^{*}$ as:
\begin{equation}
\label{eq:linear-regression-minimization}
\theta^{*} = \begin{pmatrix}
w^{*} \\ b^{*}
\end{pmatrix}
= \argmin_{(w,b)} \frac{1}{2 s^2} \sum_{j=1}^{M} |y^{(j)} - w^{\top} x^{(j)} - b|^2  .
\end{equation}
Note that the unknown standard deviation $s$ does not affect the solution to \eqref{eq:linear-regression-minimization} and can be omitted hereafter. 
Now we denote 
\begin{equation*}
X = 
\begin{pmatrix}
x^{(1)\top} & 1\\
\vdots & \vdots \\
x^{(M)\top} & 1
\end{pmatrix}
\in \mathbb{R}^{M \times (d+1)}
\quad \text{and} \quad 
Y = \begin{pmatrix}
y^{(1)} \\ \vdots \\ y^{(M)}
\end{pmatrix} \in \mathbb{R}^{M} ,
\end{equation*}
then we can rewrite the minimization problem \eqref{eq:linear-regression-minimization} in a concise form:
\begin{equation}
\label{eq:linear-regression-minimization-mtx}
\min_{\theta \in \mathbb{R}^{d+1}} \ \frac{1}{2}| Y - X\theta |^{2} .
\end{equation}
Taking the gradient of the function to be minimized in \eqref{eq:linear-regression-minimization-mtx}, we obtain 
\begin{equation}
\label{eq:linear-regression-minimization-normal-eq}
X^{\top} X \theta = X^{\top} Y .
\end{equation}
We know that minimizers of \eqref{eq:linear-regression-minimization-mtx} are among the critical points satisfying \eqref{eq:linear-regression-minimization-normal-eq}.

Note that, when $M$ is sufficiently large and $x^{(j)}$'s are sufficiently diverse, which often hold true if the problem is set up correctly and the data samples are collected properly, then $X$ has full column rank $d+1$.
(More precisely, $x^{(1)}, \dots, x^{(M)}$ span a $d$-dimensional linear subspace of $\mathbb{R}^{d+1}$ not including $1 = (1,\dots,1)^{\top} \in \mathbb{R}^{d+1}$, and consequently $X$ has full column rank.)
As a result, $X^{\top}X$ is invertible, and \eqref{eq:linear-regression-minimization-normal-eq} has only one solution
\begin{equation*}
\theta^{*} = \begin{pmatrix}
w^{*} \\ b^{*}
\end{pmatrix} 
= (X^{\top} X)^{-1} X^{\top} Y ,
\end{equation*}
which is the unique minimizer of \eqref{eq:linear-regression-minimization} or equivalently \eqref{eq:linear-regression-minimization-mtx} (since \eqref{eq:linear-regression-minimization-mtx} is clearly a convex function of $\theta$).
This is how linear regression ``learns'' the optimal value of $w$ and $b$ in \eqref{eq:linear-regression-model} using the dataset $\Dcal$.
\end{example}

\begin{example}
[Logistic classification]
\label{ex:classification}
Suppose for every $d$-dimensional feature vector $x=(x_{1},\dots,x_{d})^{\top} \in \mathbb{R}^{d}$, it has a binary-valued classification identifier $y \in \{0,1\}$ called the \emph{label}.
The label $y=1$ when the feature vector $x$ is known to be in the ``True'' class, and $y=0$ if $x$ is in the ``False'' class.

In the standard logistic classification (also called logistic regression). If a feature vector $x$ is predicted to be in the ``True'' class, i.e., $\mathrm{Pr}(y=1)$ for the label $y$ of $x$, is assumed to be
\begin{equation}
\label{eq:logistic-classification-model}
\mathrm{Pr}(y = 1) = \sigma( w^{\top} x + b) ,
\end{equation}
where
\begin{equation}
\label{eq:logistic-fn}
\sigma(z) := \frac{1}{1 + e^{-z}}, \qquad \forall \, z \in \mathbb{R}
\end{equation}
is called the \emph{logistic function}\index{Logistic function}, or more commonly known as the \emph{sigmoid function}.
Notice that the range of $\sigma$ is $(0,1)$ which can be used to describe a probability value.
We know that
\begin{equation*}
\mathrm{Pr}(y = 0) = 1 - \mathrm{Pr}(y=1) = 1 - \sigma( w^{\top} x + b) .
\end{equation*}
In other words, $y$ can be interpreted as a sample following the Bernoulli distribution with parameter $\sigma(w^{\top} x + b) \in (0,1)$. 
Recall that if $z$, which takes value either 1 or 0, is a sample following the probability distribution $\text{Bernoulli}(p)$ with $p \in (0,1)$, then $\mathrm{Pr}(Z=z) = p^{z}(1- p)^{1-z}$, which is the expression of the probability mass function of the random variable $Z \sim \text{Bernoulli}(p)$ in one formula.

For notation simplicity, we again denote
\begin{equation*}
\theta := \begin{pmatrix}
w \\ b
\end{pmatrix} \in \mathbb{R}^{d+1} .
\end{equation*}
Now we also need a dataset $\Dcal$ of $M$ i.i.d.\ samples 
\begin{equation*}
\Dcal = \Big\{ (x^{(j)}, y^{(j)}) \in \mathbb{R}^{d} \times \{0,1\}: \ j \in [M] \Big\}
\end{equation*}
with each sample $(x^{(j)}, y^{(j)})$ following the relation in \eqref{eq:logistic-classification-model}.
Then we have the likelihood function $L$ of $\theta$:
\begin{equation*}
L(\theta) = \prod_{j=1}^{M} \big( \sigma(w^{\top} x^{(j)} + b) \big)^{y^{(j)}} \big( 1 - \sigma(w^{\top} x^{(j)} + b) \big)^{1 - y^{(j)}} ,
\end{equation*}
and the negative log-likelihood function $\ell$ of $\theta$:
\begin{align}
\ell(\theta) 
& = - \log L(\theta) \label{eq:nll-logistic}\\
& = - \sum_{j=1}^{M} \Big( y^{(j)} \log \sigma(w^{\top} x^{(j)} + b) + (1 - y^{(j)}) \log (1 - \sigma(w^{\top}x^{(j)} + b) ) \Big) . \nonumber
\end{align}

To find the minimizer of $\ell$ in \eqref{eq:nll-logistic}, we first derive a few simple results. We notice that the derivative of $\sigma$ is
\begin{equation*}
\sigma'(z) = \frac{e^{-z}}{( 1 + e^{-z})^2} = \sigma(z) (1 - \sigma(z)) .
\end{equation*}
We define the function $h: \mathbb{R}^{d} \times \mathbb{R}^{d+1} \to \mathbb{R}$ by
\begin{equation*}
h(x, \theta) := \sigma( w^{\top} x + b ) .
\end{equation*}
Then there is
\begin{align*}
\nabla_{\theta} h(x,\theta) 
& = \sigma'(w^{\top} x + b)
\begin{pmatrix}
x \\ 1
\end{pmatrix}^{\top} \\
& = \sigma(w^{\top} x + b) \bigl(1 - \sigma(w^{\top} x + b) \bigr)
\begin{pmatrix}
x \\ 1
\end{pmatrix}^{\top} \in \mathbb{R}^{1 \times (d+1)} ,
\end{align*}
and hence 
\begin{align*}
\nabla_{\theta} \log h(x, \theta) & = (1 - \sigma(w^{\top} x + b) ) 
\begin{pmatrix}
x \\ 1
\end{pmatrix}^{\top} , \\
\nabla_{\theta} \log (1-h(x, \theta)) & = -\sigma(w^{\top} x + b)
\begin{pmatrix}
x \\ 1
\end{pmatrix}^{\top} .
\end{align*}
Their Hessian matrices are, respectively,
\begin{align*}
\nabla_{\theta}^{2} \log h(x, \theta) 
& = - \sigma(w^{\top} x + b) ( 1- \sigma(w^{\top} x + b)) 
\begin{pmatrix}
x \\ 1
\end{pmatrix} 
\begin{pmatrix}
x \\ 1
\end{pmatrix}^{\top}, \\
\nabla_{\theta}^{2} \log (1-h(x, \theta)) 
& = - \sigma(w^{\top} x + b) (1- \sigma(w^{\top} x + b)) 
\begin{pmatrix}
x \\ 1
\end{pmatrix} 
\begin{pmatrix}
x \\ 1
\end{pmatrix}^{\top},
\end{align*}
which are the same negative semi-definite matrix for any $x$ and $\theta$.

Given these results, we obtain from \eqref{eq:nll-logistic} that
\begin{align*}
\nabla_{\theta} \ell(\theta) 
& = - \sum_{j=1}^{M} \Big( y^{(j)} (1 - \sigma(w^{\top} x^{(j)}+b)) - (1-y^{(j)}) \sigma(w^{\top} x^{(j)}+b) \Big) \begin{pmatrix}
x^{(j)} \\ 1
\end{pmatrix}^{\top}
\end{align*}
which can be simplified in practical implementations since the label $y^{(j)}$ is either 1 or 0. 
Furthermore, we have 
\begin{align*}
\nabla_{\theta}^{2} \ell(\theta) = \sum_{j=1}^{M} \sigma(w^{\top} x^{(j)} + b) (1- \sigma(w^{\top} x^{(j)} + b)) 
\begin{pmatrix}
x^{(j)} \\ 1
\end{pmatrix} 
\begin{pmatrix}
x^{(j)} \\ 1
\end{pmatrix}^{\top}\succeq 0
\end{align*}
which implies that $\ell$ is convex in $\theta$.

Therefore, we know any critical point $\theta$ of $\ell$ satisfying
\begin{equation}
\label{eq:nll-logistic-cp}
\nabla_{\theta} \ell(\theta) = (\partial_w \ell(\theta) , \partial_b \ell(\theta) ) = 0 
\end{equation}
is a global minimizer of $\ell$, and hence also the maximum likelihood estimate given the dataset $\Dcal$.
In addition, if 
\begin{equation*}
X = 
\begin{pmatrix}
x^{(1)\top} & 1\\
\vdots & \vdots \\
x^{(M)\top} & 1
\end{pmatrix}
\in \mathbb{R}^{M \times (d+1)}
\end{equation*}
has full column rank $d+1$, then we can multiply any nonzero $z \in \mathbb{R}^{d+1}$ to the two sides of the Hessian $\nabla_{\theta}^{2} \ell(\theta)$ and find
\begin{equation*}
z^{\top} \nabla_{\theta}^{2} \ell(\theta) z = \sum_{j=1}^{M} \sigma(w^{\top} x^{(j)} + b) (1- \sigma(w^{\top} x^{(j)} + b)) \ |(x^{(j)\top}, \ 1)z|^{2} > 0 ,
\end{equation*}
which implies that $\nabla_{\theta}^{2} \ell(\theta) \succ 0$ and hence \eqref{eq:nll-logistic-cp} only admits one critical point which is the unique minimizer of $\ell$.
\end{example}

%\begin{remark}
%Notice that $\ell$ is convex with respect to $\theta$: it is easy to find by calculus that
%\begin{equation*}
%\nabla^{2}_{\theta} \Big( - \log \sigma(w^{\top} x + b) \Big) = \sigma(w^{\top}x+b) \Big( 1-  \sigma(w^{\top}x +b) \Big) \begin{pmatrix}
%x \\ 1
%\end{pmatrix}
%\begin{pmatrix}
%x^{\top} & 1
%\end{pmatrix} 
%\end{equation*}
%which is a positive semi-definite matrix for any $x \in \mathbb{R}^{d}$. Since $\ell$ is the sum of such terms with non-negative coefficients $y^{(j)}$ and $1-y^{(j)}$ in \eqref{eq:nll-logistic}, we know that $\ell$ is convex with respect to $\theta$. Therefore, the critical point satisfying \eqref{eq:nll-logistic-cp} is the global minimizer of the negative log-likelihood function $\ell$. These concepts are from the field of optimization, which we will cover in detail in Chapter \ref{chpt:opt}.
%\end{remark}

Examples \ref{ex:regression} and \ref{ex:classification} give rise to several questions about function approximation with some given dataset $\Dcal$:
\begin{enumerate}
\item What if we have the dataset $\Dcal$, but not any specified function $f$ to describe the relation between $x$ and $y$ as in \eqref{eq:linear-regression-model} and \eqref{eq:logistic-classification-model}?
\item How to find a model (a parametric function) such that, with proper parameters (e.g., $w$ and $b$), the model can (approximately) represent the unknown function $f$?

\item How to achieve a good balance between accuracy and efficiency of the model in representing $f$? Namely, the model can approximate any one in a class of functions with a pre-specified distance between functions to measure the approximation error, and meanwhile the size (the number of parameters) of this model can be as small as possible?

\item Since we do not know the true $f$ in practice, how to set criteria to evaluate whether the model has the correct parameters to approximate the desired $f$?

\item How do we find the parameters of the model in practice?

\item How large the dataset $\Dcal$ (the number of samples in $\Dcal$) we need to obtain the correct parameters of the model?
\end{enumerate}

We will keep these questions in mind when going through this book and try to find answers to them.

\section{Shallow and Deep Neural Networks}

We consider the following model, which is a hybrid of those in the linear regression and logistic classification examples (Examples \ref{ex:regression} and \ref{ex:classification}):
\begin{equation}
    \label{eq:shallow-net}
    f_{\theta}(x) := \sum_{j=1}^{d_{1}} a_{j} \sigma( w_{j}^{\top} x + b_{j}) ,
\end{equation}
where $w_{j} \in \mathbb{R}^{d}$, $a_{j}, b_{j} \in \mathbb{R}$, $d_{1} \in \mathbb{N}$, and $\theta$ is the vector concatenating all these $w_{j}$'s, $a_{j}$'s and $b_{j}$'s for $j = 1,\dots,d_{1}$.
For convenience, in the remainder of this book, we usually call $\theta$ is ``a parameter'', despite that it is the combination of all the parameters in the model $f_{\theta}$ in \eqref{eq:shallow-net}.

\begin{example}[Shallow neural network]
We illustrate the model $f_{\theta}$ in \eqref{eq:shallow-net} for the case with $d = 3$ and $d_{1} = 4$ in Figure \ref{fig:shallow-net}.
\begin{figure}
\centering
\begin{tikzpicture}[scale=1.2, transform shape]
	\Vertex[x=0, y=1, label=$x_{1}$, color=none, size=0.7]{x1}
	\Vertex[x=0, y=0, label=$x_{2}$, color=none, size=0.7]{x2}
	\Vertex[x=0, y=-1, label=$x_{3}$, color=none, size=0.7]{x3}
	\Vertex[x=2, y=1.5, label=$h_{1}$, color=none, size=0.7]{h1}
	\Vertex[x=2, y=0.5, label=$h_{2}$, color=none, size=0.7]{h2}
	\Vertex[x=2, y=-0.5, label=$h_{3}$, color=none, size=0.7]{h3}
	\Vertex[x=2, y=-1.5, label=$h_{4}$, color=none, size=0.7]{h4}
	\Vertex[x=4, y=0, label=$f_{\theta}$, color=none, size=0.7]{f}
	\Edge[Direct, lw=1pt](x1)(h1)
	\Edge[Direct, lw=1pt](x1)(h2)
	\Edge[Direct, lw=1pt](x1)(h3)
	\Edge[Direct, lw=1pt](x1)(h4)
	\Edge[Direct, lw=1pt](x2)(h1)
	\Edge[Direct, lw=1pt](x2)(h2)
	\Edge[Direct, lw=1pt](x2)(h3)
	\Edge[Direct, lw=1pt](x2)(h4)
	\Edge[Direct, lw=1pt](x3)(h1)
	\Edge[Direct, lw=1pt](x3)(h2)
	\Edge[Direct, lw=1pt](x3)(h3)
	\Edge[Direct, lw=1pt](x3)(h4)
	\Edge[Direct, lw=1pt](h1)(f)
	\Edge[Direct, lw=1pt](h2)(f)
	\Edge[Direct, lw=1pt](h3)(f)
	\Edge[Direct, lw=1pt](h4)(f)
\end{tikzpicture}
\caption{A shallow neural network $f_{\theta}$ with input layer width (dimension) $d=3$, hidden layer width $d_{1} = 4$, and output layer width $1$.}
\label{fig:shallow-net}
\end{figure}
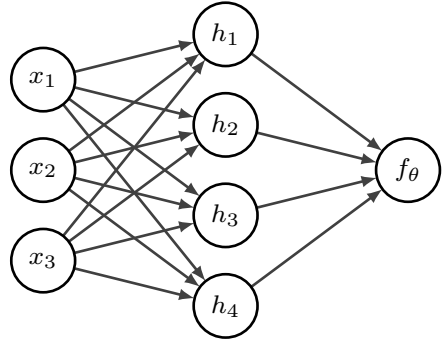

The graph shown in Figure \ref{fig:shallow-net} is a model $f_{\theta}$, which we call a \emph{shallow neural network}, where $x_{1}, x_{2}, x_{3}$ are the neurons in the \emph{input layer}\index{Layer!input}, $h_{1}, h_{2}, h_{3}, h_{4}$ are neurons in the \emph{hidden layer}\index{Layer!hidden}, and $f_{\theta}$ is the one in the \emph{output layer}\index{Layer!output}.
The relations between the values of these neurons can be expressed by
\begin{equation*}
h_{j}(x) = \sigma ( w_{j}^{\top} x + b_{j}) \in \mathbb{R}, \quad \forall\, j \in [d_{1}] ,
\end{equation*}
and
\begin{equation*}
f_{\theta} (x) = \sum_{j=1}^{d_{1}} a_{j} h_{j}(x) .
\end{equation*}
Here $\sigma$ is called the \emph{activation function}\index{Activation function}, which can be the logistic function in \eqref{eq:logistic-fn}, or any function that is not affine since otherwise $f_{\theta}$ reduces to an affine function of $x$. We will present several commonly used activation functions in modern deep learning in Section \ref{subsec:activation-fn}.
\end{example}

The seminal work \cite{cybenko1989approximation} shows that shallow neural networks\index{Neural network!shallow} of form \eqref{eq:shallow-net} are dense in the set of continuous functions defined in a compact subset $\Omega$ of $\mathbb{R}^{d}$. There have been numerous theoretical results on more general cases, e.g., using deeper networks, other activation functions, and different set of target functions to be approximated, developed in the past decades. We provide more related references in Section \ref{subsec:dnn-ref}.
%
% However, the size of this network, i.e., the dimension of $\theta$ (the total number of parameters in $f_{\theta}$) can be very large since a very wide hidden layer (very large $d_{1}$) is often needed.

%
% DNNs have enormously greater representation power and astonishing performance in almost every field of science and technology today. 
%
While shallow networks like \eqref{eq:shallow-net} have provable function approximation capacity, it was found that their compositions, generally called \emph{deep neural networks}\index{Neural network!deep}, are more powerful in function approximation with much fewer parameters than theoretical estimation. 
A simple example of deep neural networks is a direct composition of multiple shallow networks, called \emph{multilayer perceptron}\index{Multilayer perceptron} (MLP), as shown in the next example.
%
% Nowadays, DNNs refer to the general class of networks of various shapes and depths, and MLPs refer to the simplest one we describe next.

\begin{example}
[Multilayer perceptron]
A multilayer perceptron (MLP) is a composition of multiple shallow networks of form \eqref{eq:shallow-net}. In particular, an MLP can have $L$ hidden layers\index{Layer!hidden layer}, where the $l$th hidden layer may have $d_{l}$ neurons $h_{j}^{(l)}$ for $j = 1,\dots, d_{l}$ and $l=1,\dots,L$. The output layer can have one or multiple neurons as needed.

For an MLP with two hidden layers, the neurons in the first hidden layer can be expressed as 
\begin{equation*}
h_{j}^{(1)}(x) = \sigma^{(1)} (w_{j}^{(1)\top}x + b_{j}^{(1)}) , \quad w_{j}^{(1)}\in \mathbb{R}^{d}, \ b_{j}^{(1)} \in \mathbb{R}, \ j=1,\dots, d_{1} .
\end{equation*}
The neurons in the second hidden layer can be expressed as 
\begin{equation*}
h_{j}^{(2)}(x) = \sigma^{(2)} (w_{j}^{(2)\top}h_{j}(x) + b_{j}^{(2)}) , \quad w_{j}^{(2)}\in \mathbb{R}^{d_{1}}, \ b_{j}^{(2)} \in \mathbb{R}, \ j=1,\dots, d_{2} .
\end{equation*}
Here $\sigma^{(1)}$ and $\sigma^{(2)}$ are the activation functions in the first and second hidden layers, respectively.
The output layer neuron is given by
\begin{equation*}
f_{\theta}(x) = \sum_{j=1}^{d_{2}} a_{j} h_{j}^{(2)}(x) + b, \quad a_{j}, \ b\in \mathbb{R}, \ j=1,\dots,d_{2} ,
\end{equation*}
which is an affine function of the second and last hidden layer in this example.

The MLP with input layer of width $d=3$, two hidden layers of widths $d_{1}=4$ and $d_{2}=5$ respectively, and output layer of width 1 is shown in Figure \ref{fig:deep-net}.
\begin{figure}
\centering
\begin{tikzpicture}[scale=1.2, transform shape]
	\Vertex[x=0, y=1, label=$x_{1}$, color=none, size=0.7]{x1}
	\Vertex[x=0, y=0, label=$x_{2}$, color=none, size=0.7]{x2}
	\Vertex[x=0, y=-1, label=$x_{3}$, color=none, size=0.7]{x3}
	
	\Vertex[x=2, y=1.5, label=$h_{1}^{(1)}$, color=none, size=0.7]{h11}
	\Vertex[x=2, y=0.5, label=$h_{2}^{(1)}$, color=none, size=0.7]{h12}
	\Vertex[x=2, y=-0.5, label=$h_{3}^{(1)}$, color=none, size=0.7]{h13}
	\Vertex[x=2, y=-1.5, label=$h_{4}^{(1)}$, color=none, size=0.7]{h14}
	
	\Vertex[x=4, y=2, label=$h_{1}^{(2)}$, color=none, size=0.7]{h21}
	\Vertex[x=4, y=1, label=$h_{2}^{(2)}$, color=none, size=0.7]{h22}
	\Vertex[x=4, y=0, label=$h_{3}^{(2)}$, color=none, size=0.7]{h23}
	\Vertex[x=4, y=-1, label=$h_{4}^{(2)}$, color=none, size=0.7]{h24}
	\Vertex[x=4, y=-2, label=$h_{5}^{(2)}$, color=none, size=0.7]{h25}
	
	\Vertex[x=6, y=0, label=$f_{\theta}$, color=none, size=0.7]{f}
	\Edge[Direct, lw=1pt](x1)(h11)
	\Edge[Direct, lw=1pt](x1)(h12)
	\Edge[Direct, lw=1pt](x1)(h13)
	\Edge[Direct, lw=1pt](x1)(h14)
	\Edge[Direct, lw=1pt](x2)(h11)
	\Edge[Direct, lw=1pt](x2)(h12)
	\Edge[Direct, lw=1pt](x2)(h13)
	\Edge[Direct, lw=1pt](x2)(h14)
	\Edge[Direct, lw=1pt](x3)(h11)
	\Edge[Direct, lw=1pt](x3)(h12)
	\Edge[Direct, lw=1pt](x3)(h13)
	\Edge[Direct, lw=1pt](x3)(h14)
	
	\Edge[Direct, lw=1pt](h11)(h21)
	\Edge[Direct, lw=1pt](h11)(h22)
	\Edge[Direct, lw=1pt](h11)(h23)
	\Edge[Direct, lw=1pt](h11)(h24)
	\Edge[Direct, lw=1pt](h11)(h25)
	\Edge[Direct, lw=1pt](h12)(h21)
	\Edge[Direct, lw=1pt](h12)(h22)
	\Edge[Direct, lw=1pt](h12)(h23)
	\Edge[Direct, lw=1pt](h12)(h24)
	\Edge[Direct, lw=1pt](h12)(h25)
	\Edge[Direct, lw=1pt](h13)(h21)
	\Edge[Direct, lw=1pt](h13)(h22)
	\Edge[Direct, lw=1pt](h13)(h23)
	\Edge[Direct, lw=1pt](h13)(h24)
	\Edge[Direct, lw=1pt](h13)(h25)
	\Edge[Direct, lw=1pt](h14)(h21)
	\Edge[Direct, lw=1pt](h14)(h22)
	\Edge[Direct, lw=1pt](h14)(h23)
	\Edge[Direct, lw=1pt](h14)(h24)
	\Edge[Direct, lw=1pt](h14)(h25)

	\Edge[Direct, lw=1pt](h21)(f)
	\Edge[Direct, lw=1pt](h22)(f)
	\Edge[Direct, lw=1pt](h23)(f)
	\Edge[Direct, lw=1pt](h24)(f)
	\Edge[Direct, lw=1pt](h25)(f)
\end{tikzpicture}
\caption{An example of deep neural network $f_{\theta}$, called multilayer perceptron, with an input layer with width $d=3$, two hidden layers with widths $d_{1}=4$ and $d_{2}=5$, respectively, and an output layer with width $1$.}
\label{fig:deep-net}
\end{figure}
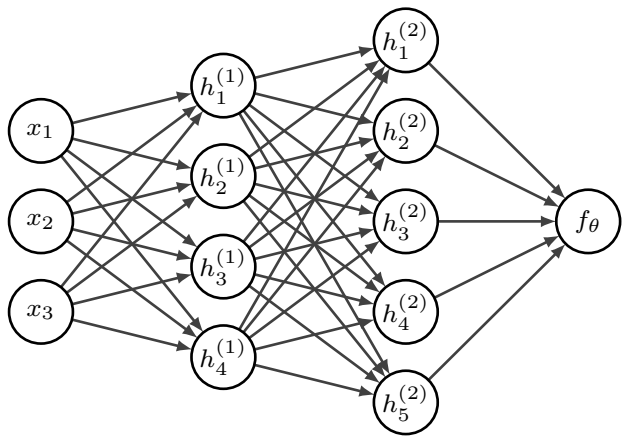
The parameter $\theta$ is a vector that concatenates all the weights and biases in this MLP, and we can count that the dimension of $\theta$ is $(d+1)d_{1} + (d_{1} + 1) d_{2} + (d_{2} + 1)$.
\end{example}

An MLP can have many hidden layers. More hidden layers yield deeper networks. 
Training a model $f_{\theta}$, which is formed as an MLP or any deep neural network of various architectures, refers to finding the parameter $\theta$, such that $f_{\theta}$ can approximate the desired unknown function.
This is in general called \emph{deep learning}\index{Deep learning}.
Deep learning is a new class of machine learning techniques and the most promising approach to general artificial intelligence (AI) because its astonishing empirical successes in an extensive range of real-world applications.

Before ending this section, we make several remarks and raise a question about deep neural networks.
First, we notice that activation functions play an important role in building a deep neural network. In particular, activation functions should not be affine---because otherwise neural networks are just compositions of affine functions, yielding affine functions only which have very limited approximation capacity.
Other than the logistic (sigmoid) function \eqref{eq:logistic-fn} we have seen earlier, another popular, simple, yet effective activation function is 
\begin{equation}
\label{eq:relu-activation-fn}
\sigma(x) = \max(0, x) ,
\end{equation}
which is called the rectified linear unit (ReLU)\index{Activation function!ReLU} function. 
%$
Notice that typical activation functions, such as \eqref{eq:logistic-fn} and \eqref{eq:relu-activation-fn}, are univariate functions, i.e., mappings from $\mathbb{R}$ to $\mathbb{R}$.
When they are applied to vectors, they work on the components of the input vector individually, yielding an output vector of the same dimension as the input vector.
We will show several other commonly used activation functions  in Section \ref{subsec:activation-fn}.

Throughout the remainder of this book, we use deep neural networks as promising parametric models in approximating complex target functions. 
In particular, we call $f_{\theta}$ with unspecified network parameter $\theta$ a \emph{network architecture}\index{Network architecture}. In a network architecture, the neurons (nodes), their connections (directed edges), and the activation functions at the neurons are fixed, whereas the values of the parameters (such as weights and biases) are not assigned.

Meanwhile, we have an important question about neural networks:
\emph{Why is a neural network guaranteed capable of approximating an unknown and complicated function $f$?}

To answer this question with sufficient mathematical rigor, we first need to specify a set $F$ of functions to which the function $f$ belongs (thus we can narrow the search in $F$), as well as a norm\index{Norm} $\|\cdot\|$ as the metric to measure the distance between two functions. 
Then, for any $\epsilon > 0$, we need to construct or at least prove the existence of a network architecture $f_{\theta}$, such that for any $f \in F$, $\| f_{\theta} - f \| \le \epsilon$ for some $\theta$. The next two sections provide more details and an answer to this question.

\section{Universal Approximation Theorem}
\label{sec:uat}

The question raised in the previous section is challenging.
To address this question, we need to have some basic assumptions on the function $f$ and where it is defined.
In the literature, $f$ is generally assumed to be continuous and its domain $\Omega$ is a bounded subset of $\mathbb{R}^{d}$.
Therefore, without loss of generality, we assume that the domain is $\Omega = \dcube$, the closed unit cube in $\mathbb{R}^{d}$, and $f$ belongs to a large class of functions that may have some smoothness property (such as differentiability) in addition to continuity.

In this and next sections, we follow \cite{yarotsky2017error} to state a theorem about the approximation property of deep ReLU networks (neural networks with all activation functions set to ReLU) and provide a proof. The theorem makes a quantitative error estimate which is representative in the literature, and the proof only requires calculus and basic mathematical analysis techniques.
Let $M>0$, $k \in \mathbb{N}$ be arbitrary, and define
\begin{equation}
\label{eq:uat-class-F}
F := \{f \in W^{k,\infty}(\dcube) : \| f \|_{W^{k,\infty}(\dcube)} \le M \} .
\end{equation}
(We will provide more details about this Sobolev space $W^{k,\infty}$ and its norm later in this section.)
Then we want to show that, for any $\epsilon> 0$, there exists a deep ReLU network architecture $f_{\theta}$ of size (we will clarify ``size'' quantitatively below) depending on $d$, $M$, $k$, and $\epsilon$, such that for any $f \in F$, there is
\begin{equation*}
\| f_{\theta} - f \|_{L^{\infty}(\dcube)} \le \epsilon 
\end{equation*}
for some parameter $\theta$ depending on $f$.
This result is formally called the \emph{Universal Approximation Theorem} of neural networks. 
(A formal presentation of this theorem and its proof will be given in the next section.)
%
% Details on the definitions of spaces and norms involved in the claim above will be given later in this section.

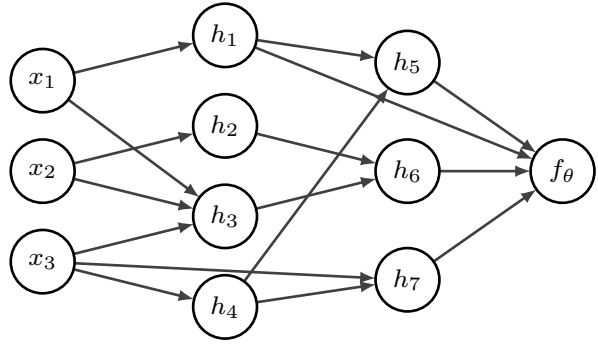
\begin{figure}
\centering
\begin{tikzpicture}[scale=1.2, transform shape]
	\Vertex[x=0, y=1, label=$x_{1}$, color=none, size=0.7]{x1}
	\Vertex[x=0, y=0, label=$x_{2}$, color=none, size=0.7]{x2}
	\Vertex[x=0, y=-1, label=$x_{3}$, color=none, size=0.7]{x3}
	
	\Vertex[x=2, y=1.5, label=$h_{1}$, color=none, size=0.7]{h11}
	\Vertex[x=2, y=0.5, label=$h_{2}$, color=none, size=0.7]{h12}
	\Vertex[x=2, y=-0.5, label=$h_{3}$, color=none, size=0.7]{h13}
	\Vertex[x=2, y=-1.5, label=$h_{4}$, color=none, size=0.7]{h14}
	
	\Vertex[x=4, y=1.2, label=$h_{5}$, color=none, size=0.7]{h21}
	\Vertex[x=4, y=0, label=$h_{6}$, color=none, size=0.7]{h22}
	\Vertex[x=4, y=-1.2, label=$h_{7}$, color=none, size=0.7]{h23}
	
	\Vertex[x=5.7, y=0, label=$f_{\theta}$, color=none, size=0.7]{f}
	\Edge[Direct, lw=1pt](x1)(h11)
	\Edge[Direct, lw=1pt](x1)(h13)
	\Edge[Direct, lw=1pt](x2)(h12)
	\Edge[Direct, lw=1pt](x2)(h13)
	%\Edge[Direct, lw=1pt](x3)(h11)
	\Edge[Direct, lw=1pt](x3)(h13)
	\Edge[Direct, lw=1pt](x3)(h14)
	\Edge[Direct, lw=1pt](x3)(h23)
	
	\Edge[Direct, lw=1pt](h11)(h21)
	\Edge[Direct, lw=1pt](h11)(f)
	\Edge[Direct, lw=1pt](h12)(h22)
	\Edge[Direct, lw=1pt](h13)(h22)
	\Edge[Direct, lw=1pt](h14)(h21)
	\Edge[Direct, lw=1pt](h14)(h23)

	\Edge[Direct, lw=1pt](h21)(f)
	\Edge[Direct, lw=1pt](h22)(f)
	\Edge[Direct, lw=1pt](h23)(f)
\end{tikzpicture}
\caption{A ReLU feed-forward network $f_{\theta}$ with $3$ input neurons $x_{1}$, $x_{2}$, and $x_{3}$. All the other neurons are computation units. The total number of weights and biases of this network is 24, which is the number of computation units (8 in this example, i.e., $h_{1},\dots,h_{7},f_{\theta}$) plus the number of edges (16 in this example).}
\label{fig:feed-forward-net}
\end{figure}

\paragraph{Feed-forward network and its size}
As the approximation theorem to be stated provides a quantitative estimate of network size regarding the error $\epsilon$, we need to clarify what network size means exactly.
To this end, we first remark that the theorem claim holds for \emph{feed-forward neural networks}\index{Feed-forward network}. A feed-forward network can be thought of as a special type of deep networks with skip connections. An example feed-forward network is illustrated in Figure \ref{fig:feed-forward-net}.

In Figure \ref{fig:feed-forward-net}, the network has 3 input neurons $x_{1}$, $x_{2}$, and $x_{3}$. The other 8 neurons are called \emph{computation units}\index{Computation unit} since their values need to be computed based on the values of their parent neurons.
For instance, the value at the neuron $h_{1}$ is given by
\begin{equation*}
h_{1}(x) = \sigma( w^{(x_{1}, h_{1})} x_{1} + w^{(x_{3},h_{1})} x_{3} + b^{h_{1}}) ,
\end{equation*}
where $w^{(i,j)} \in \mathbb{R}$ is the weight of the directed edge $(i,j)$ and $b^{j} \in \mathbb{R}$ is the bias at neuron $j$ in the network.
In this and the next sections, we also call bias as weight for convenience.
Therefore, to compute $h_{1}(x)$, we need 3 weights $w^{(x_{1}, h_{1})}$, $w^{(x_{3},h_{1})}$, and $b^{h_{1}}$. It is easy to see that the total number of weights equals the sum of the number of edges and the number of computation units. For the example network in Figure \ref{fig:feed-forward-net}, the total number of weights is 24.

\paragraph{Lebesgue and Sobolev spaces and their norms}

For a measurable set $\Omega \subset \mathbb{R}^{d}$ and a measurable function $f: \Omega \to \mathbb{R}$, the \emph{essential supremum}\index{Essential supremum} of $f$ in $\Omega$, denoted by $\|f\|_{L^{\infty}(\Omega)}$, is defined by
\begin{equation}
\label{eq:L-infty-norm}
\| f \|_{L^{\infty}(\Omega)} = \inf\Big\{ M \in \Rbb: \ 
\begin{array}{l}
\exists\,\Omega_{0} \subset \Omega\quad \text{with}\quad \mu(\Omega_{0}) = 0, \\
\mbox{s.t.}\ |f(x)| \le M, \ \forall\, x \in \Omega \setminus \Omega_{0} 
\end{array}
\Big\} .
\end{equation}
Namely, $\|f\|_{L^{\infty}(\Omega)}$ is an upper bound of $|f|$ in $\Omega$ except for some measure zero subset $\Omega_{0}$ of $\Omega$, and there is no smaller upper bound that can do the same.
The set of functions with finite $L^{\infty}$ norm\index{Norm} is called the $L^{\infty}$ \emph{Lebesgue space}.
If $\Omega$ is compact and $f$ is continuous in $\Omega$, then $\|f\|_{L^{\infty}(\Omega)} = \max_{x \in \Omega} |f(x)|$, which we will use frequently in Section \ref{sec:uat-proof}, where we use $\|\cdot\|_{\infty} := \| \cdot \|_{L^{\infty}}$ on the involved compact domain for notation simplicity.

Suppose further that $f: \Omega \to \mathbb{R}$ has all its $k$th order (weak) partial derivatives, i.e., $D^{\kbf}f$ exists for every multi-index $\kbf = (k_{1},\dots,k_{d}) \in \{0,1,\dots,k\}^{d}$ satisfying $|\kbf|_{1}:= \sumid k_{i} \le k$, then 
\begin{equation}
\|f\|_{W^{k,\infty}(\Omega)} = \max_{\kbf: |\kbf|_{1} \le k} \|D^{\kbf} f \|_{L^{\infty}(\Omega)} .
\end{equation}
(We will use boldface letters to denote vectors in this and the next sections for clear math representations. For the remainder of the book, we use standard letters for vectors as there should be no danger of confusion.)
The set of functions with finite $W^{k,\infty}$ norm\index{Norm} is called the $W^{k,\infty}$ \emph{Sobolev space}\index{Sobolev space}.
We also know that $f \in W^{k,\infty}(\Omega)$ if any only if $D^{\kbf}f$ exists in $\Omega$ for all $\kbf$ satisfying $|\kbf|_{1} \le k-1$, and $D^{\kbf}f$ are Lipschitz continuous in $\Omega$ for all $\kbf$ satisfying $|\kbf|_{1} = k-1$.
%
% The special case with $k=1$ implies that $f \in W^{1,\infty}(\Omega)$ if and only if $\partial_{i}f$ are Lipschitz continuous in $\Omega$ for all $i \in [d]$.

Since $\|\cdot \|_{L^{\infty}(\Omega)}$ and $\|\cdot \|_{W^{k,\infty}(\Omega)}$ are both absolutely homogeneous, i.e., $\| \alpha f\|_{L^{\infty}(\Omega)} = |\alpha|\, \| \alpha f\|_{L^{\infty}(\Omega)}$ for any $\alpha \in \mathbb{R}$ and $f \in L^{\infty}(\Omega)$ (the same for $\|\cdot \|_{W^{k,\infty}(\Omega)}$), it suffices to show the claim at the beginning of this section for $M=1$ in the definition of $F$ in \eqref{eq:uat-class-F}.
We present the claim as a version of the universal approximation theorem and provide its proof in the next section.

\section{Proof of Universal Approximation Theorem}
\label{sec:uat-proof}

In this section, we prove the universal approximation theorem of deep ReLU feed-forward network\index{Activation function!ReLU} (we call it ReLU net for short in this section).
A formal presentation of this theorem will be given later in this section.
Since the theorem also provides upper bounds on the depth and the numbers of weights and computation units in the network, which depend on the approximation error tolerance $\epsilon>0$, we will be tracking (the orders of) these numbers in the proof.
Since the number of weights is always larger than the number of computation units, we will only track the former.

The proof begins with the following simple proposition about approximating $x^{2}$ on the interval $[0,1]$ by a ReLU net.

\begin{proposition}
\label{prop:approx-xsquare}
For any $\epsilon>0$, the function $f(x) = x^{2}$ defined on $[0,1]$ can be approximated with error $\epsilon> 0$ by a ReLU net having $O(\log(1/\epsilon))$ weights.
\end{proposition}

\begin{proof}
We consider a base function $g: [0,1] \to [0,1]$ defined by
\begin{equation*}
g(x) =
\begin{cases}
2x, & \mbox{if}\ 0 \le x \le \frac{1}{2} , \\
2(1-x), & \mbox{if} \ \frac{1}{2} \le x \le 1,
\end{cases}
\end{equation*}
and the composition of $g$ for $s$ times:
\begin{equation*}
g_{s}(x) := \underbrace{g \circ g \circ \cdots \circ g}_{s\ \text{in total}} \; (x) 
\end{equation*}
for any $s \in \mathbb{N}$.
It is straightforward to verify that
\begin{equation}
\label{eq:gs}
g_{s}(x) =
\begin{cases}
2^{s} \big( \frac{2k}{2^{s}} - x \big), & \text{if } x \in \big[\frac{2k-1}{2^{s}}, \frac{2k}{2^{s}} \big], \ k=1,\dots, 2^{s-1} , \\
2^{s} \big( x - \frac{2k}{2^{s}} \big), & \text{if } x \in \big[ \frac{2k}{2^{s}}, \frac{2k+1}{2^{s}} \big], \ k=0,1,\dots, 2^{s-1}-1 .
\end{cases}
\end{equation}
The plots of $g_{1}(=g)$, $g_{2}$, and $g_{3}$ are shown in Figure \ref{fig:xsquare-pf-g}.
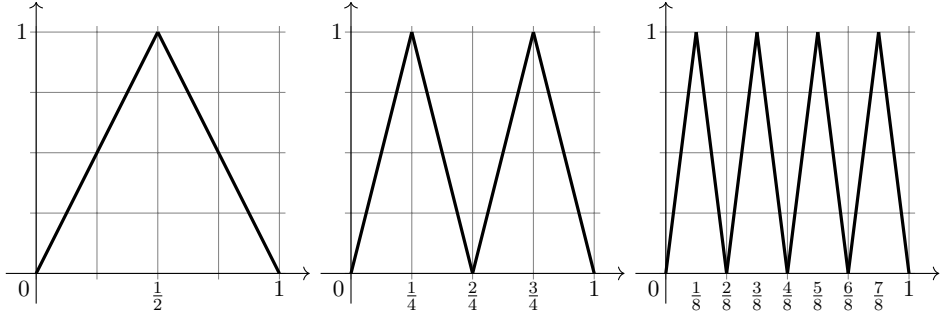
\begin{figure}
\centering
\begin{tikzpicture}[domain=-0.5:4.5, scale=.8, transform shape]
	\draw[very thin, color=gray] (-0.1,-0.1) grid (4.1,4.1);

	\draw[->] (-0.5, 0) -- (4.5, 0); %node[right] {$x$};
	\draw[->] (0, -0.5) -- (0, 4.5); %node[above] {$f(x)$};

	\draw[very thick, color=black] plot[domain=0:2](\x, {2*\x});
	\draw[very thick, color=black] plot[domain=2:4](\x, {2*(4-\x)});
	\draw[black] (-.2, 0) node[below]{$0$};
	\draw[black] (2, 0) node[below]{$\frac{1}{2}$};
	\draw[black] (4, 0) node[below]{$1$};

	\draw[black] (0, 4) node[left]{$1$};
\end{tikzpicture}
\begin{tikzpicture}[domain=-0.5:4.5, scale=.8, transform shape]
	\draw[very thin,color=gray] (-0.1,-0.1) grid (4.1,4.1);

	\draw[->] (-0.5,0) -- (4.5,0); %node[right] {$x$};
	\draw[->] (0,-0.5) -- (0,4.5); %node[above] {$f(x)$};

	\draw[very thick, color=black] plot[domain=0:1](\x, {4*\x});
	\draw[very thick, color=black] plot[domain=1:2](\x, {4*(2-\x)});
	\draw[very thick, color=black] plot[domain=2:3](\x, {4*(\x-2)});
	\draw[very thick, color=black] plot[domain=3:4](\x, {4*(4-\x)});

	\draw[black] (-.2, 0) node[below]{$0$};

	\draw[black] (1, 0) node[below]{$\frac{1}{4}$};
	\draw[black] (2, 0) node[below]{$\frac{2}{4}$};
	\draw[black] (3, 0) node[below]{$\frac{3}{4}$};
	\draw[black] (4, 0) node[below]{$1$};
		
	\draw[black] (0, 4) node[left]{$1$};
\end{tikzpicture}
\begin{tikzpicture}[domain=-0.5:4.5, scale=.8, transform shape]
	\draw[very thin, color=gray] (-0.1,-0.1) grid (4.1,4.1);

	\draw[->] (-0.5,0) -- (4.5,0); %node[right] {$x$};
	\draw[->] (0,-0.5) -- (0,4.5); %node[above] {$f(x)$};

	\draw[very thick, color=black] plot[domain=0:0.5](\x, {8*\x});
	\draw[very thick, color=black] plot[domain=0.5:1](\x, {8*(1-\x)});
	\draw[very thick, color=black] plot[domain=1:1.5](\x, {8*(\x-1)});
	\draw[very thick, color=black] plot[domain=1.5:2](\x, {8*(2-\x)});
	\draw[very thick, color=black] plot[domain=2:2.5](\x, {8*(\x-2)});
	\draw[very thick, color=black] plot[domain=2.5:3](\x, {8*(3-\x)});
	\draw[very thick, color=black] plot[domain=3:3.5](\x, {8*(\x-3)});
	\draw[very thick, color=black] plot[domain=3.5:4](\x, {8*(4-\x)});

	\draw[black] (-.2, 0) node[below]{$0$};
	
	\draw[black] (0.5, 0) node[below]{$\frac{1}{8}$};
	\draw[black] (1, 0) node[below]{$\frac{2}{8}$};
	\draw[black] (1.5, 0) node[below]{$\frac{3}{8}$};
	\draw[black] (2, 0) node[below]{$\frac{4}{8}$};
	\draw[black] (2.5, 0) node[below]{$\frac{5}{8}$};
	\draw[black] (3, 0) node[below]{$\frac{6}{8}$};
	\draw[black] (3.5, 0) node[below]{$\frac{7}{8}$};
	\draw[black] (4, 0) node[below]{$1$};

	\draw[black] (0, 4) node[left]{$1$};
\end{tikzpicture}
\caption{From left to right: $g_{1}$, $g_{2}$, and $g_{3}$ defined in \eqref{eq:gs}.}
\label{fig:xsquare-pf-g}
\end{figure}

Now for any $m = 0, 1, \cdots$, we partition $[0,1]$ into $2^{m}$ equal segments and define a piecewise linear function $f_{m}$ to approximate $f(x) = x^{2}$. In particular, we require $f_{m}$ to match $f$ at all the grid points $\frac{k}{2^{m}}$ for $k = 0,\dots, 2^{m}$:
\begin{equation*}
f_{m} \Big( \frac{k}{2^{m}} \Big) = f \Big( \frac{k}{2^{m}} \Big) = \Big( \frac{k}{2^{m}} \Big)^{2} . 
\end{equation*}
It is easy to verify that $f_{0}(x) = x$ and
\begin{equation*}
f_{m-1} (x) - f_{m} (x) = \frac{g_{m}(x)}{2^{2m}}
\end{equation*}
for all $x \in [0,1]$ and $m \in \mathbb{N}$. 
By telescoping sum in $m$, we have
\begin{equation}
\label{eq:fm-approx-xsquare}
f_{m}(x) = f_{0}(x) - \sum_{s=1}^{m} \frac{g_{s}(x)}{2^{2s}} = x - \sum_{s=1}^{m} \frac{g_{s}(x)}{2^{2s}}
\end{equation}
for all $m = 0, 1, \dots$.

Furthermore, we notice that $g$ can be implemented by a small ReLU net
\begin{equation*}
g(x) = 2 \sigma(x) - 4 \sigma \Big(x - \frac{1}{2} \Big) + 2 \sigma(x-1) .
\end{equation*}
Namely, if we define
\begin{equation*}
h_{1}^{(1)} (x) = \sigma(x), \quad h_{2}^{(1)} (x) = \sigma \Big(x - \frac{1}{2} \Big), \quad h_{3}^{(1)} (x) = \sigma(x-1) ,
\end{equation*}
where $\sigma$ is the ReLU function, then there is
\begin{equation*}
g(x) = 2 h_{1}^{(1)}(x) - 4 h_{2}^{(1)}(x) + 2 h_{3}^{(1)}(x) .
\end{equation*}
Furthermore, we define
\begin{align*}
h_{1}^{(2)} & = \sigma(g(x)) = \sigma \big(2 h_{1}^{(1)}(x) - 4 h_{2}^{(1)}(x) + 2 h_{3}^{(1)}(x) \big) ,\\
h_{2}^{(2)} & = \sigma\Big(g(x) - \frac{1}{2} \Big) = \sigma \Big(2 h_{1}^{(1)}(x) - 4 h_{2}^{(1)}(x) + 2 h_{3}^{(1)}(x) - \frac{1}{2} \Big) , \\
h_{3}^{(2)} & = \sigma(g(x) - 1) = \sigma \big(2 h_{1}^{(1)}(x) - 4 h_{2}^{(1)}(x) + 2 h_{3}^{(1)}(x) - 1\big) .
\end{align*}
Then we have
\begin{equation*}
g_{2}(x) = 2 h_{1}^{(1)}(x) - 4 h_{2}^{(2)}(x) + 2 h_{3}^{(2)}(x) .
\end{equation*}
Following this pattern, we can build a ReLU net to represent any $g_{m}$ and $f_{m}$ for $m \ge 1$.
The plot of the ReLU net $f_{3}$ is shown in Figure \ref{fig:f3}. 
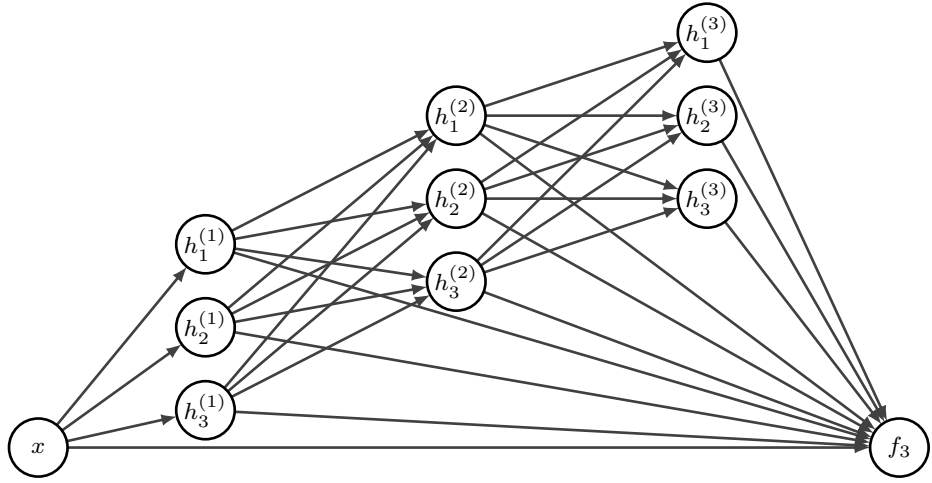
\begin{figure}
\centering
\begin{tikzpicture}[scale=1.1, transform shape]
	\Vertex[x=0, y=0, label=$x$, color=none, size=0.7]{x}
	
	\Vertex[x=2, y=2.45, label=$h_{1}^{(1)}$, color=none, size=0.7]{h11}
	\Vertex[x=2, y=1.45, label=$h_{2}^{(1)}$, color=none, size=0.7]{h12}
	\Vertex[x=2, y=0.45, label=$h_{3}^{(1)}$, color=none, size=0.7]{h13}
	
	\Vertex[x=5, y=4, label=$h_{1}^{(2)}$, color=none, size=0.7]{h21}
	\Vertex[x=5, y=3, label=$h_{2}^{(2)}$, color=none, size=0.7]{h22}
	\Vertex[x=5, y=2, label=$h_{3}^{(2)}$, color=none, size=0.7]{h23}
	
	\Vertex[x=8, y=5, label=$h_{1}^{(3)}$, color=none, size=0.7]{h31}
	\Vertex[x=8, y=4, label=$h_{2}^{(3)}$, color=none, size=0.7]{h32}
	\Vertex[x=8, y=3, label=$h_{3}^{(3)}$, color=none, size=0.7]{h33}
	
	\Vertex[x=10.3, y=0, label=$f_{3}$, color=none, size=0.7]{f3}
	
	\Edge[Direct, lw=1pt](x)(h11)
	\Edge[Direct, lw=1pt](x)(h13)
	\Edge[Direct, lw=1pt](x)(h12)
	\Edge[Direct, lw=1pt](x)(f3)
	
	\Edge[Direct, lw=1pt](h11)(h21)
	\Edge[Direct, lw=1pt](h11)(h22)
	\Edge[Direct, lw=1pt](h11)(h23)
	\Edge[Direct, lw=1pt](h11)(f3)
	\Edge[Direct, lw=1pt](h12)(h21)
	\Edge[Direct, lw=1pt](h12)(h22)
	\Edge[Direct, lw=1pt](h12)(h23)
	\Edge[Direct, lw=1pt](h12)(f3)
	\Edge[Direct, lw=1pt](h13)(h21)
	\Edge[Direct, lw=1pt](h13)(h22)
	\Edge[Direct, lw=1pt](h13)(h23)
	\Edge[Direct, lw=1pt](h13)(f3)
	
	\Edge[Direct, lw=1pt](h21)(h31)
	\Edge[Direct, lw=1pt](h21)(h32)
	\Edge[Direct, lw=1pt](h21)(h33)
	\Edge[Direct, lw=1pt](h21)(f3)
	\Edge[Direct, lw=1pt](h22)(h31)
	\Edge[Direct, lw=1pt](h22)(h32)
	\Edge[Direct, lw=1pt](h22)(h33)
	\Edge[Direct, lw=1pt](h22)(f3)
	\Edge[Direct, lw=1pt](h23)(h31)
	\Edge[Direct, lw=1pt](h23)(h32)
	\Edge[Direct, lw=1pt](h23)(h33)
	\Edge[Direct, lw=1pt](h23)(f3)

	\Edge[Direct, lw=1pt](h31)(f3)
	\Edge[Direct, lw=1pt](h32)(f3)
	\Edge[Direct, lw=1pt](h33)(f3)
\end{tikzpicture}
\caption{The ReLU net $f_{m}$ defined in \eqref{eq:fm-approx-xsquare} for $m=3$.}
\label{fig:f3}
\end{figure}

Following the ReLU net shown in Figure \ref{fig:f3} and enlarging $m$, we can see that the ReLU net $f_{m}$ has $3m+1$ computation units $\{h_{1}^{(s)}, h_{2}^{(s)}, h_{3}^{(s)}, f_{m}: \forall\,s \in [m]\}$ and $12m-5$ edges (each of $x$, $h_{1}^{(s)}$, $h_{2}^{(s)}$ and $h_{3}^{(s)}$ has 4 outgoing edges for $s \in [m-1]$, and each of $h_{1}^{(m)}$, $h_{2}^{(m)}$ and $h_{3}^{(m)}$ has 1 outgoing edge). Hence the total number of weights in $f_{m}$ is $15m - 4 = O(m)$.

It is also straightforward to find the approximation error of $f_{m}$:
\begin{align*}
\| f - f_{m} \|_{\infty}
& = \max_{x \in [0,1]} |f(x) - f_{m}(x) | \\
& = f_{m}(x_{m}^{*}) - f(x_{m}^{*}) \\
& = \frac{1}{2^{2m+2}} ,
\end{align*}
where $x_{m}^{*} = 1 - \frac{1}{2^{m+1}}$.
Therefore, in order to get $\| f - f_{m} \|_{\infty} \le \epsilon$, it suffices to have $\frac{1}{2^{2m+2}} \le \epsilon$. In other words, $f_{m}$ has $O(\log(1/\epsilon))$ depth and weights.
\end{proof}

Note that, for any $a, b \in \mathbb{R}$, there is 
\begin{equation}
\label{eq:product-ab}
a b = \frac{1}{2} \Big( (a+b)^{2} - a^{2} - b^{2} \Big) .
\end{equation}
Since we can approximate the square function $x^{2}$ by a ReLU net as in Proposition \ref{prop:approx-xsquare}, we can approximate the product of two real numbers using a ReLU net based on the identity \eqref{eq:product-ab}.
In the next proposition, we will show that how this approximation is done.

\begin{proposition}
\label{prop:product-net}
For any $M\ge 1$ and $\epsilon \in (0,1)$, there is a ReLU net $\pem: [-M,M] \times [-M,M] \to \mathbb{R}$, such that
\begin{itemize}
\item[(i)] For any $a,b \in [-M, M]$, there is $| \pem (a,b) - ab | \le \epsilon $;
\item[(ii)] If $a=0$ or $b=0$, then $\pem(a, b) = 0$; and
\item[(iii)] The depth and number of weights in $\pem$ are $O(\log(1/\epsilon) + \log M )$.
\end{itemize}
\end{proposition}

\begin{proof}
For $\delta > 0$ to be specified later, let $\qd: [0,1] \to [0,1]$ be the ReLU net approximating the square function $f(x) = x^{2}$ with approximation error $\delta$ in Proposition \ref{prop:approx-xsquare}. 
Hence $\qd(0) = 0$ and
\begin{equation*}
|\qd(x) - x^{2} | \le \delta
\end{equation*}
for all $x \in [0,1]$. Define
\begin{equation*}
\pem(a,b) = \frac{M^{2}}{2} \Big[ \qd \Big( \frac{|a+b|}{2M} \Big) - \qd \Big( \frac{|a|}{2M} \Big) - \qd \Big( \frac{|b|}{2M} \Big) \Big].
\end{equation*}
It is clear that $\pem(a, b) = 0$ if $a=0$ or $b=0$.

By setting $\delta = \frac{8\epsilon}{3M^{2}}$, we have
\begin{align*}
| \pem(a, b) - ab |
& \le \frac{M^{2}}{8} \Big[ \Big| \qd \Big( \frac{|a+b|}{2M} \Big) - \Big( \frac{|a+b|}{2M} \Big)^{2} \Big| + \Big| \qd \Big( \frac{|a|}{2M} \Big) - \Big( \frac{|a|}{2M} \Big)^{2} \Big| \\
& \qquad \qquad + \Big| \qd \Big( \frac{|b|}{2M} \Big) - \Big( \frac{|b|}{2M} \Big)^{2} \Big| \Big] \\
& \le \frac{M^{2}}{8} \cdot (3\delta) \\
& = \epsilon . 
\end{align*}
Since the depth and number of weights in $\qd$ are $O(\log(1/\delta)) = O(\log(1/\epsilon) + \log M)$, we know the same holds true for $\pem$.
\end{proof}

% For ease of understanding, all vectors will be written in boldface in the remainder of this section. Namely, $k$ is a scalar and $\kbf$ is a vector.
%
We denote the unit ball in $W^{k,\infty}(\dcube)$ by
\begin{equation}
\label{eq:W1infty-ball}
B_{k, d} = \{ f \in W^{k,\infty}(\dcube): \|f\|_{W^{k,\infty}(\dcube)}\le 1 \} .
\end{equation}
Consider the following uniform mesh grid on $\dcube$: We partition $[0,1]$ in each of the $d$ axes into $N$ equal segments, then the set of the mesh grid points is
\begin{equation*}
\Big\{ \frac{\mbf}{N} \in \dcube \ : \ \mbf = (m_{1}, \dots, m_{d}) \in N_{d} \Big\} ,
\end{equation*}
where
\begin{equation*}
N_{d} := \{0,1,\dots, N\}^{d} .
\end{equation*}
There are $(N+1)^{d}$ grid points in total.

\begin{figure}
\centering
\begin{tikzpicture}[domain=-0.5:4.5, scale=1, transform shape]
	\draw[ultra thin, color=gray] (-0.1,-0.1) grid (4.1,4.1);

	\draw[->] (-0.5,0) -- (4.5,0); %node[right] {$x$};
	\draw[->] (0,-0.5) -- (0,4.5); %node[above] {$f(x)$};

	\draw[thin, color=black] plot[domain=0:0.3333](\x, {4});
	\draw[thin, color=black] plot[domain=0.3333:0.6667](\x, {(2/3-\x)*12});
		
	\draw[thin, color=black] plot[domain=0.3333:0.6667](\x, {(\x-1/3)*12});
	\draw[thin, color=black] plot[domain=0.6667:1.3333](\x, {4});
	\draw[thin, color=black] plot[domain=1.3333:1.6667](\x, {(5/3-\x)*12});
	
	\draw[very thick, color=black] plot[domain=1.3333:1.6667](\x, {(\x-4/3)*12});
	\draw[very thick, color=black] plot[domain=1.6667:2.3333](\x, {4});
	\draw[very thick, color=black] plot[domain=2.3333:2.6667](\x, {(8/3-\x)*12});

	\draw[thin, color=black] plot[domain=2.3333:2.6667](\x, {(\x-7/3)*12});
	\draw[thin, color=black] plot[domain=2.6667:3.3333](\x, {4});
	\draw[thin, color=black] plot[domain=3.3333:3.6667](\x, {(11/3-\x)*12});

	\draw[thin, color=black] plot[domain=3.3333:3.6667](\x, {(\x-10/3)*12});
	\draw[thin, color=black] plot[domain=3.6667:4](\x, {4});
		
	\draw[black] (-.2, 0) node[below]{$0$};

	\draw[black] (1.3333, 0) node[below]{$\frac{1}{3}$};
	\draw[black] (2, 0) node[below]{$\frac{1}{2}$};
	\draw[black] (2.6667, 0) node[below]{$\frac{2}{3}$};
	\draw[black] (4, 0) node[below]{$1$};

	\draw[black] (0, 4) node[left]{$1$};
\end{tikzpicture}
\caption{From left to right: $\phi_{0}$, $\phi_{1}$, $\phi_{2}$ (thickened), $\phi_{3}$ and $\phi_{4}$ in the case where $d=1$ and $N=4$. Notice that $\sum_{m} \phi_{m}(x) = 1$ for all $x \in [0,1]$.}
\label{fig:uat-pf-phim}
\end{figure}
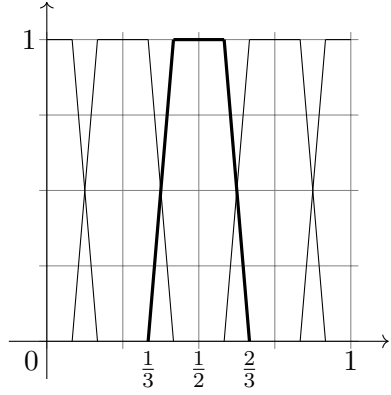

Now we define the base function $\psi: \mathbb{R} \to [0,1]$ as
\begin{equation*}
\psi(x) =
\begin{cases}
1, & \text{if} \quad |x|\le 1 , \\
2 - |x|, & \text{if} \quad 1 \le |x|\le 2 , \\
0, & \text{if} \quad |x|\ge 2 .
\end{cases}
\end{equation*}
Then for any $\mbf = (m_{1}, \dots, m_{d}) \in N_{d}$, we define
\begin{equation}
\label{eq:phim}
\phi_{\mbf}(\xbf) = \prod_{i=1}^{d} \psi \Big( 3N \big( x_{i} - \frac{m_{i}}{N} \big) \Big)
\end{equation}
for any $\xbf = (x_{1}, \dots, x_{d}) \in \dcube$.
Since $\|\psi\|_{\infty} = 1$, we know that there is also $\| \phi_{\mbf} \|_{L^{\infty}(\dcube)} \le 1$.
An example of $\phim$'s in the case of $d=1$ and $N=4$ is given in Figure \ref{fig:uat-pf-phim}.

In the following proposition, we will show that $\{ \phi_{\mbf} : \mbf \in N_{d} \}$ forms a partition of unity on $\dcube$.

\begin{proposition}
For any $N \in \mathbb{N}$, define $\phi_{\mbf}$ as in \eqref{eq:phim}. Then 
\begin{enumerate}
\item For any $\mbf \in N_{d}$ and $\xbf \in \dcube$, there is $0 \le \phi_{\mbf} ( \xbf ) \le 1$. Moreover, $\| \phi_{\mbf} \|_{L^{\infty}(\dcube)} = 1$.
\item The support of $\phi_{\mbf}$ satisfies
\begin{align*}
\supp(\phim)
& = \Big\{ \xbf \in \dcube : : x_{i} - \frac{m_{i}}{N} : \le \frac{2}{3N}, \ \forall\, i \in [d] \Big\} \\
& \subset \Big\{ \xbf \in \dcube : : x_{i} - \frac{m_{i}}{N} : \le \frac{1}{N}, \ \forall\, i \in [d] \Big\}
\end{align*}
and any $\xbf \in \dcube$ can be in the supports of at most $2^{d}$ $\phim$'s.
\item For any $\xbf \in \dcube$, there is 
\begin{equation*}
\sum_{\mbf \in N_{d}} \phimx = 1 .
\end{equation*}
\end{enumerate}
\end{proposition}

\begin{proof}
It is clear that the first claim holds true since $\|\psi \|_{\infty} = 1$ and $\phim(\frac{\mbf}{N}) = 1$.

For any $\xbf \in \dcube$, there exists $\mbf \in N_{d}$ such that for any $i \in [d]$, one and only one of the following can happen:
\begin{subequations}
\begin{align}
& \frac{3m_{i} - 2}{3N} < x_{i} < \frac{3m_{i} - 1}{3N} , \label{subeq:away-pt} \\
& \frac{3m_{i} - 1}{3N} \le x_{i} \le \frac{3m_{i}}{3N} . \label{subeq:near-pt} 
\end{align}
\end{subequations}
These imply that the second claim holds true.

When \eqref{subeq:away-pt} happens, there is
\begin{equation*}
\psii = 2 - \Big| \nximi \Big| = 2 + \nximi .
\end{equation*}
When \eqref{subeq:near-pt} happens, there is
\begin{equation*}
\psii = 1 .
\end{equation*}

Without loss of generality, let $d' \in \{0,1,\dots,d\}$ such that \eqref{subeq:away-pt} happens when $i \le d'$ and \eqref{subeq:near-pt} happens when $d'+1 \le i \le d$.

If $d'=0$, then $\psi(3N(x_{i} - \frac{m_{i}}{N})) = 1$ for all $i \in [d]$. Hence 
\begin{equation*}
\phimx = \prod_{i=1}^{d} \psii = 1 .
\end{equation*}

If $d' = 1$, then 
\begin{align*}
\sum_{\mbf \in N_{d}} \phim 
& = \phimx + \phi_{\mbf - \ebf_{1}}(\xbf) \\
& = \psi \Big( 3N \big(x_{1} - \frac{m_{1}}{N} \big) \Big) + \psi \Big( 3N \big(x_{1} - \frac{m_{1}-1}{N} \big) \Big) \\
& = (2 + 3Nx_{1} - 3m_{1}) + (2 - 3Nx_{1} + 3(m_{1}-1)) \\
& = 1 ,
\end{align*}
where
\begin{equation*}
\ebf_{i} := (0, \dots, 0, \underbrace{1}_{i\text{th}}, 0, \dots, 0) \in \{0,1\}^{d} .
\end{equation*}
In fact, it is easy to check that for any $\mbf \in \{1,\dots,N\}^{d}$ and $i \in [d]$, there is
\begin{equation}
\label{eq:phi-identity}
\psii + \psi \Big( 3N \big(x_{i} - \frac{m_{i}-1}{N} \big) \Big) = 1
\end{equation}
for any $\xbf \in \dcube$.

If $d'=2$, then
\begin{align*}
\sum_{\mbf \in N_{d}} \phim 
& = \phimx + \phi_{\mbf - \ebf_{1}}(\xbf)  + \phi_{\mbf - \ebf_{1} - \ebf_{2}}(\xbf) + \phi_{\mbf - \ebf_{2}}(\xbf)   \\
& = \psi \Big( 3N \big(x_{1} - \frac{m_{1}}{N} \big) \Big) \psi \Big( 3N \big(x_{2} - \frac{m_{2}}{N} \big) \Big) \\ 
& \qquad + \psi \Big( 3N \big(x_{1} - \frac{m_{1}-1}{N} \big) \Big) \psi \Big( 3N \big(x_{2} - \frac{m_{2}}{N} \big) \Big) \\
& \qquad + \psi \Big( 3N \big(x_{1} - \frac{m_{1}-1}{N} \big) \Big) \psi \Big( 3N \big(x_{2} - \frac{m_{2}-1}{N} \big) \Big) \\
& \qquad + \psi \Big( 3N \big(x_{1} - \frac{m_{1}}{N} \big) \Big) \psi \Big( 3N \big(x_{2} - \frac{m_{2}-1}{N} \big) \Big) \\
& = 1
\end{align*}
by using the fact \eqref{eq:phi-identity}.

Following this and \eqref{eq:phi-identity}, we find that for any $d'\in \{0,1,\dots,d\}$, there is 
\begin{equation*}
\sum_{\mbf \in N_{d}} \phimx = 1 .
\end{equation*}
This completes the proof.
\end{proof}

If $f \in W^{k,\infty}(\dcube)$, we know that $D^{\kbf}f$ exists in $\Omega$ and $\| D^{\kbf}f \|_{L^{\infty}(\dcube)} \le \|f \|_{W^{k,\infty}(\dcube)}$ for any $\kbf = (k_{1},\dots,k_{d}) \in \{0,\dots,k\}^{d}$ satisfying $|\kbf|_{1} = \sum_{i=1}^{d} k_{i} < k-1$.
For any $\mbf \in N_{d}$, since $\frac{\mbf}{N} \in \dcube$, we denote the Taylor polynomial of $f$ at $\frac{\mbf}{N}$ with degree $k-1$ as
\begin{equation}
\label{eq:Pmx}
\pmx = \sum_{\kbf: |\kbf|_{1} < k} \frac{1}{\kbf!} D^{\kbf}f \Big(\frac{\mbf}{N} \Big) \Big( \xbf - \frac{\mbf}{N} \Big)^{\kbf} ,
\end{equation}
where $\kbf! = k_{1}! \cdots k_{d}!$, $(\xbf - \frac{\mbf}{N})^{\kbf} = \prod_{i=1}^{d}(x_{i} - \frac{m_{i}}{N} )^{k_{i}}$, and
\begin{equation*}
D^{\kbf}f = \frac{\partial^{k} f}{\partial x_{1}^{k_{1}} \dots \partial x_{d}^{k_{d}}} .
\end{equation*}
Next, we show that $f \in W^{k,\infty}(\dcube)$ can be approximated by the sum of terms $\phimx \pmx$ over $\mbf \in N_{d}$.

\begin{lemma}
\label{lem:f-fN}
For any $f \in B_{k, d}$ and $\epsilon > 0$, there exist $N = N(\epsilon, k, d) \in \mathbb{N}$ and $f_{N}$ of the form
\begin{equation}
\label{eq:fN}
f_{N}( \xbf) = \sum_{\mbf \in N_{d}} \phimx \pmx,
\end{equation}
where $\pmx$ is defined as in \eqref{eq:Pmx}, such that
\begin{equation}
\label{eq:lem-f-fN-error}
\| f_{N} - f \|_{L^{\infty}(\dcube)} \le \frac{\epsilon}{2}.
\end{equation}
\end{lemma}

\begin{proof}
We notice that, for any $ \xbf \in \dcube $, there is
\begin{subequations}
\begin{align}
|f(\xbf) - f_{N}(\xbf)| 
& = \Big| \sum_{\mbf \in N_{d}} \phimx ( f(\xbf) - \pmx ) \Big| \label{subeq:f-fN-1} \\
& \le \sum_{\mbf:|x_{i} - \frac{m_{i}}{N}| < \frac{1}{N}, \forall i \in [d]} | f(\xbf) - \pmx | \label{subeq:f-fN-2} \\
& \le 2^{d} \max_{\mbf:|x_{i} - \frac{m_{i}}{N}| < \frac{1}{N}, \forall i \in [d]} | f(\xbf) - \pmx | \label{subeq:f-fN-3} \\
& \le 2^{d} \frac{1}{\kbf!} \sum_{\kbf:|\kbf|_{1} = k} \|D^{\kbf}f\|_{L^{\infty}(\dcube)} \Big| \Big( \xbf - \frac{\mbf}{N} \Big)^{\kbf} \Big| \label{subeq:f-fN-4} \\
& \le 2^{d} \cdot 1 \cdot d^{k} \cdot 1 \cdot \Big( \frac{1}{N} \Big)^{k} , \label{subeq:f-fN-5} 
\end{align}
\end{subequations}
where \eqref{subeq:f-fN-1} is due to the fact that $\{\phim: \mbf \in N_{d}\}$ is a partition of unity; \eqref{subeq:f-fN-2} due to $\|\phim\|_{L^{\infty}(\dcube)} \le 1$; \eqref{subeq:f-fN-3} due to the fact that there are at most $2^{d}$ terms in the sum in \eqref{subeq:f-fN-2}; \eqref{subeq:f-fN-4} is due to the property of $\pmx$ in \eqref{eq:Pmx} in Taylor polynomial approximation; and \eqref{subeq:f-fN-5} for $\kbf! \ge 1$, the sum in \eqref{subeq:f-fN-4} has no more than $d^{k}$ terms, $\| D^{\kbf}f \|_{L^{\infty}(\dcube)} \le \|f \|_{W^{k,\infty}(\dcube)} \le 1$, and $|x_{i} - \frac{m_{i}}{N}| \le 1$ for all $i \in [d]$. 

By choosing 
\begin{equation}
\label{eq:f-fN-pf-N-bound}
N = \Big\lceil \Big( \frac{1}{2^{d}d^{k}} \cdot \frac{\epsilon}{2} \Big)^{- \frac{1}{k}} \Big\rceil = \big\lceil 2^{-(d+1)/k} d^{-1} \epsilon^{-1/k} \big\rceil,
\end{equation}
which depends on $\epsilon$, $k$, and $d$ but not any specific $f$, we obtain the claimed bound \eqref{eq:lem-f-fN-error}.
\end{proof}

Now we are ready to prove the universal approximation theorem\index{Universal Approximation Theorem} for deep ReLU neural networks.

\begin{theorem}
[Universal approximation theorem]
\label{thm:uat}
For any $d, k \in \mathbb{N}$ and $\epsilon \in (0,1)$, there exists a ReLU net architecture $h$, which depends on $d$, $k$ and $\epsilon$ only, such that
\begin{enumerate}
\item The architecture $h$ is capable to represent any function in $B_{k, d}$ with a properly chosen parameter.
\item The architecture $h$ has depth $O(\log (1/\epsilon))$ and the number of weights at most $c \epsilon^{-d/k} \log(1/\epsilon)$ with $c$ depending on $d$ and $k$ but not $\epsilon$.
\end{enumerate}
\end{theorem}

\begin{proof}
For any $\epsilon \in (0,1)$ and $f \in B_{k, d}$, let $N$ and $f_{N}$ be given as in Lemma \ref{lem:f-fN}.
We rewrite the Taylor polynomial $\pmx$ as 
\begin{equation*}
\pmx = \sum_{\kbf: |\kbf|_{1} < k} \amkf \xmk
\end{equation*}
where the coefficients $\amkf \in \mathbb{R}$ are defined by
\begin{equation*}
\amkf = \frac{1}{\kbf!} D^{\kbf} f \Big( \frac{\mbf}{N} \Big) .
\end{equation*}
Since $f \in B_{k, d}$, we know $|\amkf| \le 1$ for all $\mbf$ and $\kbf$.

Based on \eqref{eq:fN}, we can write $f_{N}$ as
\begin{equation*}
\label{eq:fN-v2}
f_{N}(\xbf) = \sum_{\mbf \in N_{d}} \sum_{\kbf: |\kbf|_{1} <k} \amkf \phimx \xmk 
\end{equation*}
and notice that only $\amkf$ depends on the specified $f$ whereas $\phim$ and $(\xbf - \frac{\mbf}{N})^{\kbf}$ do not.

We can see from \eqref{eq:fN-v2} that $f_{N}$ has at most $(N+1)^{d} d^{k}$ terms in the sum: $N_{d}$ has $(N+1)^{d}$ elements and the size of $\{\kbf: |\kbf|_{1} < k\}$ is upper bounded by $d^{k}$.
We also observe that the function part in each term of the sum in \eqref{eq:fN-v2} has the form of 
\begin{equation}
\label{eq:uat-pf-prod}
\phimx \xmk
\end{equation}
which is the product of at most $d+k-1$ piecewise linear functions:
\begin{equation}
\label{eq:uat-pf-prod-terms}
\underbrace{ \psi\Big( 3N\big(x_{1} - \frac{m_{1}}{N} \big) \Big),\ \dots,\ \psi\Big( 3N\big(x_{d} - \frac{m_{d}}{N} \big) \Big) }_{d\ \text{piecewise linear functions}}, 
\underbrace{\xmk}_{<k\ \text{linear functions}} .
\end{equation}
For notation simplicity, we denote the product of the first $j$ terms in \eqref{eq:uat-pf-prod-terms} by $Q_{\mbf,\kbf}^{j}(\xbf)$, and the product of the last $j$ terms by $Q_{\mbf,\kbf}^{-j}(\xbf)$, for $j=1,\dots,d+k-1$. Note that $|Q_{\mbf,\kbf}^{j}(\xbf)| \le 1$ and $|Q_{\mbf,\kbf}^{-j}(\xbf)| \le 1$ for all $j$.

Next, we approximate \eqref{eq:uat-pf-prod} using a ReLU net. Let $M = d+k$ and $\pdm:[-M,M]\times [-M,M]\to \mathbb{R}$ be the ReLU net from Proposition \ref{prop:product-net} with $\delta>0$ to be determined later.
Define
\begin{equation}
\label{eq:uat-hmn}
\hmkx = \pdm \Big\{ \psi\Big(3N\big(x_{1} - \frac{m_{1}}{N} \big) \Big), \pdm\Big[\psi\Big(3N\big(x_{2} - \frac{m_{2}}{N} \big)\Big), \dots \Big] \Big\} .
\end{equation}
Namely, to compute $\hmkx$, we apply $\pdm$ to the last two terms in \eqref{eq:uat-pf-prod-terms}, then apply $\pdm$ again to the result and the third last term in \eqref{eq:uat-pf-prod-terms}, and so on, till the first term in \eqref{eq:uat-pf-prod-terms}. Therefore, $\pdm$ is applied repeatedly for at most $d+k-2$ times to obtain $\hmkx$.
For notation simplicity, we denote the result obtained by applying $\pdm$ to the last $j$ terms in \eqref{eq:uat-pf-prod-terms} by $P_{\mbf,\kbf}^{-j}(\xbf)$ for $j=2,3,\dots,d+k-1$. For example, we can see that $\hmkx = P_{\mbf,\kbf}^{-(d+|\kbf|_{1}-1)}(\xbf)$.

Note that $\psi(3N(x_{i}-\frac{m_{i}}{N}))$ can be implemented by a ReLU net with one hidden layer and $O(1)$ weights for any $i \in [d]$. The ReLU net has been shown to have $O(\log(1/\delta))$ depth and $O(\log(1/\delta) + \log M)$ weights in Proposition \ref{prop:product-net}. 
Therefore, $h_{\mbf,\kbf}$ can be implemented by a ReLU net with $O(\log(1/\delta))$ depth and $c(d,k)(\log(1/\delta) + \log M)$ weights, where $c(d,k)$ is a constant depending on $d$ and $k$ but not $\epsilon$ or $M$.

Now we estimate the approximation error of $h_{\mbf,\kbf}$. Notice that $|\psi(3N(x_{i}-\frac{m_{i}}{N}))| \le 1$ and $|x_{i}-\frac{m_{i}}{N}| \le 1$ for all $i \in [d]$ and $\xbf \in \dcube$. 
By Proposition \ref{prop:product-net}, we know for any $a,b \in \mathbb{R}$ satisfying $|a| \le 1$ and $|b| \le M$, there is 
\begin{equation*}
| \pdm(a, b) | - |b| \le |\pdm(a,b) - ab| \le \delta.
\end{equation*}
Applying this to \eqref{eq:uat-hmn}, we know the magnitudes of the arguments of all $\pdm$'s are upper bounded by $M$ since $\pdm$ is applied for at most $d+k-2\, (<M)$ times.

Consequently, we have
\begin{align*}
\Big|\hmkx & - \phimx \xmk \Big| \\
& = \Big| P_{\mbf,\kbf}^{-(d+|\kbf|_{1}-1)}(\xbf) - Q_{\mbf,\kbf}^{d+|\kbf|_{1}-1}(\xbf) \Big| \\
& \le \Big| P_{\mbf,\kbf}^{-(d+|\kbf|_{1}-1)}(\xbf) - Q_{\mbf,\kbf}^{1}(\xbf) P_{\mbf,\kbf}^{-(d+|\kbf|_{1}-2)}(\xbf) \Big| \\
& \qquad + \Big| Q_{\mbf,\kbf}^{1}(\xbf) P_{\mbf,\kbf}^{-(d+|\kbf|_{1}-2)}(\xbf) - Q_{\mbf,\kbf}^{2}(\xbf) P_{\mbf,\kbf}^{-(d+|\kbf|_{1}-3)}(\xbf) \Big|\\
& \qquad + \dots \\
& \qquad + \Big| Q_{\mbf,\kbf}^{d+|\kbf|_{1}-3}(\xbf) P_{\mbf,\kbf}^{-2}(\xbf) - Q_{\mbf,\kbf}^{d+|\kbf|_{1}-3}(\xbf) Q_{\mbf,\kbf}^{-2}(\xbf) \Big| \\
& \le (d+k) \delta ,
\end{align*}
where the last inequality is due to that there are no more than $d+k-2$ terms in the sum above it and every term is upper bounded by $\delta$.
In addition, there is $\hmkx = \phimx (\xbf - \frac{\mbf}{N})^{\kbf} = 0$ if $\xbf \notin \supp(\phim):=\mathrm{Closure}(\{\xbf: \phi_{\mbf}(\xbf)\ge 0\})$.

Now we define
\begin{equation}
\label{eq:uat-h}
h(\xbf) = \sum_{\mbf \in N_{d}} \sum_{\kbf: |\kbf|_{1} < k} \amkf \hmkx .
\end{equation}
Note that the coefficients $\amkf$ depend on $f$ whereas the functions $h_{\mbf,\kbf}$ do not.
Then for any $\xbf \in \dcube$, there is
\begin{align*}
| h(\xbf) - f_{N}(\xbf) |
& = \Big| \sum_{\mbf \in N_{d}} \sum_{\kbf: |\kbf|_{1} < k} \amkf \Big( \hmkx - \phimx \xmk \Big) \Big| \\
& = \Big| \sum_{\xbf \in \supp(\phim)} \sum_{\kbf: |\kbf|_{1} < k} \amkf \Big( \hmkx - \phimx \xmk \Big) \Big| \\
& \le 2^{d} \max_{\xbf \in \supp(\phim)} \sum_{\kbf: |\kbf|_{1} < k} \Big|  \hmkx - \phimx \xmk \Big| \\
& \le 2^{d} d^{k} (d+k)\delta ,
\end{align*}
where the first inequality is due to that $\xbf$ can be in the supports of at most $2^{d}$ $\phim$'s and $|\amkf| \le 1$, and the last inequality is because the size of $\{\kbf: |\kbf|_{1} < k\}$ is upper bounded by $d^{k}$.
By choosing
\begin{equation}
\label{eq:uat-delta}
\delta = \frac{\epsilon}{2^{d+1} d^{k}(d+k)} ,
\end{equation}
we have $|h(\xbf) - f_{N}(\xbf)| \le \frac{\epsilon}{2}$ for all $\xbf \in \dcube$. Hence
\begin{equation*}
\| h - f \|_{L^{\infty}(\dcube)} \le \| h - f_{N} \|_{L^{\infty}(\dcube)} + \| f_{N} - f \|_{L^{\infty}(\dcube)} \le \frac{\epsilon}{2} + \frac{\epsilon}{2} = \epsilon .
\end{equation*}

Finally, we notice that $h$ in \eqref{eq:uat-h} can be implemented by a linear combination of at most $(N+1)^{d} d^{k}$ ReLU nets 
\begin{equation*}
\big\{ h_{\mbf,\kbf} : \mbf \in N_{d}, \ |\kbf|_{1} \le k \} .
\end{equation*}
Each of these ReLU nets has $O(\log(1/\delta))$ depth and $O(\log(1/\delta)+\log (d+k))$ weights.
Hence, the depth of $h$ is still $O(\log(1/\delta))$, and the total number of weights in $h$ is 
\begin{equation*}
c(d,k) d^{k} (N+1)^{d} (\log(1/\delta) + \log(d+k)) .
\end{equation*}
Given that $N = \lceil 2^{-(d+1)/k}d^{-1} \epsilon^{-1/k} \rceil$ in \eqref{eq:f-fN-pf-N-bound} and $\delta$ in \eqref{eq:uat-delta}, we obtain the claimed upper bounds on the depth and the number of weights in $h$.
\end{proof}

Theorem \ref{thm:uat} shows an estimate of network size $O(\epsilon^{-d/k} \log(1/\epsilon))$ to achieve an $\epsilon$-approximation to any $f \in F$ in the sense of $L^{\infty}$ norm.
The dominant term involving $\epsilon$ in this bound is $\epsilon^{-d/k}$. The term $\log(1/\epsilon)$ is minor. 
The constant also depends on the dimension $d$ but not $\epsilon$.
This estimate appears to be typical (with minor reduction in some approaches, but the dominant term $\epsilon^{-d/k}$ remains in the estimate) in existing approximation theory of deep neural networks on Sobolev spaces.
At the end of this chapter, we provide more relevant references on this topic.

\section{Network Architecture Design}

\label{sec:network-design}

In this section, we browse through a number of important network design concepts in order to obtain some basic sense on how deep networks are used in practice. We begin with some examples of activation functions commonly used nowadays. Then we discuss a few important standard network structures which are often used as building blocks to assemble large powerful network architectures.

In addition, we introduce some useful ideas and tricks to design neural networks such that they have some prescribed properties. These make their training simple and stable, and more importantly, often lead to significant improvements on performance and efficiency in practice. We will also show some ideas of designing objective functions using physics information rather than externally given data, which led to trending emergent in modern scientific computing.

\subsection{Examples of Activation Functions}
\label{subsec:activation-fn}

We list several examples of nonlinear activation functions that are commonly used in modern deep network architectures. 
Notice that, activation functions are almost always referred to as non-affine (and hence nonlinear) functions, otherwise neural networks are merely compositions of affine functions, yielding affine functions which cannot approximate complicated functions.

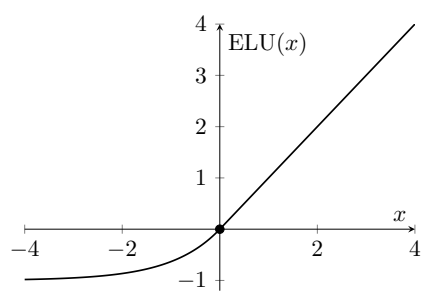
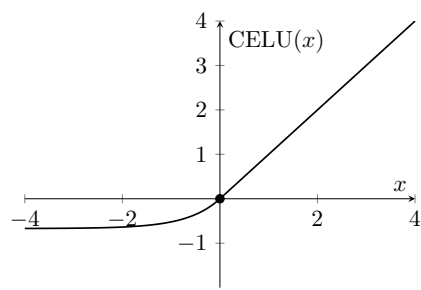
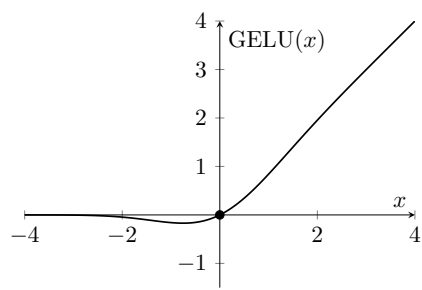
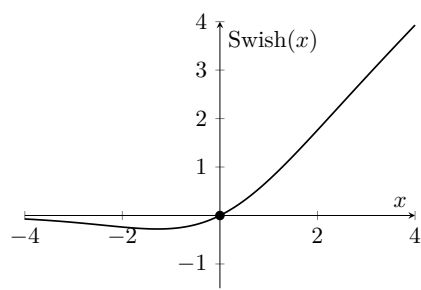
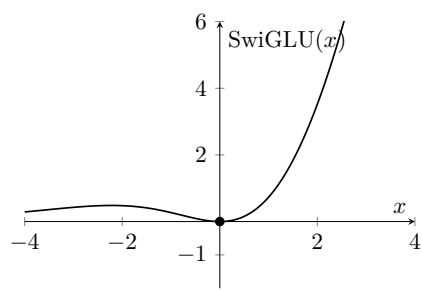
\begin{figure}
\centering
%
% -------
% Sigmoid
%
\begin{subfigure}[b]{.4\textwidth}
\centering
\begin{tikzpicture}[scale=0.8]
\begin{axis}[
    axis lines=middle,
    xlabel={$x$},
    ylabel={sigmoid$(x)$},
    xmin=-6, xmax=6,
    ymin=-0.1, ymax=1.1,
    samples=200,
    domain=-6:6,
    width=8cm,
    height=6cm,
    xtick={-6,-4,-2,0,2,4,6},
    ytick={0,0.5,1},
]
\addplot[thick] {1/(1+exp(-x))};
\addplot[only marks,mark=*] coordinates {(0,0.5)};
\end{axis}
\end{tikzpicture}
\caption{Sigmoid}
\label{fig:sigmoid}
\end{subfigure}
\qquad
%
% -------
% tanh
%
\begin{subfigure}[b]{.4\textwidth}
\centering
\begin{tikzpicture}[scale=0.8]
\begin{axis}[
    axis lines=middle,
    xlabel={$x$},
    ylabel={$\tanh(x)$},
    xmin=-4, xmax=4,
    ymin=-1.1, ymax=1.1,
    samples=200,
    domain=-4:4,
    width=8cm,
    height=6cm,
    xtick={-4,-2,0,2,4},
    ytick={-1,-0.5,0,0.5,1},
]
\addplot[thick] {tanh(x)};
\addplot[only marks,mark=*] coordinates {(0,0)};
%
%% Formula annotation (upper left)
%\node[
%    anchor=north west,
%    fill=white,
%    inner sep=3pt
%] at (axis description cs:0.02,0.98)
%{$\displaystyle
%\tanh(x)=\frac{e^x-e^{-x}}{e^x+e^{-x}}
%$};
\end{axis}
\end{tikzpicture}
\caption{tanh}
\label{fig:tanh}
\end{subfigure}
\\
\vspace{11pt}
%
% --------------------------
% RELU
%
\begin{subfigure}[b]{.4\textwidth}
\centering
\begin{tikzpicture}[scale=0.8]
\begin{axis}[
    axis lines=middle,
    xlabel={$x$},
    ylabel={$\mathrm{ReLU}(x)$},
    xmin=-4, xmax=4,
    ymin=-0.5, ymax=4,
    width=8cm,
    height=6cm,
    xtick={-4,-2,0,2,4},
    ytick={0,1,2,3,4},
]
% ReLU function
\addplot[thick,domain=-4:0] {0};
\addplot[thick,domain=0:4] {x};

% Optional: highlight the kink at zero
\addplot[only marks,mark=*] coordinates {(0,0)};
\end{axis}
\end{tikzpicture}
\caption{ReLU}
\label{fig:relu}
\end{subfigure}
\qquad
%
% --------------------------
% ELU
%
\begin{subfigure}[b]{.4\textwidth}
\centering
\begin{tikzpicture}[scale=0.8]
\begin{axis}[
    axis lines=middle,
    xlabel={$x$},
    ylabel={$\mathrm{ELU}(x)$},
    xmin=-4, xmax=4,
    ymin=-1.2, ymax=4,
    width=8cm,
    height=6cm,
    xtick={-4,-2,0,2,4},
    ytick={-1,0,1,2,3,4},
]
% ELU with alpha = 1
\addplot[thick,domain=-4:0,samples=200]
    {(exp(x)-1)};
\addplot[thick,domain=0:4]
    {x};

% Optional: emphasize continuity at zero
\addplot[only marks,mark=*] coordinates {(0,0)};

% Formula annotation (upper left)
%\node[
%    anchor=north west,
%    fill=white,
%    inner sep=3pt
%] at (axis description cs:0.02,0.98)
%{$\displaystyle
%\mathrm{ELU}(x)=
%\begin{cases}
%x, & x \ge 0,\\
%e^{x}-1, & x < 0
%\end{cases}
%$}
%;
\end{axis}
\end{tikzpicture}
\caption{ELU}
\label{fig:elu}
\end{subfigure}
\\
\vspace{11pt}
%
% --------------------------
% CELU
%
\begin{subfigure}[b]{.4\textwidth}
\centering
\begin{tikzpicture}[scale=0.8]
\begin{axis}[
    axis lines=middle,
    xlabel={$x$},
    ylabel={$\mathrm{CELU}(x)$},
    xmin=-4, xmax=4,
    ymin=-2, ymax=4,
    width=8cm,
    height=6cm,
    xtick={-4,-2,0,2,4},
    ytick={-1,0,1,2,3,4},
    declare function={alpha=1/2;},
]
% CELU (alpha = 1)
\addplot[thick,domain=-4:0,samples=200]
    {0.67*(exp(3*x/2)-1)};
\addplot[thick,domain=0:4]
    {x};

% Emphasize smooth connection at zero
\addplot[only marks,mark=*] coordinates {(0,0)};

%% Formula annotation (upper left)
%\node[
%    anchor=north west,
%    fill=white,
%    inner sep=3pt
%] at (axis description cs:0.02,0.98)
%{$\displaystyle
%\mathrm{CELU}(x)=
%\begin{cases}
%x, & x \ge 0,\\
%\alpha\!\left(e^{x/\alpha}-1\right), & x < 0
%\end{cases}
%$};

\end{axis}
\end{tikzpicture}
\caption{CELU}
\label{fig:celu}
\end{subfigure}
\qquad
%
% --------------------------
% GeLU
%
\begin{subfigure}[b]{.4\textwidth}
\centering
\begin{tikzpicture}[scale=0.8]
\begin{axis}[
    axis lines=middle,
    xlabel={$x$},
    ylabel={$\mathrm{GELU}(x)$},
    xmin=-4, xmax=4,
    ymin=-1.5, ymax=4,
    samples=300,
    domain=-4:4,
    width=8cm,
    height=6cm,
    xtick={-4,-2,0,2,4},
    ytick={-2,-1,0,1,2,3,4},
]
\addplot[thick]
{0.5*x*(1 + tanh(sqrt(2/pi)*(x + 0.044715*x^3)))};

%\node[
%    anchor=north west,
%    fill=white,
%    inner sep=3pt
%] at (axis description cs:0.02,0.98)
%{$\displaystyle
%\mathrm{GELU}(x)\approx
%\frac{x}{2}\left(1+\tanh\!\left(\sqrt{\tfrac{2}{\pi}}
%(x+0.044715x^3)\right)\right)
%$};
\addplot[only marks,mark=*] coordinates {(0,0)};
\end{axis}
\end{tikzpicture}
\caption{GELU}
\label{fig:gelu}
\end{subfigure}
\\
\vspace{11pt}
% --------------------------
% Swish
%
\begin{subfigure}[b]{.4\textwidth}
\centering
\begin{tikzpicture}[scale=0.8]
\begin{axis}[
    axis lines=middle,
    xlabel={$x$},
    ylabel={$\mathrm{Swish}(x)$},
    xmin=-4, xmax=4,
    ymin=-1.5, ymax=4,
    samples=300,
    domain=-4:4,
    width=8cm,
    height=6cm,
    xtick={-4,-2,0,2,4},
    ytick={-1,0,1,2,3,4},
]
\addplot[only marks,mark=*] coordinates {(0,0)};
% Swish activation (beta = 1)
\addplot[thick]
    {x/(1+exp(-x))};
\end{axis}
\end{tikzpicture}
\caption{Swish}
\label{fig:swish}
\end{subfigure}
\qquad
%
% --------------------------
% SwiGLU
%
\begin{subfigure}[b]{.4\textwidth}
\centering
\begin{tikzpicture}[scale=0.8]
\begin{axis}[
    axis lines=middle,
    xlabel={$x$},
    ylabel={$\mathrm{SwiGLU}(x)$},
    xmin=-4, xmax=4,
    ymin=-2, ymax=6,
    samples=300,
    domain=-4:4,
    width=8cm,
    height=6cm,
    xtick={-4,-2,0,2,4},
    ytick={-1,0,2,4,6},
]
\addplot[only marks,mark=*] coordinates {(0,0)};
%
% SwiGLU (scalar illustration)
\addplot[thick]
    {x^2/(1+exp(-x))};

\end{axis}
\end{tikzpicture}

\caption{SwiGLU}
\label{fig:swiglu}
\end{subfigure}
\caption{Examples of activation functions.}
\label{fig:activation-fn}
\end{figure}

We limit our coverage of activation functions to univariate functions in this subsection, and will mention some multivariate activation functions later in this book. 
When we say applying a univariate function to a vector, we mean by applying the activation function to each component of the vector individually and the result is a vector of the same dimension. 
All the activation functions we will cover in this subsection are plotted in Figure \ref{fig:activation-fn}. In what follows, we will provide formulations of these activation functions and briefly describe their properties.

\paragraph{Sigmoid}
The standard sigmoid\index{Activation function!sigmoid} function is defined by
\begin{equation}
\label{eq:sigmoid}
\text{sigmoid}(x) = \frac{1}{1+e^{-x}}, \qquad \forall \, x \in \mathbb{R}
\end{equation}
and shown in Figure \ref{fig:sigmoid}. In the literature, the sigmoid function refers to functions that tend to 1 as $x \to \infty$ and $0$ as $x \to -\infty$. These function can also be non-monotone. However, the function defined in \eqref{eq:sigmoid} is considered as the sigmoid function by default in the deep learning community. The sigmoid function can be used as the last operation of a neural network if the outputs are known to be in the open interval $(0,1)$. However, the derivative of \eqref{eq:sigmoid} decreases very rapidly as $x$ gets further away from 0. This is called ``gradient vanishing'' and often makes network training inefficient.

\paragraph{Hypertangent (tanh)}
The tanh function\index{Activation function!tanh} is defined by
\begin{equation}
\label{eq:tanh}
\text{tanh}(x) = \frac{e^{x}-e^{-x}}{e^{x}+e^{-x}}, \qquad \forall \, x \in \mathbb{R}
\end{equation}
and shown in Figure \ref{fig:tanh}. The tanh function \eqref{eq:tanh} can be thought of as a variation of the sigmoid function. The outputs of the tanh function is in $(-1,1)$. The tanh function also has the gradient vanishing issue. It is typically used for gating purposes in recurrent neural networks (we will see examples later in this section).

\paragraph{Rectified Linear Unit (ReLU)}
The ReLU function\index{Activation function!ReLU} is defined by
\begin{equation}
\label{eq:relu}
\text{ReLU}(x) = \max(0,x), \qquad \forall \, x \in \mathbb{R}
\end{equation}
and is shown in Figure \ref{fig:relu}. The ReLU function \eqref{eq:relu} is commonly used in hinge regression to describe relationships with distinct slopes before and after a specific hinge point. It appears in statistical modeling where a predictor's effect changes abruptly. In deep learning, ReLU is a very frequently used activation function. It does not have the gradient vanishing issue when $x\ge 0$. However, it is not differentiable at $x=0$. A deep ReLU neural network is effectively a continuous piecewise affine function, and the Lipschitz constant can be  evaluated based on the network weight parameters.

\paragraph{Exponential Linear Unit (ELU)}
The ELU activation function\index{Activation function!ELU} is defined by 
\begin{equation}
\label{eq:elu}
\text{ELU}(x) = 
\begin{cases}
x & \text{if } x \ge 0 , \\
e^{x}-1 & \text{otherwise} ,
\end{cases}
\end{equation}
and is shown in Figure \ref{fig:elu}. The ELU function \eqref{eq:elu} is a smoothed version of the ReLU function \eqref{eq:relu}. It slightly mitigates the gradient vanishing issue when $x < 0$ and sometimes can accelerate the training process. It is often used in deep networks and appears to be more stable during training at the expense of higher computational cost than the ReLU function.

\paragraph{Continuously Differentiable Exponential Linear Unit (CELU)}
The CELU activation function\index{Activation function!CELU} is defined by 
\begin{equation}
\label{eq:celu}
\text{CELU}(x) = 
\begin{cases}
x & \text{if } x \ge 0 , \\
\alpha(e^{x/\alpha}-1) & \text{otherwise} ,
\end{cases}
\end{equation}
and is shown in Figure \ref{fig:celu} with $\alpha = \frac{2}{3}$. The CELU function \eqref{eq:celu} is a direct extension of the ELU function above with a tunable hyperparameter $\alpha>0$ and can be used as a substitute of the ReLU and ELU functions.

\paragraph{Gaussian Error Linear Unit (GELU)}
The GELU activation function\index{Activation function!GELU} is defined by 
\begin{equation}
\label{eq:gelu}
\text{GELU}(x) = \frac{x}{2} \Big( 1 + \text{erf}\Big(\frac{x}{2} \Big) \Big), \quad \forall \, x \in \mathbb{R} ,
\end{equation}
where erf is the Gauss error function defined by
\begin{equation*}
\text{erf}\Big(\frac{x}{2}\Big) = \frac{2}{\sqrt{\pi}} \int_{0}^{x} e^{-t^{2}} \,dt .
\end{equation*}
The GELU activation function \eqref{eq:gelu} shown in Figure \ref{fig:gelu}. It is smooth and monotone, and is commonly used in modern transformer architectures for natural language processing and large language models. It often outperforms the ReLU activation function and becomes more popular in implementations of deep networks for many real-world applications.

\paragraph{Swish}
The swish function\index{Activation function!swish} with hyperparameter $\beta>0$ is defined by
\begin{equation}
\label{eq:swish}
\text{swish}_{\beta}(x) = x \cdot \text{sigmoid}(\beta x) = \frac{x}{1 + e^{-\beta x}}, \quad \forall \, x \in \mathbb{R} ,
\end{equation}
and reduces to the sigmoid linear unit (SiLU) function when $\beta$ is set to 1.
The SiLU function with $\beta = 1$ in \eqref{eq:swish} is shown in Figure \ref{fig:swish}.
The swish function is a smooth, non-monotone activation function that can take negative values. It often outperforms ReLU, promoting better gradient flow. 
%
% It helps deep models generalize better by avoiding the sharp zero-crossing of ReLU, leading to smoother optimization and higher accuracy in tasks like image classification, attention, and mobile models.

\paragraph{Swish Gated Linear Unit (SwiGLU)}
The SwiGLU function\index{Activation function!SwiGLU} is defined by
\begin{equation}
\label{eq:swiglu}
\text{SwiGLU}(x) = (vx+c) \cdot \text{swish}(wx+b), \quad \forall \, x \in \mathbb{R}
\end{equation}
with parameters $v, w, b, c$ and shown in Figure \ref{fig:swiglu}.
It can be extended to cases where $x$ is a vector, and then $V$ and $W$ are matrices and $b$ and $c$ are vectors, and the product becomes the componentwise product $\odot$ between vectors.
Namely, $x \odot y := (x_{1}y_{1}, \dots, x_{d} y_{d}) \in \mathbb{R}^{d}$ for any $d$-dimensional vectors $x = (x_{1},\dots, x_{d}) \in \mathbb{R}^{d}$ and $y = (y_{1},\dots, y_{d}) \in \mathbb{R}^{d}$.
The SwiGLU function is one of the most popular gate control activation functions in modern large language models. The gating mechanism allows the SwiGLU to dynamically control the flow of information, enhancing nonlinear expressiveness. The Swish function is smooth, and also unbounded above but bounded below. With the same number of parameters, SwiGLU often exhibits stronger modeling capability than ReLU, GELU, and other related variants.

\subsection{Examples of Network Blocks}

We discuss the structures of several fundamental neural network blocks. These blocks are frequently used to build large and powerful deep network architectures.

\subsubsection*{Residual Network (ResNet)}

While multilayer perceptrons (MLPs) demonstrate the clear idea of stacking multiple layers of shallow networks as deep ones, their empirical behaviors are not guaranteed in most experimental tests.
In particular, the training and testing errors often increase as the layer numbers increase in MLPs, which seem to contradict to the intuition that deeper networks are more representative in function approximation.
It is largely believed that this phenomenon is due to the increased difficulty in training deep MLPs.

\begin{figure}
\centering
\begin{tikzpicture}[scale=1.5]
\Vertex[x=0, label=$x_{l}$, color=none, size=.8]{A} 
\Vertex[x=2, shape = rectangle, label=$h_{\theta_{l}}$, color=none, size=.8]{B}
\Vertex[x=4, label=$x_{l+1}$, color=none, size=.8]{C}
\Edge[Direct, lw=1pt](A)(B)
\Edge[Direct, lw=1pt](B)(C)
\Edge[bend=50, Direct, lw=1pt](A)(C)
\draw[black] (3.5, 0.05) node[above]{\tiny $+$};
\end{tikzpicture}
\caption{A basic block in the $l$th layer of a residual network (ResNet): $x_{l+1} = x_{l} + h_{\theta_{l}}(x_{l})$, where $h_{\theta_{l}}: \mathbb{R}^{d} \to \mathbb{R}^{d}$ is an arbitrary (shallow or small multilayer) network with parameters $\theta_{l}$, and $d$ is the dimension of $x_{l}$.}
\label{fig:resnet-block}
\end{figure}
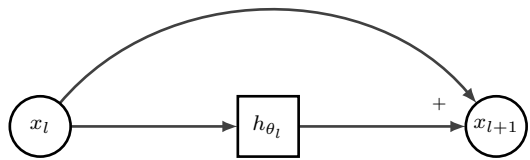

The residual network (ResNet)\index{Neural network!residual} block structures are shown very effective to overcome the aforementioned issues in a very broad range of problems and applications. 
Recall that, in the standard MLPs, the $l$th layer implements $x_{l+1} = h_{\theta_{l}}(x_{l})$, where $x_{l}$ and $x_{l+1}$ are the input and output, respectively, of the current hidden $h_{\theta_{l}}$ layer, and $\theta_{l}$ represents the trainable parameters of $h_{\theta_{l}}$.
By contrast, the $l$th layer of a ResNet, is set to 
\begin{equation}
\label{eq:resnet-block}
x_{l+1} = x_{l} + h_{\theta_{l}}(x_{l}) .
\end{equation}
In other words, there is a skip connection from $x_{l}$ to $x_{l+1}$, and only the residual (hence the name residual network) between $x_{l+1}$ and $x_{l}$ in \eqref{eq:resnet-block} is learned. 
The relation \eqref{eq:resnet-block} can be visualized in Figure \ref{fig:resnet-block}.

Notice that \eqref{eq:resnet-block} requires $h_{\theta_{l}}$ to preserve the dimension of its input, which can be implemented by properly designing the operations in such hidden layers. 
Empirically, ResNet blocks like \eqref{eq:resnet-block} can often stack over thousands of layers, and meanwhile the gradient vanishing issue of such networks appears to be largely mitigated.

As we can see, the ResNet block \eqref{eq:resnet-block} learns modifications to the identity mapping rather than entirely new transformations. If additional layers are unnecessary at a particular depth, the residual block can simply learn the zero function, causing its output to be approximately equal to its input. This property makes ResNets more robust to redundant layers and provides a form of implicit regularization. Moreover, the identity skip connection ensures that information can propagate forward and backward without distortion, provided that the Lipschitz constants of the residual blocks $h_{\theta_{l}}$ are kept sufficiently small such that the entire ResNet forms an invertible mapping. This often enables stable training in very deep ResNet architectures.

It is worth noting that the ResNet blocks can be interpreted from several different perspectives. One notable viewpoint draws connections to dynamical systems since \eqref{eq:resnet-block} resembles one time step of a forward Euler discretization in solving an ordinary differential equation (ODE) with time step size 1. This perspective motivates more advanced network architectures, such as the Neural ODE method with continuous time. We will discuss the Neural ODE method in detail in Section \ref{sec:node}. 
Another viewpoint, as we mentioned earlier, emphasizes that residual pathways in \eqref{eq:resnet-block} result in an identity matrix in the gradient of the right-hand side, which helps to reduce the risk of gradient vanishing issue with properly controlled Lipschitz property of $h_{\theta_{l}}$.

The simple design of ResNet blocks enables the training of very deep neural networks and has become a foundational element in modern deep learning architectures. These blocks are a ubiquitous tool for constructing deep neural networks across an extraordinary range of applications nowadays.

\subsubsection*{Convolutional Neural Network (CNN)}

The standard multilayer perceptrons (MLPs) have another issue of over-parameterization due to their large number of trainable weights. For example, if the $l$th and $(l+1)$th layers of an MLP have $d_{l}$ and $d_{l+1}$ computational units, respectively, then the number of weights is at least $d_{l}d_{l+1}$. This often results in many undesired behaviors in the training process and can be overly data demanding. Moreover, the structure of MLP layers overlooks the correlations between adjacent components of the inputs, reducing the efficiency in the learning processes.

Convolutional neural network (CNN)\index{Neural network!convolutional} blocks are an effective class of network structures specifically designed to process data with spatial connections between input components, such as texts, images, and videos. In particular, CNN blocks are predominant in visual recognition tasks, including image classification, object detection, semantic segmentation, and video analysis. Their success stems from an architectural bias that exploits local spatial correlations, parameter sharing, and hierarchical feature learning.

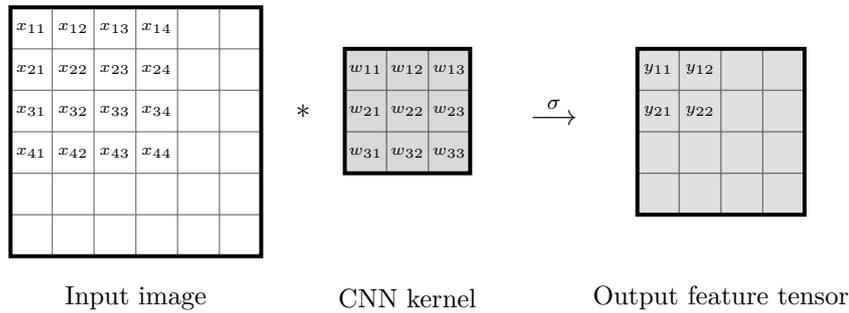
\begin{figure}[t]
\centering
\begin{tikzpicture}[scale=1.1]
\Plane[x=0,y=0,width=3,height=3,grid=5mm,color=none]
\node at (0.25,2.75){\tiny{$x_{11}$}};
\node at (0.75,2.75){\tiny{$x_{12}$}};
\node at (1.25,2.75){\tiny{$x_{13}$}};
\node at (1.75,2.75){\tiny{$x_{14}$}};
\node at (0.25,2.25){\tiny{$x_{21}$}};
\node at (0.75,2.25){\tiny{$x_{22}$}};
\node at (1.25,2.25){\tiny{$x_{23}$}};
\node at (1.75,2.25){\tiny{$x_{24}$}};
\node at (0.25,1.75){\tiny{$x_{31}$}};
\node at (0.75,1.75){\tiny{$x_{32}$}};
\node at (1.25,1.75){\tiny{$x_{33}$}};
\node at (1.75,1.75){\tiny{$x_{34}$}};
\node at (0.25,1.25){\tiny{$x_{41}$}};
\node at (0.75,1.25){\tiny{$x_{42}$}};
\node at (1.25,1.25){\tiny{$x_{43}$}};
\node at (1.75,1.25){\tiny{$x_{44}$}};

\node at (1.5,-.5){\small{Input image}};

\node at (3.5, 1.75){$*$};

\Plane[x=1, y=4, width=1.5, height=1.5, grid=5mm, color=gray]
\node at (4.25,2.25){\tiny{$w_{11}$}};
\node at (4.75,2.25){\tiny{$w_{12}$}};
\node at (5.25,2.25){\tiny{$w_{13}$}};
\node at (4.25,1.75){\tiny{$w_{21}$}};
\node at (4.75,1.75){\tiny{$w_{22}$}};
\node at (5.25,1.75){\tiny{$w_{23}$}};
\node at (4.25,1.25){\tiny{$w_{31}$}};
\node at (4.75,1.25){\tiny{$w_{32}$}};
\node at (5.25,1.25){\tiny{$w_{33}$}};

\node at (4.75,-.5){\small{CNN kernel}};

\node at (6.5, 1.75){$\stackrel{\sigma}{\longrightarrow}$};
\Plane[x=0.5, y=7.5, width=2, height=2, grid=5mm, color=white!60!black]
\node at (7.75,2.25){\tiny{$y_{11}$}};
\node at (8.25,2.25){\tiny{$y_{12}$}};
\node at (7.75,1.75){\tiny{$y_{21}$}};
\node at (8.25,1.75){\tiny{$y_{22}$}};

\node at (8.5,-.5){\small{Output feature tensor}};

\end{tikzpicture}
\caption{Illustration of a basic convolutional neural network (CNN) block applied to an image and the final output. The leftmost image is an input 2-dimensional gray-scale image of resolution $6 \times 6$, the pixel intensities of which are $\{x_{ij} \in \mathbb{R}: i,j \in [6]\}$. The middle is a kernel of size $3\times 3$ with trainable kernel values $\{w_{ij} \in \mathbb{R}: i,j \in [3]\}$. The rightmost picture is the feature tensor of size $4\times 4$ with feature values $\{y_{ij} \in \mathbb{R}: i,j\in [4]\}$, representing the output as the result of the convolution between the image and the kernel. In this example, the $3\times 3$ kernel is first applied to the $3\times 3$ patch centered at $x_{22}$ in the image and results in $y_{11} = \sum_{i=1}^{3} \sum_{j=1}^{3} \sigma(w_{ij} x_{ij})$ as the first feature value in the output, where $\sigma$ is the activation function. Additional bias can be added before applying the activation function $\sigma$. Then the same kernel applies to the $3\times 3$ patch centered at $x_{23} = \sum_{i=1}^{3} \sum_{j=1}^{3} \sigma(w_{ij}x_{i(j+1)})$ in the image and results in $y_{12}$, and so on. This convolution operation is often written simply as $y = \sigma(w * x)$ for brievity. When more than one of such blocks are stacked, the feature tensor output from the previous CNN block is used as the input to the next block. }
\label{fig:cnn}
\end{figure}

In Figure \ref{fig:cnn}, we show the effect of a simple CNN kernel (middle plot) applied to an image (left plot) and the resulting output (right plot). This idea can be easily generalized to construct more complex CNN blocks. 
For example, the input of a CNN block can be a color image with three channels of red, green, and blue, or a multi-channel feature tensor (multi-dimensional matrix). 
The CNN block itself may have $C>1$ channels, for instance, and the kernel can be represented as $\{w_{ij}^{c} \in \mathbb{R}: i, j \in [3], \, c \in [C]\}$. It can also have larger kernel size, such as $5$, $7$, or greater.
The size of the output feature tensor will be determined by the input and the kernel sizes and how the convolution is applied.

There are also many other tricks in using CNN blocks. For example, if we want to apply kernels to the boundary values of the input, we can pad artificial values, such as 0, to the outside of the input, and start applying the kernel to the patch centered at $x_{11}$ shown in Figure \ref{fig:cnn}.
If we want to compress the size of the feature tensor, we can use a larger stride when applying the kernel to the patches in the input image (the stride shown in Figure \ref{fig:cnn} is set to 1 so the feature tensor does not decrease size except for the boundary). 
There are also other commonly used operations other than adding bias and applying activation functions as mentioned in Figure \ref{fig:cnn}. These operations include batch normalization (averaging over multiple outputs) and max pooling. Numerous illustrative examples can be found on the Internet easily and are omitted here.
Most of these operations are built into the modern deep learning software packages and can be directly used in computer programming.

The key of CNN blocks is that the small local kernel can be applied to the entire input of large size. Through the convolution operation, the kernel slides across the input and computes inner products with local patches, producing a feature tensor that responds strongly to specific patterns such as edges, corners, or textures. This local connectivity drastically reduces the number of parameters compared to fully connected layers in MLPs, improving statistical efficiency and generalization, especially in high-dimensional inputs like images.

Because the kernels of CNN blocks are reused at every spatial location, they enforce a translation invariance property as a shift in the input leads to a corresponding shift in the feature tensor. This property is well aligned with natural images, where the identity of an object is largely independent of its location. As a result, CNNs can learn robust features that generalize across spatial positions while keeping the model size manageable.

By stacking multiple convolutional blocks, CNNs naturally form a hierarchical representation of data. Early layers tend to capture low-level features such as edges and simple textures, while deeper layers encode more abstract concepts like object parts and semantic structures. This hierarchical organization mirrors, to some extent, the visual processing pathway in the mammalian cortex and is a key reason for the strong empirical performance of CNNs on perceptual tasks.

From a mathematical perspective, CNNs can be viewed as structured function approximators with strong inductive biases. The convolution operation imposes locality and stationarity assumptions, while depth enables compositional representations. Numerically, CNNs can be trained very efficiently due to the convolution structures on modern hardware, particularly graphical processing units (GPUs).

Despite the rise of transformer-based models, CNNs remain highly relevant due to their efficiency, strong inductive biases, and excellent performance in scenarios with limited data or strict computational constraints. As a result, CNNs continue to play a central role in both theoretical research and real-world applications across computer vision, signal processing, and scientific computing.

\subsubsection*{Graph Neural Network (GNN)}

In a substantial amount of modern real-world applications, data are obtained from irregular domains represented by graphs, such as social networks, citation networks, molecular graphs, and physical interaction systems.
Unlike images or sequences, graphs lack a fixed grid structure, and their nodes may have varying numbers of neighbors. 
Graph neural networks (GNNs)\index{Neural network!graph} address this challenge by learning features from the nodes, edges and the overall topology of the graphs.
%
% Examples of GNNs include graph convolutional networks (GCNs), Graph SAmple and aggreGatE (GraphSAGE), Graph Attention Networks (GATs) and etc.

Let a (directed) graph be denoted by $G = (V, E)$, where $V = \{1, \dots, n\}$ is the set of nodes and $E =\{e_{ij}=(i,j) \in V \times V\}$ is the set of (directed) edges. The graph structure is represented by a weighted adjacency matrix $A = [A_{ij}]_{i,j=1}^{n} \in \mathbb{R}^{n \times n}$, where $A_{ij} > 0$ indicates an edge from the node $i$ to node $j$, i.e., $e_{ij} \in E$. In the unweighted case, we simply have $A_{ij} = 1$ if $e_{ij} \in E$ and $0$ otherwise. Each node $i$ is associated with a row feature vector $x_i \in \mathbb{R}^{1\times d}$, and stacking all node features yields a feature matrix $X \in \mathbb{R}^{n \times d}$. The goal of a GNN is to learn node-level, edge-level, or graph-level representations that are informed by both $X$ and the topology of $G$.

Most modern GNNs can be unified under the message passing neural network framework as follows. 
At the $l$th layer, each node $i$ maintains a $d_{l}$-dimensional hidden representation $h_{i}^{l} \in \mathbb{R}^{d_{l}}$. The update consists of two steps: message aggregation and hidden state update.
The message aggregation step in the $l$th layer is given by
\begin{equation*}
m_{i}^{l} = \sum_{j:e_{ji} \in E} \phi_{l}(h_{i}^{l}, h_{j}^{l}) ,
\end{equation*}
where the sum is taken over all neighbor nodes of $i$, and $\phi_{l}(\cdot,\cdot)$ is a network module that can learn transferring message (information) from the hidden representations $h_{i}^{l}$ and $h_{j}^{l}$ at the two connected nodes $i$ and $j$.
The hidden representation update step in the $l$th layer is given by
\begin{equation*}
h_{i}^{l+1} = \psi_{l}(h_{i}^{l}, m_{i}^{l}) ,
\end{equation*}
where $\psi_{l}$ can also be implemented as a neural network, such as an MLP or a gated recurrent unit (GRU).
This iterative process allows information to propagate across the graph, so that after $L$ layers, $h_{i}^{L}$ encodes information from nodes up to $L$ hops away from the node $i$.

Graph convolutional networks (GCNs) are a widely used instance of GNNs. With a properly weighted adjacency matrix $A \in \mathbb{R}^{n\times n}$ of the graph, define $\tilde{A} := A + I_{n}$ and the corresponding diagonal weighted degree matrix $\tilde{D} := \text{diag}([\tilde{D}_{11},\dots, \tilde{D}_{nn}])$ with $\tilde{D}_{ii} = \sum_{j=1}^{n} \tilde{A}_{ij}$, and $I_{n}$ is the $n$-by-$n$ identity matrix. Then the $l$th GCN layer can be set to:
\begin{equation*}
H_{l+1} := \sigma \Big(\tilde{D}^{-\frac{1}{2}} \tilde{A} \tilde{D}^{-\frac{1}{2}} H_{l} W_{l}  \Big),
\end{equation*}
where $H_{0} := X\in \mathbb{R}^{n \times d}$ is the input node feature matrix given above, and $H_{l} := [h_{1}^{l}, \dots, h_{n}^{l}]^{\top} \in \mathbb{R}^{n \times d_l}$ is the node feature matrix in the $l$th hidden layer. The symmetric normalization $\tilde{D}^{-1/2} \tilde{A} \tilde{D}^{-1/2}$ ensures numerical stability and balances contributions from nodes with different degrees. Here $W_{l} \in \mathbb{R}^{d_{l} \times d_{l+1}}$ is a trainable weight matrix, and $\sigma$ is a nonlinear activation function such as ReLU. This formulation can be interpreted as a form of Laplacian smoothing over the graph.

GNNs have been successfully applied in social network analysis, molecular property prediction, recommendation systems, and knowledge graphs. Despite their effectiveness, challenges remain, including oversmoothing, scalability to large graphs, and theoretical limits on expressive power. Ongoing research addresses these issues through improved architectures, sampling strategies, and connections to graph theory and spectral methods.

Despite their effectiveness, GNNs face challenges such as over-smoothing, where node representations become indistinguishable as depth increases, and scalability issues on very large graphs. These limitations have motivated numerous extensions, including attention-based aggregation, residual connections, sampling-based methods, and normalization techniques. Nevertheless, graph convolutional networks provide a principled and powerful framework for learning from relational data and form the foundation of many modern graph representation learning methods.

\subsubsection*{Recurrent Neural Network (RNN)}

In real-world applications, the data generated and observed often exhibit strong sequential and temporal relations, such as in texts, speech, music, video, and etc. 
Recurrent neural network (RNN)\index{Neural network!recurrent} is a class of neural architectures designed to model such data by maintaining an internal state that evolves over time. Unlike standard feed-forward neural networks, RNNs explicitly incorporate dependencies across time steps, making them suitable for tasks such as time series prediction, natural language processing, and dynamical system modeling.

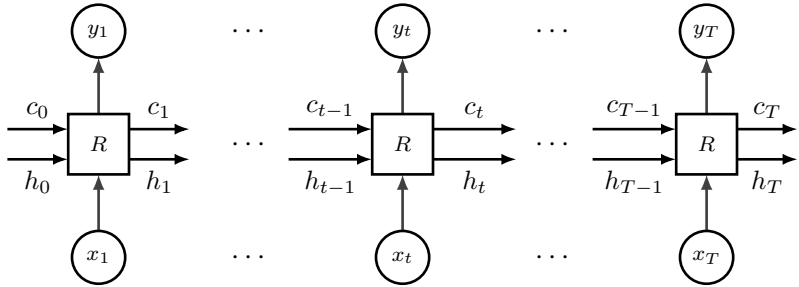
\begin{figure}[t]
\centering
\begin{tikzpicture}

\Vertex[x=0, y=0, label=$x_{1}$, color=none, size=0.7]{x1} 
\Vertex[x=0, y=1.5, shape = rectangle, label=$R$, color=none, size=0.8]{r0}
\Vertex[x=0, y=3, label=$y_{1}$, color=none, size=0.7]{y1} 
\Edge[Direct, lw=1pt](x1)(r0)
\Edge[Direct, lw=1pt](r0)(y1)

\draw[-{Latex[length=2mm]}, line width=1pt] (-1.2, 1.7) -- (-0.4,1.7) node[above, midway] {\small{$c_{0}$}};
\draw[-{Latex[length=2mm]}, line width=1pt] (-1.2, 1.3) -- (-0.4,1.3) node[below, midway] {\small{$h_{0}$}};
\draw[-{Latex[length=2mm]}, line width=1pt] (0.4, 1.7) -- (1.2,1.7) node[above, midway] {\small{$c_{1}$}};
\draw[-{Latex[length=2mm]}, line width=1pt] (0.4, 1.3) -- (1.2,1.3) node[below, midway] {\small{$h_{1}$}};
\node at (2,0) {$\cdots$};
\node at (2,1.5) {$\cdots$};
\node at (2,3) {$\cdots$};
\draw[-{Latex[length=2mm]}, line width=1pt] (2.5, 1.7) -- (3.6,1.7) node[above, midway] {\small{$c_{t-1}$}};
\draw[-{Latex[length=2mm]}, line width=1pt] (2.5, 1.3) -- (3.6,1.3) node[below, midway] {\small{$h_{t-1}$}};
\draw[-{Latex[length=2mm]}, line width=1pt] (4.4, 1.7) -- (5.5,1.7) node[above, midway] {\small{$c_{t}$}};
\draw[-{Latex[length=2mm]}, line width=1pt] (4.4, 1.3) -- (5.5,1.3) node[below, midway] {\small{$h_{t}$}};
\node at (6,0) {$\cdots$};
\node at (6,1.5) {$\cdots$};
\node at (6,3) {$\cdots$};
\draw[-{Latex[length=2mm]}, line width=1pt] (6.5, 1.7) -- (7.6,1.7) node[above, midway] {\small{$c_{T-1}$}};
\draw[-{Latex[length=2mm]}, line width=1pt] (6.5, 1.3) -- (7.6,1.3) node[below, midway] {\small{$h_{T-1}$}};

\draw[-{Latex[length=2mm]}, line width=1pt] (8.4, 1.7) -- (9.2,1.7) node[above, midway] {\small{$c_{T}$}};
\draw[-{Latex[length=2mm]}, line width=1pt] (8.4, 1.3) -- (9.2,1.3) node[below, midway] {\small{$h_{T}$}};

\Vertex[x=4, y=0, label=$x_{t}$, color=none, size=0.7]{xt} 
\Vertex[x=4, y=1.5, shape = rectangle, label=$R$, color=none, size=0.8]{rt}
\Vertex[x=4, y=3, label=$y_{t}$, color=none, size=0.7]{yt} 
\Edge[Direct, lw=1pt](xt)(rt)
\Edge[Direct, lw=1pt](rt)(yt)

\Vertex[x=8, y=0, label=$x_{T}$, color=none, size=0.7]{xT} 
\Vertex[x=8, y=1.5, shape = rectangle, label=$R$, color=none, size=0.8]{rT}
\Vertex[x=8, y=3, label=$y_{T}$, color=none, size=0.7]{yT} 
\Edge[Direct, lw=1pt](xT)(rT)
\Edge[Direct, lw=1pt](rT)(yT)

\end{tikzpicture}
\caption{Architecture of a standard recurrent neural network (RNN). The core component of the RNN is the recurrent block $R$, which is a function parameterized as a network module. At time $t$, $R$ takes the current input data $x_{t}$ (from a time sequential data $(x_{1},\dots,x_{T})$ of length $T$, e.g., a sentence of $T$ words, punctuations, and symbols) as well as the cell-state $c_{t-1}$ and history $h_{t-1}$ generated by $R$ at the previous time $t-1$, and computes the next cell-state $c_{t}$ and history $h_{t}$. It may also generate an output $y_{t}$ that can be compared with external observation values given for training its own parameters. All these inputs and outputs are represented as vectors. The RNN block $R$ shown here is repeatedly used and thus the structure and parameter values are the same for every time $t$. The initial $c_{0}$ and $h_{0}$ may be placeholders with trivial values during implementation but not needed in training. An example recurrent block $R$, known as the long short-term memory (LSTM)\index{Neural network!LSTM}, with input cell-state but not any output is given in \eqref{eq:lstm}.}
\label{fig:rnn}
\end{figure}

Let $\{x_t\}_{t=1}^T$ be an input sequence, such as a sentence consisting of $T$ words, punctuations, and symbols in order, with $x_t \in \mathbb{R}^{d}$ as a numerical representation of the $t$th term in the sequence. (Words, punctuations, and symbols are often represented as vectors and embedded into a Euclidean space in language models.) 
A basic RNN often maintains a hidden state $h_t \in \mathbb{R}^{m}$, called \emph{history}, that summarizes past information. The state update is performed using a recurrent block $R$, which is a function of the current state $x_{t}$ and the history $h_{t-1}$. This $R$ outputs an updated history $h_{t}$ which combined with $x_{t+1}$ serve as the input of $R$ at the next time $t$. The recurrent block $R$ may be designed to have other input and output, such as cell-state $c_{t}$ and observation $y_{t}$. 
An example RNN where the recurrent block $R$ has cell-state as input but no observation output is shown in Figure \ref{fig:rnn}.

From a dynamical systems perspective, an RNN block $R$ defines a nonlinear state-space model that extracts information from the past history and state to generate new history for the next state. The recurrence enables information to persist across time, allowing the network to capture temporal correlations and long-term dependencies in the input sequence.
However, repeated iterations of $R$ may lead to vanishing or exploding gradients in RNN training. 
To mitigate these issues, gated architectures such as Long Short-Term Memory (LSTM) and GRU introduce multiplicative gates that regulate information flow. These mechanisms improve gradient propagation and enable the modeling of long-range dependencies. Despite the emergence of attention-based models, RNNs remain a fundamental framework for understanding sequence modeling and learning in dynamical systems.

% https://www.geeksforgeeks.org/deep-learning/deep-learning-introduction-to-long-short-term-memory/
%
We use LSTM\index{Neural network!LSTM} as an example to illustrate the structure of RNN. In an LSTM, the block $R:=R_{F,I,O}$ is a module consisting of three gates, called the forget gate $F$, the input gate $I$, and the output gate $O$.
Each of these three gates is a small neural network: $F$ is a network with trainable weight matrix $W_{F}$ and bias vector $b_{F}$, $I$ has weight matrices $(W_{I},W_{C})$ and biases $(b_{F}, b_{C})$, and $O$ has weight matrix $W_{O}$ and bias $b_{O}$. 
The dimensions of these weights and biases are properly set so that the computations below can be carried over.

Overall, with the gates defined above, an LSTM block $R$ is commonly set to a function with input and output given by
\begin{equation}
\label{eq:lstm}
(c_{t}, h_{t}) = R_{F,I,O}(x_{t}, c_{t-1}, h_{t-1})
\end{equation}
at every time $t=1,\dots, T$, 
where the computations are implemented as
\begin{align*}
f_{t} & = \text{sigmoid} (W_{F}[h_{t-1}; x_{t}] + b_{F}) , \\
c_{t} & = f_{t} \odot c_{t-1} + \text{sigmoid} (W_{I}[h_{t-1}; x_{t}] + b_{I}) \odot \tanh (W_{C}[h_{t-1}; x_{t}] + b_{C}) , \\
h_{t} & = \text{sigmoid}(W_{O}[h_{t-1}; x_{t}] + b_{O}) \odot \tanh (c_{t}) ,
\end{align*}
and the activation functions sigmoid and tanh are applied componentwisely, $\odot$ represents the componentwise product of two vectors as we explained above, and $[h_{t-1};x_{t}]$ is the column vector concatenating $h_{t-1}$ and $x_{t}$.
If $R$ also has an output $y_{t}$, then it is used to evaluate against some observation data or given ground truth $\hat{y}_{t}$ and appears in the training objective function.

\subsection{Networks with Special Properties}
\label{subsec:special-nets}

As we can see, neural network architecture design allows substantial flexibility.
This is particularly useful in practice, because one can sophisticatedly build desired function properties into the network architectures such that these properties are automatically satisfied.

A major advantage of building desired properties into networks is that this can often effectively eliminate constraints (or at least reduce the number of constraints) in the optimization problem of learning optimal network parameters. This often results in faster and more stable computation during training, as well as improved network performance in practice.
In what follows, we introduce several examples of such design tricks for different specified properties.

\subsubsection*{Value-constrained functions}

If the function $f:\mathbb{R}^{d} \to \mathbb{R}^{n}$ to be approximated is known to have output values in some specified range, then we can explicitly enforce this into the last operation of the neural network used to parameterize the function.

For example, if $f(x) \in (0,1)^{n}$ for all $x$, then we can parameterize $f$ as the composition $\text{sigmoid}\circ \phi$, where $\phi: \mathbb{R}^{d} \to \mathbb{R}^{n}$ is an arbitrary neural network and the sigmoid function \eqref{eq:sigmoid} is applied componentwisely.
Similarly, we can use the tanh function \eqref{eq:tanh} if $f(x) \in (-1,1)^{n}$. 
If the range of $f$ is a box $(a_{1},b_{1}) \times \cdots \times (a_{n},b_{n})$ in $\mathbb{R}^{n}$ for some known $n$-dimensional vectors $a=(a_{1},\dots,a_{n})$ and $b=(b_{1},\dots,b_{n})$ with $a_{i} < b_{i}$ for all $i \in [n]$, then we can parameterize $f$ as 
\begin{equation*}
f(x) = a + (b-a) \odot \mathrm{sigmoid} (\phi(x)) ,
\end{equation*}
or 
\begin{equation*}
f(x) = \frac{1}{2} \Big( (a+b) + (b-a) \odot (\tanh (\phi(x)) \Big) .
\end{equation*}
If $f(x) \ge 0$, we can parameterize $f$ as $\max(0,\phi)$, $\max(0, \phi)^{2}$, or simply $\frac{1}{2}|\phi|^{2}$. Notice that the latter two are continuously differentiable as long as the network $\phi$ is a continuously differentiable function (e.g., $\phi$ only uses smooth activation functions).
These ideas can be generalized to more complicated cases.

If $f(x) \in \mathbb{R}^{n}$ needs to be a probability distribution over $n$ values, namely, $f(x)$ is an $n$-bin histogram, then the range of $f$ is the probability simplex defined by
\begin{equation*}
\Delta^{n}:= \Big\{(y_{1},\dots,y_{n}) \in \mathbb{R}^{n}: 0 \le y_{i} \le 1, \ \sum_{i=1}^{n}y_{i} = 1 \Big\} \subset \mathbb{R}^{n}.
\end{equation*}
(This probability simplex is also denoted by $\Delta^{n-1}$ to indicate that it is an $(n-1)$-dimensional submanifold of $\mathbb{R}^{n}$.)
In this case, we can parameterize $f$ as the composition $\text{softmax} \circ \phi$, where $\phi : \mathbb{R}^{d} \to \mathbb{R}^{n}$ is an arbitrary network and the softmax function is defined by
\begin{equation*}
\text{softmax}(w) : = \frac{1}{\sum_{i=1}^{n} e^{w_{i}}} \Big(e^{w_{1}}, \dots, e^{w_{n}} \Big) \in \mathbb{R}^{n}
\end{equation*}
at every $w \in \mathbb{R}^{n}$. Then $f(x) \in \Delta^{n}$ for all $x \in \mathbb{R}^{d}$.

In some real-world problems, the function $f$ is required to be bounded by some given function $g$ pointwisely. 
For example, $g : \mathbb{R}^{d} \to \mathbb{R}$ is given and $f$ to be found is known to satisfy $f(x) \ge g(x)$ at every $x \in \mathbb{R}^{d}$. In this case, we can parameterize $f$ as
\begin{equation*}
f(x) = g(x) + \frac{1}{2} |\phi(x)|^{2} ,
\end{equation*}
where $\phi: \mathbb{R}^{d} \to \mathbb{R}^{n}$ is an arbitrary neural network.

In some applications, there are additional requirement on the values of $f$ at specified points. For example, in some optimal control problems, we want to find the Lyapunov function $f$ of the control system.
Such functions are known to be positive definite at certain specified points called stationary points.
As a typical example, consider the standard case where the stationary point is set to the origin $0 \in \mathbb{R}^{d}$ and the Lyapunov function $f$ must satisfy $f(x) \ge 0$ at every $x \in \mathbb{R}^{d}$, and $f(x) = 0$ if and only if $x = 0$. 
In this case, we can parameterize $f$ as 
\begin{equation*}
f(x) = g(x) + \frac{1}{2} | \phi(x) - \phi(0)|^{2} ,
\end{equation*}
where $\phi: \mathbb{R}^{d} \to \mathbb{R}^{n}$ is an arbitrary neural network, and $g$ is a user-defined non-negative lower bound of $f$. Examples of $g$ include $g(x) = \frac{\alpha}{2}|x|^{2}$ and $g(x) = \alpha \log(1+|x|^{2})$, where $\alpha>0$ is a used-defined weight value to determine the scale of the lower bounding function $g$.

\subsubsection*{Even, Odd, and Periodic Functions}

If $f: \mathbb{R}^{d} \to \mathbb{R}^{n}$ is known to be even at the $i$th component $x_{i}$ of any input $x = (x_{1}, \dots, x_{d}) \in \mathbb{R}^{d}$, then we can parameterize $f$ as
\begin{equation}
\label{eq:f-odd}
f(x) = \phi \Big(x_{1},\dots,x_{i-1}, \frac{\psi(x_{i}) + \psi(-x_{i})}{2}, x_{i+1}, \dots, x_{d} \Big) ,
\end{equation}
where $\phi: \mathbb{R}^{d} \to \mathbb{R}^{n}$ and $\psi: \mathbb{R} \to \mathbb{R}$ are arbitrary neural networks.
It is easy to check that $f(x_{1},\dots,x_{i},\dots,x_{d}) = f(x_{1},\dots, -x_{i}, \dots, x_{d})$ at every $x \in \mathbb{R}^{d}$.
We can do the same if $f$ is even at more than one components.
If $f$ is odd at $x_{i}$, then we just need to replace $\frac{\psi(x_{i}) + \psi(-x_{i})}{2}$ in \eqref{eq:f-odd} with $\frac{\psi(x_{i}) - \psi(-x_{i})}{2}$.
This idea can be applied to the components of $x$ individually, such that $f$ is even at some components of $x$ and odd at some others.

If $f: \mathbb{R}^{d} \to \mathbb{R}^{n}$ is known to be $T$-periodic at the $i$th component $x_{i}$ of any $x = (x_{1}, \dots, x_{d}) \in \mathbb{R}^{d}$, then we can parameterize $f$ as
\begin{equation}
\label{eq:f-periodic}
f(x) = \phi \Big(x_{1},\dots,x_{i-1}, \cos\Big(\frac{2\pi x_{i}}{T}\Big), \sin\Big(\frac{2\pi x_{i}}{T}\Big), x_{i+1}, \dots, x_{d} \Big) ,
\end{equation}
where $\phi: \mathbb{R}^{d+1} \to \mathbb{R}^{n}$ and $\psi: \mathbb{R} \to \mathbb{R}$ are arbitrary neural networks.
This can be easily extended to the cases where $f$ is periodic at more than one components, and the periods can vary at different components.

\subsubsection*{Functions with Assigned Boundary, Initial and/or Terminal Values}

In certain real-world applications, such as solving partial differential equations (PDEs) \cite{evans1998partial}, the solutions are usually required to satisfy some given boundary, initial, and/or terminal conditions.
In these cases, we may enforce these conditions into the deep neural network used to parameterize the solutions.

For example, if the domain of the solution $u: \mathbb{R}^{d} \to \mathbb{R}$ to a PDE is $\bar{\Omega}$, which is the closure of the open unit box $\Omega := (0,1)^{d} \subset \mathbb{R}^{d}$, and the solution is known to equal to some given function $g$ defined on the boundary $\partial \Omega$, then we can parameterize $u$ as
\begin{equation}
\label{eq:u-bdry-value}
u(x) = g(x) + \phi(x) \prod_{i=1}^{d}x_{i}(1-x_{i})
\end{equation}
for every $x = (x_{1},\dots, x_{d}) \in \Omega$, where $\phi: \mathbb{R}^{d} \to \mathbb{R}$ is an arbitrary neural network.
We can manually set $g(x)$ to any value when $x$ is in the interior of $\Omega$ beforehand. It is easy to verify that \eqref{eq:u-bdry-value} has no constraints inside $\Omega$ and is always equal to $g$ on $\partial \Omega$.

While \eqref{eq:u-bdry-value} illustrates the idea of parameterizing functions with boundary conditions, we remark that one needs to be careful about the choice of $\prod_{i=1}^{d} x_{i} (1-x_{i})$. We use $d=1$ and $g=0$ as an example. If $u: \mathbb{R} \to \mathbb{R}$, and $ u(x) = O(\sqrt{x})$ as $x \to 0$, then we notice that
\begin{equation*}
\frac{u(x)}{x(1-x)} = O \Big( \frac{1}{\sqrt{x}(1-x)} \Big) \to \infty
\end{equation*}
as $x \to 0$. Therefore, it is inappropriate to parameterize $u(x) = \phi(x) \cdot x(1-x)$ as in \eqref{eq:u-bdry-value}, since we would need $\phi(x) \to \infty$ as $x\to 0$ which is infeasible as the neural network $\phi$ is continuous everywhere using typical architectures.
Back to the multivariate case, we should change \eqref{eq:u-bdry-value} to 
\begin{equation}
\label{eq:u-bdry-value-change}
u(x) = g(x) + \phi(x) \cdot \sqrt{x_{1}}(1-x_{1})\prod_{i=2}^{d}x_{i}(1-x_{i}) 
\end{equation}
or alike.
The choice of \eqref{eq:u-bdry-value-change}, as we can see, requires a pre-estimation of the order of $u(x)$ as $x$ tends to the boundary $\partial \Omega$.
In summary, we should modify \eqref{eq:u-bdry-value} such that $\phi$ only needs to approximate a uniformly bounded function---this is what neural networks are capable of as their architectures  typically only allow them to approximate continuous functions.

If the solution $u: \bar{\Omega} \times [0,\infty) \to \mathbb{R}$ of a time-evolution PDE where $\Omega$ is an open subset of $\mathbb{R}^{d}$, and $u$ is known to have initial value $f_{0}$ over $\bar{\Omega}$ at time $t = 0$, then we can parameterize $u$ as 
\begin{equation*}
u(x,t) = f_{0}(x) + t \phi(x,t)
\end{equation*}
where $\phi: \bar{\Omega} \times [0,\infty) \to \mathbb{R}$ is an arbitrary neural network to represent any function of $x$ and $t$.
This can be done similarly if the terminal value of $u$ is given.

If the solution $u: \bar{\Omega} \times [0,T] \to \mathbb{R}$ for some finite $T>0$, and both of its initial value $f_{0}$ and terminal value $f_{T}$ defined on $\bar{\Omega}$ are given, then we can parameterize $u$ as 
\begin{equation*}
u(x,t) = (T-t) f_{0}(x) + t f_{T}(x) + t(T-t)\phi(x,t) ,
\end{equation*}
which meets both initial and terminal conditions automatically. Namely, $u(x,0) = f_{0}(x)$ and $u(x,T) = f_{T}(x)$ for all $x \in \bar{\Omega}$. Meanwhile, $u(x,t)$ has no constraints when $t \in (0,T)$ since $\phi$ is an arbitrary neural network and $t(T-t)>0$. 
We may need to modify the exponent of $t$ above similarly as in \eqref{eq:u-bdry-value-change}, depending on the rate of change of $u(x,t)$ as $t$ tends to $0$ or $T$.

\subsection{Network Training Criteria}

With the problem formulated and network architecture determined, we need to specify a criterion in order to find the optimal network parameters.
This criterion is generally set as a scalar-valued function, called the \emph{objective function}\index{Objective function}, which maps any network parameter $\theta$ (the collection of all parameters in the network) to a real number.
It is also frequently called the \emph{loss function}\index{Loss function}, because in most cases this function is to be minimized.
We call it objective function hereafter since we may alternately look for maximizers or saddle points of this function in certain applications.

The technique of finding the optimal solution to a given objective function (e.g., a minimizer of a given loss function) is called optimization, which is our main topic in Chapter \ref{chpt:opt}.
In deep learning, this optimization process is also called network training, i.e., finding good parameters of the network such that it works well for the target problem.
What we will discuss in this subsection are some rules and ideas to set up appropriate objective functions for the given problems.

In supervised learning problems, the training data contain both variable $x$ and observation or label $y$, as shown in Examples \ref{ex:regression} and \ref{ex:classification}.
In these problems, the objective functions are typically formulated based on the statistical relations between $y$ and the predicted result, often denoted by $f_{\theta}(x)$, where $f_{\theta}$ is the neural network with parameter $\theta$ to estimate the function describing the relation.
For instance, in Example \ref{ex:regression} about regression, $f_{\theta}$ is manually set to an affine function $f_{\theta}(x) = w^{\top} x + b$ with $\theta=(w,b)$, and the corresponding observation $y$ follows the distribution $\Ncal(f_{\theta}(x), s^{2})$, i.e., $y$ is $f_{\theta}(x)$ plus some independent Gaussian noise with unknown variance $s^{2}$. Then we can use the maximum likelihood estimate based on Gaussian distribution to obtain the objective function of $\theta$.
In Example \ref{ex:classification} about classification, $f_{\theta}$ is manually set to $f_{\theta}(x) = \mathrm{sigmoid}(w^{\top} x + b)$ with $\theta = (w,b)$, and the corresponding label $y$ is assumed to follow the probability distribution $\text{Bernoulli}(f_{\theta}(x))$. Then the objective function is formulated by following the maximum likelihood estimate correspondingly.

In unsupervised learning and reinforcement learning problems, we usually do not have label data $y$.
In these cases, we need to set up the training objective functions based on the problem formulations and it is highly case dependent. 
In Chapters \ref{chpt:rl} and \ref{chpt:gm}, we will see many different approaches to set objective functions for the involved methods to work.

At the end of this subsection, we provide an example of objective function design in recent scientific computing research. 
The goal is to solve partial differential equations (PDEs)\index{Differential equation!partial} by parameterizing their solutions as deep neural networks.
The challenge is that the PDEs under consideration are often defined on high dimensional spaces, such as some compact set of $\mathbb{R}^{d}$ with $d \ge 5$.
Despite a substantial amount of theory and methods have been developed to solve PDEs numerically \cite{ames2014numerical,thomas2013numerical,thomas2013fdm}, such as the finite difference, finite element, and spectral methods, it is computationally challenging or infeasible to solve general PDEs using these conventional numerical PDE methods in high-dimensional cases.
The reason is that these methods typically require a refined spatial discretization of the problem domain, e.g., $[0,1]^{d}$, and approximate the solution values at all the discretized voxels.
However, the number of voxels increases at an exponential rate of the dimension $d$, known as the ``curse of dimensionality'' issue, and it is computationally infeasible for the conventional numerical PDE methods to work.

Using deep neural networks to approximate PDE solutions has shown to be a very effective approach in mitigating the dimensionality issue. 
The key idea is to design the objective functions for training these deep neural networks based on the given PDEs, which are often determined by the physics involved in the problems (PDEs are typically derived from laws of physics).
Therefore, neural networks in these methods are commonly referred to \emph{physics-informed neural networks}\index{Neural network!physics informed}.
Notice that, direct spatial discretization is avoided in these methods, and the objective function is often approximated by Monte Carlo integrations (see Appendix \ref{appsec:mc}), which appear to much more scalable for high dimensional problems.
The following examples show how these methods work.

\begin{example}
[Solving high-dimensional PDEs]
\label{ex:pinn-dr-wan}
We consider a classic PDE known as the Poisson equation.
Let $\Omega = (0,1)^{d}$ be the open unit box in $\mathbb{R}^{d}$ and $f: \Omega \to \mathbb{R}$ be given. Then the Poisson equation is written as
\begin{equation}
\label{eq:poisson}
- \Delta u(x) = f(x) , \quad \forall \, x \in \Omega ,
\end{equation}
where $\Delta$ is the Laplace operator, i.e., $\Delta u(x) = \sum_{i=1}^{d} \frac{\partial^{2}u}{\partial x_{i}^{2}} (x)$ at every $x \in \Omega$.
The Poisson equation \eqref{eq:poisson} also requires $u$ to satisfy certain boundary condition or otherwise there may be infinitely many possible solutions.
For example, in the Dirichlet boundary problem of Poisson equation, the boundary function $g: \partial \Omega \to \mathbb{R}$ is given and the solution $u$ is required to satisfy $u(x) = g(x)$ for all $x \in \partial \Omega$.
Now our approach is to parameterize $u: \bar{\Omega} \to \mathbb{R}$ as a deep neural network, denoted by $u_{\theta}$ with network parameter $\theta \in \mathbb{R}^{m}$ ($m$ is the number of parameters in this network), and determine a proper objective function to train $\theta$ such that $u_{\theta}$ satisfies the Poisson equation \eqref{eq:poisson} and equals to $g$ on $\partial \Omega$.

We have seen a simple trick to parameterize $u$ as a deep neural network which automatically satisfies the boundary condition as in \eqref{eq:u-bdry-value}. 
Let us assume $u_{\theta}$ is constructed in this way and equals to $g$ on $\partial \Omega$.
Then the remaining problem is to construct a proper objective function of $\theta$ such that $-\Delta u_{\theta}(x) =f(x)$ for all $x \in \Omega$.

A simple way to formulate the objective function is based on the error between the two sides of \eqref{eq:poisson}.
Specifically, we find the optimal parameter $\theta$ of $u$ by solving the following minimization problem of $\theta$:
\begin{equation}
\label{eq:pinn}
\min_{\theta \in \mathbb{R}^{m}} \ \int_{\Omega} | \Delta u_{\theta} (x) + f(x) |^{2} \, dx .
\end{equation}
As we can see, if $\theta$ attains objective function value $0$ in \eqref{eq:pinn}, then $u_{\theta}$ is a solution satisfying the Poisson equation \eqref{eq:poisson}.

Since an exact evaluation of the integral in \eqref{eq:pinn} is not available and the dimension of $\Omega \subset \mathbb{R}^{d}$ is high, e.g., $d \ge 5$, we approximate it using Monte Carlo integration\index{Monte Carlo integration} (see Appendix \ref{appsec:mc}).
More precisely, we notice that the integral of any (integrable) function $h: \Omega \to \mathbb{R}$ can be approximated as follows,
\begin{equation}
\label{eq:mc-int}
\int_{\Omega} h(x ) \, dx \approx \frac{1}{N} \sum_{i=1}^{N} h(x^{(i)}) ,
\end{equation}
where $\{x^{(i)} \in \Omega: i \in [N]\}$ are $N$ i.i.d.~samples drawn from the uniform distribution in $\Omega$.
It is easy to check the right-hand side above is an unbiased estimate of the left-hand side.
More details of Monte Carlo integrations can be found in Appendix \ref{appsec:mc}.

In light of \eqref{eq:mc-int}, we can approximate the objective function in \eqref{eq:pinn} by $\ell: \mathbb{R}^{m} \to \mathbb{R}$ defined as
\begin{equation}
\label{eq:pinn-loss}
\ell(\theta) := \frac{1}{N} \sum_{i=1}^{N} | \Delta u_{\theta} (x^{(i)}) + f(x^{(i)}) |^{2}, 
\end{equation}
where $x^{(i)}$'s are i.i.d.\ samples drawn from the uniform distribution in $\Omega$ as above.
All terms in \eqref{eq:pinn-loss} can be evaluated in practice.
In particular, the Laplace operator, which is the composition of the divergence operator and the gradient operator, can be computed using automatic differentiation, the details of which are provided in Section \ref{sec:autodiff}.
Then, we can apply some numerical optimization algorithm (see details in Chapter \ref{chpt:opt}) to find the minimizer $\theta$ of $\ell$ in \eqref{eq:pinn-loss}, and then $u_{\theta}$ is the approximate solution to the Poisson equation \eqref{eq:poisson}.
This method of using deep neural networks to solve high-dimensional PDEs was presented and named as physics-informed neural networks\index{Neural network!physics-informed} in \cite{raissi2019physics-informed}.
It has generated substantial attentions and numerous followups, variants, and applications since developed.

We can also design other objective functions to train the network parameter $\theta$ given the Poisson equation \eqref{eq:poisson}.
For example, the solution $u$ to the Poisson equation \eqref{eq:poisson} is known to solve the following energy minimization problem
\begin{equation}
\label{eq:deep-ritz}
\min_{u} \int_{\Omega} \Big( \frac{1}{2} |\nabla u(x)|^{2} - f(x) u(x) \Big) \, dx
\end{equation}
provided $u=g$ on $\partial \Omega$.
(This can be verified by calculus of variations in the basic PDE theory. The idea is similar to that in Section \ref{subsec:oc-el}. We omit the details here.)
Therefore, if we parameterize $u: \Omega \to \mathbb{R}$ as a neural network like \eqref{eq:u-bdry-value} such that it meets the boundary value condition, we can approximate the integral in \eqref{eq:deep-ritz} by its Monte Carlo integration
\begin{equation}
\label{eq:deep-ritz-loss}
\ell(\theta) : = \frac{1}{N} \sum_{i=1}^{N} \Big( \frac{1}{2}| \nabla u_{\theta} (x^{(i)}) |^{2} - f(x^{(i)}) u(x^{(i)}) \Big) .
\end{equation}
Minimizing $\ell$ in \eqref{eq:deep-ritz-loss} also yields some $\theta$ such that $u_{\theta}$ is an approximate solution to the Poisson equation \eqref{eq:poisson}.
This method of using energy functional as the training loss function was developed and termed as the deep Ritz method in \cite{e2018deep}.

Compared to \eqref{eq:pinn} which requires second-order partial derivatives in the Laplace operator, the advantage of \eqref{eq:deep-ritz-loss} is that only the first-order gradients of $u$ at the sampled points present.
Notice that, when we apply gradient-based optimization algorithms (as those given in Chapter \ref{chpt:opt}) to find minimizers of $\ell$ in either \eqref{eq:pinn-loss} or \eqref{eq:deep-ritz-loss}, another gradient with respect to $\theta$ is required in each iteration of the optimization process.

Note that, however, only a small number of PDEs are known to have energy form as \eqref{eq:deep-ritz}.
Alternately, theory and applications of PDEs often resort to their weak forms. 
Indeed, many PDEs in real-world applications only have solutions to their weak forms (and these solutions are called weak solutions).
In these cases, we do not expect to find any strong solution (also called classical solution) like those that are twice differentiable in \eqref{eq:poisson} as they do not exist.
(However, if a strong solution does exist, then it is also a weak solution. Therefore, we will not miss any strong solution when we consider the weak form of PDEs.)

The weak form of the Poisson equation \eqref{eq:poisson} can be derived as follows.
Again we first assume $u$ satisfies the boundary condition $u = g$ on $\partial \Omega$. By multiplying an arbitrary test function $\psi \in C_{0}^{\infty}(\Omega)$ on both sides of \eqref{eq:poisson}, where $C_{0}^{\infty}(\Omega)$ is the set of smooth functions in $\Omega$ with value zero on the boundary $\partial \Omega$, and integrating on both sides, we obtain
\begin{equation}
\label{eq:weak-poisson}
- \int_{\Omega} \Delta u(x) \psi(x) \, dx = \int_{\Omega} f(x) u(x) \, dx .
\end{equation}
Integrating by parts of the left-hand side and rearranging the equality \eqref{eq:weak-poisson}, we have the weak form of the Poisson equation \eqref{eq:poisson}:
\begin{equation}
\label{eq:weak-poisson-final}
\int_{\Omega} \Big( \nabla u(x) \cdot \nabla \psi(x) - f(x) \psi(x) \Big) \, dx = 0, \qquad \forall \, \psi \in C_{0}^{\infty}(\Omega) .
\end{equation}
Based on \eqref{eq:weak-poisson-final}, we can parameterize $\psi$ with zero boundary value on $\partial \Omega$ as an arbitrary neural network $\psi_{\eta}$ with parameter $\eta$, and try to find the optimal parameters $\theta$ and $\eta$ by solving the following saddle point problem:
\begin{equation}
\label{eq:wan-loss}
\min_{\theta} \max_{\eta} \frac{1}{2N} \Big|\sum_{i=1}^{N} \Big( \nabla u_{\theta}(x^{(i)}) \cdot \nabla \psi(x^{(i)}) - f(x^{(i)}) \psi(x^{(i)}) \Big) \Big|^{2} ,
\end{equation}	
where $\{x^{(i)} \in \Omega: i \in [N]\}$ are again $N$ i.i.d.~samples drawn from the uniform distribution in $\Omega$.
Once we found the optimal $\theta$, the corresponding $u_{\theta}$ is the weak solution to \eqref{eq:weak-poisson-final}.
In practice, there are many different numerical tricks to tackle the optimization problem \eqref{eq:weak-poisson-final} more efficiently and stably.

The benefit of \eqref{eq:wan-loss} is that only gradients appear in the objective function.
More importantly, it enables us to find the weak solutions of the Poisson equation \eqref{eq:poisson} if exist.
This method of using the weak form of a given PDE and parameterizing the test function as an adversarial network was developed and termed as the weak adversarial networks\index{Neural network!weak adversarial} in \cite{zang2020weak}.
\end{example}

\section{Remarks and References}

\subsection{Remarks on Architecture Design}

We observe a significant gap between the theoretical study of the approximation capabilities of deep neural networks and their practical design and deployment. From a theoretical perspective, research typically focuses on identifying which classes of functions can be approximated by neural networks, how such approximations should be evaluated under rigorous mathematical criteria, and how large the networks must be to achieve a desired level of accuracy. In contrast, practical research primarily emphasizes the design of efficient network architectures, often guided by heuristics, to solve specific tasks. This includes the widely adopted U-Net \cite{ronneberger2015u-net:} for image processing and transformer \cite{vaswani2017attention} for natural language processing, which are effective and powerful network architectures not discussed in this book. However, the underlying theoretical questions about their approximation errors are rarely examined in depth.

This disparity is understandable. The remarkable empirical success of deep learning has encouraged researchers to tackle a wide range of practical problems and to develop new methods that extend the applicability of neural networks to tasks previously considered intractable. These advances have been driven largely by the availability of massive datasets and the dramatic increase in computational power compared to several decades ago. Notably, many modern deep networks lack formal theoretical guarantees regarding their representation capabilities, and their sizes are often far smaller than the theoretical bounds---typically upper bounds---suggested by approximation theory. A representative example of this tension is the ongoing theoretical effort to determine whether deep networks can overcome the curse of dimensionality in general settings, while practitioners largely bypass this question and instead implement highly expressive architectures that demonstrate impressive empirical performance across numerous application domains. It remains unclear whether this gap between theory and practice will eventually be closed; however, its existence has not significantly hindered progress on either front.

Ultimately, it may be more realistic to acknowledge the necessity of designing diverse and carefully crafted network architectures to achieve practical efficiency. In nature, the diversity of animal species enables specialization in particular capabilities---such as exceptional strength, high-speed locomotion, flight, or advanced visual perception---each supported by neural systems adapted to these functions. By analogy, it is reasonable to expect that artificial neural networks should likewise be constructed through specialized and task-driven design principles to achieve optimal performance.

\subsection{References of Network Approximation Theory}
\label{subsec:dnn-ref}

Universal approximation theorem is a foundational result in neural network theory, establishing that neural networks are capable of approximating a wide class of functions to arbitrary accuracy under suitable conditions.
The seminal works \cite{cybenko1989approximation,hornik1989multilayer} showed that feed-forward neural networks can uniformly approximate any continuous function defined on compact domains. 
This result was extended to broader classes of activation functions and approximation in \cite{ellacott1994aspects,hornik1993some,hornik1994degree,hornik1991approximation}. 
In \cite{leshno1993multilayer}, it is further clarified this perspective by proving that a feed-forward network is a universal approximator if and only if its activation function is not a polynomial.
In particular, neural network approximation power was established with sigmoid activation function \cite{barron2002universal}, periodic functions \cite{mhaskar1994dimension-independent,mhaskar1995degree}, and general activation functions \cite{hornik1994degree}.

Approximation theory with explicit quantification on network size and error has been extensively studied.
In \cite{mhaskar1996neural}, such result is established for smooth and analytic functions.
For example, to approximate any function in $W^{k,p}(\Omega)$ (where $p\in[1,\infty]$ and $\Omega \subset \mathbb{R}^{d}$ is compact) with error $\epsilon >0$ in $L^{p}$-norm, it is shown that a shallow network of size $O(\epsilon^{-d/k})$ with non-polynomial smooth activation function is sufficient \cite{mhaskar1996neural}.
A similar estimate was established for shallow ReLU network in \cite{pinkus1999approximation}.
The proof given in Section \ref{sec:uat-proof} follows \cite{yarotsky2017error}, which provides such estimate for functions in $W^{k,\infty}(\Omega)$ using deep ReLU networks.
The estimates above still suffer curse of dimensionality, because the estimated network sizes increase exponentially fast in dimension $d$. 
It is shown that such issue can be lessened in sparse grid setting \cite{montanelli2019error}.
If the functions to be approximated are known to be compositional, then this issue can be substantially reduced \cite{poggio2017when}.
This field is continuously developing and improved estimates are developed.

In summary, the universal approximation theorem guarantees the expressive power of neural networks, while modern refinements elucidate the role of depth, activation functions, and approximation efficiency. Together, these results are important to form the theoretical backbone supporting the widespread empirical success of deep neural networks across diverse application domains.

\chapter{Network Training}
\label{chpt:opt}

In a deep learning project, once the network architecture and objective function for training have been specified, the next major task is to determine the optimal network parameters, for example, by minimizing the objective function (also called loss function in this case). This process is commonly referred to as \emph{network training}. The mathematical techniques used to accomplish this task fall under the broad umbrella of \emph{optimization}, a central topic in applied and computational mathematics.

Numerical optimization is a core area that focuses on the development of algorithms with provable convergence properties for solving parameterized objective functions. This chapter presents the fundamental theoretical foundations of optimization, such as optimality conditions, and automatic differentiation, which plays a crucial role in efficiently computing derivatives in modern deep learning implementations. Furthermore, we introduce several classical algorithms for unconstrained optimization, such as gradient descent (including line search strategies), Newton's method, and quasi-Newton methods, as well as methods for constrained optimization, such as Lagrangian algorithms. The discussion in this part is restricted to the deterministic setting, where the objective function and its gradients with respect to the optimization variables can be computed explicitly.

We also introduce the basic concepts of stochastic optimization, which constitute the dominant class of methods used for training deep neural networks. In particular, we discuss several widely used stochastic optimization algorithms, including stochastic gradient methods, momentum-based variants, and adaptive optimization schemes.

\paragraph{Notation clarification}

In this chapter, we follow the notations in optimization and denote the objective function\index{Objective function} by $f: \mathbb{R}^{n} \to \mathbb{R}$, the variable to be optimized by $x \in \mathbb{R}^{n}$, and the constraint set $\Omega$ is closed in $\mathbb{R}^{n}$.

It is most common that an optimization problem is formed as a minimization problem. In this case, the objective function $f$ is also called the loss function, and an optimal solution is called a minimizer. We typically require $f^{*} := f(x^{*}) = \min_{x \in \Omega} f(x)$ to be a finite number. We will provide more details in Section \ref{sec:opt-basics}.

We often employ optimization methods to find the optimal parameters of neural networks, the process of which is called \emph{network training}. Therefore, objective function $f$ and variable $x$ in this chapter often refer to loss function\index{Loss function} $\ell$ and network parameter $\theta$, respectively, in network training.

\section{Basic Concepts in Optimization}
\label{sec:opt-basics}

As we mentioned earlier, a typical optimization problem is written in the following form:
\begin{equation}
\label{eq:general-min}
\min_{x \in \Omega} \ f(x) ,
\end{equation}
where $\Omega \subset \mathbb{R}^{n}$ is called the \emph{constraint set}\index{Constraint set} (or \emph{admissible set}\index{Admissible set}), $x \in \Omega$ is called the \emph{variable}, and $f : \Omega \to \mathbb{R}$ is called the \emph{objective function}\index{Objective function} (also called the \emph{loss function}\index{Loss function} or \emph{cost function} when it is to be minimized).
The goal of \eqref{eq:general-min} is to find a minimizer $x^{*}$ of $f$ in $\Omega$, namely, $f(x^{*}) \le f(x)$ for all $x \in \Omega$.
Since maximizing an objective function $f$ is equivalent to minimizing $-f$, we generally state an optimization problem as a minimization problem in the remainder of this chapter.

There are several typical ways to categorize optimization problems, which we list below.

\paragraph{Continuous versus Discrete optimization}
Discrete optimization generally refers to the case where the constraint set $\Omega$ in \eqref{eq:general-min} is finite or countably infinite (e.g., $\Omega = \mathbb{N}$, the set of natural numbers). 
By contrast, the constraint set $\Omega$ in a continuous optimization problem is a (union of) connected region(s) in $\mathbb{R}^{n}$. In any of such regions, it is typical that all but the boundary points in $\Omega$ are interior points\index{Interior point}. Here, by $x$ being an interior point of $\Omega$, we mean that there exists a small open neighborhood of $x$ that is contained in $\Omega$. Equivalently, there exists $r>0$ such that 
\begin{equation}
\label{eq:ball-x-r}
B_{r}(x) := \{ y \in \mathbb{R}^{n} : |y - x| < r\} \subset \Omega .
\end{equation}
The set $B_{r}(x)$ is called an \emph{open ball} centered at $x$ with radius $r$. 
The set of interior points of $\Omega$ is denoted by $\mathrm{int}(\Omega)$.

Note that, constraint sets in well defined optimization problems are almost always closed, otherwise $\Omega$ may not contain a minimizer of $f$. For instance, $f(x) = x$ does not have a minimizer in $\Omega = (0,1) \subset \mathbb{R}$.
Normally, continuous optimization problems are easier to solve compared to discrete ones, because we can get some information about the points in the neighborhood (e.g., $B_{r}(x)$) of $x$ using the regularity properties of $f$ there, such as continuity, differentiability, and convexity.
We will only consider continuous optimization in this book, since almost all network training problems in deep learning can be formulated as or approximated by continuous optimization.

\paragraph{Unconstrained versus Constrained optimization}
Unconstrained optimization\index{Unconstrained optimization} refers to the case $\Omega = \mathbb{R}^{n}$. In constrained optimization\index{Constrained optimization!equality constrained}, $\Omega$ can be an explicitly given subset of $\mathbb{R}^{n}$, such as $[-1,1]^{n}$, or more generally described by a set of \emph{inequality} and/or \emph{equality} constraints as 
\begin{equation}
\label{eq:opt-constraint-set}
\Omega = \{ x \in \mathbb{R}^{n}: g(x) \le 0, \ h(x) = 0 \} .
\end{equation}
In \eqref{eq:opt-constraint-set}, we see that the constraint set $\Omega$ is determined by 
\begin{subequations}
\label{eq:ineq-eq-constraints}
\begin{align}
\text{Inequality constraint:} \qquad & g(x) \le 0 , \label{subeq:ineq-constraint}\\
\text{Equality constraint:} \qquad & h(x) = 0 , \label{subeq:eq-constraint}
\end{align}
\end{subequations}
where the functions $g: \mathbb{R}^{n} \to \mathbb{R}^{k}$ and $h: \mathbb{R}^{n} \to \mathbb{R}^{m}$ are given by
\begin{equation*}
g(x) = \begin{pmatrix}
g_{1}(x) \\ \vdots \\ g_{k}(x)
\end{pmatrix} \in \mathbb{R}^{k} 
\qquad \text{and} \qquad
h(x) = \begin{pmatrix}
h_{1}(x) \\ \vdots \\ h_{m}(x)
\end{pmatrix} \in \mathbb{R}^{m} .
\end{equation*}
Note that by $z \le 0$ ($z = 0$) we meant every component of the vector $z$ is non-positive (equal to $0$).
Notice that the equality constraint $h(x) = 0$ can be reduced to the combination of inequality\index{Constrained optimization!inequality constrained} ones $ h(x) \le 0$ and $-h(x) \le 0$. 
However, we rarely do so since equality constraints are generally easier to tackle than inequality constraints.

It is obvious that unconstrained optimization problems are easier to tackle than constrained ones in general. 
While we study the properties of both unconstrained and constrained optimization problems, we often prefer to formulate or approximate network training problems as unconstrained optimization.
This can be realized in various means and is case by case, which we will discuss in detail later in this chapter.

\paragraph{Convex versus Non-convex optimization}
We call $\Omega$ a \emph{convex set}\index{Convex!set} in $\mathbb{R}^{n}$ if for any $x, y \in \Omega$ there is
\begin{equation}
\label{eq:cvx-set-condition}
(1-t) x + t y \in \Omega, \qquad \forall\, t \in [0,1] .
\end{equation}
In other words, $\Omega$ is convex if for any two points $x, y \in \Omega$, the entire line segment between $x$ and $y$ is contained in $\Omega$. 

\begin{figure}
\centering
\begin{tikzpicture}[domain=-0.5:4.5, scale=1, transform shape]
	
	\draw[->] (-1.8, 0) -- (2.5, 0) node[right] {$x$};

	\draw[thick, color=black] plot[domain=-1.5:1.9]({\x}, {\x*\x/2+1/2}) node[right]{$f(x)$};
	\draw[thin, dashed, color=black] plot[domain=-1:1.5]({\x}, {\x/4+1.25});
	\draw[thin, color=gray] plot[domain=-0.7:0.7](\x, 0.5);
	
	\draw (-1,0) -- (-1,0.1);
	\draw (0,0) -- (0,0.1);
	\draw (1.5,0) -- (1.5,0.1);
			
	\draw[black] (-1, -0.08) node[below]{$x_{1}$};
	\draw[black] (0, 0.01) node[below]{$x^{*}$};
	\draw[black] (1.5, -0.08) node[below]{$x_{2}$};
\end{tikzpicture}
\qquad \quad
\begin{tikzpicture}[domain=-0.5:4.5, scale=1, transform shape]

	\draw[->] (-1.8, 0) -- (2.5, 0) node[right] {$x$};

	\draw[thick] plot[domain=-1.5:-0.5]({\x}, {1.35417*\x*\x*\x+3.03125*\x*\x+1.0156*\x+0.6693});
	\draw[thick] plot[domain=-0.5:1]({\x}, {\x*\x + 0.5});
	\draw[thick] plot[domain=1:1.75]({\x}, {-4*\x*\x*\x + 13*\x*\x - 12*\x +4.5});
	\draw[thick] plot[domain=1.75:2.2]({\x}, {38.6667*\x*\x*\x-211*\x*\x+380*\x-224.1667}) node[right]{$f(x)$};
	
	\draw[thin, color=gray] plot[domain=-1.5:-1.1](\x, 1.4967);
	\draw[thin, color=gray] plot[domain=-0.4:0.4](\x, 0.5);
	\draw[thin, color=gray] plot[domain=1.2:1.8](\x, 2.25);
	\draw[thin, color=gray] plot[domain=1.8:2.2](\x, 1.1669);
	
	\draw (-1.3,0) -- (-1.3,0.1);
	\draw (0,0) -- (0,0.1);
	\draw (1.5,0) -- (1.5,0.1);
	\draw (2,0) -- (2,0.1);
			
	\draw[black] (-1.3, -0.08) node[below]{$x_{3}$};
	\draw[black] (-0, 0.01) node[below]{$x^{*}$};
	\draw[black] (1.5, -0.08) node[below]{$x_{4}$};
	\draw[black] (2, -0.08) node[below]{$x_{5}$};
\end{tikzpicture}
\caption{(Left) A convex function $f$. Any critical point, such as $x^{*}$, where $f$ has a flat tangent line (i.e., $\nabla f(x^{*}) = 0$ if $f$ is differentiable at $x^{*}$), must be a global minimizer of $f$. For any $x_{1}$ and $x_{2}$, the line segment (dashed) between $(x_{1},f(x_{1}))$ and $(x_{2},f(x_{2}))$ is always on or above $f(x)$ on the line segment between $x_{1}$ and $x_{2}$; (Right) A non-convex function $f$. The critical points can be local minimizers such as $x^{*}$ and $x_{5}$, local maximizers such as $x_{3}$ and $x_{4}$, or saddle points (not shown here). Within the interval where $f$ is plotted, $x^{*}$ is the global minimizer.}
\label{fig:cvx-noncvx}
\end{figure}
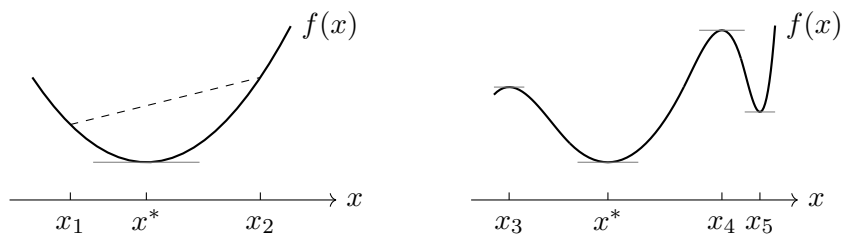

Let $\Omega$ be a convex subset of $\mathbb{R}^{n}$. Suppose $f: \Omega \to \mathbb{R}$ is a function defined on $\Omega$. Then we call $f$ a \emph{convex function} if 
\begin{equation*}
\label{eq:cvx-fn-condition}
f((1-t) x + ty) \le (1-t) f(x) + t f(y), \qquad \forall\, t \in [0,1].
\end{equation*}
In other words, a function $f$ is convex\index{Convex} if and only if for any two points $x,y \in \Omega$, the secant line segment, i.e., the straight line segment in $\mathbb{R}^{n}$ with $(x,f(x))$ and $(y,f(y))$ as the end points, is always on or above $f$ when restricted to the line segment $\{(1-t)x+ty : t\in[0,1]\}$. 
Figure \ref{fig:cvx-noncvx} shows the examples of a convex function\index{Convex!function} (left) and a non-convex (right) function.
We call $f$ a \emph{concave function}\index{Concave} if $-f$ is a convex function.

The problem \eqref{eq:general-min} is called a convex optimization\index{Convex!optimization} problem if $\Omega$ is a convex set and $f$ is a convex function. 
If $\Omega$ is determined by both inequality constraint\index{Constrained optimization!inequality constrained} $g(x)\le 0$ and equality constraint $h(x) = 0$ as in \eqref{eq:ineq-eq-constraints}, then we usually require that $g_{i}$'s are convex functions and $h_{j}$'s are affine functions for all $i \in [k]$ and $j \in [m]$. 
Such requirement makes a sufficient (but not necessary) condition for $\Omega$ to be convex. 
This convention of requirement is because the sub-level set $\{x \in \mathbb{R}^{n} : g_{i}(x)\le 0 \}$ must be convex when $g_{i}$ is a convex function, and the intersection of convex sets $\{x: g(x) \le 0\} = \cap_{i=1}^{k} \{x: g_{i}(x) \le 0\}$ is also a convex set. For the same reason, both $h_{j}$ and $-h_{j}$ need to be convex (because $h(x) = 0$ is equivalent to $h(x)\le 0$ and $-h(x) \le 0$), which means that $h_{j}$ is affine for every $j$.

Convex functions have many important and useful properties. For this reason, there are numerous theoretical and algorithmic results developed for convex optimization. However, the optimization problems (particularly network training) in deep learning applications are almost always non-convex. Therefore, we will mostly consider non-convex optimization in this book.

Convexity and non-convexity make a significant difference on our expectation to obtain the ``best'' or ``optimal'' solution to an optimization problem. 
To clarify this, we first define local and global optimality.
We call $x^{*}$ a \emph{local minimizer}\index{Minimizer!local} of the problem \eqref{eq:general-min} if there exists $r>0$ such that 
\begin{equation}
\label{eq:local-min-condition}
f(x) \ge f(x^{*}), \qquad \forall\, x \in \Omega \cap B_{r}(x^{*}) .
\end{equation}
In addition, we call $x^{*}$ a \emph{strict} local minimizer if the equality in \eqref{eq:local-min-condition} holds only if $x = x^{*}$. 
We call $x^{*}$ a \emph{global minimizer}\index{Minimizer!global} of the problem \eqref{eq:general-min} if 
\begin{equation}
\label{eq:global-min-condition}
f(x) \ge f(x^{*}), \qquad \forall\, x \in \Omega .
\end{equation}
We call $x^{*}$ a strict global minimizer of \eqref{eq:general-min} if the equality in \eqref{eq:global-min-condition} holds only if $x=x^{*}$. Similarly, we can define (strict) local and global maximizers of $f$ in $\Omega$.
Figure \ref{fig:cvx-noncvx} also shows some examples of local/global minimizers/maximizers.
It is obvious that a global minimizer must be a local minimizer, but the converse is not necessarily true.

We remark that one of the major properties of convex optimization\index{Convex!optimization} is that any local minimizer is a global minimizer, as we show in the following proposition.

\begin{proposition}
Suppose $\Omega \subset \mathbb{R}^{n}$ is a convex set and $f: \Omega \to \mathbb{R}$ is a convex function. If $x^{*} \in \Omega$ is a local minimizer of $f$, then $x^{*}$ must be a global minimizer of $f$ in $\Omega$.
\end{proposition}

\begin{proof}
Since $x^{*}$ is a local minimizer of $f$ in $\Omega$, we know by definition that there exists $r > 0$ such that \eqref{eq:local-min-condition} holds.

Now we use proof by contradiction. Suppose there exists another point $x \in \Omega$ such that $f(x) < f(x^{*})$. We choose any $\alpha \in (0, \min\{ r, |x-x^{*}|\})$, and set $t = \alpha / |x - x^{*}|$ and $y = (1-t)x^{*} + tx$. Then $t \in (0,1)$, $y \in \Omega$ (since $\Omega$ is a convex set and $x,x^{*} \in \Omega$), and
\begin{equation*}
|y - x^{*}| = t| x - x^{*}| = \alpha < r,
\end{equation*}
which means that $y \in B_{r}(x^{*})$. Therefore, we get
\begin{align*}
f(y) = f( (1-t)x^{*} + t x) \le (1-t) f(x^{*}) + t f(x) < f(x^{*}),
\end{align*}
where the first inequality is due to the convexity of $f$ and the second due to $f(x) < f(x^{*})$ and $t\in(0,1)$. This is a contradiction to \eqref{eq:local-min-condition}.
\end{proof}

\section{Optimality Conditions}
\label{sec:opt-optcond}

To prepare the discussion on optimality conditions of solutions to optimization problems, we first fix some writing formats about gradients, Hessian, and the chain rule, which make presentations more convenient in the remainder of this section.

\paragraph{Gradients and Chain Rule Format}
Suppose $f: \mathbb{R}^{n} \to \mathbb{R}^{m}$ is differentiable. If $m=1$, then we denote the \emph{gradient}\index{Gradient} of $f$ at $x$ as a row vector
\begin{equation}
\label{eq:def-gradient}
\nabla f(x) = \big( \partial_{1}f(x), \dots, \partial_{n} f(x) \big) \in \mathbb{R}^{1 \times n},
\end{equation}
where $\partial_{i}f(x) := \frac{\partial f}{\partial x_{i}}(x)$. If $m > 1$ and we denote $f(x) = (f_{1}(x), \dots, f_{m}(x))^{\top} \in \mathbb{R}^{m}$ as a column vector, then the \emph{Jacobian}\index{Jacobian} of $f$ at $x$ is an $m \times n$ matrix
\begin{equation}
\label{eq:def-jacobian}
\nabla f(x) =
\begin{pmatrix}
\nabla f_{1}(x) \\ \vdots \\ \nabla f_{m}(x)
\end{pmatrix}
=
\begin{pmatrix}
\partial_{1} f_{1}(x) & \partial_{2} f_{1}(x) & \cdots & \partial_{n} f_{1}(x) \\
\partial_{1} f_{2}(x) & \partial_{2} f_{2}(x) & \cdots & \partial_{n} f_{2}(x) \\
\vdots & \vdots & \ddots & \vdots \\
\partial_{1} f_{m}(x) & \partial_{2} f_{m}(x) & \cdots & \partial_{n} f_{m}(x)
\end{pmatrix} \in \mathbb{R}^{m \times n} .
\end{equation}
Note that the size of the Jacobian in \eqref{eq:def-jacobian} is consistent with the size of gradient in \eqref{eq:def-gradient} when $m=1$. Keeping this format will help us to track the dimensions of the involved matrices when we apply the chain rule to compositions of functions below.

Suppose in addition that $g: \mathbb{R}^{m} \to \mathbb{R}^{k}$ is another differentiable function. Then $g\circ f: \mathbb{R}^{m} \to \mathbb{R}^{k}$ is differentiable. If we apply the chain rule to $g \circ f$, we are supposed to obtain a $k \times n$ Jacobian matrix, which is written in the \emph{left-to-right} order of the gradients applied to the outside function $g$ and then the inside function $f$:\index{Chain rule}
\begin{equation}
\label{eq:chain-rule}
\underbrace{\nabla (g \circ f)(x)}_{k \times n} = \underbrace{\nabla_{f} g(f(x))}_{k \times m} \underbrace{\nabla f(x)}_{m \times n} ,
\end{equation}
where the size of each matrix is shown under the corresponding brace.
Similarly, if there is also $h : \mathbb{R}^{k} \to \mathbb{R}^{l}$, then $h \circ g \circ f : \mathbb{R}^{n} \to \mathbb{R}^{l}$ and we follow the left-to-right rule above to obtain
\begin{equation*}
\underbrace{\nabla (h \circ g \circ f)(x)}_{l \times n} = \underbrace{\nabla_{g} h (g(f(x)))}_{l \times k} \underbrace{\nabla_{f} g(f(x))}_{k \times m} \underbrace{\nabla f(x)}_{m \times n} .
\end{equation*}
The same rule applies to compositions of more than three functions.

If $f: \mathbb{R}^{n} \to \mathbb{R}$ is twice differentiable, then we denote the \emph{Hessian}\index{Hessian} of $f$ at $x$ by
\begin{equation}
\label{eq:def-hessian}
\nabla^{2}f(x) = 
\begin{pmatrix}
\partial_{11}^{2} f(x) & \partial_{12}^{2} f(x) & \cdots & \partial_{1n}^{2} f(x) \\
\partial_{21}^{2} f_{2}(x) & \partial_{22}^{2} f(x) & \cdots & \partial_{2n}^{2} f(x) \\
\vdots & \vdots & \ddots & \vdots \\
\partial_{n1}^{2} f_{m}(x) & \partial_{n2}^{2} f(x) & \cdots & \partial_{nn}^{2} f(x)
\end{pmatrix} \in \mathbb{R}^{n \times n} ,
\end{equation}
where $\partial_{ij}^{2}f(x) := \frac{\partial^{2}f}{\partial x_{i} \partial x_{j}}(x)$.
We know that $\nabla^{2}f(x)$ is symmetric if $f$ is twice continuously differentiable at $x$.

Let ``$\mathrm{tr}$'' denote the \emph{trace} of a square matrix. If $f: \mathbb{R}^{n} \to \mathbb{R}^{n}$ is a differentiable vector field, then the \emph{divergence} of $f$ at $x$ is given by
\begin{equation}
\label{eq:def-divergence}
\nabla \cdot f(x) := \mathrm{tr}(\nabla f(x)) = \sum_{i=1}^{n} \partial_{i} f_{i}(x) ,
\end{equation}
which is a scalar.
If $f : \mathbb{R}^{n} \to \mathbb{R}$ is a twice differentiable scalar-valued function, then the \emph{Laplacian} of $f$ at $x$ is defined by
\begin{equation}
\label{eq:def-laplacian}
\Delta f(x) := \mathrm{tr}(\nabla^{2} f(x)) = \sum_{i=1}^{n} \partial_{ii}^{2} f(x) .
\end{equation} 

\subsection{General Principles of Optimality Conditions}

Given a subset $\Omega$ of $\mathbb{R}^{n}$, we call a nonzero vector $v \in \mathbb{R}^{n}$ a \emph{feasible direction}\index{Feasible direction} at $x \in \Omega$ if there exists $r>0$ such that $ x + \alpha v \in \Omega$ for all $ \alpha \in [0,r)$. 
Let $f: \Omega \to \mathbb{R}$ be differentiable. 
We denote the \emph{directional derivative} of $f$ at $x$ in the direction $v$ by
\begin{equation}
\label{eq:def-directional-derivative}
D_{v} f(x) := \langle \nabla f(x), v \rangle = \frac{d}{d \alpha} f(x + \alpha v) \Big|_{\alpha = 0} = \lim_{\alpha \to 0^{+}} \frac{f(x+\alpha v) - f(x)}{\alpha} ,
\end{equation}
which is the derivative of $f$ at $x$ when restricted to the ray $\{x + \alpha v: \alpha \ge 0 \}$.

\paragraph{First-order necessary condition}

If $x \in \Omega$ is a local minimizer\index{Minimizer!local} of $f$, then it is necessary that the directional derivative of $f$ at $x$ is non-negative for any feasible direction $v$ at $x$. This is called the \emph{first-order necessary condition}\index{Necessary condition!first-order} of local minimizers for the optimization problem \eqref{eq:general-min} with differentiable $f$.

Note that, if $x$ is an interior point of $\Omega$, then $B_{r}(x) \subset \Omega$ for some $r>0$. Therefore, we know any nonzero vector $v\in \mathbb{R}^{n}$ is a feasible direction at $x \in \Omega$ because $x + \alpha v \in B_{r}(x) \subset \Omega$ for any $\alpha \in [0, r/|v|)$. 
This implies that $D_{v} f(x) = \langle \nabla f(x), v \rangle = 0$ for any $v \in \mathbb{R}^{n}$ and hence $\nabla f(x) = 0$. Therefore, an interior local minimizer must be a critical point of $f$.
This is the reason that we often seek for the critical points of $f$ in an optimization problem, because all interior local minimizers of $f$ are among them.
This fact is particularly useful in the unconstrained case $\Omega = \mathbb{R}^{n}$ where every point is an interior point.

\paragraph{Second-order necessary condition}

Suppose that $f: \mathbb{R}^{n} \to \mathbb{R}$ is twice continuously differentiable. If $x \in \Omega$ is a local minimizer of $f$ and $v$ is a feasible direction satisfying $D_{v}f(x) = \langle \nabla f(x) , v \rangle = 0$, then by Taylor's theorem, $f(x+\alpha v)$ as a function of $\alpha$ satisfies
\begin{equation}
\label{eq:sonc-with-v-pre}
0 \le f(x+ \alpha v) - f(x) = \frac{\alpha^{2}}{2} v^{\top} \nabla^{2}  f(x + \xi \alpha v) v
\end{equation}
for any sufficiently small $\alpha>0$ and some $\xi = \xi(\alpha, x, v) \in [0,1]$.
Since $\nabla^{2} f$ is continuous, we know that, as $\alpha \to 0^{+}$, \eqref{eq:sonc-with-v-pre} implies
\begin{equation}
\label{eq:sonc-with-v}
v^{\top} \nabla^{2} f(x) v \ge 0.
\end{equation}
This \eqref{eq:sonc-with-v} is called the \emph{second-order necessary condition}\index{Necessary condition!second-order} for the optimization problem \eqref{eq:general-min}.

Notice that, if $x$ is an interior point of $\Omega$ and a local minimizer of $f$, then we know $\nabla f(x) = 0$ and any direction $v$ is a feasible direction. Therefore, we deduce from \eqref{eq:sonc-with-v} that $\nabla^{2} f(x) \succeq 0$, i.e., $\nabla^{2}f(x)$ is a positive semi-definite matrix.
Therefore, $\nabla^{2} f(x) \succeq 0$ is the second-order necessary condition\index{Necessary condition!second-order} for a local minimizer $x$ when $x$ is an interior point of $\Omega$ in \eqref{eq:general-min}.
If in addition $\nabla^{2} f(x) \succ 0$, then $x$ is a strict local minimizer of $f$, and we call $\nabla^{2} f(x) \succ 0$ the second-order sufficient condition\index{Sufficient condition!second-order} for a local minimizer $x$ of $f$.

\begin{remark}
We remark that the first- and second-order necessary conditions are \emph{not} sufficient for non-convex functions in general. For example, for $f:\mathbb{R}\to\mathbb{R}$ defined by $f(x) = x^{3}$, there is $f'(0) = f''(0) = 0$, which means that $f$ satisfies both conditions at $x=0$. However $x=0$ is a saddle point of $f$, not a local minimizer or maximizer.
\end{remark}

The first- and second-order necessary conditions above provide concise characterizations of local minimizers and are particularly informative when they are interior points of $\Omega$.
However, we are not able to verify them in practice when a local minimizer $x$ is on the boundary of $\Omega$ in constrained optimization problems.
This is because it is difficult to determine the set of all feasible directions at $x$ and check these necessary conditions. 
Therefore, we need more practically useful necessary conditions for constrained optimization, as we show next.

\subsection{Necessary Optimality Conditions}

We begin by considering the first- and second-order necessary conditions for the optimization problem with equality constraint only:
\begin{subequations}
\label{eq:min-eq-constraint}
\begin{align}
\min_{x \in \mathbb{R}^{n} } \quad & f(x) , \label{subeq:min-eq-constraint-f} \\
\text{s.t.} \quad & h(x) = 0 . \label{subeq:min-eq-constraint-h}
\end{align}
\end{subequations}
We require that the constraint function $h: \mathbb{R}^{n} \to \mathbb{R}^{m}$ with $m \le n$; otherwise the constraint is either over-redundant (some constraints are redundant and can be removed) or overdetermined (there are too many constraints which result in an empty set $\Omega$), and we can modify $h$ and convert the problem \eqref{eq:min-eq-constraint} to an equivalent one with $m \le n$.
With $m \le n$, we know that the rank of the Jacobian $\nabla h(x)$ cannot be greater than $m$ at any $x$.
Now we introduce the definition of regular points.
\begin{definition}
[Regular point]
We call $x$ a \emph{regular point}\index{Regular point!with equality constraint} of \eqref{subeq:min-eq-constraint-h} if the Jacobian matrix of $h$ at $x$ has full row rank $m$.
\end{definition}

\begin{definition}
[Tangent space]
Denote the constraint set in \eqref{subeq:min-eq-constraint-h} by $\Omega = \{ x \in \mathbb{R}^{n}: h(x) = 0\}$. Then the \emph{tangent space}\index{Tangent space} of $\Omega$ at $x \in \Omega$ is defined by
\begin{equation}
\label{eq:tangent-space}
T(x) = \{ v \in \mathbb{R}^{n} : \nabla h(x) v = 0 \} \subset \mathbb{R}^{n} .
\end{equation}
Namely, $T(x)$ is the orthogonal complement of the row space of the matrix $\nabla h(x)$. 
%
% Notice that $\dim(T(x)) = n - \dim(\text{Row}(\nabla h(x))) = n - m$.
\end{definition}

\begin{lemma}
\label{lem:regular-pt}
Suppose $x \in \Omega = \{x \in \mathbb{R}^{n} : h(x) = 0 \}$ is a regular point. Then $v \in T(x)$ if and only if there exists a curve $c: (-\delta, \delta) \to \Omega$ such that $c(0) = x$ and $c'(0) = v$.
\end{lemma}

\begin{proof}
We first show the necessity. Suppose $c(t)$ is a curve satisfying the conditions mentioned above. Then $c(t) \in \Omega$, i.e., $h(c(t)) = 0$, for all $t \in (-\delta, \delta)$. Taking derivative with respect to $t$, we obtain that
\begin{equation*}
0 = \frac{d}{dt} h(c(t)) \Big|_{t = 0} = \nabla h(c(0)) c'(0) = \nabla h(x) v .
\end{equation*}
Therefore $v \in T(x)$ due to \eqref{eq:tangent-space}.

Next we show the sufficiency. Suppose $x \in \Omega$ is a regular point and $v \in T(x)$. We define $\bar{h}: \mathbb{R} \times \mathbb{R}^{m} \to \mathbb{R}^{m}$ by
\begin{equation*}
\bar{h}(t, w) = h(x + tv + \nabla h(x)^{\top} w) .
\end{equation*}
Then we have $\bar{h}(0, 0) = h(x) = 0$ and
\begin{equation*}
\partial_{w} \bar{h}(t, w) |_{(t, w) = (0, 0)} = \nabla h(x) \nabla h(x)^{\top} \succ 0 ,
\end{equation*}
since $\nabla h(x) \in \mathbb{R}^{m \times n}$ has full row rank $m$ and $m \le n$. 
By the Implicit Function Theorem, we know there exists a neighborhood $(-\delta, \delta)$ of $t=0$ for some $\delta>0$, such that there exists $u: (-\delta, \delta) \to \mathbb{R}^{m}$ satisfying $(0,u(0)) = (0,0)$ and
\begin{equation}
\label{eq:regular-pt-pf}
\bar{h}(t, u(t)) = h(x + t v + \nabla h(x)^{\top} u(t)) = 0
\end{equation}
for all $t \in (-\delta, \delta)$. 
Notice that
\begin{equation*}
0 = \frac{d}{dt} h(x + tv + \nabla h(x)^{\top} u(t)) \Big|_{t=0} = \nabla h(x)  (v + \nabla h(x)^{\top} u'(0)).
\end{equation*}
Since $\nabla h(x) v = 0$ and $\nabla h(x) \nabla h(x)^{\top} \succ 0$, we know $u'(0) = 0$.

Define the curve $c(t) = x + t v + \nabla h(x)^{\top} u(t)$, then $h(c(t)) = 0$, i.e., $c(t) \in \Omega$, for all $t \in (-\delta, \delta)$ due to \eqref{eq:regular-pt-pf}. Moreover, $c(0) = x + \nabla h(x)^{\top}u(0) = x$ and $c'(0) = v + \nabla h(x)^{\top} u'(0) = v$. This completes the proof.
\end{proof}

\begin{theorem}
[Lagrange]
\label{thm:lagrange-eq-constraint}
Suppose $f, h \in C^{1}$. If $x$ is a local minimizer\index{Minimizer!local} of \eqref{eq:min-eq-constraint} and $x \in \Omega = \{x \in \mathbb{R}^{n} : h(x) = 0 \}$ is a regular point, then there exists $\lambda \in \mathbb{R}^{m}$ such that
\begin{equation*}
\nabla f(x) + \lambda \nabla h(x) = 0 .
\end{equation*}
\end{theorem}

\begin{proof}
By Lemma \ref{lem:regular-pt}, we know for any $v \in T(x)$ there exist $\delta>0$ and a curve $c: (-\delta, \delta) \to \Omega$ such that $c(0) = x$ and $c'(0) = v$. 

Since $x$ is a local minimizer of $f$, we know 
\begin{equation*}
0 = \frac{d}{dt} f(c(t)) \Big|_{t = 0} = \nabla f(c(0)) c'(0) = \nabla f(x) v .
\end{equation*}
Since $v \in T(x)$ is arbitrary, we know $\nabla f(x)$ is in the row space of $\nabla h(x)$.
This implies that there exists (a row vector) $\lambda \in \mathbb{R}^{m}$ such that $\nabla f(x) + \lambda \nabla h(x) = 0$.
\end{proof}

\begin{remark}
Note that $x\in \Omega$ being a regular point in Theorem \ref{thm:lagrange-eq-constraint} is required: Consider the problem \eqref{eq:min-eq-constraint} with $f,h: \mathbb{R} \to \mathbb{R}$ defined by
\begin{equation*}
f(x) = x \quad \text{and} \quad h(x) = 
\begin{cases}
\frac{1}{2}x^{2}, & \text{if} \quad x < 0, \\
0, & \text{if} \quad 0 \le x \le 1, \\
\frac{1}{2}(x-1)^{2}, & \text{if} \quad x > 1 .
\end{cases}
\end{equation*}
Then it is easy to show that $x=0$ is the minimizer of $f$ under the constraint $h(x) = 0$. Moreover, $f,h \in C^{1}$, $f'(0) = 1$ and $h'(0) = 0$ (so $x=0$ is not a regular point), and the conclusion of Theorem \ref{thm:lagrange-eq-constraint} does not hold.
\end{remark}

To memorize the result of Theorem \ref{thm:lagrange-eq-constraint}, we define the \emph{Lagrange equation}\index{Lagrange!equation} of $(x,\lambda)$ as 
\begin{equation*}
L(x, \lambda) = f(x) + \lambda h(x) .
\end{equation*}
Then Theorem \ref{thm:lagrange-eq-constraint} implies that, if $x \in \Omega$ is a regular point and a local minimizer\index{Minimizer!local} of the equality constrained optimization problem \eqref{eq:min-eq-constraint} with $f \in C^{1}(\mathbb{R}^{n}; \mathbb{R})$ and $h \in C^{1}(\mathbb{R}^{n}; \mathbb{R}^{n})$, then there exists $\lambda \in \mathbb{R}^{m}$, called the \emph{Lagrange multiplier}\index{Lagrange!multiplier}, such that 
\begin{subequations}
\label{eq:lagrange-condition-eq-constraint}
\begin{align}
\nabla f(x) + \lambda \nabla h(x) & = 0 , \label{subeq:lagrange-condition-eq-constraint-dLdx} \\
h(x) & = 0 . \label{subeq:lagrange-condition-eq-constraint-dLdlambda}
\end{align}
\end{subequations}
We can memorize \eqref{eq:lagrange-condition-eq-constraint} as $(\partial_{x} L(x,\lambda), \partial_{\lambda} L(x, \lambda)) = (0,0)$.

If $f, h \in C^{2}$, then we can continue from Theorem \ref{thm:lagrange-eq-constraint} to establish the \emph{second-order necessary condition}\index{Necessary condition!second-order} for \eqref{eq:min-eq-constraint}. 

\begin{theorem}
\label{thm:sonc-eq-constraint}
Suppose $f \in C^{1}(\mathbb{R}^{n}; \mathbb{R})$ and $h \in C^{1} ( \mathbb{R}^{n} ; \mathbb{R}^{m} )$ with $m \le n$. If $x \in \Omega = \{ x \in \mathbb{R}^{n} : h(x) = 0 \}$ is a regular point and a local minimizer of \eqref{eq:min-eq-constraint}, then there exists a Lagrange multiplier $\lambda \in \mathbb{R}^{m}$ (row vector) such that 
\begin{enumerate}
\item
$\nabla f(x) + \lambda \nabla h(x) = 0$; and
\item
For any $v \in T(x)$, there is
\begin{equation}
\label{eq:sonc-eq-constraint}
v^{\top} \Big( \nabla^{2} f(x) + \sum_{i=1}^{m} \lambda_{i} \nabla^{2} h(x) \Big) v \ge 0 .
\end{equation}
\end{enumerate}
\end{theorem}

\begin{proof}
The first claim is just Theorem \ref{thm:lagrange-eq-constraint}. For the second claim, we follow Theorem \ref{thm:lagrange-eq-constraint} to obtain the curve $c: (-\delta, \delta) \to \Omega$ such that $c(0) = x$ and $c'(0) = v$. Now we define $\phi: (-\delta, \delta) \to \mathbb{R}$ by $\phi(t) := f(c(t))$. Since $x$ is a local minimizer and $c(t) \in \Omega$ for all $t \in (-\delta, \delta)$, we know that
\begin{equation}
\label{eq:sonc-eq-constraint-pf-phi}
0 \le \phi''(0) = v^{\top} \nabla^{2} f(x) v + \nabla f(x) c''(0) .
\end{equation}
We further define for every $i \in [m]$ that $\psi_{i}: (-\delta,\delta) \to \mathbb{R}$ by $\psi_{i} (x) := h_{i} (c(t))$. Then, due to $h_{i}(c(t)) = 0$ for all $t \in (-\delta, \delta)$, there is
\begin{equation}
\label{eq:sonc-eq-constraint-pf-psi}
0 = \psi_{i}''(0) = v^{\top} \nabla^{2} h_{i} (x) v + \nabla h_{i}(x)^{\top} c''(0)
\end{equation}
for every $i \in [m]$. Taking the sum of \eqref{eq:sonc-eq-constraint-pf-phi} and \eqref{eq:sonc-eq-constraint-pf-psi} weighted by $\lambda_{i}$ for all $i \in [m]$, and noticing that $\nabla f(x) + \lambda \nabla h(x) = 0$, we obtain \eqref{eq:sonc-eq-constraint}.
\end{proof}

Now we consider the first- and second-order necessary conditions\index{Necessary condition!second-order} for the optimization problem with both equality and inequality constraints\index{Constrained optimization!inequality constrained}:
\begin{subequations}
\label{eq:min-eq-ineq-constraints}
\begin{align}
\min_{x \in \mathbb{R}^{n}} \quad & f(x) , \label{subeq:min-eq-ineq-constraints-obj} \\
\text{s.t.} \quad & g(x) \le 0 , \label{subeq:min-eq-ineq-constarints-g} \\
& h(x) = 0 , \label{subeq:min-eq-ineq-constraints-h}
\end{align}
\end{subequations}
where $g=(g_{1}, \dots, g_{k}) : \mathbb{R}^{n} \to \mathbb{R}^{k}$ and $h=(h_{1}, \dots, h_{m}) : \mathbb{R}^{n} \to \mathbb{R}^{m}$. For any $x$ satisfying both $g(x) \le 0$ and $h(x) = 0$, we split $[k]$ into the \emph{active set}\index{Active set}
\begin{equation}
\label{eq:kkt-active-set}
A(x) := \{ j \in [k] : g_{j} (x) = 0 \}
\end{equation}
and the \emph{inactive set}\index{Inactive set}
\begin{equation}
\label{eq:kkt-inactive-set}
I(x) := [k] \setminus A(x) = \{	j \in [k] : g_{j}(x) < 0 \} .
\end{equation}
Let $\Omega$ denote the constraint set
\begin{equation*}
\Omega = \{ x \in \mathbb{R}^{n} : g(x) \le 0, \ h(x) = 0 \} .
\end{equation*}
Then we call $x \in \Omega$ a \emph{regular point}\index{Regular point!with both equality and inequality constraints} if the vectors
\begin{equation}
\label{eq:kkt-regular-pt-vectors}
\nabla h_{1} (x),\ \dots,\ \nabla h_{m} (x), \quad \text{and} \quad \nabla g_{j}(x), \ \forall\, j \in A(x)
\end{equation}
are linearly independent (note that these are $m + \text{Cardinality}(A(x))$ row vectors in $\mathbb{R}^{n}$).
Then the following result describes the first-order necessary condition of \eqref{eq:min-eq-ineq-constraints}, known as the \emph{Karush--Kuhn--Tucker} (KKT) theorem\index{Karush--Kuhn--Tucker!Theorem}.

\begin{theorem}
[Karush--Kuhn--Tucker]
\label{thm:kkt}
For the constrained optimization problem \eqref{eq:min-eq-ineq-constraints} with $f \in C^{1}(\mathbb{R}^{n}; \mathbb{R})$, $g \in C^{1}(\mathbb{R}^{n} ; \mathbb{R}^{k})$ and $h \in C^{1} (\mathbb{R}^{n} ; \mathbb{R}^{m})$. If $x\in \Omega$ is a regular point and a local minimizer\index{Minimizer!local} of $f$, then there exist $\lambda \in \mathbb{R}^{m}$ and $\mu \in \mathbb{R}^{k}$ (both $\lambda$ and $\mu$ are row vectors), such that 
\begin{subequations}
\label{eq:kkt}
\begin{align}
\nabla f(x) + \lambda \nabla h(x) + \mu \nabla g(x) & = 0 , \label{subeq:kkt-d-lagrange} \\
h(x) & = 0 , \label{subeq:kkt-h} \\
g(x) & \le 0 , \label{subeq:kkt-g} \\
\mu & \ge 0 , \label{subeq:kkt-mu} \\
\mu_{j} g_{j}(x) & = 0 , \quad \forall\, j \in [k] . \label{subeq:kkt-complement}
\end{align}
\end{subequations}
\end{theorem}

\begin{proof}
Notice that \eqref{subeq:kkt-h} and \eqref{subeq:kkt-g} hold since $x \in \Omega$. 
We let $\mu_{j} = 0$ for all $j \in I(x)$. Then \eqref{subeq:kkt-complement} holds since $g_{j}(x) = 0$ for all $j \in A(x)$.

Since $g_{j}$'s are $C^{1}$ and $g_{j}(x) < 0$ for all $j \in I(x)$, we know there exists $r>0$ such that $g_{j}(y) < 0$ for all $y \in B_{r}(x)$ and $j \in I(x)$.
We denote
\begin{equation*}
\Omega' := \{ y \in \Omega \cap B_{r}(x) : h(y) = 0, \ g_{j}(y) = 0, \ \forall\, j \in A(x) \}.
\end{equation*}
Since $x \in \Omega$ is a regular point and local minimizer of $f$, we know $x$ is also a regular point and local minimizer of $f$ in $\Omega'$ (because $g, h \in C^{1}$). Notice that $\Omega'$ only contains equality constraints. Therefore, we apply Theorem \ref{thm:lagrange-eq-constraint} and know that there exist $\lambda \in \mathbb{R}^{m}$ and $\mu_{j} \in \mathbb{R}$ for each $j\in A(x)$ such that \eqref{subeq:kkt-d-lagrange} holds (recall that $\mu_{l} = 0$ for all $l \in I(x)$). 

Now it remains to show that $\mu_{j} \ge 0$ for all $j \in A(x)$. We use a proof of contradiction: if not, then there exists $j \in A(x)$ such that $\mu_{j} < 0$. Define
\begin{equation}
\label{eq:kkt-pf-Omega-hat}
\hat{\Omega} := \{ x \in \mathbb{R}^{n} : h(x) = 0, \  g_{j'}(x) = 0, \ \forall \, j' \in A(x) \setminus \{j\} \} 
\end{equation}
and 
\begin{equation}
\label{eq:kkt-pf-T-hat}
\hat{T}(x) := \{ v \in \mathbb{R}^{n} : \nabla h(x) v = 0, \ \nabla g_{j'}(x) v = 0, \ \forall \, j' \in A(x) \setminus \{j\} \} .
\end{equation}
Notice that $\hat{T}(x)$ is the tangent space of $\hat{\Omega}$ at $x$ and also the orthogonal complement of 
\begin{equation}
\label{eq:kkt-pf-span}
\text{span}\{\nabla h_{i} (x), \ \nabla g_{j'}(x) \  : \  i \in [m], \  j' \in A(x) \setminus \{j\} \} .
\end{equation}
Since $x$ is a regular point, we know $\nabla g_{j}(x)$ is linearly independent of those vectors in the span \eqref{eq:kkt-pf-span} and hence $ \nabla g_{j}(x) \in \hat{T}(x)$. Hence there exists $v \in \hat{T}(x)$ (e.g., $v = - \nabla g_{j} (x)^{\top}$) such that $\nabla g_{j}(x) v < 0$.

We multiply $v$ on both sides of \eqref{subeq:kkt-d-lagrange} and obtain
\begin{equation*}
0 = \nabla f(x) v + \lambda \nabla h(x) v + \mu \nabla g(x) v = \nabla f(x) v + \mu_{j} \nabla g_{j}(x) v,
\end{equation*}
where the second equality is due to $v \in \hat{T}(x)$ defined in \eqref{eq:kkt-pf-T-hat} and $\mu_{l} =  0$ for all $l \in I(x)$. Since $\mu_{j} < 0$ and $\nabla g_{j}(x) v < 0$, we have $\nabla f(x) v < 0$.

As $x$ is a regular point in $\hat{\Omega}$ and $\hat{T}(x)$ is the tangent space of $\hat{\Omega}$ at $x$, we know by Theorem \ref{thm:lagrange-eq-constraint} that there exist $\delta > 0$ and a curve $c: (-\delta, \delta) \to \hat{\Omega}$ such that $c(0) = x$ and $c'(0) = v$. 
Moreover, 
\begin{equation*}
\frac{d}{dt} f(c(t)) \Big|_{t = 0} = \nabla f(c(0)) c'(0) = \nabla f(x) v < 0
\end{equation*}
and
\begin{equation*}
\frac{d}{dt} g_{j}(c(t)) \Big|_{t = 0} = \nabla g_{j}(c(0)) c'(0) = \nabla g_{j} (x) v < 0.
\end{equation*}
Hence, there exists $\delta' \in (0,\delta)$ sufficiently small, such that $c(t) \in B_{r}(x)$, $f(c(t)) < f(c(0)) = f(x)$, $g_{j} (c(t)) < g_{j}(c(0)) = g_{j}(x)=0$ for all $t \in (0, \delta')$. Meanwhile, there are $g_{j'} (c(t))  = 0 $ for every $j' \in A(x) \setminus \{j\}$ (because $c(t) \in \hat{\Omega}$) and $g_{l}(c(t)) < 0$ for every $l \in I(x)$ (because $c(t) \in B_{r}(x)$) for all $t \in (0, \delta')$. Hence $\{c(t) : t \in (0,\delta') \} \subset \Omega$, which implies that these points on the curve $c$ are feasible, and they attain smaller values of $f$. This is a contradiction to $x$ being a local minimizer of $f$ in $\hat{\Omega}$. Therefore, $\mu_{j} \ge 0$ for all $j \in A(x)$, and the proof is completed. 
\end{proof}

To memorize the result of the Theorem \ref{thm:kkt}, we can also define the Lagrangian equation of $(x,\lambda, \mu)$ as
\begin{equation*}
L(x, \lambda, \mu) = f(x) + \lambda h(x) + \mu g(x) .
\end{equation*}
Then Theorem \ref{thm:kkt} implies that, if $x \in \Omega$ is a regular point and a local minimizer of the constrained optimization problem \eqref{eq:min-eq-ineq-constraints} with $f \in C^{1}(\mathbb{R}^{n}; \mathbb{R})$, $g \in C^{1}(\mathbb{R}^{n}; \mathbb{R}^{k})$, and $h \in C^{1}(\mathbb{R}^{n}; \mathbb{R}^{m})$, then there exist a Lagrange multiplier $\lambda \in \mathbb{R}^{m}$ and a \emph{KKT multiplier}\index{Karush--Kuhn--Tucker!KKT multiplier} $\mu \in \mathbb{R}^{k}$ which must be nonnegative, such that $\partial_{x} L(x, \lambda, \mu) = 0$ and $\mu g(x) = 0$ (called the \emph{complementary slackness} condition). Note that, provided \eqref{subeq:kkt-g} and \eqref{subeq:kkt-mu}, we know $\mu g(x) = 0$ and \eqref{subeq:kkt-complement} are equivalent.
The set \eqref{eq:kkt} of equalities and inequalities is called the \emph{KKT conditions}\index{Karush--Kuhn--Tucker!KKT conditions}.
A point $x$ satisfying the KKT conditions is called a \emph{KKT point}\index{Karush--Kuhn--Tucker!KKT point}.

If $f, g, h \in C^{2}$, we can also establish the second-order necessary condition\index{Necessary condition!second-order} for \eqref{eq:min-eq-ineq-constraints}, as given below.

\begin{theorem}
\label{thm:sonc-eq-ineq-constraint}
Suppose there are $f \in C^{2}(\mathbb{R}^{n}; \mathbb{R})$, $g \in C^{2}(\mathbb{R}^{n}; \mathbb{R}^{k})$ and $h \in C^{2} ( \mathbb{R}^{n} ; \mathbb{R}^{m} )$ with $k+m\le n$. If $x \in \Omega = \{ x \in \mathbb{R}^{n} : g(x) \le 0, \ h(x) = 0 \}$ is a regular point and a local minimizer\index{Minimizer!local} of \eqref{eq:min-eq-ineq-constraints}, then there exist a Lagrange multiplier $\lambda \in \mathbb{R}^{m}$ and a KKT multiplier $\mu \in \mathbb{R}^{k}$ such that 
\begin{enumerate}
\item
The KKT conditions \eqref{eq:kkt} hold; and
\item
For any $v \in T(x)$, there is
\begin{equation}
\label{eq:sonc-eq-ineq-constraint}
v^{\top} \Big( \nabla^{2} f(x) + \sum_{i=1}^{m} \lambda_{i} \nabla^{2} h(x) + \sum_{j=1}^{k} \mu_{j} \nabla^{2} g_{j}(x) \Big) v \ge 0 ,
\end{equation}
where
\begin{equation*}
T(x) = \{ v \in \mathbb{R}^{n}: \nabla h(x) v = 0, \ \nabla g_{j}(x) v = 0, \ \forall\, j \in A(x) \}.
\end{equation*}
\end{enumerate}
\end{theorem}

\begin{proof}
The first claim is just Theorem \ref{thm:lagrange-eq-constraint}. For the second claim, we follow the proof of Theorem \ref{thm:sonc-eq-constraint} to obtain the curve $c: (-\delta, \delta) \to \Omega'=\{x \in \mathbb{R}^{n}: h(x) = 0, \ g_{j}(x) = 0, \ \forall\, j \in A(x) \}$, such that $c(0) = x$ and $c'(0) = v$. The rest of the proof is very similar to the one for Theorem \ref{thm:sonc-eq-constraint} and hence omitted here.
\end{proof}

\subsection{Sufficient Optimality Conditions}
\label{subsec:sosc}

In certain cases, we also need to determine whether a given point is a local minimizer or not. This requires a sufficient condition of local minimizers. In this subsection, we derive such condition by using the second-order derivatives of the objective function. Note that, it is in general not possible to have sufficient condition solely using first-order derivatives.

\begin{lemma}
\label{lem:sosc-lemma}
Suppose $P$ and $Q$ are two symmetric matrices, where $Q$ is positive semi-definite and $P$ is positive definite on the null space of $Q$, i.e., $x^{\top} P x > 0$ for any nonzero $x$ with $x^{\top} Q x = 0$, then there exists $\bar{\gamma} > 0$ such that $ P + \frac{1}{\gamma} Q \succ 0 $ for any $ \gamma \in (0, \bar{\gamma})$. 
\end{lemma}

\begin{proof}
We use a proof by contradiction. 
If the claim does not hold, then for any $k \in \mathbb{N}$, choose $\gamma_{k} = 1/k$ and then there exists $x_{k}$ with $|x_{k}| =1$ such that 
\begin{equation}
\label{eq:sosc-lem-contradiction}
x_{k}^{\top} \Big( P + \frac{1}{\gamma_{k}} Q \Big) x_{k} = x_{k}^{\top} (P + k Q ) x_{k} \le  0
\end{equation}
Since $\{x_{k} \}$ is bounded, there exists a subsequence $\{x_{k_{j}}\}$ of $\{x_{k}\}$ and $\bar{x}$ such that $x_{k_{j}} \to \bar{x}$ as $j \to \infty$. 

Note that $|\bar{x}| = 1$, i.e., $\bar{x} \ne 0$. Taking limit of \eqref{eq:sosc-lem-contradiction} for $k= k_{j}$ and $j \to \infty$, then we have
\begin{equation*}
\bar{x}^{\top} P \bar{x} + \lim_{j \to \infty} k_{j} x_{k_{j}}^{\top} Q x_{k_{j}} \le 0 .
\end{equation*}
Since $Q$ is positive semi-definite, we know $x_{k_{j}}^{\top} Q x_{k_{j}} \ge 0$ for all $j$. This implies that $\bar{x}^{\top} P \bar{x} \le 0$.

Notice that $\bar{x}^{\top} P \bar{x}$ is a fixed number and $k_{j} \to \infty$ as $j \to \infty$. Therefore, we must have
\begin{equation*}
\bar{x}^{\top} Q \bar{x} = \lim_{j\to\infty} x_{k_{j}}^{\top} Q x_{k_{j}} = 0 ,
\end{equation*}
% otherwise the right-hand side of \eqref{eq:sosc-lem-contradiction} tends to $\infty$. 
and thus $\bar{x}^{\top} P \bar{x} > 0$ because $P$ is positive definite on the null space of $Q$. This is a contradiction to $\bar{x}^{\top} P \bar{x} \le 0$ above.
\end{proof}

Now we are ready to give the sufficient condition of local minimizers\index{Minimizer!local}. We first consider the case with equality constrained optimization\index{Constrained optimization!equality constrained} \eqref{eq:min-eq-constraint}.

\begin{proposition}
[Second-order sufficient condition]
\label{prop:sosc}
Suppose $f, h \in C^{2}$, $x^{*} \in \mathbb{R}^{n}$ and $\lambda^{*} \in \mathbb{R}^{m}$ satisfy the Lagrange condition and
\begin{equation}
\label{eq:sosc}
y^{\top} \Big( \nabla^{2} f(\xs) + \sum_{i=1}^{m} \lambda_{i}^{*} \nabla^{2} h_{i}(\xs) \Big) y > 0
\end{equation}
for all nonzero $y \in \mathbb{R}^{n}$ with $\nabla h(x^{*}) y = 0$. Then $x^{*}$ is a strict local minimizer of $f$ with constraint $h(x) = 0$.
In fact, there exists $\alpha > 0$ and $r > 0$, such that 
\begin{equation*}
f(x) \ge f(x^{*}) + \frac{\alpha}{2} |x - x^{*}|^{2}
\end{equation*}
for all $x \in B_{r}(x^{*})$ satisfying $h(x) = 0$.\index{Sufficient condition!second-order}
\end{proposition}

\begin{proof}
Let $\gamma > 0$. We define the augmented Lagrangian function $l_{A}$ by
\begin{equation}
l_{A}(x, \lambda; \gamma) := f(x) + \lambda h(x) + \frac{1}{2 \gamma} |h(x)|^{2} ,
\end{equation}
Notice that $l_{A}$ is also the Lagrangian function of 
\begin{subequations}
\label{eq:min-eq-constraint-aug}
\begin{align}
\min_{x \in \mathbb{R}^{n}} \quad & f(x) + \frac{1}{2\gamma} |h(x)|^{2}, \label{subeq:min-eq-constraint-aug-f} \\
\text{s.t.} \quad & h(x) = 0 . \label{subeq:min-eq-constraint-aug-h}
\end{align}
\end{subequations}
We can see that \eqref{eq:min-eq-constraint-aug} has the same local minimizer as \eqref{eq:min-eq-constraint}. 
Moreover, we have
\begin{equation*}
\nabla_{x} l_{A} (x, \lambda; \gamma) = \nabla f(x) + \lambda \nabla h(x) + \frac{1}{\gamma} h(x)^{\top} \nabla h(x)
\end{equation*}
and
\begin{equation*}
\nabla_{xx}^{2} l_{A}(x, \lambda; \gamma) = \nabla^{2} f(x) + \sum_{i=1}^{m}\Big(\lambda_{i} + \frac{h_{i}(x)}{\gamma} \Big) \nabla^{2} h_{i}(x) + \frac{1}{\gamma} \nabla h(x)^{\top} \nabla h(x).
\end{equation*}
Since $(x^{*}, \lambda^{*})$ satisfies the Lagrange conditions, namely,
\begin{align*}
\nabla f(\xs) + \lambda^{*} \nabla h(\xs) & = 0 ,\\
h(\xs) & = 0 ,
\end{align*}
we have 
\begin{equation*}
\nabla_{x} l_{A} (x^{*}, \lambda^{*}; \gamma) = \nabla f(x^{*}) + \lambda^{*} \nabla h(x^{*}) + \frac{1}{\gamma} h(x^{*})^{\top} \nabla h(x^{*}) = 0
\end{equation*}
and
\begin{equation*}
\nabla_{xx}^{2} l_{A}(x^{*}, \lambda^{*}; \gamma) = \nabla^{2} f(\xs) + \sum_{i=1}^{m} \lambda_{i}^{*} \nabla^{2} h_{i}(\xs) + \frac{1}{\gamma} \nabla h(x^{*})^{\top} \nabla h(x^{*}) .
\end{equation*}

Notice that, for any $y$ in the null space of $\nabla h(\xs)^{\top} \nabla h(\xs)$, or equivalently, $\nabla h(x^{*}) y = 0$, the inequality \eqref{eq:sosc} holds.
Therefore, by Lemma \ref{lem:sosc-lemma}, there exists $\bar{\gamma} > 0$ such that $\nabla_{xx}^{2}l_{A}(x^{*}, \lambda^{*}; \gamma) \succ 0$ for all $\gamma \in (0, \bar{\gamma})$. 

Now we consider $l_{A}(\cdot, \lambda^{*}; \gamma)$ with both $\gamma \in (0, \bar{\gamma})$ and $\lambda^{*}$ fixed. Then $l_{A}(\cdot, \lambda^{*}; \gamma): \mathbb{R}^{n} \to \mathbb{R}$ is a function of $x$.
From previous derivations, we know $\nabla_{x} l_{A}(\cdot, \lambda^{*}; \gamma) = 0$ and $\nabla_{xx}^{2} l_{A}(\cdot, \lambda^{*}; \gamma) \succ 0$. 
Therefore, $\xs$ is a strict local minimizer of $l_{A}(\cdot, \lambda^{*}; \gamma)$.  
Moreover,
\begin{align*}
l_{A}(x, \lambda^{*}; \gamma) 
& = l_{A}(x^{*}, \lambda^{*}; \gamma) + \nabla_{x} l_{A}(x^{*}, \lambda^{*}; \gamma) \\
& \qquad + \frac{1}{2}(x-x^{*})^{\top} \nabla_{xx}^{2}l_{A}(x^{*}, \lambda^{*}; \gamma) (x-x^{*}) + o(|x-\xs|^{2}) \\
& = f(x^{*}) + \frac{1}{2}(x-x^{*})^{\top} \nabla_{xx}^{2}l_{A}(x^{*}, \lambda^{*}; \gamma) (x-x^{*}) + o(|x - x^{*}|^{2}) .
\end{align*}
As $l_{A}(x^{*}, \lambda^{*}; \gamma) \in C^{2}$, we know there exist $\alpha > 0$ and $r> 0$ such that 
\begin{equation*}
l_{A}(x, \lambda^{*};\gamma) \ge f(x^{*}) + \frac{\alpha}{2} | x - x^{*} |^{2}
\end{equation*}
for all $x \in B_{r}(x^{*})$.
If in addition $h(x) = 0$, we have $l_{A}(x, \lambda^{*}; \gamma) = f(x)$ and the inequality above reduces to $f(x) \ge f(\xs) + \frac{\alpha}{2}|x - \xs|^{2}$ for all $x \in B_{r}(x^{*})$ with $h(x) = 0$.
\end{proof}

As we can see, the second-order sufficient condition (Proposition \ref{prop:sosc}) is a characterization of strict local minimizer\index{Minimizer!local} $\xs$. Therefore, we can easily extend it to the optimization problem with both equality and inequality constraints \eqref{eq:min-eq-ineq-constraints}: We can simply ignore all inactive inequality constraints at $\xs$ as the strict inequalities hold at $\xs$, and notice that $\xs$ yields equalities in all the other constraints. Therefore, we can directly make a conclusion as Proposition \ref{prop:sosc} by including the active inequality constraints (they all become equality constraints at $\xs$ now). We omit the details here.

\section{Automatic Differentiation}
\label{sec:autodiff}

In most optimization algorithms, we need to evaluate the first- and second-order derivatives of the objective function. A simple approach is to use finite difference. 
More precisely, let $f: \mathbb{R}^{n} \to \mathbb{R}$, then we can approximate the partial derivative $\frac{\partial f(x)}{\partial x_{i}}$ as
\begin{equation}
\label{eq:dfdx-finite-diff}
\frac{\partial f(x)}{\partial x_{i}}  \approx \frac{f(x + \epsilon e_{i}) - f(x)}{\epsilon} ,
\end{equation}
where $e_{i} \in \mathbb{R}^{n}$ has $1$ as its $i$th component and $0$ elsewhere, and $\epsilon>0$ is a user-chosen step size.
However, the approximation \eqref{eq:dfdx-finite-diff} is often inaccurate for large $\epsilon$ while also suffering floating error severely for small $\epsilon$.
Hence such approximation is not suitable for large-scale optimization problems like network training.

As an alternative, back-propagation\index{Back-propagation} is a much more elegant approach to evaluating first- and second-order derivatives of functions. Back-propagation is a particular case of the general class of derivative evaluation method called automatic differentiation, which we discuss in detail in this section.

Automatic differentiation\index{Automatic differentiation} is a technique to compute analytic values of derivatives based on the representation of a function and the chain rule\index{Chain rule}.
The foundation of automatic differentiation is that any function is evaluated by performing a series of compositions of elementary operations. By elementary operation, we meant the following type of functions which take only one or two scalar-valued arguments:
\begin{itemize}
\item
Single-argument operations: $\sin$, $\cos$, $\tan$, $\exp$, $\log$, and etc.
\item
Two-argument operations: $+$, $-$, $\times$, $/$, and etc.
\end{itemize}
For any function $f$, we can construct a \emph{computational graph}\index{Automatic differentiation!computational graph} to represent $f$ using such elementary operations, where the nodes represent the scalar-valued arguments and results of these operations, and the edges indicate the relations between the nodes based on the operation executed.

The following example demonstrates how a function can be decomposed into such elementary operations on its arguments and the corresponding computational graph is constructed.

\begin{figure}
\centering
\begin{tikzpicture}[scale=1, transform shape]
	\Vertex[x=0, y=1.5, label=$x_{1}$, color=none, size=0.6]{x1}
	\Vertex[x=0, y=0, label=$x_{2}$, color=none, size=0.6]{x2}
	\Vertex[x=0, y=-1.5, label=$x_{3}$, color=none, size=0.6]{x3}
	
	\Vertex[x=1.5, y=1.5, label=$x_{4}$, color=none, size=0.6]{x4}
	\Vertex[x=1.5, y=0, label=$x_{5}$, color=none, size=0.6]{x5}
	\Vertex[x=1.5, y=-1.5, label=$x_{6}$, color=none, size=0.6]{x6}
	
	\Vertex[x=3, y=0, label=$x_{7}$, color=none, size=0.6]{x7}
	\Vertex[x=3, y=-1.5, label=$x_{8}$, color=none, size=0.6]{x8}
	\Vertex[x=4.5, y=0, label=$x_{9}$, color=none, size=0.6]{x9}

	\Vertex[x=6, y=0, label=$x_{10}$, color=none, size=0.6]{x10}
	
	\Edge[Direct, lw=1pt](x1)(x4)
	\Edge[Direct, lw=1pt](x2)(x4)
	\Edge[Direct, lw=1pt](x2)(x5)
	\Edge[Direct, lw=1pt](x2)(x6)
	\Edge[Direct, lw=1pt](x3)(x6)
	\Edge[Direct, lw=1pt](x5)(x7)
	\Edge[Direct, lw=1pt](x6)(x8)
	\Edge[Direct, lw=1pt](x7)(x9)
	\Edge[Direct, lw=1pt](x8)(x9)
	\Edge[Direct, lw=1pt](x9)(x10)
	\Edge[Direct, bend=19, lw=1pt](x4)(x10)
	
	\draw[black] (1.6, 1.6) node[above right]{\tiny $+$};
	\draw[black] (1.6, 0.1) node[above right]{\tiny $2\times$};
	\draw[black] (1.6, -1.4) node[above right]{\tiny $\times$};
	\draw[black] (3.1, 0.1) node[above right]{\tiny $\exp$};
	\draw[black] (3.2, -1.5) node[above right]{\tiny $\cos$};
	\draw[black] (4.6, 0.1) node[above right]{\tiny $\times$};
	\draw[black] (6.1, 0.1) node[above right]{\tiny $/$};
	
\end{tikzpicture}
\caption{The computational graph of the function $f: \mathbb{R}^{3} \to \mathbb{R}$ defined in \eqref{eq:ad-example-fn}. The argument variables of $f$ are represented by the nodes $x_{1}$, $x_{2}$, and $x_{3}$. The other nodes represent the intermediate scalar-valued variables. In particular, $x_{10} = f(x_{1},x_{2},x_{3})$. The elementary operation executed at each node is shown at the upper-right corner of the corresponding node.}
\label{fig:comp-graph}
\end{figure}
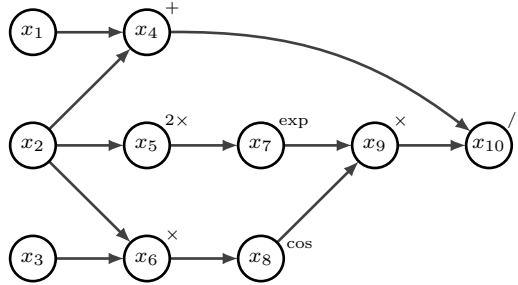

\begin{example}
\label{ex:ad-example}
Consider the function $f: \mathbb{R}^{3} \to \mathbb{R}$ as follows:
\begin{equation}
\label{eq:ad-example-fn}
f(x) = \frac{e^{2x_{2}} \cos(x_{2} x_{3})}{x_{1} + x_{2}}
\end{equation}
for $x = (x_{1}, x_{2}, x_{3}) \in \mathbb{R}^{3}$. 
In addition to the argument nodes $x_{1}$, $x_{2}$, $x_{3}$, the computational graph of $f$ also introduces new nodes $x_{4}, \dots, x_{10}$ to represent the intermediate variables:
\begin{align}
x_{4} & = x_{1} + x_{2} , \nonumber \\
x_{5} & = 2 \times x_{2} , \nonumber \\
x_{6} & = x_{2} \times x_{3} , \nonumber \\
x_{7} & = e^{x_{5}} =e^{2x_{2}} , \label{eq:ad-example-nodes} \\
x_{8} & = \cos(x_{6}) = \cos(x_{2} x_{3}) , \nonumber \\
x_{9} & = x_{7} \times x_{8} =e^{2x_{2}} \cos(x_{2} x_{3}), \nonumber \\
x_{10} & = x_{9} / x_{4} = e^{2x_{2}} \cos(x_{2} x_{3}) / (x_{1} + x_{2}) . \nonumber
\end{align}
With slight abuse of notation, we use $x_{i}$ to denote the node as well as the value at this node hereafter.
The computational graph of $f$ in \eqref{eq:ad-example-fn} is illustrated in Figure \ref{fig:comp-graph}. A directed edge from $x_{i}$ to $x_{j}$ will be denoted by $(x_{i}, x_{j})$ later in this section. In such case, we call $x_{i}$ a \emph{parent node} of $x_{j}$, and $x_{j}$ a \emph{child node} of $x_{i}$.

As we can see, each of the nodes in \eqref{eq:ad-example-nodes} is computed by some elementary operations applied to either one or two of its parent nodes. Hence, each of them has one or two incoming edges.
In Figure \ref{fig:comp-graph}, the operation executed at each node is labeled at the upper right corner of the corresponding node. Note that the graph must have the operations and connections ordered correctly according to $f$.
We also remark that all of the nodes in the computational graph are indeed functions of $(x_{1}, x_{2}, x_{3})$.

In the computational graph, once the values of all the parent nodes of $x_{i}$ are known, the value of $x_{i}$ can be computed. If we have the values of the input variables $x=(x_{1},\dots,x_{n})$, we can compute the values of all nodes in the computational graph. The process of such ordered computations is called a \emph{forward sweep}.
\end{example}

In what follows, we show how automatic differentiation computes first-order derivatives (such as gradient and Jacobian) in Section \ref{subsec:ad1} and second-order derivatives (such as Hessian) in Section \ref{subsec:ad2}. In either case, there are two modes of automatic differentiation, called the \emph{forward mode} and \emph{reverse mode}, as we will discuss in detail below.

\subsection{Computation of First-order Derivatives}
\label{subsec:ad1}

In this subsection, we consider the computation of the first-order derivatives of a function using automatic differentiation. 
We first present the forward mode which is more straightforward and easier to understand, and then the reverse mode which is often commonly used and more adaptive to deep learning tasks.
We also discuss the computation complexities and storage requirements in both cases.

\subsubsection*{Forward Mode to Compute Gradient and Jacobian}

The forward mode\index{Automatic differentiation!forward mode} of automatic differentiation can be used to compute the directional derivative $D_{v}f(x)$ if $f: \mathbb{R}^{n} \to \mathbb{R}$ is scalar-valued, or the Jacobian-vector product $\nabla f(x) v$ if $f:\mathbb{R}^{n} \to \mathbb{R}^{m}$ is vector-valued, given any $x$ in the domain of $f$ and (column) vector $v \in \mathbb{R}^{n}$. As a consequence, one can obtain the gradient (or Jacobian) $\nabla f$ by setting $v = e_{i}$ for $i\in [n]$, where $e_{i}$ is an $n$-dimensional vector taking value $1$ as its $i$th component and $0$ as all the other components.

We first consider the case when $f: \mathbb{R}^{n} \to \mathbb{R}$ is scalar-valued. In addition to the input variables $x=(x_{1}, \dots, x_{n})$ of $f$, the computational graph will introduce intermediate variables $x_{n+1}, \dots, x_{N}$, which are all functions of $x_{1}, \dots, x_{n}$. For instance, $N = 10$ in Example \ref{ex:ad-example} and Figure \ref{fig:comp-graph}.
%
% Moreover, we will also compute $D_{v} x_{j} = \langle \nabla x_{j}, v \rangle \in \mathbb{R}$. 
%
%
%
\begin{figure}
\centering
\begin{tikzpicture}[scale=1, transform shape]
	\Vertex[x=0, y=1.5, label=$x_{1}$, color=none, size=0.6]{x1}
	\Vertex[x=0, y=0, label=$x_{2}$, color=none, size=0.6]{x2}
	\Vertex[x=0, y=-1.5, label=$x_{3}$, color=none, size=0.6]{x3}
	
	\Vertex[x=1.5, y=1.5, label=$x_{4}$, color=none, size=0.6]{x4}
	\Vertex[x=1.5, y=0, label=$x_{5}$, color=none, size=0.6]{x5}
	\Vertex[x=1.5, y=-1.5, label=$x_{6}$, color=none, size=0.6]{x6}
	
	\Vertex[x=3, y=0, label=$x_{7}$, color=none, size=0.6]{x7}
	\Vertex[x=3, y=-1.5, label=$x_{8}$, color=none, size=0.6]{x8}
	\Vertex[x=4.5, y=0, label=$x_{9}$, color=none, size=0.6]{x9}

	\Vertex[x=6, y=0, label=$x_{10}$, color=none, size=0.6]{x10}
	
	\Edge[Direct, lw=1pt, position={below=0.5mm}, fontscale=0.7, label=$1$](x1)(x4)
	\Edge[Direct, lw=1pt, position={below=0.5mm}, fontscale=0.7, label=$1$](x2)(x4)
	\Edge[Direct, lw=1pt, position={below=0.5mm}, fontscale=0.7, label=$2$](x2)(x5)
	\Edge[Direct, lw=1pt, position={below=0.5mm}, fontscale=0.7, label=$\pi$](x2)(x6)
	\Edge[Direct, lw=1pt, position={below=0.5mm}, fontscale=0.7, label=$1$](x3)(x6)
	\Edge[Direct, lw=1pt, position={below=0.5mm}, fontscale=0.7, label=$e^{2}$](x5)(x7)
	\Edge[Direct, lw=1pt, position={below=0.5mm}, fontscale=0.7, label=$0$](x6)(x8)
	\Edge[Direct, lw=1pt, position={below=0.5mm}, fontscale=0.7, label=$-1$](x7)(x9)
	\Edge[Direct, lw=1pt, position={below=0.5mm}, fontscale=0.7, label=$e^{2}$](x8)(x9)
	\Edge[Direct, lw=1pt, position={below=0.5mm}, fontscale=0.7, label=$1$](x9)(x10)
	\Edge[Direct, bend=19, lw=1pt, position={below=0.5mm}, fontscale=0.7, label=$e^{2}$](x4)(x10)
	
	\draw[black] (1.6, 1.6) node[above right]{\tiny $+$};
	\draw[black] (1.6, 0.1) node[above right]{\tiny $2\times$};
	\draw[black] (1.6, -1.4) node[above right]{\tiny $\times$};
	\draw[black] (3.1, 0.1) node[above right]{\tiny $\exp$};
	\draw[black] (3.2, -1.5) node[above right]{\tiny $\cos$};
	\draw[black] (4.6, 0.1) node[above right]{\tiny $\times$};
	\draw[black] (6.1, 0.1) node[above right]{\tiny $/$};
	
	\draw[black] (0, 1.75) node[above]{\tiny $0$};
	\draw[black] (0, 0.25) node[above]{\tiny $1$};
	\draw[black] (0, -1.25) node[above]{\tiny $\pi$};
	\draw[black] (1.5, 1.75) node[above]{\tiny $1$};
	\draw[black] (1.5, 0.25) node[above]{\tiny $2$};
	\draw[black] (1.5, -1.25) node[above]{\tiny $\pi$};
	\draw[black] (3, 0.25) node[above]{\tiny $e^{2}$};
	\draw[black] (3, -1.25) node[above]{\tiny $-1$};
	\draw[black] (4.5, 0.25) node[above]{\tiny $-e^{2}$};
	\draw[black] (6, 0.25) node[above]{\tiny $-e^{2}$};
	
	\draw[black] (0, 1.25) node[below]{\tiny $2$};
	\draw[black] (0, -0.25) node[below]{\tiny $1$};
	\draw[black] (0, -1.75) node[below]{\tiny $0$};
	\draw[black] (1.5, 1.25) node[below]{\tiny $3$};
	\draw[black] (1.5, -0.25) node[below]{\tiny $2$};
	\draw[black] (1.5, -1.75) node[below]{\tiny $\pi$};
	\draw[black] (3, -0.25) node[below]{\tiny $2e^{2}$};
	\draw[black] (3, -1.75) node[below]{\tiny $0$};
	\draw[black] (4.5, -0.25) node[below]{\tiny $-2e^{2}$};
	\draw[black] (6, -0.25) node[below]{\tiny $e^{2}$};
	
\end{tikzpicture}
\caption{The forward mode of automatic differentiation to compute $f(x)$ and $D_{v}f(x)$ for the function given in \eqref{eq:ad-example-fn} with $x = (0,1,\pi)$ and $v = (2,1,0)$. The values of $x_{j}$ and $D_{v}x_{j}$ are shown above and below the corresponding node $x_{j}$. The elementary operation executed at $x_{j}$ is shown at the upper-right corner of $x_{j}$. The value of $\frac{\partial x_{j}}{\partial x_{i}}$ is shown below the directed edge $(x_{i}, x_{j})$. At the end node $x_{10}$, we obtain $f(x) = -e^{2}$ and $D_{v}f(x) = e^{2}$. }
\label{fig:comp-graph-forward}
\end{figure}
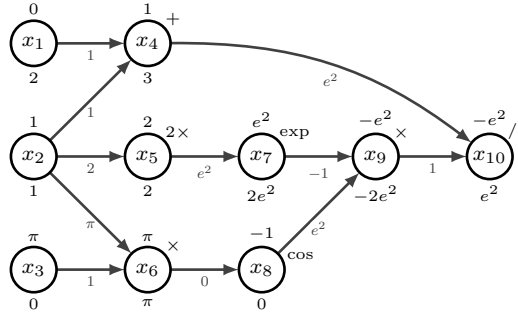

Now we proceed the elementary operations as shown in the computational graph in Figure \ref{fig:comp-graph} in the forward sweep\index{Automatic differentiation!forward sweep} (from left to right in Figure \ref{fig:comp-graph}). In addition to the value $x_{j}$, we also compute the value $D_{v}x_{j}$ by applying the chain rule:
\begin{equation}
\label{eq:forward-mode-Dvx}
D_{v}x_{j} = \sum_{x_{i} \in P_{j}} \frac{\partial x_{j}}{\partial x_{i}} D_{v} x_{i} ,
\end{equation}
where $P_{j}$ is the set of parent nodes of $x_{j}$. Note that the sum in \eqref{eq:forward-mode-Dvx} contains only one or two terms. We begin the computation by assigning input values to $x_{1},\dots,x_{n}$, and noticing that $D_{v}x_{i} = \langle \nabla x_{i}, v \rangle = \langle e_{i}, v \rangle = v_{i}$ for every $i \in [n]$.
Then we continue to evaluate their child nodes as well as the directional derivatives at these child nodes. When we finish the forward sweep, we will obtain both $f(x) = x_{N}$ and $D_{v}f(x) = D_{v} x_{N}$ at the last node $x_{N}$.
Figure \ref{fig:comp-graph-forward} shows the values of the intermediate variables in the case of $f$ given in \eqref{eq:ad-example-fn} with input $x = (x_{1},x_{2},x_{3}) = (0,1,\pi)$ and $v = (2,1,0)$. 
In Figure \ref{fig:comp-graph-forward}, the values of $x_{j}$ and $D_{v}x_{j}$ are shown on above and below the node $x_{j}$. The value of the partial derivative $\frac{\partial x_{j}}{\partial x_{i}}$ is shown under the directed edge $(x_{i}, x_{j})$.
The elementary operation executed at each intermediate variable $x_{j}$ is shown at its upper-right corner for reference.
For example, given that $(x_{7}, D_{v}x_{7}) = (e^{2}, 2e^{2})$ (these two values are shown above and below the node $x_{7}$) and $(x_{8}, D_{v}x_{8}) = (-1,0)$, we obtain 
\begin{equation*}
x_{9} = x_{7} \cdot x_{8} = e^{2} \cdot (-1) = -e^{2}
\end{equation*}
and
\begin{equation*}
D_{v}x_{9} = \frac{\partial x_{9}}{\partial x_{7}} D_{v} x_{7} + \frac{\partial x_{9}}{\partial x_{8}} D_{v}x_{8} = (-1) \cdot (2e^{2}) + e^{2} \cdot 0 = -2e^{2}.
\end{equation*}
With the values of $x$ and $v$ given above, we will eventually obtain $x_{10} = -e^{2}$ and $D_{v}x_{10} = e^{2}$. 
Manual calculations show the same result with $\nabla f(x) = (e^{2}, -e^{2}, 0)$, which verifies $D_{v}f(x) = \langle \nabla f(x) ,v \rangle = e^{2}$.

If $f = (f_{1},\dots, f_{m}) : \mathbb{R}^{n} \to \mathbb{R}^{m}$ is a vector-valued function, then the computational graph will be similar as in Figure \ref{fig:comp-graph} except that there are $m$ end nodes instead of one. We will obtain the values $(D_{v}f_{1}(x)$, $\cdots$, $D_{v}f_{m}(x))$ at these nodes, which can form the desired Jacobian-vector product
\begin{equation*}
\nabla f(x) v = 
\begin{pmatrix}
D_{v}f_{1}(x) \\ \vdots \\ D_{v} f_{m}(x)
\end{pmatrix} .
\end{equation*}
The computational cost will be $m$ times larger than the cost with a scalar-valued function.

The advantage of the forward mode is that the computer memory storing $(x_{j}, D_{v}x_{j})$ can be released once the computations at all of the child nodes of $x_{j}$ are completed. This can be beneficial in practice since the computational graph is often very large.
The drawback is that, if the full gradient or Jacobian is needed, then we have to set $v$ to each of $e_{i}$'s for $i \in [n]$. The total computational cost will be $n$ times larger than the cost of directional derivative (or Jacobian-vector product) above, which is expensive when $n$ is very large.
We will show next that the reverse mode of automatic differentiation can address this issue.

\subsubsection*{Reverse Mode to Compute Gradient and Jacobian}
We begin by directly computing the gradient $\nabla f(x)$ of a scalar-valued function $f: \mathbb{R}^{n} \to \mathbb{R}$\index{Automatic differentiation!gradient}. (The directional derivative $D_{v} f(x) = \langle \nabla f(x), v \rangle$ can be easily computed by vector inner product afterwards for any given $v$.)
The main difference between the reverse mode and the forward mode is that the reverse mode does not compute anything other than the values of $x_{j}$'s in the forward sweep. 
Instead, once the forward sweep is completed, the reverse mode\index{Automatic differentiation!reverse mode} of automatic differentiation begins to compute the value $y_{j}:= \frac{\partial f(x)}{\partial x_{j}}$ at each node $x_{j}$, starting from the end node $x_{N}$ and going the reverse directions of the directed edges until $y_{i}$ at all the input argument variables in $x = (x_{1},\dots, x_{n})$ are computed.
At that point, the gradient $\nabla f(x) = (y_{1},\dots, y_{n})$ is obtained.
This process is called the \emph{reverse sweep}\index{Automatic differentiation!reverse sweep}, which we explain in details below.

\begin{figure}
\centering
\begin{tikzpicture}[scale=1, transform shape]
	\Vertex[x=0, y=1.5, label=$x_{1}$, color=none, size=0.6]{x1}
	\Vertex[x=0, y=0, label=$x_{2}$, color=none, size=0.6]{x2}
	\Vertex[x=0, y=-1.5, label=$x_{3}$, color=none, size=0.6]{x3}
	
	\Vertex[x=1.5, y=1.5, label=$x_{4}$, color=none, size=0.6]{x4}
	\Vertex[x=1.5, y=0, label=$x_{5}$, color=none, size=0.6]{x5}
	\Vertex[x=1.5, y=-1.5, label=$x_{6}$, color=none, size=0.6]{x6}
	
	\Vertex[x=3, y=0, label=$x_{7}$, color=none, size=0.6]{x7}
	\Vertex[x=3, y=-1.5, label=$x_{8}$, color=none, size=0.6]{x8}
	\Vertex[x=4.5, y=0, label=$x_{9}$, color=none, size=0.6]{x9}

	\Vertex[x=6, y=0, label=$x_{10}$, color=none, size=0.6]{x10}
	
	\Edge[Direct, lw=1pt, position={below=0.5mm}, fontcolor=gray, fontscale=0.7, label=$1$](x1)(x4)
	\Edge[Direct, lw=1pt, position={below=0.5mm}, fontcolor=gray, fontscale=0.7, label=$1$](x2)(x4)
	\Edge[Direct, lw=1pt, position={below=0.5mm}, fontcolor=gray, fontscale=0.7, label=$2$](x2)(x5)
	\Edge[Direct, lw=1pt, position={below=0.5mm}, fontcolor=gray, fontscale=0.7, label=$\pi$](x2)(x6)
	\Edge[Direct, lw=1pt, position={below=0.5mm}, fontcolor=gray, fontscale=0.7, label=$1$](x3)(x6)
	\Edge[Direct, lw=1pt, position={below=0.5mm}, fontcolor=gray, fontscale=0.7, label=$e^{2}$](x5)(x7)
	\Edge[Direct, lw=1pt, position={below=0.5mm}, fontcolor=gray, fontscale=0.7, label=$0$](x6)(x8)
	\Edge[Direct, lw=1pt, position={below=0.5mm}, fontcolor=gray, fontscale=0.7, label=$-1$](x7)(x9)
	\Edge[Direct, lw=1pt, position={below=0.5mm}, fontcolor=gray, fontscale=0.7, label=$e^{2}$](x8)(x9)
	\Edge[Direct, lw=1pt, position={below=0.5mm}, fontcolor=gray, fontscale=0.7, label=$1$](x9)(x10)
	\Edge[Direct, bend=19, lw=1pt, position={below=0.5mm}, fontcolor=gray, fontscale=0.7, label=$e^{2}$](x4)(x10)
	
	\draw[black] (1.6, 1.6) node[above right]{\tiny $+$};
	\draw[black] (1.6, 0.1) node[above right]{\tiny $2\times$};
	\draw[black] (1.6, -1.4) node[above right]{\tiny $\times$};
	\draw[black] (3.1, 0.1) node[above right]{\tiny $\exp$};
	\draw[black] (3.2, -1.5) node[above right]{\tiny $\cos$};
	\draw[black] (4.6, 0.1) node[above right]{\tiny $\times$};
	\draw[black] (6.1, 0.1) node[above right]{\tiny $/$};
	
	\draw[black] (0, 1.75) node[above]{\tiny $0$};
	\draw[black] (0, 0.25) node[above]{\tiny $1$};
	\draw[black] (0, -1.25) node[above]{\tiny $\pi$};
	\draw[black] (1.5, 1.75) node[above]{\tiny $1$};
	\draw[black] (1.5, 0.25) node[above]{\tiny $2$};
	\draw[black] (1.5, -1.25) node[above]{\tiny $\pi$};
	\draw[black] (3, 0.25) node[above]{\tiny $e^{2}$};
	\draw[black] (3, -1.25) node[above]{\tiny $-1$};
	\draw[black] (4.5, 0.25) node[above]{\tiny $-e^{2}$};
	\draw[black] (6, 0.25) node[above]{\tiny $-e^{2}$};
	
	\draw[gray] (0, 1.25) node[below]{\tiny $e^{2}$};
	\draw[gray] (0, -0.25) node[below]{\tiny $-e^{2}$};
	\draw[gray] (0, -1.75) node[below]{\tiny $0$};
	\draw[gray] (1.5, 1.25) node[below]{\tiny $e^{2}$};
	\draw[gray] (1.5, -0.25) node[below]{\tiny $-e^{2}$};
	\draw[gray] (1.5, -1.75) node[below]{\tiny $0$};
	\draw[gray] (3, -0.25) node[below]{\tiny $-1$};
	\draw[gray] (3, -1.75) node[below]{\tiny $e^{2}$};
	\draw[gray] (4.5, -0.25) node[below]{\tiny $1$};
	\draw[gray] (6, -0.25) node[below]{\tiny $1$};
	
\end{tikzpicture}
\caption{The reverse mode of automatic differentiation to compute $\nabla f(x)$ for the function given in \eqref{eq:ad-example-fn} with $x = (0,1,\pi)$. The values of $x_{j}$ (black) and $y_{j}=\frac{\partial f(x)}{\partial x_{j}}$ (gray) are shown above and below the corresponding node $x_{j}$. The elementary operation executed at $x_{j}$ is shown at the upper-right corner of $x_{j}$. The value of $\frac{\partial x_{i}}{\partial x_{j}}$ (gray) is shown below the directed edge $(x_{j}, x_{i})$. The black values are computed in the forward sweep. The gray values are computed in the reverse sweep. The result is $\nabla f(x) = (e^{2}, e^{-2}, 0)$. }
\label{fig:comp-graph-reverse}
\end{figure}
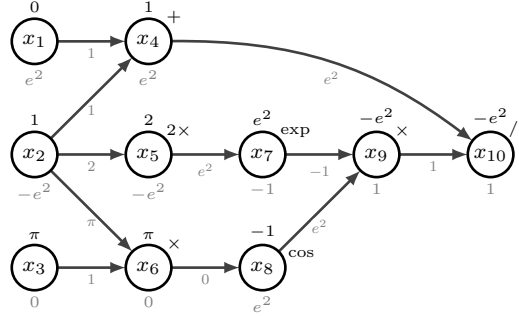

To begin the reverse sweep, we notice that $y_{N} = \frac{\partial f(x)}{\partial x_{N}} = 1$ since $f(x) = x_{N}$.
Then for any $x_{j}$, we can apply the chain rule to obtain
\begin{equation*}
y_{j} = \frac{\partial f(x)}{\partial x_{j}} = \sum_{x_{i} \in C_{j}} \frac{\partial f(x)}{\partial x_{i}} \frac{\partial x_{i}}{\partial x_{j}} = \sum_{x_{i} \in C_{j}} y_{i} \frac{\partial x_{i}}{\partial x_{j}}, 
\end{equation*}
where $C_{j}$ denotes the set of all child nodes of $x_{j}$. Note that $C_{j}$ may contain more than two nodes depending on the definition of $f$, and this is different from the case in \eqref{eq:forward-mode-Dvx}.
Figure \ref{fig:comp-graph-reverse} shows the values of $x_{j}$ (black color, above the node $x_{j})$ and $y_{j}$ (gray color, below the node $x_{j}$) in the case of $f$ given in \eqref{eq:ad-example-fn} with input $x = (x_{1},x_{2},x_{3}) = (0,1,\pi)$ and $v = (2,1,0)$. 
In Figure \ref{fig:comp-graph-reverse}, the values of $x_{j}$'s are computed in the forward sweep; and the values of $y_{j}$ at every node $x_{j}$ and $\frac{\partial x_{i}}{\partial x_{j}}$ at every directed edge $(x_{j},x_{i})$ are computed in the reverse sweep. 
The elementary operation executed at each intermediate variable $x_{j}$ is again shown at its upper-right corner for reference.
For example, given $y_{4}=e^{2}$, $y_{5} = -e^{2}$, and $y_{6} = 0$, we obtain
\begin{equation*}
y_{2} = y_{4} \frac{\partial x_{4}}{\partial x_{2}} + y_{5} \frac{\partial x_{5}}{\partial x_{2}} + y_{6} \frac{\partial x_{6}}{\partial x_{2}} = e^{2} \cdot 1 + (-e^{2}) \cdot 2 + 0 \cdot \pi = -e^{2} .
\end{equation*}
After the reverse sweep is done, we obtain $(y_{1},y_{2},y_{3}) = (e^{2}, -e^{2}, 0)$, matching the gradient $\nabla f(x)$ at $x=(0,1,\pi)$ showed earlier.

If $f = (f_{1},\dots, f_{m}) : \mathbb{R}^{n} \to \mathbb{R}^{m}$ is a vector-valued function and we want to compute the \emph{vector-Jacobian} product $u\nabla f(x) \in \mathbb{R}^{n}$ for some given $u \in \mathbb{R}^{m}$, which are both row vectors\footnote{Note that this is different from the Jacobian-vector product $\nabla f(x) v$ for some given $v \in \mathbb{R}^{n}$, which are both column vectors.}, then we can simply define
\begin{equation*}
\bar{f} := \sum_{i=1}^{m} u_{i} f_{i} \ : \ \mathbb{R}^{n} \to \mathbb{R} ,
\end{equation*}
which is a scalar-valued function, and use the reverse mode of automatic differentiation as above. 
If we want to compute the full Jacobian $\nabla f(x)$, then we need to compute a (column) vector $y_j = (y_{j}^{1},\dots,y_{j}^{m}) \in \mathbb{R}^{m}$, where $y_{j}^{k} = \frac{\partial f_{k}(x)}{\partial x_{j}}$ for $k=1,\dots,m$, at every node $x_{j}$ in the reverse sweep. 
After the reverse sweep is completed, we obtain the full Jacobian\index{Automatic differentiation!Jacobian}
\begin{equation*}
\nabla f(x) = 
\begin{pmatrix}
\nabla f_{1}(x) \\ \vdots \\ \nabla f_{m}(x)
\end{pmatrix} = 
\begin{pmatrix}
y_{1}, \dots, y_{n} 
\end{pmatrix} \in \mathbb{R}^{m \times n} .
\end{equation*}

The advantage of the reverse mode is that the computational cost of the full gradient is only a small multiple (often about 4 to 5) of what needed to evaluate a scalar-valued function $f: \mathbb{R}^{n} \to \mathbb{R}$. In contrast, the forward mode requires $O(n)$ times of the computational cost for $f$.
Note that the so-called back-propagation in deep learning community is a special case of the reverse mode automatic differentiation.
For vector-valued function $f: \mathbb{R}^{n} \to \mathbb{R}^{m}$, the computational costs to evaluate the full Jacobian $\nabla f(x)$ using the forward and reverse modes become similar as $m$ increases though.
One issue with the reverse mode is that all the values computed in the forward sweep need to be stored, and the computer memory storing $y_{j}$ cannot be released until it has been used by all the parent nodes of $x_{j}$ in the reverse sweep. Therefore, the memory requirement for reverse mode of automatic differentiation may be higher in practice.

\subsection{Computation of Second-order Derivatives}
\label{subsec:ad2}

We follow the same pattern as Section \ref{subsec:ad2} and consider the computations of second-order derivatives using forward and reverse modes in order in this subsection.

\subsubsection*{Forward Mode to Compute Hessian}
Let $f : \mathbb{R}^{n} \to \mathbb{R}$ be a scalar-valued, twice continuously differentiable function.
In this case, the Hessian of $f$ is an $n \times n$ symmetric matrix $\nabla^{2} f(x) = [\partial_{ij} f(x)]$ where $\partial_{ij}f(x) = \frac{\partial^{2}f(x)}{\partial x_{i} \partial x_{j}}$ for $i,j \in [n]$. 
We first consider the evaluation of 
\begin{equation*}
D_{uv}x_{j} := u^{\top} (\nabla^{2} x_{j}) v \in \mathbb{R}
\end{equation*}
for every node $x_{j}$ in the computational graph of $f$ and any given column vectors $u, v \in \mathbb{R}^{n}$.
Like the forward mode\index{Automatic differentiation!forward mode} to compute gradient, we only perform a forward sweep on the computational graph of $f$. 
The difference is that we also need to compute several second-order partial derivatives at every node $x_{j}$ in the forward sweep. 
When the forward sweep is completed, we obtain the desired value $u^{\top}\nabla^{2} f(x) v = u^{\top} (\nabla^{2} x_{N}) v$ at the end node $x_{N}$.

\begin{figure}
\centering
\begin{tikzpicture}[scale=1, transform shape]
	\Vertex[x=0, y=0, color=none, size=0.6, style=dashed]{xk2}
	\Vertex[x=0, y=1, color=none, size=0.6, style=dashed]{xj1}
	\Vertex[x=0, y=2, color=none, size=0.6, style=dashed]{xj2}
	\Vertex[x=1.5, y=0.5, label=$x_{k}$, color=none, size=0.6]{xk}
	\Vertex[x=1.5, y=1.5, label=$x_{j}$, color=none, size=0.6]{xj}
	\Vertex[x=3, y=1, label=$x_{i}$, color=none, size=0.6]{xi}
	
	\Edge[Direct, lw=1pt, style={dashed}](xk2)(xk)
	\Edge[Direct, lw=1pt, style={dashed}](xj1)(xj)
	\Edge[Direct, lw=1pt, style={dashed}](xj2)(xj)
	\Edge[Direct, lw=1pt](xj)(xk)
	\Edge[Direct, lw=1pt](xk)(xi)
	\Edge[Direct, lw=1pt](xj)(xi)
	
\end{tikzpicture}
\caption{The subgraph (of the computational graph of $f$) involved in the evaluations of the quantities \eqref{eq:forward-mode-Hessian-node} at $x_{i}$ in the forward mode to compute $D_{uv} f(x)$. Note that each node can have up to two parent nodes. The subgraph shown here is the most complicated case as $x_{i}$ has two parents $x_{j}$ and $x_{k}$, and $x_{j}$ is also a parent of $x_{k}$. If $x_{i}$ has only one parent, or the parents $x_{j}$ and $x_{k}$ are not a parent of each other, then the evaluations of \eqref{eq:forward-mode-Hessian-node} are similar and simpler than the case shown in this subgraph.}
\label{fig:forward-Hessian-subgraph}
\end{figure}
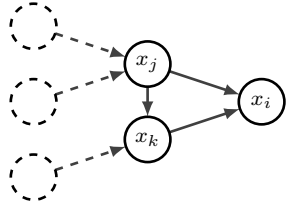

More precisely, at every node $x_{i}$ in the computational graph of $f$, we need to evaluate the following quantities:
\begin{align}
\label{eq:forward-mode-Hessian-node}
x_{i},\ \ \underbrace{\frac{\partial x_{i}}{\partial x_{j}}, \ \frac{\partial x_{i}}{\partial x_{k}},  \ D_{u} x_{i}, \ D_{v} x_{i}}_{\text{4 quantities of first-order terms}} , \qquad \qquad \qquad \qquad \qquad  & \\
\underbrace{\frac{\partial^{2} x_{i}}{\partial x_{j}^{2}}, \ \frac{\partial^{2} x_{i}}{\partial x_{k} \partial x_{j}}, \ \frac{\partial^{2} x_{i}}{\partial x_{k}^{2}}, \ D_{u}\Big(\frac{\partial x_{i}}{\partial x_{j}} \Big), \ D_{u}\Big(\frac{\partial x_{i}}{\partial x_{k}} \Big), \ D_{v}\Big(\frac{\partial x_{i}}{\partial x_{j}} \Big),\ D_{v}\Big(\frac{\partial x_{i}}{\partial x_{k}} \Big), \ D_{uv} x_{i} }_{\text{8 quantities of second-order terms}} .&  \nonumber
\end{align}
Recall that $x_{i}$ has either one or two parent nodes for every $i=n+1,\dots,N$. Without loss of generality, we consider the most complicated case where $x_{i}$ has two parent nodes $x_{j}$ and $x_{k}$, and $x_{j}$ is a parent node of $x_{k}$. 
If $x_{i}$ has one parent only, or it has two parents $x_{j}$ and $x_{k}$ which are not a parent of each other, then the derivation is similar and simpler.
The subgraph of the computational graph for $f$ for the case we consider is shown in Figure \ref{fig:forward-Hessian-subgraph}.
To evaluate the quantities in \eqref{eq:forward-mode-Hessian-node}, we notice that such quantities have already been evaluated and are stored at the parent nodes $x_{j}$ and $x_{k}$ in the forward sweep.
Since $x_{i}$ is an elementary function of $x_{j}$ and $x_{k}$ with some given form like those in \eqref{eq:ad-example-nodes}, we can directly evaluate 
\begin{equation*}
x_{i}, \ \frac{\partial x_{i}}{\partial x_{j}}, \ \frac{\partial x_{i}}{\partial x_{k}}, \ \frac{\partial^{2} x_{i}}{\partial x_{j}^{2}}, \ \frac{\partial^{2} x_{i}}{\partial x_{k} \partial x_{j}}, \ \frac{\partial^{2} x_{i}}{\partial x_{k}^{2}} .
\end{equation*}
Furthermore, by the chain rule, we can evaluate 
\begin{align*}
D_{u} x_{i} & = \frac{\partial x_{i}}{\partial x_{j}} D_{u} x_{j} + \frac{\partial x_{i}}{\partial x_{k}} D_{u} x_{k} , \\
D_{v} x_{i} & = \frac{\partial x_{i}}{\partial x_{j}} D_{v} x_{j} + \frac{\partial x_{i}}{\partial x_{k}} D_{v} x_{k} ,
\end{align*}
since all the terms on the right-hand sides are known.
Then we have
\begin{align}
D_{u} \Big( \frac{\partial x_{i}}{\partial x_{j}} \Big)
& = \frac{\partial }{\partial x_{j}}(D_{u} x_{i}) \nonumber \\
& = \frac{\partial }{\partial x_{j}} \Big( \frac{\partial x_{i}}{\partial x_{j}} D_{u} x_{j} + \frac{\partial x_{i}}{\partial x_{k}} D_{u} x_{k} \Big) \label{eq:forward-mode-Hessian-Diju}\\
& = \frac{\partial^{2} x_{i}}{\partial x_{j}^{2}} D_{u}x_{j} + \frac{\partial x_{i}}{\partial x_{j}} \frac{\partial}{\partial x_{j}}(D_{u}x_{j}) + \frac{\partial^{2} x_{i}}{\partial x_{k}\partial x_{j}} D_{u}x_{k} + \frac{\partial x_{i}}{\partial x_{j}} \frac{\partial}{\partial x_{j}}(D_{u}x_{k}). \nonumber 
\end{align}
Notice that $\frac{\partial}{\partial x_{j}}(D_{u}x_{j}) = 0$ and $\frac{\partial}{\partial x_{j}}(D_{u}x_{k}) = D_{u}(\frac{\partial x_{k}}{\partial x_{j}})$ is known when we evaluate the quantities \eqref{eq:forward-mode-Hessian-node} at $x_{k}$ (or equal to $0$ if $x_{j}$ is not a parent of $x_{k}$).
Therefore, all terms in \eqref{eq:forward-mode-Hessian-Diju} are known and $D_{u}(\frac{\partial x_{i}}{\partial x_{j}})$ can be evaluated. 
Similarly, $D_{v}(\frac{\partial x_{i}}{\partial x_{j}})$, $D_{u}(\frac{\partial x_{i}}{\partial x_{k}})$, and $D_{v}(\frac{\partial x_{i}}{\partial x_{k}})$ can be evaluated as well.
Lastly, we have 
\begin{align*}
D_{uv} x_{i}
& = D_{v} (D_{u} x_{i}) \\
& = D_{v} \Big( \frac{\partial x_{i}}{\partial x_{j}} D_{u} x_{j} + \frac{\partial x_{i}}{\partial x_{k}} D_{u} x_{k} \Big) \\
& = D_{v}\Big(\frac{\partial x_{i}}{\partial x_{j}}\Big) D_{u}x_{j} + \frac{\partial x_{i}}{\partial x_{j}} D_{uv}x_{j} + D_{v}\Big(\frac{\partial x_{i}}{\partial x_{k}}\Big) D_{u}x_{k} + \frac{\partial x_{i}}{\partial x_{k}} D_{uv}x_{k} ,
\end{align*}
where all terms in the sum are known at this point, and $D_{uv} x_{i}$ can be evaluated.
Now we have evaluated all terms in \eqref{eq:forward-mode-Hessian-node} for $x_{j}$ and can move on to the next node.
It is worth noting that, the computational costs can be much lower than they appear in the formulas above, since many terms may vanish as $x_{i}$ is merely the result of some elementary operation of $x_{j}$ and $x_{k}$.
We follow this way to complete the forward sweep and obtain the desired quantity $D_{uv}x_{N} = u^{\top} \nabla^{2}f(x) v$.

We remark that the forward mode discussed above can be used to evaluate the curvature of the curve $c(t):=f(x+tv)$ at $t=0$ for any given $x, v\in \mathbb{R}^{n}$: there are $c(0) = f(x)$, $c'(0) = D_{v}f(x)$, and
\begin{equation*}
c''(0) = v^{\top} \nabla^{2} f(x) v = D_{vv} f(x).
\end{equation*}

If we want to compute the entire Hessian $\nabla^{2} f(x)$, we can set $u = e_{i}$ and $v = e_{j}$ for all $i \le j \le n$ and $1 \le i \le n$ to obtain its upper-right triangle. By the symmetry of $\nabla^{2} f(x)$ provided $f \in C^{2}$, we obtain the full Hessian. 
This requires $\frac{1}{2} n(n+1) $ different pairs of $(u,v)=(e_{i},e_{j})$ to compute the Hessian.
If the Hessian is sparse and the locations of nonzero entries are known, then the number of pairs $(u,v)$ needed to be evaluated can be significantly reduced and the computational cost can be lowered. 

If we want to evaluate the Hessian-vector product $\nabla^{2} f(x) v$ for some given vector $v \in \mathbb{R}^{n}$, then we only need to vary $u=e_{i}$ for $i \in [n]$ in $u^{\top} \nabla^{2} f(x) v$. The total computational cost will reduce to $O(n)$ in this case, which is still high for large $n$.
Next, we show that the reverse mode can address this issue.

\subsubsection*{Reverse Mode to Compute Hessian}
We directly consider the computation of the Hessian-vector product\index{Automatic differentiation!Hessian-vector} $\nabla^{2} f(x) v$ for $f \in C^{2}(\mathbb{R}^{n}; \mathbb{R})$ and any vector $v \in \mathbb{R}^{n}$.
We first perform the forward sweep and compute the values of $x_{j}$ and $D_{v}x_{j}$ at every node $x_{j}$ in the computational graph of $f$. 
At the end node $x_{N}$, we obtain the values of both $f(x) = x_{N}$ and $D_{v}f(x) = D_{v}x_{N}$. 
Then we perform the reverse sweep and compute $z_{j} := \frac{\partial }{\partial x_{j}}(D_{v} f(x))$ at every node $x_{j}$\index{Automatic differentiation!reverse mode}. 
Notice that 
\begin{equation*}
z_{N} = \frac{\partial }{\partial x_{N}}(D_{v} f(x)) = D_{v}\Big(\frac{\partial f(x)}{\partial x_{N}} \Big) = 0
\end{equation*}
since $x_{N} = f(x)$.
Then, by the chain rule, we know that
\begin{equation}
\label{eq:reverse-mode-Hessian-z}
z_{j} = \frac{\partial }{\partial x_{j}} (D_{v} f(x)) = \sum_{x_{j} \in C_{j}} \frac{\partial }{\partial x_{i}} (D_{v} f(x)) \frac{\partial x_{i}}{\partial x_{j}} = \sum_{x_{i} \in C_{j}} z_{i} \frac{\partial x_{i}}{\partial x_{j}} ,
\end{equation}
where $z_{i}$ at every child node $x_{i}$ of $x_{j}$ has been evaluated at this point.
When the reverse sweep is done, we obtain 
\begin{equation*}
z_{j} = \frac{\partial }{\partial x_{j}} (D_{v} f(x)) = \frac{\partial }{\partial x_{j}} \Big( \sum_{i=1}^{n} v_{i} \partial_{i}f(x) \Big) = \sum_{i=1}^{n} v_{i} \partial^{2}_{ij} f(x)
\end{equation*}
at each $x_{j}$ of the input variables $x=(x_{1}, \dots, x_{n})$. Then we have
\begin{equation*}
\nabla^{2} f(x) v = 
\begin{pmatrix}
\partial^{2}_{11}f(x) & \cdots & \partial^{2}_{1n}f(x) \\
\vdots & \ddots & \vdots \\
\partial^{2}_{n1}f(x) & \cdots & \partial^{2}_{nn}f(x)
\end{pmatrix}
\begin{pmatrix}
v_{1} \\ \vdots \\ v_{n}
\end{pmatrix} 
= 
\begin{pmatrix}
z_{1} \\ \vdots \\ z_{n}
\end{pmatrix} \in \mathbb{R}^{n} .
\end{equation*}
The computational cost of $\nabla^{2}f(x)v$ using the reverse mode described above is only a moderate multiple (about 12) of the cost to compute $f$. This is significantly lower than the $O(n)$ cost of forward mode for the same task.
The only issue is that the reverse mode requires all $x_{j}$'s evaluated in the forward sweep to be stored, and the memory storing $z_{j}$ cannot be released until $z_{j}$ has been used by all the parent nodes of $x_{j}$ in the reverse sweep.

If the full Hessian\index{Automatic differentiation!Hessian} $\nabla^{2}f(x)$ is needed, we can follow the procedure above with $v=e_{i}$ for all $i \in [n]$. The total computational cost is $O(n)$. Again, this can be further reduced if the sparsity structure of the Hessian is known.

\section{Deterministic Optimization Algorithms}
\label{sec:deterministic-opt}

In this section, we present several representative algorithms for deterministic optimization\index{Deterministic optimization}, namely, the objective function\index{Objective function} has an analytic form without any randomness or expectation, and its derivatives at all orders can be accurately evaluated, say by automatic differentiation.
This may be more clear by comparing the deterministic optimization \eqref{eq:deterministic-min} and stochastic optimization \eqref{eq:soa-loss} below.

The general form of deterministic optimization is written as
\begin{equation}
    \label{eq:deterministic-min}
    \min_{x \in \Omega} \ f(x) ,
\end{equation}
where $\Omega$ is a closed subset of $\mathbb{R}^{n}$ and the objective function $f: \mathbb{R}^{n} \to \mathbb{R}$ is possibly non-convex.
We assume that $f$ is differentiable and its gradient $\nabla f$ is Lipschitz continuous with some Lipschitz constant $L>0$ (often referred to as ``smooth'' in the optimization community).
It is worth noting that we usually assume $f$ is globally Lipschitz in $\mathbb{R}^{n}$ with $L$, but this can be often relaxed to being Lipschitz on a bounded domain of interest, as we will explain in the derivations later.
This assumption is mild and held in the majority of problems we encounter in deep learning.

With the smoothness assumption, we have the following important inequality which is frequently used in convergence proofs of optimization algorithms.

\begin{proposition}
Suppose $f: \mathbb{R}^{n} \to \mathbb{R}$ is continuously differentiable and $\nabla f$ is Lipschitz continuous with Lipschitz constant $L > 0$. Then 
\begin{equation}
\label{eq:grad-L-ineq}
f(y) \le f(x) + \langle \nabla f(x), y-x \rangle + \frac{L}{2} |y - x|^{2}
\end{equation}
for all $x , y \in \mathbb{R}^{n}$.
\end{proposition}

\begin{proof}
For any $x, y \in \mathbb{R}^{n}$, we define $\phi : \mathbb{R} \to \mathbb{R}$ by
\begin{equation*}
\phi(t) = f((1-t) x + ty) .
\end{equation*}
Since $f$ is differentiable, we have
\begin{equation*}
\phi(1) = \phi(0) + \int_{0}^{1} \phi'(t) \,dt,
\end{equation*}
which implies that
\begin{equation*}
f(y) = f(x) + \int_{0}^{1} \langle \nabla f((1-t)x + ty), y-x \rangle \, dt .
\end{equation*}
Subtracting $f(x) + \langle f(x), y-x \rangle$ on both sides, we obtain
\begin{align*}
f(y) - f(x) - \langle \nabla f(x), y-x \rangle 
& = \int_{0}^{1} \langle \nabla f((1-t)x + ty) - \nabla f(x) , y - x \rangle \, dt \\
& \le \int_{0}^{1} | \nabla f((1-t)x + ty) - \nabla f(x) | |y-x| \, dt \\
& \le L |y - x|^{2} \int_{0}^{1} t \,dt \\
& = \frac{L}{2} |y-x|^{2} ,
\end{align*}
where the second inequality is because $\nabla f$ is Lipschitz continuous.
\end{proof}

%Throughout this book, we always assume that $f$ has at least one minimizer denoted by $x^{*}$ in $\mathbb{R}^{n}$, and its value $f^{*} := f(x^{*}) = \min_{x \in \mathbb{R}^{n}} f(x)$ is finite, i.e., $f^{*} > -\infty$.
%%
%We remark that this assumption is almost always needed in optimization algorithms. It is an assumption that the optimization problem is properly set up rather than an artificial requirement for the algorithms to work.

\subsection{Algorithms for Unconstrained Optimization}
\label{subsec:alg-uncon}

If the optimization problem \eqref{eq:deterministic-min} is unconstrained\index{Unconstrained optimization}, namely, $\Omega = \mathbb{R}^{n}$, our major task is to find a critical point of $f$ according to the first-order necessary condition.
To this end, we consider the following iterative scheme:
\begin{equation}
\label{eq:descent-iteration}
x_{k+1} = x_{k} + \alpha_{k} v_{k} ,
\end{equation}
where $k=0,1,2,\dots$ is called the \emph{iteration counter}, $\alpha_{k} > 0$ the \emph{step size}\index{Step size} (or \emph{learning rate} which is commonly used in the deep learning community), and $v_{k}$ is a descent direction\index{Descent direction} of $f$ at $x_{k}$ in the $k$th iteration. 
Here by descent direction we meant that $\langle \nabla f(x_{k}), v_{k} \rangle < 0$ (we will discuss descent direction in detail later in this section).
In the developments of optimization algorithms like \eqref{eq:descent-iteration}, the key is to determine proper $v_{k}$ and $\alpha_{k}$ such that the sequence $\{x_{k}\}$ generated by \eqref{eq:descent-iteration} satisfies $|\nabla f(x_{k})| \to 0$ as $k \to \infty$.
%
% With some mild conditions which we will show later, we can claim that the accumulation points\footnote{We call $\hat{x}$ an accumulation point of the sequence $\{x_{k}\}$ if $\{x_{k}\}$ has a subsequence, say $\{x_{k_{j}}\}$, converging to $\hat{x}$ as $j \to \infty$.} of $\{x_{k}\}$ are critical points of $f$.

\subsubsection*{Gradient Descent Method and Line Search}

We first show the construction of the gradient descent method, with an automated step size selection strategy called backtracking line search, to illustrate the ideal features and potential issues of an optimization algorithm.

In the gradient descent method, the easiest step is to set up the descent direction $v_{k}$. 
We temporarily assume that $v_{k}$ is a unit vector since our goal is to determine a direction for the moment. In particular, we want to find $v_{k}$ such that the directional derivative $D_{v_{k}} f(x_{k})$ can be negative and as small as possible, so we can attain fastest decay of $f$ along this direction. This easily renders the following conclusion:
\begin{align*}
v_{k}
& = \argmin_{|v| = 1} D_{v} f(x_{k}) \\
& = \argmin_{|v| = 1} \langle \nabla f(x_{k}), v \rangle \\
& = - \nabla f(x_{k}) / |\nabla f(x_{k}) |.
\end{align*}
Therefore, we can just use $v_{k} = -\nabla f(x_{k})$ and let the step size $\alpha_{k}$ be responsible of the scaling as we show in detail later.
Since we use the negative gradient as the descent direction, we have the iterative scheme
\begin{equation}
    \label{eq:gd}
    x_{k+1} = x_{k} - \alpha_{k} \nabla f(x_{k}) ,
\end{equation} 
and call this iterative scheme \eqref{eq:gd} the \emph{gradient descent method}\index{Gradient descent algorithm}.

We remark that the choice $v_{k} = - \nabla f(x_{k})$ in the gradient descent method \eqref{eq:gd} just ensures $v_{k}$ to be a descent direction, which merely simplifies our discussion below. 
In fact, we can prescribe any $c>0$ and choose any $v_{k}$ as long as $\nabla f(x_{k}) v_{k} \le -c | \nabla f(x_{k})|^{2}$ for all $k$, our derivations below still work.

Determining a suitable step size $\alpha_{k}$ in \eqref{eq:descent-iteration} and \eqref{eq:gd} is particularly important, as it affects the convergence and stability of the algorithm in practice.
Large step sizes result in faster progress but may cause the sequence $\{x_{k}\}$ in \eqref{eq:gd} to diverge or highly fluctuate.
Small step sizes make the sequence $\{x_{k}\}$ progress slowly and the algorithm work very inefficiently.

To find the proper step size at iteration $k$, it is intuitive to look for the $\alpha$ that minimizes $f$ along the ray $x_{k} + \alpha v_{k}$ for $\alpha \ge 0$. To this end, we can set the line search function $\phi_{k} : \mathbb{R} \to \mathbb{R}$ as $\phi_{k}(\alpha) := f(x_{k} + \alpha v_{k})$ and solve this one-dimensional optimization problem to obtain $\alpha_{k} = \argmin_{\alpha \ge 0} \phi_{k}(\alpha)$.
A gradient descent method with such chosen step sizes is called \emph{steepest descent method}\index{Steepest descent method}.
However, despite $\phi_{k}$ is a univariate function, finding its minimizer involves evaluations of $f$ and $\nabla f$, and the minimization problem is not easy to solve.

Another common approach is using a constant step size $\alpha > 0$ such that $\alpha_{k}= \alpha$ for all $k$ yields convergence.
In fact, we can show that for any $\alpha \in (0, 2/L)$, where $L$ is the Lipschitz constant of $\nabla f$, the sequence $\{x_{k}\}$ generated by \eqref{eq:gd} satisfies $|\nabla f(x_{k})| \to 0$.
However, the value of $L$ is often unknown and difficult to estimate, and we cannot determine the proper interval for $\alpha$ in practice.

There are a number of different strategies for the algorithm to determine the step sizes automatically without much knowledge about the objective function $f$. 
We consider one of the widely used strategies called \emph{backtracking line search}\index{Backtracking line search} to find $\alpha_{k}$. 
The goal of this line search strategy is to find a step size $\alpha_{k}$ such that it satisfies the following \emph{Armijo condition}\index{Armijo condition} for some prescribed $c \in (0,1)$:
\begin{equation}
\label{eq:armijo}
f( x_{k} + \alpha_{k} v_{k} ) \le f(x_{k}) + c \alpha_{k} \nabla f(x_{k}) v_{k}.
\end{equation}
In the gradient descent method \eqref{eq:gd}, we have $v_{k} = -\nabla f(x_{k})$ and hence the last term in \eqref{eq:armijo} reduces to $-c \alpha_{k} |v_{k}|^{2}$.

In general, as long as $v_{k}$ is a descent direction, we know $\nabla f(x_{k}) v_{k} < 0$, and \eqref{eq:armijo} implies that $f( x_{k} + \alpha_{k} v_{k} ) \le f(x_{k})$.
Hence, \eqref{eq:armijo} is also called the \emph{sufficient descent condition}\index{Sufficient descent condition}.
It is useful to interpret the Armijo condition \eqref{eq:armijo} in the following way: By using the line search function $\phi_{k}(\alpha) = f(x_{k} + \alpha v_{k})$ again, we can see that the Armijo condition \eqref{eq:armijo} effectively requires 
\begin{equation}
\label{eq:armijo-interpret}
\phi_{k}(\alpha) \le \phi_{k}(0) + c \alpha \phi_{k}'(0) .
\end{equation}
The right-hand side of \eqref{eq:armijo-interpret} is a straight line passing through $(0,\phi_{k}(0))$ with slope $c\phi_{k}'(0)<0$ (note that the tangent line of $f$ at $(x_{k}, f(x_{k}))$ corresponds to the straight line passing through $(0,\phi_{k}'(0))$ with more steep slope $\phi_{k}'(0)<0$). 
Since $f(x)$ is bounded below by $f^{*}$, this line $\phi_{k}(0) + c \alpha \phi_{k}'(0)$ must intercept with $\phi(\alpha)$ at some $\hat{\alpha}>0$, and the step size $\alpha_{k}$ should be between $0$ and this $\hat{\alpha}$.
Therefore, the Armijo condition \eqref{eq:armijo} ensures decreasing the objective function value and meanwhile prevents the step size $\alpha_{k}$ from being too large.

From \eqref{eq:armijo-interpret}, we can see that it holds true as long as $\alpha$ is sufficiently small. However, the step size $\alpha_{k}$ should not be too small in order to just satisfy \eqref{eq:armijo-interpret}. This is because that small step sizes cause slow improvements of the iterates $x_{k}$'s and the algorithm needs many more iterations to converge, reducing the overall computational cost efficiency. 
Therefore, a line search strategy may also include a \emph{curvature condition}\index{Curvature condition} such as 
\begin{equation}
\label{eq:line-search-curvature}
\phi_{k}'(\alpha) \ge \hat{c} \phi_{k}'(0)
\end{equation}
for some prescribed $\hat{c} \in (c,1)$. This condition prevents $\alpha_{k}$ from being too small.
As a consequence, a line search strategy also need an implementable procedure to find a step size $\alpha_{k}$ that satisfies both \eqref{eq:armijo-interpret} and \eqref{eq:line-search-curvature}.
%
% However, the curvature condition \eqref{eq:line-search-curvature} is generally more difficult to implement.

The combination of the Armijo condition \eqref{eq:armijo} and the curvature condition \eqref{eq:line-search-curvature} is called the \emph{Wolfe condition}\index{Wolfe condition}, which is one of the most commonly used conditions for line search.
Next, we show how backtracking line search works to automatically find such step sizes.

% Backtracking line search is a simple but effective method to find $\alpha_{k}$ satisfying \eqref{eq:armijo}, or equivalently \eqref{eq:armijo-interpret}, but not too small. 
%
In backtracking line search, we manually set an upper bound $\bar{\alpha}>0$ of step sizes in the gradient descent method and the backtracking rate $\rho \in (0,1)$ arbitrarily.
Then, at any iteration $k$, we set $\alpha_{k} = \bar{\alpha} \rho^{l_{k}}$ where
\begin{equation}
\label{eq:backtracking-lk}
l_{k} = \argmin \Big\{ l \in \{0,1,\dots\} : f(x_{k} + \bar{\alpha} \rho^{l} v_{k}) \le f(x_{k}) - c \bar{\alpha} \rho^{l} |v_{k}|^{2} \Big\} .
\end{equation}
To implement the process of finding $l_{k}$ in \eqref{eq:backtracking-lk}, we initialize $\alpha_{k} = \bar{\alpha}$ and check whether the Armijo condition is satisfied:
\begin{equation}
\label{eq:armijo-vk}
f(x_{k} + \alpha_{k} v_{k} ) \le f(x_{k}) - c \alpha_{k} |v_{k}|^{2} .
\end{equation}
If \eqref{eq:armijo-vk} is satisfied, we decide to use this $\alpha_{k}$; otherwise, we replace $\alpha_{k}$ with $\rho \alpha_{k}$, commonly written as $\alpha_{k} \leftarrow \rho \alpha_{k}$ in pseudo code, and try \eqref{eq:armijo-vk} again.
This process is repeated until \eqref{eq:armijo-vk} is satisfied and we will accept that $\alpha_{k}$ as our step size at iteration $k$.
The gradient descent algorithm with this backtracking line search strategy is summarized in Algorithm \ref{alg:gd-backtracking}.

\begin{algorithm}
\caption{Gradient descent method with backtracking line search}
\label{alg:gd-backtracking}
\begin{algorithmic}[1]
\REQUIRE $\bar{\alpha}, \epsilon_{\text{tol}} > 0$, $\rho, c \in (0,1)$, $x_{0} \in \mathbb{R}^{n}$, $k=0$
\REPEAT
	\STATE $v_{k} \leftarrow - \nabla f(x_{k})$
	\STATE $\alpha_{k} \leftarrow \bar{\alpha}$
	\WHILE{$f(x_{k}+\alpha_{k}v_{k}) > f(x_{k}) - c \alpha_{k}|v_{k}|^{2}$}
		\STATE $\alpha_{k} \leftarrow \rho \alpha_{k}$
	\ENDWHILE
	\STATE $x_{k} \leftarrow x_{k} + \alpha_{k} v_{k}$
	\STATE $k \leftarrow k+1$
\UNTIL $|v_{k}| < \epsilon_{\text{tol}}$
\ENSURE $x_{k}$
\end{algorithmic}
\end{algorithm}

Now several questions for Algorithm \ref{alg:gd-backtracking} remain: (i) Can the line search (Line 4--6) terminate in finitely many times (and if yes, what is the upper bound on the number of searches)? and (ii) Can the termination criterion\index{Termination criterion} (Line 9) be satisfied with finite $k$ (and if yes, how large $k$ needs to be)? 

We use the following lemma and theorem to answer questions (i) and (ii) above, respectively.

\begin{lemma}
\label{lem:gd-backtracking-ls-bound}
Suppose $f: \mathbb{R}^{n} \to \mathbb{R}$ is $C^{1}$ and $\nabla f$ is Lipschitz continuous with constant $L > 0$ in the sub-level set $\{x \in \mathbb{R}^{n} : f(x) \le f(x_{0}) \}$, where $x_{0} \in \mathbb{R}^{n}$ is any initial guess. For any $\bar{\alpha} > 0$ and $\rho, c \in (0,1)$, denote 
\begin{equation}
\label{eq:backtracking-l}
\bar{l} := \biggl\lceil \frac{ \log (2(1-c) \bar{\alpha}^{-1}L^{-1})}{ \log \rho } \biggr\rceil .
\end{equation}
Then the number of line searches (Line 4--6) is at most $\bar{l}$ in any iteration $k$ of Algorithm \ref{alg:gd-backtracking}. Namely, $l_{k} \le \bar{l}$ for all $k$, where $l_{k}$ is defined in \eqref{eq:backtracking-lk}.
\end{lemma}

\begin{proof}
We use a proof by contradiction. If $l_{k} > \bar{l}$, then
\begin{equation}
\label{eq:armijo-violated}
f(x_{k}) - c \alpha |v_{k}|^{2} < f(x_{k} + \alpha v_{k}) 
\end{equation}
with $\alpha := \bar{\alpha} \rho^{\bar{l}}$.
By the definition of $\bar{l}$ in \eqref{eq:backtracking-l}, we know
\begin{equation*}
\bar{l} \ge \frac{ \log (2(1-c) \bar{\alpha}^{-1}L^{-1})}{ \log \rho } ,
\end{equation*}
and due to $\rho \in (0,1)$ there is
\begin{equation*}
\alpha = \bar{\alpha} \rho^{\bar{l}} \le \frac{2(1-c)}{L} .
\end{equation*}

On the other hand, by applying \eqref{eq:grad-L-ineq}, we have
\begin{align*}
f(x_{k} + \alpha v_{k})
& \le f(x_{k}) + \langle \nabla f(x_{k}), \alpha v_{k} \rangle + \frac{L \alpha^{2}}{2} |v_{k}|^{2} \\
& = f(x_{k}) - \alpha \Big(1 - \frac{L \alpha}{2} \Big) |v_{k}|^{2} \\
& \le f(x_{k}) - c \alpha |v_{k}|^{2},
\end{align*}
which contradicts to \eqref{eq:armijo-violated}.
\end{proof}

Lemma \ref{lem:gd-backtracking-ls-bound} implies a uniform upper bound $\bar{l} = O(\log L)$ on the number of line searches needed in each iteration.
Note that each line search criterion check (Line 4 in Algorithm \ref{alg:gd-backtracking}) requires one evaluation of $f$ at the candidate point $x_{k} + \alpha_{k}v_{k}$.

Next, we show that the sequence $\{x_{k}\}$ generated by Algorithm \ref{alg:gd-backtracking} satisfies $|\nabla f(x_{k})| \to 0$ as $k \to \infty$ from any initial guess $x_{0}$.

\begin{theorem}
[Convergence of Algorithm \ref{alg:gd-backtracking}]
\label{thm:gd-convergence}
    For any $x_{0} \in \mathbb{R}^{n}$, $\bar{\alpha} >0$, and $\rho, c\in (0,1)$, the sequence $\{x_k\}$ generated by Algorithm \ref{alg:gd-backtracking}\index{Convergence} satisfies
\begin{enumerate}
\item The series of $|\nabla f(x_{k})|^{2}$ is bounded, i.e.,
\begin{equation}
\label{eq:gd-grad-series}
	\sum_{k=0}^{\infty} | \nabla f(x_{k}) |^2 < \infty \ .
\end{equation}
As a consequence, there is $| \nabla f(x_{k})| \to 0$ as $k\to\infty$.
\item For any $K > 1$, there is 
\begin{equation}
	\min_{0 \le k \le K-1} | \nabla f(x_{k}) |^2 \le \frac{L(f(x_{0}) - f^{*})}{2c(1-c)\rho K} = O \Big( \frac{1}{K} \Big) .
\end{equation}
\end{enumerate}
\end{theorem}

\begin{proof}
With the step size $\alpha_{k}$ obtained from the backtracking line search, we know the Armijo condition \eqref{eq:armijo} is satisfied and there is
\begin{equation*}
c \alpha_{k} | \nabla f(x_{k}) |^{2} \le f(x_{k}) - f(x_{k+1})
\end{equation*}
Taking sum of $k=0,1,\dots,K-1$ and noticing that $\alpha_{k} = \bar{\alpha} \rho^{l_{k}} \ge \bar{\alpha} \rho^{\bar{l}}$, we obtain
\begin{equation}
\label{eq:gd-pf-telescope}
c \bar{\alpha} \rho^{\bar{l}} \sum_{k=0}^{K-1} |\nabla f(x_{k}) |^{2} \le f(x_{0}) - f(x_{K}) \le f(x_{0}) - f^{*} < \infty
\end{equation}
since $f(x_{K}) \ge f^{*} > -\infty$. This justifies \eqref{eq:gd-grad-series} and the first claim.

Due to the definition of $\bar{l}$ in \eqref{eq:backtracking-l}, we know
\begin{equation*}
\bar{l} - 1 \le \frac{ \log (2(1-c) \bar{\alpha}^{-1}L^{-1})}{ \log \rho } 
\end{equation*}
and hence
\begin{equation*}
\rho^{\bar{l} - 1} \ge \frac{2(1-c)}{\bar{\alpha} L}.
\end{equation*}
Therefore, from \eqref{eq:gd-pf-telescope}, we have
\begin{equation}
\min_{0 \le k \le K-1} |\nabla f(x_{k}) |^{2} \le \frac{1}{K}\sum_{k=0}^{K-1} |\nabla f(x_{k})|^{2} \le \frac{L(f(x_{0}) - f^{*})}{2c(1-c)\rho K}.
\end{equation}
This justifies the second claim.
\end{proof}

Theorem \ref{thm:gd-convergence} provides us with confidence that the iterates $\{x_{k}\}$ generated by Algorithm \ref{alg:gd-backtracking} converges to a critical point at a sublinear rate\index{Convergence!sublinear} $O(1/k)$. This result is often considered the best convergence result for general smooth non-convex optimization problems.

Meanwhile, the algorithm itself does not require much knowledge about the problem: We do not need to struggle with the choice of step sizes because the line search can find them automatically. Moreover, the convergence is guaranteed regardless of the hyperparameter\index{Hyperparameter} (the parameters that need manually selected in an algorithm) values as long as they are in the specified ranges, which are independent of the problem.
By contrast, recall that if we use a constant step size $\alpha_{k} \equiv \alpha \in (0,2/L)$, then we need to know the Lipschitz constant $L$ of $\nabla f$, which is difficult to estimate and varies with different $f$.

Despite the aforementioned promising properties of Algorithm \ref{alg:gd-backtracking}, we are often more interested in the practical efficiency of an optimization algorithm. 
The reason is that, from practical perspective, it is more appealing to have a significantly faster empirical convergence (as this reduces time, resources, and power to obtain a solution) at the cost of some freedom in hyperparameter setting or convergence rate guarantee.
Gradient descent method is known to converge extremely slowly with ill-conditioned\index{Ill-conditioned} objective function $f$, even if $f$ is strongly convex\index{Convex!strongly} (in which case the ratio $L/\mu$, where $\mu$ is the strong convexity parameter of $f$, is large).
Therefore, extensive works have been devoted to build more efficient optimization algorithms, particularly for large-scale optimization problems.
Next, we consider several optimization methods to mitigate the ill-conditioning issue of optimization problems by exploiting second-order information of objective functions.

\subsubsection*{Newton's Method and Quasi-Newton Methods}

The gradient descent method is easy to implement as the descent direction $v_{k}$ is simply chosen to be the negative gradient $- \nabla f(x_{k})$ at each iterate $x_{k}$. 
However, the gradient descent method can be very slow in convergence even if the step sizes are chosen as in the steepest descent method.
This is mostly due to the ill-conditioning of the Hessian of the objective function.
We now consider its opposite extreme, the \emph{Newton's method}\index{Newton's method}, which is more structurally complicated than most descent methods but is one of the fastest optimization methods in terms of convergence rate.

Newton's method proceeds with the typical form descent methods as
\begin{equation}
\label{eq:newton-iteration}
x_{k + 1} = x_{k} + \alpha_{k} v_{k} ,
\end{equation}
where the key is in the selection of the \emph{Newton's direction}\index{Newton's direction}, which we denote by:
\begin{equation}
\label{eq:newton-direction}
v_{k}^{N} := - (\nabla^{2} f(x_{k}))^{-1} \nabla f(x_{k}) .
\end{equation}
The idea of the Newton's direction is that the update $x_{k+1}$ should be the minimizer of a second-order approximation to the original objective function at $x_{k}$:
\begin{equation*}
f(x) \approx \tilde{f}_{k}(x) := f(x_{k}) + \nabla f(x_{k}) (x - x_{k}) + \frac{1}{2} (x - x_{k})^{\top} \nabla^{2} f(x_{k}) (x - x_{k}) .
\end{equation*}
Taking the gradient of $\tilde{f}_{k}$ above with respect to $x$, we can see that $x_{k+1}$ given in \eqref{eq:newton-iteration} is indeed the stationary point of $\tilde{f}_{k}$, provided that $\nabla^{2} f(x_{k})$ is invertible.
As the Hessian provides the second-order information of the objective function near $x_{k}$, it is expected that the Newton's direction $v_{k}$ in \eqref{eq:newton-direction} is more effective than only using the first-order information as in gradient descent methods.

As we can see, the Newton's direction \eqref{eq:newton-direction} involves the inverse of the Hessian $\nabla^{2} f(x_{k})$.
Therefore, if $\nabla^{2} f(x_{k})$ is singular, the Newton's method cannot work.
Furthermore, if $\nabla^{2} f(x_{k})$ is not positive definite, then it is possible that $\vkN$ defined in \eqref{eq:newton-direction} is not a descent direction.
These issues can often be remedied by using line search strategy of step sizes, trust-region technique, Gauss--Newton approximation, and etc.
As our goal here is to motivate the idea and demonstrate the convergence of the Newton's method and its important variant later, we simply assume that $\vkN$ exists and is a descent direction.

We also notice that the Newton's step $\vkN$ in \eqref{eq:newton-direction} can be very expensive to compute as it requires the inverse of the Hessian matrix, as the computational cost is typically cubic in $n$. We will address this issue later and just temporarily assume that we have $\vkN$ for each iteration.

Now we consider the local convergence\index{Convergence!local} behavior of the Newton's method with $\vkN$ and constant step size $\alpha_{k}$.
By local convergence\index{Convergence!local}, we meant that for a local minimizer\index{Minimizer!local} $x^{*}$ satisfying the sufficient optimality condition (i.e., $\nabla f(x^{*}) = 0$ and $\nabla^{2} f(x^{*}) \succ 0$), we can show that the Newton's method converges if the initial guess is close enough to $\xs$.
%
% In addition, it is required that the initial $x_{0}$ is sufficiently close to $x^{*}$. 
%
We admit that these conditions seem strong and not practical, however, they often appear to be the best convergence guarantee that have been established for non-convex problems.
This specific convergence result of the Newton's method is given as follows.

\begin{theorem}
[Local quadratic convergence of Newton's method]
\label{thm:newtown-convergence}
Suppose $f: \mathbb{R}^{n} \to \mathbb{R}$ is $C^{2}$ and $\nabla^{2} f$ is Lipschitz continuous, namely, there exists $M > 0$ such that
\begin{equation*}
| \nabla^{2} f(\hat{x}) - \nabla^{2} f(x) | \le M |\hat{x} - x|
\end{equation*}
for any $x, \hat{x} \in \mathbb{R}^{n}$, where $|A| := \max_{x:|x|\le 1} |Ax|$ is the matrix 2-norm.
Let $x^{*}$ be a local minimizer with $\nabla f(x^{*}) = 0$ and $\nabla^{2} f(x^{*}) \succ 0$, and $\{x_{k}\}$ be the sequence generated by the Newton's method with Newton direction $\vkN$ and step size $\alpha_{k}=1$ for every $k$, then the following statements hold:
\begin{enumerate}
\item
If $x_{0}$ is sufficiently close to $x^{*}$, then $x_{k} \to x^{*}$ as $k\to \infty$ at a quadratic rate (to be specified in the proof).

\item
Moreover, $|\nabla f(x_{k})| \to 0$ as $k\to \infty$ at a quadratic rate.
\end{enumerate}
\end{theorem}

\begin{proof}
We notice that
\begin{align}
x_{k+1} - \xs
& = x_{k} - \vkN - x^{*} \nonumber \\
& = x_{k} - ( \nabla^{2} f(x_{k}) )^{-1} \nabla f(x_{k}) - \xs \label{eq:newton-pf-1}\\
& = (\nabla^{2} f(x_{k}))^{-1} ( \nabla^{2} f(x_{k}) (x_{k} - \xs) - \nabla f(x_{k}) ) . \nonumber
\end{align}
By Taylor's theorem, we know
\begin{equation}
\label{eq:newton-pf-2}
\nabla f(x_{k}) - \nabla f(\xs) 
= \int_{0}^{1} \nabla^{2} f(x_{k} + t(\xs - x_{k})) (x_{k} - \xs) \, dt .
\end{equation}
Hence there is
\begin{align}
| \nabla^{2} f(x_{k})&(x_{k} - \xs)  - ( \nabla f(x_{k}) - \nabla f(\xs) ) |  \nonumber \\
& = \Big| \int_{0}^{1} \Big(\nabla^{2} f(x_{k}) - \nabla^{2} f(x_{k}+ t(x_{k} - \xs)) \Big) (x_{k} - \xs) \, dt\Big| \nonumber \\
& \le M |x_{k} - \xs |^{2} \int_{0}^{1} t \, dt \label{eq:newton-pf-3} \\
& = \frac{M}{2} |x_{k} - \xs |^{2}  \nonumber
\end{align}
where the first equality is due to \eqref{eq:newton-pf-2} and the inequality is because $\nabla^{2} f$ is $M$-Lipschitz continuous.

Since $\nabla^{2} f(\xs) \succ 0$ and $f \in C^{2}$, we know there exists a neighborhood of $x^{*}$ such that $\nabla^{2} f(x)$ is invertible at every $x$ in the neighborhood.
Moreover, there exists $r>0$ such that 
\begin{equation}
\label{eq:newton-pf-4}
| (\nabla^{2} f(x))^{-1} | \le 2 | (\nabla^{2} f(\xs))^{-1} |
\end{equation}
for all $x \in B_{r}(\xs)$. 

Combining \eqref{eq:newton-pf-1}, \eqref{eq:newton-pf-3}, and \eqref{eq:newton-pf-4}, we obtain
\begin{align*}
| x_{k+1} - \xs | 
& \le \frac{M}{2} \cdot 2 \, | (\nabla^{2} f(\xs))^{-1} | \, |x_{k} - \xs|^{2} \\
& = M \, | (\nabla^{2} f(\xs))^{-1} | \, |x_{k} - \xs|^{2} .
\end{align*}
By choosing $x_{0}$ such that
\begin{equation*}
| x_{0} - \xs | \le \min \Big( r, \frac{1}{2M | (\nabla^{2} f(\xs))^{-1}|} \Big) ,
\end{equation*}
we can show by induction that there exists $C>0$ such that 
\begin{equation}
\label{eq:newton-pf-5}
|x_{k+1} - \xs | \le C |x_{k} - \xs|^{2} \quad \text{and} \quad |x_{k} - \xs | < 1/C
\end{equation}
for all $k$. This implies that $x_{k} \to \xs$, and the first condition \eqref{eq:newton-pf-5} is the so-called \emph{quadratic convergence rate}\index{Convergence!quadratic}.

Furthermore, by the choice of $\vkN$ in \eqref{eq:newton-direction}, we have
\begin{equation*}
\nabla f(x_{k}) + \nabla^{2} f(x_{k}) \vkN = 0
\end{equation*}
and thus
\begin{align*}
| \nabla f(x_{k+1}) |
& = | \nabla f(x_{k+1}) - \nabla f(x_{k}) - \nabla^{2} f(x_{k}) \vkN | \\
& = \Big| \int_{0}^{1} \nabla^{2} f(x_{k} + t(x_{k+1}-x_{k})) (x_{k+1} - x_{k}) \, dt - \nabla^{2} f(x_{k}) \vkN \Big| \\
& = \Big| \int_{0}^{1} \nabla^{2} f(x_{k} + t \vkN) \vkN \, dt - \nabla^{2} f(x_{k}) \vkN \Big| \\
& \le \int_{0}^{1} | \nabla^{2} f(x_{k} + t \vkN) - \nabla^{2} f(x_{k}) | |\vkN| \, dt  \\
& \le \frac{M}{2} |\vkN|^{2} \\
& \le \frac{M}{2} | (\nabla^{2} f(x_{k}))^{-1} |^{2} |\nabla f(x_{k})|^{2} \\
& \le 2 M | (\nabla^{2} f(\xs))^{-1} |^{2} |\nabla f(x_{k})|^{2} ,
\end{align*}
which implies the quadratic rate of $|\nabla f(x_{k})|$ converging to $0$.
\end{proof}

Now we return to the unsolved issue earlier about computing $\vkN$ in \eqref{eq:newton-direction}.
For large-scale problems, i.e., $n$ is at millions or higher, it is intractable to find the inverse of $\nabla^{2} f(x_{k})$ as this requires a computational cost of $O(n^{3})$. 
Instead of directly finding the inverse, we can just compute $\vkN$ by solving 
\begin{equation}
\label{eq:newton-v-eq}
\nabla^{2} f(x_{k}) v = - \nabla f(x_{k}) .
\end{equation}
This is in the form of a standard linear system 
\begin{equation}
\label{eq:Qxb}
Q x = b
\end{equation}
where $Q$ is symmetric positive definite (when we assume $\nabla^{2} f(x_{k})$ in \eqref{eq:newton-v-eq} is positive definite, which is true when $x_{k}$ is sufficiently close to $\xs$ where $\nabla^{2} f(\xs) \succ 0$).

There have been a substantial amount of numerical methods developed to solve large linear system like \eqref{eq:Qxb}.
Here we only present the method called \emph{conjugate gradient}\index{Conjugate gradient} (CG), which is one of the most classic and efficient methods for solving \eqref{eq:Qxb}. 
The standard CG method for solving linear systems is given in Algorithm \ref{alg:cg}.
\begin{algorithm}
\caption{Conjugate gradient (CG) method for solving $Qx = b$}
\label{alg:cg}
\begin{algorithmic}[1]
\REQUIRE Initial $x_{0} \in \mathbb{R}^{n}$, $g_{0}=Qx_{0} - b$, $d_{0} = -g_{0}$.
\FOR{$j=0,1,,\dots,n-1$}
	\STATE 
	\begin{subequations}
	\label{eq:cg}
	\begin{align}
	\alpha_{j} & = \frac{g_{j} \cdot d_{j}}{g_{j} \cdot (Qd_{j})}, \label{eq:cg-alpha} \\
	x_{j+1} & = x_{j} + \alpha_{j} d_{j}, \label{eq:cg-x} \\
	g_{j+1} & = Q x_{j+1} - b = g_{j} + \alpha_{j} (Q d_{j}), \label{eq:cg-g} \\
	\beta_{j} & = \frac{g_{j} \cdot (Q d_{j})}{d_{j} \cdot (Qd_{j})}, \label{eq:cg-beta} \\
	d_{j+1} & = - g_{j+1} + \beta_{j} d_{j} . \label{eq:cg-d}
	\end{align}
	\end{subequations}
\ENDFOR
\ENSURE $x_{n}$.
\end{algorithmic}
\end{algorithm}
Note that the CG method converges in at most $n$ iterations provided all computations are exact, and the dominant per-iteration computational cost is a matrix-vector multiplication $Qd_{j}$ in \eqref{eq:cg-alpha}, which is reused to compute other quantities in the $j$th iteration.
In practice, we often get very close to the true solution of $Qx = b$ in $\sqrt{n}$ iterations, depending on the condition number\index{Condition number} of $Q$ (the ratio of the largest and smallest eigenvalues of $Q$).
If the condition number is too large, there are also many pre-conditioning modifications to the CG method such that the ill-conditioning issue can be mitigated \cite{nocedal2006numerical}.
If the Hessian is sparse, the computational cost can be further reduced.
Hence, CG is a practically viable approach to estimate $\vkN$ by solving \eqref{eq:newton-v-eq}.

In summary, while the Newton's method is computationally intractable in its original form for large-scale problems, there are many workarounds that produce \emph{inexact Newton's methods}\index{Inexact Newton's method} which are efficient for such problems.
The example using CG to estimate the Newton step above, which is called the Newton--CG\index{Newton--CG method} method, is one of such methods.
We remark that, these inexact Newton's methods generally do not compute or store the Hessian matrix, but only some Hessian-vector multiplications. Therefore, they are often termed as Hessian-free methods despite they use second-order derivatives of $f$.

There is another class of Hessian-free methods known as \emph{quasi-Newton method}\index{Quasi-Newton method}, which are often considered as the first choice for solving smooth non-convex optimization problems of small to moderate sizes.
The main idea of quasi-Newton methods is as follows.
Suppose the iterate $x_{k+1}$ is close to $x_{k}$, then we have
\begin{align*}
\nabla f(x_{k}) 
& = \nabla f(x_{k+1}) + \nabla^{2} f(x_{k+1}) (x_{k} - x_{k+1}) + o(|x_{k} - x_{k+1}|) \\
& \approx \nabla f(x_{k+1}) + \nabla^{2} f(x_{k+1}) (x_{k} - x_{k+1}) .
\end{align*}
Rearranging this relation and assume the approximation is accurate, we obtain
\begin{equation}
\label{eq:secant-motivation}
\nabla^{2} f(x_{k+1}) (x_{k+1} - x_{k}) = \nabla f(x_{k+1}) - \nabla f(x_{k}) .
\end{equation}
If we could find a matrix $B_{k} \in \mathbb{R}^{n \times n}$ to mimic the role of the Hessian $\nabla^{2} f(x_{k})$ for every $k$, then we can rewrite \eqref{eq:secant-motivation} as 
\begin{equation}
\label{eq:secant}
B_{k+1} (\underbrace{x_{k+1} - x_{k}}_{=:\, s_{k}} ) = \underbrace{\nabla f(x_{k+1}) - \nabla f(x_{k}) }_{=:\, y_{k}} .
\end{equation}
The equation \eqref{eq:secant} is called the \emph{secant equation}\index{Secant equation}, and optimization algorithms satisfying \eqref{eq:secant} are called \emph{quasi-Newton} methods.
If such $B_{k}$ is available at each iteration, then the quasi-Newton method has the following update rule:
\begin{equation}
\label{eq:quasi-newton-update}
x_{k+1} = x_{k} - \alpha_{k} B_{k}^{-1} \nabla f(x_{k}), 
\end{equation}
where the step size $\alpha_{k}$ is typically found by line search such that it satisfies the Wolfe condition \eqref{eq:armijo} and \eqref{eq:line-search-curvature}.

In practice, a working quasi-Newton method\index{Quasi-Newton method} does not only require \eqref{eq:secant} but also that $B_{k}$ can be easily updated and its inverse (or $B_{k}^{-1}b$ for any $b \in \mathbb{R}^{n}$) is cheap to compute.
%
% There are several sophisticated quasi-Newton methods that satisfy all these conditions.
%
The Broyden--Fletcher--Goldfarb--Shannon (BFGS)\index{BFGS method} method is one of the most notable quasi-Newton methods that satisfy all these conditions.
The BFGS update rule of $B_{k}$ is (notice that $y_{k}$ is a row vector and $s_{k}$ is a column vector):
\begin{equation}
\label{eq:bfgs}
B_{k+1} = B_{k} + \frac{y_{k}^{\top} y_{k}}{y_{k}s_{k}} - \frac{(B_{k}s_{k})(B_{k}s_{k})^{\top}}{s_{k}^{\top}(B_{k}s_{k})}, 
\end{equation}
where all denominators are scalars and numerators are matrices of rank 1.

In BFGS implementations, we also need to update $H_{k}:= B_{k}^{-1}$ in every iteration so that \eqref{eq:quasi-newton-update} can be computed easily using $H_{k}$. To this end, we recall the Sherman--Morrison--Woodbury\index{Sherman--Morrison--Woodbury formula} formula: If a matrix $A \in \mathbb{R}^{n \times n}$ is invertible, and two column vectors $u,v \in \mathbb{R}^{n}$ satisfy $1 + v^{\top} A u \ne 0$, then
\begin{equation}
\label{eq:sherman-morrison-woodbury}
( A + u v^{\top})^{-1} = A^{-1} - \frac{(A^{-1} u)(v^{\top} A^{-1})}{1 + v^{\top} A^{-1} u} .
\end{equation}
This formula can be verified by direct calculation.

By applying the Sherman--Morrison--Woodbury formula \eqref{eq:sherman-morrison-woodbury} repeatedly, we obtain the update rule of $H_{k}$:
\begin{align*}
H_{k+1} 
& = B_{k+1}^{-1} \\
& = \Big( B_{k} + \frac{y_{k}^{\top} y_{k}}{y_{k}s_{k}} - \frac{(B_{k}s_{k})(B_{k}s_{k})^{\top}}{s_{k}^{\top}(B_{k}s_{k})} \Big)^{-1} \\
& = H_{k} + \Big( 1 + \frac{y_{k} H_{k} y_{k}^{\top}}{y_{k}s_{k}} \Big) \frac{s_{k}s_{k}^{\top}}{y_{k}s_{k}} - \frac{H_{k}(s_{k}y_{k}^{\top}) + (s_{k}y_{k}^{\top}) H_{k}}{y_{k}s_{k}} .
\end{align*}
This is the update rule of $H_{k}$ in the BFGS method\index{BFGS method}.

A drawback of the BFGS method (and all quasi-Newton methods) is that they need to store a possibly dense $n \times n$ matrix $H_{k}$ (even if the true Hessian is sparse), which hinders their application to large-scale problems. 
There is a variant called limited-memory BFGS (L-BFGS) that uses curvature information from only the most recent iterations (between 3 and 20) to construct the approximation of $H_{k}$. 
We refer interested readers to \cite{nocedal2006numerical} for more details.
When an iterate $x_{k}$ is sufficiently close to a local minimizer $x^{*}$ with $\nabla^{2} f(\xs) \succ 0$, the BFGS method is shown to converge at a superlinear rate\index{Convergence!superlinear}, i.e., 
\begin{equation*}
\lim_{k\to\infty}\frac{|x_{k+1}-\xs|}{|x_{k}-\xs|} = 0 .
\end{equation*} 
By contrast, the L-BFGS method has linear convergence rate, namely,
\begin{equation*}
\lim_{k\to\infty}\frac{|x_{k+1}-\xs|}{|x_{k}-\xs|} = \gamma
\end{equation*}
for some constant $\gamma \in (0,1)$ \cite{nocedal2006numerical}.

\subsection{Algorithms for Constrained Optimization}

In this subsection, we present several algorithms for constrained optimization\index{Constrained optimization}.
We remark that they are rarely used in network training because we often avoid to form such large-scale optimization problems as constrained ones which are much more challenging.
However, these algorithms provide inspiring ideas of leveraging unconstrained optimization algorithms for solving constrained ones.
They can be useful when we want to enforce additional constraint and regularization on network parameters or design adaptive network architectures.

A major class of approaches to solving constrained optimization problems is replacing the problem with a sequence of unconstrained optimization problems. These unconstrained problems are designed such that their solutions converge to a solution of the constrained one.
We first present a commonly used algorithm in this case.

\subsubsection*{Quadratic Penalty Method}

We consider optimization problems with equality constraints \eqref{eq:min-eq-constraint}, which we recall below:
\begin{subequations}
\label{eq:min-eq-constraint-alg}
\begin{align}
\min_{x \in \mathbb{R}^{n} } \quad & f(x) , \label{subeq:min-eq-constraint-alg-f} \\
\text{s.t.} \quad & h(x) = 0 . \label{subeq:min-eq-constraint-alg-h}
\end{align}
\end{subequations}
The first algorithm we introduce is called the \emph{quadratic penalty method}\index{Quadratic penalty method}.
This method modifies the constrained problem \eqref{eq:min-eq-constraint-alg} as an unconstrained problem
\begin{equation}
\label{eq:min-quad-penalty}
\min_{x \in \mathbb{R}^{n}} Q_{\gamma}(x) ,
\end{equation}
where the \emph{quadratic penalty function}\index{Quadratic penalty function} $Q_{\gamma}$ is defined by
\begin{equation}
\label{eq:quad-penalty-fn}
Q_{\gamma} (x) : = f(x) + \frac{1}{2 \gamma} | h(x) |^{2}
\end{equation}
and $\gamma > 0$ is called the \emph{penalty parameter}.
This penalty parameter $\gamma$ is used to penalize the violation of the constraint $h(x) = 0$: The smaller $\gamma$ is, the heavier the penalty is. 
Therefore, it makes sense to set a sequence of penalty parameters $\{\gamma_{k}\}$ such that $\gamma_{k} \to 0^{+}$ as $k \to \infty$. Meanwhile, for each $k$, we solve the unconstrained minimization problem \eqref{eq:min-quad-penalty} with $\gamma = \gamma_{k}$ and some initial guess to obtain a solution $x_{k}$, and expect that the sequence of solutions $\{x_{k}\}$ converges to a solution of the original constrained problem \eqref{eq:min-eq-constraint-alg}.
If we choose $\gamma_{k}$ and an initial guess wisely, it is possible that the unconstrained problem $\min_{x}Q_{\gamma_{k}}$ can be solved quickly.

Now we describe the convergence property of the aforementioned quadratic penalty method. 

\begin{theorem}
\label{thm:quad-penalty-alg-global}
Suppose $x_{k}$ is the global minimizer of $Q_{\gamma_{k}}$ and that $\gamma_{k} \to 0^{+}$. Then every accumulation point\index{Accumulation point}\footnote{We call $\hat{x}$ an accumulation point of the sequence $\{x_{k}\}$ if $\{x_{k}\}$ has a subsequence, say $\{x_{k_{j}}\}$, converging to $\hat{x}$ as $j \to \infty$. An accumulation point is also called a limit point.} $x^{*}$ of $\{ x_{k} \}$ is a global minimizer\index{Minimizer!global} of the constrained problem \eqref{eq:min-eq-constraint-alg}.
\end{theorem}

\begin{proof}
Let $\bar{x}$ be a global minimizer of \eqref{eq:min-eq-constraint-alg}, i.e., $h(\bar{x}) = 0$ and
\begin{equation*}
f(\bar{x}) \le f(x)
\end{equation*}
for all $x$ satisfying $h(x) = 0$. On the other hand, since $x_{k}$ is a global minimizer of $Q_{\gamma_{k}}$ on $\mathbb{R}^{n}$, we know $Q_{\gamma_{k}}(x_{k}) \le Q_{\gamma_{k}}(\bar{x})$, i.e., 
\begin{equation}
\label{eq:quad-penalty-alg-opt}
f(x_{k}) + \frac{1}{2 \gamma_{k}} |h(x_{k})|^{2} \le f(\bar{x}) + \frac{1}{2 \gamma_{k}} | h(\bar{x}) |^{2} = f(\bar{x}) ,
\end{equation}
or equivalently 
\begin{equation*}
|h(x_{k})|^{2} \le 2 \gamma_{k} (f(\bar{x}) - f(x_{k}) ) .
\end{equation*}
As it is assumed that $f$ is lower bounded by some finite $f^{*}$, we have
\begin{equation*}
|h(x_{k})|^{2} \le 2 \gamma_{k} ( f(\bar{x}) - f(x_{k}) ) \le 2 \gamma_{k} (f(\bar{x}) - f^{*} ) ,
\end{equation*}
where $f(\bar{x}) - f^{*}$ is a finite constant.

Let $x^{*}$ be an accumulation point of $\{ x_{k} \}$, then there exists a subsequence $\{ x_{k_{j}} \}$ of $\{ x_{k} \}$, such that $x_{k_{j}} \to x^{*}$ as $j \to \infty$. 

Due to the continuity of $h$, we obtain
\begin{equation*}
| h(x^{*} ) |^{2} = \lim_{j \to \infty} |h(x_{k_{j}})|^{2} \le \lim_{j} 2 \gamma_{k_{j}} (f(\bar{x}) - f^{*}) = 0 ,
\end{equation*}
because $\gamma_{k_{j}} \to 0$ as $j \to \infty$. Therefore $h(x^{*}) = 0$, which implies that $x^{*}$ is a feasible point.

Furthermore, from \eqref{eq:quad-penalty-alg-opt}, we have that $f(x_{k}) \le f(\bar{x})$ for all $k$.
Since $f$ is continuous, we have 
\begin{equation*}
f(x^{*}) = \lim_{j \to \infty} f(x_{k_{j}}) \le f(\bar{x}) .
\end{equation*}
Since $\bar{x}$ is a global minimizer, we know $f(x^{*}) = f(\bar{x})$ and $x^{*}$ is also a global minimizer.
\end{proof}

While Theorem \ref{thm:quad-penalty-alg-global} makes a very desirable convergence property, it is very difficult to obtain a global minimizer of the non-convex function $Q_{\gamma}$ in practice.

Next, we present a more practical algorithm that does not need global minimizer of $Q_{\gamma_{k}}$ for any $k$, but still ensures convergence to a Lagrangian point\index{Lagrangian!point} (a point satisfies the Lagrange conditions but not necessarily a global minimizer) of \eqref{eq:min-eq-constraint-alg} under certain conditions. This algorithm is presented in Algorithm \ref{alg:quad-penalty}\index{Quadratic penalty method}.

\begin{algorithm}[t]
\caption{Quadratic penalty method for solving \eqref{eq:min-eq-constraint-alg}}
\label{alg:quad-penalty}
\begin{algorithmic}[1]
\REQUIRE Initial penalty parameter $\gamma_{0}>0$, tolerance $\epsilon_{0}$ and guess $x_{0,0} \in \mathbb{R}^{n}$.
\FOR{$k=0,1,2,\dots$}
	\STATE Start from $x_{k,0}$, find $x_{k}$ such that $| \nabla Q_{\gamma_{k}}(x_{k}) | \le \epsilon_{k}$.
	\IF{$x_{k}$ is ``satisfactory''}
		\STATE Terminate.
	\ENDIF
	\STATE Choose $\gamma_{k+1} \in (0, \gamma_{k})$.
	\STATE Choose $\epsilon_{k+1} \in (0, \epsilon_{k})$.
	\STATE Choose an initial guess $x_{k+1,0}$ for the next iteration.
\ENDFOR
\ENSURE $x_{k}$.
\end{algorithmic}
\end{algorithm}

\begin{remark}
We have several remarks about Algorithm \ref{alg:quad-penalty}. First, it is practical that $x_{k}$ in line 2 of Algorithm \ref{alg:quad-penalty} can be found provided certain smoothness conditions on $f$ and $h$. 
More specifically, notice that $\gamma_{k}>0$ and 
\begin{equation*}
\nabla Q_{\gamma_{k}}(x) = \nabla f(x) + \frac{1}{\gamma_{k}} h(x)^{\top} \nabla h(x) .
\end{equation*}
If both $f$ and $h^{\top}\nabla h$ are Lipschitz continuous, then so is $Q_{\gamma_{k}}$. Thus, there exists an algorithm, such as the gradient descent method with backtracking line search (Algorithm \ref{alg:gd-backtracking}), that can find $x_{k}$ in finitely many steps from any initial guess $x_{k,0}$. 
Note that, however, $h^{\top} \nabla h(x)$ may not be Lipschitz continuous in many cases, e.g., $h(x) = x^{2}$. It is Lipschitz if $h$ is an affine function though.

Second, we often choose $\gamma_{k}$ to be decreasing and $\epsilon_{k}$ non-increasing, but eventually need $\gamma_{k} \to 0^{+}$ and $\epsilon_{k} \to 0^{+}$ as $k \to \infty$. A commonly used method is to scale $\gamma_{k}$ and $\epsilon_{k}$ by some prescribed constants in $(0,1)$.

Third, a typical choice of the initial guess is $x_{k+1,0} = x_{k}$. But better choices may exist.

Last but not the least, we notice that Algorithm \ref{alg:quad-penalty} is terminated when $x_{k}$ is ``satisfactory'' which sounds very subjective. In practice, we terminate when all of $\gamma_{k}$, $|h(x_{k})|$, and $| \nabla Q_{\gamma_{k}}(x_{k}) |$ (all these quantities can be evaluated) are sufficiently small.

\end{remark}

We see that Algorithm \ref{alg:quad-penalty} is a promising approach: As $\epsilon_{k} \to 0$, we have 
\begin{equation*}
| \nabla Q_{\gamma_{k}} (x_{k}) |  = \Big| \nabla f(x_{k}) + \frac{1}{\gamma_{k}} h(x_{k})^{\top} \nabla h(x_{k}) \Big| \le \epsilon_{k} .
\end{equation*}
If we denote $\lambda_{k} := h(x_{k})^{\top} / \gamma_{k}$, then we may have $\nabla f(x_{k}) + \lambda_{k} \nabla h(x_{k}) $ close to $0$. Therefore, if $(x_{k}, \lambda_{k})$ has an accumulation point $(x^{*}, \lambda^{*})$, then we would have
\begin{equation*}
\nabla f(x^{*}) + \lambda^{*} \nabla h(x^{*}) = \lim_{j \to \infty} \nabla f(x_{k_{j}}) + \lambda_{k_{j}} \nabla h(x_{k_{j}}) = 0 
\end{equation*}
for some subsequence $\{(x_{k_{j}}, \lambda_{k_{j}})\}$, and hence $(x^{*}, \lambda^{*})$ satisfies the first Lagrange condition\index{Lagrange!condition}.
Hence, it remains to show $h(x^{*}) = 0$, i.e., the second Lagrange condition.
The following theorem provides a complete result.

\begin{theorem}
\label{thm:quad-penalty-alg-convergence}
Suppose $\epsilon_{k} \to 0^{+}$ and $\gamma_{k} \to 0^{+}$ as $k \to \infty$ in Algorithm \ref{alg:quad-penalty}, then for any accumulation point $x^{*}$ of $\{ x_{k} \}$ and $x^{*}$ is regular, we have that $x^{*}$ is a Lagrangian point of \eqref{eq:min-eq-constraint-alg}. Moreover, for any subsequence $\{ x_{k_{j}} \}$ of $\{ x_{k} \}$ that converges to $x^{*}$, denote $\lambda_{k_{j}} := \nabla h(x_{k_{j}}) / \gamma_{k_{j}} $, then $\lambda_{k_{j}}$ also converges to some $\lambda^{*}$ as $j \to \infty$, and $(x^{*}, \lambda^{*})$ satisfies the Lagrange conditions.
\end{theorem}

\begin{proof}
Let $\{ x_{k_{j}} \}$ be any subsequence of $\{ x_{k} \}$ such that $x_{k_{j}} \to x^{*} $ as $j \to \infty$.
Recall that $\nabla Q_{\gamma} (x) = \nabla f(x) + \frac{1}{\gamma} h(x)^{\top} \nabla h(x)$ for any fixed $\gamma > 0$. 
By the property of $x_{k}$ in line 1 of Algorithm \ref{alg:quad-penalty}, we have
\begin{equation*}
\Big| \nabla f(x_{k_{j}}) + \frac{1}{\gamma_{k_{j}}} h(x_{k_{j}})^{\top} \nabla h(x_{k_{j}}) \Big| \le \epsilon_{k_{j}} ,
\end{equation*}
which implies that 
\begin{equation*}
| h(x_{k_{j}})^{\top} \nabla h(x_{k_{j}}) | \le \gamma_{k_{j}} ( \epsilon_{k_{j}} + | \nabla f(x_{k_{j}}) | ) .
\end{equation*}
By the continuity of $f$, $h$, and $\nabla h$, as well as $x_{k_{j}} \to x^{*}$, $\epsilon_{k_{j}} \to 0$, $\gamma_{k_{j}} \to 0$ as $j \to \infty$, we know 
\begin{equation*}
| h(x^{*})^{\top} \nabla h(x^{*}) | = \lim_{j\to \infty} | h(x_{k_{j}})^{\top} \nabla h(x_{k_{j}}) | = 0 .
\end{equation*}
Since $x^{*}$ is a regular point, i.e., $\nabla h(x^{*})$ has full row rank, we know $h(x^{*}) = 0$ and hence $x^{*}$ is a feasible point. 

Next, we show that $\lambda_{k_{j}}$ converges to some $\lambda^{*}$ as $j \to \infty$.
To this end, we notice that $\nabla h$ and the determinant $\det(\cdot)$ are continuous functions, and $\nabla h(x^{*})$ has full row rank. 
Therefore, $x^{*}$ is an interior point of the set
\begin{equation*}
\{x \in \mathbb{R}^{n} : \nabla h(x) \text{ has full row rank}\} = \{ x\in \mathbb{R}^{n} : \det(\nabla h(x) \nabla h(x)^{\top} ) > 0 \} .
\end{equation*}
As a result, for any sufficiently large $j$, we know $\nabla h(x_{k_{j}}) \nabla h(x_{k_{j}})^{\top}$ has full rank and hence is invertible. 

We let $\lambda_{k_{j}} : = h(x_{k_{j}})^{\top} / \gamma_{k_{j}}$. Then there is
\begin{equation}
\label{eq:quad-penalty-alg-pf-lambda}
\lambda_{k_{j}} \nabla h(x_{k_{j}}) = \nabla Q_{\gamma_{k_{j}}} (x_{k_{j}}) - \nabla f(x_{k_{j}}).
\end{equation}
As $\nabla h(x_{k_{j}}) \nabla h(x_{k_{j}})^{\top}$ is invertible, we have
\begin{equation*}
\lambda_{k_{j}} = \big(\nabla Q_{\gamma_{k_{j}}} ( x_{k_{j}} ) - \nabla f(x_{k_{j}}) \big) \nabla h(x_{k_{j}})^{\top} \big( \nabla h(x_{k_{j}}) \nabla h(x_{k_{j}}) \big)^{-1}.
\end{equation*}
Taking $j \to \infty$, we have $|\nabla Q_{\gamma_{k_{j}}}(x_{k_{j}})| \le \epsilon_{k_{j}} \to 0$ as $j\to \infty$ and 
\begin{align*}
\lim_{j \to \infty} \lambda_{k_{j}} 
& = \lim_{j \to \infty} \Big(\nabla Q_{\gamma_{k_{j}}} ( x_{k_{j}} ) - \nabla f(x_{k_{j}}) \Big) \nabla h(x_{k_{j}})^{\top} \Big( \nabla h(x_{k_{j}}) \nabla h(x_{k_{j}})^{\top} \Big)^{-1} \\
& = - \nabla f(x^{*}) \nabla h(x^{*})^{\top} (\nabla h(x^{*}) \nabla h(x^{*})^{\top})^{-1} \\
& =: \lambda^{*} .
\end{align*}
This implies that $\{ \lambda_{k_{j}} \}$ converges to $\lambda^{*}$ defined above.

Moreover, taking $j \to \infty$ in \eqref{eq:quad-penalty-alg-pf-lambda}, we obtain
\begin{equation*}
\nabla f(x^{*}) + \lambda^{*} \nabla h(x^{*}) = 0 .
\end{equation*}
Hence $(x^{*}, \lambda^{*})$ satisfies the Lagrange conditions \eqref{eq:lagrange-condition-eq-constraint}.
\end{proof}

\begin{remark}
The condition that $x^{*}$ is regular in Theorem \ref{thm:quad-penalty-alg-convergence} is necessary, or otherwise we cannot get $h(x^{*}) = 0$, which means that $x^{*}$ may not be feasible. However, we can still show that $h(x^{*})^{\top} \nabla h(x^{*}) = 0$, which means that $x^{*}$ is a stationary point of $|h(x)|^{2}$.
\end{remark}

We also need to be aware of the ill-conditioning issue when we minimize $Q_{\gamma_{k}}$ as $\gamma_{k}$ tends to $0$. Notice that
\begin{equation*}
\nabla Q_{\gamma_{k}}(x) = \nabla f(x) + \frac{1}{\gamma_{k}} h(x)^{\top} \nabla h(x) = \nabla f(x) + \frac{1}{\gamma_{k}} \sum_{i=1}^{m} h_{i}(x) \nabla h_{i}(x)
\end{equation*}
and
\begin{equation*}
\nabla^{2} Q_{\gamma_{k}}(x) = \nabla^{2} f(x) + \frac{1}{\gamma_{k}} \sum_{i=1}^{m} h_{i}(x) \nabla^{2} h_{i}(x) + \frac{1}{\gamma_{k}} \nabla h(x)^{\top} h(x) .
\end{equation*}
If we substitute $x$ by $x_{k_{j}}$ and $\gamma_{k}$ by $\gamma_{k_{j}}$, and notice that $x_{k_{j}} \to x^{*}$ and $\lambda_{k_{j}} = h(x_{k_{j}})^{\top}/\gamma_{k_{j}} \to \lambda^{*}$ as $j \to \infty$, we would expect that $\nabla^{2} Q_{\gamma_{k_{j}}} (x_{k_{j}})$ approaches
\begin{equation}
\label{eq:quad-penalty-Q-Hessian}
\nabla^{2} f(x^{*}) + \sum_{i=1}^{m} \lambda_{i}^{*} \nabla^{2} h_{i}(x^{*}) + \lim_{j\to\infty} \frac{1}{\gamma_{k_{j}}} \nabla h(x_{k_{j}})^{\top} \nabla h(x_{k_{j}}) .
\end{equation}
The sum of the first two terms approaches the Hessian of the Lagrangian function $l(x,\lambda) = f(x) + \lambda h(x)$ with respect to $x$ at $(x^{*},\lambda^{*})$, i.e., $\partial_{x}^{2} l(x,\lambda)$.
This Hessian has fixed eigenvalues, which are positive if $\xs$ is a strict local minimizer. 
However, the last term in \eqref{eq:quad-penalty-Q-Hessian} is troublesome: We know that $\nabla h(x_{k_{j}})^{\top} \nabla h(x_{k_{j}})$ approaches $\nabla h(x^{*})^{\top} \nabla h(x^{*}) \in \mathbb{R}^{n\times n}$, which typically is singular since $m < n$. In this case, $1/\gamma_{k_{j}}$ will scale up all the positive eigenvalues, making the condition number of $\nabla^{2}Q_{\gamma_{k_{j}}}$ very large at $x^{*}$ as $\gamma_{k_{j}} \to 0$.
This means that minimizing $Q_{\gamma_{k_{j}}}$ can be very difficult in practice.
There is a remedy to mitigate this issue; see \cite[Chapter 17.1]{nocedal2006numerical}.

\subsubsection*{Lagrangian Method}

We consider an alternative method for equality constrained optimization\index{Constrained optimization!equality constrained} problems based on the Lagrangian function\index{Lagrangian!function}:
\begin{equation}
l(x, \lambda) = f(x) + \lambda h(x),
\end{equation}
where $f, h \in C^{2}$. The \emph{Lagrangian method}\index{Lagrangian!method} is given by
\begin{subequations}
\label{eq:lagrange-method}
\begin{align}
x_{k+1} & = x_{k} - \alpha_{k} (\nabla f(x_{k}) + \lambda_{k} \nabla h(x_{k}))^{\top} , \label{subeq:lagrange-method-x} \\
\lambda_{k+1} & = \lambda_{k} + \beta_{k} h(x_{k})^{\top}, \label{subeq:lagrange-method-lambda}
\end{align}
\end{subequations}
where $k$ is the iteration counter and $\alpha_{k}, \beta_{k} > 0$ are step sizes.
The transposes ${}^{\top}$ are to make vector shapes consistent to our convention, i.e., $x \in \mathbb{R}^{n}$ is a column vector and $\lambda \in \mathbb{R}^{m}$ is a row vector.

Before we discuss the property of the Lagrangian method \eqref{eq:lagrange-method}, we show an intuitive way to memorize this method as follows.
For any $x \in \mathbb{R}^{n}$, we notice that
\begin{equation*}
\max_{\lambda \in \mathbb{R}^{m}} l(x,\lambda) = 
\begin{cases}
f(x), & \text{if } h(x) = 0, \\
+\infty, & \text{if } h(x) \ne 0.
\end{cases}
\end{equation*}
Therefore, we can rewrite the equality constrained optimization problem \eqref{eq:min-eq-constraint-alg} as
\begin{equation*}
\min_{x \in \mathbb{R}^{n}} \{ f(x) : h(x) = 0 \} = \min_{x \in \mathbb{R}^{n}} \max_{\lambda \in \mathbb{R}^{m}} l(x,\lambda) .
\end{equation*}
Therefore, the Lagrangian method \eqref{eq:lagrange-method} can be thought of as gradient descent for minimizing $l(x,\lambda)$ with respect to $x$ and gradient ascent for maximizing it with respect to $\lambda$.
Since only gradients and Jacobians are needed in \eqref{eq:lagrange-method}, we call it a first-order Lagrangian method.

Now we want to examine the convergence property of the Lagrangian method\index{Lagrangian!method} \eqref{eq:lagrange-method}. We simplify the analysis by assuming constant step sizes $\alpha_{k}=\alpha$ and $\beta_{k} = \beta$ for some $\alpha, \beta > 0$. By rescaling $\lambda$, we can also assume $\beta = \alpha$ without loss of generality.
In this case, the Lagrangian method \eqref{eq:lagrange-method} can be thought of as a \emph{fixed point iteration}\index{Fixed point iteration} of form $z_{k+1} = F(z_{k})$, where $z_{k} := (x_{k}, \lambda_{k}) \in \mathbb{R}^{n+m}$ is a column vector that concatenates $x$ and $\lambda$, and 
\begin{equation}
\label{eq:lagrange-method-F}
F(z) : = \begin{pmatrix}
x - \alpha ( \nabla f(x) + \lambda \nabla h(x))^{\top} \\
\lambda^{\top} + \alpha h(x)
\end{pmatrix} 
\end{equation}
for any $z = (x, \lambda) \in \mathbb{R}^{n+m}$. 
Here, by fixed point of $F$, we meant a point $z$ satisfying $z = F(z)$. 
With this definition, we have the following lemma, for which the proof is trivial and hence omitted here.
\begin{lemma}
\label{lem:lagrangian-method-fp}
A point $z^{*} = (x^{*}, \lambda^{*})$ satisfies the Lagrange conditions\index{Lagrange!condition}
\begin{subequations}
\label{eq:lagrange-method-condition}
\begin{align}
\nabla f(x^{*}) + \lambda^{*} \nabla h(x^{*}) & = 0 , \label{eq:lagrange-method-condition-x} \\
h(x^{*}) & = 0 \label{eq:lagrange-method-condition-lambda}
\end{align}
\end{subequations}
if and only if $z^{*}$ is a fixed point of $F$.
\end{lemma}

Next, we show that if $(x^{*}, \lambda^{*})$ satisfies the second-order sufficient conditions, i.e., $(x^{*}, \lambda^{*})$ satisfies \eqref{eq:lagrange-method-condition} and 
\begin{equation}
\label{eq:lagrangian-method-2nd-order}
\nabla^{2} f(x^{*}) + \sum_{i=1}^{m} \lambda_{i}^{*} \nabla^{2} h_{i}(x^{*}) \succ 0,
\end{equation}
then the sequence $\{(x_{k}, \lambda_{k})\}$ generated by the Lagrangian method \eqref{eq:lagrange-method} will converge to $(x^{*}, \lambda^{*})$ provided that the initial guess $(x_{0}, \lambda_{0})$ is close enough to $(x^{*}, \lambda^{*})$ and $\alpha$ is sufficiently small.

\begin{theorem}
Suppose $(x^{*}, \lambda^{*})$ satisfies the second-order sufficient conditions, namely, \eqref{eq:lagrange-method-condition} and \eqref{eq:lagrangian-method-2nd-order}, of the equality constrained optimization problem \eqref{eq:min-eq-constraint-alg} and $\alpha,\beta > 0$ are sufficiently small, then there exists a neighborhood of $(x^{*}, \lambda^{*})$, such that for any initial $(x_{0}, \lambda_{0})$ in this neighborhood, the sequence $\{(x_{k}, \lambda_{k})\}$ generated by the Lagrangian method \eqref{eq:lagrange-method} converges to $(x^{*}, \lambda^{*})$.
\end{theorem}

\begin{proof}
As we mentioned above, we assume $\alpha = \beta$ by properly rescaling $\lambda$. Notice that the Lagrangian method is $z_{k+1} = F(z_{k})$. By Lemma \ref{lem:lagrangian-method-fp}, we just need to show that $z_{k}$ converges to a fixed point of $F$. Since $z^{*} = (x^{*}, \lambda^{*})$ is a fixed point of $F$, there is
\begin{equation*}
|z_{k+1} - z^{*} | = |F(z_{k}) - F(z^{*})| .
\end{equation*}
Notice that
\begin{equation*}
\nabla F(z) = I + \alpha 
\begin{pmatrix}
-H(x, \lambda) & - \nabla h(x)^{\top} \\
\nabla h(x) & 0 
\end{pmatrix} \in \mathbb{R}^{(n+m) \times (n+m)} ,
\end{equation*}
where
\begin{equation*}
H(x,\lambda) := \nabla^{2} f(x) + \sum_{i=1}^{m} \lambda_{i} \nabla^{2} h_{i}(x) .
\end{equation*}
On the other hand, by the Mean Value Theorem, there exists a matrix-valued function $M$ of $z_{k}$ and $z^{*}$ ($M$ depends on $F$) such that
\begin{equation*}
F(z_{k}) - F(z^{*}) = M(z_{k}, z^{*}) (z_{k} - z^{*}) ,
\end{equation*}
where $(M(z_{k},z^{*}))_{i}$, the $i$th row of $M(z_{k}, z^{*})$, is the value of $\nabla F_{i}$ at some point in the line segment connecting $z_{k}$ and $z^{*}$. Taking norms on both sides, we obtain
\begin{equation}
\label{eq:lagrange-method-pf-iter-ineq}
|F(z_{k}) - F(z^{*})| \le | M(z_{k}, z^{*})| | z_{k} - z^{*} | .
\end{equation}

Now we claim that for $\alpha > 0$ sufficiently small, there is $| \nabla F(z^{*}) | < 1$.
To see this, we notice that $\nabla F(z^{*}) = I + \alpha M^{*}$ where
\begin{equation*}
M^{*} : = 
\begin{pmatrix}
-H(x^{*}, \lambda^{*}) & - \nabla h(x^{*})^{\top} \\
\nabla h(x^{*}) & 0
\end{pmatrix} .
\end{equation*}
Therefore, it suffices to show that all eigenvalues of $M^{*}$ have negative real parts, since that ensures all eigenvalues of $\nabla F(z^{*}) = I + \alpha M^{*}$ have real part in $(0,1)$ when $\alpha > 0$ is sufficiently small.

Now let $\xi$ be any eigenvalue of $M^{*}$ and $z = (x, \lambda)$ is the corresponding eigenvector.
(Note that by the definition of eigenvectors $z \ne 0$.) 
Then $M^{*} z = \xi z$. Multiplying $z^{\top}$\footnote{Note that since $z$ can be a complex vector, the notation ${}^{\top}$ stands for complex conjugate transpose.}, we obtain $z^{\top} M^{*} z = \xi |z|^{2}$. Hence 
\begin{equation}
\label{eq:lagrange-method-pf-real-part}
\text{Re}(z^{\top} M^{*} z) = \text{Re}(\xi) |z|^{2}
\end{equation}
where $\text{Re}(\cdot)$ stands for the real part of the argument. 
In addition, we notice that
\begin{align}
\text{Re}(z^{\top} M^{*} z) 
& = - \text{Re}(x^{\top} H(x^{*}, \lambda^{*}) x ) - \text{Re}(x^{\top} \nabla h(x^{*})^{\top} \lambda^{*}) \nonumber \\
& \qquad \qquad + \text{Re}(\lambda^{*\top}\nabla h(x^{*}) x)  \label{eq:lagrange-method-pf-real-part-matrix} \\
& = - \text{Re}(x^{\top} H(x^{*}, \lambda^{*}) x ) . \nonumber 
\end{align}
Since $H(x^{*}, \lambda^{*}) \succ 0$ due to \eqref{eq:lagrangian-method-2nd-order}, we know that $\text{Re}(x^{\top} H(x^{*}, \lambda^{*}) x) > 0$ as long as $x \ne 0$.

Now we see that, if $x\ne 0$, then by comparing \eqref{eq:lagrange-method-pf-real-part} and \eqref{eq:lagrange-method-pf-real-part-matrix} and noticing $z \ne 0$, there is $\text{Re}(\xi) < 0$.

If $x = 0$, then $z = (0,\lambda)$. By comparing the first $n$ components of the two sides of $M^{*} z = \xi z$, we have $\nabla h(x^{*})^{\top} \lambda = 0$. Since $x^{*}$ is a regular point, we know $\nabla h(x^{*})$ has full row rank and therefore $\lambda = 0$. However, this implies $z = 0$, which is a contradiction to $z \ne 0$. 
Now we know all eigenvalues of $M^{*}$ have negative real part.

At this point, we know that for $\alpha> 0$ sufficiently small, there is
\begin{equation*}
|\nabla F(z^{*}) | = | I + \alpha M^{*} | < 1 .
\end{equation*} 
By the continuity of $\nabla F$ and norms, there exist $r > 0$ and $\kappa \in (0,1)$ such that $M(z, z^{*}) \le \kappa$ for all $z \in \bar{B}_{r}(z^{*}) := \{z : | z - z^{*} | \le r \}$.

If $z_{0} = (x_{0},\lambda_{0}) \in \bar{B}_{r}(z^{*})$, then by \eqref{eq:lagrange-method-pf-iter-ineq} there is
\begin{equation*}
|z_{1} - z^{*} | \le | M(z_{0}, z^{*}) | \, | z_{0} - z^{*} | \le \kappa |z_{0} - z^{*} | \le r,
\end{equation*}
and hence $|M(z_{1}, z^{*}) | \le \kappa \in (0,1)$ as well.
If $z_{k} \in \bar{B}_{r}(z^{*})$, then by the same way we have
\begin{equation*}
| z_{k+1} - z^{*} | \le \kappa |z_{k} - z^{*} | \le r
\end{equation*}
and $|M(z_{k+1}, z^{*})| \le \kappa \in (0,1)$. By induction, we know $z_{k}$ converges to $z^{*}$ at a linear rate\index{Convergence!linear}.
\end{proof}

\subsubsection*{Augmented Lagrangian Method}

In Section \ref{subsec:sosc}, we introduced a variant of the Lagrange function, called the \emph{augmented Lagrangian function}, to derive the second-order sufficient conditions of constrained optimization\index{Constrained optimization!equality constrained} problems.
There is also an associated method, called the \emph{augmented Lagrangian method}\index{Augmented Lagrangian method}, to solve constrained optimization, which we discuss next.

It is worth noting that the augmented Lagrangian method is also related to the quadratic penalty method. 
The difference is that, with the Lagrange multiplier, the augmented Lagrangian method can effectively mitigate the ill-conditioning issue of the quadratic penalty method.

To gain some inspiration about the construction of the augmented Lagrangian method, we first recall the equality constrained optimization problem:
\begin{subequations}
\label{eq:min-eq-constraint-alm}
\begin{align}
\min_{x \in \mathbb{R}^{n}} \quad & f(x) , \label{subeq:min-eq-constraint-alm-f} \\
\text{s.t.} \quad & h(x) = 0 . \label{subeq:min-eq-constraint-alm-h}
\end{align}
\end{subequations}
We also recall that, in Theorem \ref{thm:quad-penalty-alg-convergence} where we showed the convergence of the quadratic penalty method, the accumulation point of the sequence $\{ h(x_{k})^{\top} / \gamma_{k} \}$ is in fact the Lagrange multiplier $\lambda^{*}$ corresponding to the accumulation point $x^{*}$ of the sequence $\{x_{k}\}$.
Hence, we would expect $h(x_{k}) \approx \gamma_{k} \lambda^{*}$ for $k$ sufficiently large.
In the quadratic penalty method, we need $\gamma_{k} \to 0^{+}$ (in order to obtain $h(x_{k}) \to 0$) which causes the ill-conditioning of $\nabla^{2} Q_{\gamma_{k}}$. 
In the augmented Lagrangian method, this issue can be mitigated so that $\gamma_{k}$ does not need to approach to $0$ very fast while still ensuring that $h(x_{k})$ tends to $0$ quickly.

To see this, we recall that the augmented Lagrangian function\index{Augmented Lagrangian function} is defined by the following ``hybrid'' of the Lagrangian function and the quadratic penalty function corresponding to the problem \eqref{eq:min-eq-constraint-alm}:
\begin{equation}
\label{eq:alm-fn}
l_{A}(x, \lambda; \gamma) := f(x) + \lambda h(x) + \frac{1}{2\gamma} |h(x)|^{2} .
\end{equation}
Taking gradient of $l_{A}$ with respect to $x$, we have
\begin{align*}
\nabla_{x} l_{A} (x, \lambda; \gamma) 
& = \nabla f(x) + \lambda \nabla h(x) + \frac{1}{\gamma} h(x)^{\top} \nabla h(x) \\
& = \nabla f(x) + \Big( \lambda + \frac{1}{\gamma} h(x)^{\top} \Big) \nabla h(x) .
\end{align*}
Suppose $k$ is sufficiently large, and we use $x_{k}$ to denote the minimizer of $l_{A}(x, \lambda_{k}; \gamma_{k})$, then we would expect to get
\begin{equation*}
\lambda_{k} + \frac{h(x_{k})^{\top}}{\gamma_{k}} \approx \lambda^{*} ,
\end{equation*}
or equivalently
\begin{equation*}
h(x_{k})^{\top} \approx \gamma_{k} (\lambda^{*} - \lambda_{k}) ,
\end{equation*}
when $(x_{k},\lambda_{k})$ is close to $(\xs, \lambda^{*})$. 
If $\lambda_{k} \to \lambda^{*}$, then $\lambda^{*} - \lambda_{k} \to 0$, and therefore we may not need $\gamma_{k} \to 0$ as fast as in the quadratic penalty method. This also suggests that we make the following update
\begin{equation}
\label{eq:alm-lambda-update}
\lambda_{k+1} = \lambda_{k} + \frac{h(x_{k})^{\top}}{\gamma_{k}} 
\end{equation}
every time after we update $x_{k}$ by
\begin{equation}
\label{eq:alm-x-update}
x_{k} = \argmin_{x \in \mathbb{R}^{n}} \ l_{A}(x, \lambda_{k}; \gamma_{k}) .
\end{equation}
If we keep the setting of the quadratic penalty method, i.e., $\gamma_{k} \to 0^{+}$,  and assume that $\{\lambda_{k}\}$ is bounded, then we can prove that the augmented Lagrangian method\index{Augmented Lagrangian method} also ensures convergence, as shown below.

\begin{proposition}
\label{prop:alm-global}
Let $(\bar{x}, \bar{\lambda})$ be any accumulation point of $\{(x_{k}, \lambda_{k})\}$ generated by \eqref{eq:alm-lambda-update} and \eqref{eq:alm-x-update} with $\gamma_{k} \to 0$. Then $\bar{x}$ is a global minimizer\index{Minimizer!global} of \eqref{eq:min-eq-constraint-alm}.
\end{proposition}

\begin{proof}
Denote $f^{*} := \min \{ f(x) : h(x) = 0, \ x \in \mathbb{R}^{n} \}$. Then by \eqref{eq:alm-x-update} for every $k$, we have
\begin{equation*}
l_{A} (x_{k}, \lambda_{k}; \gamma_{k}) \le l_{A} (x, \lambda_{k}; \gamma_{k}) 
\end{equation*}
for any $x$.
Taking the infimum of the right-hand side over all $x$ satisfying $h(x) = 0$, we obtain
\begin{equation}
\label{eq:limit-pt-alm}
l_{A}(x_{k}, \lambda_{k}; \gamma_{k}) \le f^{*} .
\end{equation}
Suppose the subsequence $\{(x_{k_{j}}, \lambda_{k_{j}}) \}$ converges to $(\bar{x}, \bar{\lambda})$ as $j \to \infty$. Then by taking $j \to \infty$ we obtain
\begin{equation}
\label{eq:alm-limit-pt-f-upper}
f(\bar{x}) + \bar{\lambda} h(\bar{x}) + \lim_{j \to \infty} \frac{1}{2\gamma_{k_{j}}} |h(x_{k_{j}})|^{2} \le f^{*} .
\end{equation}
Notice that $f(\bar{x}) + \bar{\lambda} h(\bar{x})$ is some constant. Since $\gamma_{k} \to 0$ and $|h(x_{k})| \ge 0$, we must have
\begin{equation*}
|h(\bar{x})|^{2} = \lim_{j \to \infty} |h(x_{k_{j}})|^{2}  = 0 ,
\end{equation*}
which implies that $h(\bar{x}) = 0$ and thus $\bar{x}$ is feasible.
Moreover, from \eqref{eq:alm-limit-pt-f-upper}, we have $f(\bar{x}) \le f^{*}$. Therefore $\bar{x}$ is a global minimizer of \eqref{eq:min-eq-constraint-alm}.
\end{proof}

While the convergence result given in Proposition \ref{prop:alm-global}, we notice that \eqref{eq:alm-x-update} is difficult to implement in practice since $l_{A}(\cdot, \lambda_{k}; \gamma_{k})$ can be non-convex and we are generally unable to obtain its global minimizer $x_{k}$.
As an alternative, we can approximate the critical points of $l_{A}(\cdot, \lambda_{k}; \gamma_{k})$ for all $k$. We summarize the augmented Lagrangian method\index{Augmented Lagrangian method} in Algorithm \ref{alg:alm}. In the next proposition, we show that if $\{x_{k}\}$ generated by Algorithm \ref{alg:alm} converges a regular point $\xs$, then this $\xs$ satisfies the Lagrange conditions of the problem \eqref{eq:min-eq-constraint-alm}.
\begin{algorithm}[t]
\caption{Augmented Lagrangian method for \eqref{eq:min-eq-constraint-alm}}
\label{alg:alm}
\begin{algorithmic}[1]
\REQUIRE Initial $\gamma_{0}>0$, $\epsilon_{0} >0$, $x_{0,0} \in \mathbb{R}^{n}$, $\lambda_{0} \in \mathbb{R}^{m}$.
\FOR{$k=0,1,2,\dots$}
	\STATE Start from $x_{k,0}$, find $x_{k}$ such that $| \nabla_{x} l_{A}(x_{k},\lambda_{k}; \gamma_{k}) | \le \epsilon_{k}$.
	\IF{$x_{k}$ is ``satisfactory''}
		\STATE Terminate.
	\ENDIF
	\STATE Set $\lambda_{k+1} = \lambda_{k} + \frac{\nabla h(x_{k})^{\top}}{\gamma_{k}}$.
	\STATE Choose $\gamma_{k+1} \in (0, \gamma_{k})$.
	\STATE Choose $\epsilon_{k+1} \in (0, \epsilon_{k})$.
	\STATE Choose an initial guess $x_{k+1,0}$ for the next iteration.
\ENDFOR
\ENSURE $x_{k}$.
\end{algorithmic}
\end{algorithm}

\begin{proposition}
\label{prop:alm-local}
Let $f, h \in C^{1}$, $\{ \lambda_{k} \}$ be bounded, and $\gamma_{k}, \epsilon_{k} \to 0^{+}$. Assume $\{x_{k}\}$ satisfies $| \nabla_{x} l_{A} (x_{k}, \lambda_{k}; \gamma_{k} ) | \le \epsilon_{k} $ for all $k$ and $x_{k} \to x^{*}$ where $x^{*}$ is regular (note that we do not assume $x^{*}$ is a local minimizer). Then $h(x^{*}) = 0$ and $\tilde{\lambda}_{k} := \lambda_{k} + \frac{h(x_{k})^{\top}}{\gamma_{k}}  \to \lambda^{*}$, where $(x^{*}, \lambda^{*})$ satisfies the Lagrange conditions\index{Lagrange!condition} \eqref{eq:lagrange-method-condition}. 
\end{proposition}

\begin{proof}
As $x_{k} \to \xs$ and $\xs$ is a regular point, i.e., $\nabla h(\xs)$ has full row rank, we assume that $\nabla h(x_{k})$ has full row rank and $\nabla h(x_{k}) \nabla h(x_{k})^{\top}$ is invertible for all $k$.
Notice that
\begin{align}
\nabla_{x} l_{A}(x_{k}, \lambda_{k}; \gamma_{k}) 
& = \nabla f(x_{k}) + \lambda_{k} \nabla h(x_{k}) + \frac{1}{\gamma_{k}} h(x_{k})^{\top} \nabla h(x_{k}) \nonumber \\
& = \nabla f(x_{k}) + \tilde{\lambda}_{k} \nabla h(x_{k}) \label{eq:alm-pf-dl} ,
\end{align}
where
\begin{equation*}
\tilde{\lambda}_{k} := \lambda_{k} + \frac{h(x_{k})^{\top}}{\gamma_{k}}.
\end{equation*}
Right-multiplying $\nabla h(x_{k})^{\top} (\nabla h(x_{k}) \nabla h(x_{k})^{\top})^{-1}$ to both sides of \eqref{eq:alm-pf-dl} and rearranging, we obtain
\begin{equation*}
\tilde{\lambda}_{k} = (\nabla_{x} l_{A}(x_{k}, \lambda_{k}; \gamma_{k}) - \nabla f(x_{k}) ) \nabla h(x_{k})^{\top} (\nabla h(x_{k}) \nabla h(x_{k})^{\top})^{-1} .
\end{equation*}
Since $x_{k} \to x^{*}$ and $|\nabla_{x} l_{A}(x_{k}, \lambda_{k}; \gamma_{k})| \to 0$ as $k \to \infty$, we know 
\begin{equation*}
\tilde{\lambda}_{k} \to \lambda^{*} := - \nabla f(\xs) \nabla h(\xs)^{\top} ( \nabla h(\xs) \nabla h(\xs)^{\top})^{-1} .
\end{equation*}
Taking limit \eqref{eq:alm-pf-dl} for $k \to \infty$, we know that 
\begin{equation*}
\nabla f(x^{*}) + \lambda^{*} \nabla h(x^{*}) = 0 .
\end{equation*}

Since $\lambda_{k}$ is bounded and $\lambda_{k} + \frac{h(x_{k})^{\top}}{\gamma_{k}} = \tilde{\lambda}_{k} \to \lambda^{*}$, we know the sequence $\{ \frac{h(x_{k})^{\top}}{\gamma_{k}} \}$ is bounded.
Since $\gamma_{k} \to 0^{+}$, we know $h(x_{k}) \to 0$.
As $x_{k} \to x^{*}$ and $h$ is continuous, we have $h(x^{*}) = 0$ .
\end{proof}

We notice that Proposition \ref{prop:alm-local} still requires $\gamma_{k} \to 0$. In practice, there are three possible situations as follows.

The first is that, for some $k$, we are not able to find $x_{k}$ that satisfies $| \nabla_{x} l_{A} (x_{k}, \lambda_{k}; \gamma_{k}) | \le \epsilon_{k}$. This usually happens when $l_{A} (\cdot, \lambda_{k}; \gamma_{k}) $ is not lower bounded.

The second situation is that $\{x_{k}\}$ satisfies $| \nabla_{x} l_{A} (x_{k}, \lambda_{k}; \gamma_{k}) | \le \epsilon_{k} $ for every $k$ but $\{ x_{k} \}$ does not converge, or it converges to a point $x^{*}$ that is infeasible. The usually happens when the original problem (equality constrained) does not have feasible solution. In this case, $\{x_{k}\}$ converges to an infeasible $x^{*}$, which is a stationary point of $| h(x) |^{2}$, i.e., $h(x^{*})^{\top} \nabla h(x^{*}) = 0$. 

The third situation is that $\{x_{k}\}$ exists and converges to point $x^{*}$ which is a local minimizer but not regular, i.e., $\nabla h(x^{*})$ does not have full row rank. In this case $\nabla h(x^{*}) \nabla h(x^{*})^{\top}$ is not invertible, and $\lambda_{k} + \frac{h(x_{k})}{\gamma_{k}}$ may not converge.

We remark that the augmented Lagrangian method may not work efficiently sometimes. This is usually because $\nabla_{xx}^{2} l_{A}(\cdot, \lambda_{k}; \gamma_{k})$ has very large condition number as $\gamma_{k} \to 0$.

\paragraph{Application to inequality constrained optimization}
We can apply the augmented Lagrangian method to optimization problems with both equality and inequality constraints\index{Constrained optimization!inequality constrained}. To this end, we can convert the inequality constraints to equality constraints. This can be done by introducing the \emph{squared slack variables} as follow.

Consider the general optimization problem with equality and inequality constraints
\begin{subequations}
\label{eq:alm-eq-ineq}
\begin{align}
\min_{x \in \mathbb{R}^{n}} \quad & f(x) , \label{subeq:alm-eq-ineq-f} \\
\text{s.t.} \quad & h(x) = 0 , \label{subeq:alm-eq-ineq-h} \\
& g_{j}(x) \le 0, \quad j = 1,\dots, m, \label{subeq:alm-eq-ineq-g}
\end{align}
\end{subequations}
We can write \eqref{eq:alm-eq-ineq} in an equivalent form 
\begin{subequations}
\label{eq:alm-eq-ineq-slack}
\begin{align}
\min_{x \in \mathbb{R}^{n}, y \in \mathbb{R}^{m}} \quad & f(x) , \label{subeq:alm-eq-ineq-slack-f} \\
\text{s.t.} \quad & h(x) = 0 , \label{subeq:alm-eq-ineq-slack-h} \\
& g_{j}(x) + y_{j}^{2} = 0, \quad j = 1,\dots, m, \label{subeq:alm-eq-ineq-slack-g}
\end{align}
\end{subequations}
by introducing the squared slack variables $y=(y_{1},\dots,y_{m})^{\top} \in \mathbb{R}^{m}$.
The augmented Lagrangian function\index{Augmented Lagrangian function} of \eqref{eq:alm-eq-ineq-slack} is 
\begin{equation}
\label{eq:alm-eq-ineq-slack-lagrangian}
l_{A}(x, y, \lambda, \mu; \gamma) = f(x) + \lambda h(x) + \frac{1}{2\gamma} |h(x)|^{2} + \mu \tilde{g}(x,y) + \frac{1}{2 \gamma} |\tilde{g}(x,y)|^{2} ,
\end{equation}
where
\begin{equation*}
\tilde{g}(x,y) := \Big( \tilde{g}_{1}(x,y_{1}), \dots, \tilde{g}_{m}(x,y_{m}) \Big)^{\top} \in \mathbb{R}^{m}
\end{equation*}
and
\begin{equation*}
\tilde{g}_{j}(x,y_{j}) := g_{j}(x) + y_{j}^{2} , \quad \text{for} \ j = 1,\dots, m .
\end{equation*}
When $\lambda$, $\mu$, and $\gamma$ are fixed, we need to solve the unconstrained optimization subproblem of $(x,y)$:
\begin{equation}
\label{eq:alm-eq-ineq-slack-lagrangian-subproblem}
\min_{x,y} \ l_{A}(x, y, \lambda, \mu; \gamma) .
\end{equation}
If we fix $x$ in \eqref{eq:alm-eq-ineq-slack-lagrangian-subproblem}, then $y$ has a closed form solution: First, the components of $y=(y_{1},\dots,y_{m})$ are separated and each $y_{j}$ solves
\begin{equation*}
\min_{y_{j}} \ \mu_{j} (g_{j}(x) + y_{j}^{2}) + \frac{1}{2 \gamma} ( g_{j}(x) + y_{j}^{2})^{2} ,
\end{equation*}
which is equivalent to
\begin{equation*}
\min_{z_{j} \ge 0} \ \mu_{j} (g_{j}(x) + z_{j}) + \frac{1}{2 \gamma} ( g_{j}(x) + z_{j})^{2} .
\end{equation*}
We can complete the square of $z_{j}$ above, eliminate the constant independent of $z_{j}$, and obtain an equivalent optimization problem:
\begin{equation*}
\min_{z_{j} \ge 0} \ \frac{1}{2\gamma} z_{j}^{2} + \Big( \frac{1}{\gamma} \nabla g_{j}(x)^{\top} + \mu_{j} \Big) z_{j} ,
\end{equation*}
which has its minimizer in a closed form:
\begin{equation*}
z_{j}^{*} = \max(0, -(g_{j}(x) + \gamma \mu_{j})) .
\end{equation*}
Note that $z_{j}^{*}$ depends on $x$, $\mu_{j}$ and $\gamma$. Denote
\begin{equation*}
g_{j}^{+} (x, \mu_{j}; \gamma) := g_{j}(x) + z_{j}^{*} = \max( g_{j}(x) , - \gamma \mu_{j} ) .
\end{equation*}
Substituting $y_{j}^{2} = z_{j}^{*}$ back into the augmented Lagrangian function \eqref{eq:alm-eq-ineq-slack-lagrangian}, we obtain a simplified augment Lagrangian function without $y$:
\begin{align*}
l_{A}(x, \lambda, \mu; \gamma) 
& := f(x) + \lambda h(x) + \frac{1}{2 \gamma } |h(x)|^{2} \\
& \qquad +\sum_{j=1}^{m} \Big( \mu_{j} g_{j}^{+} (x, \mu; \gamma) + \frac{1}{2\gamma} (g_{j}^{+}(x, \mu; \gamma) )^{2} \Big) .
\end{align*}
We can further simplify $l_{A}(x, \lambda, \mu; \gamma)$ by completing the squares in the sum:
\begin{align*}
\mu_{j} g_{j}^{+} (x, \mu; \gamma) + \frac{1}{2\gamma} (g_{j}^{+}(x, \mu; \gamma) )^{2}
& = \frac{1}{2\gamma} \Big( (g_{j}^{+}(x, \mu_{j}; \gamma) + \gamma \mu_{j})^{2} - \gamma^{2} \mu_{j}^{2} \Big) \\
& = \frac{1}{2 \gamma} \Big( \max(g_{j}(x) + \gamma \mu_{j}, 0 )^{2} - \gamma^{2} \mu_{j}^{2} \Big) \\
& = \frac{\gamma}{2} \Big( \max \Big(\frac{1}{\gamma} g_{j}(x) + \mu_{j}, 0 \Big)^{2} -  \mu_{j}^{2} \Big) .
\end{align*}
The remainder is the same as the augmented Lagrangian method (Algorithm \ref{alg:alm}) with $(\lambda, \mu)$ as the multiplier.
However, we notice that the problem is $\max(\cdot,0)^{2}$ is not twice differentiable. This can be a problem when we minimize $l_{A}(\cdot, \lambda_{k}, \mu_{k}; \gamma_{k})$ to find $x_{k}$. 

Nevertheless, following the analysis above, we still expect to have
\begin{equation*}
\lambda_{k} + \frac{1}{\gamma_{k}} h(x_{k})^{\top} \to \lambda^{*} 
\end{equation*}
and
\begin{equation*}
\mu_{k} + \frac{1}{\gamma_{k}} g^{+} (x_{k}, \mu_{k}; \gamma_{k}) = \max \Big( \mu_{k} + \frac{1}{\gamma_{k}} g(x_{k}) , 0 \Big) \to \mu^{*}
\end{equation*}
as $k \to \infty$, where the maximum is taken componentwisely, and $(\lambda^{*}, \mu^{*})$ is the Lagrange multiplier corresponding to $h(x^{*}) = 0$ and $g(x^{*}) \le 0$. Notice that $\mu^{*} \ge 0$, which is consistent with the KKT conditions.

\section{Stochastic Optimization Algorithms}
\label{sec:stochastic-opt}

Deep learning often requires large scale of data samples, and as a consequence the objective functions are usually formulated in distinctive settings where stochastic optimization\index{Stochastic optimization} algorithms become more adaptive than conventional deterministic ones.
In this section, we present consider stochastic optimization algorithms, including their formulations, motivations, advantages and drawbacks, and several widely used algorithms in network training nowadays.
In particular, we focus on their applications in solving non-convex unconstrained optimization problems which are the prevalent formulation in training neural networks.

\subsection{Fundamentals of Stochastic Optimization}
\label{subsec:fundamentals-so}

Recall that, in the standard supervised learning problem, we are often given a dataset $\Dcal = \{(x_{i}, y_{i}) \in \mathbb{R}^{d} \times \mathbb{R}^{m}: i \in [N]\}$ consisting of $N$ sample pairs, where $N$ can be very large like $10^{6}$ or greater.
Then our goal is to find the mapping $u: \mathbb{R}^{d} \to \mathbb{R}^{m}$, parameterized as a deep neural network $u_{\theta}$ with parameter $\theta \in \mathbb{R}^{n}$, such that $u_{\theta}(x_{i})$ is approximately equal to $y_{i}$ for all $i \in [N]$. 
Suppose for simplicity we set the loss function\index{Loss function} $\ell: \mathbb{R}^{n} \to \mathbb{R}$ of $\theta$ as the averaged squared differences between $u_{\theta}(x_{i})$ and $y_{i}$, and find the optimal $\theta$ by solving
\begin{equation}
\label{eq:soa-eloss-theta}
\min_{\theta \in \mathbb{R}^{n}}\ \ell(\theta) := \frac{1}{N} \sum_{i=1}^{N} f_{i}(\theta), \quad \text{where} \quad f_{i}(\theta) = \frac{1}{2} |u_{\theta}(x_{i}) - y_{i}|^{2} .
\end{equation}
Since the gradient $\nabla_{\theta} f_{i}(\theta)$ and the Hessian $\nabla_{\theta}^{2} f_{i}(\theta)$ can be calculated by automatic differentiation as we explained in Section \ref{sec:autodiff}, we know how to compute the gradient and Hessian of $\ell(\theta)$.
Therefore, in theory, we may apply any of the unconstrained optimization algorithms introduced in Section \ref{subsec:alg-uncon}, such as the gradient descent method, the Newton's method, and the quasi-Newton methods, to find a (local) minimizer of $\ell$. 
However, it turns out to be infeasible, or at least inefficient empirically, to employ a deterministic optimization algorithm with a loss function of form \eqref{eq:soa-eloss-theta}. Let us dive into this and see the reasons next.

To be consistent with the notations used throughout this chapter, we still denote by $x$ the variable to be optimized (instead of $\theta$) and by $F_{N}$ the loss function to indicate that it is the average of $N$ terms, and write the optimization problem \eqref{eq:soa-eloss-theta} as
\begin{equation}
\label{eq:soa-eloss}
\min_{x \in \mathbb{R}^{n}} \ F_{N}(x) := \frac{1}{N} \sum_{i=1}^{N} f_{i}(x).
\end{equation}
We remark that $F_{N}$ is also called the \emph{finite-sum} objective, which is a kind of objective function\index{Objective function} often appears in deep learning problems. 
It is a special case of the general form of (unconstrained) stochastic optimization problem, which is typically of form
\begin{equation}
\label{eq:soa-loss}
\min_{x \in \mathbb{R}^{n}} \ F(x) := \mathbb{E}_{\xi} [f(x; \xi)] ,
\end{equation}
where $\xi$ is a random variable whose sample (realization) may be one or a subset of samples from $\Dcal$ consisting of $N$ samples independently drawn from some (possibly unknown) data probability distribution.

For example, if we have a dataset $\Dcal=\{z_{i}: i \in [N]\}$ called a \emph{batch}, where each sample $z_{i}$ renders a function $f_{i}$ of $x$, and $\xi$ is set to be a sample drawn from $\Dcal$ with uniform distribution, then \eqref{eq:soa-loss} reduces to \eqref{eq:soa-eloss}.
One can also interpret \eqref{eq:soa-eloss} as an approximation of \eqref{eq:soa-loss} by Monte Carlo integration (see Appendix \ref{appsec:mc}) with dataset $\Dcal$.
The loss functions $F$ and $F_{N}$ are often called the \emph{expected loss}\index{Loss function!expected} and \emph{empirical loss}\index{Loss function!empirical}, respectively.
We generally focus on minimizing $F_{N}$ as \eqref{eq:soa-eloss} in the remainder of this section, and draw some connections to  the problem of minimizing $F$ as in \eqref{eq:soa-loss} when needed.

If we consider solving \eqref{eq:soa-eloss} as a deterministic optimization problem, then the standard gradient descent method proceeds as follows in the $k$th iteration:
\begin{equation}
\label{eq:batch-gd}
x_{k+1} = x_{k} - \frac{\alpha_{k}}{N} \sum_{i=1}^{N} \nabla f_{i}(x_{k}) ,
\end{equation}
where $\alpha_{k} > 0$ is the step size.
We call \eqref{eq:batch-gd} the \emph{batch gradient}\index{Batch gradient method} (or \emph{full gradient}) method.
This is effectively applying the gradient descent method to \eqref{eq:soa-eloss}.

To obtain insights about stochastic optimization, we consider the basic stochastic update for solving \eqref{eq:soa-eloss} as follows in the $k$th iteration:
\begin{equation}
\label{eq:basic-sgd}
x_{k+1} = x_{k} - \alpha_{k} \nabla f_{i_{k}}(x) ,
\end{equation}
where $i_{k}$ is an index randomly drawn from $\{1,\dots, N\}$ and independent of any other $i_{j}$'s for all $j < k$.
We call \eqref{eq:basic-sgd} the \emph{basic stochastic gradient}\index{Basic stochastic gradient} method hereafter.
Note that, $x_{k}$ in \eqref{eq:basic-sgd} depends not only on $F_{N}$, the initial guess $x_{0}$, and the step sizes $\alpha_{k}$'s, but also the random sequence $\{i_{j}: j < k\}$. Indeed, $\{x_{k}: k \ge 1\}$ can be thought of as a stochastic process.

By comparing the batch gradient \eqref{eq:batch-gd} and the basic stochastic gradient \eqref{eq:basic-sgd}, we may have the feeling that the former, deterministic optimization approach is more appealing.
This is understandable as there are many sophisticated deterministic optimization algorithms available, and we can expect to obtain satisfactory convergence rate and solution quality by employing these algorithms.
Moreover, the structure of $F_{N}$ in \eqref{eq:soa-eloss} allows us to leverage parallel computing technique which can significantly reduce per-iteration computational time in practice.

However, the stochastic gradient method \eqref{eq:basic-sgd} has proven to be more efficient in deep learning applications from multiple perspectives \cite{bottou2018optimization}.
For example, we consider a network training problem where the given dataset $\Dcal$ is poorly constructed: $\Dcal$ consists of ten identical copies of a small dataset $\Dcal_{\mathrm{sub}}$.
(Although we often try to avoid such situation, it often occurs that the data samples collected are similar and repetitive in practice. Nevertheless, we use this example just to provide an intuitive motivation that the batch gradient method may not be as effective as the basic stochastic gradient method.)
In this case, the batch gradient method generates the same iterates by just using $\Dcal_{\mathrm{sub}}$, but it literally takes 10 times of computational cost because it is using $\Dcal$ in practice. 
By contrast, the computational cost and convergence behavior remain the same for the basic stochastic gradient method.
This suggests that using all samples from $\Dcal$ as in the batch gradient method \eqref{eq:batch-gd} can be less efficient than the basic stochastic gradient method \eqref{eq:basic-sgd}.

Indeed, there have been numerous reports on the comparison of the basic stochastic gradient method and advanced deterministic optimization method in the literature. For example, some results posted by these reports show that the basic stochastic gradient method with a properly tuned constant step size dramatically outperforms the limited-memory BFGS\index{BFGS method} (L-BFGS) method, one of the most efficient and widely adopted optimization algorithm as a variant of the BFGS method (Section \ref{subsec:alg-uncon}). 
More specifically, the basic stochastic gradient method is significantly faster than L-BFGS and attains satisfactory result in the first few epochs (the two methods are compared using objective function value versus epoch number---one epoch\index{Epoch} means the method uses all samples in $\Dcal$ in expectation---which is a fair comparison that takes into account the difference of the per-iteration computational costs using the two methods) in a binary classification problem with a logistic loss function and the RCV1 dataset, while L-BFGS catches up and surpasses the basic stochastic gradient method a lot later \cite{bottou2018optimization}.

This result appears to be surprising and counter-intuitive. 
However, the following example provided by \cite{bertsekas2015convex} provides a convincing explanation of this phenomenon. 

\begin{example}
\label{ex:batch-vs-sg}
Let $N$ be an odd integer and $M = \frac{N-1}{2}$. We partition $[-1,1]$ into $2M$ segments of equal length $\frac{1}{M} = \frac{2}{N-1}$. Denote the $N=2M+1$ grid points of the partition by
\begin{equation}
\label{eq:ex-sgd-zi}
z_{i} = -1 + \frac{2(i-1)}{N-1}, \quad i = 1,\dots, N .
\end{equation}
Notice that $z_{1} = -1$ and $z_{N} = 1$. 
Now we define
\begin{equation}
\label{eq:ex-sgd-fi}
f_{i}(x) := (x-z_{i})^{2}
\end{equation}
for $i=1,\dots, N$, and
\begin{equation}
\label{eq:ex-sgd-FN}
F_{N}(x) := \frac{1}{N}\sum_{i=1}^{N} f_{i}(x) = \frac{1}{N}\sum_{i=1}^{N} (x-z_{i})^{2} .
\end{equation}
Figure \ref{fig:batch-vs-sg} provides an illustration of $z_{i}$'s and $f_{i}$'s for the case $N=5$ and $M=2$.
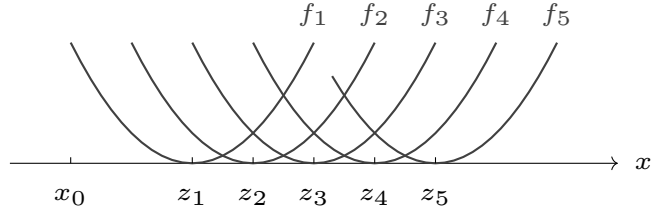
\begin{figure}[t]
\centering
\begin{tikzpicture}[scale=1.6, transform shape]
	
	\draw[->] (-2.5, 0) -- (2.5, 0) node[right] {{\tiny $x$}};
	
	\draw[thick, color=white!25!black] plot[domain=-2:0]({\x}, {\x*\x + 2*\x + 1}) node[above]{{\tiny $f_{1}$}};
	\draw[thick, color=white!25!black] plot[domain=-1.5:0.5]({\x}, {\x*\x + \x + 0.25}) node[above]{{\tiny $f_{2}$}};
	\draw[thick, color=white!25!black] plot[domain=-1:1]({\x}, {\x*\x}) node[above]{{\tiny $f_{3}$}};
	\draw[thick, color=white!25!black] plot[domain=-0.5:1.5]({\x}, {\x*\x - \x + 0.25}) node[above]{{\tiny $f_{4}$}};
	\draw[thick, color=white!25!black] plot[domain=0.15:2]({\x}, {\x*\x - 2*\x + 1}) node[above]{{\tiny $f_{5}$}};	
	
	\draw (-2,0) -- (-2,0.05);
	
	\draw (-1,0) -- (-1,0.05);
	\draw (-0.5,0) -- (-0.5,0.05);
	\draw (0,0) -- (0,0.05);
	\draw (0.5,0) -- (0.5,0.05);
	\draw (1,0) -- (1,0.05);
			
	\draw[black] (-2, -0.08) node[below]{{\tiny $x_{0}$}};
	\draw[black] (-1, -0.08) node[below]{{\tiny $z_{1}$}};
	\draw[black] (-0.5, -0.08) node[below]{{\tiny$z_{2}$}};
	\draw[black] (0, -0.08) node[below]{{\tiny$z_{3}$}};
	\draw[black] (0.5, -0.08) node[below]{{\tiny$z_{4}$}};
	\draw[black] (1, -0.08) node[below]{{\tiny$z_{5}$}};
\end{tikzpicture}
\caption{Illustration of $z_{i}$ in \eqref{eq:ex-sgd-zi} and $f_{i}$ in \eqref{eq:ex-sgd-fi} with $N=5$ and $M=\frac{N-1}{2}=2$. Here $z_{1},\dots,z_{N}$ are the grid points of the equal partition of $[-1,1]$ into $2M=4$ segments, with $z_{1}=-1$ and $z_{N} = 1$. The functions $f_{i}(x) = (x-z_{i})^{2}$ are quadratic functions shown by the curves, and the average $F_{N}(x):=\frac{1}{N} \sum_{i=1}^{N}f_{i}(x) = x^{2} + \frac{1}{N} \sum_{i=1}^{N}z_{i}^{2}$ (not shown in this figure) has a unique minimizer at $x=0$ (also the point $z_{3}$ in this plot). We can examine the convergence behavior when applying the batch gradient descent method \eqref{eq:batch-gd} and the basic stochastic gradient method \eqref{eq:basic-sgd} with initial guess $x_{0}$. The basic stochastic gradient descent method converges to the minimizer $x=0$ at the beginning with lower computational cost than the batch gradient descent method, but it can get confused when $x_{k}$ is close to $x=0$ as it only has the gradient of a randomly selected $f_{i}$ in each iteration.}
\label{fig:batch-vs-sg}
\end{figure}

Notice that $f_{i}(x) = (x-z_{i})^{2}$ is a quadratic function with minimum value at $z_{i}$. Meanwhile, the average $F_{N}$ is also a simple quadratic function:
\begin{equation}
\label{eq:ex-batch-FN}
F_{N}(x) = x^{2} + \underbrace{\frac{1}{N} \sum_{i=1}^{N} z_{i}^{2}}_{\text{constant}} , 
\end{equation}
which has a unique minimum value at $x=0$.

If we apply the batch gradient descent method \eqref{eq:batch-gd} or more advanced deterministic optimization algorithm like L-BFGS to \eqref{eq:ex-sgd-FN} starting from an initial guess $x_{0}$ far away from the minimizer $0$ (see Figure \ref{fig:batch-vs-sg}), they are effectively minimizing \eqref{eq:ex-batch-FN} and can generate a sequence of iterates $\{x_{k}\}$ that stably converges to the minimizer $x=0$.
By contrast, if we use the basic stochastic gradient method \eqref{eq:basic-sgd} with this initial $x_{0}$, we can expect that the generated sequence of iterates $\{x_{k}\}$ still converge to $0$ rapidly at in the first few epochs but at a much lower per-iteration computational cost: The batch gradient descent method computes the gradients of all $f_{i}$'s, and it is $N$ times of the cost by the basic stochastic gradient descent method which only computes the gradient of a randomly selected $f_{i}$ in each iteration.

This example also suggests that the basic stochastic gradient method may only ensure convergence to a neighborhood of the minimizer if a constant step size is used. 
In other words, to ensure convergence to the minimizer, a diminishing step size policy may have to be employed.

From theoretical perspective, we can also provide an estimate on the upper bounds of the total computational costs required by batch gradient descent and basic stochastic gradient for minimizing $F_{N}$.
Notice that $F_{N}$ is a strongly convex smooth function, the batch gradient descent is effectively the standard gradient descent method and known to have a linear convergence rate.
More precisely, one can show that the batch gradient method converges at a linear rate\index{Convergence!linear}:
\begin{equation*}
F_{N}(x_{k}) - F_{N}^{*}  \le O(\gamma^{k}) ,
\end{equation*}
where $F_{N}^{*}> \infty$ is the (possibly unknown) minimum value of $F_{N}$ over $\mathbb{R}$ and $\gamma \in (0,1)$ depends on the ratio $L/\mu \ge 1$.
Here $L$ is the Lipschitz constant of $\nabla F_{N}$ and $\mu$ is the strong convexity parameter of $F_{N}$, i.e.,
\begin{equation*}
F_{N}(\hat{x}) \ge F_{N}(x) + \nabla F_{N}(x) (\hat{x} - x) + \frac{\mu}{2} | \hat{x} - x |^{2} 
\end{equation*}
for all $x, \hat{x} \in \mathbb{R}^{n}$.
We know that $F_{N}$ is more ill-conditioned and $\gamma$ is closer to 1 if $L/\mu$ is larger. 
(In this particular example with $F_{N}$ defined in \eqref{eq:ex-batch-FN}, $L=\mu=1$. However, we remark that the ratio $L/\mu$ can be very large even if $F_{N}$ is strongly convex in real-world applications.)

Due to the linear convergence rate above, one can easily deduce that to obtain an $\epsilon$-optimal solution $x_{k}$ of $F_{N}$, namely,
\begin{equation*}
F_{N}(x_{k}) - F_{N}^{*} \le \epsilon
\end{equation*}
for any user-selected error tolerance $\epsilon > 0$, the iteration number $k$ is at the order $O(\log(1/\epsilon))$. 
As the per-iteration computational cost for the batch gradient descent method is $N$ (as it needs to compute the gradients of $f_{1},\dots,f_{N}$ to get the gradient of $F_{N}$), the total computational cost to obtain an $\epsilon$-optimal solution is proportional to $N \log (1/\epsilon)$.

On the other hand, we can check the computational cost of the basic stochastic gradient descent method \eqref{eq:basic-sgd} when applied to minimize $F_{N}$ given in \eqref{eq:ex-sgd-FN}.
As $F_{N}$ is strongly convex, one can show that the basic stochastic gradient method converges at a sublinear rate\index{Convergence!sublinear}:
\begin{equation*}
\mathbb{E}[F_{N}(x_{k})] - F_{N}^{*} = O(1/k) ,
\end{equation*}
where $x_{k}$ is the $k$th iterate generated by the basic stochastic gradient descent method. 
The convergence rate above is only sublinear, which is much slower than linear \emph{asymptotically}. 
Nevertheless, the sublinear rate above does not depend on the sample size $N$.
As a result, the number of iterations needed to achieve an $\epsilon$-optimal solution is $O(1/\epsilon)$.
In terms of per-iteration cost, we know it is $1$ as the basic stochastic gradient descent method only needs to compute the gradient of a randomly selected function among $f_{1},\dots,f_{N}$.
In summary, the total computational cost is proportional to $1/\epsilon$, which is again independent of $N$.

Now we see that computational costs needed by the batch gradient descent \eqref{eq:batch-gd} and the basic stochastic gradient descent \eqref{eq:basic-sgd} are proportional to $N \log (1/\epsilon)$ and $1/\epsilon$, respectively.
For fixed $N$, we know $N \log (1/\epsilon)$ is much smaller than $1/\epsilon$ asymptotically as $\epsilon \to 0$.
However, the practical situation is that $\epsilon > 0$ is often set to a sufficiently small number in advance, and the sample size $N$ plays an important role in the convergence rate, especially at the early stage of the optimization process.
This explains why the basic stochastic gradient descent can be even faster than L-BFGS in convergence in the first few epochs.

\end{example}

We also remark that an additional advantage of the stochastic optimization algorithms, such as the basic stochastic gradient descent method \eqref{eq:basic-sgd}, is their applicability even when minimizing an expected loss function $F$ as in \eqref{eq:soa-loss}.
By contrast, deterministic optimization algorithms, such as the batch gradient algorithm \ref{eq:batch-gd}, become infeasible since we cannot compute the exact gradient of $F$ in practice.
These advantages are the reasons why stochastic optimization algorithms have become prevalent in deep learning research and applications nowadays.

Next, we consider the convergence properties and computation complexities of stochastic gradient methods. Comparing the batch gradient method \eqref{eq:batch-gd} and the basic stochastic gradient method \eqref{eq:basic-sgd}, we see that their per-iteration computational costs are $N$ versus $1$, because \eqref{eq:batch-gd} computes $N$ gradients whereas \eqref{eq:basic-sgd} only computes one.
As a convention in deep learning, we say a method completes one epoch\index{Epoch} when it uses up all $N$ samples in the given dataset $\Dcal$ on expectation or equivalent. Therefore, the batch gradient method \eqref{eq:batch-gd} completes one epoch in one iteration, whereas the basic stochastic gradient method \eqref{eq:basic-sgd} completes one epoch in every $N$ iterations.
We will also see the so-called mini-batch stochastic gradient method\index{Mini-batch stochastic gradient} which uses $N_{k}$ samples of $\Dcal$ (called a mini-batch) in the $k$th iteration where $N_{k}$ is between 1 and $N$ (for simplicity we assume $N_{k}$ does not change in $k$), and therefore it completes one epoch every $N/N_{k}$ iterations.

\subsubsection*{Convergence of Stochastic Gradient Method}

Now we investigate the convergence behavior of stochastic gradient method. 
As we pointed out earlier, stochastic gradient methods, such as \eqref{eq:basic-sgd}, can be applied to find minimizers of either empirical loss \eqref{eq:soa-eloss} or expected loss \eqref{eq:soa-loss}.
For the latter case, we denote by $p_{\xi}$ the probability distribution of $\xi$.
Therefore, we use $F: \mathbb{R}^{n} \to \mathbb{R}$ to denote the loss function\index{Loss function} that can be either of them in the remainder of this section.
The general form of stochastic gradient (SG) method is given in Algorithm \ref{alg:sg}. We will provide more details about the steps in this algorithm next.
\begin{algorithm}
\caption{Stochastic Gradient Method for Solving \eqref{eq:soa-eloss} or \eqref{eq:soa-loss}}
\label{alg:sg}
\begin{algorithmic}[1]
\REQUIRE Initial guess $x_{1} \in \mathbb{R}^{n}$.
	\FOR{$k=1,2,\dots$}
		\STATE Generate a realization of $\xi_{k}$.
		\STATE Compute $g(x_{k}, \xi_{k})$.
		\STATE Choose a step size $\alpha_{k} > 0$.
		\STATE Set $x_{k+1} \leftarrow x_{k} - \alpha_{k} g(x_{k}, \xi_{k})$.
	\ENDFOR
% \ENSURE $x_{k}$. 
\end{algorithmic}
\end{algorithm}

We first explain how Lines 2 and 3 of Algorithm \ref{alg:sg} are implemented. 
In particular, we consider three scenarios in the implementation of SG method in either the expected loss case or the empirical loss case: The basic SG, the mini-batch SG, and the mini-batch SG with second-order information (we called it second-order mini-batch SG\index{Mini-batch stochastic gradient!second-order} for short hereafter).
We remark that, in either of the three scenarios, $\{\xi_{k}: k \ge 1\}$ is a sequence of jointly independent random variables.
Once $\xi_{k}$ is determined, the vector $g(x_{k},\xi_{k})$ is defined according to the specified SG method:
\begin{equation}
\label{eq:sgd-g}
g(x_{k}, \xi_{k}) = 
\begin{cases}
\nabla_{x} f(x_{k}; \xi_{k}) & \text{(Basic SG)} \\
\displaystyle{\frac{1}{N_{k}} \sum_{i=1}^{N_{k}} \nabla_{x} f(x_{k}; \xi_{k,i})} & \text{(Mini-batch SG)} \\
\displaystyle{H_{k} \frac{1}{N_{k}} \sum_{i=1}^{N_{k}} \nabla_{x} f(x_{k}; \xi_{k,i})} & \text{(Second-order mini-batch SG)}
\end{cases}
\end{equation}

In the basic SG scenario, we have showed the update scheme in \eqref{eq:basic-sgd} for the finite-sum case \eqref{eq:soa-eloss}.
In this case, $\xi_{k}$ refers to a sample index drawn from $\{1,\dots,N\}$ with uniform distribution.
If the sampled index is $i_{k} \in \{1,\dots,N\}$, then we have $\xi_{k} = i_{k}$ in Line 2 and $g(x_{k},\xi_{k}) = \nabla f_{i_{k}}(x_{k})$ in Line 3 of Algorithm \ref{alg:sg}.
If the basic SG is applied to the expected loss $F$ defined in \eqref{eq:soa-loss}, then $\xi_{k}$ is a sample drawn from $p_{\xi}$ and we set $g(x_{k}, \xi_{k}) = \nabla_{x} f(x_{k}, \xi_{k})$.

In the mini-batch SG scenario, $\xi_{k}$ refers to $N_{k}$ indices ($N_{k}$ is called the mini-batch size which is user-selected in advance) drawn from $\{1,\dots,N\}$ uniformly for the empirical loss case, and $g(x_{k},\xi_{k})$ is the average of the gradients of $f_{i}$ at the selected indices in $\xi_{k}$. See \eqref{eq:sgd-g} for the explicit form.
Similarly, in the expected loss case, $\xi_{k}$ is a set of $N_{k}$ samples independently drawn from $p_{\xi}$, and $g(x_{k},\xi_{k})$ is the averaged gradients on the samples specified by $\xi_{k}$.

The second-order mini-batch SG scenario is similar to the previous one except for an additional symmetric positive definite scaling matrix $H_{k}$.
This matrix is computed based on different strategies of incorporating second-order information.
A general rule on the selection of $H_{k}$ is that $-g(x_{k}, \xi_{k})$ should be a descent direction under expectation over $\xi_{k}$.
We will provide more details on the construction of $H_{k}$ later.

Now we list a few assumptions required to establish the convergence of the stochastic gradient descent method in Algorithm \ref{alg:sg}.

\begin{assumption}
\label{ass:sgd}
Suppose the loss function $F: \mathbb{R}^{n} \to \mathbb{R}$ is continuously differentiable and the following assumptions hold.
\begin{enumerate}
\item
$F^{*} := \inf_{x \in \mathbb{R}^{n}} F(x)$ is finite.

\item
The gradient $\nabla F$ is $L$-Lipschitz continuous\index{Lipschitz continuous}, namely,
\begin{equation*}
| \nabla F(\hat{x}) - \nabla F(x) | \le L | \hat{x} - x|, \quad \forall \, x, \hat{x} \in \mathbb{R}^{n} .
\end{equation*}

\item
There exist scalars $\mu$ and $\mu_{G}$ with $0 < \mu \le \mu_{G}$ such that for all $k$
\begin{align}
\nabla F(x_{k}) \cdot \mathbb{E}_{\xi_{k}}[g(x_{k},\xi_{k})] & \ge \mu |\nabla F(x_{k})|^{2} , \label{eq:sgd-descent} \\
| \mathbb{E}_{\xi_{k}}[g(x_{k}, \xi_{k})] | & \le \mu_{G} | \nabla F(x_{k}) | . \label{eq:sgd-gF} 
\end{align}

\item
There exist scalars $M, M_{G} \ge 0$ such that for all $k$
\begin{equation}
\label{eq:sgd-2nd}
\mathbb{E}_{\xi_{k}}[ |g(x_{k}, \xi_{k})|^{2}] \le M + M_{G} |\nabla F(x_{k}) |^{2} .
\end{equation}
\end{enumerate}
\end{assumption}

The first two conditions in Assumption \ref{ass:sgd} are mild and often satisfied in real-world applications.
The condition \eqref{eq:sgd-descent} requires that, in expectation on $\xi_{k}$, the angle between $g(x_{k}, \xi_{k})$ and $\nabla F(x_{k})$ should be less than $\pi/2$.
This is a typical requirement for $-g(x_{k},\xi_{k})$ to be a descent direction in expectation.
The condition \eqref{eq:sgd-gF} effectively requires the expectation (the first moment) of $g(x_{k},\xi_{k})$ is at order $O(|\nabla F(x_{k})|)$ for every $k$.
Combining with \eqref{eq:sgd-descent}, we see that $|g(x_{k},\xi_{k})|$ in expectation on $\xi_{k}$ should be proportional to $|\nabla F(x_{k})|$.
The condition \eqref{eq:sgd-2nd} requires the second moment of $g(x_{k},\xi_{k})$ to be upper bounded by some constant $M>0$ plus $O(|\nabla F(x_{k})|^{2})$ for every $k$.

Note that, if $g(x_{k},\xi_{k})$ is an unbiased estimate of $F(x_{k})$, namely, 
\begin{equation*}
\mathbb{E}_{\xi_{k}}[g(x_{k}, \xi_{k})] = \nabla F(x_{k}) ,
\end{equation*}
then both of the conditions \eqref{eq:sgd-descent} and \eqref{eq:sgd-gF} hold with $\mu = \mu_{G} = 1$.
In the standard settings of many SG methods, such as the basic SG method, it can be shown that $g(x_{k},\xi_{k})$ is indeed an unbiased estimate of $\nabla F(x_{k})$.

Under Assumption \ref{ass:sgd}, we have a simple bound on the difference between $\mathbb{E}_{\xi_{k}}[F(x_{k+1})]$ and $F(x_{k})$ at the $k$th iteration, as shown in the following lemma.

\begin{lemma}
Suppose the conditions in Assumption \ref{ass:sgd} hold true. Let $x_{k}$ be the sequence of iterates generated by the stochastic gradient method in Algorithm \ref{alg:sg} from any initial $x_{0}$. Then
\begin{equation}
\label{eq:lemma-sg-bound}
\mathbb{E}_{\xi_{k}}[F(x_{k+1})] - F(x_{k})
\le - \Big(\mu - \frac{1}{2} \alpha_{k} L M_{G}\Big) \alpha_{k} | \nabla F(x_{k}) |^{2} + \frac{1}{2} \alpha_{k}^{2} L M .
\end{equation}
\end{lemma}

\begin{proof}
Since $\nabla F$ is $L$-Lipschitz continuous, we have
\begin{align*}
F(x_{k+1}) - F(x_{k})
& \le \nabla F(x_{k}) \cdot (x_{k+1} - x_{k}) + \frac{L}{2} | x_{k+1} - x_{k} |^{2} \\
& = - \alpha_{k} \nabla F(x_{k}) \cdot g(x_{k},\xi_{k}) + \frac{\alpha_{k}^{2}L}{2} | g(x_{k},\xi_{k}) |^{2} .
\end{align*}
Taking expectation of $\xi_{k} \sim p_{\xi}$ on both sides, and noticing that $x_{k}$ is independent of $\xi_{k}$ ($\xi_{k}$ is generated independently after $x_{k}$ is computed), we obtain
\begin{align*}
\mathbb{E}_{\xi_{k}}[F(x_{k+1})] - F(x_{k})
\le - \alpha_{k} \nabla F(x_{k}) \cdot \mathbb{E}_{\xi_{k}}[g(x_{k},\xi_{k})] + \frac{\alpha_{k}^{2}L}{2} \mathbb{E}_{\xi_{k}}[| g(x_{k},\xi_{k}) |^{2}] .
\end{align*}
With the conditions \eqref{eq:sgd-descent} and \eqref{eq:sgd-2nd} applied to the first and second terms, respectively, on the right-hand side above, we obtain
\begin{align*}
\mathbb{E}_{\xi_{k}}[F(x_{k+1})] - F(x_{k})
& \le - \alpha_{k} \mu |\nabla F(x_{k})|^{2} + \frac{\alpha_{k}^{2}L}{2}\Big( M + M_{G} [| \nabla F(x_{k}) |^{2}] \Big) \\
& = - \Big(\mu - \frac{1}{2} \alpha_{k} L M_{G}\Big) \alpha_{k} | \nabla F(x_{k}) |^{2} + \frac{1}{2} \alpha_{k}^{2} L M ,
\end{align*}
which completes the proof.
\end{proof}

Now we are ready to show the convergence of stochastic gradient method in Algorithm \ref{alg:sg}.
We consider the convergence of this method in two settings of the step size $\alpha_{k}$.
The first setting is using a constant step size for all iterations, and the second setting is using pre-specified automatically diminishing step sizes.

\begin{theorem}
[SG with constant step size]
\label{thm:sg-constant-step}
Suppose the conditions in Assumption \ref{ass:sgd} hold true. Let $x_{k}$ be the sequence of iterates generated by the stochastic gradient method in Algorithm \ref{alg:sg} from any initial $x_{0}$ with a constant step size $\alpha_{k} = \alpha$ for all $k$, where
\begin{equation}
\label{eq:sg-constant-step}
0 < \alpha \le \frac{\mu}{L M_{G}} .
\end{equation}
Then there is
\begin{equation}
\label{eq:sg-constant-step-sum}
\mathbb{E}\Big[ \sum_{k=1}^{K} | \nabla F(x_{k})] |^{2} \Big]
\le \frac{K \alpha L M}{\mu} + \frac{2(F(x_{1}) - F^{*})}{\mu \alpha} .
\end{equation}
Consequently, there is
\begin{equation}
\label{eq:sg-constant-step-sum-K}
\mathbb{E}\Big[ \frac{1}{K}\sum_{k=1}^{K} | \nabla F(x_{k})] |^{2} \Big]
\le \frac{\alpha L M}{\mu} + \frac{2(F(x_{1}) - F^{*})}{\mu \alpha K} .
\end{equation}
which converges to the positive scalar $\alpha L M / \mu$ as $K \to \infty$.
\end{theorem}

\begin{proof}
Taking expectation of \eqref{eq:lemma-sg-bound} in $\xi_{1},\dots,\xi_{k}$ (notice that $x_{k}$ is determined by $\xi_{1},\dots, \xi_{k-1}$ for every $k$) and using the bound of $\alpha$ in \eqref{eq:sg-constant-step}, we obtain
\begin{align*}
\mathbb{E}[F(x_{k+1})] - \mathbb{E}[F(x_{k})]
& \le - \Big(\mu - \frac{1}{2} \alpha L M_{G}\Big) \alpha | \nabla F(x_{k}) |^{2} + \frac{1}{2} \alpha^{2} L M \\
& \le - \frac{1}{2} \mu \alpha | \nabla F(x_{k}) |^{2} + \frac{1}{2} \alpha^{2} L M .
\end{align*}
Taking the sum of both sides for $k=1,2,\dots,K$, we obtain
\begin{equation*}
F^{*} - F(x_{1}) \le \mathbb{E}[F(x_{k+1})] - F(x_{1}) \le - \frac{1}{2} \mu \alpha \sum_{k=1}^{K}| \nabla F(x_{k}) |^{2} + \frac{1}{2} \alpha^{2} L M ,
\end{equation*}
where the first inequality is due to $F(x_{k+1}) \ge F^{*}$ for all $k$.
Rearranging this inequality, we obtain \eqref{eq:sg-constant-step-sum}. Dividing both sides of \eqref{eq:sg-constant-step-sum} by $K$ yields \eqref{eq:sg-constant-step-sum-K}.
\end{proof}

We can see that the bound obtained in \eqref{eq:sg-constant-step-sum-K} is pessimistic as it does not tend to 0 as $K \to \infty$ unless $M = 0$. 
However, from \eqref{eq:sgd-2nd} we see that $M = 0$ means the second moment $\mathbb{E}_{\xi_{k}}[|g(x_{k}, \xi_{k})|^{2}]$ must decrease as fast as $|\nabla F(x_{k})|^{2}$.
Even if $g(x_{k},\xi_{k})$ is an unbiased estimate of $\nabla F(x_{k})$, this is still difficult because this effectively requires the noise (variance) in $g(x_{k}, \xi_{k})$ tends to 0, which is almost never achievable in real-world applications.

Nevertheless, we can still obtain some insights about the estimate \eqref{eq:sg-constant-step-sum-K}. The convergence to a positive number depending on $M$ and the step size $\alpha$ explains the phenomenon we saw in Example \ref{ex:batch-vs-sg} above: The basic SG method can be efficient at the beginning, but may end up lingering in a neighborhood of the minimizer $x=0$ rather than converging to $0$.
The size of this neighborhood can be reduced by using a smaller step size $\alpha$, but this obviously slows down the entire convergence and may not be worth it.

Next, we consider the convergence of the SG method in Algorithm \ref{alg:sg} with diminishing step size. 
We employ the step size rule of $\alpha_{k}$ characterized in the seminal work \cite{robbins1951stochastic}:
\begin{equation}
\label{eq:rm-step}
\sum_{k=1}^{\infty} \alpha_{k} = \infty \qquad \text{and} \qquad  
\sum_{k=1}^{\infty} \alpha_{k}^{2} < \infty .
\end{equation}
Then we obtain the following result.

\begin{theorem}
[SG with diminishing step size]
\label{thm:sg-diminish-step}
Suppose the conditions in Assumption \ref{ass:sgd} hold true. Let $x_{k}$ be the sequence of iterates generated by the stochastic gradient method in Algorithm \ref{alg:sg} from any initial $x_{0}$ with step sizes $\alpha_{k}$ following the rule \eqref{eq:rm-step}. 
Then there is
\begin{equation}
\label{eq:sg-diminish-step-sum}
\lim_{K \to \infty} \mathbb{E}\Big[ \sum_{k=1}^{K} \alpha_{k} | \nabla F(x_{k})] |^{2} \Big] < \infty ,
\end{equation}
and therefore
\begin{equation}
\label{eq:sg-diminish-step-sum-K}
\lim_{K \to \infty} \mathbb{E}\Big[ \frac{1}{A_{K}}\sum_{k=1}^{K} \alpha_{k} | \nabla F(x_{k})] |^{2} \Big] = 0 ,
\end{equation}
where $A_{K} := \sum_{k=1}^{K} \alpha_{k}$.
\end{theorem}

\begin{proof}
Notice that the condition $\sum_{k=1}^{\infty} \alpha_{k}^{2} < \infty$ in \eqref{eq:rm-step} implies that $\alpha_{k} \to 0$ as $k \to \infty$. 
Therefore, without loss of generality, we assume $\alpha_{k} \le \frac{\mu}{L M_{G}}$ for all $k$. 
Then, by taking expectation of \eqref{eq:lemma-sg-bound} in $\xi_{1},\dots,\xi_{k}$, we obtain
\begin{align*}
\mathbb{E}[F(x_{k+1})] - \mathbb{E}[F(x_{k})]
& \le - \Big(\mu - \frac{1}{2} \alpha_{k} L M_{G}\Big) \alpha_{k} | \nabla F(x_{k}) |^{2} + \frac{1}{2} \alpha_{k}^{2} L M \\
& \le - \frac{1}{2} \mu \alpha_{k} | \nabla F(x_{k}) |^{2} + \frac{1}{2} \alpha_{k}^{2} L M .
\end{align*}
Taking sum of both sides for $k=1,2,\dots,K$, we obtain
\begin{equation*}
F^{*} - F(x_{1}) \le \mathbb{E}[F(x_{k+1})] - F(x_{1}) \le - \frac{1}{2} \mu \sum_{k=1}^{K} \alpha_{k}| \nabla F(x_{k}) |^{2} + \frac{1}{2} L M \sum_{k=1}^{K} \alpha_{k}^{2} .
\end{equation*}
Rearranging this inequality, we obtain 
\begin{equation}
\label{eq:sg-diminish-step-bound}
\sum_{k=1}^{K} \alpha_{k}| \nabla F(x_{k}) |^{2} \le \frac{2(F(x_{1}) - F^{*})}{\mu} + \frac{L M}{\mu} \sum_{k=1}^{K} \alpha_{k}^{2} .
\end{equation}
Again, due to $\sum_{k=1}^{\infty} \alpha_{k}^{2} < \infty$ in \eqref{eq:rm-step}, we know the right-hand side is increasing and upper bounded by a finite number as $K \to \infty$, which implies \eqref{eq:sg-diminish-step-sum}.

Furthermore, dividing both sides of \eqref{eq:sg-diminish-step-bound} by $A_{K}$ and noticing the condition $A_{K} = \sum_{k=1}^{K} \alpha_{k} \to \infty$ as $K \to \infty$ in \eqref{eq:rm-step}, we obtain \eqref{eq:sg-diminish-step-sum-K}.
\end{proof}

Based on Theorem \ref{thm:sg-diminish-step}, we can easily make the following claim: 
\begin{equation*}
\liminf_{k \to \infty} \mathbb{E}[ | \nabla F(x_{k}) |^{2} ] = 0 .
\end{equation*}
In other words, $\{\mathbb{E}[ | \nabla F(x_{k}) |^{2} ] : k \ge 1 \}$ must have subsequence that converges to $0$.

\subsubsection*{Second-order Stochastic Optimization Methods}

As pointed out in \cite{bottou2018optimization}, second-order information (also called curvature information) can be very helpful to eliminate some critical issues (such as ill-conditioning) with SG methods and significantly improve their performance.
This is inspired by the substantial successes of second-order methods, such as the class of quasi-Newton methods, in solving a broad range of deterministic optimization problems.
We remark again that, by using second-order information, it does not mean that one needs to compute or store the Hessian of $F$, which can be infeasible, or at least extremely expensive in computation, for deep learning problems where the dimension $n$ is at millions or even billions.

We use a variant of the inexact Newton method to illustrate the use of second-order information in stochastic optimization algorithms. 
In particular, we consider the subsampled Hessian-free inexact Newton method as an example.
This method is summarized in Algorithm \ref{alg:inexact-newton}.
\begin{algorithm}[t]
\caption{Subsampled Hessian-free Inexact Newton Method}
\label{alg:inexact-newton}
\begin{algorithmic}[1]
\REQUIRE Initial guess $x_{1} \in \mathbb{R}^{n}$, constants $\rho,\gamma,\eta \in (0,1)$, $K_{\mathrm{CG}} \in \mathbb{N}$.
	\FOR{$k=1,2,\dots$}
		\STATE Generate realizations of $\xi_{k}$ and $\xi_{k}^{H}$.
		\STATE Compute $g(x_{k}, \xi_{k})$.
		\STATE Compute $s_{k}$ by applying Hessian-free CG to solve 
			\begin{equation}
			\label{eq:inexact-newton-normal}
			H(x_{k}; \xi_{k}^{H}) s = - g(x_{k}; \xi_{k})
			\end{equation}
			and terminate CG if the maximum iteration $K_{\mathrm{CG}}$ is reached or
			\begin{equation}
			\label{eq:inexact-newton-cg-error}
			| H(x_{k}; \xi_{k}^{H})s + g(x_{k},\xi_{k})| \le \rho |g(x_{k};\xi_{k})| 
			\end{equation}
			is satisfied.
		\STATE Choose a step size $\alpha_{k} > 0$ by backtracking line search with rate $\gamma$ such that
		\begin{equation}
		\label{eq:inexact-newton-ls}
		f(x_{k+1}; \xi_{k}) \le f(x_{k}; \xi_{k}) + \eta\, \alpha_{k}\, g(x_{k}; \xi_{k}) \cdot s_{k}
		\end{equation}
		where $x_{k+1} = x_{k} + \alpha_{k} s_{k}$.
	\ENDFOR
% \ENSURE $x_{k}$. 
\end{algorithmic}
\end{algorithm}

To implement this method, we note that in addition to $\xi_{k}$ of size $N_{k}$ for $g(x_{k}, \xi_{k})$ as in \eqref{eq:sgd-g}, we need to sample another realization $\xi_{k}^{H}$ of size $N_{k}^{H}$ which yields a stochastic Hessian estimate
\begin{equation}
\label{eq:inexact-newton-H}
H(x_{k}; \xi_{k}^{H}) := \frac{1}{N_{k}^{H}} \sum_{i=1}^{N_{k}^{H}} \nabla^{2} f(x_{k}; \xi_{k,i}^{H}) .
\end{equation}
It was found that this method appears not sensitive to the noise in the estimate $H(x_{k};\xi_{k}^{H})$ as to $g(x_{k}; \xi_{k})$. Therefore, $N_{k}^{H}$ is often chosen to be smaller than $N_{k}$, making the computational cost lower.
Note that, $H(x_{k};\xi_{k}^{H})$ is not explicitly computed and stored here.
Instead, the method employs the conjugate gradient (CG) algorithm (Algorithm \ref{alg:cg}) to solve for $s_{k} \in \mathbb{R}^{n}$ from the linear system \eqref{eq:inexact-newton-normal}, which only requires a number of matrix-vector multiplications (see Section \ref{sec:autodiff} for details of computing such multiplications by automatic differentiation).
This is the reason that the method is Hessian-free.

While the CG algorithm is proved to achieve the exact solution of \eqref{eq:inexact-newton-normal} in finite steps provided $H(x_{k};\xi_{k}^{H})$ is positive definite, one usually terminates CG early once a prescribed maximum iteration number $K_{\mathrm{CG}}$ is reached or the difference between the two sides of \eqref{eq:inexact-newton-normal} is sufficiently small as specified by \eqref{eq:inexact-newton-cg-error}.
The solution $s_{k}$ obtained by the CG algorithm for \eqref{eq:inexact-newton-normal} plays the role of a descent direction, and a standard backtracking line search strategy is applied to determine the step size $\alpha_{k}$ and consequently the next iterate $x_{k+1}$ with $\xi_{k}$ fixed.

\subsection{Practical Stochastic Optimization Algorithms}
\label{subsec:practical-soa}

While the standard stochastic gradient descent methods introduced above remain the fundamental approach, their limitations have motivated the development of adaptive optimization methods. 
In this subsection, we present several widely used adaptive optimizers in deep learning with some comments on their motivations and properties. 
However, at the time of writing, these methods do not have complete convergence guarantee when solving stochastic optimization problems with smooth non-convex objective functions under mild regularity assumptions.
The investigation of their convergence behaviors is an active area of research nowadays.

Nevertheless, these method appear to be faster and more efficient (and also stable) than the standard basic/mini-batch SG methods empirically.
Therefore, they are often the first choice for training networks in deep learning research and industrial applications.
Meanwhile, it is very likely that more efficient and stable stochastic optimization algorithms will be invented continuously in the future.

% Adaptive optimizers such as AdaGrad, RMSProp, Adam, and AdamW have greatly simplified the training of deep neural networks by reducing the need for manual learning-rate tuning. While Adam and AdamW dominate practical applications, recent empirical and theoretical studies suggest that no single optimizer is universally superior. Understanding the trade-offs between convergence speed, stability, and generalization remains an active area of research.

\subsubsection*{Momentum Method}

For notation simplicity, we denote by $g_{k}$ the stochastic descent direction such as those in \eqref{eq:sgd-g}:
\begin{equation*}
g_{k} := g(x_{k}; \xi_{k}) .
\end{equation*}
One of the two critical techniques to accelerate convergence is using the momentum. 
The idea of momentum is to incorporate a weighted moving average of past gradients to improve $g_{k}$. 
This helps smoothing out the trajectory of the optimization, allowing the algorithm to converge faster by reducing oscillations.
A basic stochastic gradient descent method with \emph{momentum}\index{Momentum} is given by
\begin{subequations}
\label{eq:momentum}
\begin{align}
m_{k+1} & = \beta_{1} m_{k} + (1 - \beta_{1}) g_{k} , \label{subeq:momentum-m}\\
x_{k+1} & = x_{k} - \alpha_{k} m_{k+1} , \label{subeq:momentum-x}
\end{align}
\end{subequations}
where $m_{k} \in \mathbb{R}^{n}$ is the momentum at the $k$th iteration, and $\alpha_{k}$ is the step size (learning rate).
The momentum $m_{1}$ initialized as $0$, and the weight parameter $\beta_{1}$ is often set to $0.9$ in practice.

Despite the speed improvement of \eqref{eq:momentum}, the step sizes $\alpha_{k}$ still depend on the Lipschitz constant of the gradient of the objective function to be minimized, and they need to follow a diminishing policy for convergence.
These bring additional trouble for practitioners to tune these step sizes for optimal performance.
The \emph{adaptive methods}\index{Adaptive method} that we present below have the feature of cancelling the Lipschitz constant and can often work properly with a fixed step size. 
These adaptive methods are effectively using a diagonal scaling\index{Diagonal scaling} matrix $H_{k}$, cf.~\eqref{eq:sgd-g}, to incorporate second-order information of the objective function.

\subsubsection*{Adaptive Gradient (AdaGrad)}

AdaGrad is one of the earliest adaptive methods proposed for large-scale and sparse optimization problems \cite{duchi2011adaptive}. 
It introduces a raw second-order moment estimate in the form of a vector $v_{k} \in \mathbb{R}^{n}$ in each iteration.
Then the components of $v_{k}$ are used to scale their corresponding ones in $g_{k}$.
The AdaGrad\index{AdaGrad} scheme is given by
\begin{subequations}
\label{eq:adagrad}
\begin{align}
v_{k+1} & = v_{k} + g_{k} \odot g_{k} , \label{subeq:adagrad-v}\\
x_{k+1} & = x_{k} - \alpha_{k} \frac{g_{k}}{\sqrt{v_{k+1}} + \epsilon} , \label{subeq:adagrad-x}
\end{align}
\end{subequations}
where $\odot$ is the componentwise product of two vectors, which results in a vector of the same dimension, and the parameter $\epsilon > 0$ is usually set to $10^{-8}$ to bound the denominator away from $0$.
The square root and division in \eqref{subeq:adagrad-x} are also applied componentwisely.
Therefore, the components of $g_{k}$ are scaled individually in \eqref{subeq:adagrad-x}, and this ratio is invariant to the scaling of the objective function by any positive constant (which changes the Lipschitz constant of the objective function).
As a result, the step sizes $\alpha_{k}$ are not affected by such scaling and hence can remain fixed during iterations.

AdaGrad is particularly effective in settings with sparse features, such as natural language processing, since infrequent parameters receive relatively larger updates. However, its main drawback is the denominator monotonically increases, which makes the ratio in \eqref{subeq:adagrad-x} tend to 0 rapidly. This often leads to premature convergence and stalled training progresses.

\subsubsection*{Root Mean Square Propagation (RMSProp)}

RMSProp\index{RMSProp} \cite{tieleman2012rmsprop:} was proposed as a practical improvement over AdaGrad to mitigate its aggressive learning-rate decay issue. The RMSProp scheme is given by
\begin{subequations}
\label{eq:rmsprop}
\begin{align}
v_{k+1} & = \beta_{2}v_{k} + (1-\beta_{2}) g_{k} \odot g_{k} , \label{subeq:rmsprop-v}\\
x_{k+1} & = x_{k} - \alpha_{k} \frac{g_{k}}{\sqrt{v_{k+1}} + \epsilon} , \label{subeq:rmsprop-x}
\end{align}
\end{subequations}
where $\beta_{2}$ is often set to $0.99$, and $v_{1}$ is initialized to $0$. As we can see, the weighted sum in \eqref{subeq:rmsprop-v} mitigate the issue of accumulating squared gradient, and hence the denominator vector in \eqref{subeq:rmsprop-x} does not increase aggressively as in Adagrad.

By focusing on recent gradients, RMSProp performs well in many applications and has been widely used for training recurrent neural networks. However, RMSProp seems sensitive to hyperparameter choices.

\subsubsection*{Adaptive Momentum Estimate (Adam)}

Adam\index{Adam} optimizer \cite{kingma2015adam} combines the momentum and RMSprop techniques to provide a more balanced and efficient optimization process. The key equations governing Adam are as follows: Adam combines ideas from momentum methods and RMSProp by maintaining exponential moving averages of both first- and second-order moments of the gradients. 
Specifically, in the $k$th iteration, Adam updates the first moment estimate $m_{k}$ and raw second moment estimate $v_{k}$ by
\begin{subequations}
\label{eq:adam-mv}
\begin{align}
m_{k+1} & = \beta_{1}m_{k} + (1-\beta_{1}) g_{k} , \label{subeq:adam-m}\\
v_{k+1} & = \beta_{2}v_{k} + (1-\beta_{2}) g_{k} \odot g_{k} , \label{subeq:adam-v}
\end{align}
\end{subequations}
which have initials $m_{1} = v_{1} = 0$. 
Since such initials make $m_{k}$ and $v_{k}$ biased towards $0$, Adam attempts to correct such biases using their corrected versions
\begin{subequations}
\label{eq:adam-mv-correct}
\begin{align}
\hat{m}_{k+1} & = \frac{m_{k+1}}{1 - \beta_{1}^{k}} , \label{subeq:adam-m-correct}\\
\hat{v}_{k+1} & = \frac{v_{k+1}}{1 - \beta_{2}^{k}} , \label{subeq:adam-v-correct}
\end{align}
\end{subequations}
where both of the correction denominators in \eqref{eq:adam-mv-correct} tend to 1 as $k\to\infty$ provided $\beta_{1},\beta_{2} \in (0,1)$.
Then Adam updates $x_{k}$ by
\begin{equation}
\label{eq:adam-x}
    x_{k+1} = x_{k} - \alpha_{k} \frac{\hat{m}_{k+1}}{\sqrt{\hat{v}_{k+1}} + \epsilon}.
\end{equation}
The hyperparameters are often set the same as before: $\beta_{1} = 0.9$, $\beta_{2} = 0.999$, $\epsilon = 10^{-8}$.
The step size $\alpha_{k}$ can be fixed as a constant and often set to $10^{-3}$ which generally works well, but occasionally can be as small as $10^{-5}$.

Adam is widely adopted due to its fast convergence and robustness across a broad range of architectures and datasets. However, subsequent studies have shown that Adam may generalize worse than SG with momentum in some scenarios and may fail to converge in certain theoretical settings \cite{reddi2018convergence}.

\subsubsection*{Adam with Decoupled Weight Decay (AdamW)}

A typical approach to control the value of $x$ and stabilize optimization process is simply adding a Tikhonov regularization\index{Regularization} to the objective function.
Namely, one can apply an optimization algorithm, say Adam, to minimize $f(x) + \frac{\lambda}{2} |x|^{2}$ instead of $f(x)$.
Here $\lambda$ is a hyperparameter to weigh the term $|x|^{2}$.
As we can see, the Tikhonov regularization tends to reduce the magnitudes of the components in $x$.
Such modification, however, changes the objective function, its first- and second-moments, and most importantly, the optimal solution.

AdamW\index{AdamW} \cite{loshchilov2017decoupled} is an alternative approach by simply adding a decay term when updating $x$. More specifically, AdamW proceeds with the same computations \eqref{eq:adam-mv}--\eqref{eq:adam-mv-correct} as in Adam, but only changes \eqref{eq:adam-x} to 
\begin{equation}
\label{eq:adamw-x}
    x_{k+1} = x_k - \alpha_{k} \lambda x_{k} - \alpha_{k} \frac{\hat{m}_{k+1}}{\sqrt{\hat{v}_{k+1}} + \epsilon} ,
\end{equation}
which differs from \eqref{eq:adam-x} by an additional term $\alpha_{k} \lambda x_{k}$.
Here $\lambda$ is referred to as the decay weight and often set to $0.01$ in practical implementations.

In various empirical tests, AdamW appears to improve generalization and stability, particularly in large-scale models such as transformers, when compared to Adam. As a result, AdamW has become the default optimizer in many modern deep learning frameworks.

\subsubsection*{Momentum Orthogonalized by Newton-Schulz (Muon)}

Muon\index{Muon} \cite{jordan2024muon:} is an optimization approach that harnesses the singular values of the weight parameters in deep neural networks.
Recall that a majority of building blocks in deep neural networks have affine\index{Affine function} operations of form $Wx + b$, where $W$ is a weight matrix.
For ease of presentation, we call $W$ a 2-dimensional (2D) weight matrix and assume its size is $d_{\mathrm{out}} \times d_{\mathrm{in}}$ hereafter.

Notice that, in all optimization algorithms we discussed above, $W$ is reshaped as a vector and concatenated with $b$ as a part of the total parameter vector $\theta \in \mathbb{R}^{n}$ (we use $x \in \mathbb{R}^{n}$ in this chapter but it refers to the parameter $\theta \in \mathbb{R}^{n}$ in network training).
Therefore, the matrix structure of $W$ is ignored during the optimization process of network training.

However, the weight matrices play important roles in affine transformation and have strong impact to the Lipschitz constant of the network which is effectively a mapping from its inputs to the corresponding outputs. 
Therefore, Muon proposes to regularize the gradients with respect to the weight matrices by orthogonalization. 
To this end, Muon employ a few Newton--Schulz (NS) iterations to the matrix momentum $\mu M_{k}+g_{k}$ to normalize all singular values to 1.
The Muon scheme is given by
\begin{subequations}
\label{eq:muon}
\begin{align}
M_{k+1} & = \mu M_{k} + g_{k} , \label{subeq:muon-M}\\
O_{k+1} & = \mathrm{NS}(M_{k+1}; (a,b,c), \epsilon, K_{\mathrm{NS}}) , \label{subeq:muon-O} \\
x_{k+1} & = x_{k} - \alpha \lambda x_{k} - \alpha O_{k} , \label{subeq:muon-x}
\end{align}
\end{subequations}
where the weight parameter $\mu$ is often set to $0.95$, $\lambda$ is set to $0.1$, and the step size $\alpha$ is set to $0.001$.

Now we explain the terms in \eqref{eq:muon}. First, we note that $g_{k}$ in Muon \eqref{eq:muon} refers to the (stochastic) gradient of the objective function with respect to each individual weight matrix in the neural network. 
Muon can also be applied to update 4D convolutional parameters by flattening three of the four dimensions properly such that they become 2D matrices like the weights. The gradients with respect to other parameters in the form of scalar or vector are still updated using the previously presented algorithms, such as Adam and AdamW. 
Hence, we assume hereafter that $g_{k}$ is a 2D matrix of the same size as its corresponding weight matrix $W \in \mathbb{R}^{d_{\mathrm{out}} \times d_{\mathrm{in}}}$.
The momentum $M_{k}$ is a matrix of the same size.

The key step in the Muon scheme is \eqref{subeq:muon-O}. Taking any matrix $M$ of size $m \times n$ (without loss of generality, we assume $m \ge n$ hereafter) as input, this step returns a matrix $O$ of the same size as an approximation of $U V^{\top}$, where $M=U\Sigma V^{\top}$ is the \emph{singular value decomposition}\index{Singular value decomposition} (SVD) of $M$.
We know that, SVD can factorize any real or complex matrix $M$ in the form of $U\Sigma V^{\top}$, where $U \in \mathbb{R}^{m \times n}$ has $n$ orthonormal vectors in $\mathbb{R}^{m}$ (i.e., the columns of $U$ is a subset of an orthonormal basis of $\mathbb{R}^{m}$), $V$ is an $n\times n$ orthogonal matrix, and $\Sigma$ is an $n\times n$ diagonal matrix with non-negative diagonal entries called the \emph{singular values}\index{Singular value} of $M$. Consequently, the columns of $U$ and $V$ are called the left and right \emph{singular vectors} of $M$, respectively. See details about SVD in \cite{trefethen1997numerical, golub1996matrix}. 
Therefore, the effect of the step \eqref{subeq:muon-O} is removing $\Sigma$ from $M=U \Sigma V^{\top}$ and approximating the orthogonal version $O = U V^{\top}$ of $M$.
Namely, $O$ is the solution to
\begin{equation}
\label{eq:muon-min}
\min_{O \in \mathbb{R}^{m \times n}} \frac{1}{2} \| O - M \|_{F}^{2} ,
\end{equation}
where $\| A \|_{F}:= (\sum_{i=1}^{m} \sum_{j=1}^{n} |A_{ij}|^{2})^{1/2}$ is the Frobenius norm\index{Norm!Frobenius} of the $m$-by-$n$ matrix $A=[A_{ij}]_{i,j}$. 
This is effectively reshaping $A$ as an $mn$-dimensional vector and applying the standard vector 2-norm.
Moreover, Frobenius norm is the norm induced by the inner product $\langle A, B \rangle_{F} := \text{tr}(A^{\top} B)$ for any $A, B \in \mathbb{R}^{m \times n}$.
Due to these properties of Frobenius norm, one can easily find that the minimizer of \eqref{eq:muon-min} is $U V^{\top}$.
This is called the orthogonalization in Muon.

However, SVD is computationally expensive and one may want to avoid using it to solve \eqref{eq:muon-min} in practice.
The NS iteration is a method that can quickly approximate the solution to \eqref{eq:muon-min} and hence is employed by Muon in \eqref{subeq:muon-O}.

The idea of the NS iteration is as follows.
We define a quintic polynomial $\phi: \mathbb{R} \to \mathbb{R}$ of $\sigma$ (here $\sigma$ refers to a positive singular value of a matrix) defined by
\begin{equation}
\label{eq:muon-phi}
\phi(\sigma) := a \sigma + b \sigma^{3} + c \sigma^{5} .
\end{equation}
We also extend this definition and apply $\phi$ to a matrix $M$:
\begin{align}
\phi(M)
& := a M + b(M M^{\top}) M + c (M M^{\top})^{2} M \nonumber \\
& \;= \Big( a I_{m} + b(M M^{\top})  + c (M M^{\top})^{2} \Big) M \label{eq:muon-phi-mtx} \\
& \;= \Big( a I_{m} + b(U \Sigma^{2} U^{\top})  + c (U \Sigma^{4} U^{\top}) \Big) U \Sigma V^{\top} \nonumber \\
& \;= U\Big( a \Sigma + b \Sigma^{3}  + c \Sigma^{5} \Big) V^{\top} , \nonumber 
\end{align}
where we used the SVD $M = U \Sigma V^{\top}$ with $U^{\top} U = I_{n}$ repeatedly to obtain the equalities.
We see that $\phi(M)$ in \eqref{eq:muon-phi-mtx} effectively applies \eqref{eq:muon-phi} to each singular value $\sigma$ of $M$ individually.

Therefore, for the NS iteration to normalize all singular values to $1$ rapid, we just need to find the coefficients $(a,b,c)$ of \eqref{eq:muon-phi} such that a few iterations of $\phi$ can map any positive number $\sigma$ to 1.
In other words, we need to find $(a,b,c)$ in \eqref{eq:muon-phi} such that
\begin{equation*}
\phi^{(j)} \, (\sigma) \approx 1, \quad \forall \, \sigma > 0
\end{equation*}
with $j$ as small as possible, where $\phi^{(j)}$ denotes the composition of $\phi$ for $j$ times.
If such $(a,b,c)$ can be found, we can approximate $O=U V^{\top}$ using $\phi^{(j)}(M)$ for any $M = U\Sigma V^{\top}$.

In practice, the simple choice $(a,b,c) = (2,-1.5,0.5)$ appears to work just fine.
With some subtle optimization, the choice $(a,b,c) = (3.4445, -4.775, 2.0315)$ seems to work even better and hence is currently the default setting in practical implementations in popular deep learning programming packages such as PyTorch \cite{paszke2019pytorch:}.
In the Muon method, the NS iteration in \eqref{subeq:muon-O} is often terminated if the maximum iteration number $K_{\mathrm{NS}}$ (usually 5 is enough) is reached or the relative change of $(a,b,c)$ is smaller than $\epsilon$ (usually set to $10^{-7}$).

We can also incorporate the idea of Nesterov's momentum into the Muon method \eqref{eq:muon}.
This is often implemented by replacing $B_{k+1}$ in \eqref{subeq:muon-O} with $g_{k} + \mu B_{k+1}$ where $\mu$ is set to $0.95$ by default. 

There are also several modifications to the step size in practical implementations.
In particular, the step size $\alpha$ is multiplied by $(\max (1, d_{\mathrm{out}}/d_{\mathrm{in}}))^{1/2}$ in \cite{jordan2024muon:} and by $0.2 (\max (d_{\mathrm{out}}, d_{\mathrm{in}}))^{1/2}$ in \cite{liu2025muon}, which appear to be efficient in large language model training empirically.

\section{Remarks and References}
\label{sec:opt-remark}

\subsection{Remarks on Optimization for Network Training}

This chapter covers most of the fundamental concepts in optimization theory, as well as a number of optimization algorithms and techniques. The focus is on non-convex smooth optimization problems, which are prevalent in modern deep learning. Indeed, there have been a substantial amount of results on deterministic optimization methods developed in the past decades, including comprehensive convergence analysis and mature software packages. However, they seem to be faltered in solving modern large-scale optimization problems \cite{bottou2018optimization}. By contrast, new stochastic optimization algorithms are being continuously invented in the deep learning community to address the challenges of such problems. Despite these algorithms generally do not have complete and rigorous convergence guarantees yet, their extraordinary efficiency makes them dominate in large deep network training nowadays.

Besides the comments made in \cite{bottou2018optimization}, we have several additional remarks on this phenomenon from the practical perspective. 
Training large-scale neural networks is known to be extremely resource demanding. It requires a substantial number of graphical processing units (GPUs) and huge amount of electricity power. 
Some large networks, such as those designed for modern large language models, each have several billions to hundreds of billions of parameters, and the training often takes months on clustered systems of GPUs to complete. 
Given the current surge of artificial intelligence, rapid developments of new deep learning techniques, and fierce competitions between different entities, it is not surprising that people prefer algorithms that are faster and more efficient to obtain satisfactory results in short time at the cost of slightly reduced solution quality.
By contrast, traditional approaches in the field of optimization focus on crafted algorithmic design, rigorous convergence analysis, tight complexity estimates with upper/lower complexity bounds, and flexible hyperparameter settings.
\begin{figure}[t]
\centering
\begin{tikzpicture}[domain=-0.1:4, scale=1, transform shape]
	
	\draw[->] (-0.1, 0) -- (4.8, 0) node[right]{\# epochs};
	\draw[->] (0, -0.1) -- (0, 3.5) node[left]{loss};
	
	\draw[line width=0.2mm, color=black] plot[domain=0:4.2]({\x}, {3/(10*\x + 1)});
	\draw[line width=0.2mm, dashed, color=black] plot[domain=0:4.2]({\x}, {3*0.35^(\x)});
		
	\draw[black] (3.3, -0.05) -- (3.3, 0.05);
	\draw[black] (1, -0.05) -- (1, 0.05);
	
	\draw[black] (3.3, 0) node[below]{\small{$k_{2}$}};
	\draw[black] (1, 0) node[below]{\small{$k_{1}$}};

\end{tikzpicture}
\caption{An illustrative comparison between the loss function value decays using two optimization algorithms. The first algorithm (solid curve) converges at a sublinear rate\index{Convergence!sublinear} $O(1/k)$, where $k$ is the number of epochs, and the second algorithm (dashed curve) converges at a linear rate\index{Convergence!linear} $O(\gamma^{k})$ for some $\gamma \in (0,1)$. The first algorithm has a fast initial convergence and probably can reach a satisfactory solution after $k_{1}$ epochs, whereas the second algorithm has a faster asymptotic convergence and will surpass the first algorithm after $k_{2}$ epochs.  }
\label{fig:convergence-rate}
\end{figure}
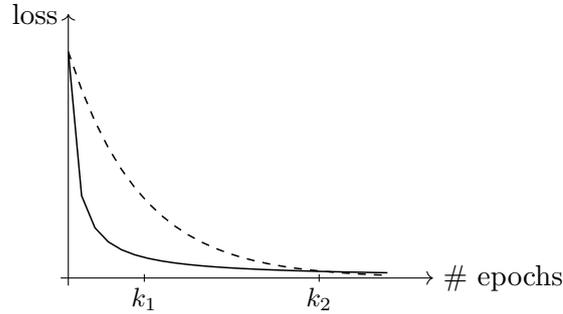

Figure \ref{fig:convergence-rate} shows an illustrative plot to explain the situation.
In Figure \ref{fig:convergence-rate}, the horizontal line is the number of epochs which represent the practical computational cost, and the vertical line is the loss function value on training dataset.
The solid curve shows the performance of an optimization algorithm with sublinear convergence rate $O(1/k)$, whereas the dashed curve is of an algorithm with linear convergence rate $O(\gamma^{k})$ with some $\gamma \in (0,1)$.  
In theory, it is natural that we prefer the second algorithm as its linear convergence is superior to sublinear rate and will converge to the minimum loss function value much faster asymptotically.

However, given the two decay curves in Figure \ref{fig:convergence-rate}, a practitioner may prefer the first algorithm with sublinear rate (solid curve) due to its fast convergence at the initial stage.
This is particularly reasonable if the result provided by this algorithm at the $k_{1}$th epoch is already satisfactory.
Despite the algorithm with linear rate (dashed curve) will outperform the first one after $k_{2}$ epochs, the improvement of solution quality may not be enough to justify the substantial investment in additional computations. 
For example, a few percent improvement of evaluation measure on training dataset in a recommender system or generative pre-trained transformer (GPT) may not be noticeable to human, but the training time can be a 4 months ($k_{2}$) versus 1 month ($k_{1}$) difference in practice. 
In this situation, most practitioners would choose to use (at least first try) the algorithm with sublinear convergence rate.

Due to the large scale and complex structure of deep neural networks, numerous factors can strongly influence the performance of a deep learning model.
Among these, the choice of methodology and network architecture is often the most critical.
For example, a moderately trained convolutional neural network can achieve much higher image classification accuracy than a dedicately trained fully connected network.
In addition, the quality of the data samples, the choice of initialization, and the preprocessing procedures all have a significant impact on the empirical performance of network training.
Consequently, practitioners tend to prioritize these aspects when designing and training deep learning models, while typically selecting a widely used optimizer (such as Adam or AdamW) that is known for its efficiency and fast convergence in practice, without paying close attention to its theoretical convergence guarantees.

\subsection{References of Numerical Optimization}

Comprehensive introduction to optimization theory and applications can be found in \cite{chong2013introduction}. 
For the fundamental theoretical results in convex analysis and optimization algorithms, we refer to \cite{rockafellar1996convex,bertsekas2015convex,nesterov2014introductory,borwein2006convex,boyd2004convex,bertsekas2003convex,ben-tal1987lectures}. 
General continuous optimization theory and algorithms are given in the classic books \cite{bertsekas2016nonlinear,nocedal2006numerical}.

The discussions on basic optimization theory and optimality conditions in Sections \ref{sec:opt-basics} and \ref{sec:opt-optcond} largely follow \cite{chong2013introduction}.
Section \ref{sec:autodiff} on automatic differentiation follows \cite{nocedal2006numerical}, with some extra details and explanations throughout, particularly on the computations of second-order derivatives.
Section \ref{sec:deterministic-opt} on algorithms for deterministic optimizations mostly follows \cite{chong2013introduction,nocedal2006numerical,bertsekas2016nonlinear}. 
Fundamentals of stochastic optimization given in Section \ref{subsec:fundamentals-so} follow \cite{bottou2018optimization}. 
We provide an overview of the practical stochastic optimization algorithms in Section \ref{subsec:practical-soa} based on online discussions and short tutorials as they just emerged in recent years.

\chapter{Deep Optimal Control}
\label{chpt:oc}

Optimal control provides a rigorous mathematical framework for designing sequences of decisions that steer dynamical systems toward desired objectives, such as maximizing reward (or equivalently, minimizing cost), subject to physical and operational constraints. It plays a central role in a broad range of real-world applications.

In this chapter, we begin with a systematic study of the fundamental concepts and core results of optimal control. In particular, we present the foundational theory of optimal control, including the Pontryagin Maximum Principle and the Hamilton--Jacobi--Bellman equation, together with their proofs. We then extend these classical results to the optimal control of probability densities, also known as mean-field control, motivated by the fact that modern deep learning systems often involve controlling a massive number of agents. As a result, many problems of interest in these situations can naturally be formulated as control problems over probability distributions.

Despite the importance and widespread use of optimal control, classical methods suffer from severe scalability limitations in high-dimensional settings or in the presence of complex nonlinear dynamics. To address these challenges, we introduce a deep learning-based approach, known as the neural ordinary differential equation (Neural ODE) method, as a powerful tool for solving a broad class of optimal control problems. This framework can be adapted to handle both standard optimal control problems and the probability control problems. Finally, we demonstrate the strong potential of combining deep learning with optimal control for tackling emerging scientific computing problems, using the task of learning solution operators for evolution partial differential equations as a representative example.

\section{Optimal Control Theory}
\label{sec:oc-theory}

In this section, we present a concise and systematic review of the foundational results in optimal control theory. We begin with the calculus of variations and its role in deriving the Euler--Lagrange equations. We then introduce Hamiltonian dynamics, which offer a unifying and insightful perspective on general control systems. Finally, we present the Pontryagin Maximum Principle and the Hamilton--Jacobi--Bellman equation, together with rigorous proofs.

\subsection{Euler--Lagrange Equations}
\label{subsec:oc-el}

Optimal control is a complex optimization problem with very different nature compared to standard optimization discussed in Chapter \ref{chpt:opt}.
In an optimal control problem, the subject to be optimized is a function, called \emph{control}\index{Control}, which is often represented as a mapping from time $[0,T]$ (can be extended to $[0,\infty)$ or $\mathbb{R}$, but we restrict most of our discussions to finite time window as this has sufficient generality) to a subset $\Ucal$ of some multi-dimensional Euclidean space $\mathbb{R}^{m}$.
This subset $\Ucal$ is called the \emph{admissible set of control}\index{Control!admissible set}.

While optimal control shares a similar goal of seeking for the solution that minimizing or maximizing certain objective function, the theory and solution methods are very different from (but still related to) those in standard optimization given in Chapter \ref{chpt:opt}.
As the subject to be optimized is a time-dependent control function $u: [0,T] \to \Ucal \subset \mathbb{R}^{m}$, we need new theory, analysis, and computational tools to solve optimal control problems.

% Since the subjects to be optimized are functions in optimal control rather than vectors, we need to consider a generalization of calculus to the cases where the ``variables'' are now functions.
%
The basic approach is called the \emph{calculus of variations}\index{Calculus of variations}, and such method is known as the \emph{variational method}\index{Variational method}.
We use a classical example to introduce calculus of variations and Lagrangian functions, which further yield the Euler--Lagrange equations.

Suppose that we are given a smooth function $L: \mathbb{R}^{d} \times \mathbb{R}^{d} \rightarrow \mathbb{R}$. We want to find a curve $ x^{*}: [0, T] \rightarrow \mathbb{R}^{d}$ that minimizes the functional
\begin{equation}
\label{eq:oc-int-lagrangian}
I[ x ]:=\int_{0}^{T} L( x(t), \xdot(t)) \, d t ,
\end{equation}
with constraints $ x(0)=x_{0}$ and $ x(T)=x_{1}$ for some prescribed $x_{0},x_{1} \in \mathbb{R}^{d}$.
In \eqref{eq:oc-int-lagrangian}, $\xdot$ is the time-derivative of $x$ (we can think of $\xdot$ as the control for now).
We often interpret $x(t)$ as the position (in general we refer it as the state) and $\xdot(t)$ as the velocity at time $t$.
Similar to optimization, our first step is to find the optimality conditions of $\xs$. Namely, suppose that $\xs$ is a minimizer of $I$ defined in \eqref{eq:oc-int-lagrangian}, we want to know the conditions that $\xs$ must satisfy.
These conditions to optimal control are analogues to the Lagrange conditions and KKT conditions to constrained optimization in Chapter \ref{chpt:opt}.

In the literature, the integrand $L$ in \eqref{eq:oc-int-lagrangian} is called the \emph{Lagrangian function}\index{Lagrangian!function}. Note that this should be distinguished from the Lagrangian function of equality constrained optimization problems in Chapter \ref{chpt:opt}, as they have different definitions and purposes.
We often write $L$ as a function of $x, v \in \mathbb{R}^{d}$ (where $v$ substitutes $\xdot$ which is interpreted as the velocity).
In the remainder of this section, we also use the following notations to denote partial derivatives:
\begin{align*}
\partial_{x} L & :=\left(\partial_{x_{1}} L, \ldots, \partial_{x_{d}} L \right), \\
\partial_{v} L & :=\left(\partial_{v_{1}} L, \ldots, \partial_{v_{d}} L \right),
\end{align*}
where
\begin{equation*}
\partial_{x_{i}} L = \frac{\partial L}{\partial x_{i}} \quad \text{and} \quad \partial_{v_{i}} L = \frac{\partial L}{\partial v_{i}}
\end{equation*}
for every $i = 1,\dots, d$.
The following theorem shows that, if $\xs$ is an optimal solution to \eqref{eq:oc-int-lagrangian}, then it must satisfy the so-called \emph{Euler--Lagrange equation}\index{Euler--Lagrange equation}.
%
% In other words, satisfying this Euler--Lagrange equation is a necessary condition for $\xs$ to be a minimizer of \eqref{eq:oc-int-lagrangian}.
%
The proof of this theorem uses the idea of calculus of variations, which can be seen below.

\begin{theorem}[Euler--Lagrange equation]
\label{thm:oc-el}
Suppose $x$ is an optimal solution to \eqref{eq:oc-int-lagrangian} with the boundary constraints, then $x$ solves the Euler--Lagrange equation:
\begin{equation}
\label{eq:oc-el}
\frac{d}{d t}\Big(\partial_{v} L ( x(t), \xdot(t) ) \Big)=\partial_{x} L ( x(t), \xdot(t)) .
\end{equation}
\end{theorem}

\begin{proof}
Recall that the solution to a smooth optimization problem without constraint is among the critical points where the gradient of the objective function is zero.
We would like to follow a similar approach and try to find those curves where the ``gradient'' of the functional $I$ defined in \eqref{eq:oc-int-lagrangian} is zero.
However, we cannot directly take such gradient as it is with respect to a curve rather than a vector.
Therefore, we first convert the problem to one that the standard calculus can be applied.

To this end, we choose any smooth curve $y: [0, T] \rightarrow \mathbb{R}^{d}$ with $y(0)= y(T)=0$.
We usually call $y$ a perturbation of $x$. 
Now we see that for any $\epsilon \in \mathbb{R}$, the curve $x+ \epsilon y: [0, T] \rightarrow \mathbb{R}^{d}$ still satisfies the boundary constraints $x(0)+\epsilon y(0) = x_{0}$ and $x(T)+\epsilon y(T) = x_{1}$ at $t=0$ and $t=T$, respectively.
Define $\phi: \mathbb{R} \to \mathbb{R}$ by
\begin{equation}
\label{eq:phi-epsilon}
\phi(\epsilon) := I[ x + \epsilon y] = \int_{0}^{T} L(x(t)+\epsilon y(t), \xdot(t) + \epsilon \ydot(t)) \, dt.
\end{equation}
Since $x$ is a minimizer of \eqref{eq:oc-int-lagrangian}, we have
\begin{equation}
\label{eq:oc-el-pf-phi-chain}
\phi(0) = I[x] \le I[ x + \epsilon y ] = \phi(\epsilon)
\end{equation}
for any $\epsilon \in \mathbb{R}$.

Since $L: \mathbb{R}^{d} \times \mathbb{R}^{d} \to \mathbb{R}$ is a smooth function, we know $\phi$ is smooth and by \eqref{eq:oc-el-pf-phi-chain} there is
\begin{equation*}
\phi'(0) = 0 .
\end{equation*}

Now we need to find the expression of $\phi'$. To this end, we recall \eqref{eq:phi-epsilon}
% \begin{equation*}
% \phi(\epsilon) = \int_{0}^{T} L \Big( x(t)+\epsilon  y(t), \xdot(t)+ \epsilon \ydot(t) \Big) \, d t
% \end{equation*}
and find
\begin{align*}
\phi'(\epsilon) & = \int_{0}^{T} \Big[  \partial_{x} L \Big( x(t)+\epsilon  y(t), \xdot(t)+ \epsilon \ydot(t) \Big) \cdot y(t) \\
& \qquad \qquad + \partial_{v} L \Big( x(t)+\epsilon  y(t), \xdot(t)+ \epsilon \ydot(t) \Big) \cdot \ydot(t)  \Big] dt .
\end{align*}
Then, at $\epsilon=0$, we obtain
\begin{align}
\label{eq:oc-el-pf-dphi0}
0 = \phi'(0) = \int_{0}^{T} \Big[  \partial_{x} L(x(t), \xdot(t)) \cdot y(t) + \partial_{v} L(x(t),\xdot(t)) \cdot \ydot(t)  \Big] dt .
\end{align}

Now we investigate what \eqref{eq:oc-el-pf-dphi0} implies provided that it holds true for all $y: [0,T] \to \mathbb{R}^{d}$ with $y(0) = y(T) = 0$. Suppose we choose any $i$ from $1,\dots,d$, and set
\begin{equation}
\label{eq:oc-el-pf-y-special}
y(t) = ( 0, \dots, 0, \underbrace{z(t)}_{i\text{th}}, 0, \dots, 0) \in \mathbb{R}^{d}
\end{equation}
for all $t \in [0,T]$, where $z: [0,T] \to \mathbb{R}$ is a smooth function with $z(0) = z(T) = 0$.
Then $y$ defined in \eqref{eq:oc-el-pf-y-special} also satisfies the constraints at the times $0$ and $T$.
Substituting this $y$ defined in \eqref{eq:oc-el-pf-y-special} into \eqref{eq:oc-el-pf-dphi0}, we obtain
\begin{equation*}
0=\int_{0}^{T} \Big( \partial_{x_{i}} L( x(t), \xdot(t)) z(t) + \partial_{v_{i}} L( x(t), \xdot(t)) \zdot(t) \Big) \, d t .
\end{equation*}
Taking integration by parts of the second term in the integral on the right-hand side above, we obtain
\begin{equation}
\label{eq:oc-el-pf-zi}
0=\int_{0}^{T} \Big( \partial_{x_{i}} L( x(t), \xdot(t)) - \frac{d}{d t} \partial_{v_{i}} L( x(t), \xdot(t)) \Big) z(t) \, d t .
\end{equation}
for all smooth curves $z$ with $z(0)=z(T)=0$.

Given \eqref{eq:oc-el-pf-zi}, we claim that 
\begin{equation}
\label{eq:oc-el-pf-i}
\partial_{x_{i}} L( x(t), \xdot(t)) - \frac{d}{d t} \partial_{v_{i}} L( x(t), \xdot(t)) = 0
\end{equation}
for all $t \in [0,T]$.
To see \eqref{eq:oc-el-pf-i}, we can use a proof of contradiction: If $\partial_{x_{i}} L( x(t), \xdot(t)) - \frac{d}{d t} \partial_{v_{i}} L( x(t), \xdot(t)) > 0$ at some $t \in [0,T]$, then there exists a subinterval $(t-\delta, t+\delta)$ (or half of this subinterval if $t=0$ or $t=T$) where $\partial_{x_{i}} L( x(t), \xdot(t)) - \frac{d}{d t} \partial_{v_{i}} L( x(t), \xdot(t)) > 0$ due to the smoothness of $L$. Taking $z>0$ in the interval $(t-\delta,t+\delta)$ and $0$ elsewhere, we have 
\begin{align*}
0 & < \int_{t-\delta}^{t+\delta} \Big( \partial_{x_{i}} L( x(t), \xdot(t)) - \frac{d}{d t} \partial_{v_{i}} L( x(t), \xdot(t)) \Big) z(t) \, d t \\
& = \int_{0}^{T} \Big( \partial_{x_{i}} L( x(t), \xdot(t)) - \frac{d}{d t} \partial_{v_{i}} L( x(t), \xdot(t)) \Big) z(t) \, d t 
\end{align*}
which contradicts to \eqref{eq:oc-el-pf-zi}.

Since \eqref{eq:oc-el-pf-i} holds for any $i=1,\dots,d$, by rearranging \eqref{eq:oc-el-pf-i} we obtain the Euler--Lagrange equation \eqref{eq:oc-el}.
\end{proof}

Notice that the Euler--Lagrange equation \eqref{eq:oc-el} is a nonlinear system of $d$ second-order ordinary differential equations (ODEs)\index{Differential equation!ordinary} of $x$. These ODEs are of form \eqref{eq:oc-el-pf-i}:
\begin{equation*}
\frac{d}{d t} \partial_{v_{i}} L( x(t), \xdot(t)) = \partial_{x_{i}} L( x(t), \xdot(t)) 
\end{equation*}
for $i = 1,\dots, d$.

We also remark that, the smoothness requirements of $L$ and $x$ can be lessen (e.g., the involved gradients/derivatives exist except for some measure zero set on $[0,T]$). In such cases, a rigorous proof can be obtained by using measure theory, while the conclusions remain the same.

\subsection{Hamiltonian Dynamics}

The Euler--Lagrange equation \eqref{eq:oc-el} provides a necessary condition of minimizers of \eqref{eq:oc-int-lagrangian}, where the constraints of $x$ are set at the endpoints of $[0,T]$. The velocity $\xdot$ can be thought of as the control of $x$.

Next, we introduce another important concept, called Hamiltonian dynamics\index{Hamiltonian!dynamics}, which is closely related to the Euler--Lagrange equation \eqref{eq:oc-el} but more prevalent in general optimal control problems.
In contrast to \eqref{eq:oc-el} which is a system of $d$ second-order ODEs, the Hamiltonian dynamics form a system of $2d$ first-order ODEs.

To this end, let $x: [0,T] \to \mathbb{R}^{d}$ be given, we define another curve $p: [0,T] \to \mathbb{R}^{d}$, called the \emph{generalized momentum}\index{Generalized momentum}, as follows:
\begin{equation}
\label{eq:oc-p}
p(t):=\partial_{v} L( x(t), \xdot(t))
\end{equation}
for all $t \in [0,T]$. 
Our goal is to rewrite \eqref{eq:oc-el} as a system of $2d$ first-order ODEs of $(x,p)$.

Note that, the conversion from the Euler--Lagrange equations to a Hamiltonian dynamics generally requires that 
\begin{equation}
\label{eq:oc-p-req}
p = \partial_{v} L(x, v)
\end{equation}
can determine a unique $v$ given $(x,p)$. This can be achieved by assuming that, for example, $\partial_{vv}^{2}L(x,v)$ is nonsingular at every $x$ and $v$, so that the Implicit Function Theorem can be applied to show this requirement is satisfied.
In the remainder of this section, we assume that this requirement on \eqref{eq:oc-p-req} is satisfied and we have its solution denoted by $v(x, p)$. Namely, $p = \partial_{v} L(x, v(x,p))$.

\begin{definition}
[Hamiltonian]
\label{def:oc-hamiltonian}
Let $x, p \in \mathbb{R}^{d}$ and $v(x,p)$ be the solution to \eqref{eq:oc-p-req}, then we define $H: \mathbb{R}^{d} \times \mathbb{R}^{d} \to \mathbb{R}$ by
\begin{equation}
\label{eq:oc-hamiltonian}
H(x, p)=p \cdot  v(x, p)-L(x,  v(x, p)) .
\end{equation}
We call $H$ the \emph{Hamiltonian}\index{Hamiltonian} function.
\end{definition}

Similar to the Lagrangian function $L$ above, we denote the partial derivatives of $H$ as
\begin{align*}
\partial_{x} H & :=\left(\partial_{x_{1}} H, \ldots, \partial_{x_{d}} H \right), \\
\partial_{p} H & :=\left(\partial_{p_{1}} H, \ldots, \partial_{p_{d}} H \right),
\end{align*}
where
\begin{equation*}
\partial_{x_{i}} H = \frac{\partial H}{\partial x_{i}} \quad \text{and} \quad \partial_{p_{i}} H = \frac{\partial H}{\partial p_{i}}
\end{equation*}
for every $i = 1,\dots, d$.
Now we ready to show that the Euler--Lagrange equation \eqref{eq:oc-el} can be converted to Hamiltonian dynamics using $H$.

\begin{theorem}
[Hamiltonian dynamics]
\label{thm:hamiltonian-ode}
Let $x$ be the solution to the Euler--Lagrange equation \eqref{eq:oc-el} and define $p$ as in \eqref{eq:oc-p}. Then the pair $(x,p): [0,T] \to \mathbb{R}^{d} \times \mathbb{R}^{d}$ solves the Hamiltonian dynamics\index{Hamiltonian dynamics}:
\begin{equation}
\label{eq:hamiltonian-ode}
\begin{cases}
\xdot(t)=\partial_{p} H( x(t),  p(t)) , \\
\pdot(t)=-\partial_{x} H( x(t),  p(t)) .
\end{cases}
\end{equation}
Moreover, $H( x(t),  p(t))$ is a constant function of $t$.
\end{theorem}

\begin{proof}
By the definition of Hamiltonian $H$ in \eqref{eq:oc-hamiltonian}, where $v= v(x, p)$ solves $p=\partial_{v} L(x, v)$, we have 
\begin{align}
\partial_{x} H(x, p) & = p \cdot \partial_{x}  v(x,p) - \partial_{x} L(x,  v(x, p))-\partial_{v} L(x,  v(x, p)) \cdot \partial_{x}  v(x,p) \nonumber \\
& =-\partial_{x} L(x,  v(x, p)) \label{eq:oc-dhdx-dldx}
\end{align}
since $p=\partial_{v} L(x,v(x,p))$. 
Therefore, we have
\begin{align*}
\pdot(t) 
& = \frac{d}{d t} \Big( \partial_{v} L( x(t), \xdot(t)) \Big) \\
& = \partial_{x} L( x(t), \xdot(t)) \\
& =\partial_{x} L( x(t),  v( x(t),  p(t))) \\
& =-\partial_{x} H( x(t),  p(t)) ,
\end{align*}
where the first equality is obtained by taking derivatives on both sides of \eqref{eq:oc-p} with respect to $t$,
the second equality is due to the Euler--Lagrange equation \eqref{eq:oc-el}, the third equality is because $ p(t)=\partial_{v} L( x(t), \xdot(t))$ which implies $\xdot(t)= v( x(t),  p(t))$, and the last equality is due to \eqref{eq:oc-dhdx-dldx}.

Furthermore, we have from the definition of $H$ in \eqref{eq:oc-hamiltonian} that
\begin{align*}
\partial_{p} H(x, p) & = v(x, p)+p \cdot \partial_{p}  v(x,p) -\partial_{v} L(x,  v(x, p)) \cdot \partial_{p}  v(x,p) \\
& = v(x, p)
\end{align*}
since $p=\partial_{v} L(x,  v(x, p))$. Therefore, we have
\begin{equation*}
\partial_{p} H( x(t),  p(t))= v( x(t),  p(t)) 
\end{equation*}
for every $t \in [0,T]$. 
Again, since $p(t)=\partial_{v} L( x(t), \xdot(t))$ implies $\xdot(t)= v( x(t),  p(t))$, we get
\begin{equation*}
\xdot(t)=\partial_{p} H( x(t),  p(t)) .
\end{equation*}
At this point, we have shown that $(x(t),p(t))$ satisfies the system of ODEs \eqref{eq:hamiltonian-ode}.

To show that $H(x(t),p(t))$ is a constant, we take its derivative with respect to $t$:
\begin{align*}
\frac{d}{d t} H( x(t),  p(t))
& = \partial_{x} H(x(t),p(t)) \cdot \xdot(t)+\partial_{p} H(x(t),p(t)) \cdot \pdot(t) \\
& = - \pdot(t) \cdot \xdot(t) + \xdot(t) \cdot \pdot(t) \\
& = 0 ,
\end{align*}
where the second equality is due to \eqref{eq:hamiltonian-ode}.
This completes the proof.
\end{proof}

\begin{example}
We consider an example of classical mechanics where the Lagrangian function is given by
\begin{equation}
\label{eq:oc-ex-L}
L(x, v)=\frac{1}{2}m|v|^{2} - V(x),
\end{equation}
where $x$ is the position, $m$ is the mass, and $v$ is the velocity of an object. 
The function $V: \mathbb{R}^{d} \to \mathbb{R}$ is the potential energy.
Hence, $L$ in \eqref{eq:oc-ex-L} is the difference between the kinetic energy and the potential energy.

We also see that
\begin{equation*}
\partial_{x} L(x,v)=-\nabla V(x) \quad \text{and} \quad \partial_{v} L(x,v)=m v .
\end{equation*}
Plugging these into the Euler--Lagrange equation \eqref{eq:oc-el}, we obtain
\begin{equation*}
m \ddot{x} (t) = - \nabla V(x(t)) .
\end{equation*}
Therefore, the Euler--Lagrange equation of the function \eqref{eq:oc-ex-L} is the second-order ODE describing the Newton's second law of motion. Furthermore, the generalized momentum is exactly the momentum in classical mechanics:
\begin{equation*}
p=\partial_{v} L(x, v)=m v,
\end{equation*}
which implies 
\begin{equation*}
v = v(x,p) = \frac{p}{m} .
\end{equation*}
Plugging $x$ and $v(x,p) = \frac{p}{m}$ into the Hamiltonian $H$ \eqref{eq:oc-hamiltonian}, we obtain
\begin{align*}
H(x, p)
& =p \cdot \frac{p}{m} - L \Big(x, \frac{p}{m}\Big) \\
& =\frac{|p|^{2}}{m}-\frac{1}{2} m \frac{|p|^{2}}{m^{2}} +V(x) \\
& =\frac{|p|^{2}}{2 m}+V(x) ,
\end{align*}
which is the sum of the kinetic energy and the potential energy. 
If we substitute $p(t) = m v(t)$ back into $H$ above, then we obtain
\begin{equation*}
H(x(t),p(t)) = \frac{1}{2} m |v(t)|^{2} + V(x(t)) .
\end{equation*}
By Theorem \ref{thm:hamiltonian-ode}, we know $H(x(t),p(t))$ remains as a constant over $t$.
This means that the sum of the kinetic energy and the potential energy does not change over time, namely, the total energy is conserved.

We also notice that the Hamiltonian dynamics are
\begin{equation*}
\begin{cases}
\xdot(t)= \partial_{p}H(x(t),p(t)) = \frac{ p(t)}{m} , \\
\pdot(t)= - \partial_{x}H(x(t),p(t)) = -\nabla V( x(t)) .
\end{cases}
\end{equation*}
If $V(x):=mgx$, where $x$ is the height and $g$ is the acceleration due to gravity (approximately $9.81 \text{m}/\text{sec}^{2}$ on earth surface), then there is $\pdot(t) = -mg$. 
\end{example}

\subsection{Pontryagin Maximum Principle}

The discussions above reveal some fundamental connections between Lagrangian mechanics and Hamiltonian mechanics. 
Meanwhile, the Hamiltonian dynamics demonstrate significant importance in control theory: The necessary condition of an optimal control can be described using the Hamiltonian dynamics. This is our goal in this subsection.

To begin with, let us consider an illustrative example of optimal control problem on commodity trading.
Suppose we are a wheat trader with $x_{1}(t)$ cash on hand and $x_{2}(t)$ units of wheat in stock at time $t$.
At the initial time $t=0$, we have $x_{1}(0) = x_{1,0}$ and $x_{2}(0) = x_{2,0}$ for some given $x_{1,0}, x_{2,0} \ge 0$.
Let $c>0$ be the cost of storing a unit of wheat for a unit of time.
Assume for simplicity that we know the price $q(t)$ of a unit amount of wheat at any time $t$. 
Now we need to find the optimal $u(t)$, which is the rate of buying or selling wheat at time $t$ ($u(t)>0$ if buying and $<0$ if selling), such that the total holdings at the prescribed end time $T$ is maximized.

To form this commodity trading problem mathematically, we notice that the dynamics of $x(t) = (x_{1}(t), x_{2}(t)) \in \mathbb{R}^{2}$ is given by
\begin{equation}
\label{eq:oc-ex-trade-ode}
\begin{cases}
\dot{x}_{1}(t) = - c x_{2}(t) - q(t) u(t) , \\
\dot{x}_{2}(t) = u(t) ,
\end{cases}
\end{equation}
with initial values $x_{1}(0) = x_{1,0}$ and $x_{2}(0) = x_{2,0}$. 
The reason of \eqref{eq:oc-ex-trade-ode} is that, the rate of spending money, $\dot{x}_{1}(t)$, is equal to the cost of storing the wheat in stock $x_{2}(t)$ and expense of buying (or income by selling) wheat at the rate $u(t)$; and the change rate of wheat in stock $\dot{x}_{2}(t)$ is equal $u(t)$.
The total holdings at time $T$ is given by
\begin{equation}
\label{eq:oc-ex-trade-obj}
I[u] := x_{1}(T) + q(T) x_{2}(T) ,
\end{equation}
which is the sum of cash on hand $x_{1}(T)$ and the worth of wheat in stock $q(T)x_{2}(T)$ at time $T$.
Therefore, the optimal control problem is to find the rate $u: [0,T] \to \mathbb{R}$ such that the total holdings $I$ defined in \eqref{eq:oc-ex-trade-obj} is maximized under the constraint \eqref{eq:oc-ex-trade-ode} with the initial value of $x(0)=(x_{1}(0),x_{2}(0))=(x_{1,0},x_{2,0})$.
Here we allow $u(t)$ to be any number for simplicity.
In practice, we often require $u(t) \in \Ucal$ at every $t$, where $\Ucal$ is a subset of $\mathbb{R}$ (or $\mathbb{R}^{m}$ if $u(t)$ is an $m$-dimensional vector).
In what follows, we use $\Ucal_{T} := \{ u: [0,T] \to \Ucal \subset \mathbb{R}^{m} \}$ to denote the admissible set of the control $u$ over the entire time window $[0,T]$.

The commodity trading example above represents a typical formation of general optimal control\index{Optimal control} problems. 
More precisely, these optimal control problems are formulated as
\begin{subequations}
\label{eq:oc}
\begin{align}
\max_{u \in \Ucal_{T}} \quad & I[u] := \int_{0}^{T} r(x(t),u(t)) \, dt + g(x(T)) , \label{eq:oc-obj} \\
\text{s.t.} \quad & \xdot(t) = f(x(t), u(t)) , \label{eq:oc-ode} \\
& x(0) = x_{0} , \label{eq:oc-init}
\end{align}
\end{subequations}
where the variable to be optimized is the control trajectory $u \in \Ucal_{T}$.

By comparing the commodity trading example above with \eqref{eq:oc}, we see that the dynamics \eqref{eq:oc-ex-trade-ode} corresponds to the ODE constraint \eqref{eq:oc-ode}, the initial value $x(0)=(x_{1,0},x_{2,0})$ corresponds to \eqref{eq:oc-init}, and the total holdings $x_{1}(T) + q(T) x_{2}(T)$ in \eqref{eq:oc-ex-trade-obj} is $g(x(T))$ in \eqref{eq:oc-obj}, which is called the \emph{terminal reward}\index{Reward!terminal} (or terminal payoff).
In \eqref{eq:oc-obj}, we have an additional term $\int_{0}^{T} r(x(t),u(t)) \, dt$ that did not appear in the commodity trading example \eqref{eq:oc-ex-trade-obj}. This is the integral of the so-called \emph{running reward}\index{Reward!running} $r: \mathbb{R}^{d} \times \mathbb{R}^{m} \to \mathbb{R}$, which can be easily merged into the terminal reward $g$ as we will show later.

Note that we did not include the time variable $t$ in the ODE defining function $f$ in \eqref{eq:oc-ode} explicitly and make it in the form of $f(t,x(t),u(t))$.
This is because the time variable $t$ can be easily merged into $x(t)$ by introducing an additional component $t$ with constant derivative $1$. More precisely, we can define 
\begin{align*}
\bar{x}(t) & := (t, x(t)) , \\
\bar{f}(\bar{x}(t),u(t)) & := (1, f(t,x(t),u(t))) ,
\end{align*}
then there is 
\begin{equation*}
\xdot(t) = f(t,x(t),u(t)) \quad \Longleftrightarrow \quad \dot{\bar{x}}(t) = \bar{f}(\bar{x}(t),u(t))
\end{equation*}
and the right-hand side is still of the form \eqref{eq:oc-ode}.

The optimal control problem \eqref{eq:oc} is often termed as a \emph{fixed end-time and free endpoint}\index{Optimal control!fixed end-time and free endpoint} problem, because the terminal time $T$ is fixed and the endpoint $x(T)$ is not.
There is another variation of optimal control, called fixed endpoint and free end-time problem, which has similar optimality theory as \eqref{eq:oc}.
We will cover that variation in this section as well.

Next, we represent one of the most significant results about optimality condition of solutions to \eqref{eq:oc}. This result is known as the \emph{Pontryagin Maximum Principle}\index{Pontryagin Maximum Principle}. We first give the definition of control Hamiltonian function, which will be used in the Pontryagin Maximum Principle and its proof. 

\begin{definition}
[Control Hamiltonian]
\label{def:oc-control-hamiltonian}
Let $f$ and $r$ be the ODE defining function and the running reward function, respectively, in \eqref{eq:oc}. The \emph{control Hamiltonian}\index{Control Hamiltonian} function $H: \mathbb{R}^{d} \times \mathbb{R}^{d} \times \Ucal \to \mathbb{R}$ is defined by
\begin{equation}
\label{eq:oc-control-hamiltonian}
H(x, p, u) := p \cdot f(x,u) + r(x, u) 
\end{equation}
for any $(x,p,u) \in \mathbb{R}^{d} \times \mathbb{R}^{d} \times \Ucal$.
\end{definition}

Comparing with the Hamiltonian function in \eqref{eq:oc-hamiltonian}, we see that the control Hamiltonian function in \eqref{eq:oc-control-hamiltonian} has the control $u \in \Ucal \subset \mathbb{R}^{m}$ as another variable, the control velocity $v$ in \eqref{eq:oc-hamiltonian} is replaced with a general ODE defining function $f$ controlled by $u$ in \eqref{eq:oc-ode}, and the negative Lagrangian $-L$ (where the integral \eqref{eq:oc-int-lagrangian} of $L$ is to be minimized) is replaced with the running reward function $r$ (to be maximized) in \eqref{eq:oc-obj}.

Now we provide two lemmas that are crucial in the Pontryagin Maximum Principle and its proof. 
%
% The first lemma is particularly important as it also provides a means to train deep neural networks for solving optimal control problems in Section \ref{sec:node}.

\begin{lemma}
\label{lem:pmp-lemma1-x-ode}
Suppose $x: [0,T] \to \mathbb{R}^{d}$ solves the initial value problem 
\begin{equation}
\label{eq:pmp-lemma1-x-ode}
\begin{cases}
\xdot (t)= f (x(t), u(t)) , & \forall \, t \in [0,T] , \\
x(0) = x_{0} &
\end{cases}
\end{equation}
for some given defining function $f:\mathbb{R}^{d} \times \mathbb{R}^{m} \to \mathbb{R}^{d}$, control function $u:[0,T] \to \mathbb{R}^{m}$, and initial value $x_{0} \in \mathbb{R}^{d}$. Assume that $f(\cdot,u(t))$ is $L_{f}$-Lipschitz continuous and $C^{2}$ in $x$ for every $t$. Let $y_{0} \in \mathbb{R}^{d}$ be arbitrary and fixed. For any $\epsilon > 0$, let $\xeps$ solve the initial value problem
\begin{equation}
\label{eq:pmp-lemma1-xeps-ode}
\begin{cases}
\xepsdot (t)= f (\xeps(t), u(t)) , \\
\xeps(0) = x_{0} + \epsilon y_{0} + o(\epsilon).
\end{cases}
\end{equation}
Then there is
\begin{equation}
\label{eq:pmp-lemma1-xeps}
\xeps(t) = x(t) + \epsilon y(t) + o(\epsilon)
\end{equation}
uniformly in $t$ on $[0,T]$, namely, 
\begin{equation*}
\lim_{\epsilon \to 0} \sup_{t \in [0,T]}  \frac{|\xeps(t) - x(t) - \epsilon y(t)|}{ \epsilon } = 0 ,
\end{equation*}
where $y$ is the solution to the initial value problem
\begin{equation}
\label{eq:pmp-lemma1-y}
\begin{cases}
\ydot (t)= \partial_{x} f (x(t), u(t)) y(t), \\
y(0) = y_{0} .
\end{cases}
\end{equation}
\end{lemma}

\begin{proof}
We define 
\begin{equation}
\delta(t) := \frac{1}{2} |\xeps(t) - x(t)|^{2} .
\end{equation}
Then there is
\begin{align*}
\dot{\delta}(t) 
& = (\xeps(t) - x(t) ) \cdot (\xepsdot(t) - \xdot(t)) \\
& = (\xeps(t) - x(t) ) \cdot (f(\xeps(t),u(t)) - f(x(t),u(t))) \\
& = (\xeps(t) - x(t) ) \cdot \partial_{x} f(\xi^{\epsilon}(t), u(t)) (\xeps(t) - x(t)) \\
& \le 2 L_{f} \delta(t) ,
\end{align*}
where $\xi^{\epsilon}(t)$ is a point on the line segment between $x(t)$ and $x^{\epsilon}(t)$ due to Taylor theorem.
By Gr\"{o}nwall inequality, we have for all $t \in [0,T]$ that
\begin{equation*}
\delta(t) = \delta(0) e^{2L_{f}t} \le \delta(0) e^{2L_{f}T}.
\end{equation*}
Therefore, 
\begin{equation*}
|\xeps(t) - x(t)| = \sqrt{2 \delta(t)} \le \sqrt{2 \delta(0)} e^{L_{f}T} = |\epsilon y_{0} + o(\epsilon)| e^{L_{f}T} \le C \epsilon |y_{0}| e^{L_{f}T} 
\end{equation*}
for all $t \in [0,T]$ when $\epsilon$ is sufficiently small, where $C$ is some constant independent of $\epsilon$ and $t$.
Therefore, $\| \xeps - x \|_{L^{\infty}([0,T])} \to 0$ as $t\to 0$, i.e., $\xeps(t) - x(t) = O(\epsilon)$ and converges to 0 as $\epsilon \to 0$ uniformly on $[0,T]$.

Define for every $t \in [0,T]$ that
\begin{equation*}
y(t) := \lim_{\epsilon \to 0} \frac{\xeps(t) - x(t)}{\epsilon} .
\end{equation*}
Then clearly
\begin{equation*}
y(0) = \lim_{\epsilon \to 0} \frac{\xeps(0) - x(0)}{\epsilon} = \lim_{\epsilon \to 0} \Big(y_{0} + \frac{o(\epsilon)}{\epsilon} \Big) = y_{0} ,
\end{equation*}
and
\begin{align*}
\ydot(t)
& = \lim_{\epsilon \to 0} \frac{\xepsdot(t) - \xdot(t)}{\epsilon} \\
& = \lim_{\epsilon \to 0} \frac{f(\xeps(t),u(t)) - f(x(t),u(t))}{\epsilon} \\
& = \lim_{\epsilon \to 0} \frac{1}{\epsilon} \Big( \partial_{x} f(x(t),u(t)) (\xeps(t) - x(t)) \\
& \qquad \qquad + \frac{1}{2} (\xeps(t) - x(t)) \partial_{xx}^{2}f(\eta^{\epsilon}(t), u(t)) (\xeps(t) - x(t)) \Big)\\
& = \lim_{\epsilon \to 0} \Big( \partial_{x} f(x(t),u(t)) \frac{\xeps(t) - x(t)}{\epsilon} \\
& \qquad \qquad + \frac{\epsilon}{2} \frac{\xeps(t) - x(t)}{\epsilon} \partial_{xx}^{2}f(\eta^{\epsilon}(t), u(t)) \frac{\xeps(t) - x(t)}{\epsilon} \Big)\\
& = \partial_{x} f(x(t),u(t)) y(t) ,
\end{align*}
where $\eta^{\epsilon}(t)$ is a point on the line segment between $x(t)$ and $x^{\epsilon}(t)$.
Note that the first equality requires $\partial_{\epsilon} \partial_{t} \xeps (t) = \partial_{t} \partial_{\epsilon}\xeps(t)$. A sufficient condition for this to hold is the Clairaut theorem: If both second-order partial derivatives exist and continuous, the order of partial derivatives can be exchanged. This holds when $f$ is sufficiently smooth, e.g., $f$ is $C^{1}$.

To conclude, we have $\xeps(t) = x(t) + \epsilon y(t) + o(\epsilon)$ uniformly on $[0,T]$, where $y$ solves the initial value problem \eqref{eq:pmp-lemma1-y}. 
\end{proof}

In order to prove the Pontryagin Maximum Principle rigorously, we need to be inclusive about all possible variations of a control trajectory. Theoretically speaking, the optimal control problem does not assume the control $u$ to be continuous in $t$. Therefore, we consider the following so-called simple variation\index{Simple variation} which allows discontinuous changes of a control function $u$.

\begin{definition}
[Simple variation]
For any $\tau \in (0,T]$ and $\epsilon$ sufficiently small such that $0 < \tau - \epsilon < \tau \le T$, define the \emph{simple variation} (also called \emph{needle-shaped} or \emph{needle-like variation}) of $u(t)$ as
\begin{equation}
\label{eq:simple-var}
\ueps(t):=
\begin{cases}
w, & \text{if} \ \tau - \epsilon < t < \tau , \\
u(t) , & \text{otherwise} .
\end{cases}
\end{equation}
\end{definition}

With the simple variation $\ueps$, we let $\xeps$ (note that this is no longer the one defined in \eqref{eq:pmp-lemma1-xeps-ode} above) be the solution to the initial value problem
\begin{equation}
\label{eq:oc-x-perturb}
\begin{cases}
\xepsdot (t) = f(\xeps(t), \ueps(t)) , \quad \forall\, t \in [0,T] ,\\
\xeps(0) = x_{0} .
\end{cases}
\end{equation}

It is important to quantify the difference between the original trajectory $x$ and this $\xeps$.
To this end, we require that $f \in C^{1}(\mathbb{R}^{d} \times \mathbb{R}^{m}; \mathbb{R}^{d})$ hereafter.
Then we show that the function $y$ defined below captures the first-order term of their difference.

\begin{definition}
[Perturbation]
\label{def:oc-y-perturb}
Let $y(t) = 0$ for $t \in [0,\tau)$ and $y(t)$ solves the initial value problem
\begin{equation}
\label{eq:oc-y-perturb}
\begin{cases}
\ydot(t) = \partial_{x} f(x(t), u(t)) y(t) , \\
y(\tau) = f(x(t),w) - f(x(t),u(t))
\end{cases}
\end{equation}
for $t \in [\tau, T]$. We call $y$ a \emph{perturbation} function\index{Perturbation function}.
\end{definition}

\begin{lemma}
\label{lem:pmp-lemma2-y}
Suppose that $\xeps$ solves the problem \eqref{eq:oc-x-perturb} with the simple variation $\ueps$, then
\begin{equation}
\xeps(t) = x(t) + \epsilon y(t) + o(\epsilon) 
\end{equation}
as $\epsilon \to 0$ uniformly on $[0,T]$, where $y:[0,T] \to \mathbb{R}^{d}$ is the perturbation function in Definition \ref{def:oc-y-perturb}.
\end{lemma}

\begin{proof}
Let $t \in (0,\tau)$ be arbitrary and fixed, then for any $\epsilon \in (0, \tau -t)$, there is $0 < t < \tau-\epsilon$ and hence $\xeps(t) = x(t)$.

For $t \in [\tau, T]$, we have
\begin{align*}
\xeps(t) - x(t)
& = \int_{\tau - \epsilon}^{t} \Big( f(x(s), w) - f(x(s), u(s)) \Big) \, ds \\
& = \Big( f(x(\tau), w) - f(x(\tau), u(\tau)) \Big) \epsilon + o(\epsilon)
\end{align*}
as $\epsilon \to 0$
by the continuity of $f$ in $u$. Since $x(t)$ and $\xeps(t)$ solve the same ODE with defining function $f$ and $u$ over $[\tau, T]$ but only differ in the initial 
\begin{equation*}
\xeps(\tau) = x(\tau) + \Big( f(x(\tau), w) - f(x(\tau), u(\tau)) \Big) \epsilon + o(\epsilon)
\end{equation*} 
at time $\tau$. By Lemma \ref{lem:pmp-lemma1-x-ode}, we know $\xeps(t) = x(t) + \epsilon y(t) + o(\epsilon)$ for all $t$ as $\epsilon \to 0$ where $y$ is the perturbation function in Definition \ref{def:oc-y-perturb}.
\end{proof}

Before we prove the Pontryagin Maximum Principle, we also introduce the concept of co-state $p: [0,T] \to \mathbb{R}^{d}$ for the state $x: [0,T] \to \mathbb{R}^{d}$ in a control system.
It is convenient to think $p(t) \in \mathbb{R}^{d}$ as a row vector (notice that it has the same dimension as $x(t)$) at every $t$ and behaves like the Lagrange multiplier to the constraint formed as the initial value problem \eqref{eq:pmp-lemma1-x-ode}.
The co-state is also called the adjoint function, and it satisfies the so-called adjoint equation, as shown below.

\begin{definition}
[Co-state and adjoint equation]
Let $x$ be the state solving the initial value problem \eqref{eq:pmp-lemma1-x-ode} with control $u$. Then $p: [0,T] \to \mathbb{R}^{d}$ is called the \emph{co-state}\index{Co-state} if it solves the following terminal value problem 
\begin{equation}
\label{eq:adj-p}
\begin{cases}
\pdot(t) = - p(t) \partial_{x} f(x(t), u(t)) - \partial_{x} r(x(t), u(t)), & \forall \, t \in [0,T] , \\
p(T) = \nabla g(x(T)) .
\end{cases}
\end{equation}
The ODE in \eqref{eq:adj-p} is called the \emph{adjoint equation}\index{Adjoint equation}.
\end{definition}

\subsubsection*{Pontryagin Maximum Principle without Running Reward}

The Pontryagin Maximum Principle provides a necessary condition of the optimal control. Therefore, its establishment requires the assumption that the given control is optimal (just like we assume a local minimizer attains the smallest objective function value in its neighbor in optimization when we derive the necessary optimality condition). 
Now our objective function $I[u]$ is given in \eqref{eq:oc-obj}. 
We begin with the case where $I$ only has the terminal reward $g(x(T))$.

\begin{lemma}
The variation of terminal reward $I[u] = g(x(T))$ (assuming there is no running reward) in \eqref{eq:oc-obj} is given by
\begin{equation}
\label{eq:pmp-lemma3-dI}
\frac{d}{d \epsilon} I[ \ueps ] \Big|_{\epsilon = 0} = p(\tau) \Big( f(x(\tau), w ) - f(x(\tau),u(\tau)) \Big) .
\end{equation}
\end{lemma}

\begin{proof}
According to Lemma \ref{lem:pmp-lemma2-y}, we have
\begin{equation*}
I[\ueps] = g(\xeps(T)) = g(x(T) + \epsilon y(T) + o(\epsilon) ) ,
\end{equation*}
where $y$ is the perturbation function in Definition \ref{def:oc-y-perturb}. 
Therefore,
\begin{equation}
\label{eq:pmp-lemma3-pf-1}
\frac{d}{d \epsilon} I[ \ueps ] \Big|_{\epsilon = 0} = \nabla g(x(T)) y(T).
\end{equation}
On the other hand, we have by the adjoint equation \eqref{eq:adj-p} without $r$ that
\begin{align*}
\frac{d}{d t} (p(t) y(t)) 
& = \pdot(t) y(t) + p(t) \ydot(t) \\
& = - p(t) \partial_{x} f(x(t), u(t)) y(t) + p(t) \partial_{x} f(x(t), u(t)) y(t) \\
& = 0 ,
\end{align*}
which implies that $p(t)y(t)$ is a constant in $t$.
Hence there is 
\begin{equation}
\label{eq:pmp-lemma3-pf-2}
\nabla g(x(T)) y(T) = p(T) y(T) = p(\tau) y(\tau) ,
\end{equation}
where the first equality is due to the terminal condition of $p$ in \eqref{eq:adj-p}.
Notice that the value $y(\tau)$ is given in \eqref{eq:oc-y-perturb}, we obtain \eqref{eq:pmp-lemma3-dI} by combining \eqref{eq:pmp-lemma3-pf-1} and \eqref{eq:pmp-lemma3-pf-2}.
\end{proof}

Now we are ready to present and prove the Pontryagin Maximum Principle\index{Pontryagin Maximum Principle}. Again, we consider the optimal control problem \eqref{eq:oc} without running reward in the objective function.

\begin{theorem}
[Pontryagin Maximum Principle]
\label{thm:oc-pmp}
Let $ u(\cdot)$ be the optimal solution to \eqref{eq:oc} and $x$ the corresponding trajectory satisfying \eqref{eq:oc-ode} and \eqref{eq:oc-init}.
Then there exists a function $p:[0, T] \rightarrow \mathbb{R}^{n}$ such that $(x,p)$ satisfies the control Hamiltonian dynamics:
\begin{equation}
\label{eq:oc-control-h-ode}
\begin{cases}
\xdot(t)=\partial_{p} H\left( x(t),  p(t),  u(t)\right) , \\
\pdot(t)=-\partial_{x} H\left( x(t),  p(t),  u(t)\right) ,
\end{cases}
\end{equation}
with initial condition of $x$ and terminal condition of $p$ given by
\begin{equation*}
x(0) = x_{0} \quad \text{and} \quad p(T) = \nabla g(x(T)) .
\end{equation*}
Moreover, the following identity, known as the \emph{Pontryagin Maximum Principle}\index{Pontryagin Maximum Principle}, holds true for every $t$:
\begin{equation}
\label{eq:pmp}
H( x(t),  p(t),  u(t)) \,= \, \max _{w \in \Ucal} H( x(t),  p(t), w ).
\end{equation}
%
% In addition, $H(x(t),p(t),u(t))$ is a constant function of $t$.
\end{theorem}

\begin{proof}
Without the running reward $r$, we have the control Hamiltonian \eqref{eq:oc-control-hamiltonian} as
\begin{equation*}
H(x,p,u) = p \cdot f(x,u) .
\end{equation*}
Hence \eqref{eq:oc-control-h-ode} is the combination of the ODE of $x$ given in \eqref{eq:oc-ode} and the adjoint equation of $p$ given in \eqref{eq:adj-p}. The initial condition of $x$ and terminal condition of $p$ are given in \eqref{eq:oc-init} and \eqref{eq:adj-p}, respectively. 

Let $t \in (0,T)$ be fixed and $w \in \Ucal$ be arbitrary, then $\epsilon \mapsto I[\ueps]$ has a maximum at $\epsilon = 0$ provided that $u$ is optimal. Hence
\begin{equation}
0 \ge \frac{d}{d \epsilon} I[\ueps] \Big|_{\epsilon = 0} = p(\tau) \Big( f(x(t), w) - f(x(t), u(t)) \Big) .
\end{equation}
Therefore, we obtain
\begin{align*}
H(x(t), p(t), w) 
& = p(t) f(x(t), w) \\
& \le p(t) f(x(t), u(t)) \\
& = H(x(t), p(t), u(t)) .
\end{align*}
In addition, since $u(t) \in \Ucal$, we obtain the maximum principle \eqref{eq:pmp}.
\end{proof}

% \paragraph{Connection to Hamiltonian system without control}
%
Notice that the Hamiltonian dynamics with control \eqref{eq:oc-control-h-ode} has the same form as the one \eqref{eq:hamiltonian-ode} without control.
To memorize these dynamics, just keep in mind that they are nonlinear systems of first-order ODEs of the state and co-state pair $(x,p)$, and the defining function is $(\partial_{p}H, -\partial_{x} H)$. 
Such dynamics structure is said to be in the \emph{symplectic form}.

\subsubsection*{Pontryagin Maximum Principle with Running Reward}

Now we consider the general case where the objective function $I[u]$ includes a running reward:
\begin{equation}
\label{eq:pmp-obj-w-r}
I[ u ] : = \int_{0}^{T} r( x(t),  u(t)) \, d t + g(x(T)) .
\end{equation}
In this case, the control Hamiltonian has its original form \eqref{eq:oc-control-hamiltonian}:
\begin{equation*}
H(x, p, a)= p \cdot f(x, a)+r(x, u) .
\end{equation*}
To derive the Pontryagin Maximum Principle for this case, we can simply introduce an auxiliary variable $x_{d+1}: [0,T] \to \mathbb{R}$ to record the running reward, and then the objective function can be written without any running reward, reducing to the case we discussed above. 

To this end, we consider $x_{d+1}$ defined by
\begin{equation}
\label{eq:pmp-x-aux}
\begin{cases}
\dot{x}_{d+1}(t)=r( x(t),  u(t)), & \forall \, t \in [0,T] ,  \\ 
x_{d+1}(0)=0 , & \end{cases}
\end{equation}
where $ x$ solves the original initial value problem:
\begin{equation}
\label{eq:pmp-x}
\begin{cases}
\xdot(t) = f(x(t), u(t)), & \forall \, t \in [0,T] ,  \\ 
x(0)=x_{0} . & 
\end{cases}
\end{equation}
Concatenating $x$ and $x_{d+1}$ as a single augmented state variable, we have
\begin{equation}
\label{eq:pmp-aug-x}
\bar{x} = 
\begin{pmatrix}
x \\ x_{d+1}
\end{pmatrix}
=
\begin{pmatrix}
x_{1} \\ \vdots \\ x_{d} \\ x_{d+1}
\end{pmatrix}
\in \mathbb{R}^{d+1} .
\end{equation}
Following this, we denote the defining function $\bar{f}: \mathbb{R}^{d+1} \times \Ucal \to \mathbb{R}^{d+1}$ of the augmented state variable $\bar{x}$ in \eqref{eq:pmp-aug-x} and the original control $u$ as
\begin{equation}
\label{eq:pmp-f-aug}
\bar{f}(\bar{x}, u) := 
\begin{pmatrix}
f(x,u) \\ r(x,u)
\end{pmatrix}
=
\begin{pmatrix}
f_{1}(x,u) \\ \vdots \\ f_{d}(x,u) \\ r(x,u)
\end{pmatrix}
\in \mathbb{R}^{d+1} 
\end{equation}
and the terminal reward function $\bar{g}: \mathbb{R}^{d+1} \to \mathbb{R}$ as
\begin{equation}
\label{eq:pmp-g-aug}
\bar{g}(\bar{x}):=g(x)+x_{d+1} .
\end{equation}
Notice that $\bar{f}(\bar{x},u)$ in \eqref{eq:pmp-f-aug} does not depend on $x_{d+1}$ and there is
\begin{equation*}
\partial_{\bar{x}} \bar{f}(\bar{x}, u) = 
\begin{pmatrix}
\partial_{x} f(x,u), & \partial_{x_{d+1}} f(x,u) \\
\partial_{x} r(x,u), & \partial_{x_{d+1}} r(x,u) 
\end{pmatrix}
=
\begin{pmatrix}
\partial_{x} f(x,u), & 0 \\
\partial_{x} r(x,u), & 0
\end{pmatrix}
\in \mathbb{R}^{(d+1) \times (d+1)} .
\end{equation*}

Consequently, our control problem is converted to an equivalent one with no running reward by using the terminal reward function $\bar{g}$ in \eqref{eq:pmp-g-aug}:
\begin{equation}
\label{eq:I-pmp-aug}
\min_{u \in \Ucal_{T}}\ \bar{I}[u]:=\bar{g}(\bar{x}(T)) = g(x(T)) + x_{d+1}(T) 
\end{equation}
subject to the ODE and initial value constraints by combining \eqref{eq:pmp-x-aux} and \eqref{eq:pmp-x}:
\begin{equation}
\begin{cases}
\dot{\bar{x}}(t) = \bar{f}(\bar{x}(t),  u(t)), & \forall \, t \in [0,T] ,\\ 
\bar{x}(0)=\bar{x}_{0} . & 
\end{cases}
\end{equation}

Applying Theorem \ref{thm:oc-pmp} to the control problem with the reward $\bar{I}$ in \eqref{eq:I-pmp-aug}, we obtain the Pontryagin Maximum Principle for the control Hamiltonian
\begin{equation*}
\bar{H}(\bar{x}, \bar{p}, u)=\bar{p} \cdot \bar{f}(\bar{x}, u) 
\end{equation*}
Then there exists a co-state function $\bar{p}:[0, T] \rightarrow \mathbb{R}^{d+1}$ that satisfies the adjoint equation
\begin{align}
\Bigl( \pdot(t),\ \pdot_{d+1}(t) \Bigr) 
& = \dot{\bar{p}}(t) \nonumber \\
& = - \partial_{\bar{x}} \bar{H} \Bigl(\bar{x}(t), \bar{p}(t), u(t) \Bigr) \nonumber \\
& = - \bar{p}(t) \cdot \partial_{\bar{x}} \bar{f} \Bigl(\bar{x}(t), \bar{p}(t), u(t)\Bigr) \label{eq:pmp-p-aug-ode} \\
& = - \Bigl(p(t),\ p_{d+1}(t) \Bigr) \cdot \begin{pmatrix}
\partial_{x} f(x(t),u(t)), & 0 \\
\partial_{x} r(x(t),u(t)), & 0
\end{pmatrix} \nonumber \\
& = \begin{pmatrix}
- p(t) \cdot \partial_{x} f(x(t),u(t)) - p_{d+1} \partial_{x} r(x(t),u(t)) , \ 0
\end{pmatrix} \nonumber 
\end{align}
and the terminal value
\begin{equation}
\label{eq:pmp-p-aug-terminal}
\bar{p}(T) = (p(T), \ p_{d+1}(T)) = (\nabla g(x(T)), \ 1 ) .
\end{equation}
From \eqref{eq:pmp-p-aug-ode} and \eqref{eq:pmp-p-aug-terminal}, we can see that $p_{d+1}: [0,T] \to \mathbb{R}$ satisfies the terminal value problem
\begin{equation*}
\begin{cases}
\pdot_{d+1}(t)  = 0, \quad \forall \, t \in [0, T] , \\
p_{d+1}(T) = 1 , 
\end{cases}
\end{equation*}
from which we conclude that $p_{d+1}(t) = 1$ for all $t \in [0,T]$.

With $p_{d+1}(t) = 1$, we have
\begin{equation}
\label{eq:control-H-w-running}
\bar{H}(\bar{x}, \bar{p}, u) = \bar{p} \cdot \bar{f}(\bar{x}, u) = p \cdot f(x, u) + r(x, u) = H(x, p, u) ,
\end{equation}
which is the control Hamiltonian with running reward. Meanwhile, from \eqref{eq:pmp-p-aug-ode} and \eqref{eq:pmp-p-aug-terminal}, we see that the adjoint $p$ solves
\begin{equation*}
\begin{cases}
\pdot(t)  = - p(t) \cdot \partial_{x} f(x(t),u(t)) - \partial_{x} r(x(t),u(t))  , \quad \forall \, t \in [0,T] , \\
p(T) = \nabla g(x(T)) , 
\end{cases}
\end{equation*}
where the ODE of $p$ above is exactly $\pdot(t) = - \partial_{x}H(x(t),p(t),u(t))$ with the control Hamiltonian $H$ defined in \eqref{eq:control-H-w-running}.
In summary, the Pontryagin Maximum Principle in Theorem \ref{thm:oc-pmp} holds for the optimal control problem \eqref{eq:oc} with running reward as well.

\subsubsection*{Pontryagin Maximum Principle for Fixed Endpoint Problem}

Now we consider the optimal control problem with fixed endpoint (fixed arrival state) and free end-time.
The general form of such problem is given by
\begin{subequations}
\label{eq:oc-fix-end}
\begin{align}
\max_{u \in \Ucal_{T}} \quad & I[u] := \int_{0}^{T} r(x(t),u(t)) \, dt , \label{eq:oc-fe-obj} \\
\text{s.t.} \quad & \xdot(t) = f(x(t), u(t)) , \label{eq:oc-fe-ode} \\
& x(0) = x_{0} , \label{eq:oc-fe-init} \\
& x(T) = x_{1} , \label{eq:oc-fe-end}
\end{align}
\end{subequations}
where $x_{0}, x_{1} \in \mathbb{R}^{d}$ are some given initial value and terminal value, respectively, and $T$ is the first arrival time at $x_{1}$ and depends on the control $u$.

We remark that (i) the only requirement for $u$ is that $u(t) \in \Ucal$ at any time $t$, but the terminal time $T$ can vary; (ii) the objective function in \eqref{eq:oc-fe-obj} no longer has a terminal reward; and (iii) the first arrival time $T$ is assumed to be finite by setting the running reward properly (e.g., setting $r \le -\delta$ uniformly on $\mathbb{R}^{d} \times \Ucal$ for some prescribed $\delta > 0$), such that late arrival always tend to cause reduction of accumulated running reward.

To derive the Pontryagin Maximum Principle for the fixed endpoint optimal control problem \eqref{eq:oc-fix-end}, we need to introduce more complicated control variations as follows.

\begin{definition}
[Multiple variations of a control]
For any $K \in \mathbb{N}$, choose time points $0< \tau_{1} < \tau_{2} < \cdots < \tau_{K}$, positive numbers $\lambda_{1}, \cdots, \lambda_{K} > 0$, and control variables $w_{1}, w_{2}, \dots, w_{K} \in \Ucal$.
For $\epsilon > 0$ sufficient small such that the intervals $[\tau_{k}- \lambda_{k}\epsilon, \tau_{k}]$ for $k\in [K]:=\{1,\dots,K\}$ do not overlap, we define the \emph{multiple variations} of the control $u: [0,T] \to \Ucal$ by\index{Multiple variations}
\begin{equation}
\label{eq:multi-var}
\ueps(t) := 
\begin{cases}
w_{k}, & \text{if } \tau_{k}-\lambda_{k} \epsilon \le t < \tau_{k} \ \text{for any} \ k \in [K] , \\  
u(t), & \text{otherwise}.
\end{cases}
\end{equation}
\end{definition}

For the multiple variations $\ueps$ defined in \eqref{eq:multi-var}, we let $\xeps$ be the solution to the initial value problem
\begin{equation}
\label{eq:multi-var-x}
\begin{cases}
\xepsdot(t) = f ( \xeps(t),  \ueps(t) ) & \text{for any } t \ge 0 ,\\  
\xeps(0)=x_{0} . & 
\end{cases}
\end{equation}

To ease the presentation below, we let the matrix-valued function $\Phi(t,s)$ denote the solution to the problem
\begin{equation}
\begin{cases}
\frac{d}{d t} \Phi(t,s) = \partial_{x} f ( x(t), u(t) ) & \text{for any } t \ge s \ge 0 ,\\  
\Phi(s,s)=I_{d} . & 
\end{cases}
\end{equation}
We also denote 
\begin{equation*}
y_{\tau_{k}} = f(x(\tau_{k}), w_{k}) - f(x(\tau_{k}), u(\tau_{k}) )  
\end{equation*}
for $k = 1,\dots, K$ according to the multiple variations $\ueps$ defined in \eqref{eq:multi-var}.
Then, it is straightforward to obtain the lemma below by following Lemma \ref{lem:pmp-lemma2-y}.
We omit its proof here.

\begin{lemma}
[Perturbation function]
\label{lem:oc-perturb}
Let $\xeps$ be the solution to $\ueps$ defined in \eqref{eq:multi-var-x} and $x$ the solution to the original control $u$. Then for any finite $T>0$, there is
\begin{equation*}
\max_{0 \le t \le T} \frac{| \xeps(t) - x(t) - \epsilon y(t) |}{\epsilon} \to 0 
\end{equation*}
as $\epsilon \to 0$, where $y$ is the perturbation function\index{Perturbation function}:
\begin{equation*}
y(t) =
\begin{cases}
0, & \text{if } 0 \le t < \tau_{k} , \\
\sum_{k = 1}^{k'} \lambda_{k} \Phi(t, \tau_{k}) y_{\tau_{k}}, & \text{if } \tau_{k} \le t < \tau_{k+1}, \ k' = 1, \dots, K-1 , \\
\sum_{k = 1}^{K} \lambda_{k} \Phi(t, \tau_{k}) y_{\tau_{k}}, & \text{if } \tau_{k} \le t .
\end{cases}
\end{equation*}
\end{lemma}

\begin{definition}
[Cone of variations]
We define the \emph{cone of variations}\index{Cone of variations} at time $t$ as the following set
\begin{equation}
\label{eq:cone-var}
C(t):=\left\{ \sum_{k=1}^{K} \lambda_{k}  \Phi(t, \tau_{k}) y_{\tau_{k}} \, : \, 
\begin{array}{l}
K \in \mathbb{N}, \text{ and } \lambda_{k}>0, w_{k} \in \Ucal, \\
0< \tau_{1} \le \cdots \le \tau_{K} <t, \ \forall\, k \in [K]
\end{array}
\right\} .
\end{equation}
\end{definition}

Notice that $C(t)$ is a convex cone in $\mathbb{R}^{d}$ since $\lambda_{k} \ge 0$ for all $k$, and it consists of all changes in the state $x(t)$ up to order $\epsilon$ (i.e., changes at the order $O(\epsilon)$) by applying some multiple variations of the control $u$.

The property of $C(t)$ is important for us to develop the Pontryagin Maximum Principle for the fixed endpoint problem \eqref{eq:oc-fix-end}. 
To investigate the property of $C(t)$, we need the following topological lemma.

\begin{lemma}
[Zeros of a vector field]
\label{lem:zeros-vf}
Let $S$ denote a compact (closed and bounded) and convex subset of $\mathbb{R}^{n}$, and $q$ be an interior point of $S$. Suppose $\phi: S \to \mathbb{R}^{n}$ is a continuous vector field that satisfies the following strict inequalities:
\begin{equation}
\label{eq:oc-topo-ineq}
| \phi(x) - x | < | q - x |, \quad \forall\, x \in \partial S .
\end{equation}
Then there exists $\xs \in S$ such that $\phi(\xs) = q$.
\end{lemma}

\begin{proof}
We first consider a simple case where $S = \bar{B}_{1}(0):=\{x \in \mathbb{R}^{n}: |x|\le 1\}$, which is the closed unit ball with center $0$ and radius $1$ in $\mathbb{R}^{n}$, and $q = 0 \in \mathbb{R}^{n}$ is an interior point of $\bar{B}_{1}(0)$. 
In this case, \eqref{eq:oc-topo-ineq} reduces to
\begin{equation*}
|\phi(x) - x| < |x|, \quad \forall\, x \in \partial \bar{B}_{1}(0) .
\end{equation*}
Squaring both sides above, we obtain
\begin{equation*}
|\phi(x)|^{2} - 2 \phi(x) \cdot x < 0 ,
\end{equation*}
which implies that $\phi(x) \cdot x > \frac{1}{2} |\phi(x)|^{2} \ge 0$ for all $x$ with $|x| = 1$.

Since $\bar{B}_{1}(0)$ and $\partial \bar{B}_{1}(0)$ are both compact in $\mathbb{R}^{n}$ and $\phi$ is continuous, we know there exist $M,m >0$ such that 
\begin{equation*}
\phi(x) \cdot x \ge m \quad \text{and} \quad |\phi(x) | \le M
\end{equation*}
for all $x \in \bar{B}_{1}(0)$ (including the boundary of $\bar{B}_{1}(0)$).
Choosing $\alpha \in (0, 2m/M)$ and defining $\psi(x) := x - \alpha \phi(x)$, we obtain
\begin{align*}
|\psi(x)|^{2}
& = |x - \alpha \phi(x)|^{2} = |x|^{2} - 2 \alpha \phi(x) \cdot x + \alpha^{2}|\phi(x)|^{2} \\
& \le |x|^{2} - 2m \alpha + \alpha^{2} M = |x|^{2} - \alpha (2m - \alpha M) < |x|^{2} \le 1
\end{align*}
for all $x \in \bar{B}_{1}(0)$.
By Brouwer's fixed point theorem, there exists $\xs \in \bar{B}_{1}(0)$ such that $\psi(\xs) = \xs$.
Namely, $\xs - \alpha \phi(\xs) = \xs$ or $\phi(\xs) = 0 = q$.

In the general case, we can always assume that after a translation of $S$ we have its interior point $q$ at $0$, and then a radial dilation of $S$ centered at $q=0$ converts $S$ into $\bar{B}_{1}(0)$. 
The combination of these two operations is a homeomorphism (a continuous mapping from $\mathbb{R}^{n}$ to $\mathbb{R}^{n}$ with continuous inverse), and we convert the general case to the simple case with $S=\bar{B}_{1}(0)$ as above.
\end{proof}

Now let us consider the optimal control problem with some given fixed endpoint $x_{1} \in \mathbb{R}^{d}$. 
Again let $T > 0$ be the first time that the trajectory $x$ satisfying the initial value problem
\begin{equation*}
\begin{cases}
\xdot(t)= f( x(t),  u(t)) , \quad \forall\, t \ge 0\\
x(0)=x_{0} 
\end{cases}
\end{equation*}
arrives at the endpoint $x_{1}$. 
Then the total reward functional is given by
\begin{equation*}
I[ u ] = \int_{0}^{T} r( x(t),  u(t)) \, d t ,
\end{equation*}
where $r: \mathbb{R}^{d} \times \Ucal \rightarrow \mathbb{R}$ is a given running reward function.

We again consider the auxiliary function $x_{d+1}: [0,T] \to \mathbb{R}$ satisfying the initial value problem
\begin{equation*}
\begin{cases}
\xdot_{d+1}(t)= r( x(t),  u(t)) , \quad \forall\, t \ge 0 , \\
x_{d+1}(0)=0 . 
\end{cases}
\end{equation*}
We denote 
\begin{equation*}
\bar{x} :=
\begin{pmatrix}
x \\ x_{d+1}
\end{pmatrix} \in \mathbb{R}^{d+1}, \quad 
\bar{x}_{0} :=
\begin{pmatrix}
x_{0} \\ 0
\end{pmatrix} \in \mathbb{R}^{d+1},
\end{equation*}
and
\begin{equation*}
\bar{f}(\bar{x}, u) := 
\begin{pmatrix}
f(x,u) \\ r(x,u) 
\end{pmatrix} \in \mathbb{R}^{d+1}, 
\end{equation*}
and the terminal reward functional as $\bar{g}(\bar{x}) : = x_{d+1}$.
Then the optimal control problem \eqref{eq:oc-fix-end} with fixed endpoint $x_{1} \in \mathbb{R}^{d}$ (recall that $x_{0}, x_{1} \in \mathbb{R}^{d}$ are the given initial and end points of $x:[0,T] \to \mathbb{R}^{d}$, and $x_{1}$ is not the first component of $x(t)$; We will not discuss individual components of $x(t)$ in this proof) can be written as
\begin{subequations}
\label{eq:oc-endpoint}
\begin{align}
\max_{u} \quad & I[u] := \bar{g}(\bar{x}(T)) = x_{d+1}(T) , \label{eq:oc-endpoint-obj} \\
\text{s.t.} \quad & 
\begin{cases}
\dot{\bar{x}}(t) = \bar{f}(\bar{x}(t), u(t) ), \quad \forall\, t \in [0,T] , \label{eq:oc-endpoint-ode} \\
\bar{x}(0) = \bar{x}_{0}, \\
x(T) = x_{1} .
\end{cases}
\end{align}
\end{subequations}
where the admissible set of $u$ is $\{u: [0,\infty) \to \Ucal\}$.

To establish the Pontryagin Maximum Principle for the optimal control problem \eqref{eq:oc-endpoint}, we let $u$ denote the optimal control and $x$ the corresponding optimal trajectory.
Our first goal is to construct the co-state $p$ for the Pontryagin Maximum Principle.
To this end, we will use the notations $\xbar$, $\fbar$ and $\gbar$ above.
Specifically, we define the cone of variations in this case by following \eqref{eq:cone-var} as
\begin{equation}
\label{eq:cone-var-endpoint}
C:=C(T)=\left\{ \sum_{k=1}^{K} \lambda_{k}  \bar{\Phi}(T, \tau_{k}) \ybar_{\tau_{k}} \, : \, 
\begin{array}{l}
K \in \mathbb{N}, \text{ and } \lambda_{k} \ge 0, w_{k} \in \Ucal, \\
0< \tau_{1} \le \cdots \le \tau_{K} <T, \ \forall\, k \in [K]
\end{array}
\right\} .
\end{equation}
where
\begin{equation*}
\ybar_{\tau_{k}} = \fbar(\xbar(\tau_{k}), w_{k}) - \fbar(\xbar(\tau_{k}), u(\tau_{k}))  ,
\end{equation*}
and $\bar{\Phi}(t,s)$ satisfies
\begin{equation*}
\begin{cases}
\frac{d}{d t} \bar{\Phi}(t,s) = \partial_{\xbar} \fbar(\xbar(t), u(t)), \quad \text{for any } t \ge s \ge 0\\
\bar{\Phi}(s,s) = I_{d+1} .
\end{cases}
\end{equation*}
Then the following lemma indicates a critical property of $C$.

\begin{lemma}
[Geometry of the cone of variations]
\label{lem:oc-lemma-cone-geometry}
The unit vector $e_{d+1}=(0,\dots,0,1) \in \mathbb{R}^{d+1}$ is not an interior point of $C$ in \eqref{eq:cone-var-endpoint}.
\end{lemma}

\begin{proof}
We use a proof by contradiction. Assume $e_{d+1}$ is an interior point of $C$, then there exist linearly independent vectors $z_{1}, \dots, z_{d+1} \in \mathbb{R}^{d+1}$ and positive numbers $\lambda_{1}, \dots, \lambda_{d+1} > 0$ such that
\begin{equation*}
e_{d+1} = \sum_{k=1}^{d+1} \lambda_{k} z_{k} ,
\end{equation*}
where
\begin{equation*}
z_{k} = \bar{\Phi}(T, \tau_{k}) \ybar_{\tau_{k}}
\end{equation*}
for $0 < \tau_{1} < \dots < \tau_{d+1} < T$ (we can safely assume these time points are distinct since $e_{d+1}$ is an interior point of $C$ and we are able to make $\tau_{1},\dots,\tau_{d+1}$ different). 
% and 
%\begin{equation*}
%\bar{y}_{\tau_{k}} = \fbar(\xbar(\tau_{k}), w_{k}) - \fbar(\xbar(\tau_{k}), u(\tau_{k})) .
%\end{equation*}

Now we let $\eta > 0$ and define 
\begin{equation*}
S := \Big\{ \sum_{k=1}^{d+1} \lambda_{k} z_{k} : 0 \le \lambda_{k} \le \eta, \ \forall\, k \in [d+1] \Big\} .
\end{equation*}
Since $z_{1}, \dots, z_{d+1}$ are linearly independent, we know $S$ has interior points.
Moreover, $S$ is a compact convex polytope with all edges of length proportional to $\eta$ and a vertex at $0 \in \mathbb{R}^{d+1}$.

Since $e_{d+1}$ is an interior point of $C$, we know there exists sufficiently small $\mu > 0$ such that $q := \mu e_{d+1}$ is an interior point of $S$. 
In addition, there exists $\delta := \delta(\eta,\mu) > 0$, such that $|\ybar - q | > \delta$ for all $\ybar \in \partial S$.

For any small $\epsilon > 0$, we define $\phi^{\epsilon}: S \to \mathbb{R}^{d+1}$ by
\begin{equation}
\label{eq:oc-phi-eps}
\phi^{\epsilon}(z) := \frac{\xbar^{\epsilon}(T) - \xbar(T)}{\epsilon} 
\end{equation}
for any $z \in S$, where $\xbar$ solves the initial value problem
\begin{equation*}
\begin{cases}
\dot{\xbar}(t) = \fbar(\xbar(t),u(t)), \quad \forall\, t \in [0,T] , \\
\xbar(0) = \xbar_{0} ,
\end{cases}
\end{equation*}
and $\xbar^{\epsilon}$ solves the initial value problem
\begin{equation*}
\begin{cases}
\dot{\xbar}^{\epsilon}(t) = \fbar(\xbar^{\epsilon}(t),u^{\epsilon}(t)), \quad \forall\, t \in [0,T] , \\
\xbar^{\epsilon}(0) = \xbar_{0} ,
\end{cases}
\end{equation*}
and $\ueps$ is the multiple variations of the control $u$ defined by
\begin{equation*}
\ueps(t) =
\begin{cases}
w_{k}, & \text{if } \tau_{k} - \lambda_{k} \epsilon \le t < \tau_{k}, \ k = 1, \dots, K, \\
u(t) , & \text{otherwise} .
\end{cases}
\end{equation*}
Then by Lemma \ref{lem:oc-perturb}, we have
\begin{equation*}
\max_{0 \le t \le T} \frac{|\xbar^{\epsilon}(t) - \xbar(t) - \epsilon \ybar(t)|}{\epsilon} \to 0
\end{equation*}
as $\epsilon \to 0$ for any finite $T$, where $\ybar$ is the first-order perturbation between $\xbar^{\epsilon}$ and $\xbar$ as defined in Lemma \ref{lem:oc-perturb} but with ${\Phi}(t,\tau_{k})$ and $y_{\tau_{k}}$ there replaced by the corresponding ones determining $z$ in \eqref{eq:oc-phi-eps} here. 
Therefore, 
\begin{equation*}
| \phi^{\epsilon}(z) - \ybar(T) | = \Big| \frac{\xbar^{\epsilon}(T) - \xbar(T)}{\epsilon} - \ybar(T) \Big|
= \Big| \frac{\xbar^{\epsilon}(T) - \xbar(T) - \epsilon \ybar(T)}{\epsilon} \Big| \le |O(\epsilon)| .
\end{equation*}
We can choose $\epsilon$ small enough such that $|O(\epsilon)| < \delta = \delta(\eta,\mu)$, which implies
\begin{equation*}
|\phi^{\epsilon}(z) - \ybar(T)| \le |O(\epsilon)| < \delta = \delta(\eta,\mu) \le | \ybar - q |, \quad \forall\, \ybar \in \partial S .
\end{equation*}

By Lemma \ref{lem:zeros-vf}, we know there exists $z^{*} \in S$, such that $\phi^{\epsilon}(z^{*}) = q = \mu e_{d+1}$.
Let $\ueps$ be the multiple variations of the control $u$ corresponding to this $z^{*}$ and $\xeps$ be the trajectory following $\ueps$. Then $\phi^{\epsilon}(z^{*}) = \frac{1}{\epsilon}(\xbar^{\epsilon}(T) - \xbar(T)) = \mu e_{d+1}$, namely
$\xeps(T) = x(T) = x_{1}$ and $\xeps_{d+1}(T) - x_{d+1}(T) = \mu \epsilon > 0$.
This implies that $\xeps(T) = x_{1}$ at time $T$ and earns a greater reward:
\begin{equation*}
\int_{0}^{T} r(\xeps(t), \ueps(t)) \, dt = x_{d+1}^{\epsilon}(T) > x_{d+1}(T) = \int_{0}^{T} r(x(t),u(t)) \, dt ,
\end{equation*}
which is a contradiction to the optimality of $u$ and $x$. Therefore, $e_{d+1}$ is not an interior point of $C$.
\end{proof}

Lemma \ref{lem:oc-lemma-cone-geometry} implies that $e_{d+1}$ is either on the boundary of the convex cone $C$ or outside of $C$. If $e_{d+1} \in \partial C$, then this is the so-called \emph{abnormal problem}\index{Abnormal problem} where the Pontryagin Maximum Principle\index{Pontryagin Maximum Principle} is useless. If $e_{d+1} \notin C$, then the principle holds as presented below.

\begin{theorem}
[Pontryagin Maximum Principle for fixed endpoint problem]
\label{thm:pmp-endpoint}
Suppose the optimal control problem \eqref{eq:oc-fix-end} is not abnormal, and $u$ is an optimal solution to the problem and $x$ is the trajectory corresponding to $u$. Then there exists a function $p: [0, T] \rightarrow \mathbb{R}^{d}$ satisfying the adjoint equation, and the maximization principle as in Theorem \ref{thm:oc-pmp} holds true with this triple $(x,p,u)$.
\end{theorem}

\begin{proof}
We have $e_{d+1} \notin C$ when \eqref{eq:oc-fix-end} is not abnormal. Notice that $C$ is closed and convex.
By the supporting hyperplane theorem, there exists $v = (v_{1},\dots, v_{d+1}) \in \mathbb{R}^{d+1}$ such that $v_{d+1} > 0$ and $v \cdot z \le 0$ for all $z \in C$. Without loss of generality, we assume $v_{d+1} = 1$ by setting $v$ to $v/v_{d+1}$.

Suppose $\pbar = (p, p_{d+1}): [0,T] \to \mathbb{R}^{d+1}$ solves the terminal value problem
\begin{equation}
\label{eq:pmp-fix-end-pbar-terminal-value}
\begin{cases}
\dot{\pbar} (t) = - \pbar(t) \partial_{\xbar} \fbar(\xbar(t), u(t)), \quad \forall\, t \in [0,T] ,\\
\pbar(T) = v .
\end{cases}
\end{equation}
Therefore, $\dot{p}_{d+1}(t) = 0$ for all $t \in [0,T]$ and $p_{d+1}(T) = v_{d+1} = 1$, which imply that $p_{d+1}(t) = 1$ for all $t \in [0,T]$.
As a result, $p$ satisfies the adjoint equation 
\begin{align*}
\pdot(t) 
& = - \pbar(t) \partial_{x} \bar{f}(\bar{x}(t),u(t)) \\
& = - p(t) \partial_{x} f(x(t),u(t)) - p_{d+1}(t) \partial_{x} r(x(t),u(t)) \\
& = - p(t) \partial_{x} f(x(t),u(t)) - \partial_{x} r(x(t),u(t)) 
\end{align*}
on $[0,T]$ according to \eqref{eq:pmp-fix-end-pbar-terminal-value}.

For any $\tau \in [0, T)$ and $w \in \Ucal$, define
\begin{equation*}
\ybar_{\tau} := \fbar(\xbar(\tau), w) - \fbar(\xbar(\tau), u(\tau)) \ybar(t)
\end{equation*}
and $\ybar(T):= \bar{\Phi}(T,\tau) \ybar_{\tau}$ which is the solution to the initial value problem
\begin{equation*}
\begin{cases}
\dot{\ybar}(t) = \partial_{\xbar} \fbar(\xbar(t), u(t)), \quad \forall\, t \in [\tau, T] , \\
\ybar(\tau) = \ybar_{\tau} .
\end{cases}
\end{equation*}
Furthermore, by the definition of $C$ in \eqref{eq:cone-var-endpoint}, we know $\ybar(T) \in C$.

Now we notice that, by the geometric property of $C$ given at the beginning of this proof and the terminal condition $\pbar(T) = v$, there is
\begin{equation}
\label{eq:pmp-ep-opt}
0 \ge v \cdot \ybar(T) = \pbar(T) \cdot \ybar(T) .
\end{equation}
Moreover, there is
\begin{align*}
\frac{d}{d t} (\pbar(t) \cdot \ybar(t))
& = \dot{\pbar}(t) \cdot \ybar(t) + \pbar(t) \cdot \dot{\ybar}(t) \\
& = - \pbar(t) \partial_{\xbar} \fbar(\xbar(t),u(t)) \cdot \ybar(t) + \pbar(t) \cdot \partial_{\xbar} \fbar(\xbar(t),u(t)) \ybar(t) \\
& = 0 ,
\end{align*}
and therefore $\pbar(t)\cdot \ybar(t)$ is constant in $t$, which implies that
\begin{equation}
\label{eq:pmp-ep-py}
\pbar(T)\cdot \ybar(T) = \pbar(\tau) \cdot \ybar(\tau) .
\end{equation}
Combining \eqref{eq:pmp-ep-opt} and \eqref{eq:pmp-ep-py}, we have $0 \ge \pbar(\tau) \ybar(\tau) = \pbar(\tau) \ybar_{\tau}$, namely,
\begin{equation*}
0 \ge \pbar(\tau) \cdot ( \fbar(\xbar(\tau), w) - \fbar(\xbar(\tau), u(\tau)) .
\end{equation*}
Hence, there is
\begin{align*}
\bar{H}(\xbar(\tau), \pbar(\tau), w)
& = \pbar(\tau) \cdot \fbar(\xbar(\tau), w) \\
& \le \pbar(\tau) \cdot \fbar(\xbar(\tau), u(\tau)) \\
& = \bar{H}(\xbar(\tau), \pbar(\tau), u(\tau)) .
\end{align*}
As $w \in \Ucal$ and $\tau \in [0,T)$ are arbitrary, and $u(\tau) \in \Ucal$, we obtain 
\begin{equation}
\label{eq:pmp-fix-end-bar}
\bar{H}(\xbar(\tau), \pbar(\tau), u(\tau)) = \max_{w \in \Ucal} \bar{H}(\xbar(\tau), \pbar(\tau), w)
\end{equation}
at every $\tau$. Notice that, by the definition of $\bar{H}$, there is
\begin{equation}
\label{eq:pmp-fix-end-Hbar}
\bar{H}(\xbar, \pbar, u) = \pbar \cdot \fbar(\xbar,u) = p \cdot f(x,u) + r(x,u)
\end{equation}
since $p_{d+1} = v_{d+1} = 1$. Substituting \eqref{eq:pmp-fix-end-Hbar} into \eqref{eq:pmp-fix-end-bar}, we obtain the maximum principle for the optimal control problem with fixed endpoint \eqref{eq:oc-endpoint} or equivalently \eqref{eq:oc-fix-end}.
\end{proof}

\subsection{Hamilton--Jacobi--Bellman Equation}

\label{subsec:hjb}

In this section, we consider a different approach to solving optimal control problems. This approach is based on the idea of dynamical programming (DP), which is to embed the given problem into a larger class of problems and solve this entire class all at once.
Then the solution to the entire class can directly imply the solution to the original problem.

Recall that the optimal control problem with fixed time $T$ and free endpoint is given by
\begin{subequations}
\label{eq:oc-hjb-original}
\begin{align}
\max_{u \in \Ucal_{T}} \quad & I[u] := \int_{0}^{T} r(z(s), u(s)) \, ds + g(z(T)), \\
\text{s.t.} \quad &
\begin{cases}
\zdot(s) = f(z(s), u(s)), \quad \forall\, s \in [0,T] ,\\
z(0) = x ,
\end{cases}
\end{align}\end{subequations}
with some given initial value $x \in \mathbb{R}^{d}$. 
We can embed this problem \eqref{eq:oc-hjb-original} into a larger class of problems with arbitrary the initial value $x \in \mathbb{R}^{d}$ and start time $t \in [0,T]$:
\begin{subequations}
\label{eq:oc-hjb-general}
\begin{align}
\max_{u} \quad & I_{x,t}[u] := \int_{t}^{T} r(z(s), u(s)) \, ds + g(z(T)), \\
\text{s.t.} \quad &
\begin{cases}
\zdot(s) = f(z(s), u(s)), \quad \forall\, s \in [t,T] ,\\
z(t) = x ,
\end{cases}
\end{align}
\end{subequations}
where $u$ is only required to satisfy $u(s) \in \Ucal$ for every $s \in [t, T]$.
Then the original optimal control problem \eqref{eq:oc-hjb-original} is a special case of \eqref{eq:oc-hjb-general} with $x$ being a specified point and $t = 0$.

We define the \emph{value function}\index{Value function} $v: \mathbb{R}^{d} \times [0,T] \to \mathbb{R}$ by
\begin{equation}
\label{eq:oc-value-fn}
v(x,t) := \sup_{u} I_{x,t}[u] .
\end{equation}
It is clear that $v(x,T) = g(x)$ for all $x \in \mathbb{R}^{d}$.

Now the question is whether we can find $v(x,t)$ by constructing a problem of $v$ and solving it, and then figure out the optimal control $u$ using the found value function $v$. This is the typical strategy of DP.
% is it is better to be smart at the beginning than to be stupid for a time and then become smart.

\begin{theorem}
[Hamilton--Jacobi--Bellman equation]
\label{thm:hjb}
Assume $v$ is $C^{1}$ in $(x,t)$, then $v$ solves the following nonlinear PDE, called the \emph{Hamilton--Jacobi--Bellman} (HJB) equation\index{Hamilton--Jacobi--Bellman equation}:
\begin{equation}
\label{eq:hjb}
\partial_{t} v(x,t) + \max_{w \in \Ucal} \Big\{ \nabla v(x,t) \cdot f(x,w) + r(x, w) \Big\} = 0
\end{equation}
with terminal value $v(x,T) = g(x)$ for all $x \in \mathbb{R}^{d}$ and $t \in [0,T]$.
\end{theorem}

\begin{proof}
We first prove that \eqref{eq:hjb} holds with supremum and inequality (no greater than 0), and then show that it in fact holds in its original form provided the existence of optimal control $u$.

To this end, we follow our previous notation by defining
\begin{equation}
\label{eq:hjb-UT}
\Ucal_{T} := \{ u : [0,T] \to \Ucal \text{ measurable} \}
\end{equation}

For any $x \in \mathbb{R}^{d}$, $t \in [0,T)$, $\epsilon > 0$ with $t+\epsilon \le T$, and $w \in \Ucal$, suppose we solve the initial value problem on the interval $[t, t+\epsilon]$:
\begin{equation}
\begin{cases}
\zdot(s) = f(z(s), w) , \quad \forall\, s \in [t, t+ \epsilon] , \\
z(t) = x .
\end{cases}
\end{equation}
Then the reward received during $[t, t+\epsilon]$ is $\int_{t}^{t+\epsilon} r(z(s),w) ds$. Let $x(t+s) = z(t+s)$, then the maximal reward that can be obtained during $[t+\epsilon, T]$ is $v(x(t+\epsilon), t+\epsilon)$ by the definition of the value function $v$.

On the other hand, since $v(x,t)$ is the maximal reward at $(x,t)$, its optimality implies
\begin{equation}
\label{eq:hjb-v-opt}
v(x,t) = v(z(t), t) \ge \int_{t}^{t+\epsilon} r(z(s),w)\,ds + v(x(t+\epsilon),t+\epsilon)
\end{equation}
Rearranging \eqref{eq:hjb-v-opt} and dividing both sides by $\epsilon$, we obtain
\begin{equation*}
\frac{v(x(t+\epsilon),t+\epsilon) - v(x,t)}{\epsilon} + \frac{1}{\epsilon} \int_{t}^{t+\epsilon} r(z(s), w) \, ds \le 0 .
\end{equation*}
Letting $\epsilon \to 0$, we have $x(t+s) = z(t+s) \to z(t) = x$ and therefore
\begin{equation*}
\partial_{t} v(x,t) + \nabla v(x,t) \cdot \zdot(t) + r(x,w) \le 0 .
\end{equation*}
Since $\zdot(t) = f(z(t),w)$, we have
\begin{equation}
\label{eq:hjb-ineq}
\partial_{t} v(x,t) + \nabla v(x,t) \cdot f(x,w) + r(x,w) \le 0 .
\end{equation}
Since \eqref{eq:hjb-ineq} holds for any $w \in \Ucal$, we obtain
\begin{equation}
\label{eq:hjb-ineq-gen}
\partial_{t} v(x,t) + \sup_{w \in \Ucal} \Big\{ \nabla v(x,t) \cdot f(x,w) + r(x,w) \Big\} \le 0 .
\end{equation}

Next, we show that the supremum is indeed maximum and the equality holds in \eqref{eq:hjb-ineq-gen}. 
To see these, suppose we continuously use the optimal control $u$ since time $t$, then we obtain 
\begin{equation*}
v(x,t) = \int_{t}^{t+\epsilon} r(x(s),u(s)) \, ds + v(x(t+\epsilon), t+\epsilon)
\end{equation*}
Rearranging the equality above and dividing both sides by $\epsilon$, we obtain
\begin{equation*}
\frac{v(x(t+\epsilon),t+\epsilon) - v(x,t)}{\epsilon} + \frac{1}{\epsilon} \int_{t}^{t+\epsilon} r(x(s), u(s)) \, ds = 0 .
\end{equation*}
Taking $\epsilon \to 0$, we get
\begin{equation}
\label{eq:hjb-eq}
\partial_{t} v(x,t) + \nabla v(x,t) \cdot f(x(t),u(t)) + r(x(t),u(t)) = 0 .
\end{equation}
Since $u(t) \in \Ucal$ for every $t$, we know the supremum in \eqref{eq:hjb-ineq-gen} is indeed maximum and the equality holds true, yielding the HJB equation \eqref{eq:hjb}.
\end{proof}

Now let us see how to use the HJB equation to find the optimal control.
Suppose we can solve the HJB and find its solution $v$, then we define $\pi: \mathbb{R}^{d} \times [0,T] \to \Ucal$ by
\begin{equation}
\pi(x,t) := \argmax_{w \in \Ucal} \{ \nabla v(x,t) \cdot f(x,w) + r(x,w) \} .
\end{equation}
In this case, we have
\begin{equation}
\label{eq:hjb-at-max}
\partial_{t} v(x,t) + \nabla v(x,t) \cdot f(x,\pi(x,t)) + r(x, \pi(x,t)) = 0 .
\end{equation}
Then for any $x \in \mathbb{R}^{d}$ and $t \in [0, T)$, we can solve the following initial value problem (assuming $\pi(\cdot, t)$ is sufficiently regular) to obtain $x^{*}$:
\begin{equation}
\label{eq:hjb-x-star-ode}
\begin{cases}
\xdot^{*}(s) = f(\xs(s), \pi(\xs(s), s) ) , \quad \forall\, s \in [t,T] , \\
\xs(t) = x .
\end{cases}
\end{equation}
Then we can claim that the optimal control $u^{*}$ to the problem \eqref{eq:oc-hjb-general} is
\begin{equation*}
u^{*}(s) := \pi(\xs(s), s)
\end{equation*}
at every $s \in [t,T]$. To see this $u^{*}$ is indeed optimal, we notice that
\begin{align*}
I_{x,t}[u^{*}]
& = \int_{t}^{T} r(\xs(s), u^{*}(s)) \,ds + g(\xs(T)) \\
& = - \int_{t}^{T} \Big( \partial_{s} v(\xs(s),s) + \partial_{x} v(\xs(s),s) \cdot f(\xs(s),u^{*}(s)) \Big) ds + g(\xs(T)) \\
& = - \int_{t}^{T} \Big( \partial_{s} v(\xs(s),s) + \partial_{x} v(\xs(s),s) \cdot \dot{x}^{*}(s) \Big) ds + g(\xs(T)) \\
& = - \int_{t}^{T} \frac{d}{d s}\Big( v(\xs(s),s) \Big) ds + g(\xs(T)) \\
& = v(\xs(t),t) - v(\xs(T),T) + g(\xs(T)) \\
& = v(x,t) \\
& = \sup_{u} I_{x,t}[u] ,
\end{align*}
where the second equality is due to \eqref{eq:hjb-at-max}, the third due to the ODE in \eqref{eq:hjb-x-star-ode}, the sixth due to the terminal condition $v(x,T) = g(x)$ for all $x$. This implies that $u^{*}$ is an optimal control to \eqref{eq:oc-hjb-general}.

At this point, we see that solving the HJB equation \eqref{eq:hjb} can be an alternative approach to solving the optimal control problem, and the steps have been explained as above.

However, we remark that the HJB equation \eqref{eq:hjb} is a nonlinear first-order PDE of $v$, and it is further complicated due to the presence of maximum which requires solving an optimization problem on $\Ucal$ at every time $t$. In some special cases, the maximizer as a closed form and the maximum in \eqref{eq:hjb} can be eliminated. 
However, even without the maximum, this PDE can still be infeasible or at least difficult to solve due to its high dimension (the spatial dimension is $d$ according to $x$) in practice. Nevertheless, the strong connections between the HJB equation (from the DP view) and the optimal control provide significant insights and analysis tools to study and solve control problems.

\section{Optimal Probability Density Control Theory}
\label{sec:odc-theory}

Deep learning problems generally involve massive amount of data points or agents. 
For example, in a large multi-agent control system, the state of each agent can be represented as a vector with values describing its location, orientation, velocity, and acceleration. 
In many cases, these agents can be thought of as samples following some specific probability distributions.
Typically, the learning problems are formulated as evolving these samples in time, which is equivalent to the evolution of the probability distribution they follow, such that certain objective function can be minimized or maximized in a given time horizon $[0,T]$. 
Therefore, such problems can be formulated as optimal control of probability density.
In this section, we extend the theoretical results of classical optimal control in Section \ref{sec:oc-theory} to the optimal probability control scenario.

% Given the large scale of these systems, a probabilistic description of the state is often the most effective approach \cite{jimenez2020optimal}. Consequently, a standard method for solving these multi-agent control problems is through mean-field modeling, which serves as a continuum approximation of the discrete multi-agent system. Here, the probability density of the agent states represents the entire system. Notably, the state of an agent may reside in a high-dimensional space, as it encompasses combined variables such as location, orientation, velocity, and acceleration. In this regard, optimal probability control and multi-agent control are unified into the same problem framework, and the goal is to find the control vector field that maximizes a given total reward function describing the probability density, or equivalently the collective behaviors of the agents, with consideration of inter-agent collaboration and competitions.

Let $\Omega$ be an open subset of $\Rbb^{d}$, where $d$ is the dimension of state vector (such as the concatenation of location, orientation, and acceleration of an agent mentioned above).
In the remainder of this section, we generally refer a sample to be an agent.
Suppose there are $N$ agents with states $\{x_{i} \in \Omega: i = 1,\dots, N\}$ which follow some initial distribution $p$ on $\Omega$, i.e., $x_{i} \sim p$. 
Under a time-evolving control vector field $u: \Omega \times [0,T] \to \Rbb^{d}$ over a fixed time horizon $[0,T]$, the states of the agents change according to 
\begin{equation}
    \label{eq:agent-ode}
    \dot{x}_{i}(t) = u(x_{i}(t), t), \quad i=1,\dots, N .
\end{equation}
Each agent accumulates some running rewards during $[0,T]$ and receives a terminal reward at time $T$.
Unlike classical optimal control, these rewards \emph{depend not only this specific agent but also the collective behaviors of all agents.} 
We provide a few examples of running and terminal reward functions below.

We use $\rho(\cdot,t)$ to denote the time marginal of the states $x_{i}(t)$ for every $t$, namely, $x_{i}(t) \sim \rho(\cdot, t)$. Then \eqref{eq:agent-ode} corresponds to the continuity equation
\begin{equation}
    \label{eq:ct-eq}
    \partial_{t} \rho(x,t) + \nabla \cdot \bigr(\rho(x, t) u(x,t) \bigr) = 0
\end{equation}
in $\Omega \times [0,T]$. 
For notation simplicity, we write $\rho(\cdot,t)$ and $u(\cdot,t)$ as $\rho_{t}$ and $u_{t}$ respectively, and often omit $(x)$ in the integrals hereafter unless it is needed for clarification.

%
%Let $\Omega$ be an open subset of $\Rbb^d$, where $d$ is the state space dimension. Most notably, we focus on the case of high dimensions, e.g., $d\ge 5$. 
%%
%% We can also extend our theory and computation methodology to more general cases of $\Omega$ where the density functions and controls have zero Dirichlet and Neumann boundary conditions, which often hold true in practice since they usually vanish outside a bounded region and our derivations are still valid.
%%
%We also remark that our goal is to control probability density functions defined on $\Omega$, and these functions form an infinite-dimensional function space. This is very different from classical control problems which are defined on finite-dimensional (such as Euclidean) spaces. 

We define the space of probability density functions on $\Omega$:
\begin{equation} 
\label{eq:P}
P = \Bigl\{ p: \Omega \to [0,\infty) : \int_{\Omega} p(x) \, dx = 1 \Bigr\}.
\end{equation}
% We also assume necessary smoothness of $p$ for our derivations below, e.g., $p$ is differentiable in $\Omega$.
%
%For notation simplicity, we omit specifying $\Omega$, $(x)$, and $dx$ hereafter unless there is a chance of confusion.
%
Furthermore, for the given terminal time $T>0$, we define the space of time-evolving probability density functions as:
\begin{equation*} 
%\label{eq:PT}
P_{T} = \bigl\{ \rho : \Omega \times [0,T] \to \Rbb : 
     \rho(\cdot,t) \in P, \ \forall \, t \in [0,T] \bigr\},
\end{equation*}
% where $L^1(0,T)$ is the space of Lebesgue integrable functions defined on $[0,T]$.
%
% (Note: The $L^1$ requirement in time $t$ is a sufficient condition that may be relaxed, as discussed in Section [X]).
%
We assume that the admissible set of control vector field at any fixed time is 
\begin{equation}
\label{eq:U}
U = \bigl\{ w \in C(\bar{\Omega}; \Rbb^d) : \mathrm{Lip}(w) \le M_{U},\, (w \cdot n)|_{\partial \Omega} = 0 \bigr\}, 
\end{equation}
for some $M_{U} > 0$, namely, $w$ is $M_{U}$-Lipschitz continuous in $\bar{\Omega}$ and satisfies the boundary condition $w(x) \cdot n(x) = 0$, where $n(x) \in \Rbb^{d}$ is the outer normal of $\partial \Omega$ at $x$ for all $x \in \partial \Omega$.
%
% where $W_{0}^{1,\infty}(\Omega;\Rbb^{d})$ is the $(1,\infty)$-Sobolev space of vector-valued functions with zero boundary condition on $\partial \Omega$.
%
% Note that \eqref{eq:U} holds if and only if $w$ is essentially bounded by $M_{U}$ (bounded by $M_{U}$ except for a measure zero subset of $\Omega$) and $M_{U}$-Lipschitz continuous on $\Omega$ with $w=0$ on $\partial \Omega$. 
%
The boundary condition of $w$ conserves the total probability density as 1.
Then we define the admissible set of time-evolving control vector fields as
\begin{equation}
\label{eq:UT}
    U_T= \bigl\{ u: \bar{\Omega} \times [0,T] \to \Rbb^{d} : u(\cdot, t) \in U,\ \forall\, t \in [0,T] \bigr\}.
\end{equation}
Note that \eqref{eq:UT} does not require $u(x,\cdot)$ to be a continuous function of $t$, allowing $u$ to change direction instantaneously and yielding broader range of applications of the results obtained below.

Now we have the optimal probability control problem in its general form as
\begin{subequations}
\label{eq:control-problem}
\begin{align}
    \max_{u\in U_T} \quad & I[u] := \int_0^T R(\rho_t,u_t) \, dt + G(\rho_T) , \label{eq:odc-obj} \\
    \mathrm{s.t.} \quad & \partial_t\rho_t + \nabla \cdot (\rho_t u_t) = 0,\quad \forall\, (t,x) \in [0,T] \times \Omega, \label{eq:odc-ct-eq} \phantom{\frac{}{}} \\
    & \rho_0=p,\quad \forall \, x \in \Omega, \label{eq:odc-init}
\end{align}
\end{subequations}
where $I$ is the total reward functional, $U_{T}$ is the admissible set of control vector fields, and $p$ is a given initial probability distribution, i.e., $x_{i}(0) \sim p$. 
In \eqref{eq:control-problem}, $R: P \times U \to \mathbb{R}$ and $G: P \to \mathbb{R}$ are the running and terminal functionals, respectively.
We assume $I^{*}:= \max_{u \in U_{T}} I[u]$ to be finite.

We have a few assumptions on the probability distributions and controls as follows, which ensure the derivations later can be carried out rigorously.
%
% In addition to the definitions above, we assume $R: P \times R \to \Rbb$ and $G: P \to \Rbb$ are Fr\'{e}chet differentiable with respect to their arguments.  
%
% We also have a differentiability assumption on $u(x,\cdot)\rho(x,\cdot)$, which is a function of $t$ at every fixed $x \in \Omega$, as follows.
%
%\vspace{6pt}

\begin{assumption}
\label{assump:rho}
    We make the following assumptions:
    \begin{enumerate}
        \item[(i)] For all $x \in \Omega$, $\rho(x,\cdot)$ is differentiable in $t$ on $[0,T]$.
        \item[(ii)] For all $t \in [0,T]$, $\rho(\cdot,t) u(\cdot,t)$ is differentiable in $\Omega$. 
        \item[(iii)] For all $x \in \Omega$, $\rho(x,\cdot) u(x,\cdot)$ has Lebesgue points in $(0,T)$ everywhere.
        \item[(iv)] For all $w \in U$, $R(\cdot, w)$ is Fr\'{e}chet differentiable in $P$ in the $L^{2}$ sense.
        \item[(v)] $G(\cdot)$ is Fr\'{e}chet differentiable in $P$ in the $L^{2}$ sense.
    \end{enumerate}
\end{assumption}
%
%\vspace{6pt}

Assumption \ref{assump:rho} (i) and (ii) can be relaxed to weakly differentiable as long as \eqref{eq:ct-eq} is valid.
Assumption \ref{assump:rho} (iii) ensures $\frac{d}{dt} \int_{0}^{t}\rho(x,s)u(x,s)\,ds = \rho(x,t)u(x,t)$ for all $t \in (0,T)$. This is used in a step in the proof of Lemma \ref{lem:uniform-approx} below (we will specify it in the proof). 
Assumption \ref{assump:rho} (iv) and (v) allow functional derivatives of $R$ and $G$ at any $p \in P$, which are needed in the establishment of control Hamiltonian dynamics later.
Notice that running and terminal reward functionals in real-world applications typically satisfy Assumption \ref{assump:rho} (iv) and (v).
%
%For example, the running reward in \eqref{eq:r} has functional derivative
%\begin{equation*}
%    \frac{\delta}{\delta \rho_{t}} R(\rho_{t},u_{t})(x) = -\frac{1}{2} |u_{t}(x)|^{2} - \int_{\Omega} \frac{\gamma  \rho_{t}(y)}{|y-x|^{2}+\epsilon}\,dy
%\end{equation*}
%and the terminal reward in \eqref{eq:g} has functional derivative
%\begin{equation*}
%    \frac{\delta}{\delta \rho_{T}} G(\rho_{T})(x) = - \frac{1}{2} |x - x^{*}|^{2} .
%\end{equation*}
%%
%We will show experimental results on \eqref{eq:control-problem} with other running and terminal reward functionals in Section \ref{sec:results}.

\subsection{Maximum Principle for Probability Control}
\label{subsec:mp}

Now we establish a maximum principle (MP) for the optimal probability control problem \eqref{eq:control-problem} with any initial value $p \in P$ in this subsection. As an analogue to the Pontryagin MP for classic optimal control in Section \ref{sec:oc-theory}, the MP to be established here is a time pointwise necessary condition of the optimal solution $u$ to \eqref{eq:control-problem} defined on the infinite-dimensional spaces $P_{T}$ and $U_{T}$. 
%
% As mentioned before, this is equivalent to finding the optimal control vector field for a massive amount of indistinguishable particles in the mean-field sense. 
%
To develop this MP, we first introduce the definition of adjoint partial differential equation (PDE) and control Hamiltonian functional.
%
%\vspace{6pt}
%
\begin{definition}
    [Adjoint PDE]
    \label{def:adj}
    Let $u \in U_T$ be a control vector field and $\rho \in P_T$ the corresponding evolutionary probability function solving \eqref{eq:odc-ct-eq} with initial \eqref{eq:odc-init}. Define an evolution PDE by
    \begin{equation}
        \label{eq:adj}
        \partial_{t} \phi_t + u_t \cdot \nabla \phi_t = - \frac{\delta}{\delta \rho_{t}} R(\rho_t,u_t) 
    \end{equation}
    in $\Omega \times [0,T]$.
    We call \eqref{eq:adj} the \emph{adjoint PDE} associated to $(\rho,u)$. We call $\phi$ the \emph{adjoint function} of $(\rho,u)$ if $\phi: \Omega \times [0,T] \to \Rbb$ solves \eqref{eq:adj} with terminal condition $\phi_{T} = \frac{\delta}{\delta \rho_{T}} G(\rho_T)$.
\end{definition}
%
%\vspace{6pt}

Throughout this work, we assume $\frac{\delta}{\delta \rho} R(\cdot,w)$, $\frac{\delta}{\delta \rho} G(\cdot)$, and $u$ in the control problem \eqref{eq:control-problem} are sufficiently smooth for any $w \in U$ such that the following assumption holds. 
%
%\vspace{6pt}

\begin{assumption}
\label{assump:phi}
    An adjoint function $\phi: \Omega \times [0,T] \to \Rbb$ for \eqref{eq:control-problem} with any initial $p \in P$ exists.
\end{assumption}
%
%\vspace{6pt}

% We see that Assumption \ref{assump:phi} holds for typical application problems, such as those with the running and terminal reward functionals defined in \eqref{eq:r} and \eqref{eq:g}.

Note that the adjoint function $\phi$ has spatial variable $x$ and time variable $t$, and the adjoint equation is an evolution PDE. By contrast, in classical optimal control, the adjoint function is a function of $t$ only, and the adjoint equation is an ODE.

% With the definition of adjoint equation, we can define the Hamiltonian functional as follows.
%
%\vspace{6pt}

\begin{definition}
    [Hamiltonian functional]
    \label{def:Hamiltonian}
    Let $p \in P$, $w \in U$, and $f: \Omega \to \Rbb$ be a differentiable function, we define the \emph{Hamiltonian functional} by
    \begin{equation}
        \label{eq:Hamiltonian}
        H(p, f, w) = \langle p, \ w \cdot \nabla f \rangle + R(p, w).
    \end{equation}
    where $R$ is the running reward functional in \eqref{eq:control-problem}.
\end{definition}
%
%\vspace{6pt}

% Throughout the remainder of this paper, we write $\frac{\delta}{\delta \rho}H(\rho_t, \phi_t, u_t)$ as the first variation of $H$ with respect to its first argument $\rho_t$ in the $L^2$ sense at time $t$. 
%
Now we can verify that $(\rho,\phi)$ induced by any $u \in U_{T}$ satisfies the symplectic control Hamiltonian dynamics, as summarized in the following proposition.
%
%\vspace{6pt}
%
\begin{proposition}
    [Control Hamiltonian dynamics]
    \label{prop:control-Hamiltonian}
    Let $u \in U_T$ be a control vector field, $\rho \in P_T$ the corresponding evolutionary probability, and $\phi$ the adjoint function of $(\rho,u)$. Then $(\rho, \phi)$ satisfies the control Hamiltonian dynamics:
    \begin{equation}
    \label{eq:hamiltonian-dynamics}
        \begin{cases}
            \partial_t \rho_t = \displaystyle\frac{\delta}{\delta \phi_{t}}H(\rho_t, \phi_t, u_t) = - \nabla \cdot (\rho_t u_t) , \\
            \partial_t \phi_t = -\displaystyle\frac{\delta}{\delta \rho_{t}}H(\rho_t, \phi_t, u_t) = - u_t \cdot \nabla \phi_t - \frac{\delta}{\delta \rho_{t}} R(\rho_t,u_t) 
        \end{cases}
    \end{equation}
    for all $t$.
\end{proposition}
%
%\vspace{6pt}

The dynamics \eqref{eq:hamiltonian-dynamics} are easy to verify using \eqref{eq:ct-eq}, \eqref{eq:adj}, and \eqref{eq:Hamiltonian}, and hence we omit the proof of Proposition \ref{prop:control-Hamiltonian}.

In what follows, we write $u^*$ as an optimal control vector field of \eqref{eq:control-problem} and $\rho^*$ the corresponding probability. 
Then we define a needle-like variant $u^{\epsilon}$ of $u^*$ as follows: For any time $\tau \in (0,T)$ and $\epsilon \in (0, \tau)$, and $w \in U$, we define
\begin{equation}
\label{eq:ue}
\ue_t:= \begin{cases}
    w, & \tau - \epsilon < t \le \tau,\\
    u^*_t, & \text{otherwise}.
\end{cases}    
\end{equation}
%
% Note that $w : \Omega \to \Rbb^{d}$ is a fixed vector field, i.e., $w(x)$ is time-independent on $(\tau-\epsilon,\tau)$ for each $x \in \Omega$. Since $ w \in U$, we have $u^{\epsilon}\in U_T$. 
%
We denote by $\rhoteps \in P_T$ the evolutionary probability determined by \eqref{eq:ct-eq} with control vector field set to $u^{\epsilon} \in U_T$ and initial $\rho_{0}^{\epsilon} = p$.
Next, we introduce the perturbation function that describes the first-order approximation of $\rhoteps$ to $\rho_{t}$.

%
%\vspace{6pt}
%
\begin{definition}[Perturbation function]
\label{def:sigma}
    For any $\tau \in (0,T)$, we define the function $\sigma :[0,T]\times \Omega \to \Rbb$ as follows: 
    If $t \in [0,\tau)$, then
    \begin{equation}
        \label{eq:sigma-pre-tau}
        \sigma_t=0 \,;
    \end{equation}
    If $t \in [\tau,T]$, then $\sigma$ is defined as the solution to the initial value problem:
    \begin{equation}
        \label{eq:sigma-ivp}
        \begin{cases}
        \partial_t \sigma_t = - \nabla \cdot \bigl(\utstar \sigmat \bigr) , & \quad \forall\, t > \tau, \\
        \sigmatau = -\nabla \cdot \bigl( \rhotaustar(w- \utaustar) \bigr) . & 
        \end{cases}
    \end{equation}
\end{definition}
%
%\vspace{6pt}

Note that $\sigmat(x)$ is not necessarily differentiable at $t= \tau$ for a fixed $x$, and $\sigmat$ depends on $\tau$ and $w$ but \emph{not} $\epsilon$. 
%
% In this next lemma, we show that $\sigmat$ is the first-order approximation of $\rhoteps$ with respect to $\rhotstar$ in the $L^2$ sense for all $t \in [0,T]$.
%
We show that the first-order approximation of $\rhoteps$ to $\rho_{t}$ is $\sigmat$ in the following lemma.
\begin{lemma}
    \label{lem:uniform-approx}
    Suppose $\tau \in (0,T)$, $\epsilon \in (0,\tau)$, $w \in U$, and $\uteps$ is defined in \eqref{eq:ue}. Let $\sigmat$ be the perturbation function in Definition \ref{def:sigma}. Then for every $t \in [0,T]$, there is
   \begin{equation}
    \label{eq:o-eps}
    \lim_{\epsilon \to 0} \frac{1}{\epsilon} \| \rhoteps - \rhotstar - \epsilon \sigmat \|_2 = 0.
    \end{equation}
\end{lemma}
%
%\vspace{6pt}

\begin{proof}
We prove \eqref{eq:o-eps} in three cases: $t \in [0,\tau)$, $t = \tau$, and $t \in (\tau, T]$. 

For any $t \in [0,\tau)$, let $\epsilon \in (0, \tau -t )$, then there is $t < \tau - \epsilon$. By \eqref{eq:sigma-pre-tau}, we know $\sigmat= 0$. By \eqref{eq:ue}, we have  $\rhoseps = \rhosstar$ for all $s < \tau-\epsilon$. Therefore $\rhoteps = \rhotstar$.

For $t = \tau$, we only need to show that $\sigmatau$ defined by the initial condition in \eqref{eq:sigma-ivp} is equal to $\lim_{\epsilon \to 0} \frac{\rhotaueps - \rhotaustar}{\epsilon}$ almost everywhere in $\Omega$. To this end, let $\psi \in H_0^1(\Omega)$ be arbitrary, then there is
\begin{align}
    \int_{\Omega} \Big( \lim_{\epsilon \to 0} \frac{\rhotaueps - \rhotaustar}{\epsilon} \Big) \psi \,dx 
    = & \ \lim_{\epsilon \to 0} \frac{1}{\epsilon} \int_{\Omega} (\rhotaueps - \rhotaustar) \psi \,dx \nonumber \\
    = & \ \lim_{\epsilon \to 0} \frac{1}{\epsilon} \int_{\Omega} \Big( \int_{\tau-\epsilon}^{\tau} (\partial_{s} \rhoseps - \partial_{s} \rhosstar) \, ds \Big) \psi \, dx \nonumber \\
    = & \ \lim_{\epsilon \to 0} \frac{1}{\epsilon} \int_{\tau-\epsilon}^{\tau} \int_{\Omega} (w\rhoseps - \ustar_{s}\rhosstar)\cdot \nabla \psi  \,dx ds \label{eq:pf-t-tau} \\
    = & \ \int_{\Omega} \Big( \lim_{\epsilon \to 0} \frac{1}{\epsilon} \int_{\tau-\epsilon}^{\tau} (w - \ustar_s)\rhosstar ds \Big) \cdot \nabla \psi\,dx \nonumber \\
    = & \ \int_{\Omega}  (w - \ustar_\tau)\rhotaustar  \cdot \nabla \psi\,dx \nonumber \\
    = & \ - \int_{\Omega} \nabla \cdot \bigl((w - \ustar_\tau)\rhotaustar \bigr) \psi \,dx \nonumber \\
    = & \ \int_{\Omega} \sigmatau \psi \,dx \nonumber 
\end{align}
where the first equality of \eqref{eq:pf-t-tau} is due to the Lebesgue dominated convergence theorem; the second due to $\rhotaueeps = \rhotauestar$ and integration in time; the third by the continuity equations of $\rhoeps$ and $\rhostar$, integration by parts on $\Omega$, and $w, u_{\tau}^{*} \in U$ (so that $(w\cdot n)|_{\partial \Omega} = 0$ and $(u_{\tau}^{*}\cdot n)|_{\partial \Omega} = 0$); the fourth by the Lebesgue dominated convergence theorem; the fifth by Assumption \ref{assump:rho} (iii); the sixth is due to integration by parts and $w, u_{\tau}^{*} \in U$; and the last equality by the initial condition in \eqref{eq:sigma-ivp}.

For $t> \tau$, define $\delta(t):= \frac{1}{2}\| \rhoteps - \rhotstar - \epsilon \sigmat\|_2^2$, then there is
\begin{align}
\dot{\delta}(t)
&= \int_{\Omega} (\rho^{\epsilon}_t-\rho^*_t-\epsilon \sigmat) \ \partial_t(\rho^{\epsilon}_t-\rho^*_t-\epsilon \sigmat) \,dx \nonumber \\
&=  - \int_{\Omega} (\rho^{\epsilon}_t-\rho^*_t-\epsilon \sigmat)\ \nabla\cdot \bigl(u^*_t(\rho^{\epsilon}_t-\rhotstar-\epsilon \sigmat) \bigr)\, dx \nonumber \\
 &= \int_{\Omega}u^*_t\cdot \nabla \Big( \frac{1}{2}|\rho^{\epsilon}_t-\rho^*_t-\epsilon \sigmat|^2 \Big) \, dx \label{eq:delta-pf}\\
 & \leq  d M_{U} \delta(t), \nonumber
\end{align}
where the second equality is due to the continuity equations of $\rhoteps$, $\rhotstar$ and $\sigmat$ on $(\tau,T]$; the third is due to integration by parts and $(u_{t}^{*} \cdot n)|_{\partial \Omega} = 0$; and the last equality is due to $\mathrm{Lip}(u_{t}) \le M_{U}$ yielding $\| \nabla \cdot \utstar\|_{\infty} \le d M_U$.
By Gr\"{o}nwall inequality, we have $\delta(t) \le \delta(\tau) e^{d M_{U} (T-\tau)} \le \delta(\tau) e^{d M_{U} T}$ for all $t \in (\tau,T]$. 
As we have shown that \eqref{eq:o-eps} holds true at $t = \tau$, namely, $\delta(\tau) = o(\epsilon^2)$, we obtain $\delta(t) = o(\epsilon^{2})$, which implies \eqref{eq:o-eps} for all $t \in (\tau, T]$.
\end{proof}
%
%\vspace{6pt}

In the next lemma, we derive the first variation of $I$, which is critical in the proof of the maximum principle that leverages the optimality of $u^{*}$.
%
%\vspace{6pt}

\begin{lemma}
\label{lem:cost-derivative}
    Let $u^{\epsilon}$ be defined in \eqref{eq:ue} and $I$ defined in \eqref{eq:control-problem} with control vector field $u^{\epsilon}$, then
    \begin{align*}
    \frac{d}{d \epsilon}I[u^{\epsilon}] \Big|_{\epsilon=0}
    & = R(\rho^*_s, w)-R(\rho^*_s,u^*_s) + \Bigl\langle \frac{\delta}{\delta \rho^{*}_{T}} G(\rhostar_{T}), \sigma_{T} \Bigr\rangle +\int_{0}^{T} \Bigl\langle \frac{\delta}{\delta \rho_{t}^{*}} R(\rho^*_t,u^*_t), \sigmat \Bigr\rangle \, dt
    \end{align*} 
    where $\sigmat$ is given in Definition \ref{def:sigma}.
\end{lemma}
%
%\vspace{6pt}

\begin{proof}
    First of all, we have 
    \begin{align*}
    I[u^{\epsilon}]
    & = \int_0^{\tau-\epsilon} R(\rho^{\epsilon}_t,u^*_t) \,dt + \int_{\tau-\epsilon}^{\tau} R(\rho^{\epsilon}_t,w) \, dt + \int_{\tau}^{T} R(\rho^{\epsilon}_t,u^*_t) \,dt + G(\rho^{\epsilon}_T). \label{eq:I-decompose}
    \end{align*} 
    Differentiating with respect to $\epsilon$, using Lemma \ref{lem:uniform-approx}, and setting $\epsilon = 0$, we find that
\begin{align*}
        \frac{d}{d \epsilon}I[u^{\epsilon}] \Big|_{\epsilon=0} 
        & = \int_{0}^{\tau} \Bigl\langle \frac{\delta}{\delta \rho^{*}_{t}} R(\rho^*_t,u^*_t), \sigmat \Bigr\rangle \, dt -R(\rhotaustar,\utaustar) + R(\rhotaustar,w) \\
        & \qquad + \int_{\tau}^{T} \Bigl\langle \frac{\delta}{\delta \rho^{*}_{t}} R(\rho^*_t,u^*_t), \sigmat \Bigr\rangle \, dt + \Bigl\langle \frac{\delta}{\delta \rho^{*}_{T}} G(\rhostar_{T}), \sigma_{T} \Bigr\rangle \\
        & = R(\rhotaustar,w) - R(\rhotaustar,\utaustar) + \int_{0}^{T} \Bigl\langle \frac{\delta}{\delta \rho^{*}_{t}} R(\rho^*_t,u^*_t), \sigmat \Bigr\rangle \, dt + \Bigl\langle \frac{\delta}{\delta \rho^{*}_{T}} G(\rhostar_{T}), \sigma_{T} \Bigr\rangle ,
\end{align*}
    which completes the proof. 
\end{proof}
%
%\vspace{6pt}

Now we are ready to establish the maximum principle which provides a time pointwise necessary condition of any optimal control vector field $u^{*}$. 
%
%\vspace{6pt}

\begin{theorem}[Maximum Principle]
    \label{thm:maximum-principle}
    Let $u^*$ be an optimal solution to \eqref{eq:control-problem} with corresponding density $\rho^*$ and $\phi^{*}$ the adjoint function of $(\rho^{*}, u^{*})$. Then $(\rho^{*}, \phi^{*}, u^{*})$ satisfies the control Hamiltonian dynamics \eqref{eq:hamiltonian-dynamics} with $\rho_{0}^{*}=p$ and $\phi^{*}_{T} = \frac{\delta}{\delta \rho_{T}} G(\rho_{T})$, and the following holds for all $\tau \in [0,T]$:
    \begin{equation}
    \label{eq:pmp-odc}
        H(\rhotaustar,\phi_\tau^*,u^*_\tau)\ = \ \max_{w \in U} \, H(\rho^*_{\tau},\phi_{\tau}^*,w) .
    \end{equation}
    % where $\phi^*: [0,T] \times \Omega$ is the adjoint function associated with $(\rho^*,u^*)$ satisfying \eqref{eq:adj}. 
    % In addition, if $\ustar \in \intr(U_T)$, then $H(\rhotstar, \phi_t^*, \utstar)$ is constant in time $t$.
\end{theorem}
%
%\vspace{6pt}

\begin{proof}
    Let $\tau \in [0,T]$ and $w \in U$ and define $u^{\epsilon}$ as in \eqref{eq:ue} with $\tau$ and $w$. Noting that $u^*$ is optimal for \eqref{eq:control-problem} and applying Lemma \ref{lem:cost-derivative}, we find
    \begin{align}
    0 \ge \frac{d}{d \epsilon}I[u^{\epsilon}] \big|_{\epsilon=0} =R(\rho^*_{\tau}, w)-R(\rho^*_{\tau},u^*_{\tau})+  \Bigl\langle \frac{\delta}{\delta \rho^{*}_{T}}  G(\rho^*_T), \sigma_T \Bigr\rangle +\int_0^T\Bigl\langle \frac{\delta}{\delta \rho^{*}_{t}} R(\rho^*_t,u^*_t), \sigma_t \Bigr\rangle \, dt . \label{eq:cost-derivative}
    \end{align} 
    On the other hand, we have
    \begin{align}
    \label{eq:phi-sigma} 
    \langle \phi^*_T,\sigma_T \rangle 
    = \langle \phi^*_{\tau},\sigma_{\tau} \rangle + \int_{\tau}^T \Bigl(\langle \partial_t \phi^*_t,\sigmat\rangle + \langle\phi^*_t,\partial_t \sigmat \rangle \Bigr) \, dt .
    \end{align}
    Substituting the adjoint PDE \eqref{eq:adj} for $(\rhostar,\ustar)$ and the conditions \eqref{eq:sigma-ivp} for $\sigma$ into \eqref{eq:phi-sigma}, and integrating by parts, we obtain
    \begin{align}
        \langle \phi^*_T,\sigma_T \rangle 
    &=\langle \phi^*_{\tau},\sigma_{\tau} \rangle -\int_{\tau}^T \Bigl\langle \frac{\delta}{\delta \rho^{*}_{t}} R(\rho_t^*,u_t^*),\sigmat \Bigr\rangle \, dt \label{eq:phi-sigma-new} \\
    &=\langle \phi^*_{\tau},\sigma_{\tau} \rangle -\int_{0}^T \Bigl\langle \frac{\delta}{\delta \rho^{*}_{t}} R(\rho_t^*,u_t^*),\sigmat \Bigr\rangle \, dt, \nonumber 
    \end{align}
    where the second equality is due to $\sigmat = 0$ for all $t \in [0,\tau)$. 
    
    We combine \eqref{eq:cost-derivative} and \eqref{eq:phi-sigma-new} as well as the terminal condition $\phi_T^* = \frac{\delta}{\delta \rho^{*}_{T}}  G(\rho_T^*)$ to conclude
    \begin{equation}
    \label{eq:opt-first}
        0 \ge  R(\rho^*_{\tau},w)-R(\rho^*_{\tau},u^*_{\tau})+\langle \phi^*_{\tau},\sigma_{\tau} \rangle .
    \end{equation}
    On the other hand, we have
    \begin{align}
    \langle \phi^*_{\tau},\sigma_{\tau} \rangle 
    & = \bigl\langle \phi^*_{\tau},-\nabla\cdot \bigl(w-u_{\tau}^*)\rho_{\tau}^*\bigr) \bigr\rangle \label{eq:phi-sigma-at-tau} \\
    & = \langle \utaustar \cdot \nabla \phi^*_{\tau},\rho_{\tau}^* \rangle - \langle w \cdot \nabla \phi^*_{\tau},\rho_{\tau}^* \rangle, \nonumber
    \end{align}
    where the first equality is due to the value of $\sigmatau$ in the initial condition in \eqref{eq:sigma-ivp}, and the last equality is due to integration by parts.
    Combining \eqref{eq:opt-first} and \eqref{eq:phi-sigma-at-tau}, we have
    \begin{align*}
        0 
        & \ge  \bigl( R(\rho^*_{\tau},w)+ \langle \utaustar \cdot \nabla \phi^*_{\tau},\rho_{\tau}^* \rangle \bigr) - \bigl(R(\rho^*_{\tau},u^*_{\tau}) + \langle w \cdot \nabla \phi^*_{\tau},\rho_{\tau}^* \rangle \bigr) \\
        & = H(\rhotaustar,\phi_{\tau}^*,w) - H(\rhotaustar,\phi_{\tau}^*,\utaustar) .
    \end{align*}
    As this is true for each $ \tau \in [0,T]$ and $w, \utaustar \in U$, the claim \eqref{eq:pmp-odc} is proved.
    % Lastly, to show $H(\rhotstar, \phi_t^*, \utstar)$ is constant in $t$ when $\ustar \in \intr(U_T)$, we have 
    % \begin{align*}
    %     \frac{d}{dt} H(\rhotstar, \phi_t^*, \utstar) 
    %     & = \langle \frac{\delta}{\delta \rho} H(\rhotstar, \phi_t^*, \utstar), \partial_t \rhotstar \rangle 
    %     + \langle \delta_\phi H(\rhotstar, \phi_t^*, \utstar), \partial_t \phi_t^* \rangle 
    %     + \langle \frac{\delta}{\delta u} H(\rhotstar, \phi_t^*, \utstar), \partial_t \utstar \rangle \\
    %     & = \langle -\partial_t \phi_t^*, \partial_t \rhotstar \rangle 
    %     + \langle \partial_t \rhotstar, \partial_t \phi_t^* \rangle + 0  \\
    %     & = 0,
    % \end{align*}
    % where the second equality is due to the property of the control Hamiltonian system \eqref{eq:Hamiltonian} applied to the first two terms, while the last term vanishes is due to the optimality of $\utstar$ justified in \eqref{eq:pmp-odc} that implies $\frac{\delta}{\delta u} H(\rhotstar, \phi_t^*, \utstar) = 0$ when $\ustar \in \intr(U_T)$. This completes the proof.
\end{proof}
%
%\vspace{6pt}

\begin{example}
\label{ex:mp}
    Let $T>0$ be arbitrary and fixed, and $\Omega = \Rbb^{d}$. Consider the control problem \eqref{eq:control-problem} with running reward
    \begin{equation}
        \label{eq:ex-R}
        R(\rho_{t}, u_{t}) = - \frac{1}{2} \int_{\Omega} |u_{t}(x)|^{2} \rho_{t}(x) \, dx
    \end{equation}
    and terminal reward
    \begin{equation}
        \label{eq:ex-G}
        G(\rho_{T}) = - \frac{1}{2} \int_{\Omega} |x|^{2} \rho_{T}(x) \, dx .
    \end{equation}
    We can choose any positive $p \in P$, namely $p(x)>0$ for all $x \in \Omega$, as the initial $\rho_{0}$ in \eqref{eq:control-problem}.

    We can set the terminal reward \eqref{eq:ex-G} to $G(\rho_{T}) = - \frac{1}{2} \int_{\Omega} |x - x^{*}|^{2} \rho_{T}(x) \, dx$ for any $x^{*}$, and the calculations below are similar. We use $x^{*}=0$ for simplicity here.

    By Definition \ref{def:Hamiltonian}, the Hamiltonian functional is
    \begin{equation}
        \label{eq:ex-H}
        H(\rho_{t},\phi_{t}, u_{t}) = \langle \rho_{t}, u_{t} \cdot \nabla \phi_{t}\rangle - \frac{1}{2} \int_{\Omega} |u_{t}|^{2} \rho_{t} \, dx . \nonumber
    \end{equation}
    From \eqref{eq:hamiltonian-dynamics}, we have the control Hamiltonian dynamics of $(\rho,\phi)$ as
    \begin{equation}
        \label{eq:ex-HD}
        \begin{cases}
            \partial_{t} \rho_{t} = - \nabla \cdot (\rho_{t} u_{t}) , \\
            \partial_{t} \phi_{t} = - u_{t} \cdot \nabla \phi_{t} + \frac{1}{2} |u_{t}|^{2} ,
        \end{cases}
    \end{equation}
    with initial value of $\rho$ as
    \begin{equation}
        \label{eq:ex-rho-initial}
        \rho_{0} = p
    \end{equation}
    and terminal value of $\phi$ as
    \begin{equation}
        \label{eq:ex-phi-terminal}
        \phi_{T} = \frac{\delta}{\delta \rho_{T}}G(\rho_{T}) = - \frac{1}{2}|x|^{2} .
    \end{equation}

    By the maximum principle \eqref{eq:pmp-odc}, there is
    \begin{align}
        H(\rho_{t}, \phi_{t}, u_{t}) 
        & = \max_{w \in U} H(\rho_{t}, \phi_{t}, w) \label{eq:ex-control-H} \\
        & = \max_{w \in U} \Bigl\{ \langle \rho_{t}, w \cdot \nabla \phi_{t} \rangle - \frac{1}{2}\int_{\Omega}|w|^{2}\rho_{t}\, dx\Bigr\} . \nonumber
    \end{align}
    Completing the square in the maximization above, we deduce that the maximizer $u_{t}$ satisfies $\rho_{t}(u_{t} - \nabla \phi_{t}) = 0$.
    Assuming $\rho_{t} > 0$, we have
    \begin{equation}
        \label{eq:mp-u}
        u_{t}  = \nabla \phi_{t}.
    \end{equation}
    Plugging \eqref{eq:mp-u} into the control Hamiltonian dynamics \eqref{eq:ex-HD}, and combining with \eqref{eq:ex-rho-initial} and \eqref{eq:ex-phi-terminal}, we know that the optimal $\rho$ solves the initial value problem 
    \begin{equation}
        \label{eq:ex-rho-ivp}
        \begin{cases}
            \partial_{t} \rho_{t} + \nabla \cdot (\rho_{t} \nabla \phi_{t}) = 0, & \forall \, t \in [0,T], \\
            \rho_{0} = p , & 
        \end{cases}
    \end{equation}
    and its corresponding adjoint function $\phi$ solves the terminal value problem
    \begin{equation}
        \label{eq:ex-phi-tvp}
        \begin{cases}
            \partial_{t}\phi_{t} + \frac{1}{2} |\nabla \phi_{t}|^{2} = 0, & \forall\, t \in [0,T], \\
            \phi_{T} = - \frac{1}{2}|x|^{2} .  &
        \end{cases}
    \end{equation}
    We can verify that the solution to \eqref{eq:ex-phi-tvp} is 
    \begin{equation}
        \label{eq:ex-phi-sol}
        \phi_{t}(x) = \frac{|x|^{2}}{2(t-T-1)} .
    \end{equation}
    This implies that the optimal control vector field is
    \begin{equation}
        \label{eq:ex-u-sol}
        u_{t}(x) = \frac{x}{t - T - 1}
    \end{equation}
    for all $x$ and $t$. We will continue with this example to see its value functional and HJB equation in the next subsection.
\end{example}

\subsection{HJB Equation for Probability Control}
\label{subsec:hjb-odc}

In this subsection, we derive the Hamilton--Jacobi--Bellman (HJB) equation associated with the optimal probability control problem \eqref{eq:control-problem}.
To this end, we need to define value functional, which is an analogue to the value function in the classic optimal control.
Note that this is a time-evolving functional defined on the cross space $P \times [0,T]$. 
%
% Therefore, we need to assume that the functional derivative in $P$ and time derivative exist.
%
The definition of this value functional is as follows.
\begin{definition}[Value functional]
    \label{def:value-fnl}
    For any $p \in P$ and $t \in [0,T]$, define the time-evolving functional $V$ by
    \begin{equation}
    \label{eq:value-fn}
        V(p,t)=\sup_{u\in U_{T}} \Big\{ \int_t^T R(\rho_s,u_s) \, ds + G(\rho_T) \Bigr\}
    \end{equation}
    where $\rho_{s}$ solves the initial value problem:
    \begin{equation}
        \label{eq:value-fn-ivp}
        \begin{cases}
            \partial_s \rho_s + \nabla \cdot ( u_s \rho_s) = 0, \quad \forall\, s \in [t,T], & \\
            \rho_t = p. &
        \end{cases}
    \end{equation}
    We call $H$ the \emph{value functional} of \eqref{eq:control-problem}.
\end{definition}
%
%\vspace{6pt}

Note that, for any $t$, $V(\cdot,t): P \to \Rbb$ is a functional on the infinite-dimensional space of probability density functions on $\Omega$.
By contrast, value function in classical optimal control is function on some subset of finite-dimensional Euclidean space.

Now we are ready to establish the HJB equation of the value functional $V$ as follows.

%
%\vspace{6pt}

\begin{theorem}[HJB equation for probability control]
    \label{thm:hjb-odc}
    If $V(\cdot,t)$ is continuously Fr\'{e}chet differentiable in $P$ for all $t$ and $V(p,\cdot)$ is $C^{1}$ in $[0,T]$ for all $p \in P$, then $V: [0,T] \times P \to \Rbb$ satisfies
    \begin{equation}
        \label{eq:hjb-odc}
        \partial_t V(p,t)+\max_{w \in U } \Bigl\{ \Bigl\langle w \cdot \nabla  \frac{\delta}{\delta p} V(p,t), p \Bigr\rangle + R(p,w) \Bigr\} =0,    
    \end{equation}
    with terminal value $V(\cdot,T)=G(\cdot)$ in $\Omega$.
\end{theorem}

%
%\vspace{6pt}

\begin{proof}
    It is clear that $V(p,T) = G(p)$ for any $p \in P$ by \eqref{eq:value-fn} and \eqref{eq:value-fn-ivp}.
    Now let $ p \in P$, $t \in [0,T)$. For any $\epsilon > 0$, we consider the constant control $w\in U$ in the interval $[t,t+ \epsilon] \subset [0,T)$ and the associated evolution $\rho_s$ given by 
    \[
    \begin{cases}
        \partial_s \rho_s(x) + \nabla\cdot(w \rho_s)=0, & \forall \, s \in (t,t+\epsilon),\\
        \rho_t = p. &
    \end{cases}
    \]
    From the definition of the value functional \eqref{eq:value-fn}, we know
    \[
    V(p,t) \ge \int_t^{t+\epsilon}R(\rho_s,w) \, ds + V(\rho_{t+\epsilon}, t+\epsilon) .
    \]
    We can rearrange the inequality above and divide by $\epsilon$ to find
    \[
    0 \ge \frac{1}{\epsilon} \int_t^{t+\epsilon}R(\rho_s, w)\,ds + \frac{V(\rho_{t+\epsilon}, t+\epsilon)-V(p,t)}{\epsilon}.
    \] Taking $\epsilon \to 0$ and noticing $\rho_t = p$, we have
    \[
    0 \ge R(p,w)+\partial_t V(p,t)+\Bigl\langle \frac{\delta}{\delta p} V(p,t),\partial_t \rho_t \Bigr\rangle.
    \]
    Realizing $\partial_t \rho_t = -\nabla\cdot (w \rho_t) = - \nabla \cdot (w p)$ and using integration by parts, we have
    \[
    0 \ge R(p,w)+\partial_t V(p,t)+\Bigl\langle w \cdot \nabla \frac{\delta}{\delta p} V(p,t), p \Bigr\rangle.
    \] 
    As we chose $w$ arbitrarily, we can conclude
    \begin{equation}
        \label{eq:value-geq-zero}
        0 \ge \partial_t V(p,t)+\max_{w \in U}\Bigl\{\Bigl\langle w \cdot \nabla \frac{\delta}{\delta p} V(p,t),p \Bigr\rangle + R(p, w)\Bigr\}.
    \end{equation}

    Now we show the inequality in \eqref{eq:value-geq-zero} is actually an equality. To this end, we consider any $(t,p) \in [0,T) \times P$, and let $\ustar : [t,T] \times \Omega \to \Rbb$ and $\rhostar: [t,T] \times \Omega \to \Rbb$ be the optimal solution to the control problem defined by the value functional at $V(p,t)$ in \eqref{eq:value-fn}. Then for any $\epsilon>0$ we know by the optimality of $\rhostar$ and $\rhotstar = p$, there is
    \[
    V(\rhotstar,t)=\int_t^{t+\epsilon} R(\rho_s^*,u_s^*)\, ds+V(\rho_{t+\epsilon}^*, t+\epsilon).
    \] 
    We follow the same steps as before by dividing by $\epsilon$, taking a limit $\epsilon \to 0$, and using integration by parts to find
    \begin{equation}
        \label{eq:value-leq-zero}
        0=\partial_t V(\rhotstar,t)+\Bigl\langle \utstar \cdot \nabla \frac{\delta}{\delta p} V(\rhotstar,t), \rhotstar \Bigr\rangle+R(\rhotstar,u_t^*).
    \end{equation}
   Combining both \eqref{eq:value-geq-zero} and \eqref{eq:value-leq-zero}, $\rhotstar = p$, and $\utstar \in U$, we conclude the maximum can be attained and the equality in \eqref{eq:value-geq-zero} holds. This completes the proof.
\end{proof}
%
%\vspace{6pt}

\begin{example}
    \label{ex:hjb}
    We continue with Example \ref{ex:mp} with the running reward \eqref{eq:ex-R}, terminal reward \eqref{eq:ex-G}, and arbitrary initial value $\rho_{0} = p>0$. 

    Recall from \eqref{eq:hjb-odc} that the HJB equation is
    \begin{equation}
        \label{eq:ex-hjb}
        \partial_{t} V(p,t) + \max_{w \in U} \Bigl\{ \Bigl\langle p, w \cdot \nabla \frac{\delta}{\delta p}V(p,t) \Bigr\rangle - \frac{1}{2} \int_{\Omega} |w|^{2}p \, dx \Bigr\} = 0 .
    \end{equation}
    Inspired by the square completion in \eqref{eq:ex-control-H} and the optimal solution \eqref{eq:ex-u-sol}, we attempt $w = \frac{x}{t-T-1}$ as the maximizer in \eqref{eq:ex-hjb}. Completing the square, we obtain $\nabla \frac{\delta}{\delta p} V(p,t) = w$, from which we deduce that
    \begin{equation}
        \label{eq:ex-value-fn}
        V(p,t) = \frac{1}{2(t-T-1)} \int_{\Omega} |x|^{2} p(x) \, dx .
    \end{equation}
    We can verify that $V(p,t)$ in \eqref{eq:ex-value-fn} indeed satisfies the HJB equation \eqref{eq:ex-hjb} with the maximum attained at $w = \frac{x}{t-T-1}$ for all $t$. Moreover, the terminal value of $V(p,t)$ is
    \begin{equation}
        \label{eq:ex-value-fn-terminal}
        V(p,T) = - \frac{1}{2} \int_{\Omega} |x|^{2} p(x) \, dx  = G(p).
    \end{equation}
    Notice that \eqref{eq:ex-value-fn} also suggests the optimal control to be
    \begin{equation}
        \label{eq:ex-u-sol-value}
        u_{t}(x) = \nabla \frac{\delta}{\delta \rho_{t}} V(\rho_{t},t) = \frac{x}{t-T-1},
    \end{equation}
    which agrees with \eqref{eq:ex-u-sol}.
\end{example}

\subsection{Variations of Initial Perturbation}
\label{subsec:var-init}

In this subsection, we consider the Fr\'{e}chet derivative of $G(\rho_{T})$ with respect to the initial value $p$, where $\rho: \Omega \times [0,T] \to \Rbb$ solves the continuity equation \eqref{eq:ct-eq} with initial value $\rho_{0} = p$.
The result demonstrates an important property of the adjoint function $\phi$.

Suppose the initial states $\{x_{i} \in \Omega: i = 1,\dots,N\}$ of agents are perturbed, changing the initial probability $p$ to
\begin{equation}
    \label{eq:perturb-init}
    p^{\epsilon} := p + \epsilon h + o(\epsilon) \in P, 
\end{equation}
with some $h, o(\epsilon): \Omega \to \Rbb$ satisfying
\begin{equation}
    \label{eq:h-condition}
    \int_{\Omega} h(x)\,dx = 0, \quad \int_{\Omega} o(\epsilon) \, dx = 0 .
\end{equation}
Note that \eqref{eq:h-condition} is necessary for $p^{\epsilon} \in P$.
% and $p^{\epsilon} \in P$. 
%
% Note that $h(x)=O(p(x))$ locally near $x$ with $p(x) \to 0$, as $p$ is approaching to the boundary of $P$ in \eqref{eq:P}, which is a bounded convex subset of $L^{2}(\Omega)$, and the neighborhood of $p$ becomes smaller.
%
Let $\rhoeps$ solve the initial value problem:
\begin{equation}
    \label{eq:var-ivp}
    \begin{cases}
        \partial_{t} \rhoteps + \nabla \cdot ( \rhoteps u) = 0, & \forall\, t \in [0,T] , \\
        \rhoeps_{0} = p^{\epsilon}. &
    \end{cases}
\end{equation}
We are interested in the difference between $\rhoteps$ from \eqref{eq:var-ivp} and $\rho_{t}$ from \eqref{eq:odc-ct-eq}--\eqref{eq:odc-init}, which use the same control vector field $u \in U_{T}$ but different initial values $p^{\epsilon}$ and $p$, respectively.

%
%\vspace{6pt}

\begin{lemma}
    \label{lem:var-g}
    Let $\rhoteps$ and $\rho_{t}$ solve \eqref{eq:var-ivp} with initials $p^{\epsilon}$ and $p$, respectively. Then 
    \begin{equation}
        \label{eq:var-rho-limit}
        \lim_{\epsilon \to 0} \sup_{0 \le t \le  T} \frac{1}{\epsilon} \| \rhoteps - \rho_{t} - \epsilon s_{t} \|_{2} = 0,
    \end{equation}
    where $s_{t}$ solves the initial value problem:
    \begin{equation}
        \label{eq:var-s-ivp}
        \begin{cases}
            \partial_{t} s_{t} + \nabla \cdot ( s_{t} u_{t}) =  0, & \forall\, t \in [0,T], \\
            s_{0} = h. &
        \end{cases}
    \end{equation}
\end{lemma}

%\vspace{6pt}

\begin{proof}
    Define $\delta(t) := \frac{1}{2}\| \rhoteps - \rho_{t} - \epsilon s_{t} \|_{2}^{2}$. Mimicking \eqref{eq:delta-pf} with $\sigma_{t}$ replaced by $s_{t}$, we obtain $\delta(t) \le \delta(0) e^{d M_{U}T}$ for all $t \in [0,T]$.
    Since $\delta(0) = \frac{1}{2}\|p^{\epsilon} - p \|_{2}^{2} = o(\epsilon^{2})$, we know 
    \begin{equation*}
        \lim_{\epsilon \to 0} \sup_{0 \le t \le T} \frac{(2\delta(t))^{1/2}}{\epsilon} \le \lim_{\epsilon \to 0} \frac{(2\delta(0))^{1/2}}{\epsilon} e^{d M_{U} T/2}= 0, 
    \end{equation*}
    which justifies \eqref{eq:var-rho-limit}.
\end{proof}

%\vspace{6pt}

\begin{proposition}
    \label{prop:var-g}
    For any $T > 0$, $u \in U_{T}$, initial value $p \in P$, let $\rho_{t}$ be solution to the initial value problem 
    \begin{equation}
        \label{eq:ex-var-rho-ivp}
        \begin{cases}
            \partial_{t} \rho_{t} + \nabla \cdot (\rho_{t} u_{t}) = 0, & \forall\, t \in [0,T] , \\
            \rho_{0} = p .
        \end{cases}
    \end{equation}
    Then for any functional $G: P \to \Rbb$ Fr\'{e}chet differentiable in $P$, there is
    \begin{equation}
        \label{eq:var-g}
        \frac{\delta}{\delta p} G(\rho_{T}) = \phi_{0} ,
    \end{equation}
    where $p=\rho_{0}$ is the initial value of $\rho$, and $\phi$ solves the terminal value problem:
    \begin{equation}
        \label{eq:var-phi-tvp}
        \begin{cases}
            \partial_{t} \phi_{t} + u_{t} \cdot \nabla \phi_{t} = 0, & \forall\, t \in [0,T], \\
            \phi_{T} = \frac{\delta}{\delta \rho_{T}} G(\rho_{T}) .
        \end{cases}
    \end{equation}
\end{proposition}
%
%\vspace{6pt}

\begin{proof}
    For any $h$ with \eqref{eq:h-condition}, let $p^{\epsilon}$ be the initial value defined in \eqref{eq:perturb-init} and $\rho^{\epsilon}$ solve the initial value problem \eqref{eq:var-ivp}, we know by Lemma \ref{lem:var-g} that  
    \begin{equation}
        G(\rho_{T}^{\epsilon}) = G\bigr(\rho_{T} + \epsilon s_{T} + o(\epsilon)\bigr) ,
    \end{equation}
    where $s_{t}$ is the solution to \eqref{eq:var-s-ivp}.
    Therefore
    \begin{align}
        \label{eq:var-G-terminal}
        \frac{d}{d \epsilon} G(\rho_{T}^{\epsilon}) \Big|_{\epsilon = 0} = \Bigl\langle \frac{\delta}{\delta \rho_{T}} G(\rho_{T}), s_{T} \Bigr\rangle = \langle\phi_{T}, s_{T}\rangle ,
    \end{align}
    where the second equality is due to the terminal condition of $\phi_{t}$ in \eqref{eq:var-phi-tvp}. 
    
    On the other hand, notice that
    \begin{align*}
        \frac{d}{dt} \langle\phi_{t}, s_{t}\rangle
        & = \langle \partial_{t}\phi_{t}, s_{t}\rangle + \langle\phi_{t}, \partial_{t} s_{t}\rangle \\
        & = - \langle u_{t} \cdot \nabla \phi_{t}, s_{t}\rangle - \langle \phi_{t}, \nabla \cdot (s_{t}u_{t}) \rangle \\
        & = 0 ,
    \end{align*}
    where the second equality is due to the PDE of $\phi$ in \eqref{eq:var-phi-tvp} and the PDE of $s$ in \eqref{eq:var-s-ivp}; and the third equality is due to integration by parts and $(u_{t}\cdot n)|_{\partial \Omega}=0$ on $\partial \Omega$.
    Therefore, $\langle\phi_{t}, s_{t}\rangle$ is constant in $t$ and hence 
    \begin{equation}
        \label{eq:var-ip-0T}
        \langle\phi_{T}, s_{T}\rangle = \langle\phi_{0}, s_{0}\rangle = \langle\phi_{0}, h\rangle .
    \end{equation}

    Combining \eqref{eq:var-G-terminal} and \eqref{eq:var-ip-0T}, and noticing that $\frac{\delta}{\delta p} G(\rho_{T}) = \frac{d}{d \epsilon} G(\rho_{T}^{\epsilon})|_{\epsilon = 0}$, we obtain 
    \begin{equation}
        \Bigl\langle\frac{\delta}{\delta p} G(\rho_{T}), h \Bigr\rangle = \langle\phi_{0}, h\rangle .
    \end{equation}
    Since $h$ is arbitrary, we know \eqref{eq:var-g} holds.
\end{proof}
%
%\vspace{6pt}

\begin{example}
    \label{ex:fn-var}
    Let $T>0$ and $p \in P$ be given arbitrarily and $\Omega = \Rbb^{d}$. Define $u \in U_{T}$ by
    \begin{equation}
        \label{eq:ex-var-u}
        u_{t}(x) = \frac{x}{2(t-T-1)} 
    \end{equation}
    for all $x \in {\Omega}$ and $t \in [0,T]$.
    Let $\rho$ be the solution to \eqref{eq:ex-var-rho-ivp} with the vector field $u$ in \eqref{eq:ex-var-u} and initial value $p$. Define a functional $G: P \to \Rbb$ such that 
    \begin{equation}
        \label{eq:ex-var-G}
        G(q) = -\frac{1}{2} \int_{\Omega} |x|^{2} q(x) \, dx, \quad \forall \, q \in P .
    \end{equation}
    Notice that $\frac{\delta}{\delta q}G(q) = - \frac{1}{2}|x|^{2}$ for all $q$.
    
    We can verify that
    \begin{equation}
        \label{eq:ex-var-phi-sol}
        \phi_{t}(x) = \frac{|x|^{2}}{2(t-T-1)}
    \end{equation}
    is the solution to the terminal value problem \eqref{eq:var-phi-tvp} with $u$ in \eqref{eq:ex-var-u} and terminal value $\phi_{T} = \frac{\delta}{\delta \rho_{T}}G(\rho_{T}) = - \frac{1}{2}|x|^{2}$. 
    
    Now, from Proposition \ref{prop:var-g}, we claim that
    \begin{equation}
        \label{eq:ex-G-var}
        \frac{\delta}{\delta p} G(\rho_{T}) = \phi_{0} = - \frac{|x|^{2}}{2(T+1)} .
    \end{equation}

    To justify the claim \eqref{eq:ex-G-var}, we first notice that 
    \begin{equation*}
        \label{eq:ex-var-rho}
        \rho_{t}(x) = \Bigl(\frac{T+1}{T+1-t}\Bigr)^{d/2} p\biggl(\Bigl(\frac{T+1}{T+1-t}\Bigr)^{1/2}x\biggr) 
    \end{equation*}
    is the solution to the initial value problem \eqref{eq:ex-var-rho-ivp} with the vector field $u$ in \eqref{eq:ex-var-u} and the given initial value $p$. Therefore
    \begin{equation*}
        \label{eq:ex-var-rhoT}
        \rho_{T}(x) = (T+1)^{d/2} p\bigl( \sqrt{T+1} x \bigr) .
    \end{equation*}
    Hence we obtain
    \begin{align*}
        G(\rho_{T})
        & = - \frac{1}{2} \int_{\Omega} |x|^{2} \rho_{T}(x) \, dx \\
        & = - \frac{1}{2} \int_{\Omega} |x|^{2} (T+1)^{d/2} p\bigl( \sqrt{T+1} x \bigr) \, dx \\
        & = - \frac{1}{2(T+1)} \int_{\Omega} |y|^{2} p(y) \, dy ,
    \end{align*}
    where the last equality is due to the change of variables $y = \sqrt{T+1}x$. Therefore $\frac{\delta}{\delta p} G(\rho_{T}) = - \frac{|x|^{2}}{2(T+1)}$, which justifies \eqref{eq:ex-G-var}.
\end{example}

\section{Neural ODE Method}
\label{sec:node}

While we have seen many important theoretical results for optimal control above, it remains a question on how to solve general control problems numerically in practice.
In this section, we introduce an approach, called the \emph{neural ordinary differential equation} method, to tackle this problem. We call it the Neural ODE method\index{Neural ODE method} for short hereafter.

We focus on the following general optimal control problem with fixed end-time $T$ and free endpoint given in Section \ref{sec:oc-theory}:
\begin{subequations}
\label{eq:node-oc-problem}
\begin{align}
\max_{u \in \Ucal_{T}} \quad & I[u] := g(x(T)), \\
\text{s.t.} \quad & 
\begin{cases}
\xdot(t) = f(t, x(t), u(t)), \quad \forall \, t \in [0,T] , \\
x(0) = x_{0}
\end{cases}
\end{align}
\end{subequations}
where the terminal reward function $g: \mathbb{R}^{d} \to \mathbb{R}$, the defining function $f: [0,T] \times \mathbb{R}^{d} \times \Ucal \to \mathbb{R}^{d}$ including the control as an argument, and some initial value $x_{0} \in \mathbb{R}^{d}$ are given.

The reason that we focus on \eqref{eq:node-oc-problem} is that it covers the majority of optimal control problems in real-world applications. It is also partially due to a practical issue as we will use deep neural networks to approximate the control $u(t)$, and it is sometimes difficult for such approximation to satisfy certain constraints exactly (such as arriving at some specified state precisely in the fixed endpoint optimal control problem).

We also remark that the time variable $t$ is back to $f$ as an explicit input in \eqref{eq:node-oc-problem} for ease of presentation below. Meanwhile, we do not have the running reward $r$ in \eqref{eq:node-oc-problem} as we can simply add an auxiliary variable as in \eqref{eq:pmp-x-aux} to take the running reward into account.

The idea of the Neural ODE method is to parameterize the control function $u$ as a deep neural network $u_{\theta}: [0,T] \to \Ucal \subset \mathbb{R}^{m}$, where $\theta \in \mathbb{R}^{n}$ represents the vector consisting of its $n$ network parameters.

\begin{figure}
\centering
\begin{tikzpicture}[scale=1]
\Vertex[x=0, y=2, label=$x(t)$, color=none, size=0.7]{A} 
\Vertex[x=2, y=2, shape = rectangle, label=$f$, color=none, size=0.7]{B}
\Vertex[x=5, y=2, label=$\dot{x}(t)$, color=none, size=0.7]{C}
\Vertex[x=0, y=0, label=$t$, color=none, size=0.7]{D}
\Vertex[x=2, y=0, shape = rectangle, label=$u_{\theta}$, color=none, size=0.7]{E}
\Edge[Direct, lw=1pt](A)(B)
\Edge[Direct, lw=1pt](B)(C)
\Edge[Direct, lw=1pt](D)(B)
\Edge[Direct, lw=1pt](D)(E)
\Edge[Direct, lw=1pt](E)(B)
\Edge[bend=310, Direct, lw=1pt](C)(A)
\draw (2.5, 0.7) node[above]{{\scriptsize $u_{\theta}(t)$}};
\draw (3.5, 2) node[above]{{\scriptsize $f(t,x(t),u_{\theta}(t))$}};
\node at (2.5, 3.8) [align=center]{{\scriptsize $\dot{x}(t)$ determines the} \\ {\scriptsize instantaneous change of $x(t)$}};
\end{tikzpicture}
\caption{The control system solved by the Neural ODE method. The control ODE is given by $\dot{x}(t) = f(t,x(t),u(t))$, where $f: [0,T] \times \mathbb{R}^{d} \times \Ucal \to \mathbb{R}^{d}$ is determined by control problem, and the control $u:[0,T] \to \Ucal$ is parameterized as a deep neural network $u_{\theta}$ with parameter $\theta$. In the original Neural ODE method \cite{chen2018neural}, the function $f:[0,T] \times \mathbb{R}^{d} \to \mathbb{R}^{d}$ is parameterized as a deep neural network $f_{\theta}$ with parameter $\theta$, and the ODE is $\dot{x}(t) = f_{\theta}(t,x(t))$.}
\label{fig:node}
\end{figure}
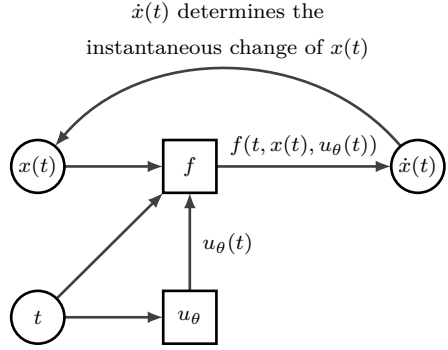

Now the optimal control problem \eqref{eq:node-oc-problem} reduces to
\begin{subequations}
\label{eq:node-oc-theta}
\begin{align}
\max_{\theta \in \mathbb{R}^{n}} \quad & J(\theta) := I[u_{\theta}] = g(x(T)), \label{subeq:node-oc-theta-obj} \\
\text{s.t.} \quad & 
\begin{cases}
\xdot(t) = f(t, x(t), u_{\theta}(t)), \quad \forall \, t \in [0,T] , \label{subeq:node-oc-theta-ivp} \\
x(0) = x_{0} .
\end{cases}
\end{align}
\end{subequations}
Note that the variable to be optimized is no longer the time-dependent control function $u: [0,T] \to \Ucal$.
Instead, it is the vector $\theta \in \mathbb{R}^{n}$ to be optimized, just like in standard optimization.
Once the optimal value $\theta$ is found, the neural network $u_{\theta}: [0,T] \to \Ucal$ is (approximately) the desired optimal solution $u$ to the control problem \eqref{eq:node-oc-problem}. 
The control system with control function $u_{\theta}$ is illustrated in Figure \ref{fig:node}.

Given that the variable to be optimized is a vector $\theta$, it is natural to employ some gradient-based optimization algorithm we introduced in Section \ref{sec:deterministic-opt}. The key ingredient when using such algorithm is the gradient of the objective function $J$ with respect to $\theta$. 
Therefore, our main goal is to find $\nabla_{\theta} J(\theta)$ for the function $J$ defined in \eqref{eq:node-oc-theta}.
If we can compute $\nabla_{\theta} J(\theta)$ at any $\theta$, we can supply it to any gradient-based optimization algorithm to find the (local) maximizers of $J(\theta)$, which are (sub-)optimal network parameter $\theta$ of $u_{\theta}$.
The Neural ODE method provides a means to compute $\nabla_{\theta} J(\theta)$.

To show how the Neural ODE method works, we first introduce two trivial auxiliary functions. The first one $\tau: [0,T] \to [0,T]$ is the identity function:
\begin{equation*}
\tau(t)=t
\end{equation*}
for all $t \in [0,T]$ and hence it solves the initial value problem
\begin{equation}
\begin{cases}
\dot{\tau}(t) = 1 , \quad \forall \, t \in [0,T] , \\
\tau(0) = 0 .
\end{cases}
\end{equation}
The second one $\sigma: [0,T] \to \mathbb{R}^{n}$ is a constant function in time: 
\begin{equation*}
\sigma(t) = \theta
\end{equation*}
for all $t \in [0,T]$ and hence it solves the initial value problem
\begin{equation}
\begin{cases}
\dot{\sigma}(t) = 0 , \quad \forall \, t \in [0,T] , \\
\sigma(0) = \theta .
\end{cases}
\end{equation}
Furthermore, we denote the augmented function $z: [0,T] \to \mathbb{R}^{d+1+n}$ as 
\begin{equation}
z(t) := 
\begin{pmatrix}
x(t) \\ \tau(t) \\ \sigma(t) 
\end{pmatrix} \in \mathbb{R}^{d+1+n}, \quad \text{with} \quad
z(0) := 
\begin{pmatrix}
x_{0} \\ 0 \\ \theta
\end{pmatrix} \in \mathbb{R}^{d+1+n} ,
\end{equation}
the augmented ODE defining function $h$ of $z$ as
\begin{equation}
h(z(t)) := 
\begin{pmatrix}
f(\tau(t), x(t), u_{\sigma(t)}(\tau(t)) \\ 1 \\ 0
\end{pmatrix} \in \mathbb{R}^{d+1+n} ,
\end{equation}
and the augmented terminal reward function $\bar{g}: \mathbb{R}^{d+1+n} \to \mathbb{R}$ as
\begin{equation*}
\bar{g}(z) := g(x)
\end{equation*}
for any $z = (x, \tau, \sigma) \in \mathbb{R}^{d+1+n}$.
Note that we know the expressions of $u_{\theta}$ (a neural network architecture of our choice) and $f$ (the given ODE defining function of the control problem \eqref{eq:node-oc-problem}).
Moreover, we see that the ODE $\zdot(t) = h(z(t))$ is autonomous\index{Autonomous} ($h$ does not depend on $t$ explicitly).

Now we can rewrite the optimal control problem \eqref{eq:node-oc-theta} in an equivalent form:
\begin{subequations}
\begin{align}
\max_{\theta \in \mathbb{R}^{n}} \quad & J(\theta) := \bar{g}(z(T)), \\
\text{s.t.} \quad & 
\begin{cases}
\zdot(t) = h(z(t)), \quad \forall \, t \in [0,T] , \\
z(0) = z_{0} .
\end{cases}
\end{align}
\end{subequations}

For any $y_{0} \in \mathbb{R}^{d+1+n}$ and $\epsilon > 0$, consider $\zeps: [0, T] \to \mathbb{R}^{d+1+n}$ solving the initial value problem
\begin{equation*}
\begin{cases}
\dot{\zeps}(t) = h(\zeps(t)), \quad \forall \, t \in [0,T] , \\
\zeps(0) = z_{0} + \epsilon y_{0} .
\end{cases}
\end{equation*}
Then by Lemma \ref{lem:pmp-lemma1-x-ode}, we have
\begin{equation*}
\zeps(t) = z(t) + \epsilon y(t) + o(\epsilon)
\end{equation*}
as $\epsilon \to 0$ uniformly on $[0,T]$, namely,
\begin{equation*}
\sup_{0 \le t \le T} \frac{|\zeps(t) - z(t) - \epsilon y(t)|}{\epsilon} \to 0 
\end{equation*}
as $\epsilon \to 0$, where $y:[0,T] \to \mathbb{R}^{d+1+n}$ is the solution to the initial value problem
\begin{equation*}
\begin{cases}
\ydot(t) = \nabla_{z} h(z(t)) y(t), \quad \forall \, t \in [0, T] , \\
y(0) = y_{0} .
\end{cases}
\end{equation*}
Hence there is
\begin{equation*}
\frac{d}{d \epsilon} \gbar(\zeps(T)) \Big|_{\epsilon = 0} = \nabla_{z} \bar{g}(z(T)) \cdot y(T) .
\end{equation*}
Moreover, we know the co-state $p:[0,T] \to \mathbb{R}^{1 \times (d+1+n)}$ ($p(t)$ is a row vector) corresponding to the state $z$ solves the terminal value problem:
\begin{equation*}
\begin{cases}
\pdot(t) = - p(t) \nabla_{z} h(z(t)), \quad \forall \, t \in [0, T] , \\
p(T) = \nabla_{z} \gbar(z(T)) 
\end{cases}
\end{equation*}
and as usual there is
\begin{align*}
\frac{d}{d t} \Big( p(t) \cdot y(t) \Big) 
& = \pdot(t) \cdot y(t) + p(t) \cdot \ydot(t) \\
& = - p(t) \nabla_{z} h(z(t)) \cdot y(t) + p(t) \cdot \nabla_{z} h(z(t)) y(t) \\
& = 0
\end{align*}
for all $t$, which implies that $p(t) \cdot y(t)$ is constant in $t$.
Hence there is
\begin{align*}
\frac{d}{d \epsilon} \gbar(\zeps(T)) \Big|_{\epsilon = 0} = \nabla_{z} \gbar(z(T)) \cdot y(T)  = p(T) \cdot y(T) = p(0) \cdot y_{0} .
\end{align*}

We notice that $p(t)$ has three compartments
\begin{equation*}
p(t) = (p_{x}(t), p_{\tau}(t), p_{\sigma}(t)) \in \mathbb{R}^{1 \times (d+1+n)}
\end{equation*}
and denote
\begin{equation*}
z :=
\begin{pmatrix}
x \\ \tau \\ \sigma
\end{pmatrix} \in \mathbb{R}^{d+1+n} \quad \text{and} \quad
h(z) :=
\begin{pmatrix}
f(\tau, x, u_{\sigma}(\tau)) \\ 1 \\ 0
\end{pmatrix}\in \mathbb{R}^{d+1+n} .
\end{equation*}
Then we have
\begin{equation*}
\partial_{z} h(z) = 
\begin{pmatrix}
\partial_{x} f  & \partial_{\tau} f + \partial_{u} f \partial_{\tau} u & \partial_{u} f \partial_{\sigma}u \\
0 & 0 & 0 \\
0 & 0 & 0
\end{pmatrix} \in \mathbb{R}^{(d+1+n) \times (d+1+n)} .
\end{equation*}
Notice that $\partial_{\tau} f = \partial_{t} f$, $\partial_{\tau} u = \partial_{t} u$, and $\partial_{\sigma}u = \partial_{\theta} u$ since $\tau(t)=t$ and $\sigma(t)=\theta$ for all $t \in [0,T]$.

Hence, the co-state $p=(p_{x}, p_{\tau}, p_{\sigma}) : [0, T] \to \mathbb{R}^{1 \times (d+1+n)}$ solves the terminal value problem with the adjoint equation
\begin{equation*}
\pdot(t) = -p(t) \nabla_{z} h(z(t)),
\end{equation*}
namely,
\begin{equation}
\label{eq:node-adjoint}
\begin{cases}
\pdot_{x}(t) = - p_{x}(t) \partial_{x} f(t, x(t), u_{\theta}(t)) \in \mathbb{R}^{1 \times d} \\
\pdot_{\tau}(t) = - p_{x}(t) \Big(\partial_{t} f(t, x(t), u_{\theta}(t)) + \partial_{u} f(t,x(t),u_{\theta}(t)) \partial_{t} u_{\theta}(t) \Big) \in \mathbb{R}^{1 \times 1} \\
\pdot_{\sigma}(t) = - p_{x}(t) \Big( \partial_{u} f(t, x(t), u_{\theta}(t)) \partial_{\theta} u_{\theta}(t) \Big) \in \mathbb{R}^{1 \times n}
\end{cases}
\end{equation}
and terminal value
\begin{equation*}
p(T) = \gbar(z(T)) = g(x(T)),
\end{equation*}
namely,
\begin{equation}
\label{eq:node-terminal}
\begin{cases}
p_{x}(T) = \nabla g(x(T)) \in \mathbb{R}^{1 \times d} , \\
p_{\tau}(T) = 0 \in \mathbb{R}^{1 \times 1} , \\
p_{\sigma}(T) = 0 \in \mathbb{R}^{1 \times n} .
\end{cases}
\end{equation}

Suppose we choose $\eta \in \mathbb{R}^{n}$ arbitrarily and set $y_{0} = (0, 0, \eta^{\top})^{\top} \in \mathbb{R}^{d+1+n}$, then the derivations above yield
\begin{align*}
\nabla_{\theta} J(\theta) \cdot \eta 
= \frac{d}{d\epsilon} J(\theta + \epsilon \eta) \Big|_{\epsilon = 0} 
= \frac{d}{d\epsilon} \gbar(\zeps(T)) \Big|_{\epsilon = 0} = p(0) \cdot y_{0} = p_{\sigma}(0) \cdot \eta .
\end{align*}
As $\eta$ is arbitrary, we know
\begin{equation}
\label{eq:node-dJ}
\nabla_{\theta} J(\theta) = p_{\sigma}(0) .
\end{equation}

Now we see that, to compute $\nabla_{\theta} J(\theta)$ for any $\theta$, we can just solve the initial value problem \eqref{subeq:node-oc-theta-ivp} with this $\theta$ for $x(T)$, and then solve the terminal value problem \eqref{eq:node-adjoint}--\eqref{eq:node-terminal} for $p_{\sigma}(0)$. Then we obtain \eqref{eq:node-dJ}.
We summarize the procedure to compute $\nabla_{\theta} J(\theta)$ in Algorithm \ref{alg:node-grad}.
Note that both of the initial and terminal value problems can be solved by well-developed numerical ODE solvers\index{Numerical ODE solver}, such as the midpoint method and the Runge--Kutta method \cite{burden2011numerical}.
As $\nabla_{\theta} J(\theta)$ can be evaluated, we can employ any gradient-based optimization algorithm to find a (local) maximizer of $J(\theta)$.
\begin{algorithm}
\caption{Compute $\nabla_{\theta} J(\theta)$ for \eqref{eq:node-oc-theta} in the Neural ODE method}
\label{alg:node-grad}
\begin{algorithmic}[1]
\REQUIRE Current $\theta \in \mathbb{R}^{n}$.
	\STATE Solve $x(T)$ from the initial value problem:
	\begin{equation*}
	\begin{cases}
	\xdot(t) = f(t, x(t), u_{\theta}(t)), \quad \forall \, t \in [0, T] , \\
	x(0) = x_{0} .
	\end{cases}
	\end{equation*}
	\STATE Solve $p(0) = (p_{x}(0), p_{\tau}(0), p_{\sigma}(0))$ from the terminal value problem:
	\begin{equation*}
	\begin{cases}
	\pdot(t) = - p(t) \nabla_{z} h(z(t)), \quad \forall \, t \in [0, T] , \\
	p(T) = (\nabla g(x(T)), 0, 0 ) ,
	\end{cases}
	\end{equation*}
	where the explicit form of the adjoint equation is given in \eqref{eq:node-adjoint}.
\ENSURE $\nabla_{\theta} J(\theta) = p_{\sigma}(0)$. 
\end{algorithmic}
\end{algorithm}

\subsubsection*{Remarks on the Neural ODE Method}

We have several remarks on the Neural ODE method in solving optimal control problems.

First, the Neural ODE method parameterizes the control $u:[0,T] \to \Ucal$ as a neural network $u_{\theta}$, the solution control is continuous in time since we almost always design neural networks as continuous functions in practice.
The issue is that we cannot find the optimal control if it is not continuous in time, but a close approximation may be possible if the network architecture is expressive enough.
Meanwhile, an advantage of using deep network to parameterize the control is that we can design the network properly such that it has strong representation power while its outputs are always in $\Ucal$ (e.g., $\Ucal$ is a bounded set in $\mathbb{R}^{m}$) automatically, avoiding explicit constraint enforcement when learning the control. Some examples of neural network design are given in Section \ref{subsec:special-nets}.

Second, the Neural ODE method was originally proposed in \cite{chen2018neural} as a continuous-time version of the residual network (ResNet) architecture, which can be thought of as the Euler scheme for solving ODEs with defining function $f$ parameterized as deep networks. More specifically, the ODE in \cite{chen2018neural} has a simple yet flexible form where $f$ is directly parameterized as a deep neural network $f_{\theta}: [0,T] \times \mathbb{R}^{d} \to \mathbb{R}^{d}$ with parameter $\theta \in \mathbb{R}^{n}$:
\begin{equation}
\label{eq:node-oc}
\begin{cases}
\xdot(t) = f_{\theta}(t, x(t)) , \quad \forall \, t \in [0, T] , \\
x(0) = x_{0} .
\end{cases}
\end{equation}
The adjoint equation of the co-state $p$ can be derived similarly as we did above and is in a form simpler than \eqref{eq:node-adjoint}:
\begin{equation}
\label{eq:node-adjoint-original}
\begin{cases}
\pdot_{x}(t) = - p_{x}(t) \partial_{x} f_{\theta}(t, x(t)) \in \mathbb{R}^{1 \times d} , \\
\pdot_{\tau}(t) = - p_{x}(t) \partial_{t} f_{\theta}(t, x(t)) \in \mathbb{R}^{1 \times 1} , \\
\pdot_{\sigma}(t) = - p_{x}(t) \partial_{\theta} f_{\theta}(t, x(t)) \in \mathbb{R}^{1 \times n} ,
\end{cases}
\end{equation}
and the terminal value is again $p(t) = (\nabla g(x(T)), 0, 0) \in \mathbb{R}^{1 \times (d+1+n)}$.
Then we can specify a terminal reward $g(x(T))$ as the objective function $J(\theta)$, and the gradient $\nabla_{\theta} J(\theta)$ can be computed in the same way as in Algorithm \ref{alg:node-grad} with the initial value and terminal value problems replaced with \eqref{eq:node-oc} and \eqref{eq:node-adjoint-original}, respectively.

Third, the Neural ODE method can be cast as a general deep learning approach. 
For example, suppose we have a standard supervised learning problem with a given dataset $\Dcal = \{(x_{i,0}, x_{i}^{*}) \in \mathbb{R}^{d} \times \mathbb{R}^{d}: i \in [M] \}$, and our goal is to find the mapping $F: \mathbb{R}^{d} \to \mathbb{R}^{d}$ such that $F(x_{i,0}) \approx x_{i}^{*}$ for all $i \in [M]$. Instead of directly parameterizing $F$ as a deep neural network, which may be difficult to train, we can set $F$ as the mapping from the initial $x_{i}(0) := x_{i,0}$ to its terminal $x_{i}(T)$, i.e., $F(x_{i,0}) := x_{i}(T)$ for all $i \in [M]\}$, by using the Neural ODE method with the ODE defining function parameterized as a neural network $f_{\theta}$, and train $\theta$ such that $x_{i}(T) \approx x_{i}^{*}$ for all $i \in [M]$. In this case, the network training problem is formulated as
\begin{subequations}
\label{eq:node-sl}
\begin{align}
\max_{\theta \in \mathbb{R}^{n}} \quad & J(\theta) := -\frac{1}{2 M} \sum_{i=1}^{M} |x_{i}(T) - x_{i}^{*} |^{2} , \label{subeq:node-sl-obj} \\
\text{s.t.} \quad & 
\begin{cases}
\xdot_{i}(t) = f_{\theta}(t, x_{i}(t)), \quad \forall \, t \in [0,T] , \\
x_{i}(0) = x_{i,0}, \quad i = 1,\dots,M.
\end{cases} \label{subeq:node-sl-ivp}
\end{align}
\end{subequations}
We can apply the Neural ODE method to solve many other problems as long as they can be formulated as optimization problems with ODE constraint.

Fourth, we can also leverage the Neural ODE method to solve many probability control problems. The idea is that we use a sufficient number of moving particles $\{x_{i}(t) \in \mathbb{R}^{d}: i \in [M]\}$ to represent the probability density function $\rho(\cdot, t)$ on $\mathbb{R}^{d}$ at any time $t$. The particle number $M$ may be artificially set or determined by the data available in the problem. When we parameterize the vector field as a deep neural network $f_{\theta}: [0,T] \times \mathbb{R}^{d} \to \mathbb{R}^{d}$, then these particles will move according to the ODE \eqref{subeq:node-sl-ivp}. Correspondingly, the evolution of $\rho(\cdot, t)$ is given by the following continuity equation:
\begin{equation}
\label{eq:node-rmk-ce}
\partial_{t} \rho(x,t) = - \nabla \cdot ( \rho(x,t) f_{\theta}(t,x) ) .
\end{equation}
Therefore, we can use the particles as the representation of $\rho$, and learn the optimal $f_{\theta}$ that makes $\rho(\cdot,t)$ maximize or minimize the given objective function. A concrete example of this approach is given in Section \ref{sec:probability-density-control} of Chapter \ref{chpt:gm}.

Fifth, when we use the Neural ODE method for probability control, we sometimes need to estimate the value of $\rho(x,t)$.
For example, as we will show in Section \ref{sec:probability-density-control}, we need to estimate the negative entropy
\begin{equation*}
\int \rho(x,t) \log \rho(x,t) dx \approx \frac{1}{M} \log \rho(x_{i}(t),t). 
\end{equation*}
where the approximation on the right-hand side is due to Monte Carlo integration (see Section \ref{appsec:mc} for details about this integration approximation method) and $x_{i}(t) \sim \rho(\cdot, t)$. 
Therefore, we need to estimate $\log \rho(x,t)$ using the moving particles. 
To this end, we can employ the following ODE:
\begin{equation}
\label{eq:node-logp}
\frac{d}{d t} \log \rho(x(t), t) = - \nabla \cdot f_{\theta}(t, x(t)) ,
\end{equation}
where $d/dt$ on the left-hand side is the full gradient with respect to $t$, i.e., the gradient with respect to $t$ is applied to both $x(t)$ and $t$ in $\rho(x(t),t)$, and $\nabla$ and $\nabla\cdot$ denote the gradient and divergence with respect to $x$, respectively.
To see \eqref{eq:node-logp}, we notice that
\begin{align*}
\frac{d}{d t} \rho(x(t), t) 
& = \partial_{t} \rho(x(t), t) + \nabla \rho(x(t), t) \cdot \xdot(t) \\
& = \partial_{t} \rho(x(t), t) + \nabla \rho(x(t), t) \cdot f_{\theta}(t, x(t)) \\
& = - \nabla \cdot ( \rho(t,x) f_{\theta}(t,x) ) + \nabla \rho(x(t), t) \cdot f_{\theta}(t, x(t)) \\
& = - \rho(x(t), t) \nabla \cdot f_{\theta}(t, x(t)) ,
\end{align*}
where the second equality is due to the ODE $\xdot(t) = f_{\theta}(t,x(t))$, and the third equality is due to the continuity equation \eqref{eq:node-rmk-ce}.
Diving both sides by $\rho(x(t),t)$ (assuming $\rho$ is positive everywhere), we obtain \eqref{eq:node-logp}.

Notice that \eqref{eq:node-rmk-ce} only provides a means to compute the change of $\log \rho(x(t),t)$. In practice, we may not know the initial $\rho(\cdot,0)$ (or the terminal $\rho(\cdot,T)$ at some given $T>0$). However, this is usually not an issue in learning $\theta$, because the initial or terminal is often given in advance and does not depend on $\theta$. Therefore, we can often eliminate it from the objective function during modeling without affecting the solution $\theta$ in the network training process.

%We want to add two additional remarks on \eqref{eq:node-logp}. In particular, in certain computations, the divergence $\nabla \cdot f_{\theta}(x(t),t)$ can be estimated using an estimation.
%%
%For example, we notice that
%\begin{equation}
%\nabla \cdot f_{\theta}(x(t),t) = \mathrm{tr}(\nabla f_{\theta}(x(t),t))
%\end{equation}
%where the term in the trace is the Jacobian of $f$ with respect to $x(t)$. This trace can be estimated by
%\begin{equation*}
%\mathrm{tr}(\nabla f_{\theta}(x(t),t) ) \approx \mathbb{E}_{z}[z^{\top} \nabla f_{\theta}(x(t),t) z ]
%\end{equation*}
%with random samples $z \in \mathbb{R}^{d}$ drawn from simple distributions with mean $0$ and variance $I_{d}$, such as the standard normal distribution $N(0, I_{d})$.
%%
%To see this, we denote $A = \nabla f_{\theta}(x(t),t)$ for simplicity and notice that
%\begin{align*}
%\mathbb{E}_{z} [z^{\top} A z ]
%& = \mathbb{E}_{z} \Big[ \sum_{i,j=1}^{d} a_{ij} z_{i} z_{j} \Big] = \sum_{i,j=1}^{d} a_{ij} \mathbb{E}_{(z_{i}, z_{j})} \\
%& = \sum_{i,j=1}^{d} a_{ij} \delta_{ij} = \sum_{i=1}^{d} a_{ii} = \mathrm{tr}(A) .
%\end{align*}

\section{Remarks and References}

\subsection{Application in PDE Solution Operator Learning}
\label{subsec:node-so}

In this subsection, we present an application of deep optimal control in learning solution operators of evolution partial differential equations (PDEs)\index{Differential equation!partial}. 
We first provide an overview of evolution PDEs and the definition of solution operators.
Then we show how to learn solution operators using deep optimal control. 
We focus on the critical components and key ideas during the discussions below and briefly mention extensions to general cases.

An evolution PDE refers to a PDE that describes how a system changes over time, modeling dynamic processes like heat flow, waves, fluid and etc. 
Suppose $\Omega$ is an open subset of $\mathbb{R}^{d}$ and $F$ is a differential operator defining an evolution PDE. Then the evolution PDE is written as
\begin{equation}
\label{eq:evo-pde}
\partial_t v(x,t) = F [v](x,t), \quad \forall \, (x,t) \in \Omega \times [0,T],
\end{equation}
where $T>0$ is some prescribed terminal time.

Here are a few examples of evolution PDEs (where $\gamma>0$ is some given constant) that appear in a broad range of real-world problems:
\begin{itemize}
\item
Heat equation:
\begin{equation}
\label{eq:heat-example}
\partial_{t} v(x,t) = F[v](x,t) := \Delta v(x,t) .
\end{equation}
\item
Hamilton--Jacobi equation:
\begin{equation}
\label{eq:hj-example}
\partial_{t} v(x,t) = F[v](x,t) := \Delta v(x,t) + \frac{\gamma}{2} |\nabla v(x,t)|^{2} .
\end{equation}
\item
Allen--Cahn equation:
\begin{equation}
\label{eq:ac-example}
\partial_{t} v(x,t) = F[v](x,t) := \Delta v(x,t) - \gamma v(x,t)(v(x,t)^{2} - 1) .
\end{equation}
\end{itemize}
In the examples above, $\nabla$ and $\Delta$ are respectively the gradient and Laplacian of $u$ with respect to $x$ as usual, and $v: \bar{\Omega} \times [0,T] \to \mathbb{R}$ is called a solution to the equation if it satisfies the corresponding PDE, subject to some initial and boundary condition which we will explain in detail soon. 
A differential operator $F$ refers to the mapping from $v$ to a function determined by $v$ and its gradient up to certain order. 
The differential operators $F$ in the examples above are all second-order due to the Laplacian.

We remark that the differential operator $F$ can be nonlinear, such as in the Hamilton--Jacobi equation \eqref{eq:hj-example} and the Allen--Cahn equation \eqref{eq:ac-example}.
Nonlinearity often cause significant challenge in solving these PDEs because the solutions to such PDEs may become stiff and unstable as time changes.

Another major challenge in solving PDEs numerically (there are very few PDEs that have analytic solutions and we almost always need some numerical solver to solve PDEs) is the dimensionality.
If the dimension $d$ is large (e.g., $d \ge 5$), then the classic numerical PDE solvers, such as the finite difference method, finite element method, and spectral method, which are well studied in the literature of numerical PDEs \cite{bartels2015numerical}, often become computationally infeasible due to the issue known as curse of dimensionality. This is because these methods typically require spatial discretization of the domain $\Omega \subset \mathbb{R}^{d}$ (or in a different transform domain), and the number of unknown variables in the discretized versions grows exponentially fast in $d$, making the discretized problem intractable for large $d$.

We also notice that, in the examples \eqref{eq:heat-example}, \eqref{eq:hj-example}, and \eqref{eq:ac-example} above, the differential operators are \emph{autonomous}\index{Autonomous} since they do not explicitly depend on $t$. Instead, they depend on $t$ implicitly through $v(x,t)$. However, we can easily extend the method to be introduced later to tackle PDEs with non-autonomous $F$ by using the augmented variable $\bar{x} := (t,x)$ and converting the PDEs into autonomous ones. This is straightforward and we omit the details here.

Furthermore, it is worth pointing out that we can also handle evolution PDEs with higher-order time derivatives using the method to be introduced. 
As an example, for the evolution PDE with second-order time derivative
\begin{equation}
\label{eq:dtt-example}
\partial_{tt}^{2} v(x,t) = F[v](x,t) ,
\end{equation}
we can convert \eqref{eq:dtt-example} to an equivalent PDE with only first-order time derivative:
\begin{equation*}
\begin{cases}
\partial_{t} v(x,t) = w(x,t) , \\
\partial_{t} w(x,t) = F[v](x,t) ,
\end{cases}
\end{equation*}
and the solution $(w,v): \bar{\Omega} \times [0,T] \to \mathbb{R}^{2}$ satisfying the system of two PDEs above is to be found.

For simplicity, we assume that $F$ is autonomous and $\Omega = \mathbb{R}^{d}$ below. 
We will not consider boundary condition for simplicity in this case.
Recall that evolution PDEs often require some initial condition to ensure uniqueness of solution. For instance, in \eqref{eq:heat-example} and \eqref{eq:hj-example}, we see that $v+C$ with arbitrary constant $C \in \mathbb{R}$ solves the same PDE as long as $v$ is a solution to the evolution PDE.

The initial condition is determined by some given initial value $g: \bar{\Omega} \to \mathbb{R}$. 
We let $v^{g}: \bar{\Omega} \times [0,T] \to \mathbb{R}$ denote the solution to the initial value problem 
\begin{equation}
\label{eq:so-ivp}
\begin{cases}
\partial_{t} v(x,t) = F[v](x,t) , \quad \forall \, (x,t) \in \Omega \times [0,T] , \\
v(x,0) = g(x), \quad x \in \bar{\Omega} .
\end{cases}
\end{equation}
Then, we call $S_{F}: C^{1}(\bar{\Omega};\mathbb{R}) \to C^{2,1}(\bar{\Omega} \times [0,T];\mathbb{R})$ defined by
\begin{equation}
\label{eq:so}
S_{F}(g) := v^{g}.
\end{equation}
the \emph{solution operator associated with $F$}\index{Solution operator} to the problem \eqref{eq:so-ivp}, where the range $C^{2,1}(\bar{\Omega} \times [0,T];\mathbb{R})$ is the set of real-valued functions defined on $\bar{\Omega} \times [0,T]$ that are twice continuously differentiable with respect to $x$ and continuously differentiable with respect to $t$ in the interior points.
In other words, the solution operator $S_{F}$ associated with $F$ is the mapping from any initial value $g$ to its corresponding solution $v^{g}$ of \eqref{eq:so-ivp}.
Our goal is to develop a method that can find (an approximation of) $S_{F}$ for any given $F$.

We remark that finding the solution operator $S_{F}$ is a problem substantially different from solving \eqref{eq:so-ivp} with a given and fixed $g$. 
The solution operator $S_{F}$ can produce the solution $v^{g}$ at very low computational cost for any $g$.
By contrast, solving \eqref{eq:so-ivp} means that one employs a method (such as the PINN method in Example \ref{ex:pinn-dr-wan}) with a given initial $g$, and needs to start from scratch when another initial $g$ is given, which is computationally intensive if we need the solutions for many different initials.
The benefit of having the solution operator $S_{F}$ becomes more appealing when the dimension $d$ is high, in which case solving \eqref{eq:so-ivp} with one instance of $g$ is computationally expensive.

Now we show how deep optimal control provides an approach to approximating the solution operator $S_{F}$ for any given $F$. 
To this end, notice that if we parameterize a function $v: \mathbb{R}^{d} \to \mathbb{R}$ as a deep neural network $u_{\theta}$ with parameter $\theta \in \mathbb{R}^{n}$, then we effectively turn an $n$-dimensional vector $\theta$ as the representative of the function $v$ which is an element of an infinite-dimensional function space. 
If $v: \mathbb{R}^{d} \times [0, T] \to \mathbb{R}^{d}$ evolves in time (as those solutions to evolution PDEs), then we can use a trajectory of parameters $\theta: [0,T] \to \mathbb{R}^{n}$ to approximately represent $v$. 
Namely, we use $u_{\theta(t)}(\cdot)$ as an approximation of $v(\cdot,t)$ for any $t \in [0,T]$ by steering (controlling) $\theta(t)$ properly.

At this point, we would like to hold on and ask the following question: Is a deep neural network $u_{\theta(t)}: \mathbb{R}^{d} \to \mathbb{R}$ with time-evolving parameter $\theta(t)$ equally expressive as a deep neural network $\hat{u}_{\eta}: \mathbb{R}^{d} \times [0,T] \to \mathbb{R}$ with a fixed parameter $\eta$?
By $u_{\theta(t)}$ being equally expressive as $\hat{u}_{\eta}$, we meant that $u_{\theta(t)}(\cdot)$ equals to $\hat{u}_{\eta}(\cdot,t)$ for any $t$.
If the answer is yes, then we know $u_{\theta(t)}$ is a function approximator of the same representation capability as $\hat{u}_{\eta}$, where the latter is known to have the capability of representing a large class of functions (e.g., $W^{k,\infty}$ functions on a compact subset of an Euclidean space) by the universal approximation theorem (Theorem \ref{thm:uat}). 
The following lemma gives a positive answer to this question.

\begin{lemma}
\label{lem:evolution_params}
For any differentiable feed-forward neural network architecture $\hat{u}_{\eta}: \mathbb{R}^{d} \times [0,T] \to \mathbb{R}$ defined by $(x,t) \mapsto \hat{u}_{\eta}(x,t)$ with parameter $\eta$, there exists a feed-forward neural network $u_{\theta(t)} : \mathbb{R}^{d} \to \mathbb{R}$ defined by $x \mapsto u_{\theta(t)}(x)$ with $\theta(t)$ as its parameter at time $t$, where $\theta(t)$ is differentiable in time, such that $u_{\theta(t)}(x) = \hat{u}_{\eta}(x,t)$ for all $x \in \mathbb{R}^{d}$ and $t \in \mathbb{R}$.
\end{lemma}
\begin{proof}
We denote the feed-forward network $\hat{u}_{\eta}$ as 
\begin{equation*}
\hat{u}_{\eta}(x,t)=\hat{w}^{\top}_L \hat{h}_{L-1}(x,t) + \hat{b}_L
\end{equation*}
with
\begin{equation*}
\hat{h}_{l}(x,t) = \sigma(\hat{W}_{l} \hat{h}_{l-1}(x,t)+\hat{b}_{l}) , \quad l = 1,\ \ldots,\ L-1,
\end{equation*}
where $\sigma$ is a differentiable activation function, 
\begin{equation*}
\hat{h}_0(x,t) = \sigma(\hat{W}_0x+\hat{w}_0t+\hat{b}_0) ,
\end{equation*}
and the collection of parameters is
\begin{equation*}
\eta = (\hat{w}_L,\ \hat{b}_L,\ \hat{W}_{L-1},\ \dots,\ \hat{W}_0,\ \hat{w}_0,\ \hat{b}_0) .
\end{equation*}
Note that all components of $\eta$ are constants independent of $x$ and $t$.

Similarly, we denote the network $u_{\theta(t)}$ as 
\begin{equation*}
u_{\theta(t)}(x)=w_l(t)^{\top} h_{l-1}(x,t)+ b_l(t)
\end{equation*}
with
\begin{equation*}
h_{l}(x,t) = \sigma(W_{l}(t) h_{l-1}(x,t)+ b_{l}(t)), \quad l = 1,\ \ldots,\ L-1,
\end{equation*}
where $h_0(x,t)=\sigma(W_0(t)x + b_0(t))$ and
\begin{equation*}
\theta(t) = (w_L(t),\ b_L(t),\ W_{L-1}(t),\ \dots,\ W_0(t),\ b_{0}(t)) .
\end{equation*}
Note that $\theta(t)$ is time-evolving. Hence, we can set $b_{0}(t) = \hat{w}_0 t+\hat{b}_0$ for all $t$ and all other components of $\theta(t)$ as constants identical to the corresponding ones in $\eta$, i.e.,
\begin{equation*}
w_L(t)\equiv \hat{w}_L, \quad b_L(t)\equiv \hat{b}_L, \quad \dots, \quad b_1(t) \equiv \hat{b}_1, \quad W_{0}(t) \equiv \hat{W}_{0} 
\end{equation*}
for all $t$.
Then it is clear that $u_{\theta(t)}(x) = \hat{u}_{\eta}(x,t)$ for all $x$ and $t$, and $\theta(t)$ is differentiable. 
\end{proof}
\begin{remark}
\label{rmk:static-evolving-nn}
While we state Lemma \ref{lem:evolution_params} for feed-forward neural networks, it is easy to see that the result holds for general neural networks as well. To see this, note that for any neuron $\hat{y}(x,t)$ we can rewrite it as $y_{\eta(t)}(x)$ as shown in Lemma \ref{lem:evolution_params}. For any neuron $z(t,x,\hat{y}(x,t))$ we can write it as 
\begin{equation*}
\hat{z}(t,x,\hat{y}(x,t))= \hat{z}(t,x,y_{\eta(t)}(x)) = z_{(\eta(t),\xi(t))}(x)
\end{equation*}
for some neural network $z_{\eta, \xi}$ with input $x$ and parameters $(\eta,\xi)$ evolving in $t$. Applying this repeatedly justifies the claim. 
\end{remark}

Now we see that it is safe to use a deep neural network $u_{\theta(t)}$ with fixed architecture but time-evolving parameter $\theta(t)$ to approximate the solution $v(\cdot,t)$ of an evolution PDE. 
The next question is how to control $\theta(t)$ such that $u_{\theta(t)}$ can behave like a solution to the initial value problem \eqref{eq:so-ivp}.
To meet the initial condition in \eqref{eq:so-ivp} for any $g$, we can directly set
\begin{align}
\theta(0) 
& = \argmin_{\theta_{0}} \| u_{\theta_{0}} - g \|_{L^{2}(\Omega)}^{2} \nonumber \\
& = \argmin_{\theta_{0}} \int | u_{\theta_{0}} (x) - g(x) |^{2} \, dx . \label{eq:so-init}
\end{align}
Notice that the integral in \eqref{eq:so-init} is typically approximated by Monte Carlo integration in practice, particularly in high-dimensional cases, and the minimization is generally easy to solve as we are approximating a known function $g$ using a neural network.

To satisfy the PDE in \eqref{eq:so-ivp}, we substitute the solution $v(x,t)$ by its approximation $u_{\theta(t)}(x)$ and obtain
\begin{equation}
\label{eq:nn-pde}
\partial_t \utt(x) = F[\utt](x), \quad \forall \, (x,t) \in \Omega \times [0,T] .
\end{equation}
Note that we eliminate $t$ from the parenthesis on the right-hand side of \eqref{eq:nn-pde} as we assumed $F$ to be autonomous. 
On the other hand, by the chain rule, we have
\begin{equation}
\label{eq:nn-cr}
\partial_t \utt(x) = \nabla_{\theta}u_{\theta(t)}(x) \cdot \dot{\theta}(t) .
\end{equation}
Combining \eqref{eq:nn-pde} and \eqref{eq:nn-cr}, we see that, in order for $u_{\theta(t)}$ to be a solution of the evolution PDE in \eqref{eq:so-ivp}, the parameter $\theta$ should satisfy 
\begin{equation}
\label{eq:so-theta}
\nabla_{\theta}u_{\theta(t)}(x) \cdot \dot{\theta}(t) = F[\utt](x) 
\end{equation}
for all $t$.
Therefore, $\theta(t)$ should be controlled in a way such that the difference between the two sides of \eqref{eq:so-theta} can be minimized at every $t$.

As our goal is to find the solution operator $S_{F}$ that can be applied to any initial $g$, we need a control vector field that can steer $\dot{\theta}(t)$ to minimize the difference between the two sides of \eqref{eq:so-theta} regardless of the initial $\theta(0)$ which is determined by $g$.
To this end, we shall collect a set $\{ g_{i}: \bar{\Omega} \to \mathbb{R}: i \in [M]\}$ consisting of $M$ initial functions (ideally we should choose them to be similar to those we may encounter in practical applications of $S_{F}$). 
Then we find their corresponding initials $\theta_{i,0}$ as in \eqref{eq:so-init}:
\begin{equation}
\label{eq:so-initials}
\theta_{i}(0) = \argmin_{\theta_{i,0} \in \mathbb{R}^{n}} \int | u_{\theta_{i,0}}(x) - g_{i}(x) |^{2} \, dx ,
\end{equation}
which can be solved in parallel.
We denote by $\Dcal_{\mathrm{init}} = \{\theta_{i}(0)\in \mathbb{R}^{n}: i \in [M]\}$ the set of initial conditions.
Then we try to find a control vector field $h: \mathbb{R}^{n} \to \mathbb{R}^{n}$, which is also parameterized as a deep neural network $h_{\eta}$ with parameter $\eta$, such that \eqref{eq:so-theta} always holds (approximately) starting from any initial $\theta_{i}(0)$ in $\Dcal_{\mathrm{init}}$. 
Then we set train this control vector field $h_{\eta}$ by solving the following optimal control problem:
\begin{subequations}
\label{eq:so-training}
\begin{align}
\max_{\eta} \ & \quad J(\eta) := - \frac{1}{M} \sum_{i=1}^{M} r_{i}(T) , \label{eq:so-training-obj}\\
\text{s.t.} \ & 
\begin{cases}
\dot{\theta}_{i}(t) = h_{\eta}(\theta_{i}(t)), \quad \forall \, t\in [0,T] , \ i \in [M] ,\\ 
\dot{r}_{i}(t) = \| \nabla_{\theta} u_{\theta_{i}(t)} \cdot h(\theta_{i}(t)) - F[u_{\theta_{i}(t)}] \|_{L^{2}(\Omega)}^{2} , \\
\theta_{i}(0) \in \Dcal_{\mathrm{init}}, \quad r_{i}(0) = 0 , \quad i \in [M] .
\end{cases} \label{eq:so-training-ivp}
\end{align}
\end{subequations}
In \eqref{eq:so-training-ivp}, $r_{i}(T)$ is the total squared difference between the two sides of \eqref{eq:so-theta} generated by $\theta_{i}(t)$ over $[0,T]$.
We can construct the adjoint equation of \eqref{eq:so-training} as shown in the Neural ODE method in Section \ref{sec:node} and compute $\nabla_{\eta} J(\eta)$, and apply any gradient-based optimization algorithm to train the neural network $h_{\eta}$ representing the control vector field.

The solution operator $S_{F}$ is implemented as follows once the network $h_{\eta}$ is trained.
For any new initial condition $g: \bar{\Omega} \to \mathbb{R}$, we first compute $\theta(0)$ from \eqref{eq:so-init}, which is simple and straightforward as it is fitting a neural network to a given function $g$ on $\Omega$. Then we can solve for $\theta(t)$ from the following ODE starting from $\theta(0)$:
\begin{equation*}
%\label{eq:so-testing-ivp}
\dot{\theta}(t) = h_{\eta}(\theta(t)), \quad \forall\, t \in [0,T] ,
\end{equation*}
which can be solved accurately and rapidly using any numerical ODE solver. Then the network $u_{\theta(t)}(\cdot)$ with parameter set to $\theta(t)$ is an approximation of the solution $v^{g}(\cdot, t)$ to the evolution PDE with initial $g$ in \eqref{eq:so-ivp}. 
The computational cost of finding $\theta(0)$ dominates this two-step process of implementing $S_{F}$, however, it is much lower than the cost of solving the problem \eqref{eq:so-ivp} from scratch.

In practice, there are several tricks to improve the accuracy of the control vector field $h_{\eta}$ and accelerate its training. 
For example, we can increase the quality of data by choosing the set of initial values $\{g_{i}: \bar{\Omega} \to \mathbb{R}: i \in [M]\}$ more carefully, such that they are more relevant to the initials to which the trained solution operator $S_{F}$ would be applied.
We can also solve individual $\theta_{i}(t)$ for each of the training initials in $\Dcal_{\mathrm{init}}$ and generate sample points $\theta_{i}(t_{j})$ with multiple time points $t_{j} \in (0,T]$.
These sample points can be formed as a regularization term added to the ODE of $r_{i}(t)$ in \eqref{eq:so-training-ivp}.
We refer interested readers to \cite{gaby2025approximation,gaby2024neural} for more details.

\subsection{References of Optimal Control}

Optimal control is a core discipline in applied mathematics and engineering and has been extensively studied in the past decades. There are numerous excellent books that cover the theory and computation of optimal control in the literature \cite{pontryagin2018mathematical,fleming1975deterministic,macki2012introduction,bardi1997optimal,hocking1991optimal,evans2024introduction}. 
Note that the original versions of \cite{pontryagin2018mathematical,macki2012introduction,evans2024introduction} were published decades before the years shown in the bibliography and the modern literature database.

Historically, optimal control grew out of the calculus of variations, which dates back several centuries, but it matured into a distinct field in the middle of the 20th century following advances in aerospace and computational technologies in the 1960s and 1970s. Early work connected the minimization of integral cost functionals with variations of control functions, formalizing necessary conditions for optimality that generalize classical Euler--Lagrange equations \cite{sargent2000optimal}.
This chapter does not cover stochastic optimal control, where the ODE is replaced with a stochastic differential equation (SDE) and the reward function is in the expectation form. The main ideas are similar to deterministic optimal control, but the diffusion effect and related analysis need to be taken into account carefully. We refer interested readers to \cite{fleming1975deterministic,evans2024introduction} for more details.

A foundational result in the canonical literature is the Pontryagin Maximum Principle formulated in the 1950s, which provides necessary conditions for optimal controls in terms of a Hamiltonian system and an adjoint (co-state) evolution. The significance of this principle is that it reduces infinite-dimensional optimization over control functions to a pointwise condition, simplifying analysis and implementation for many problems. 
In parallel, the Bellman dynamic programming principle introduced an alternative framework focused on the value function and recursive optimality, foundational to both continuous and discrete optimal control and later to reinforcement learning (we will cover more details in the next chapter).
The core ideas and results of these two foundational results have covered in Section \ref{sec:oc-theory}, the large part of which follows \cite{evans2024introduction} with some expansions on the proof details. 
Extensions of these two results to optimal probability density control are presented in Section \ref{sec:odc-theory}, which follows \cite{gaby2025hamiltonian}.
The Neural ODE method as a continuous-time variant of residual networks was proposed in \cite{chen2018neural}, and it turns out to be a powerful tool to solve standard optimal control and many other general control and learning problems, as we explained in Section \ref{sec:node}. We also present its application in learning PDE solution operators in Section \ref{subsec:node-so} by following \cite{gaby2025approximation,gaby2024neural} and later in probability density control method for generative modeling in Section \ref{sec:probability-density-control}.

\chapter{Deep Reinforcement Learning}
\label{chpt:rl}

Reinforcement learning addresses sequential decision-making problems in which an agent interacts with an environment by taking actions based on observed states. The objective is to design an optimal policy that maximizes the expected cumulative reward obtained from the environment over time.

This chapter focuses on two central aspects of reinforcement learning. The first concerns its mathematical foundations. We introduce the fundamental components of reinforcement learning systems and show how they are formalized within the framework of Markov decision processes. We present a collection of core theoretical results, most notably the Bellman operators and Bellman equations, that underpin reinforcement learning. These results are established with rigorous mathematical proofs.

The second aspect addresses reinforcement learning algorithms in both model-based and model-free approaches. Model-based methods are strongly grounded in the Bellman theory, and we present two canonical algorithms: policy iteration and value iteration. Recognizing the complexity of real-world reinforcement learning problems, we further discuss a series of model-free algorithms that have demonstrated impressive empirical success. During the presentations, we illustrate how deep learning techniques can be integrated to address large-scale and high-dimensional reinforcement learning problems.

\section{Fundamentals of Reinforcement Learning}
\label{sec:rl-fundamentals}

\subsection{Basic Concepts in Reinforcement Learning}

In reinforcement learning, an \emph{agent}\index{Agent} usually refers to a subject which can change state, perform actions, and receive feedbacks from the environment at specified time points. %
Example agents include robots, automobiles, drones, and game players. 
The \emph{state}\index{State} of an agent usually refers to (the combination of) its location, speed, direction and etc.
An \emph{action}\index{Action} made by the agent can be acceleration, braking, robot arm rotation, etc.
The \emph{environment}\index{Environment} can be the surrounding physical world, including human, which provides rewards to the agents according to their states and actions. The environment also determines the state transitions of the agents.
A \emph{reward}\index{Reward} is often formed as a numerical value based on which the agent can learn how to perform optimally and earn the maximal total rewards over time.
In reinforcement learning, we usually follow a discrete time setting as the rewards and measurements of state are often taken at discrete time points.

In this chapter, we use $x$ to denote the state of the agent, $u$ as the action\footnote{In the reinforcement learning literature, the state and action are often denoted by $s$ and $a$, respectively. We use $x$ and $u$ here to indicate their equivalences to the state and control in optimal control discussed in Chapter \ref{chpt:oc}.}, and $r$ as the reward. They are represented as numbers or vectors such that numerical computations can be carried out. 
We let $\Xcal$ and $\Ucal$ denote the \emph{state space} and \emph{action space}, respectively.
Note that $\Xcal$ and $\Ucal$ can be either finite sets, sets with countably many elements, or subsets of Euclidean spaces. 
We also have the following definition of stationary policy\index{Policy!stationary}, which describes the action to be made by the agent at any state $x$.
\begin{definition}
[Stationary policy]
\label{def:stationary-policy}
We call $\pi$ a \emph{stationary policy} (or just \emph{policy}) on $\Xcal$ if $\pi(\cdot | x )$ is a time-independent probability distribution on the action space $\Ucal$ at every $x \in \Xcal$, i.e.,
$\pi(u|x) \ge 0$ for any $u \in \Ucal$, and 
\begin{equation*}
\int_{\Ucal} \pi( u | x) \, du = 1 .
\end{equation*}
If $\Ucal$ is a discrete set, such as a finite set or a set of countably many actions, then the integral should be replaced with sum over $\Ucal$.
We use $\Pi$ to denote the set of stationary policies defined on $\Xcal$.
\end{definition}
In practice, there are applications where the policies are non-stationary\index{Policy!non-stationary}, namely, we have $\pi_{t}(\cdot |x)$ which also depends on time $t$. However, it is the standard setting of using stationary policies in most reinforcement learning research and applications, and therefore we will follow this setting and only consider stationary policies in this chapter.

\begin{figure}
\centering
\begin{tikzpicture}[scale=1, transform shape, node distance=1cm, >=Latex]

\node at (0,8) [draw, align=left] {\qquad \quad \qquad \qquad \qquad \qquad \qquad Environment \\ 
$\bullet$ The relevant physical world surrounding the agent. \\
$\bullet$ May include non-biological and biological (such as human) subjects. \\
$\bullet$ May provide a numerical reward value $r \in \mathbb{R}$ according to the current
\\ \ \, state $x \in \Xcal$ of and action $u \in \Ucal$ taken by the agent. \\
$\bullet$ Can move the agent to a new state $x' \in \Xcal$ according to its current 
\\ \ \, state $x \in \Xcal$ and the action $u \in \Ucal$ it takes. Here $x' \in \Ucal$ is determined 
\\ \ \, by some (possibly unknown) transition probability $p(\cdot|x,u)$ on $\Ucal$.
};

%\draw[-{Latex[length=2mm]}, line width=.6pt] (-1, 2.5) -- (-1,6) node[left, midway] 
%{At time $t$, the agent is \\ 
%at state $x_{t}$ and \\
%takes action $u_{t}$. \\
%The environment \\
%observes $(x_{t},u_{t})$.};

\draw[-{Latex[length=2mm]}, line width=.6pt] (-1,2.5) -- (-1, 6) ;

\node at (-3.1, 4.2) [align=left] {At time $t$, the agent\\ 
is at state $x_{t}\in \Xcal $ and \\
takes action $u_{t} \in \Ucal$ by \\
following policy $\pi(\cdot|x_{t})$. \\
The environment \\
observes $(x_{t},u_{t})$.};

\draw[-{Latex[length=2mm]}, line width=.6pt] (1, 6) -- (1,2.5) ;

\node at (3.3, 4.2) [align=left] {After observing $(x_{t},u_{t})$, \\ 
the environment moves \\
the agent to $x_{t+1} \in \Xcal$ \\
according to $p(\cdot|x_{t},u_{t})$ \\
and (possibly) provides a \\
reward $r_{t+1} \in \mathbb{R}$ based \\
on $(x_{t},u_{t})$.};

\node at (0,0) [draw, align=left] {\qquad \quad \qquad \qquad \qquad \qquad \qquad Agent \\ 
$\bullet$ A subject that can interact with the environment. \\
$\bullet$ Usually refers to an intelligent robot, automobile, drone, and etc. \\
$\bullet$ Can acquire its current state $x \in \Xcal$. \\
$\bullet$ Can take an action $u \in \Ucal$ based on $x$ by following a policy $\pi(\cdot|x)$. \\
$\bullet$ If $\pi(\cdot|x)$ is deterministic, then $u \in \Ucal$ is directly selected. \\
$\bullet$ If $\pi(\cdot|x)$ is stochastic (a probability on $\Ucal$), then $u \sim \pi(\cdot|x)$. \\
$\bullet$ The goal of RL is to find an optimal policy $\pi(\cdot|x)$ for every $x \in \Xcal$
\\ \ \, such that the agent can maximize the total reward received from 
\\ \ \, the environment over time.
};

\end{tikzpicture}
\caption{Structure of a standard reinforcement learning (RL) system. This system consists of an agent and an environment. The functionalities of the agent and environment are explained in their corresponding boxes. The agent can interact with the environment by taking actions following a policy $\pi$. At any time $t$, the environment can provide a reward value $r_{t+1}$ and moves the agent to a new state $x_{t+1}$ according to the current state $x_{t}$ of the agent and the action $u_{t}$ that the agent takes. The goal of RL is to find an optimal policy $\pi$ such that the agent can gain maximal total reward over time.}
\label{fig:rl}
\end{figure}
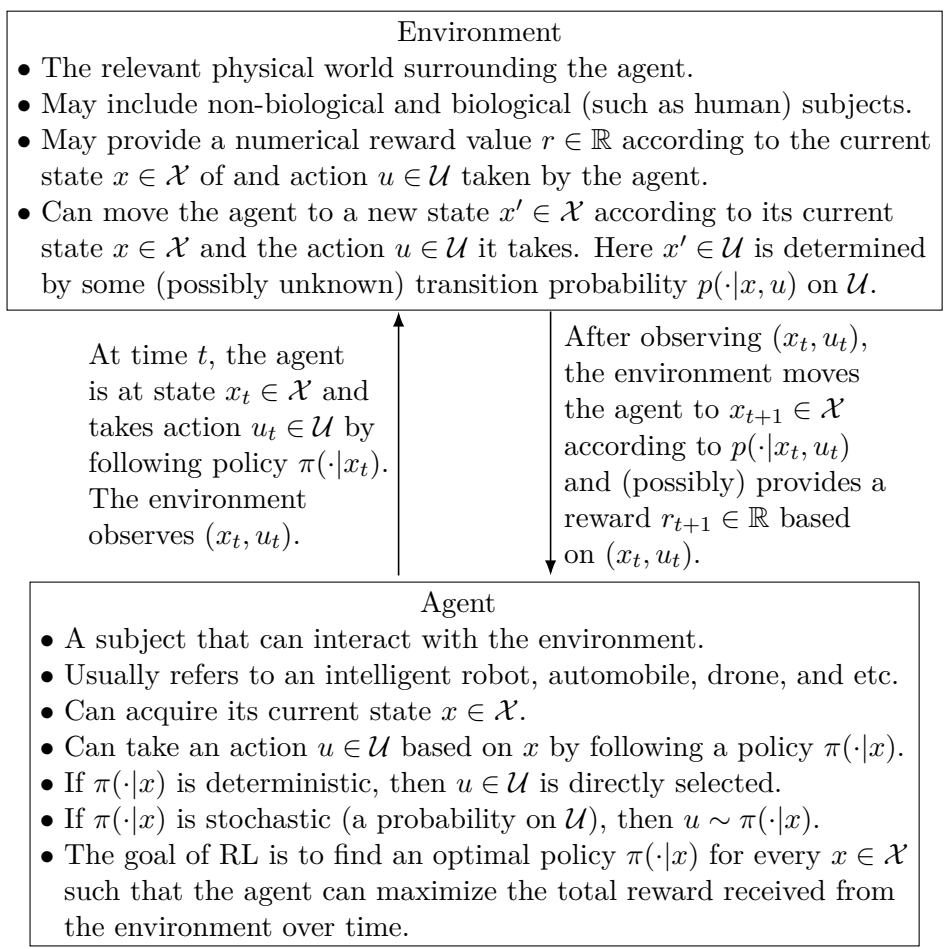

\subsection{Markov Decision Processes}

When a stationary policy $\pi$ is given, the agent can start from an initial state $X_{0}$ drawn from some initial state distribution $p_{0}$ on the state space $\Xcal$ at time $t=0$, and implement the policy $\pi$ to make action $U_{0}$. Then the environment provides a reward $R_{1}$ based on $X_{0}$ and $U_{0}$, and moves the agent to a new state $X_{1}$. This process will continue in time and eventually forms a trajectory:
\begin{equation}
\label{eq:mdp}
X_{0}, \ U_{0}, \ R_{1}, \ X_{1}, \ U_{1}, \ R_{2}, \ \cdots, \ X_{t}, \ U_{t}, \ R_{t+1}, \ X_{t+1}, \ \cdots
\end{equation}
Note that we use capital letters in \eqref{eq:mdp} since all these values are random in general (except for cases like  $p_{0}$ and $\pi$ are deterministic, i.e., they only map to specific values and do not have any randomness) in a reinforcement learning problem.

In the setting of reinforcement learning, the trajectory \eqref{eq:mdp} is assumed or set up as a \emph{Markov process}\index{Markov process}. More precisely, for any time $t$, the random variables $U_{t}, R_{t+1}, X_{t+1}, \cdots $ depend only on the value of $X_{t}$, not any values of the past random variables $X_{0},\cdots, X_{t-1}, U_{t-1}, R_{t}$.
In this case, \eqref{eq:mdp} is called a \emph{Markov decision process}\index{Markov decision process}.

From the probability perspective, the Markov decision process \eqref{eq:mdp} is determined by the following policy and transition probability distributions. For any $x \in \Xcal$, there is
\begin{equation}
\label{eq:policy-transition}
U_{t} \sim \pi(\, \cdot \,| X_{t} = x) ,
\end{equation}
where the $\pi \in \Pi$ is a stationary policy. For any $x \in \Xcal$ and $u \in \Ucal$, the next state $X_{t+1}$ follows the transition probability 
\begin{equation}
\label{eq:state-transition}
X_{t+1} \sim p( \,\cdot\, | X_{t} = x, U_{t} = u) ,
\end{equation}
Note that this transition probability $p(\,\cdot\,|x,u)$ is completely determined by the environment and is not affected by any policy.
When a policy $\pi$ is given, we also frequently use the following state-to-state transition probability
\begin{equation}
\label{eq:ss-transition}
p^{\pi}(x'|x) := \int_{\Ucal} \pi(u|x) p(x'|u,x) \, du 
\end{equation}
for any $x' \in \Xcal$ during our derivations in this chapter.

It is worth noting that, if a policy $\pi$ is given and the state-to-state transition probability $p^{\pi}$ in \eqref{eq:ss-transition} is known, then the Markov decision process \eqref{eq:mdp} reduces to a \emph{Markov reward process}\index{Markov reward process}:
\begin{equation}
\label{eq:mrp}
X_{0}, \ R_{1}, \ X_{1}, \ R_{2}, \ \cdots, \ X_{t}, \ R_{t+1}, \ X_{t+1}, \ \cdots
\end{equation}
since the actions will be taken automatically by the policy $\pi$, and the state transition is determined by $p^{\pi}$. 
A Markov reward process \eqref{eq:mrp} can be observed, but the policy is supposed to be fixed during this process.

In some reinforcement learning problems, the reward $R_{t+1}$ is also a random variable following some conditional probability $p_{\tiny{R}}(\cdot|X_{t} = x, U_{t}=u)$ (again $p_{\tiny{R}}$ is determined by the environment and not affected by any policy). 
However, it is mostly the case that only the expectation of $R_{t+1}$ is needed to solve the reinforcement learning problem, and hence for simplicity we will just set $R_{t+1} = r(X_{t}, U_{t})$, where $r(x,u): \Xcal \times \Ucal \to \mathbb{R}$ is deterministic (and given by the environment) and can be evaluated given $x$ and $u$.

\subsection{Bellman Operators and Equations}
\label{subsec:rl-bellman}

With all the concepts given above, we are ready to present the mathematical formulation of a reinforcement learning problem. To this end, we first introduce an important term called the state value function (which can be thought as an analogue of the value function in optimal control). 
In reinforcement learning, the state value function\index{Value function!state} maps every state $x \in \Xcal$ to the sum of discounted rewards received during a trajectory \eqref{eq:mdp} started from $X_{0} = x$ in expectation, as in the following definition.

\begin{definition}
[State value function]
\label{def:value-fn}
Let $\pi \in \Pi$ be a stationary policy on the state space $\Xcal$ and action space $\Ucal$, and $\gamma \in (0,1)$  be a \emph{discount factor}. Then the \emph{state value function} (or \emph{value function} for short), denoted by $v^{\pi}$, is a mapping from $\Xcal$ to $\mathbb{R}$ defined by
\begin{equation}
\label{eq:v-pi}
v^{\pi}(x) = \mathbb{E} \Big[ \sum_{t=0}^{\infty} \gamma^{t} R_{t+1} \Big| X_{0} = x \Big] ,
\end{equation}
where $\{R_{t+1}: t\ge 0\}$ are the rewards received during the trajectory of a Markov decision process \eqref{eq:mdp} which starts from $X_{0} = x$ and follows the policy distribution \eqref{eq:policy-transition} and state transition probability \eqref{eq:state-transition}.
\end{definition}

We have several remarks about the discount factor $\gamma$ introduced in Definition \ref{def:value-fn}.
This discount factor is often in $(0,1)$ because we are usually not fully certain about the future situations. We are also not sure how long a process like \eqref{eq:mdp} may last, and a discount factor $\gamma \in (0,1)$ ensures the total reward is always bounded under mild conditions (e.g., $R_{t}$ is upper bounded by some constant $M_{R}>0$ for all $t$).  
In practical applications, we use a discount factor because human and animal behaviors appear to prefer immediate rewards.
In finance, a discount factor is often considered because immediate rewards may earn more interests and profits than delayed rewards. 
Last but not least, a discount factor $\gamma \in (0,1)$ allows us to construct strong mathematical theory and numerical algorithms more easily for reinforcement learning.

It is still possible to consider reinforcement learning under the Markov decision process with $\gamma =1$. This is usually the case when there is a termination state $\bar{x} \in \Xcal$, such that the agent reaches $\bar{x}$ in finite time (under certain probability sense) almost surely and will stay at $\bar{x}$ afterwards, i.e., $p(\bar{x}|\bar{x},u) = 1$ for all $u \in \Ucal$.
Nevertheless, we will mostly consider the standard case with discount factor $\gamma \in (0,1)$ in this chapter.

Now we notice an important property of the state value function $\vpi$ defined in \eqref{eq:v-pi}: For every $x \in \Xcal$, there is
\begin{align}
\vpi (x)
& = \mathbb{E}[R_{1}|X_{0} = x] + \gamma \, \mathbb{E} \Big[ \sum_{t=1}^{\infty} \gamma^{t-1} R_{t+1} \Big| X_{0} = x \Big] \nonumber \\
& = \int \pi(u_{0}|x) r(x, u_{0}) \, du_{0} \nonumber \\
& \quad + \gamma \iint \pi(u_{0}|x) p(x_{1}|x, u_{0}) \mathbb{E} \Big[ \sum_{t=1}^{\infty} \gamma^{t-1} R_{t+1} \Big| X_{1} = x_{1} \Big] d x_{1} du_{0}  \label{eq:v-pi-self} \\
& = \int \pi(u_{0}|x) \Big[ r(x, u_{0}) + \gamma \int p(x_{1}|x, u_{0}) \vpi(x_{1})\, d x_{1} \Big] du_{0} \nonumber \\
& = \mathbb{E}_{U_{0} \sim \pi(\cdot|x)} \Big[ r(x,U_{0}) + \gamma \mathbb{E}_{X_{1}\sim p(\cdot|x, U_{0})}[\vpi (X_{1})] \Big], \nonumber 
\end{align}
where we used the policy and transition probability distributions \eqref{eq:policy-transition} and \eqref{eq:state-transition}, as well as
\begin{equation*}
\vpi(x_{1}) = \mathbb{E} \Big[ \sum_{t=1}^{\infty} \gamma^{t-1} R_{t+1} \Big| X_{1} = x_{1} \Big]
\end{equation*}
which is based on Definition \ref{def:value-fn} (as if the Markov decision process \eqref{eq:mdp} started from $x_{1}$).
The equation \eqref{eq:v-pi-self} is called the \emph{Bellman equation} of $\vpi$.

At this point, we have seen that any policy $\pi \in \Pi$ can determine a Markov decision process like \eqref{eq:mdp} as well as a state value function $\vpi: \Xcal \to \mathbb{R}$ in Definition \ref{def:value-fn}.
We also notice a property of the state value function given in \eqref{eq:v-pi-self}: $\vpi$ is equal to some operation applied to itself, as shown on the right-hand side of \eqref{eq:v-pi-self}.
This indicates the importance of this operation in reinforcement learning.
In order to describe this operation more clearly, we first consider the set of all uniformly bounded functions on $\Xcal$ as follows:
\begin{equation}
B(\Xcal) := \Big\{ v: \Xcal \to \mathbb{R} : \ \sup_{x \in \Xcal}\, |v(x)| < \infty \Big\}.
\end{equation}
% where $M_{v}$ is a number depending on $v$ but not $x$.
%
It is easy to see that $B(\Xcal)$ is a real linear space, i.e, $a_{1} v_{1} + a_{2} v_{2} \in B(\Xcal)$ for any $v_{1},v_{2} \in  B(\Xcal)$ and $a_{1},a_{2} \in \mathbb{R}$.
Moreover, we define the $\infty$-norm on $B(\Xcal)$ by $\| v\|_{\infty} := \sup_{x \in \Xcal} |v(x)|$ for any $v \in B(\Xcal)$. Then it is easy to check that $B(\Xcal)$ with this $\infty$-norm is a Banach space (see Appendix \ref{appsec:banach}).
%
% Notice that the supremum can be replaced with maximum if $\Xcal$ is a finite set, which is common in many reinforcement learning applications.
%
Now we are ready to give the formal definition of the operation on the right-hand side of \eqref{eq:v-pi-self}.

\begin{definition}
[Bellman operator associated to a given policy]
\label{def:bellman-T-pi}
Let $\pi \in \Pi$ be a stationary policy on $\Xcal$. Then the \emph{Bellman operator $\Tpi$ associated with $\pi$} is a mapping from $B(\Xcal)$ to $B(\Xcal)$ given by $\Tpi: v \mapsto \Tpi v$, where
\begin{equation}
\label{eq:bellman-T-pi}
\Tpi v (x) := \mathbb{E}_{U \sim \pi(\cdot | x)} \Big[ r(x, U) + \gamma \mathbb{E}_{X' \sim p(\cdot|x, U)}[v(X')] \Big]
\end{equation}
for every $x \in \Xcal$.
\end{definition}

Note that $\Tpi v(x)$ should be always interpreted as $(\Tpi v)(x)$\index{Bellman operator}.
Now, with the definition of Bellman operator $\Tpi$ in \eqref{eq:bellman-T-pi}, we can write the Bellman equation\index{Bellman equation} \eqref{eq:v-pi-self} of $\vpi$ in a concise form:
\begin{equation*}
\vpi = \Tpi \vpi .
\end{equation*}
This implies that $\vpi$ is a fixed point of $\Tpi$.
There are rich results about operators and their fixed points in Banach spaces. 
In particular, if an operator is $\gamma$-contractive for some $\gamma \in (0,1)$, then it has many important properties due to the fixed point theory (see Appendix \ref{appsec:banach}).
In what follows, we will show that $\Tpi$ is indeed $\gamma$-contractive.

\begin{proposition}
[$\Tpi$ is $\gamma$-contractive]
For any $\pi \in \Pi$ and $\gamma \in (0,1)$, the Bellman operator $\Tpi$ associated with $\pi$ is a $\gamma$-contraction. Namely, for any $v, \hat{v} \in B(\Xcal)$, there is 
\begin{equation}
\label{eq:T-pi-contraction}
\| \Tpi v - \Tpi \hat{v} \|_{\infty} \le \gamma \| v - \hat{v} \|_{\infty} .
\end{equation}
\end{proposition}

\begin{proof}
For any $x \in \Xcal$, there is
\begin{align*}
| \Tpi v (x) - \Tpi	\hat{v} (x) | 
& = \Big| \mathbb{E}_{U \sim \pi(\cdot | x)} \Big[ r(x, U) + \gamma \mathbb{E}_{X' \sim p(\cdot|x, U)}[v(X')] \Big] \\
& \qquad - \mathbb{E}_{U \sim \pi(\cdot | x)} \Big[ r(x, U) + \gamma \mathbb{E}_{X' \sim p(\cdot|x, U)}[\hat{v}(X')] \Big] \Big| \\
& = \Big| \gamma\, \mathbb{E}_{U \sim \pi(\cdot | x)} \Big[ \mathbb{E}_{X' \sim p(\cdot|x, U)}[v(X') - \hat{v}(X')] \Big] \Big| \\
& = \gamma\, \Big | \iint \pi(u|x) p(x' | x, u) (v(x') - \hat{v}(x'))  \, dx' du \Big|\\
& = \gamma\, \Big | \int p^{\pi}(x' | x) (v(x') - \hat{v}(x'))  \, dx' \Big|\\
& \le \gamma \int p^{\pi}(x' | x) |v(x') - \hat{v}(x')|  \, dx' \\
& \le \gamma\, \| v - \hat{v}\|_{\infty} \int p^{\pi}(x'|x) \, dx' \\
& = \gamma\, \| v - \hat{v}\|_{\infty} ,
\end{align*}
where we used the fact that expectation is linear in the second equality;
$p^{\pi}(x'|x) = \int \pi(u|x) p(x'|x,u) du$ in the fourth equality;
and $\int p^{\pi}(x'|x) \, dx' = 1$ in the last equality.
Since $x$ is arbitrary, the claim \eqref{eq:T-pi-contraction} is proved.
\end{proof}

Next, we present another Bellman operator on $\Xcal$ that is useful to derive the optimality conditions of policies in reinforcement learning. Note that this Bellman operator\index{Bellman operator} does not require any policy defined beforehand.

\begin{definition}
[Bellman operator]
\label{def:bellman-T}
The \emph{Bellman operator $\Ts$ on $\Xcal$} is a mapping from $B(\Xcal)$ to $B(\Xcal)$ given by $\Ts: v \mapsto \Ts v$, where
\begin{equation}
\label{eq:bellman-T}
\Ts v (x) := \sup_{\pi \in \Pi} \Big\{ \mathbb{E}_{U \sim \pi(\cdot | x)} \Big[ r(x, U) + \gamma \mathbb{E}_{X' \sim p(\cdot|x, U)}[v(X')] \Big] \Big\}
\end{equation}
for every $x \in \Xcal$.
\end{definition}

We remark that $\Tpi$ is a linear operator on $B(\Xcal)$, whereas $\Ts$ is nonlinear in general due to the presence of supremum in \eqref{eq:bellman-T}. Now we want to show that $\Ts$ is also $\gamma$-contractive, just like $\Tpi$. However, due to the nonlinearity, we need a small lemma before the proof.

\begin{lemma}
\label{lem:rl-sup-lemma}
Let $A$ be any set and $f, g : A \to \mathbb{R}$ be two real-valued functions defined on $A$. Then
\begin{equation}
\label{eq:rl-sup-lemma}
\Big| \sup_{a \in A} f(a) - \sup_{a \in A} g(a) \Big| \le \sup_{a \in A} | f(a) - g(a) | .
\end{equation}
\end{lemma}

\begin{proof}
First, we notice that for any two real-valued functions $f_{1}, f_{2}: A \to \mathbb{R}$, there is 
\begin{equation}
\label{eq:rl-sup-lemma-base}
\sup_{a \in A} (f_{1}(a) + f_{2}(a) ) \le \sup_{a \in A} f_{1}(a) + \sup_{a \in A} f_{2}(a) .
\end{equation}
This inequality \eqref{eq:rl-sup-lemma-base} can be easily verified: For any $a \in A$, there is 
\begin{equation*}
f_{1}(a) + f_{2}(a)  \le \sup_{a \in A} f_{1}(a) + \sup_{a \in A} f_{2}(a) ,
\end{equation*}
and then taking the supremum of the left-hand side with respect to $a$ over $A$ yields \eqref{eq:rl-sup-lemma-base}.

Now we prove \eqref{eq:rl-sup-lemma}. We see that
\begin{align*}
\sup_{a \in A} f(a)
& = \sup_{a \in A} \Big( (f(a) - g(a)) + g(a) \Big) \\
& \le \sup_{a \in A} (f(a) - g(a)) + \sup_{a \in A} g(a) \\
& \le \sup_{a \in A} |f(a) - g(a)| + \sup_{a \in A} g(a) ,
\end{align*}
where we used the fact \eqref{eq:rl-sup-lemma-base} in the first inequality, and therefore
\begin{align*}
\sup_{a \in A} f(a) - \sup_{a \in A} g(a) \le \sup_{a \in A} |f(a) - g(a)| .
\end{align*}
Similarly, there is
\begin{align*}
\sup_{a \in A} g(a) - \sup_{a \in A} f(a)  \le \sup_{a \in A} |f(a) - g(a)| .
\end{align*}
Combining the previous two inequalities, we obtain \eqref{eq:rl-sup-lemma}.
\end{proof}

Now we are ready to show that $\Ts$ defined in \eqref{eq:bellman-T} is also a $\gamma$-contraction.
\begin{proposition}
[$\Ts$ is $\gamma$-contractive]
For any $\gamma \in (0,1)$, the Bellman operator $\Ts$ is a $\gamma$-contraction. Namely, for any $v, \hat{v} \in B(\Xcal)$, there is 
\begin{equation}
\label{eq:T-contraction}
\| \Ts v - \Ts \hat{v} \|_{\infty} \le \gamma \| v - \hat{v} \|_{\infty} .
\end{equation}
\end{proposition}

\begin{proof}
For any $x \in \Xcal$, there is
\begin{align*}
| \Ts v (x) - \Ts	\hat{v} (x) | 
& = \Big| \sup_{\pi \in \Pi} \mathbb{E}_{U \sim \pi(\cdot | x)} \Big[ r(x, U) + \gamma \mathbb{E}_{X' \sim p(\cdot|x, U)}[v(X')] \Big] \\
& \qquad - \sup_{\pi \in \Pi} \mathbb{E}_{U \sim \pi(\cdot | x)} \Big[ r(x, U) + \gamma \mathbb{E}_{X' \sim p(\cdot|x, U)}[\hat{v}(X')] \Big] \Big| \\
& \le \sup_{\pi \in \Pi} \Big| \gamma \mathbb{E}_{U \sim \pi(\cdot | x)} \Big[ \mathbb{E}_{X' \sim p(\cdot|x, U)}[v(X') - \hat{v}(X')] \Big] \Big| \\
& = \gamma \sup_{\pi \in \Pi} \Big | \int p^{\pi}(x' | x) (v(x') - \hat{v}(x'))  \, dx' \Big|\\
& \le \gamma \sup_{\pi \in \Pi} \int p^{\pi}(x' | x) |v(x') - \hat{v}(x')|  \, dx' \\
& \le \gamma \| v - \hat{v}\|_{\infty} \sup_{\pi \in \Pi} \int p^{\pi}(x'|x) \, dx' \\
& = \gamma \| v - \hat{v}\|_{\infty}
\end{align*}
where we used Lemma \ref{lem:rl-sup-lemma} in the first inequality.
Since $x$ is arbitrary, the claim \eqref{eq:T-contraction} is proved.
\end{proof}

Since $\Tpi$ and $\Ts$ are $\gamma$-contractive, the Banach fixed point theorem (Theorem \ref{thm:banach-fixed-pt}) implies that each of them has a unique fixed point, and the corresponding fixed point iteration\index{Fixed point iteration} always converge to that fixed point from any initial. These are summarized in the following theorem.

\begin{theorem}
[Bellman operator has unique fixed point]
\label{thm:T-fixed-pt}
The following two statements about Bellman operators on $\Xcal$ with $\gamma \in (0,1)$ hold true:
\begin{itemize}
\item
For any stationary policy $\pi \in \Pi$ on $\Xcal$, the Bellman operator $\Tpi$ associated with $\pi$ has a unique fixed point $\vpi$. Moreover, for any $v \in B(\Xcal)$, let $\{v_{k}: k=0,1,\dots\}$ be the sequence generated by the fixed point iteration $v_{k+1} = \Tpi	v_{k}$ with $v_{0} = v$, then 
\begin{equation}
\| (\Tpi)^{k} v - \vpi \|_{\infty} = \| v_{k} - \vpi \|_{\infty} \to 0
\end{equation}
as $k \to \infty$ at a linear rate\index{Convergence!linear} of $\gamma$.

\item
The Bellman operator $\Ts$ has a unique fixed point $\vs$. Moreover, for any $v \in B(\Xcal)$, let $\{v_{k}: k=0,1,\dots\}$ be the sequence generated by the fixed point iteration $v_{k+1} = \Tpi	v_{k}$ with $v_{0} = v$, then 
\begin{equation}
\| (\Ts)^{k} v - \vs \|_{\infty} = \| v_{k} - \vs \|_{\infty} \to 0
\end{equation}
as $k \to \infty$ at a linear rate of $\gamma$.
\end{itemize}
\end{theorem}

\begin{proof}
Since $\Tpi$ and $\Ts$ are $\gamma$-contractive on $B(\Xcal)$, both the existence of a unique fixed point and the convergence of fixed point iteration immediately follow Theorem \ref{thm:banach-fixed-pt}.
\end{proof}

\begin{definition}
[Partial ordering in $B(\Xcal)$]
Let $v, \hat{v} \in B(\Xcal)$. We say
\begin{equation}
v \le \hat{v}
\end{equation}
if $v(x) \le \hat{v}(x)$ for every $x \in \Xcal$. We say $v = \hat{v}$ if $v \le \hat{v}$ and $\hat{v} \le v$.
\end{definition}

Now we have three important monotonicity properties about Bellman operators. The results will be frequently used later.

\begin{theorem}
\label{thm:rl-T-monotone}
Let $v, \hat{v} \in B(\Xcal)$ (which are not necessarily value functions of any policies in $\Pi$), the following statements hold true:
\begin{itemize}
\item[(i)]
Let $\pi \in \Pi$ be any stationary policy and $\Tpi$ be its associated Bellman operator, then
\begin{equation}
\Tpi v \le \Ts v .
\end{equation}

\item[(ii)]
If $v \le \hat{v}$, then for any stationary policy $\pi \in \Pi$, the Bellman operator $\Tpi$ associated with $\pi$ satisfies
\begin{equation}
\Tpi v \le \Tpi \hat{v} .
\end{equation}

\item[(iii)]
If $v \le \hat{v}$, then the Bellman operator $\Ts$ satisfies
\begin{equation}
\Ts v \le \Ts \hat{v} .
\end{equation}
\end{itemize}
\end{theorem}

\begin{proof}
Let $x \in \Xcal$ be arbitrary.
For (i), we notice that there is
\begin{align*}
\Tpi v(x) 
& = \mathbb{E}_{U \sim \pi(\cdot | x)} \Big[ r(x, U) + \gamma \mathbb{E}_{X' \sim p(\cdot|x, U)}[v(X')] \Big] \\
& \le \sup_{\pi \in \Pi} \mathbb{E}_{U \sim \pi(\cdot | x)} \Big[ r(x, U) + \gamma \mathbb{E}_{X' \sim p(\cdot|x, U)}[v(X')] \Big] \\ 
& = \Ts v(x) .
\end{align*}

For (ii), we have 
\begin{align*}
\Tpi v(x) 
& = \mathbb{E}_{U \sim \pi(\cdot | x)} \Big[ r(x, U) + \gamma \mathbb{E}_{X' \sim p(\cdot|x, U)}[v(X')] \Big] \\
& \le \mathbb{E}_{U \sim \pi(\cdot | x)} \Big[ r(x, U) + \gamma \mathbb{E}_{X' \sim p(\cdot|x, U)}[\hat{v}(X')] \Big] \\
& = \Tpi \hat{v}(x)
\end{align*}
where we used $v(x) \le \hat{v}(x)$ for every $x \in \Xcal$ to obtain the inequality.

Similarly, for (iii), we have 
\begin{align*}
\Ts v(x) 
& = \sup_{\pi \in \Pi} \mathbb{E}_{U \sim \pi(\cdot | x)} \Big[ r(x, U) + \gamma \mathbb{E}_{X' \sim p(\cdot|x, U)}[v(X')] \Big] \\
& \le \sup_{\pi \in \Pi} \mathbb{E}_{U \sim \pi(\cdot | x)} \Big[ r(x, U) + \gamma \mathbb{E}_{X' \sim p(\cdot|x, U)}[\hat{v}(X')] \Big] \\
& = \Ts \hat{v}(x) ,
\end{align*}
again because $v(x) \le \hat{v}(x)$ for every $x \in \Xcal$.

Since $x \in \Xcal$ above is chosen arbitrarily, we can claim that all the three statements hold true.
\end{proof}

Since the goal of reinforcement learning is to find the optimal policy (we have not defined it mathematically yet; it will be given soon below, in Definition \ref{def:rl-optimal-pi}), it is intuitive to consider those policies that are ``greedy'' in some sense. The following definition provides a concrete definition of greedy policies given a function in $B(\Xcal)$.

\begin{definition}
[Greedy policy]
\label{def:rl-greedy-pi}
For any $v \in B(\Xcal)$, $\pi$ is called \emph{greedy with respect to $v$} if
\begin{equation}
\label{eq:greedy-pi}
\Tpi v = \Ts v ,
\end{equation}
where $\Tpi$ is the Bellman operator associated with $\pi$. Note that $v \in B(\Xcal)$ may have more than one greedy policies\index{Policy!greedy}.
\end{definition}

An important problem is that, for any $v \in B(\Xcal)$, does there always exist a stationary policy $\pi$ that is greedy with respect to $v$? 
Due to the Markov property of stationary policies, the answer to this question is yes if and only if for every $x \in \Xcal$ there exists $u(x) \in \Ucal$ (does not need to be unique) satisfying
\begin{align}
\label{eq:greedy-exist}
r(x, u(x)) + \gamma & \mathbb{E}_{X' \sim p(\cdot|x, u(x))}[\hat{v}(X')]  \\
& \quad = \sup_{\pi \in \Pi } \mathbb{E}_{U \sim \pi(\cdot | x)} \Big[ r(x, U) + \gamma \mathbb{E}_{X' \sim p(\cdot|x, U)}[\hat{v}(X')] \Big] . \nonumber
\end{align}
In this case, we can set up a policy $\pi$ such that it maps $x$ to $u(x)$ at every $x$ (if there are more than one $u(x)$ satisfying \eqref{eq:greedy-exist}, we can choose any of them). Then this $\pi$ is greedy with respect to $v$ by Definition \ref{def:rl-greedy-pi}. 

We can see that there exists such $u(x)$ for \eqref{eq:greedy-exist} at every $x \in \Xcal$ if and only if the supremum can be attained in the action space $\Ucal$. 
It is easy to see that this happens when (i) $\Ucal$ is a finite set; or (ii) $\Ucal \subset \mathbb{R}^{m}$ is compact and the function
\begin{equation}
\label{eq:rl-max-exist-U}
r(x, \cdot) + \gamma \int p(x'|x,\cdot) v(x') \, dx' \ : \ \Ucal \to \mathbb{R}
\end{equation}
is continuous (because a continuous function defined on a compact domain in $\mathbb{R}^{m}$ has its maximizer(s) in the domain).
Both (i) and (ii) are sufficient conditions but not necessary. In practical settings, we often have a policy greedy with respect to any specified $v \in B(\Xcal)$.

\begin{theorem}
Let $\vs$ be the fixed point of $\Ts$. If $\pi$ is greedy with respect to $\vs$, then $\vpi = \vs$.
\end{theorem}

\begin{proof}
Since $\pi$ is greedy with respect to $\vs$, we have by Definition \ref{def:rl-greedy-pi} that
\begin{equation}
\Tpi \vs = \Ts \vs .
\end{equation}
On the other hand, we have $\Ts \vs = \vs$ since $\vs$ is the fixed point of $\Ts$. 
Therefore, we know $\Tpi \vs = \vs$, which means that $\vs$ is also a fixed point of $\Tpi$.
Since $\vpi$ is the unique fixed point of $\Tpi$, we know $\vpi = \vs$.
\end{proof}

We have seen a function of state $x$ and action $u$ which evaluates the expected total sum of rewards provided a state value function $v$, such as in \eqref{eq:greedy-exist}. This function is also useful the algorithmic developments in reinforcement learning. We give a formal definition of this function next.

\begin{definition}
[Q-function]
For any $v \in B(\Xcal)$, we define the \emph{quality function}, or simply the \emph{Q-function}\index{Q-function} (also called the \emph{state-action value function}\index{Value function!state-action}), associated with $v$ by
\begin{equation}
q(x,u) := r(x,u) + \gamma \mathbb{E}_{X' \sim p(\cdot|x,u)} [ v(X') ]
\end{equation}
for every $x \in \Xcal$ and $u \in \Ucal$. Notice that $q: \Xcal \times \Ucal \to \mathbb{R}$. 

In particular, we denote the Q-function associated with $\vpi$ (the unique fixed point of $\Tpi$) by
\begin{equation}
\label{eq:rl-q-pi}
q^{\pi}(x,u) := r(x,u)+ \gamma \mathbb{E}_{X' \sim p(\cdot|x,u)} [ \vpi(X') ],
\end{equation}
and the Q-function associated with $\vs$ (the unique fixed point of $\Ts$) by
\begin{equation}
\label{eq:rl-q-star}
q^{*}(x,u) := r(x,u)+ \gamma \mathbb{E}_{X' \sim p(\cdot|x,u)} [ \vs(X') ]
\end{equation}
for every $x \in \Xcal$ and $u \in \Ucal$. 
\end{definition}

We also notice that, conversely, $\vpi$ and $\vs$ can be represented by their Q-functions $q^{\pi}$ and $q^{*}$, respectively.  Namely, 
\begin{equation*}
\vpi(x) = \mathbb{E}_{U \sim \pi(\cdot|x)} [ q^{\pi}(x, U) ]
\end{equation*}
and
\begin{equation}
\label{eq:rl-vs-qs}
\vs(x) = \sup_{\pi \in \Pi} \mathbb{E}_{U \sim \pi(\cdot|x)} [ q^{*}(x,U) ] 
\end{equation}
for every $x \in \Xcal$. If the supremum in \eqref{eq:rl-vs-qs} can be attained in $\Pi$ for every $x \in \Xcal$ (for example, $\Ucal$ is a finite set), then \eqref{eq:rl-vs-qs} reduces to
\begin{equation*}
\vs(x) = \max_{u \in \Ucal} q^{*}(x,u) .
\end{equation*}

\subsection{Policy Optimality and Improvement Theory}

Now we are ready to define formally the meaning of optimal policies in reinforcement learning.

\begin{definition}
[Optimal policy]
\label{def:rl-optimal-pi}
We call $\pis \in \Pi$ an \emph{optimal policy}\index{Policy!optimal} if 
\begin{equation}
\label{eq:rl-optimal-pi}
\mathbb{E}_{U \sim \pis(\cdot|x)} [ \qs(x,U) ] = \vs(x)
\end{equation}
at every $x \in \Xcal$, where $\vs$ is the unique fixed point of $T^{*}$ defined in \eqref{eq:bellman-T} and $q^{*}$ is defined in \eqref{eq:rl-q-star}.
\end{definition}
It is clear that, if $\pis$ is greedy with respect to $\vs$, then $\pis$ is an optimal policy, because \eqref{eq:rl-optimal-pi} means that $T^{\pis} \vs = \Ts \vs = \vs$. We will formalize this as one of the results in a theorem below.

We also define a special function $\bar{v} \in B(\Xcal)$ as follows: for every $x \in \Xcal$, 
\begin{equation}
\label{eq:rl-v-bar}
\bar{v}(x) := \sup_{\pi \in \Pi} \vpi(x) .
\end{equation}
Note that, we do not know whether $\bar{v}$ is the value function of any policy $\pi \in \Pi$. However, we know by the definition \eqref{eq:rl-v-bar} that
\begin{equation}
\label{eq:v-pi-v-bar}
\vpi \le \bar{v}
\end{equation}
for any $\pi \in \Pi$. 
We can also define the Q-function associated with $\bar{v}$ by
\begin{equation}
\bar{q}(x,u) := r(x,u) + \gamma \mathbb{E}_{X' \sim p(\cdot|x,u)} [ \bar{v}(X') ] .
\end{equation}
Due to \eqref{eq:rl-v-bar}, we see that
\begin{equation}
\label{eq:rl-q-bar}
\bar{q}(x,u) = \sup_{\pi \in \Pi} q^{\pi}(x,u) .
\end{equation}
Moreover, we have the following important lemma about $\bar{v}$ and its relation to $\vs$.

\begin{lemma}
\label{lem:rl-v-bar-v-star}
If there exists a stationary policy $\bar{\pi} \in \Pi$ greedy with respect to $\bar{v}$, then 
\begin{equation}
\label{eq:v-bar-v-star}
\bar{v} = v^{\bar{\pi}} \le \vs ,
\end{equation}
where $v^{\bar{\pi}}$ is the unique fixed point of $T^{\bar{\pi}}$ associated with $\bar{\pi}$, and $\vs$ is the unique fixed point of $\Ts$.
\end{lemma}

\begin{proof}
For any $\hat{\pi} \in \Pi$, we have its associated Bellman operator $T^{\hat{\pi}}$ and the unique fixed point $v^{\hat{\pi}}$ such that $T^{\hat{\pi}} v^{\hat{\pi}} = v^{\hat{\pi}}$. 

Now for any $x \in \Xcal$, we have
\begin{align*}
v^{\hat{\pi}}(x)
& = T^{\hat{\pi}} v^{\hat{\pi}}(x) \\
& = \mathbb{E}_{U \sim \hat{\pi}(\cdot|x)} \Big[ r(x,U) + \gamma \mathbb{E}_{X' \sim p(\cdot|x,U)} [ v^{\hat{\pi}}(X') ] \Big] \\
& \le \mathbb{E}_{U \sim \hat{\pi}(\cdot|x)} \Big[ r(x,U) + \gamma \mathbb{E}_{X' \sim p(\cdot|x,U)} [ \bar{v}(X') ] \Big] \\
& \le \sup_{\pi \in \Pi}\mathbb{E}_{U \sim \pi(\cdot|x)} \Big[ r(x,U) + \gamma \mathbb{E}_{X' \sim p(\cdot|x,U)} [ \bar{v}(X') ] \Big] \\
& = \Ts \bar{v}(x) \\
& = T^{\bar{\pi}} \bar{v}(x) ,
\end{align*}
where the second equality is due to Definition \ref{def:bellman-T-pi} with the policy $\hat{\pi}$; 
the first inequality is due to \eqref{eq:v-pi-v-bar}; 
the second last equality is due to the definition of $\Ts$; 
and the last equality is because that $\bar{\pi}$ is greedy with respect to $\bar{v}$ and hence $T^{\bar{\pi}} \bar{v} = \Ts \bar{v}$.

Since $\hat{\pi} \in \Pi$ and $x \in \Xcal$ above are arbitrary, we know 
\begin{equation}
\label{eq:rl-lemma-v-bar-T-pi-bar}
\bar{v} \le T^{\bar{\pi}} \bar{v} ,
\end{equation}
because $\bar{v}(x) = \sup_{\pi \in \Pi} \vpi(x)$ as defined in \eqref{eq:v-pi-v-bar}.
Applying $T^{\bar{\pi}}$ to both sides of \eqref{eq:rl-lemma-v-bar-T-pi-bar} and using Theorem \ref{thm:rl-T-monotone}(ii), we obtain
\begin{equation}
\label{eq:rl-lemma-v-bar-chain1}
T^{\bar{\pi}} \bar{v} \le (T^{\bar{\pi}})^{2} \bar{v} .
\end{equation}
Combining \eqref{eq:rl-lemma-v-bar-T-pi-bar} and \eqref{eq:rl-lemma-v-bar-chain1}, we obtain $\bar{v} \le (T^{\bar{\pi}})^{2} \bar{v}$. 
Continuing applying $T^{\bar{\pi}}$ to both sides of \eqref{eq:rl-lemma-v-bar-chain1}, and combining with \eqref{eq:rl-lemma-v-bar-T-pi-bar}, we obtain 
\begin{equation*}
\bar{v} \le T^{\bar{\pi}} \bar{v}  \le \dots \le (T^{\bar{\pi}})^{k} \bar{v}
\end{equation*}
for any $k$.
By letting $k \to \infty$, and recalling that $(T^{\bar{\pi}})^{k} \bar{v} \to v^{\bar{\pi}}$ in the $\infty$-norm, we obtain 
\begin{equation}
\bar{v} \le v^{\bar{\pi}} .
\end{equation}
Since $\bar{\pi} \in \Pi$, we know from \eqref{eq:v-pi-v-bar} that $v^{\bar{\pi}} \le \bar{v}$.
Therefore, $v^{\bar{\pi}} = \bar{v}$.

Now we show $v^{\bar{\pi}} \le \vs$. Notice that
\begin{equation}
\label{eq:rl-lemma-T-star-chain0}
v^{\bar{\pi}} = T^{\bar{\pi}} v^{\bar{\pi}} \le \Ts v^{\bar{\pi}},
\end{equation}
where the inequality is due to Theorem \ref{thm:rl-T-monotone}(i). 
Applying $\Ts$ on both sides, we obtain 
\begin{equation}
\label{eq:rl-lemma-T-star-chain1}
\Ts v^{\bar{\pi}} \le (\Ts)^{2} v^{\bar{\pi}} .
\end{equation}
%
% Combining the previous two orderings, we know $v^{\bar{\pi}} \le (\Ts)^{2} v^{\bar{\pi}}$.
%
Continuing applying $T^{*}$ to both sides of \eqref{eq:rl-lemma-T-star-chain1}, and using Theorem \ref{thm:rl-T-monotone}(iii),  we obtain 
\begin{equation*}
v^{\bar{\pi}} \le \Ts v^{\bar{\pi}} \le \dots \le (\Ts)^{k} v^{\bar{\pi}}
\end{equation*}
for any $k$, where the first inequality is due to \eqref{eq:rl-lemma-T-star-chain0}. 
By letting $k \to \infty$, and recalling that $(\Ts)^{k} \bar{v} \to \vs$ in the $\infty$-norm, we obtain 
\begin{equation}
v^{\bar{\pi}} \le \vs .
\end{equation}
In conclusion, we have $\bar{v} = v^{\bar{\pi}} \le \vs$.
\end{proof}

\index{Policy!Improvement Theorem}
\begin{theorem}
[Policy improvement theorem]
\label{thm:policy-improvement}
The following statements hold true:
\begin{itemize}
\item[(i)]
For any $\pi_{0} \in \Pi$, let $\pi$ be greedy with respect to $v^{\pi_{0}}$ (which is the unique fixed point of the Bellman operator $T^{\pi_{0}}$ associated with $\pi_{0}$), namely, $\Tpi v^{\pi_{0}} = \Ts v^{\pi_{0}}$, then
\begin{equation}
v^{\pi_{0}} \le \vpi .
\end{equation}
We call $\pi$ an \emph{improvement upon $\pi_{0}$}.

\item[(ii)]
If, in addition, $v^{\pi_{0}}(x) < \Ts v^{\pi_{0}}(x)$ for some $x \in \Xcal$, then $v^{\pi_{0}}(x) < \vpi(x)$. In this case, we say $\pi$ \emph{strictly improves $\pi_{0}$ at $x$}.

\item[(iii)]
If $\Ts v^{\pis} = v^{\pis}$, then $\pis$ is an optimal policy.
\end{itemize}
\end{theorem}

\begin{proof}
For (i), we have 
\begin{equation*}
v^{\pi_{0}} = T^{\pi_{0}} v^{\pi_{0}} \le \Ts v^{\pi_{0}} = \Tpi v^{\pi_{0}},
\end{equation*}
where the inequality is due to Theorem \ref{thm:rl-T-monotone}(i), and the last equality is because $\pi$ is greedy with respect to $v^{\pi_{0}}$.
Continuing applying $\Tpi$ on both sides and employing Theorem \ref{thm:rl-T-monotone}(ii), we obtain
\begin{equation*}
v^{\pi_{0}} \le \Tpi v^{\pi_{0}} \le \dots \le (\Tpi)^{k} v^{\pi_{0}}
\end{equation*}
for all $k$. Letting $k \to \infty$ and noticing that $(\Tpi)^{k} v^{\pi_{0}} \to \vpi$ in the $\infty$-norm, we obtain $v^{\pi_{0}} \le \vpi$.

For (ii), we follow the same deductions as in (i) and obtain
\begin{equation*}
v^{\pi_{0}}(x) < \Ts v^{\pi_{0}}(x) \le (\Tpi)^{k} v^{\pi_{0}}(x)
\end{equation*}
for all $k$ at this $x$. Taking $k \to \infty$, we get $ v^{\pi_{0}}(x) < \Ts v^{\pi_{0}}(x) \le \vpi(x) $.

For (iii), since $\Ts v^{\pis} = v^{\pis}$, we know $v^{\pis} = \vs$, the unique fixed point of $\Ts$. Noticing that $v^{\pis}$ is also the fixed point of $T^{\pis}$, we have $T^{\pis} \vs = \vs$, namely
\begin{equation*}
\mathbb{E}_{U \sim \pis(\cdot|x)} [\qs(x,U)] = \vs(x)
\end{equation*}
for every $x \in \Xcal$. By Definition \ref{def:rl-optimal-pi}, we know $\pis$ is an optimal policy.
\end{proof}

Now we have a more clear picture of what an optimal policy may look like. We connect the optimality condition of a policy, its relation to $\bar{v}$, and the greediness property in the following theorem.

\begin{theorem}
\label{thm:rl-equiv}
The following statements are equivalent:
\begin{itemize}
\item[(i)]
$\pis$ is an optimal policy, namely, $\mathbb{E}_{U \sim \pis(\cdot|x)} [\qs(x,U)] = \vs(x)$. 

\item[(ii)]
$\vs = v^{\pis} = \bar{v}$.

\item[(iii)]
$\pis$ is greedy with respect to $\vs$.
\end{itemize}
\end{theorem}

\begin{proof}
If (i) holds true, we have $T^{\pis} \vs = \vs$, which means that $\vs$ is the unique fixed point of $T^{\pis}$. Therefore $\vs = v^{\pis}$. Notice that $v^{\pis} \le \bar{v}$ by \eqref{eq:v-pi-v-bar}. Combining Lemma \ref{lem:rl-v-bar-v-star} which shows $\bar{v} \le \vs$, we obtain (ii).

If (ii) holds true, then we have
\begin{equation*}
T^{\pis} \vs = T^{\pis} v^{\pis} = v^{\pis} = \vs = \Ts \vs,
\end{equation*}
where the first and third equalities are due to (ii), the second equality is because $\pis$ is the fixed point of $T^{\pis}$, and the last equality is because $\vs$ is the fixed point of $\Ts$.

If (iii) holds true, then
\begin{equation*}
T^{\pis} \vs = \Ts \vs = \vs ,
\end{equation*}
where the first equality is because that $\pis$ is greedy with respect to $\vs$. This shows that $\pis$ is an optimal policy according to Definition \ref{def:rl-optimal-pi}.
\end{proof}

\begin{corollary}
If $\pis$ is an optimal policy\index{Policy!optimal}, then $\pis$ is greedy with respect to $\bar{v}$\index{Policy!greedy}.
\end{corollary}

\begin{proof}
If $\pis$ is an optimal policy, we know from Theorem \ref{thm:rl-equiv} that $v^{\pis} = \bar{v} = \vs$. 
Therefore,
\begin{equation*}
T^{\pis} \bar{v} = T^{\pis} v^{\pis} = v^{\pis} = \vs = \Ts \vs = \Ts \bar{v},
\end{equation*}
which implies that $\pis$ is greedy with respect to $\bar{v}$.
\end{proof}

\paragraph{Summary}
Now we take a brief summary of the theoretical results developed in this section.
Under the standard setting with infinite time horizon and a discount factor $\gamma \in (0,1)$, we know that every stationary policy $\pi \in \Pi$ yields a Bellman operator $T^{\pi}: B(\Xcal) \to B(\Xcal)$, which has a unique fixed point $v^{\pi}$. Here $v^{\pi}: \Xcal \to \mathbb{R}$ is the state value function of $\pi$, meaning that $v^{\pi}(x)$ is the total future reward in expectation since the agent is at the state $x \in \Xcal$. The Q-function (state-action value function) $q^{\pi}(x,u)$ is the total future reward in expectation since a specific action $u \in \Ucal$ is taken by the agent when it is at the state $x \in \Xcal$.

% We define an ``optimal'' Bellman operator $T^{*}: B(\Xcal) \to B(\Xcal)$, which has a unique fixed point $\vs$ and consequently a Q-function $q^{*}$. However, we do not know if there is any policy in $\Pi$ corresponding to $\vs$ just yet.

We have a conceptually ``pointwise optimal'' value function $\bar{v}: \Xcal \to \mathbb{R}$ that is defined to have the supremum value over all stationary policies in $\Pi$ at every $x \in \Xcal$. However, $\bar{v}$ itself does not suggest whether it has a corresponding policy in $\Pi$, and consequently it is unknown whether it corresponds to any Bellman operator.

Meanwhile, we also have a conceptually ``optimal'' Bellman operator $T^{*}: B(\Xcal) \to B(\Xcal)$. It has a unique fixed point $\vs \in B(\Xcal)$ and consequently a Q-function $q^{*}$, but it is not always true that $\vs$ is the state value function of some policy in $\Pi$. 
However, under the mild condition that for every $x \in \Xcal$, a maximizer of the function
\begin{equation*}
r(x,\cdot) + \gamma \int p(x'|x, \cdot) \vs(x') \, dx' \quad : \quad \Ucal \to \mathbb{R} ,
\end{equation*}
can be attained in $\Ucal$, then $\vs$ has a greedy policy $\pi^{*} \in \Pi$.
(We mentioned that if $\Ucal$ is finite, or $\Ucal$ is compact and the real-valued function on $\Ucal$ defined above is continuous, then a maximizer can be attained in $\Ucal$.) 
More importantly, it turns out that the Bellman operator of $\pi^{*}$ is $T^{*}$, whose unique fixed point is $\vs$, and $\bar{v}$ is just $\vs$.
This implies that $\pi^{*}$ is an optimal policy by definition and $\vs$ is the optimal state value function.

\section{Reinforcement Learning Algorithms}
\label{sec:rl-alg}

Reinforcement learning algorithms can be categorized in many different ways. 
This is because the applications domains and problem settings where reinforcement learning algorithms are employed are very diverse.
In this section, we present reinforcement learning algorithms into two classes following one of the most typical categorizations in this community: model-based method and model-free method. Model-based method in general consider problems where we have (approximately) accurate model to describe the the reward and state transition probability determined by the environment, whereas in model-free method we do not have such knowledge and need to use data generated by the agent (through its interactions with the environment) to solve the problem. 
We will provide details about these two classes of algorithms in the two subsections below.

\subsection{Model-based Reinforcement Learning Algorithms}
\label{subsec:rl-model-based}

Model-based reinforcement learning algorithms refer to those applied to problems where we have the following information:
\begin{itemize}
\item
The reward function $r(x,u)$, or the conditional reward probability distribution $p_{\tiny{R}}(\cdot| x,u)$ defined on $\mathbb{R}$ such that we can compute the expected value of reward
\begin{equation*}
\mathbb{E}_{R \sim p_{\tiny{R}}(\cdot|x,u)} [ R ] = \int r p_{\tiny{R}}(r|x,u) \, dr,
\end{equation*}
for every state-action pair $(x,u) \in \Xcal \times \Ucal$. 
For ease of discussion, we assume the function $r$ is known when we discuss model-based algorithms below.

\item 
The state transition probability 
\begin{equation*}
p(\,\cdot \, | \, x,u)
\end{equation*}
on $\Xcal$, which is the probability of the next state given that the current state is $x$ and an action $u$ is performed, for every state-action pair $(x,u) \in \Xcal \times \Ucal$. 
\end{itemize}

Notice that, the information above is determined by the environment in the problem setting. Specifically, we know exactly how much reward can be received and what is the next state (or the probability distribution of the next state) if the current state is $x$ and an action $u$ is taken. This is the reason we call such reinforcement learning problems \emph{model-based}.

When we work on a model-based reinforcement learning problem, we often can directly solve for the optimal policy in the reinforcement learning problem, and do not need the agent to interact with the environment and collect necessary data to estimate these quantities.
In many of such cases, we can use the idea of dynamic programming or alike to find the optimal policy. In what follows, we provide details about some commonly used model-based reinforcement learning algorithms. These algorithms heavily depend on the properties of Markov decision processes and Bellman operators we discussed in Section \ref{sec:rl-fundamentals}.

\subsubsection*{Policy Iteration}
Policy iteration is an iterative scheme that generates a sequence of continuously improving policies for model-based reinforcement learning. The sequence of policies can be proved to converge to the optimal policy under certain conditions (to be specified later). The theory behind the policy iteration algorithm is Policy Improvement Theorem (Theorem \ref{thm:policy-improvement}).

The typical procedure of a policy iteration algorithm is as follows. We choose an arbitrary initial guess $v_{0} \in B(\Xcal)$ (not necessarily a value function) and find the policy $\pi_{0}$ that is greedy with respect to $v_{0}$. 
Now at the $k$th iteration with any $k\ge 0$, our first step is to find the value function $v^{\pi_{k}}$, called the \emph{policy evaluation}\index{Policy!evaluation} step as we try to find the value function $v^{\pi_{k}}$ corresponding to the current policy $\pi_{k}$; and then in the second step, which is called \emph{policy improvement}\index{Policy!improvement}, we find a policy $\pi_{k+1}$ which is greedy with respect to $v^{\pi_{k}}$, and can prove that $\pi_{k+1}$ is an improvement over $\pi_{k}$. 
Then we increase the iteration counter $k$ by 1 and continue to the next iteration.
We summarize the policy iteration\index{Policy iteration} algorithm in Algorithm \ref{alg:policy-iteration}.
\begin{algorithm}
\caption{Policy Iteration for Model-based Reinforcement Learning}
\label{alg:policy-iteration}
\begin{algorithmic}[1]
\REQUIRE $v_{0} \in B(\Xcal)$ and $\pi_{0} \in \Pi$ greedy with respect to $v_{0}$, $k=0$.
\REPEAT
	\STATE (Policy evaluation) Find $v^{\pi_{k}}$ such that $T^{\pi_{k}} v^{\pi_{k}} = v^{\pi_{k}}$. 
	\STATE (Policy improvement) Find $\pi_{k+1} \in \Pi$ greedy with respect to $v^{\pi_{k}}$.
%	\STATE $\pi_{k} \leftarrow \pi_{k+1}$.
	\STATE $k \leftarrow k+1$
\UNTIL Convergent.
\ENSURE $\pi_{k}$.
\end{algorithmic}
\end{algorithm}

When we execute the policy iteration algorithm (Algorithm \ref{alg:policy-iteration}), we obtain a sequence of policies \
\begin{equation}
\label{eq:policy-iteration-iters}
\pi_{0}, \ \pi_{1}, \ \dots, \ \pi_{k}, \ \dots .
\end{equation}
By the definition of greedy policies (Definition \ref{def:rl-greedy-pi}), we know that the policy improvement step results in $T^{\pi_{k+1}} v^{\pi_{k}} = \Ts v^{\pi_{k}}$.
Furthermore, by Policy Improvement Theorem (Theorem \ref{thm:policy-improvement}(i)), we have
\begin{equation*}
v^{\pi_{k}} \le v^{\pi_{k+1}}
\end{equation*}
for every $k \ge 0$.

If both the state space $\Xcal$ and the action space $\Ucal$ in are finite sets, then we know there are only finitely many greedy policies, and the iterations \eqref{eq:policy-iteration-iters} must stop in finitely many iterations with $\pi_{k} = \pi_{k+1}$ for some $k$.
In this case, we have
\begin{equation*}
v^{\pi_{k}} = T^{\pi_{k}} v^{\pi_{k}} = T^{\pi_{k+1}} v^{\pi_{k}} = \Ts v^{\pi_{k}} 
\end{equation*}
where the first equality is because that $v^{\pi_{k}} $ is the fixed point of $T^{\pi_{k}}$, and the second equality is due to $\pi_{k} = \pi_{k+1}$. Therefore, we have $v^{\pi_{k}} = \vs$ since $\vs$ is the unique minimizer of $\Ts$. From Theorem \ref{thm:rl-equiv}, we know $\pi_{k}$ is an optimal policy.

Now we need to consider the practical implementation of the policy iteration algorithm (Algorithm \ref{alg:policy-iteration}). More specifically, we need to find ways to solve the policy evaluation (Line 2) and policy improvement (Line 3) subproblems. Depending on the problem size, we need different approaches to solve these subproblems, which we detail below.

\paragraph{Small problems}
By small problems, we mean the state space $\Xcal$ and the action space $\Ucal$ are small finite sets (with $10^{3}$ elements or alike). In this case, for any $\pi$, we can compute the state reward
\begin{equation}
\label{eq:pe-r-pi}
r^{\pi}(x) := \mathbb{E}_{U \sim \pi(\cdot | x) }[ r(x,U) ]
\end{equation}
and the state-to-state transition probability
\begin{equation}
\label{eq:pe-ss-transition}
p^{\pi}(x'|x) := \sum_{u \in \Ucal} \pi(u|x) p(x'|x,u) 
\end{equation}
for all $x, x' \in \Xcal$.
Let $n= |\Xcal|$ be the number of states in $\Xcal$. Then both $\vpi$ and $r^{\pi}$ on $\Xcal$ can be represented as $n$-dimensional column vectors, and the transition probabilities can be formed as an $n$-by-$n$ matrix $P^{\pi} = [p^{\pi}(x'|x)]_{x,x' \in \Xcal}$.
Therefore, the Bellman equation of $\vpi$
\begin{equation*}
\vpi(x) = \Tpi \vpi (x) = \mathbb{E}_{U \sim \pi(\cdot|x)} \Big[ r(x,U) + \gamma \mathbb{E}_{X' \in p(\cdot|x,U)}[\vpi(X')] \Big] 
\end{equation*}
over all $x \in \Xcal$ can be written as a linear system
\begin{equation}
\label{eq:pe-small-linear}
\vpi = I_{n} + \gamma P^{\pi} \vpi ,
\end{equation}
where $I_{n}$ is the $n$-by-$n$ identity matrix.

By Gershgorin circle theorem, we know that all the eigenvalues of $P^{\pi}$ are in the following subset of $\mathbb{R}$:
\begin{equation}
\label{eq:rl-pi-gcir}
\bigcup_{x \in \Xcal} D \Big(p^{\pi}(x|x), \sum_{x' \ne x} p^{\pi}(x'|x) \Big)
\end{equation}
where $D(c,r):=[c-r,c+r]$ is a closed interval on $\mathbb{R}$. Since $\sum_{x' \in \Xcal} p^{\pi}(x'|x) = 1$ and $p^{\pi}(x'|x) \ge 0$ for all $x,x' \in \Xcal$, we know the set \eqref{eq:rl-pi-gcir} is contained in $[-1,1]$. Therefore, the maximum magnitude of all eigenvalues of $P^{\pi}$ is no greater than 1. As a result, we know $I_{n}-\gamma P^{\pi}$ is invertible provided $\gamma <1$, and \eqref{eq:pe-small-linear} has a unique solution 
\begin{equation}
\label{eq:pe-small-sol}
\vpi = (I_{n} - \gamma P^{\pi})^{-1} r^{\pi} .
\end{equation}
In practice, the solution \eqref{eq:pe-small-sol} can be computed by using existing numerical linear system solvers. The computational cost is $O(n^{3})$ in general (or lower if $P^{\pi}$ has special structure).
This is the cost of policy evaluation (Line 2 in Algorithm \ref{alg:policy-iteration}) in each iteration.

For policy improvement (Line 3 of Algorithm \ref{alg:policy-iteration}), we can directly set $\pi_{k+1}$ as the following deterministic greedy policy: For any $x \in \Xcal$, we set
\begin{equation}
\label{eq:pi-piux}
\pi_{k+1}(u|x) =
\begin{cases}
1, & \text{if } u = \argmax_{u' \in \Ucal} q^{\pi_{k}}(x,u') , \\
0, & \text{otherwise},
\end{cases}
\end{equation}
where $q^{\pi_{k}}$ is the Q-function associated with $\pi_{k}$ as defined in \eqref{eq:rl-q-pi}. If $q^{\pi_{k}}(x,\cdot)$ has more than one maximizers in $\Ucal$, we can choose any one of them in \eqref{eq:pi-piux}.

\paragraph{Moderate problems} In these cases, we often have state space size $|\Xcal| \approx 10^{6}$ or alike. It is computationally expensive to directly solve \eqref{eq:pe-small-linear} in general. One feasible approach is to solve the policy evaluation subproblem (Line 2 in Algorithm \ref{alg:policy-iteration}) for the given policy $\pi$ by applying the fixed point iteration $v_{j+1} = \Tpi v_{j}$ for $j \ge 0$ with some initial guess $v_{0}$. 
Since $\Tpi$ is $\gamma$-contractive, we know by Theorem \ref{thm:T-fixed-pt} that
\begin{equation*}
\| v_{j} - \vpi \|_{\infty} \le \gamma^{j} \| v_{0} - \vpi \|_{\infty}
\end{equation*}
for all $j$. Therefore, we can approximate $\vpi$ at a linear rate of $\gamma$ for any $\pi$.
In practice, we can terminate the fixed point iteration once $\|v_{j+1} - v_{j} \|_{\infty} < \varepsilon_{\text{tol}}$ for some prescribed tolerance (a user-defined hyperparameter) $\varepsilon_{\text{tol}} > 0$, and consider the policy evaluation (Line 2 in Algorithm \ref{alg:policy-iteration}) approximately solved and move on to policy improvement (Line 3 in Algorithm \ref{alg:policy-iteration})

To solve the policy improvement subproblem, we can set $\pi_{k+1}$ to the deterministic greedy policy as in \eqref{eq:pi-piux}. This is computationally feasible when $\Ucal$ is a finite set of small to moderate size.

\paragraph{Large problems} By large problems, we mean those where $\Xcal$ has $10^{10}$ states or even much larger, or $\Xcal$ is a continuous region in $\mathbb{R}^{d}$ with $d \ge 5$ such that a moderate space discretization results in over $10^{10}$ states (this is due to the well-known curse-of-dimensionality issue). 
In practice, these problems often have much larger sizes, which make storing the value function or the policy in the memory of a modern computer infeasible. 
Therefore, we often use function approximators to represent the value function and the policy as an alternative. 
Deep neural networks are shown to be very effective function approximators in these cases.

More specifically, in the policy evaluation step (Line 2 in Algorithm \ref{alg:policy-iteration}), we have the current estimated policy $\pi_{k}$. To find its value function $v^{\pi_{k}}: \Xcal \to \mathbb{R}$, we parameterize it as a deep neural network and train it by solving the following minimization problem:
\begin{equation}
\label{eq:pe-nn}
v^{\pi_{k}} = \argmin_{v} \frac{1}{2} \int |v(x) - T^{\pi_{k}}v(x)|^{2}\, dx
\end{equation}
where $T^{\pi_{k}}$ is the Bellman operator associated with the policy $\pi_{k}$ as defined in \eqref{eq:bellman-T-pi}. The minimization \eqref{eq:pe-nn} is with respect to the network parameters of $v$.
Since $T^{\pi_{k}}$ is $\gamma$-contractive, it has a unique fixed point, and therefore the problem \eqref{eq:pe-nn} has a unique minimizer which is the desired $v^{\pi_{k}}$.

The policy improvement step (Line 3 in Algorithm \ref{alg:policy-iteration}) requires finding the policy $\pi_{k+1}$ greedy with respect to $v^{\pi_{k}}$, as seen in \eqref{eq:pi-piux}. 
It reduces to solving the maximization problem for a greedy action $u \in \Ucal$ at each state $x \in \Xcal$.
The difficulty depends on whether such maximization problems are easy to solve or not, which can vary across different application problems. Therefore, method employed for solving this policy improvement subproblem is often case by case.

Nevertheless, there are some applications where the state space $\Xcal$ is extremely large but the action space is very small. For example, in the Go board game (called Weiqi in China, Igo in Japan, and Baduk in Korea), which is an ancient Chinese strategy board game, the state space $\Xcal$ contains all possible board positions and $|\Xcal| \approx 10^{170}$, while $|\Ucal|$ is only a few hundreds (the number of legal actions in $\Ucal$ is often smaller depending on the current state $x$). 
Moreover, the transition from $(x,u)$ to the next state $x'$ is deterministic: once an action $u$ is performed (placing a stone on the board) at state $x$ (the current board position), the next state $x'$ is known. 
In this case, the maximization problem in \eqref{eq:pi-piux} can be easily solved by simple search using a computer for any specific $x$.

However, it is still challenging to solve the policy improvement subproblem even if $|\Ucal|$ is small: The state space $\Xcal$ is still too large, and it is impossible to find an action $u$ for every $x \in \Xcal$. In practice, we can parameterize the policy $\pi$ (which is a conditional probability on $\Ucal$ given any state in $\Xcal$) as a deep neural network (called the policy network), and train this network along with the deep neural network (called the value network) which parameterizes the value function. For example, with the current value function $\vpi$, we can sample $M$ states $\{x_{i}: i \in [M]\}$ from $\Xcal$, and solve
\begin{equation*}
u_{i} = \argmax_{u \in \Ucal} \Big\{ r(x_{i}, u) + \gamma \mathbb{E}_{X' \sim p(\cdot| x_{i}, u)} [ \vpi(X') ] \Big\}
\end{equation*}
for each sampled state $x_{i}$. 
Then we use these sample pairs $\{(x_{i}, u_{i}): i \in [M]\}$ to train the policy network.
This is certainly not always simple and practically implementable. But it demonstrates a general strategy on handling the policy improvement step. We will also present several practical algorithms to tackle this issue in Section \ref{subsec:rl-model-free}.

\subsubsection*{Value Iteration}

Value iteration\index{Value iteration} is another iterative scheme for solving model-based reinforcement learning problems. 
The idea of value iteration is to generate a sequence of value functions which directly approximates the optimal value function $\vs$.
When the (approximation of) value function $\vs$ is found, we directly obtain the optimal policy which is the one greedy with respect to $\vs$.
By contrast, policy iteration needs to generate a sequence of continuously improving policy during its iterations.

Since $\Ts$ is $\gamma$-contractive, we know $\vs$ is its unique fixed point and can be approximated by using the fixed point iteration $v_{k+1} = \Ts v_{k}$ from any initial $v_{0}$. 
The value iteration algorithm is effectively implementing this fixed point iteration scheme.
We summarize the value iteration algorithm in Algorithm \ref{alg:value-iteration}.
\begin{algorithm}[t]
\caption{Value Iteration for Model-based Reinforcement Learning}
\label{alg:value-iteration}
\begin{algorithmic}[1]
\REQUIRE $v_{0} \in B(\Xcal)$, $k=0$.
\REPEAT
	\STATE (Value improvement) Set $v_{k+1} = \Ts v_{k}$.
	\STATE $k \leftarrow k+1$
\UNTIL Convergent.
\STATE Find $\pi$ greedy with respect to $v_{k}$.
\ENSURE $\pi$.
\end{algorithmic}
\end{algorithm}

To implement the value improvement\index{Value improvement} step (Line 2 of Algorithm \ref{alg:value-iteration}), we set
\begin{equation}
v_{k+1}(x) = \max_{u \in \Ucal} \Big\{ r(x,u) + \gamma \mathbb{E}_{X' \sim p(\cdot|x, u)} [v_{k}(X')] \Big\}
\end{equation}
for every $x$ and $k$. 
This is usually feasible for problems of small to moderate sizes.
However, for problems of large sizes, we need to use function approximators, such as deep neural networks, to approximate $v_{k}$.
To this end, for every given $v_{k}$, we can sample $M$ states $\{x_{i}: i \in [m]\}$ from $\Xcal$ and find their corresponding $u_{i}$'s by solving
\begin{equation*}
u_{i} = \argmax_{u \in \Ucal} \Big\{ r(x_{i}, u) + \gamma \mathbb{E}_{X' \sim p(\cdot| x_{i}, u)} [ v_{k}(X') ] \Big\} 
\end{equation*}
for every $i \in [M]$.
Then we use these sample pairs $\{(x_{i},u_{i}): i \in [M]\}$ to train $v_{k+1}$:
\begin{equation*}
v_{k+1} = \argmin_{v} \frac{1}{2M} \sum_{i=1}^{M} \Big| v(x_{i}) - \Big( r(x_{i},u_{i}) + \gamma \mathbb{E}_{X' \sim p(\cdot|x_{i},u_{i})} [v_{k}(X')] \Big) \Big|^{2} ,
\end{equation*}
where the minimization is with respect to the network parameters of $v$.
By Theorem \ref{thm:T-fixed-pt}, we know $v_{k}$ converges to $\vs$ in the $\infty$-norm at a linear rate of $\gamma$.
Upon convergence of $v_{k}$, we can choose $\pi$ greedy with respect the output $v_{k}$ by setting $\pi$ the same way as in
\eqref{eq:pi-piux}.
Notice that, only at this point of the value iteration algorithm (Algorithm \ref{alg:value-iteration}), we compute a policy and it is an approximation to the optimal policy $\pi^{*}$.

\subsection{Model-free Reinforcement Learning Algorithms}
\label{subsec:rl-model-free}

In many real-world applications, we may not have accurate models of the reward function $r$ (or the reward probability $p_{\tiny{R}}(\cdot|x,u)$) and the state transition probability $p(\cdot|x,u)$ due to the complexity of the environment or the problem setting.
In these cases, we need to use the agent to perform actions and interact with the environment. The agent can collect data during these interactions, with which we can estimate the reward and state transition probability, and find the optimal policy that yields the maximal total reward.
This class of methods are called the \emph{model-free} reinforcement learning algorithms.

In this subsection, we present a series of model-free reinforcement learning algorithms. 
These methods share lots of similarities, but they are also different due to the diversity of problem settings and data availability.
Nevertheless, we should keep the fundamental results developed in Section \ref{sec:rl-fundamentals} in mind which help to understand these algorithms.

\subsubsection*{Monte Carlo Method}
\label{subsubsec:rl-mc}

The original Monte Carlo method\index{Monte Carlo method} is a model-free method and often applied to problems of small to moderate sizes.
Recall that, for a given policy $\pi \in \Pi$, its value function is defined by
\begin{equation*}
\vpi(x) = \mathbb{E}_{\pi, p} \Big[ \sum_{t=0}^{\infty} \gamma^{t} R_{t+1} \Big| X_{0} = x \Big] ,
\end{equation*}
for every $x \in \Xcal$, where $R_{t}$ is the reward received at time $t$. 
In problems up to moderate size, $\Xcal$ is a finite set of size $10^{6}$ or alike, and therefore the value function $\vpi: \Xcal \to \mathbb{R}$ can be stored as a \emph{tabular} (as an array of size $|\Xcal|$) in computer memory and easily updated according to observation data, which we detail below.

Since we do not know the reward function $r$ and the transition probability $p$ in the model-free setting, we need to let the agent perform actions in the environment and collect sample data to estimate these quantities.
In this case, suppose we let the agent interacts with the environment by following a current estimated $\pi$, then we will observe a Markov decision process according to this $\pi$:
\begin{equation}
\label{eq:rl-mc-mdp}
X_{0}, \ U_{0}, \ R_{1}, \ \dots, \ X_{t}, \ U_{t}, \ R_{t+1}, \ X_{t+1}, \ U_{t+1}, \dots
\end{equation}
To extend the concept of value function $\vpi$, we define the \emph{gross reward} by
\begin{equation}
\label{eq:rl-mc-G}
G_{t}^{\pi}(x) := \sum_{\tau=t}^{\infty} \gamma^{\tau - t} R_{\tau+1},
\end{equation}
which is the total reward received since time $t$ with $X_{t} = x$, assuming the Markov decision process started from time $t$ instead of $0$. We also define the expected gross reward by
\begin{equation}
\label{eq:rl-mc-g}
g_{t}^{\pi} (x) :=  \mathbb{E}_{\pi, p} \Big[ G_{t}^{\pi}(x) \Big| X_{t} = x \Big] .
\end{equation}
Then $g_{t}^{\pi}(x)$ with any $t$ can be used to estimate $\vpi(x)$ with proper scaling factor $\gamma^{t}$. 
In the Monte Carlo method, we need to use the samples of $g_{t}^{\pi}(x)$, namely the samples of $G_{t}^{\pi}(x)$, to approximate the value function $\vpi$.

To be more specific, we let the agent follow the current policy $\pi$ and perform actions in the environment. Then we observe  a sample trajectory, called an \emph{episode}, of the Markov reward process.
An episode is just a part of the sampled Markov decision process \eqref{eq:rl-mc-mdp} without actions:
\begin{equation}
\label{eq:rl-mc-mrp}
x_{0}, \ r_{1}, \ x_{1}, \ r_{2}, \ \dots,\ x_{t}, \ r_{t+1}, \ x_{t+1}, \ \dots, x_{T}, \ r_{T+1} ,
\end{equation}
which terminates at some finite time $T$ in practice.
If we observe $x_{t} = x$ for at time $t$, then we take 
\begin{equation}
\label{eq:rl-mc-g-sample}
\sum_{\tau = t}^{T} \gamma^{\tau - t} r_{\tau+1}
\end{equation}
as one sample of $G_{t}^{\pi}(x)$. 
Notice that, we may obtain zero, one, or multiple samples like \eqref{eq:rl-mc-g-sample} for $x \in \Xcal$ in one episode, depending on the number of times that the state $x$ appears in the episode.
In practice, we can also run multiple episodes and take the average of samples like \eqref{eq:rl-mc-g-sample}. With a slight abuse of notation, we denote this average as $G_{t}^{\pi}(x)$ ($t$ is to imply that the gross reward is received since we reached the state $x$ at time $t$; but the actual value of $t$ does not matter because $G_{t}^{\pi}(x)$ is an estimate of $\vpi(x)$ for any $t$). In this case, the Monte Carlo method updates $\vpi(x)$ by
\begin{equation}
\label{eq:rl-mc-vpi-update}
\vpi(x) \leftarrow \vpi(x) + \alpha ( G_{t}^{\pi}(x) - \vpi(x)),
\end{equation}
where $\alpha > 0$ is the step size.

The Monte Carlo method is known to yield unbiased estimate of the value function $\vpi$ if the episode is infinitely long, but the variance of the estimate is often large in general. After $\vpi$ is estimated, we can find the policy $\pi'$ greedy with respect to $\vpi$, and repeat the process of the Monte Carlo method using the updated policy $\pi'$ and estimate the new value function $v^{\pi'}$, and so on.
As a result, the Monte Carlo method generates a sequence of improving policies as in the value iteration algorithm (Algorithm \ref{alg:value-iteration}), but without knowledge of the reward function $r$ and the transition probability $p$.

\subsubsection*{Temporal Difference Method}

The Temporal difference (TD)\index{Temporal difference method} is another model-free method closely related to the Monte Carlo method. 
The difference is that, for the current policy $\pi$, the tabular of $\vpi$ in the TD method is continuously updated when going through the sampled Markov reward process (an episode) of form \eqref{eq:rl-mc-mrp}. By contrast, this tabular is only updated at the completion of every episode in the Monte Carlo method.

More specifically, suppose $x_{t} = x$, and the agent receives a reward $r_{t+1}$ and move to state $x_{t+1}$, then we can update $\vpi(x)$ by
\begin{equation}
\label{eq:rl-td0}
\vpi(x) \leftarrow \vpi(x) + \alpha \Big( \underbrace{\overbrace{r_{t+1} + \gamma \vpi(x_{t+1})}^{\text{TD target}} - \vpi(x)}_{\text{TD error}} \Big)
\end{equation}
where $\alpha>0$ is the step size.
As we can see, the TD method \eqref{eq:rl-td0} uses the so-called \emph{TD target} $r_{t+1} + \gamma \vpi(x_{t+1})$ to replace $G_{t}^{\pi}(x)$ in the Monte Carlo method \eqref{eq:rl-mc-vpi-update}.
The goal is to make the \emph{TD error} in \eqref{eq:rl-td0} become zero, yielding the Bellman equation of $\vpi$ in \eqref{eq:v-pi-self}.
Note that the TD method uses part of the current estimate, i.e., $\vpi(x_{t+1})$, to estimate the value $\vpi(x)$. This strategy is called \emph{bootstrapping}. 
Comparing to the Monte Carlo method, the estimate of $\vpi$ obtained in the TD method is biased, but the variance is often smaller than that in the Monte Carlo method.

The TD method discussed above is often called in short as the TD(0) method. It can be generalized by using the $n$-step gross rewards for $n=1,2,\dots$ as the TD target. To this end, we denote
\begin{align*}
G_{t,1}^{\pi}(x) & := R_{t+1} + \gamma \vpi(X_{t+1}) , \\
G_{t,2}^{\pi}(x) & := R_{t+1} + \gamma R_{t+2} + \gamma^{2} \vpi(X_{t+2}) , \\
& \ \ \vdots \\
G_{t,n}^{\pi}(x) & := R_{t+1} + \gamma R_{t+2} + \gamma^{2} R_{t+3} + \dots + \gamma^{n-1} R_{t+n} +  \gamma^{n} \vpi(X_{t+n}) , \\
& \ \ \vdots \\
G_{t,\infty}^{\pi}(x) & := G_{t}^{\pi},  \qquad \text{(Assuming $R_{t} = 0$ when $t > T+1$)}.
\end{align*}
Then for any $\lambda \in [0,1]$, we define
\begin{equation}
\label{eq:rl-td-G}
G_{t}^{\pi,\lambda} (x) := \sum_{n=1}^{\infty} (1-\lambda) \lambda^{n-1} G_{t,n}^{\pi}(x) .
\end{equation}
Notice that $(1-\lambda) \lambda^{n-1}$ plays the role of weight in the sum \eqref{eq:rl-td-G}, and the total weight is $\sum_{n=1}^{\infty} (1-\lambda) \lambda^{n-1} = 1$.
Then the $\text{TD}(\lambda)$ method, as an extension of the TD(0) method, updates $\vpi(x)$ by
\begin{equation}
\label{eq:rl-td-lambda}
\vpi(x) \leftarrow \vpi(x) + \alpha ( G_{t}^{\pi, \lambda}(x) - \vpi(x)) .
\end{equation}
In fact, the $\text{TD}(\lambda)$ method is just an interpolation of the TD(0) method and the Monte Carlo method: When $\lambda = 0$, we have
\begin{equation*}
G_{t}^{\pi, 0}(x) = G_{t,0}^{\pi}(x) = R_{t+1} + \gamma v^{\pi}(X_{t+1})
\end{equation*}
and hence $\text{TD}(\lambda)$ is exactly the TD(0) method; when $\lambda \to 1$, all the weights in \eqref{eq:rl-td-G} approximate 1, and the $\text{TD}(\lambda)$ becomes the Monte Carlo method asymptotically.

For any policy $\pi$, the value $\vpi(x)$ updated by \eqref{eq:rl-td0} iteratively is provably convergent to the true value function of $\pi$ provided proper assignments of step sizes and infinite amount of episode data. We omit the proofs here.

Same as the Monte Carlo method, after the value function $\vpi$ is updated, we need to find a policy $\pi'$ greedy with respect to $\vpi$, and then repeat the process of the TD update \eqref{eq:rl-td0} using $\pi'$ to estimate the new function $v^{\pi'}$, and so on. Therefore, the TD method is similar to the Monte Carlo method, but considered more efficient in using data than the latter in practice.

\subsubsection*{The SARSA Algorithm}

The SARSA\index{SARSA algorithm} algorithm is also a model-free method that has a lot of similarity as the TD method and applicable to problems of small to moderate sizes.
The difference is the SARSA algorithm learns the quality function (the state-action value function) $q^{\pi}$ instead of the state value function $\vpi$ in the TD method.
Given a policy $\pi$, the agent can perform actions and generate a Markov decision process
\begin{equation}
\label{eq:sarsa-mdp}
X_{0}, \ U_{0}, \ R_{1}, \ \dots, \ X_{t}, \ U_{t}, \ R_{t+1}, \ X_{t+1}, \ U_{t+1}, \ \dots
\end{equation}
To learn $q^{\pi}$, the SARSA algorithm also keeps a tabular of $\qpi$ on $\Xcal \times \Ucal$ in computer memory. Then by sampling an episode of \eqref{eq:sarsa-mdp}, we observe a sample trajectory of the Markov decision process
\begin{equation*}
x_{0}, \ u_{0}, \ r_{1}, \ \dots, \ x_{t}, \ u_{t}, \ r_{t+1}, \ x_{t+1}, \ u_{t+1}, \ \dots
\end{equation*}
and updates $\qpi(x,u)$ by
\begin{equation}
\label{eq:sarsa}
\qpi(x_{t},u_{t}) \leftarrow \qpi(x_{t},u_{t}) + \alpha (r_{t+1} + \gamma \qpi(x_{t+1}, u_{t+1}) - \qpi(x_{t},u_{t})) .
\end{equation}
We can see a lot of similarities between \eqref{eq:rl-td0} and \eqref{eq:sarsa} (we directly use $q^{\pi}(x_{t},u_{t})$ instead of $q^{\pi}(x,u)$ in \eqref{eq:sarsa} but the idea is the same as $\vpi(x_{t})$ in \eqref{eq:rl-td0}).
The goal of \eqref{eq:sarsa} is to make $r_{t+1} + \gamma \qpi(x_{t+1}, u_{t+1})$ equal to $\qpi(x_{t},u_{t})$ in expectation provided $x_{t+1}$ is a sample following $p(\cdot|x_{t},u_{t})$, yielding \eqref{eq:rl-q-pi}.
As each of such updates requires a tuple $(x_{t}, u_{t}, r_{t+1}, x_{t+1}, u_{t+1})$ extracted from the sample trajectory, the SARSA algorithm got its name as a brevity of (State, Action, Reward, State, Action).

For any policy $\pi$, the function $\qpi$ updated by the SARSA scheme \eqref{eq:sarsa} iteratively can also be proved to converge to the true Q-function of $\pi$. Then we can update the policy to the one greedy with respect to $\pi$, namely,
\begin{equation}
\label{eq:rl-greedy-pi}
\pi'(u|x) =
\begin{cases}
1, & \text{if } u = \argmax_{u' \in \Ucal} \qpi(x,u') , \\
0, & \text{otherwise} .
\end{cases}
\end{equation}
The greedy policy above is chosen this way because
\begin{equation*}
T^{\pi'} \vpi(x) = T^{*} \vpi(x) = \max_{\hat{\pi} \in \Pi} \mathbb{E}_{U' \sim \hat{\pi}(\cdot|x)}[ \qpi(x,U') ] = \max_{u' \in \Ucal} \qpi(x,u') ,
\end{equation*}
provided a maximizer can be attained in $\Ucal$ for every $x \in \Xcal$.
Once we have $\pi'$, we can generate new sample trajectories following $\pi'$, and apply the SARSA scheme \eqref{eq:sarsa} on the trajectories to obtain $q^{\pi'}$, and so on.
This process generates a sequence of policies $\pi_{1}, \pi_{2},\dots, \pi_{k},\dots$, where $\pi_{k+1}$ is greedy with respect to $v^{\pi_{k}}$ for every $k$.
By Policy Improvement Theorem (Theorem \ref{thm:policy-improvement}), we know $\pi_{k}$ converges to $\pis$.

In practice, the SARSA algorithm does not need a very long episode to update $q$, because it can be updated continuously during an episode as in \eqref{eq:sarsa} and terminated any time.

A main drawback of the SARSA algorithm is that it requires tuple $(x_{t}, u_{t}, r_{t+1}, x_{t+1}, u_{t+1})$ to implement the update \eqref{eq:sarsa}. If some state-action pair $(x_{t+1}, u_{t+1})$ is not sampled, which often happens when the current estimate of $q$ suggests that certain action $u_{t+1}$ is unlikely valuable at the state $x_{t+1}$, then we may never be able to update $q(x_{t+1}, u_{t+1})$.

This drawback is a common issue in model-free algorithms, and the key behind is about the tradeoff between \emph{exploitation}\index{Exploitation} and \emph{exploration}\index{Exploration} in reinforcement learning.
Exploitation means that the algorithm always takes the best action according to the previously obtained state-action value $q$, whereas exploration reserves some probability to try other actions and may discover even better actions.
A real-life analogue is that by exploitation we always go to our current favorite restaurant and try all items on its menu, versus by exploration we occasionally try other restaurants randomly and see if they have better dishes.

In many model-free reinforcement learning algorithms, exploration is implemented by the \emph{$\epsilon$-greedy policy}\index{Policy!$\epsilon$-greedy}
\begin{equation}
\label{eq:rl-eps-policy}
\pi(u|x) =
\begin{cases}
\frac{\epsilon}{|\Ucal|} + 1 - \epsilon, & \text{if } u = \argmax_{u' \in \Ucal} q^{\pi}(x,u') , \\
\frac{\epsilon}{|\Ucal|}, & \text{otherwise} .
\end{cases}
\end{equation}
for some small $\epsilon \in (0,1)$, where $|\Ucal|<\infty$, e.g., $\Ucal$ is a finite set or a continuous compact subset in an Euclidean space.
(We can use probabilities other than uniform distribution for cases where $\Ucal$ is unbounded.)
To implement this, we can first draw a Bernoulli sample with two possible outcomes: Head with probability $1-\epsilon$ versus Tail with probability $\epsilon$, then we take the maximizer $u$ when we get the Head and randomly choose one from $\Ucal$ with uniform distribution when we get the Tail.

Similar to the $n$-step temporal difference method, the SARSA algorithm may also use a longer tuple
\begin{equation*}
(x_{t}, \ u_{t},\ r_{t+1},\ x_{t+1},\ u_{t+1},\ \dots,\ r_{t+n},\ x_{t+n},\ u_{t+n})
\end{equation*}
in a sample trajectory of the Markov decision process. In this case, we have the variant of the SARSA algorithm, called the \emph{$n$-step SARSA} algorithm which updates $q$ by
\begin{align*}
q^{\pi}(x_{t}, u_{t}) \leftarrow &\ q^{\pi}(x_{t}, u_{t}) \\
& \quad + \alpha \Big( r_{t+1} + \dots + \gamma^{n-1} r_{t+n} + \gamma^{n} q^{\pi}(x_{t+n}, u_{t+n}) - q^{\pi}(x_{t}, u_{t}) \Big) .
\end{align*}

\subsubsection*{The Q-learning Algorithm}

The Q-learning\index{Q-learning algorithm} algorithm is a variant of the SARSA algorithm by changing the goal of learning the Q-function $q^{\pi}$ of the current estimate of policy $\pi$ to learning the optimal Q-function $\qs$ directly.
This is similar to the value iteration algorithm (Algorithm \ref{alg:value-iteration}) which learns the optimal state value function $\vs$).
Recall from \eqref{eq:rl-q-star} that the optimal state-action value function $\qs$ satisfies the Bellman equation
\begin{align}
\qs(x,u) 
& = r(x,u) + \mathbb{E}_{X' \sim p(\cdot|x,u)} [ \vs(X') ] \nonumber \\
& = r(x,u) + \mathbb{E}_{X' \sim p(\cdot|x,u)} [ \max_{u' \in \Ucal} \qs(X', u') ] . \label{eq:rl-q-learning-bellman}
\end{align}
The transition probability $p(\cdot|x,u)$ is determined by the environment, not any policy. 
Therefore, if we observe a transition triple $(x_{t},u_{t},x_{t+1})$ in an episode, we can take $\vs(x_{t+1})$ as an unbiased estimate of $\mathbb{E}_{X' \sim p(\cdot|x_{t},u_{t})}[\vs(X')]$.
Similar to the update rule in the SARSA algorithm, the update of $q$ in the Q-learning algorithm is 
\begin{equation}
\label{eq:rl-q-learning-update}
q(x_{t},u_{t}) \leftarrow q(x_{t},u_{t}) + \alpha \Big(r_{t+1} + \gamma \max_{u' \in \Ucal} q(x_{t+1},u') - q(x_{t},u_{t}) \Big) .
\end{equation}
The convergence of $q$ updated above to $\qs$ can also be verified under certain conditions. The rest part of the Q-learning algorithm is the same as the SARSA algorithm. 
Once $\qs$ is learned by iterating \eqref{eq:rl-q-learning-update}, we can find a policy greedy with respect to $\qs$ as in \eqref{eq:rl-greedy-pi} and it will be an optimal policy.

Compared to the SARSA algorithm, the major advantage of the Q-learning algorithm is that only tuples of form $(x_{t},u_{t},r_{t+1},x_{t+1})$ are used as samples. 
They do not include $u_{t+1}$ and hence is less data demanding than the SARSA algorithm.

\paragraph{Summary} 
At this point, we can take a quick summary about the model-free methods we discussed so far, including the Monte Carlo method, the temporal difference (TD) method, the SARSA and Q-learning algorithms.
More specifically, given a policy $\pi$, the Monte Carlo method uses complete episodes to estimate the value function $\vpi$, the TD(0) method uses the current estimation $\vpi$ to itself (a bootstrapping strategy), and the $\text{TD}(\lambda)$ method is an interpolation between the TD(0) method and the Monte Carlo method.
For both Monte Carlo and TD methods, a new policy greedy with respect to $\vpi$ is obtained and the process is repeated. They can be regarded as model-free versions of the policy iteration algorithm (Algorithm \ref{alg:policy-iteration}).

By contrast, the SARSA algorithm learns the Q-function $q^{\pi}$ for the current policy $\pi$, and then update $\pi$ to the policy greedy with respect to $q^{\pi}$. Then the sequence of policies generated by the SARSA algorithm is expected to converge to the optimal policy $\pis$. Therefore, the SARSA algorithm can also be thought of as a model-free version of the policy iteration algorithm, but it learns $q^{\pi}$ instead of $\vpi$ in each iteration.

The Q-learning algorithm directly learns the optimal Q-function $\qs$ by updating $q$ in the rule \eqref{eq:rl-q-learning-update}. Upon convergence, (an approximation of) $\qs$ is obtained and the optimal policy is set to the one greedy with respect to $\qs$. Therefore, the Q-learning algorithm can be considered as a model-free version of the value iteration algorithm (Algorithm \ref{alg:value-iteration}), except that it is approximating $\qs$ instead of $\vs$.

All these methods can be applied to problems of small to moderate sizes, as they need to store either $\vpi$ or $q^{\pi}$ as a tabular in computer memory. We consider their variants that can be used for large problems next.

\subsubsection*{Deep Q-Network (DQN) Algorithm}

%The SARSA and Q-learning algorithms above apply to problems of small to moderate size.
%
%In such cases, they can store the Q-function as a tabular over $\Xcal \times \Ucal$.
%
%The entries of the tabular Q-function are updated according to these methods.
%
We consider a variant of the Q-learning algorithm, called the deep Q-network (DQN)\index{DQN algorithm} algorithm, that can solve model-free reinforcement problems with large state space $\Xcal$ and action space $\Ucal$.
The key in the DQN algorithm is using a function approximator to approximate the optimal Q-function $\qs$.
We use deep neural networks as such approximators.

Recall that the Bellman equation of $\qs$ is given by \eqref{eq:rl-q-learning-bellman}.
Therefore, it is natural to consider the loss function for training $\qs$ as
\begin{equation}
\label{eq:rl-dqn-loss}
\frac{1}{2} \Big|\qs(x_{t},u_{t}) - \Big( r_{t+1} + \gamma \max_{u' \in \Ucal} \qs(x_{t+1},u') \Big) \Big|^{2} 
\end{equation}
for a sample tuple $(x_{t},u_{t},r_{t+1},x_{t+1})$.
By running many episodes, we can get many sample tuples of form $(x_{t},u_{t},r_{t+1},x_{t+1})$, and take the average of them as the loss function of the Q-network.
In each iteration, we can take an $\epsilon$-greedy policy $\pi$ with respect to the current estimate of $\qs$, and run episodes again to update $\qs$.

In practice, the DQN algorithm uses an \emph{experience replay} buffer to store previously obtained $(x_{t},u_{t},r_{t+1},x_{t+1})$. Note that, once $(x_{t},u_{t})$ is determined, $r_{t+1} = r(x_{t},u_{t})$ or $r_{t+1} \sim p_{\tiny{R}}(\cdot|x_{t},u_{t})$, and $x_{t+1} \sim p(\cdot|x_{t},u_{t})$ are independent of policies, so the data obtained from a policy earlier can be stored in the experience replay buffer, and reused in the training of $\qs$ for a new policy. 
This mitigates the issue of sampling dependency if we only use samples generated by following the current policy, because the same samples can be reused by different policies and the overall sampling complexity can be improved.

Notice that an issue with the DQN setting above is that $\qs$ appears at both places in the loss \eqref{eq:rl-dqn-loss}. This often causes the training unstable because once the current estimate $\qs$ prefers a specific state-action sequel, then it will be further encouraged during the later training and cause larger estimation error.
To mitigate this issue, we can set two DQNs, denoted by $q$ and $\hat{q}$, which are called the training DQN and the target DQN, respectively.
When we update $q$, we keep $\hat{q}$ fixed and solve 
\begin{equation}
\label{eq:dqn-training-q}
\min_{q} \frac{1}{2} \iiint \Big| q(x,u) - \Big(r(x,u) + \gamma \max_{u' \in \Ucal} \hat{q}(x',u') \Big) \Big|^{2} \, dx' du dx 
\end{equation}
After several iterations of \eqref{eq:dqn-training-q} in updating $q$, we set $\hat{q} \leftarrow q$ and repeat \eqref{eq:dqn-training-q} for several more interations. This often improves the computational stability empirically.
However, the error accumulation issue persists.

\subsubsection*{Double Deep Q-Networks Algorithm}

Notice that, in the DQN algorithm above, we fix $\hat{q}$ when we update $q$. Namely, the target used to update $q$ is given by the following function:
\begin{equation*}
r(x,u) + \gamma \max_{u' \in \Ucal} \hat{q}(x',u') .
\end{equation*}
If we assume a greedy $u' \in \Ucal$ in the maximization problem above, the target is effectively
\begin{equation}
\label{eq:dqn-target-equiv}
r(x,u) + \gamma \hat{q}(x', \argmax_{u' \in \Ucal} \hat{q}(x',u')) .
\end{equation}
Therefore, if $\hat{q}$ overestimates $(x,u)$, then this will still accumulate error.

In the double deep Q-networks (Double DQNs)\index{Double DQNs algorithm} algorithm, we simply change \eqref{eq:dqn-target-equiv} to the following one:
\begin{equation*}
r(x,u) + \gamma \hat{q}(x', \argmax_{u' \in \Ucal} q(x',u')) ,
\end{equation*}
which is used as the target to train $q$. When $q$ and $\hat{q}$ are updated alternately after every few iterations, it is found that the error accumulation issue can be greatly mitigated empirically.

\subsubsection*{Policy Gradient Algorithm}
\label{subsubsec:policy-grad}

The Monte Carlo algorithm, the temporal difference algorithm, and the SARSA algorithm all aim at learning the state value function $\vpi$ or the state-action value function $q^{\pi}$ for some given policy $\pi$.
On the other hand, the Q-learning algorithm, the DQN method and their variants attempt to learn $\qs$. Only when the optimal $\vs$ is $\qs$ is learned, these methods set the optimal policy to be the one greedy to these value function.

The \emph{policy gradient algorithm}\index{Policy gradient} takes a different approach by learning the optimal policy $\pis$ directly. 
In the policy gradient algorithm, we also parametrize the policy as a deep neural network. 
This network is either a mapping from the state space $\Xcal$ to the action space $\Ucal$, or a probability distribution on $\Ucal$ for every $x \in \Xcal$.
However, we need a scalar-valued objective function to train this policy $\pis$. 
Specifically, we need to compute the gradient of the objective function with respect to the policy (more precisely, the network parameters of this policy network).

Suppose we currently have an estimated policy $\pi$, then we can generate a sample episode of arbitrary length $t$:
\begin{equation}
\label{eq:rl-pg-z}
z_{t} = (x_{0},\ u_{0},\ x_{1},\ u_{1},\ \dots,\ x_{t},\ u_{t}) ,
\end{equation}
where $x_{0} \sim p_{0}$ for some initial state distribution (which is given beforehand and independent on any specific policy), and $u_{t} \sim \pi(\cdot|x_{t})$ determined by the policy $\pi$ and $x_{t+1} \sim p(\cdot|x_{t}, u_{t})$ by the environment at any time $t$.
Here we assume that the episode may terminate at a finite time and the discount factor $\gamma \in [0,1]$, but the derivation we showed below also applies to cases for infinite time horizon and the discount factor $\gamma \in (0,1)$.

For our derivation below, we define
\begin{equation}
\label{eq:rl-pg-rho-pi}
\rho_{t}^{\pi}(z_{t}) := p_{0}(x_{0}) \pi(u_{0}|x_{0}) p(x_{1}|x_{0},u_{0}) \dots p(x_{t}|x_{t-1},u_{t-1}) \pi(u_{t}|x_{t})
\end{equation}
for any time $t$. The total reward of the episode $z_{T}$ terminated at a finite time $T$ is given by
\begin{equation}
\label{eq:rl-pg-R}
R(z_{T}) := \sum_{t=0}^{T} \gamma^{t} r_{t+1} = \sum_{t=0}^{T} \gamma^{t} r(x_{t},u_{t})
\end{equation}
% for the trajectory $z_{T}$ defined in \eqref{eq:rl-pg-z}.
%
Then we denote the objective function, which is the expected total reward of $z_{T}$ (determined by the policy $\pi$), as
\begin{equation}
\label{eq:rl-pg-J}
J(\pi) := \mathbb{E}_{Z_{T} \sim \rho^{\pi}_{T}} [R(Z_{T})] = \int R(z_{T}) \rho_{T}^{\pi}(z_{T}) \, dz_{T} .
\end{equation}
Now our goal is to find $\nabla_{\pi} J(\pi)$, the gradient of $J$ with respect to $\pi$.
Notice that, $\nabla_{\pi}$ hereafter should again be interpreted as the gradient with respect to the network parameters of $\pi$.

To write the gradient $\nabla_{\pi} J(\pi)$ as a practically implementable objective function, we first notice that
\begin{equation}
\label{eq:rl-pg-J-decomp}
J(\pi) = \int R(z_{T}) \rho_{T}^{\pi}(z_{T}) \, dz_{T} = \sum_{t=0}^{T} \gamma^{t} \underbrace{\int r(x_{t}, u_{t}) \rho^{\pi}(z_{T}) \, dz_{T}}_{=:\, J_{t}(\pi)} 
\end{equation}
by combining \eqref{eq:rl-pg-R} and \eqref{eq:rl-pg-J}.
We see that $J_{t}(\pi)$ can be rewritten as
\begin{align}
J_{t}(\pi)
& = \int r(x_{t},u_{t}) \rho_{T}^{\pi}(z_{T}) \, dz_{T} \nonumber \\
& = \iint r(x_{t},u_{t}) p_{0}(x_{0}) \pi(u_{0}|x_{0}) \cdots \pi(u_{T}|x_{T}) \, (du_{T}\cdots dx_{t+1}) dz_{t} \nonumber \\
& = \int \Big[r(x_{t},u_{t}) p_{0}(x_{0}) \cdots \pi(u_{t}|x_{t}) \label{eq:rl-pg-Jt} \\
& \qquad \qquad \cdot \Big( \int p(x_{t+1}|x_{t},u_{t}) \cdots \pi(u_{T}|x_{T}) \, du_{T}\cdots dx_{t+1} \Big) \Big] dz_{t} \nonumber \\
& = \int r(x_{t},u_{t}) \underbrace{p_{0}(x_{0}) \cdots \pi(u_{t}|x_{t})}_{=\, \rho_{t}^{\pi}(z_{t})} \, dz_{t} , \nonumber
\end{align}
where the second equality is due to the definition of $\rho_{T}^{\pi}(z_{T})$ in \eqref{eq:rl-pg-z} and $dz_{T} = (du_{T} dx_{T}\cdots dx_{t+1})dz_{t}$; the third equality due to the integration over $(u_{T},x_{T},\dots,x_{t+1})$ first; and the last equality is because of
\begin{equation}
\label{eq:rl-pg-int1}
\int p(x_{t+1}|x_{t},u_{t}) \cdots \pi(u_{T}|x_{T}) \, du_{T}\cdots dx_{t+1} = 1 .
\end{equation}

Now we first check $\nabla_{\pi} J_{t}(\pi)$, which is given by
\begin{align}
\nabla_{\pi} J_{t}(\pi)
& = \nabla_{\pi} \Big( \int r(x_{t}, u_{t}) \rho_{t}^{\pi}(z_{t}) \, dz_{t} \Big) \nonumber \\
& = \int r(x_{t}, u_{t}) \rho_{t}^{\pi}(z_{t}) \Big( \nabla_{\pi} \log \rho_{t}^{\pi}(z_{t})\Big) \, dz_{t} \label{eq:rl-pg-dJt} \\
& = \int r(x_{t}, u_{t}) \rho_{T}^{\pi}(z_{T}) \Big( \nabla_{\pi} \log \rho_{t}^{\pi}(z_{t})\Big) \, dz_{T} \nonumber
\end{align}
where we used the fact that $\nabla_{\pi} \rho_{t}^{\pi}(z_{t}) = \rho_{t}^{\pi}(z_{t}) \nabla_{\pi} \log \rho_{t}^{\pi}(z_{t})$ in the second equality, and \eqref{eq:rl-pg-int1} again in the last equality.
We take a closer look at $\nabla_{\pi} \log \rho_{t}^{\pi}(z_{t})$ and notice that by \eqref{eq:rl-pg-rho-pi} there is
\begin{align}
\nabla_{\pi} \log \rho_{t}^{\pi}(z_{t})
& = \nabla_{\pi} \Big( \log p_{0}(x_{0}) + \sum_{s=0}^{t} \log \pi(u_{s}|x_{s}) + \sum_{s=0}^{t} \log p(x_{s+1}|x_{s},u_{s}) \Big) \nonumber \\
& = \sum_{s=0}^{t} \nabla_{\pi} \log \pi(u_{s}|x_{s}) \label{eq:rl-pg-drhot}
\end{align}
where the last equality is due to the fact that only the middle term in the parenthesis above contains $\pi$ (the first term only contains the initial distribution $p_{0}$ and the last term only contains the state transition distributions determined by the environment).
Hence we have the final form of $\nabla_{\pi} J_{t}(\pi)$ by combining \eqref{eq:rl-pg-dJt} and \eqref{eq:rl-pg-drhot}:
\begin{equation}
\label{eq:rl-pg-dJt-final}
\nabla_{\pi} J_{t}(\pi) = \int r(x_{t},u_{t}) \Big( \sum_{s=0}^{t} \nabla_{\pi} \log \pi(u_{s}|x_{s}) \Big) \rho_{T}^{\pi}(z_{T}) \, dz_{T} .
\end{equation}
Now we have
\begin{align}
\nabla_{\pi} J(\pi)
& = \nabla_{\pi} \Big( \sum_{t=0}^{T} \gamma^{t} J_{t}(\pi) \Big) \nonumber \\
& = \sum_{t=0}^{T} \gamma^{t} \nabla_{\pi} J_{t}(\pi) \nonumber \\
& = \int \sum_{t=0}^{T} \sum_{s=0}^{t} \gamma^{t} r(x_{t},u_{t}) \nabla_{\pi} \log \pi(u_{s}|x_{s}) \rho_{T}^{\pi}(z_{T}) \, dz_{T} \label{eq:rl-pg-dJ-final}  \\
& = \int \sum_{t=0}^{T} \Big(\sum_{s=t}^{T} \gamma^{s} r(x_{s},u_{s})\Big) \nabla_{\pi} \log \pi(u_{t}|x_{t}) \rho_{T}^{\pi}(z_{T}) \, dz_{T} \nonumber \\
& = \mathbb{E}_{Z_{T} \sim \rho_{T}^{\pi}} \Big[ \sum_{t=0}^{T} \Big(\sum_{s=t}^{T} \gamma^{s} R_{s+1}\Big) \nabla_{\pi} \log \pi(U_{t}|X_{t}) \Big] , \nonumber
\end{align}
where the fourth equality is due to the exchange of the order of summations over $t$ and $s$ and then swapping the symbols $s$ and $t$, and the last equality is due to $R_{s+1} = r(X_{s},U_{s})$ for all $s$.

Notice that the right-hand side of \eqref{eq:rl-pg-dJ-final} can be approximated in practice. If we take the Monte Carlo approach, then a sample episode $z_{T} \sim \rho_{T}^{\pi}$ (the time $T$ depends on the actual length of $z_{T}$ and is not fixed beforehand) can be sampled by letting the agent behave in the environment following the current estimated policy $\pi$. In this case, the reward $r_{s+1} = r(x_{t},u_{t})$ can be stored and used in the estimation of \eqref{eq:rl-pg-dJ-final}.
If we want to approximate the reward $\sum_{s=t}^{T} \gamma^{s} r_{s+1}$ using an estimated Q-function $q(x_{t},u_{t})$ (which can be approximated by a deep neural network with input from $\Xcal \times \Ucal$ as in the DQN and double DQN algorithms above), then we see that $\sum_{s=t}^{T} \gamma^{s} r_{s+1} = \gamma^{t} q(x_{t},u_{t})$.

Now we see that $\log \pi$ is the only term not yet determined in \eqref{eq:rl-pg-dJ-final}. If $\Xcal$ is a large finite set or a continuous subset of an Euclidean space, we can parameter $\pi(\cdot|x)$ as a probability on $\Ucal$ given any $x \in \Xcal$. More specifically, if $\Ucal = \{u_{1}, \dots, u_{m}\}$ is a finite set of $m$ actions, then we can parameterize $\pi: \Xcal \to \mathbb{R}^{m}$ as a deep neural network and set the last operation in this network to be the softmax function:
\begin{equation}
\text{softmax}(w) := \frac{1}{\sum_{i=1}^{m} e^{w_{i}} } ( e^{w_{1}}, e^{w_{2}}, \dots, e^{w_{m}}) \in \mathbb{R}^{m} 
\end{equation}
for any $w \in \mathbb{R}^{m}$.
In this case, we know $\pi(u_{i}|x) > 0$ and $\sum_{i=1}^{m} \pi(u_{i}|x) = 1$ for every $i \in [m]$ and $x \in \Xcal$, and the actual action $u$ at the state $x$ should be taken by following the distribution $\pi(\cdot|x)$ on $\Ucal$. 
In this case, it is straightforward to find $\nabla_{\pi} \pi(u|x)$, the gradient of $\pi(u|x)$ with respect to the network parameters of $\pi$, for any $(x,u) \in \Xcal \times \Ucal$.

If $\Ucal$ is a continuous subset of $\mathbb{R}^{m}$ for some $m \in \mathbb{N}$, then we may set two deep neural networks, $\mu_{\pi}: \Xcal \to \mathbb{R}^{m}$ and $\sigma_{\pi}: \Xcal \to (0,\infty)$ (we can easily generalize $\sigma_{\pi}$ to have range in $(0,\infty)^{m}$ so $\sigma_{\pi}$ can be the diagonal entries of a diagonal covariance matrix, or an $m$-by-$m$ symmetric positive definition matrix directly as the covariance matrix), and set $\pi(\cdot\,|x) = \Ncal(\,\cdot\,;\,\mu_{\pi}(x), \sigma_{\pi}(x)^{2} I_{m})$. Then the closed form of $\nabla_{\pi} \pi(u|x)$ is also easy to obtain.

Once we know how to compute $\nabla_{\pi} J(\pi)$ for any $\pi$, we can apply gradient ascent method or alike to update the network parameters of the policy $\pi$ and approximate the optimal policy $\pis$.

\subsubsection*{The Actor-Critic Algorithm}

Notice that the algorithms we discussed above either learn the state (or state-action) value function (such as the DQN and double DQN algorithms) or the policy (such as the policy gradient algorithm).
The actor-critic (AC)\index{Actor-Critic algorithm} algorithm is a new one that learns both of the state value function $\vpi$ and the actor $\pi$ jointly. In the AC algorithm, both $\pi$ and $\vpi$ are parameterized as deep neural networks, and they are called the actor network and the critic network, respectively.
The goal of the critic network is to evaluate the value function $\vpi$ of the current policy determined by the actor network $\pi$, whereas the goal of the actor network is to improve $\pi$ based on the feedback from $\vpi$ given by the critic network.
More specifically, recall that
\begin{equation}
\label{eq:ac-dJpi}
\nabla_{\pi} J(\pi) = \mathbb{E}_{Z_{T}\sim \rho_{T}^{\pi}} \Big[ \sum_{t=0}^{T} \Big( \sum_{s=t}^{T} \gamma^{s} R_{s+1} \Big) \nabla_{\pi} \log \pi(u_{t}|x_{t}) \Big] .
\end{equation}
In the AC algorithm, we do not sample episodes to update $\pi$. Instead, we use a similar idea as in the TD method with sample tuple $(x_{t}, u_{t}, r_{t+1}, x_{t+1})$ to replace $\sum_{s=t}^{T} \gamma^{s} R_{s+1}$ with 
\begin{equation*}
\gamma^{t} \sum_{s=t}^{T} \gamma^{s-t} R_{s+1} = \gamma^{t} (r_{t+1}) + \gamma \vpi(x_{t+1}) ,
\end{equation*}
where $\vpi$ is the critic network. Once the actor network $\pi$ is updated by the gradient ascent algorithm or alike on \eqref{eq:ac-dJpi}, we can turn to update the critic network $\vpi$ by minimizing
\begin{equation}
\label{eq:rl-ac-update-critic}
\frac{1}{2} |r_{t+1} + \gamma \vpi(x_{t+1}) - \vpi(x_{t})|^{2} .
\end{equation}
In practical implementations, $r_{t+1} + \gamma \vpi(x_{t+1})$ above is set as the target state value function and the gradient of \eqref{eq:rl-ac-update-critic} with respect to the parameter of $v^{\pi}$ is only applied to the second $\vpi$. Namley, the approximate gradient of \eqref{eq:rl-ac-update-critic} is set to $- (r_{t+1} + \gamma \vpi(x_{t+1}) - \vpi(x_{t})) \cdot \nabla_{\vpi} \vpi(x_{t})$ empirically in the AC algorithm.

\subsubsection*{Trust-Region Policy Optimization (TRPO) Algorithm}

While the actor-critic (AC) algorithm establish a new framework of alternately updating the policy (the actor) and its value function (the critic), its empirical computations often behave unstably. One of the major reasons is that the gradients with respect to the network parameters usually appear overly aggressive and it may need better restrains to mitigate the stability issue.

We recall that the policy gradient algorithm considered the gradient of the following function
\begin{equation*}
J(\pi) = \mathbb{E}_{Z_{T} \sim \rho_{T}^{\pi}}[R(Z_{T})]
\end{equation*}
with respect to the current policy $\pi$, where we also assume here that the sampled episode is infinitely long and the total reward is given by $R(z_{T}) = \sum_{t=0}^{\infty} \gamma^{t} r(x_{t},u_{t})$ for ease of discussion below. 
We also notice that this function can be written as
\begin{equation}
\label{eq:trpo-unmodified}
J(\pi) = \mathbb{E}_{X \sim p_{0}} [\vpi(X)] = \mathbb{E}_{X \sim p_{0}} \Big[ \mathbb{E}_{U \sim \pi(\cdot|X)} [\qpi (X, U)] \Big] .
\end{equation}
As we can see, the unstable computation of the policy gradient step is mainly because the estimation error in $\vpi$ or $\qpi$ when they are parameterized deep neural networks.
In this case, it can be expensive to use $\pi$ to generate new episodes as many risky sample actions may occur, raising the practical cost of data generations.

To overcome this issue, the trust-region policy optimization (TRPO) algorithm modifies the objective function in \eqref{eq:trpo-unmodified} by introducing a previously estimated policy $\pi_{k}$ (assuming it behaved well during the sample episode generation processes):
\begin{equation}
\label{eq:trpo-obj-modified}
J(\pi) = \mathbb{E}_{X \sim p_{0}} \Big[ \mathbb{E}_{U \sim \pi_{k}(\cdot|X)} \Big[ \frac{\pi(U|X)}{\pi_{k}(U|X)} q^{\pi}(X,U) \Big] \Big] .
\end{equation}
It is easy to verify that the new objective function \eqref{eq:trpo-obj-modified} is identical to the one in \eqref{eq:trpo-unmodified}.

As the TRPO algorithm requires that the gradient with respect to the network parameters of the policy $\pi$ should not be overly aggressive, it proposes to constrain the difference between the previous $\pi_{k}(\cdot|x)$ and the new $\pi(\cdot|x)$ at every $x \in \Xcal$, and assumes that the unknown Q-function $\qpi$ can be well approximated by the previously obtained $q^{\pi_{k}}$.
More precisely, the TRPO algorithm proposes the following scheme to update the policy $\pi$:
\begin{subequations}
\label{eq:trpo}
\begin{align}
\max_{\pi} \quad & \tilde{J}(\pi) := \mathbb{E}_{X \sim p_{0}} \Big[ \mathbb{E}_{U \sim \pi_{k}(\cdot|X)} \Big[ \frac{\pi(U|X)}{\pi_{k}(U|X)} q^{\pi_{k}}(X,U) \Big] \Big] , \label{eq:trpo-obj} \\
\text{s.t.}\quad & \mathbb{E}_{X \sim p_{0}} \Big[ \kl( \pi_{k}(\cdot|X), \, \pi(\cdot|X) ) \Big] \le \delta , \label{eq:trpo-kl}
\end{align}
\end{subequations}
where $\delta>0$ is some user-selected hyperparameter.
As we can see, this process produces a sequence of policies $\{\pi_{k}: k \ge 1\}$ and Q-functions $\{q^{\pi_{k}}: k \ge 1\}$. The advantage is that the adjacent policies are required to be close in the sense of KL divergence \eqref{eq:trpo-kl} in order to avoid unstable updates of the network parameters of $\pi_{k}$.

One can design many variants of the TRPO algorithms based on the idea presented above. For the original TRPO algorithm \eqref{eq:trpo}, the objective function in \eqref{eq:trpo-obj} can be approximated by an affine function of $\pi$, and \eqref{eq:trpo-kl} can be approximated as an inequality constraint on a quadratic function of the network parameters of $\pi$. With these approximations, the computational cost of \eqref{eq:trpo} can be considerably lower, but is still a bit expensive compared to some other variants we will introduce later.

As a final remark related to the TRPO algorithm and its variants, we note that an empirically promising approach to further stabilize the numerical computation is to introduce the so-called \emph{baseline} into the objective function.
The reason is that, while the objective function such as \eqref{eq:trpo-obj-modified} appears to be an unbiased estimate of the true objective function \eqref{eq:trpo-unmodified}, the sample variances are often large and can cause undesired convergence issues. 
The baseline function is designed to be another unbiased estimator of the true objective function, and when it is subtracted from \eqref{eq:trpo-obj-modified}, the estimate remains unbiased. 
Meanwhile, the goal of this baseline function is to reduce the sample variance and improve the empirical convergence stability of the TRPO algorithm and its variants.

There are multiple choices of designing the baseline function. One typical choice is to implement the policy network $\pi(\cdot|x)$ as a normal distribution given any $x \in \Xcal$, so that the KL divergence in \eqref{eq:trpo-kl} has closed form expression and is easier to handle. Meanwhile, we design the baseline function to be subtracted from \eqref{eq:trpo-obj} as
\begin{align*}
\mathbb{E}_{X \sim p_{0}}[v^{\pi_{k}}(X)]
& = \int p_{0}(x) v^{\pi_{k}}(x) \, dx \\
& = \int p_{0}(x) v^{\pi_{k}}(x) \Big( \int \pi(u|x) \, du \Big)\, dx \\
& = \iint p_{0}(x) \pi_{k}(u|x) \frac{\pi(u|x)}{\pi_{k}(u|x)} v^{\pi_{k}}(x) \, du dx \\
& = \mathbb{E}_{X \sim p_{0}} \Big[ \mathbb{E}_{U \sim \pi_{k}(\cdot|X)} \Big[ \frac{\pi(U|X)}{\pi_{k}(U|X)} v^{\pi_{k}}(X) \Big] \Big] ,
\end{align*}
which is also an unbiased estimate of $J(\pi)$ (assuming that $\vpi$ can be closely approximated by $v^{\pi_{k}}$).
As a result, the Q-function $q^{\pi_{k}}(X,U)$ in the objective function \eqref{eq:trpo-obj} is replaced with $q^{\pi_{k}}(X,U) - v^{\pi_{k}}(X)$. This is shown to reduce the computation instability issue in many numerical experiments.

\subsubsection*{Proximal Policy Optimization (PPO) Algorithm}

The proximal policy optimization (PPO)\index{PPO algorithm} algorithm is a followup variant of the TRPO algorithm above.
With some moderate changes to the TRPO, the PPO algorithm often presents stronger learning performance and faster numerical computations.
The idea of the PPO algorithm is to use a relaxed version of the Lagrangian method to handle the constraint \eqref{eq:trpo-kl}. By relaxed version it means that the Lagrangian multiplier is updated in a heuristic manner.

More specifically, the penalty version of the PPO algorithm modifies the scheme \eqref{eq:trpo} in the TRPO algorithm as an unconstrained problem:
\begin{equation*}
\tilde{J}(\pi) := \mathbb{E}_{X \sim p_{0}} \Big[ \mathbb{E}_{U \sim \pi_{k}(\cdot|X)} \Big[ \frac{\pi(U|X)}{\pi_{k}(U|X)} a^{\pi_{k}}(X,U) - \lambda \kl\Big(\pi_{k}(\cdot|X), \pi(\cdot|X)\Big) \Big] \Big] ,
\end{equation*}
where $a^{\pi_{k}}(x,u) := q^{\pi_{k}}(x,u) - v^{\pi_{k}}(x)$ is called the \emph{advantage function}\index{Advantage function} (notice that it can be either positive or negative, depending whether the action $u$ in $q^{\pi_{k}}$ brings over-average value compared to its expected value $v^{\pi_{k}}$), which is commonly used in many model-free reinforcement learning algorithms, $\lambda$ plays the role of the Lagrange (KKT) multiplier. 
In the PPO algorithm, $\lambda$ is simply updated to $\lambda/2$ if the KL divergence is smaller than $2\delta/3$; to $2\lambda$ if the KL is larger than $3\delta/2$; and remains unchanged otherwise, where $\delta>0$ is again a user-chosen hyperparameter.

Another simple yet effective version of the PPO algorithm is called the PPO-Clip. 
In PPO-Clip, the objective function is changed to the following one:
\begin{equation*}
\tilde{J}(\pi) := \mathbb{E}_{X \sim p_{0}} \Big[ \mathbb{E}_{U \sim \pi_{k}(\cdot|X)} \Big[ \min \Big(\xi_{k}(X,U) a^{\pi_{k}}(X,U), \, \bar{\xi}_{k}^{\pi}(X,U) a^{\pi_{k}}(X,U) \Big) \Big] \Big]
\end{equation*}
where the ratio $\xi_{k}$ is defined by
\begin{equation*}
\xi_{k}^{\pi}(x,u) : = \frac{\pi(U|X)}{\pi_{k}(U|X)} ,
\end{equation*}
and its clipped value $\bar{\xi}_{k}^{\pi}(x,u)$ is restricted to the interval $[1-\epsilon, 1+\epsilon]$:
\begin{equation*}
\bar{\xi}_{k}^{\pi}(x,u) : =
\begin{cases}
1-\epsilon, & \text{if}\ \xi_{k}^{\pi}(x,u) \le 1-\epsilon , \\
1+\epsilon, & \text{if}\ \xi_{k}^{\pi}(x,u) \ge 1+\epsilon , \\
\xi_{k}^{\pi}(x,u), & \text{otherwise} , \\
\end{cases}
\end{equation*}
for some user-chosen hyperparameter $\epsilon \in (0,1)$.
Notice that the clipped value is set in this way because the advantage function value $a^{\pi_{k}}$ can be either positive or negative.
This PPO-Clip version is simpler to implement than the relaxed Lagrangian approach and appears to be more very effective in many empirical tests.

\subsubsection*{Deep Deterministic Policy Gradient (DPG) Algorithm}

The deep deterministic policy gradient (DPG)\index{DPG algorithm} algorithm is an approach that avoids sampling actions in the learning process.
By assuming the policy to be greedy, the DPG algorithm can directly merge the policy network (the actor network) into the value network (the critic network) in the actor-critic algorithm.

Since greedy policies are all deterministic, the actor in the DPG algorithm can be set to a deep neural network $w: \Xcal \to \Ucal$.
In other words, the deterministic policy $\pi$ reduces to the following one-hot probability:
\begin{equation*}
\pi(u|x) :=
\begin{cases}
1, \text{ at }w(x)\ \text{where} \ w(x) = \argmax_{u \in \Ucal} q(x,u), \\
0, \text{ at every } u \in \Ucal \setminus \{w(x)\}.
\end{cases}
\end{equation*}
If there are multiple maximizers, then $w(x)$ can be set to any one of them.
Since the actor network $w$ always outputs a maximizer in $\Ucal$, the critic in the DPG algorithm can be set to the optimal Q-function $\qs$, which is also parameterized as a deep neural network defined on $\Xcal \times \Ucal$. 
Assuming the critic network $\qs$ is differentiable with respect to the argument $u \in \Ucal$, we have
\begin{equation*}
\partial_{u} \qs(x, w(x)) \nabla_{w} w(x) ,
\end{equation*}
where $\nabla_{w} w(x)$ should be interpreted as the gradient of the actor network $w$ with respect to its network parameters.

Notice that the use of the greedy policy $w(x)$ results in 
\begin{equation*}
\qs(x,w(x)) = \max_{\pi \in \Pi} \mathbb{E}_{U \sim \pi(\cdot|x)}[\qs(x,U)] = \vs(x) .
\end{equation*}
Therefore, we can define an objective function $J$ of $w$ to update the parameters of $w$ as follows:
\begin{equation*}
J(w) := \mathbb{E}_{X \sim p_{0}}[\vs(X)] = \mathbb{E}_{X \sim p_{0}}[\qs(X,w(X))].
\end{equation*}
We see that the gradient of $J(w)$ with respect to $w$ can be evaluated by
\begin{equation*}
\nabla_{w} J(w) = \nabla_{w} \mathbb{E}_{X \sim p_{0}}[ \qs(X, w(X))] = \mathbb{E}_{X \sim p_{0}}[ \partial_{u} \qs (X, w(X)) \nabla_{w} w(X)] .
\end{equation*}

The update of the optimal Q-function $\qs$ in the DPG algorithm is similar to the Q-learning algorithm. More specifically, we need to sample tuples of form $(x_{t}, u_{t}, r_{t+1}, x_{t+1})$ by letting the agent behave in the environment following the current estimated policy $\pi$, namely, $w$. Then for each sample tuple, we would like to minimize the error
\begin{equation*}
\frac{1}{2} | r_{t+1} + \gamma \qs(x_{t+1}, w(x_{t+1})) - \qs(x_{t},u_{t}) |^{2} .
\end{equation*}
% Again, to avoid overestimation of $\qs$ which often causes accumulated error i
In the training process, the DPG algorithm approximates the gradient of the quantity above with respect to the network parameters of $\qs$ as 
\begin{equation*}
-\Big( r_{t+1} + \gamma \qs(x_{t+1}, w(x_{t+1})) - \qs(x_{t},u_{t}) \Big) \cdot \nabla_{\qs} \qs(x_{t}, u_{t}) ,
\end{equation*}
where $\nabla_{\qs} \qs(x_{t}, u_{t})$ should be interpreted as the gradient of $\qs$ with respect to its network parameters at the given sample pair $(x_{t}, u_{t})$.
As we can see, the term $r_{t+1} + \gamma \qs(x_{t+1}, w(x_{t+1}))$ is considered as the target and the gradient $\nabla_{\qs}$ is not applied to the $\qs$ in this term.
In practice, the gradient above is averaged over all sampled tuples to update $\qs$.

There are also many variants of the DPG algorithms. For example, we can parameterize $\pi$ as some typical probability distribution, such as the normal distribution when $\Ucal$ is a continuous subset of some Euclidean space. This improves the the exploration ability of the learning process and may discover better actions compared to learning greedy actions only.

\subsubsection*{The Soft Actor-Critic (SAC) Algorithm}

The deep deterministic policy gradient (DPG) algorithm sometimes demonstrates problematic convergence behaviors since the gradients with respect to the network parameters of $\qs$ are often not accurate. 
One resolution to tackle this issue is adding a regularization to the current estimated policy $\pi$.
To this end, recall that the objective function of $\pi$ can be defined by 
\begin{equation*}
J(\pi) := \mathbb{E}_{X \sim p_{0}} \Big[ \mathbb{E}_{U \sim \pi(\cdot|X)} [q^{\pi}(X,U)] \Big] .
\end{equation*}
We want to add the weighted entropy function 
\begin{equation*}
\beta H(\pi(\cdot|X)) = -\beta \mathbb{E}_{U \sim \pi(\cdot|X)}[\log \pi(U|X)]
\end{equation*}
for every $X \sim p_{0}$, where $\beta>0$ is a user-chosen weight of $H$, to the objective function above. 
Then the objective function $J$ to be maximized with respect $\pi$ becomes
\begin{equation}
\label{eq:sac-Jpi}
J(\pi) = \mathbb{E}_{X \sim p_{0}} \Big[ \mathbb{E}_{U \sim \pi(\cdot|X)} [q^{\pi}(X,U) - \beta \log \pi(U|X)] \Big] .
\end{equation}
Theoretically, the maximizer of this objective function $J$ has a closed form
\begin{equation}
\label{eq:sac-pi}
\pi(u|x) = \frac{e^{q^{\pi}(x,u)/\beta}}{\int e^{q^{\pi}(x,u)/\beta} du}
\end{equation}
for every $x$. However, this closed form only works when $\Xcal$ and $\Ucal$ are of small to moderate sizes and their values $q^{\pi}(x,u)$ can be stored in a tabular.

For large problems, we need to parameterize the Q-function $q^{\pi}: \Xcal \times \Ucal \to \mathbb{R}$ as a deep neural network. 
For a sample tuple $(x_{t}, u_{t}, r_{t+1}, x_{t+1})$ in a generated episode, the following error is expected to be minimized:
\begin{equation}
\label{eq:sac-tuple-error}
\frac{1}{2} \Big| \Big(r_{t+1} + \gamma \mathbb{E}_{U_{t+1} \sim \pi(\cdot|x_{t+1})}q^{\pi}(x_{t+1}, U_{t+1})\Big) - \qpi(x_{t},u_{t}) \Big|^{2} .
\end{equation}
In practice, we need use the averaged errors over all sampled tuples of form $(x_{t}, u_{t}, r_{t+1}, x_{t+1})$, and take the gradient of the averaged errors with respect to the network parameters of $q^{\pi}$ to update it.

In the implementation of the SAC algorithm\index{SAC algorithm}, one should also consider the problem of overestimating $\qpi$. 
One remedy to mitigate this issue is using two deep neural networks $\qpi_{1}$ and $\qpi_{2}$, and choose the following value as the target (instead of the term in the big parenthesis within the square) in \eqref{eq:sac-tuple-error}:
\begin{equation}
\label{eq:sac-target}
r_{t+1} + \gamma \min_{j=1,2}\mathbb{E}_{U_{t+1} \sim \pi(\cdot|x_{t+1})} [\qpi_{j}(x_{t+1}, U_{t+1})] .
\end{equation}
The gradient of \eqref{eq:sac-tuple-error} (with the target set to \eqref{eq:sac-target}) with respect to the network parameters of $\qpi(x_{t},u_{t})$ is used in both $\qpi_{1}$ and $\qpi_{2}$ for the next iteration.

Although we can update the parameters of the Q-function network $\qpi$, it remains an issue to compute the maximizer of $J(\pi)$ in \eqref{eq:sac-Jpi}. The reason is that the normalizing constant in the closed form \eqref{eq:sac-pi} is not easy to compute as it may require an integration over a large action space $\Ucal$.
In such cases, we can employ the re-parameterization trick on $\pi$, such that $\pi(\cdot|x)$ is modeled as a normal distribution $\Ncal(\cdot; \mu_{\pi}(x), \sigma_{\pi}^{2} I_{m})$ when $\Ucal$ is a continuous subset of $\mathbb{R}^{m}$.
When $\Ucal$ is a large finite action space, we can also parameterize $\pi(\cdot|x)$ as a probability distribution on $\Ucal$ by setting $\pi$ as a deep neural network with the last operation as the softmax function at every $x \in \Xcal$, similar to what is done in the policy gradient algorithm above. In either case, we can evaluate $\nabla_{\pi} J(\pi)$, the gradient of $J$ with respect to the network parameters of $\pi$, provided the current estimated Q-function $\qpi$.

\section{Remarks and References}

\subsection{Reinforcement Learning and Optimal Control}

Reinforcement learning and optimal control are two closely related frameworks for sequential decision making. 
Both frameworks study how an agent should act over time in order to optimize specified cumulative and/or terminal performance objectives, and the methodologies are also extended to multi-agent cases. 
Historically, optimal control originates from control theory and applied mathematics, while reinforcement learning emerged from artificial intelligence and psychology. 
Despite their shared foundations, the two fields differ in modeling assumptions, solution methodologies, and areas of application.

In the standard optimal control setting, the control systems are typically modeled as controlled dynamical system described by an ODE. There are two major classes of optimal control problems: One is with fixed end-time $T$ but free endpoint $x(T)$, while the other one is fixed endpoint $x(T)$ but free end-time $T$, as we have shown in Section \ref{sec:oc-theory}.
In either case, the formulation of the ODE as well as the running and/or terminal rewards (or equivalently the costs) are known, the optimality conditions of the optimal control can be characterized by the Pontryagin Maximum Principle and the Hamilton--Jacobi--Bellman equation in continuous time \cite{bryson2018applied, bertsekas2012dynamic}.

Reinforcement learning, on the other hand, typically models decision making as a Markov decision process as we shown in Section \ref{sec:rl-fundamentals}. 
In reinforcement learning problems, the transition dynamics $p(x'|x,u)$ and the reward function can be unknown, and they need to learned during the solution process in finding the optimal control (called optimal policy in reinforcement learning).
For these problems, a class of different methods, typically model-free algorithms, are widely used.
We have shown some representative and influential algorithms in Section \ref{sec:rl-alg}.

It is known that the theory of reinforcement learning and optimal control are strongly connected. For instance, when the system dynamics are known and the state and action spaces are continuous, reinforcement learning reduces to stochastic optimal control. %
In this case, dynamic programming provides the unifying framework for both fields and Bellman theory underpins the optimality conditions and algorithmic developments in both fields \cite{bertsekas1995neuro-dynamic}.
Moreover, the value function in reinforcement learning is directly analogous to its counterpart in optimal control, and the policy gradient methods have strong connections to optimal control via Pontryagin Maximum Principle and trajectory optimization. The literature has shown equivalences between certain reinforcement learning objectives and stochastic control formulations, including entropy-regularized control and path integral control \cite{kappen2005path, todorov2006linearly-solvable}.

Despite the connections above, important differences between reinforcement learning and optimal control remain.
For instance, as we mentioned, optimal control often assumes known dynamics, while reinforcement learning emphasizes learning from data with minimal prior knowledge.
As a result, solution quality of optimal control problems heavily rely on accurate models, whereas reinforcement learning algorithms often need to trade off between computation efficiency and data (both provided and generated) quality.
Moreover, exploration is central to reinforcement learning but largely absent in the standard optimal control where uncertainty is handled via robust or stochastic formulations.

Nevertheless, the boundary between reinforcement learning and optimal control has started to blur after recent research advances \cite{recht2019tour}. 
In particular, deep learning has become a powerful tool to tackle large-scale and high-dimensional problems in both fields. 
The common issues raised by using deep neural networks, such as stability, robustness, and generalizability, may be addressed by further theoretical and algorithmic developments. 
To that end, understanding their relationship can provide valuable insights for designing efficient and reliable decision-making systems.

\subsection{References of Reinforcement Learning}

A comprehensive introduction to reinforcement learning is presented in \cite{sutton2018reinforcement}. 
The work \cite{szepesvari2022algorithms} provides a systematic review of classic reinforcement learning algorithms. 
Section \ref{sec:rl-fundamentals} largely follows the line of theory part presented in \cite{szepesvari2022algorithms}, with significant expansions on the proof details and new main results through Lemma \ref{lem:rl-v-bar-v-star} and Theorem \ref{thm:rl-equiv} which clarify the relations between $\bar{v}$ and the fixed point $v^{*}$ of the optimal Bellman operator $T^{*}$. 
Reinforcement learning and optimal control as well as their connections are studied in depth in \cite{bertsekas2019reinforcement}. 
Deep reinforcement learning has emerged in the past decade. Recent surveys can be found in \cite{arulkumaran2017deep,francois-lavet2018introduction,wang2022deep}.

Classic reinforcement algorithms, such as policy iteration, value iteration, and temporal difference method, are discussed in detail in \cite{sutton2018reinforcement, szepesvari2022algorithms}. 
The Q-learning method was developed in \cite{watkins1989learning} and showed to be convergent to a fixed point of the Bellman operator in \cite{watkins1992q-learning}.
The connection between dynamic programming and reinforcement learning was first shown 
in \cite{watkins1989learning}. 
The SARSA algorithm was developed by \cite{rummery1994on-line}. Early work on Monte Carlo methods was presented in \cite{singh1996reinforcement}. 
The original policy gradient method \cite{sutton1999policy} and actor-critic method \cite{konda1999actor-critic} were proposed much earlier than deep learning, but they have substantial influences to the developments of modern reinforcement learning algorithms. 
The deep Q-learning algorithm pioneering deep reinforcement learning was shown to achieve human-level performance across a benchmark of Atari games \cite{mnih2015human-level}. 
Another breakthrough demonstrated by deep reinforcement learning around the same time is the AlphaGo algorithm \cite{silver2016mastering}, which beat the world champion at Go, a classic board game considered to be impossible for computer to win human masters before deep learning emerged. 
Deep reinforcement learning was also employed to faster ways to multiply matrices using fewer multiplication operations \cite{fawzi2022discovering}.

There are substantial amount of works conducted for improved deep reinforcement learning algorithms in the past years. 
The dueling Q-network architecture was proposed in \cite{wang2016dueling}, where two heads of the same network predict 
the state value function and the advantage function.
The standard Q-learning shows systematic overestimation of the state values due to the max operation. 
The double Q-network architecture was proposed to remedy this overestimation issue \cite{van-hasselt2016deep}.
In \cite{fortunato2018noisy}, the noisy deep Q-network is introduced to add noise for stochasticity and encourage exploration, and the network can learn to decrease the noise over time as it converges. 
Distributional deep Q-network architecture was proposed in \cite{bellemare2017distributional,dabney2018distributional} aims to learn more complete information about the distribution of reward than just the expectation. 
A deep network architecture combining several aforementioned Q-network features was developed to improve both the training speed and the final performance on the ATARI benchmark \cite{hessel2018rainbow:}.
The prioritized experience replay method to improve training stability and speed was developed in \cite{schaul2016prioritized}.
Extensions to advantage functions were also developed in \cite{schulman2015high-dimensional}. 
These methods are variants of integrating deep learning into approximation of value functions, including state value functions, Q-functions (state-action value functions) and advantage functions.

Deep neural networks were also employed to learn optimal policies of reinforcement learning problems. The trust region policy optimization (TRPO) algorithm was developed in \cite{schulman2015trust}. 
The sequential work called the proximal policy optimization (PPO) algorithm was proposed in \cite{schulman2017proximal}. 
In \cite{silver2014deterministic}, the deep deterministic policy gradient (DDPG) algorithm was developed.
An actor-critic model-free algorithm based on the DDPG that can operate over continuous action spaces was introduced in \cite{lillicrap2015continuous}.
Soft actor-critic algorithms were developed in \cite{haarnoja2018soft,haarnoja2017reinforcement}.
%
% We discussed the main ideas of most of these algorithms in Section \ref{sec:rl-alg}. 
%
This field is undergoing very rapid research advancements.
More powerful, stable, and robust deep reinforcement learning algorithms are expected to developed in near future.

\chapter{Generative Models}
\label{chpt:gm}

This chapter introduces \emph{generative models}, an important class of deep learning methods that have emerged over the past decade for modeling complex data distributions and synthesizing new samples. They have vast real-world applications and can be used to generate realistic texts, music, images, videos and so on, called as artificial intelligence generated content (AIGC).

The prototype mathematical problem underlying generative modeling can be formulated as follows. Suppose we are given a dataset $\Dcal = \{x_{i} \in \mathbb{R}^{d}: i \in [M]\}$ consisting of $M$ samples independently drawn from some unknown probability distribution $\pdata$ on $\mathbb{R}^d$. The objective of generative modeling is to learn how to generate additional samples that follow the same distribution $\pdata$.

This task is challenging for several reasons. First, only the finite dataset $\Dcal$ is available, while the analytical form of $\pdata$ is unknown. Second, the dimension $d$ can be extremely large. For example, each $x_i$ may represent an image, in which case $d$ often reaches $10^4$ or higher.

The problem becomes even more involved in the setting of conditional generation, where the goal is to generate samples from an unknown conditional distribution $\pdata(\cdot \,|\, y)$ for any user-specified conditioning variable (often called a \emph{prompt}) $y$. The purpose is to generate samples with properties aligned with the need or preference of the user.

In this chapter, we focus primarily on unconditional generative models, as their theoretical foundations and methodological developments are more mature. Moreover, many existing conditional generative models build directly upon the theoretical insights and computational techniques originally developed for the unconditional setting.

Generative modeling has undergone rapid advances in recent years, yet there is still no unified framework that fully resolves the problem. This chapter presents several of the most influential generative modeling paradigms in the literature and examines their constructions and properties in detail. We begin with the classical variational autoencoder framework and the closely related generative adversarial networks approach, which emerged around the same time. We then discuss more recent developments, including diffusion models, probability density control, and flow matching methods, which leverage stochastic or deterministic differential equations for sample generation. Finally, we address practical and theoretical issues associated with these models and briefly discuss extensions and modifications for conditional generation.

\section{Variational Autoencoder}
\label{sec:vae}

Variational autoencoder (VAE)\index{Variational autoencoder} is a network architecture that can generate new samples of $\pdata$ by using a dataset $\Dcal$ consisting of finitely many of its independently drawn samples.
The VAE model consists of two functions, called the encoder and the decoder. The encoder and decoder are typically parameterized as deep neural networks, denoted by $\geta$ and $\ftheta$ with network parameters $\eta$ and $\theta$, respectively.

The \emph{encoder}\index{Variational autoencoder!encoder} $\geta$ learns to map every data sample $x \in \Dcal$ to a \emph{latent variable} $z$, which lies in the \emph{latent space} $\mathbb{R}^{m}$. 
Instead of directly determining $z$ with a given input $x$, the output $\geta(x)$ is often set to be the parameters of some probability distribution.
The typical choice is that $\geta(x)$ is a pair of a vector $\mu_{\eta}(x) \in \mathbb{R}^{m}$ and a positive scalar $\sigma_{\eta}(x) > 0$ (or a positive vector in $\mathbb{R}^{m}$ as the diagonal of a diagonal covariance matrix, or even directly an $m$-by-$m$ symmetric positive definite matrix as a covariance matrix, but we will only consider the simple case with scalar-valued $\sigma_{\eta}$ given that the generalizations are straightforward).
Then $\mu_{\eta}(x)$ and $\sigma_{\eta}(x)$ are set to be the mean and standard deviation, respectively, of a Gaussian distribution, from which the latent variable $z$ is drawn.
More precisely, $z$ is a sample drawn from the conditional probability distribution $q_{\eta}(\,\cdot\,|\,x) := \Ncal(\,\cdot\,;\, \mu_{\eta}(x), \sigma_{\eta}(x)^{2} I_{m})$.
This can be easily implemented as $z = \mu_{\eta}(x) + \sigma_{\eta}(x) \epsilon$ with $\epsilon \sim \Ncal(0,I_{m})$ in practice.

The \emph{decoder}\index{Variational autoencoder!decoder} $\ftheta$ aims at mapping any $z \in \mathbb{R}^{m}$ to a new sample $x' \in \mathbb{R}^{d}$.
Again, instead of directly determining $x'$ based on $z$, the output $\ftheta(z)$ for each input $z$ is set to be a pair of a vector $\mu_{\theta}(z)\in \mathbb{R}^{d}$ and a positive scalar $\sigma_{\theta}(z) > 0$. Then the new sample $x'$ is set to be a sample drawn from the conditional probability distribution $p_{\theta}(\cdot|z) := \Ncal(\,\cdot\,;\, \mu_{\theta}(z), \sigma_{\theta}(z)^{2}I_{d})$. 
This is implemented as $x' = \mu_{\theta}(z) + \sigma_{\theta}(z) \epsilon$ with $\epsilon \sim \Ncal(0,I_{d})$ in practice.

One additional requirement on the encoder $\geta$ is that the marginal distribution 
\begin{equation*}
q_{\eta}(z) := \int q_{\eta}(z|x) \pdata(x) \, dx
\end{equation*}
is approximately the standard Gaussian $\Ncal(0,I_{m})$, such that, after $\geta$ and $\ftheta$ are trained, we can easily draw samples from this marginal $\qeta$ and apply the decoder $\ftheta$ on them to obtain new samples following $\pdata$.

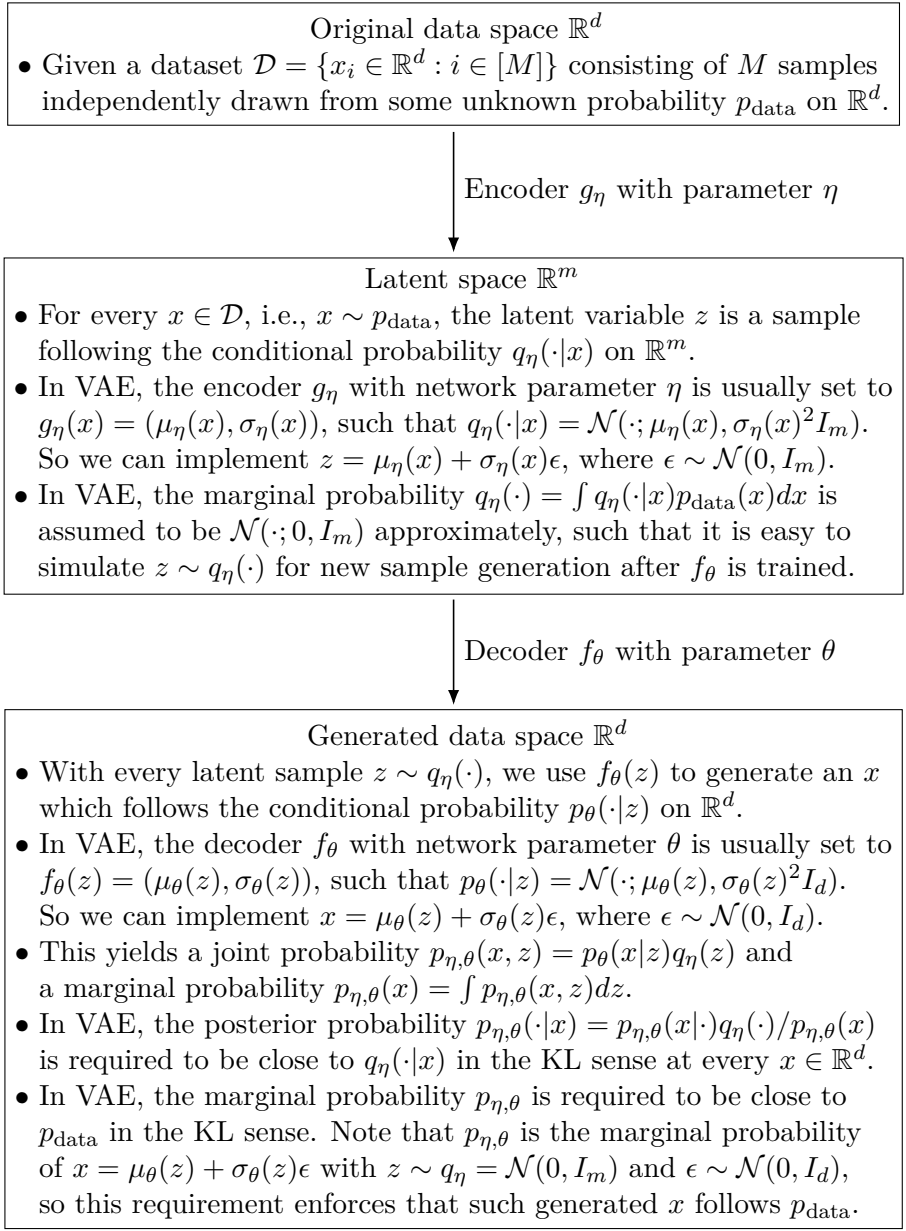
\begin{figure}
\centering
\begin{tikzpicture}[scale=1, transform shape, node distance=1cm, >=Latex]

\node at (0,12) [draw, align=left] {\qquad \qquad \qquad \qquad \qquad Original data space $\mathbb{R}^{d}$ \\ 
$\bullet$ Given a dataset $\Dcal=\{x_{i} \in \mathbb{R}^{d}: i\in [M]\}$ consisting of $M$ samples
\\ \ \, independently drawn from some unknown probability $\pdata$ on $\mathbb{R}^{d}$.
};

\draw[-{Latex[length=2mm]}, line width=.6pt] (0, 11.1) -- (0,9.5) node[right, midway] {Encoder $\geta$ with parameter $\eta$};

\node at (0,7.2) [draw, align=left] {\qquad \qquad \qquad \qquad \qquad \qquad Latent space $\mathbb{R}^{m}$ \\ 
$\bullet$ For every $x \in \Dcal$, i.e., $x \sim \pdata$, the latent variable $z$ is a sample 
\\ \ \, following the conditional probability $\qeta(\cdot|x)$ on $\mathbb{R}^{m}$. \\
$\bullet$ In VAE, the encoder $\geta$ with network parameter $\eta$ is usually set to 
\\ \ \, $\geta(x) = (\mu_{\eta}(x), \sigma_{\eta}(x))$, such that $\qeta(\cdot|x) = \Ncal(\cdot; \mu_{\eta}(x), \sigma_{\eta}(x)^{2}I_{m})$.
\\ \ \, So we can implement $z = \mu_{\eta}(x) + \sigma_{\eta}(x) \epsilon$, where $\epsilon \sim \Ncal(0,I_{m})$. \\
$\bullet$ In VAE, the marginal probability $\qeta(\cdot)= \int \qeta(\cdot|x)\pdata(x) dx$ is 
\\ \ \, assumed to be $\Ncal(\cdot; 0, I_{m})$ approximately, such that it is easy to
\\ \ \, simulate $z \sim \qeta(\cdot)$ for new sample generation after $\ftheta$ is trained.
};

\draw[-{Latex[length=2mm]}, line width=.6pt] (0, 4.9) -- (0,3.5) node[right, midway] {Decoder $\ftheta$ with parameter $\theta$};

\node at (0,0) [draw, align=left] {\qquad \qquad \qquad \qquad \qquad Generated data space $\mathbb{R}^{d}$ \\ 
$\bullet$ With every latent sample $z \sim \qeta(\cdot)$, we use $\ftheta(z)$ to generate an $x$ 
\\ \ \, which follows the conditional probability $\ptheta(\cdot|z)$ on $\mathbb{R}^{d}$. \\
$\bullet$ In VAE, the decoder $\ftheta$ with network parameter $\theta$ is usually set to 
\\ \ \, $\ftheta(z) = (\mu_{\theta}(z), \sigma_{\theta}(z))$, such that $\ptheta(\cdot|z) = \Ncal(\cdot; \mu_{\theta}(z), \sigma_{\theta}(z)^{2}I_{d})$. 
\\ \ \, So we can implement $x = \mu_{\theta}(z) + \sigma_{\theta}(z) \epsilon$, where $\epsilon \sim \Ncal(0,I_{d})$. \\
$\bullet$ This yields a joint probability $\pet(x,z)= \ptheta(x|z) \qeta(z)$ and 
\\ \ \, 
a marginal probability $\pet(x) = \int \pet(x,z) dz$. \\
$\bullet$ In VAE, the posterior probability $\pet(\cdot|x)= \pet(x|\cdot)\qeta(\cdot)/\pet(x)$
\\ \ \, is required to be close to $\qeta(\cdot|x)$ in the KL sense at every $x \in \mathbb{R}^{d}$. \\
$\bullet$ In VAE, the marginal probability $\pet$ is required to be close to
\\ \ \, $\pdata$ in the KL sense. Note that $\pet$ is the marginal probability
\\ \ \, of $x= \mu_{\theta}(z) + \sigma_{\theta}(z) \epsilon$ with $z \sim \qeta = \Ncal(0,I_{m})$ and $\epsilon \sim \Ncal(0,I_{d})$,
\\ \ \, so this requirement enforces that such generated $x$ follows $\pdata$.
};

\end{tikzpicture}
\caption{Spaces, mappings (encoder and decoder), and probability distributions in the variational autoencoder (VAE) framework.}
\label{fig:vae}
\end{figure}

Now we have several probability distributions in hand: The data probability distribution $\pdata$ (which we do not know the analytic form but only a number of its i.i.d.\ samples) on $\mathbb{R}^{d}$, the conditional distribution $q_{\eta}(\cdot|x)$ and the marginal distribution $q_{\eta}(\cdot)$ on the latent space $\mathbb{R}^{m}$, and the conditional probability $p_{\theta}(\cdot|z)$ on $\mathbb{R}^{d}$. As a result, we also have the marginal distribution $\pet(\cdot)$ on $\mathbb{R}^{d}$:
\begin{equation}
\label{eq:vae-pet}
\pet (x) = \int p_{\theta}(x|z) q_{\eta}(z)  \, dz , \quad \forall\, x \in \mathbb{R}^{d} ,
\end{equation}
where $\pet(x,z):=\ptheta(x|z)\qeta(z)$ is the joint distribution of $(x,z)$.
By the Bayes' rule, we can also obtain the posterior distribution $\pet(z|x)$ for any $x \in \mathbb{R}^{d}$:
\begin{equation}
\label{eq:vae-p-posterior}
\pet(z|x) = \frac{p_{\theta}(x|z) q_{\eta}(z)}{\pet(x)} , \quad \forall \, z \in \mathbb{R}^{m} .
\end{equation}

Setting the conditional distributions above as Gaussians not introduces diversities (compared to using deterministic mappings) and makes practical implementation convenient. For example, the Kullback--Leibler (KL) divergence between two Gaussians have closed form and the loss function for network training can be implemented easily. We will show the details later in this section.

Other than the settings above, we also need the following two important agreements between the probability distributions induced by the encoder and the decoder:
\begin{itemize}
\item
The marginal probability distribution of the generated samples should be equal to the probability distribution of the original data. Namely, 
\begin{equation}
\label{eq:vae-marginal-agree}
\pet(x) = \pdata(x)
\end{equation}
for all $x \in \mathbb{R}^{d}$. % This ensures the generated samples also follow $\pdata$.

\item
For any $x \in \mathbb{R}^{d}$, the posterior distribution of the latent variable $z$ given $x$ pretending that $x$ is a newly generated sample, should match the conditional probability of the latent variable $z$ given the same $x$ as if $x$ is one of the given samples following $\pdata$. Namely, for any $x \in \mathbb{R}^{d}$, 
\begin{equation}
\label{eq:vae-posterior-agree}
\pet(z | x) = \qeta( z | x )
\end{equation}
for all $z \in \mathbb{R}^{m}$. % This ensures that the posterior distribution of the latent variable $z$ given $x$, where $x$ is either a true sample or a generated sample, should be the same.
\end{itemize}

In the VAE framework, both \eqref{eq:vae-marginal-agree} and \eqref{eq:vae-posterior-agree} are realized by minimizing the KL divergences between the two sides.

The relationships between the data space $\mathbb{R}^{d}$, the latent space $\mathbb{R}^{m}$, the encoder $\geta$ and decoder $\ftheta$, as well as all the involved probability distributions discussed above, are summarized in Figure \ref{fig:vae}.

Now we begin to derive the mathematical formulations needed to formulate the objective function of the encoder network parameter $\eta$ and decoder network parameter $\theta$ using the aforementioned KL divergences.
More precisely, to obtain \eqref{eq:vae-marginal-agree}, we want to minimize
\begin{align}
\kl ( \pdata, \pet ) 
& = \int \pdata (x) \log \frac{\pdata(x)}{ \pet(x)} \, dx \nonumber \\
& = - H(\pdata) - \int \pdata (x) \log \pet(x) \, dx \label{eq:vae-marginal-kl} \\
& = - H(\pdata) - \mathbb{E}_{X \sim \pdata} [\log \pet(X)] , \nonumber
\end{align}
where $H(\pdata)$ is the entropy of $\pdata$ and independent of $\theta$ and $\eta$.

Therefore, if we want to minimize $\kl ( \pdata, \pet ) $ with respect to $(\theta,\eta)$, we only need to solve
\begin{equation}
\label{eq:vae-expect-evidence}
\max_{\theta, \eta} \ \mathbb{E}_{X \sim \pdata} [\log \pet(X)] .
\end{equation}
The value $\log \pet(x)$ is called the \emph{evidence}\index{Evidence} at $x$.
The goal of \eqref{eq:vae-expect-evidence} is to maximize the expectation of all evidence values of the given samples following $\pdata$.
We will take a closer look of the evidence and show its connection to the two probability distributions given in \eqref{eq:vae-posterior-agree} soon later.

To obtain the agreement between the two probability distributions in \eqref{eq:vae-posterior-agree}, we also try to minimize their KL divergence. That is, for every $x \in \mathbb{R}^{d}$, we want to minimize
\begin{equation}
\label{eq:vae-posterior-kl}
\kl ( \qeta ( \cdot | x), \ \pet ( \cdot | x) ) = \int \qeta(z | x) \log \frac{\qeta(z|x)}{\pet(z|x)} \, dz
\end{equation}
with respect to $\theta$ and $\eta$. If we want to minimize this KL over all samples following $\pdata$, we need to solve 
\begin{equation}
\label{eq:vae-posterior-kl-min}
\min_{\theta,\eta} \ \mathbb{E}_{X \sim \pdata} [\kl ( \qeta ( \cdot | X), \ \pet ( \cdot | X) ) ].
\end{equation}

Now we have two problems, i.e., \eqref{eq:vae-expect-evidence} and \eqref{eq:vae-posterior-kl-min}, to solve. However, they are closely related and can be merged into one problem.

To see this, we need to investigate the relationship between the evidence $\log \pet(x)$ and $\kl ( \qeta ( \cdot | x), \ \pet ( \cdot | x) )$, the two functions whose expectations are taken in \eqref{eq:vae-expect-evidence} and \eqref{eq:vae-posterior-kl-min}: for every $x \in \mathbb{R}^{d}$, there is
\begin{align}
\kl( & \qeta(\cdot |x), \ \pet(\cdot |x) ) \nonumber \\
& = \int \qeta(z|x) \log \frac{\qeta(z|x)}{ \pet(z|x)} \, dz \nonumber \\
& = \int \qeta(z|x) \log \Big(\frac{\qeta(z|x)}{\qeta(z)} \frac{\pet(x)}{\ptheta(x|z)} \Big) \, dz \label{eq:vae-kl-evidence} \\
& = \int \qeta(z|x) \log \frac{\qeta(z|x)}{\qeta(z)} \, dz - \int \qeta(z|x) \log \ptheta(x|z) \, dz + \log \pet(x) \nonumber \\
& = \kl( \qeta(\cdot |x), \ \qeta ) - \int \qeta(z|x) \log \ptheta(x|z) \, dz + \log \pet(x) \nonumber ,
\end{align}
where the second equality is due to the Bayes' rule
\begin{equation*}
\pet (z |x) = \frac{\ptheta(x|z) \qeta(z)}{\pet(x)} .
\end{equation*}
Rearranging \eqref{eq:vae-kl-evidence} yields
\begin{align}
\log \pet(x) & = \kl( \qeta(\cdot|x), \ \pet(\cdot|x) ) \nonumber \\
& \qquad + 
\underbrace{\int \qeta(z|x) \log \ptheta(x|z) \,dz - \kl( \qeta(\cdot|x), \ \qeta(\cdot) )}_{=: \, \elbo_{\eta,\theta}(x)} . \label{eq:vae-evidence-kl-elbo}
\end{align}
Therefore, the evidence $\log \pet(x)$ is just the sum of $\kl( \qeta(\cdot|x), \ \pet(\cdot|x) )$ and the so-called \emph{evidence lower bound}\index{Evidence lower bound (ELBO)} (ELBO) defined by
\begin{equation}
\label{eq:elbo}
\elbo_{\eta,\theta}(x) := \int \qeta(z|x) \log \ptheta(x|z) \,dz - \kl (\qeta(\cdot|x), \ \qeta(\cdot) ) .
\end{equation}
We notice that $\elbo_{\eta,\theta}(x)  \le \log \pet(x)$ at any $x$, because according to \eqref{eq:vae-evidence-kl-elbo} their difference is $\kl( \qeta(\cdot|x), \ \pet(\cdot|x) )$, which is always non-negative.

ELBO originally got its name because it was often derived in the following way in the literature: for any $x \in \mathbb{R}^{d}$, there is
\begin{align*}
\log \pet(x) 
& = \log \int \pet(x,z) \, dz \\
& = \log \int \frac{\pet(x,z)}{\qeta(z|x)} \qeta(z|x) \, dz \\
& \ge \int \qeta(z|x) \log \frac{\pet(x,z)}{\qeta(z|x)} \, dz \\
& = \int \qeta(z|x) \log \Big( \frac{\qeta(z)}{\qeta(z|x)} \ptheta(x|z) \Big) \, dz \\
& = \int \qeta(z|x) \log \ptheta(x|z) \,dz - \kl( \qeta(\cdot|x), \ \qeta(\cdot) ) \\
& = \elbo_{\eta,\theta}(x) ,
\end{align*}
where the inequality is due to Jensen's inequality as logarithm is a concave function, and the third equality is due to $\pet(x,z) = \ptheta(x|z) \qeta(z)$.
Therefore, if we want to maximize the evidence $\log \pet(x)$, we can try to maximize $\elbo_{\eta,\theta}(x)$.

Now we see that \eqref{eq:vae-evidence-kl-elbo} also provides an additional interpretation of the ELBO: Maximizing $\elbo_{\eta,\theta}(x)$ will not only increase the evidence $\log \pet(x)$, but also reduce $\kl( \qeta(\cdot|x), \pet(\cdot|x) )$ for every $x$.
More precisely, taking expectations on both sides of \eqref{eq:vae-evidence-kl-elbo}, we have
\begin{align*}
\mathbb{E}_{X \sim \pdata}[ \elbo_{\eta,\theta}(X) ]
& = \mathbb{E}_{X \sim \pdata}[\log \pet(X)] \\
& \qquad - \mathbb{E}_{X \sim \pdata}[\kl( \qeta(\cdot|X), \ \pet(\cdot|X) )] .
\end{align*}
Therefore, the maximization problem \eqref{eq:vae-expect-evidence} and the minimization problem \eqref{eq:vae-posterior-kl-min}, can be achieved simultaneously by the same means---solving the ELBO maximization problem:
\begin{equation}
\label{eq:max-elbo}
\max_{\theta,\eta} \ \mathbb{E}_{X \sim \pdata}[ \elbo_{\eta,\theta}(X) ] .
\end{equation}

In the remainder of this section, we will look into the structure of $\elbo_{\eta,\theta}$ and write \eqref{eq:max-elbo} into an implementable optimization problem using the given data $\Dcal$.

Recall the definition of ELBO in \eqref{eq:elbo}, we know that the objective function in \eqref{eq:max-elbo} can be written as
\begin{align}
\mathbb{E}_{X \sim \pdata}[ \elbo_{\eta,\theta}(X) ]
& = \underbrace{\iint \pdata(x) \qeta(z|x) \log \ptheta(x|z) \, dz dx }_{=:\, A(\theta,\eta)} \label{eq:max-elbo-AB} \\
& \qquad - \underbrace{\int \pdata(x) \kl ( \qeta(\cdot|x), \qeta(\cdot)) \, dx }_{=:\, B(\eta)} \nonumber
\end{align}

Due to the setting of the encoder $g_{\eta}$, we have
\begin{equation}
\qeta(z | x) = \Ncal(z; \mu_{\eta} (x), \sigma_{\eta}(x)^{2} I_{m}) .
\end{equation}
Therefore, we can sample $\epsilon$ from the standard Gaussian $\Ncal(0, I_{m})$, and set 
\begin{equation}
\label{eq:elbo-A-z}
z_{\eta}(x, \epsilon) = \mu_{\eta}(x) + \sigma_{\eta}(x) \epsilon .
\end{equation}
Then we know $z_{\eta}(x, \epsilon)$ is a sample drawn from the probability $\qeta( \cdot | x)$. 
Furthermore, for any $z \in \mathbb{R}^{m}$, by the setting of the decoder $f_{\theta}$, we have
\begin{equation}
\label{eq:elbo-A-logp}
p_{\theta} (x|z) = \Ncal(x; \mu_{\theta}(z), \sigma_{\theta}(z)^{2} I_{d} ).
\end{equation}
Hence, there is
\begin{equation*}
\log p_{\theta}(x|z) = - \frac{|x - \mu_{\theta}(z)|^{2}}{2 \sigma_{\theta}(z)^{2}} - \frac{d}{2} \log \big( 2\pi \sigma_{\theta}(z)^{2d} \big) .
\end{equation*}
Now we can rewrite $A(\theta,\eta)$ defined in \eqref{eq:max-elbo-AB} as
\begin{equation}
\label{eq:elbo-A}
A(\theta, \eta) = \iint \pdata(x) \Ncal(\epsilon; 0, I_{m}) \log p_{\theta}(x | z_{\eta}(x,\epsilon)) \, d\epsilon dx .
\end{equation}
where $z_{\eta}(x,\epsilon)$ and $\log p(x|z)$ are given in \eqref{eq:elbo-A-z} and \eqref{eq:elbo-A-logp}, respectively.
In practical implementations, we can approximate $A(\theta, \eta)$ in \eqref{eq:elbo-A} using Monte Carlo integration as we are only given some samples drawn from $\pdata$: For every sample $x \in \Dcal$ which consists of $M$ independent samples drawn from $\pdata$, we sample an $\epsilon \sim \Ncal(0, I_{d})$, and approximate $A(\theta,\eta)$ in \eqref{eq:elbo-A} using
\begin{equation*}
\frac{1}{M}\sum_{(x,\epsilon)} \Ncal(\epsilon; 0, I_{m}) \log p_{\theta}(x | z_{\eta}(x,\epsilon)) ,
\end{equation*}
where the sum is over $M$ pairs of such $(x,\epsilon)$ sample pairs.

Next, we see that $\kl ( \qeta(\cdot|x), \qeta(\cdot))$ in the integrand of $B(\eta)$ defined in \eqref{eq:max-elbo-AB} has a closed form because both $\qeta(\cdot|x)$ and $\qeta(\cdot)$ are Gaussian distributions.
In particular, there are
\begin{equation*}
\qeta(z|x) = \Ncal(z; \mu_{\eta}(x), \sigma_{\eta}(x)^{2} I_{m} )
\end{equation*}
and
\begin{equation*}
\qeta(z) = \Ncal(z; 0, I_{m}) .
\end{equation*}
Therefore, we have the closed form of their KL divergence (the KL divergence between two Gaussian distributions of the same dimension is given in \eqref{eq:kl-gauss}):
\begin{equation*}
\kl ( \qeta(\cdot|x), \qeta(\cdot)) = \frac{1}{2} \Big( m \sigma_{\eta}(x)^{2} + |\mu_{\eta}(x)|^{2} - m - 2m \log \sigma_{\eta}(x) \Big) .
\end{equation*}
Hence, there is
\begin{equation}
\label{eq:elbo-B}
B(\eta) = \frac{1}{2} \int \pdata(x) \Big( m \sigma_{\eta}(x)^{2} + |\mu_{\eta}(x)|^{2} - m - 2m \log \sigma_{\eta}(x) \Big) \, dx ,
\end{equation}
which can again be approximated by the Monte Carlo integration 
\begin{equation*}
\frac{1}{2} \sum_{x \in \Dcal} \Big( m \sigma_{\eta}(x)^{2} + |\mu_{\eta}(x)|^{2} - m - 2m \log \sigma_{\eta}(x) \Big) .
\end{equation*}

Now we have the expressions \eqref{eq:elbo-A} and \eqref{eq:elbo-B} of $A(\theta,\eta)$ and $B(\eta)$, respectively, as well as their implementable forms, we can maximize the expectation of ELBO defined in \eqref{eq:max-elbo-AB} with respect to $(\theta,\eta)$.

After $(\theta,\eta)$ are optimized, we can generate new samples from the unknown probability distribution $\pdata$ as follows.
We first simulate $z \sim \Ncal(0,I_{m})$ and $\epsilon \sim \Ncal(0, I_{d})$, and set $x = \mu_{\theta}(z) + \sigma_{\theta}(z) \epsilon$, then $x$ is a new sample following the probability $\pdata$.
We can simulate any number of pairs $(z,\epsilon)$ independently and simultantenously as such and compute the corresponding $x$'s, which are all new samples following the probability distribution $\pdata$.

\section{Generative Adversarial Networks}
\label{sec:gan}

Generative adversarial networks (GANs)\index{Neural network!generative} is a deep network architecture developed around the same time as the variational autoencoder framework.
The idea of GANs is straightforward and simple.
Let $\Dcal = \{x_{i} \in \mathbb{R}^{d} : i \in [M] \}$ be a given dataset consisting of $M$ samples independently drawn from some unknown probability distribution $\pdata$ on $\mathbb{R}^{d}$ as before.
The idea of GANs is to construct a \emph{generator}\index{Generator} function $g_{\theta}: \mathbb{R}^{d} \to \mathbb{R}^{d}$, parameterized as a deep neural network with parameter $\theta$, such that $g_{\theta}(z)$ is a sample following the probability $\pdata$ for any $ z \sim \Ncal(0,I_{d})$.

Now the question is how to determine the parameter $\theta$ such that $g_{\theta}(z)$ indeed follows the unknown $\pdata$.
To this end, we notice that we have the real samples from $\Dcal$ and generated samples $g_{\theta}(z)$ with $z \sim \Ncal(0,I_{d})$. 
Therefore, we can assign binary labels to these samples, i.e., we assign $1$ to every sample from $\Dcal$ and $0$ to every generated (fake) sample $g_{\theta}(z)$.
More specifically, we denote
\begin{equation}
\label{eq:gan-real-data}
\Dcal_{\text{real}} : = \Big\{ (x_{i}, 1): x_{i} \in \Dcal, \ i \in [M] \Big\}
\end{equation}
and
\begin{equation}
\label{eq:gan-fake-data}
\Dcal_{\text{fake}} : = \Big\{ (g_{\theta}(z_{j}), 0) : z_{j} \sim \Ncal(0,I_{d}), \ j \in [N] \Big\}, 
\end{equation}
where $1$ and $0$ represent the labels of the samples.
In \eqref{eq:gan-fake-data}, we assumed an arbitrary number $N$ of generated samples.

\begin{figure}
\centering
\begin{tikzpicture}[scale=1, transform shape, node distance=1cm, >=Latex]

\node at (0,2) [draw, align=left] {$\Dcal_{\text{real}} = \Big\{ (x_{i}, 1) : x_{i} \in \Dcal, \ i \in [M] \Big\}$
};

\draw[-{Latex[length=2mm]}, line width=.6pt] (3.1, 2) -- (4.8,1.2);

\node at (0,0) [draw, align=left] {$\Dcal_{\text{fake}} = \Big\{ (g_{\theta}(z), 0): z \sim \Ncal(0,I_{d})\Big\}$
};

\draw[-{Latex[length=2mm]}, line width=.6pt] (3.1, 0) -- (4.8,0.8);

\node at (6,1) [draw] {Classifier $a_{\eta}$};

\end{tikzpicture}
\caption{A schematic view of the generative adversarial networks (GANs). The dataset $\Dcal_{\text{real}}$ consists of the $M$ given real samples in $\Dcal$ with label 1. The dataset $\Dcal_{\text{fake}}$ consists of generated samples $g_{\theta}(z)$'s with label $0$, where $g_{\theta}: \mathbb{R}^{d} \to \mathbb{R}^{d}$ is the generator network with parameter $\theta$, and $z \sim \Ncal(0,I_{d})$. The classifier $a_{\eta}: \mathbb{R}^{d} \to (0,1)$ is an adversarial network with parameter $\eta$ and assigns every input sample with a probability value in $(0,1)$ that it believes the sample is real.}
\label{fig:gan}
\end{figure}
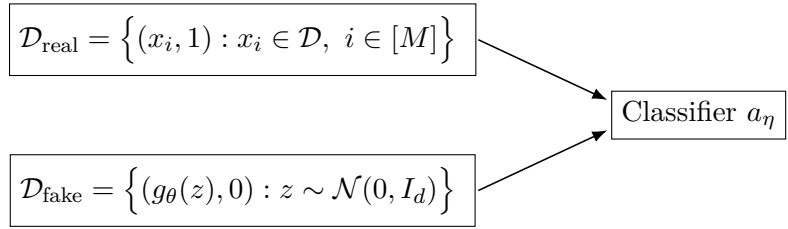

With the two datasets $\Dcal_{\text{real}}$ in \eqref{eq:gan-real-data} and $\Dcal_{\text{fake}}$, GANs further introduce a classifier, also called a \emph{discriminator}\index{Discriminator}, which is parameterized as an \emph{adversarial network} $a_{\eta}: \mathbb{R}^{d} \to (0,1)$ with parameter $\eta$.
This adversarial network $a_{\eta}$ is to be trained in order to distinguish between the generated samples from $\Dcal_{\text{fake}}$ in \eqref{eq:gan-fake-data} and the real ones from $\Dcal_{\text{real}}$ in \eqref{eq:gan-real-data}.
A schematic view of the GANs is provided in Figure \ref{fig:gan}.

The functionality of $a_{\eta}$ is based on the Bernoulli probability distribution. Recall that for any $p \in (0,1)$ and $X \sim \text{Bernoulli}(p)$, the probability of $X=1$ is $p$ and the probability of $X=0$ is $1-p$.
Therefore, the probability mass function of $X$ can be written as $\text{Prob}(X=x) = p^{x}(1-p)^{1-x}$ for $x=0$ and $1$. 
In light of this formulation, we can interpret $a_{\eta}(x) \in (0,1)$ as the probability of $x$ being a real sample. 
Specifically, this probability mass function is $a_{\eta}(x_{i})$ for any $x_{i}$ from $\Dcal_{\text{real}}$, and is $a_{\eta}(g_{\theta}(z_{j}))$ for any $g_{\theta}(z_{j})$ from $\Dcal_{\text{fake}}$. 
Since all samples from $\Dcal_{\text{real}}$ and $\Dcal_{\text{fake}}$ are independently drawn and the labels are already known to be 1 and 0 respectively, we can form their joint probability distribution as
\begin{equation}
\label{eq:gan-joint-p}
\Big( \prod_{i=1}^{M} a_{\eta}(x_{i}) \Big)^{1/M} \Big( \prod_{j=1}^{N} \big(1 - a_{\eta}(g_{\theta}(z_{j}))\big) \Big)^{1/N} ,
\end{equation}
where we removed all terms equal to 1 (e.g., $(1-a_{\eta}(x_{i}))^{0} = a_{\eta}(g_{\theta}(z_{j}))^{0} = 1$) and normalized this quantity by exponents based on the data sizes $M$ and $N$.
Taking logarithm on both sides of \eqref{eq:gan-joint-p}, we obtain
\begin{equation}
\label{eq:gan-obj}
E(\theta,\eta) : = \frac{1}{M} \sum_{i=1}^{M} \log a_{\eta}(x_{i}) + \frac{1}{N} \sum_{j=1}^{N} \log ( 1 - a_{\eta}(g_{\theta}(z_{j}))) .
\end{equation}

Now let us interpret the meaning of $E(\theta,\eta)$ in \eqref{eq:gan-obj}.
If the generator $g_{\theta}$ is fixed, then the classifier $a_{\eta}$ should be trained to assign $1$ (or very close to 1) to all $x_{i}$'s and $0$ (or very close to $0$) to all $g_{\theta}(z_{j})$'s. 
This is equivalent to maximizing $E(\theta,\eta)$ with respect to $\eta$ when $\theta$ is held fixed.
By this means, the classifier $a_{\eta}$ can excel its discrimination ability and tell correctly whether a sample is real or generated.

Conversely, if the classifier $a_{\eta}$ is fixed, then the generator $g_{\theta}$ should be trained in order to confuse the classifier, i.e., attaining a parameter $\theta$ such that $a_{\eta}(x_{i}) = a_{\eta}(g_{\theta}(z_{j})) = \frac{1}{2}$ for all $x_{i}$ and $z_{j}$.
This is equivalent to minimizing $E(\theta,\eta)$ with respect to $\theta$ when $\eta$ is held fixed.
By this means, the generator $g_{\theta}$ can excel its generation ability so that $g_{\theta}(z_{j})$ looks no different from a real sample to the classifier $a_{\eta}$, and $a_{\eta}$ can only assign probability $\frac{1}{2}$ to whatever it receives, i.e., $a_{\eta}$ cannot tell whether the sample is real or generated at all.

Based on the interpretation above, it is natural to set the problem of training $\theta$ and $\eta$ as the following saddle point (minimax) problem\index{Saddle point problem}:
\begin{equation}
\label{eq:gan-minmax}
\min_{\theta} \max_{\eta} \ E(\theta, \eta) .
\end{equation}
In practice, this saddle point problem \eqref{eq:gan-minmax} can be solved by alternately updating $\theta$ by gradient descent and $\eta$ by gradient ascent.
The convergence can be unstable but there are some remedies to mitigate this issue in practical implementations.

We can dive further into the saddle point problem \eqref{eq:gan-minmax} and see its meaning.
Suppose $\theta$ is fixed and the generator $g_{\theta}: \mathbb{R}^{d} \to \mathbb{R}^{d}$ is an invertible mapping.
Let $\ptheta$ be the probability distribution of the generated samples $g_{\theta}(z)$ with $z \sim \Ncal(0,I_{d})$.
Then by change of variables, we know that
\begin{equation}
\label{eq:gan-pgen}
\ptheta(x) = \Ncal(g_{\theta}^{-1}(x); 0, I_{d}) \det( \nabla g_{\theta}^{-1}(x)) .
\end{equation}
Suppose $M, N \to \infty$, namely, we have infinite many real and generated samples, then $E(\theta,\eta)$ in \eqref{eq:gan-obj} becomes
\begin{equation}
\label{eq:gan-cts-E}
E(\theta,\eta) = \int \pdata(x) \log a_{\eta}(x) \, dx + \int \ptheta (x) \log (1 - a_{\eta}(x)) \, dx .
\end{equation}
For any fixed $\theta$, we find that the critical point $\eta^{*}$ of $E(\theta,\cdot)$ in \eqref{eq:gan-cts-E} should satisfy
\begin{equation*}
\frac{\pdata(x)}{a_{\eta^{*}}(x)} - \frac{\ptheta (x)}{1-a_{\eta^{*}}(x)} = 0, 
\end{equation*}
which implies
\begin{equation*}
a_{\eta^{*}}(x) = \frac{\pdata(x)}{\pdata(x) + \ptheta(x)} 
\end{equation*}
for every $x$. Substituting this optimal $\eta^{*}$ back into $E(\theta,\eta)$, we obtain
\begin{align*}
E(\theta, \eta^{*}) 
& = \int \pdata(x) \log a_{\eta^{*}}(x) \, dx + \int \ptheta (x) \log (1 - a_{\eta^{*}}(x)) \, dx \\
& = 2 \, \js(\pdata, \ptheta) - 2 \log 2 ,
\end{align*}
where $\js(p,q):=(\kl(p,r) + \kl(q,r))/2$ is called the \emph{Jensen--Shannon} (JS) divergence between two probability distributions $p$ and $q$, and $r := (p+q)/2$ is the pointwise average of the two distributions. (See Appendix \ref{appsec:info-theory} for more details about the JS divergence.)
Therefore, the saddle point problem \eqref{eq:gan-minmax} is equivalent to
\begin{equation}
\label{eq:gan-js}
\min_{\theta} \ \js(\pdata, \ptheta) ,
\end{equation}

The formulation \eqref{eq:gan-js} suggests the possibility of using a broad range of differences or divergences between two probability distributions as the objective function for learning the parameter $\theta$ of the generator network $g_{\theta}$.
One popular choice following up the original work of GANs is using the Wasserstein-1 distance, i.e., changing \eqref{eq:gan-js} to
\begin{equation}
\label{eq:wgan}
\min_{\theta} \ W_{1}( \pdata, \ptheta) ,
\end{equation}
where the Wasserstein-1 distance between two probability distributions $p$ and $q$ is defined by
\begin{equation}
\label{eq:w1-dist}
W_{1}(p,q) := \min_{ \pi \in \Pi(p, q)} \iint |x-y| \pi(x,y) \, dx dy
\end{equation}
and $\Pi(x,y)$ is the set of joint distributions on $\mathbb{R}^{d} \times \mathbb{R}^{d}$ with marginal probability distributions as $p$ and $q$, given by
\begin{equation*}
\Pi(p, q) := \Big\{ \pi: \mathbb{R}^{d} \times \mathbb{R}^{d} \to [0,\infty) : 
\begin{array}{c}
p(x) = \int \pi(x,y) \, dy,\\
q(y) = \int \pi(x,y)\, dx,
\end{array}\ \forall \, x, y \in \mathbb{R}^{d} \Big\} .
\end{equation*}

Wasserstein distance have many important mathematical properties, including being a well-defined metric in the space of probability distributions on $\mathbb{R}^{d}$.
It is one of the key concepts in the area of optimal transport, which has been extensively studied in the past decades. 
For more details about Wasserstein distances, see Appendix \ref{appsec:info-theory}.
However, the main issue with Wasserstein distances is that they are complicated for numerical computations due to their constrained structures and dimensionality.

To solve the minimization \eqref{eq:wgan}, we can convert the Wasserstein-1 distance to its dual form, which is relatively easier to handle.
The dual form of the Wasserstein-1 distance $W_{1}(p,q)$ is given by
\begin{equation}
\label{eq:w1-dual}
W_{1}(p,q) = \max_{f \in \text{Lip}_{1}(\mathbb{R}^{d})} \int f(x) p(x) \, dx - \int f(x) q(x)\, dx ,
\end{equation}
where $\text{Lip}_{1}(\mathbb{R}^{d})$ is the set of all $1$-Lipschitz continuous functions defined on $\mathbb{R}^{d}$, i.e., $f \in \text{Lip}_{1}(\mathbb{R}^{d})$ if $|f(x) - f(y)| \le |x-y|$ for all $x,y \in \mathbb{R}^{d}$.

With the dual of Wasserstein-1 distance in \eqref{eq:w1-dual}, we can rewrite the minimization problem \eqref{eq:wgan} in an equivalent form, which is the minimization problem in the so-called Wasserstein GAN\index{Neural network!Wasserstein generative} model for learning the optimal parameter $\theta$ of the generator network $g_{\theta}$:
\begin{equation}
\label{eq:wgan-dual}
\min_{\theta} \max_{f \in \text{Lip}_{1}(\mathbb{R}^{d})} \int f(x) \pdata(x)\, dx - \int f(x) \ptheta(x)\, dx .
\end{equation}
The second integral in \eqref{eq:wgan-dual} can be written in a more accessible form
\begin{equation*}
\int f(x) \ptheta(x)\, dx = \int f(g_{\theta}(z)) \Ncal(z; 0, I_{d}) \, dz
\end{equation*}
due to change of variables \eqref{eq:gan-pgen}.

In practical implementations, the integrals in \eqref{eq:wgan-dual} are replaced with Monte Carlo integrations, and $f: \mathbb{R}^{d} \to \mathbb{R}$ is parameterized as a deep neural network with proper constraints on its parameters such that $f$ approximately satisfies $f \in \text{Lip}_{1}(\mathbb{R}^{d})$.
In this sense, $f$ plays a similar role as the adversarial network $a_{\eta}$ in the original work of GANs.
Interested readers may refer to the relevant literature listed at the end of this chapter for more details.

After the generator network $g_{\theta}$ is trained, we can first simulate as many samples as we want from $\Ncal(0,I_{d})$, and then apply $g_{\theta}$ to these standard Gaussian samples to obtain new samples, which are new samples following the probability distribution $\pdata$.

\section{Diffusion Models}
\label{sec:diffusion-models}

Diffusion models is a new class of generative models. 
The goal is again learning how to generate new samples following an unknown probability distribution $\pdata$ on $\mathbb{R}^{d}$, provided only a dataset $\Dcal = \{ x_{i} \in \mathbb{R}^{d}: i \in [M]\}$ consisting of $M$ samples independently drawn from $\pdata$.
Same as before, we do not know the analytic form of $\pdata$.

Diffusion models\index{Diffusion model} take a novel approach of sample generation by constructing a \emph{stochastic differential equation}\index{Differential equation!stochastic} (SDE) defined in $\mathbb{R}^{d}$ over the time interval $[0, T]$, where $T> 0$ is arbitrarily chosen:
\begin{equation}
\label{eq:dm-gen-sde-prototype}
d Z_{t} = \mu_{t} (Z_{t}) dt + (\Sigma_{t}(Z_{t}))^{1/2} dW_{t},
\end{equation}
where
\begin{equation*}
\mu: \ \mathbb{R}^{d} \times [0,T] \to \mathbb{R}^{d}
\end{equation*}
is called the \emph{drift} and 
\begin{equation*}
\Sigma: \ \mathbb{R}^{d} \times [0,T] \to \mathbb{R}^{d \times d} 
\end{equation*}
is called the \emph{diffusion}\index{Diffusion} coefficient\index{Diffusion!coefficient}. Both $\mu$ and $\Sigma$ are time-dependent.
The property of this SDE is that, from a simple initialization probability $\pinit$ on $\mathbb{R}^{d}$ which is often set to $\Ncal(0,I_{d})$ or alike, this SDE \eqref{eq:dm-gen-sde-prototype} can drive any initial value $Z_{0}$ drawn from $\pinit$ at time $t=0$ forward in time to $Z_{T}$ at $t=T$, such that $Z_{T}$ follows the probability $\pdata$, i.e., $Z_{T}$ is a sample drawn from $\pdata$.

Note that, since the goal is to generate any number of new samples $Z_{T}$'s following $\pdata$, we need the ability of easily drawing initials $Z_{0}$'s following $\pinit$. We also need the conditional probability distribution to have a closed form in order to implement the training loss function conveniently, which we will detail later. Due to these reasons, the initialization probability $\pinit$ is often chosen to be a simple one so that it is easy to draw samples from it. 
As mentioned above, a typical choice of $\pinit$ is the standard Gaussian distribution $\Ncal(0, I_{d})$ on $\mathbb{R}^{d}$, or some trivial variant of it with shifted mean or scaled variance.

For implementation simplicity and computation efficiency, $\Sigma$ is often set to a state-independent scalar multiple of the identity matrix $I_{d}$, namely, $\sigma I_{d}$, where $\sigma: [0,T] \to [0,\infty)$ depends on time $t$ only.
This simplification does not affect the core of diffusion models and can be generalized easily.

Now the goal of diffusion models is summarized as follows: Find a drift $\mu: \mathbb{R}^{d} \times [0,T] \to \mathbb{R}^{d}$ and a diffusion coefficient $ \sigma: [0, T] \to [0, \infty)$, such that for any trajectory $Z_{t}$ with initial value $Z_{0} \sim \pinit = \Ncal(0, I_{d})$ and following the SDE (for notation simplicity, we denote $\mu_{t}(z):=\mu(z,t) \in \mathbb{R}^{d} $ and $\sigma_{t}:=\sigma(t) \in \mathbb{R}$ hereafter):
\begin{equation}
\label{eq:dm-gen-sde}
d Z_{t} = \mu_{t}(Z_{t}) dt + \sigma_{t} d W_{t}
\end{equation}
over $[0,T]$, there is $Z_{T} \sim \pdata$. 
Once such $\mu$ and $\sigma$ are found, we can draw $Z_{0} \sim \pinit$ as many as we want, and numerically solve the SDE \eqref{eq:dm-gen-sde} (using numerical SDE solvers such as the Euler-Maruyama method) to obtain a new sample $Z_{T} \sim \pdata$ from each $Z_{0}$.

As we can see, the main task now is to find such $\mu_{t}$ and $\sigma_{t}$ for \eqref{eq:dm-gen-sde}. 
To this end, diffusion models first try to determine another SDE 
\begin{equation}
\label{eq:forward-time-sde}
d X_{t} = f_{t}(X_{t}) dt + g_{t} d W_{t}
\end{equation}
with some manually designed $f: \mathbb{R}^{d} \times [0, T] \to \mathbb{R}^{d}$ and diffusion coefficient $g: [0,T] \to [0, \infty)$,
so that \eqref{eq:forward-time-sde} can do exactly the opposite of \eqref{eq:dm-gen-sde}:
Namely, for any initial $X_{0}$ chosen to be one of the given samples following $\pdata$, we can numerically solve \eqref{eq:forward-time-sde} and get $X_{T} \sim \pinit$.
In this sense, the SDE \eqref{eq:forward-time-sde} can transform $\pdata$ to $\pinit$.
The purpose of designing this new SDE \eqref{eq:forward-time-sde} is to simulate its trajectories and learn the corresponding time-evolving probability. By doing so, we can learn the desired $\mu_{t}$ and $\sigma_{t}$, as will be shown later in this section.

Before moving on, we remark that finding the proper $f$ and $g$ for \eqref{eq:forward-time-sde} such that $X_{T} \sim \pinit$ from $X_{0} \sim \pdata$ is highly non-trivial in general. We will discuss this issue further in this section. Nevertheless, it is easy to find $f$ and $g$ such that $X_{T}$ can approximately follow $\pinit$ when $T$ is sufficiently large. For example, a common choice is using the Ornstein--Uhlenbeck (OU) process\index{Ornstein--Uhlenbeck process} with a diffusion coefficient compatible with the drift:
\begin{equation}
\label{eq:ou}
d X_{t} =  - a X_{t} dt + \sqrt{2a}\, dW_{t} .
\end{equation}
That is, 
\begin{equation}
\label{eq:ou-f-g}
f_{t}(x) = - ax \quad \text{and} \quad g_{t} \equiv \sqrt{2a}
\end{equation}
for all $t$ and $x$, where $a>0$ is a user-chosen constant.
For this SDE \eqref{eq:ou}, it is known that the conditional probability of $X_{t}$ given its initial value $X_{0} = x_{0} \sim \pdata$ is a Gaussian:
\begin{equation*}
\Ncal ( x_{0} e^{-at}, (1 - e^{-2at}) I_{d} ) 
\end{equation*}
for any $t \ge 0$. Therefore, we have
\begin{equation*}
X_{T}|(X_{0} = x_{0}) \ \sim \ \Ncal( x_{0} e^{-aT}, (1 - e^{-2aT}) I_{d} ) \ \approx \ \ \pinit := \Ncal (0, I_{d})
\end{equation*}
for sufficiently large $T$ with any $x_{0}$, regardless of what $\pdata$ is.
For now, we will accept the OU process \eqref{eq:ou} as our choice \eqref{eq:forward-time-sde} and that $X_{0} \sim \pdata$ and $X_{T} \sim \pinit = \Ncal (0,I_{d})$.

Now the questions are how these simple $f$ and $g$ in \eqref{eq:ou-f-g} relate to the desired $\mu$ and $\sigma$ in \eqref{eq:dm-gen-sde}, and how to use $f$ and $g$ to construct $\mu$ and $\sigma$.

To answer these two questions, we notice that the Fokker--Planck equation\index{Fokker--Planck equation} of the SDE \eqref{eq:forward-time-sde} is given by
\begin{equation}
\label{eq:forward-fpe}
\partial_{t} \rho_{t} = - \nabla \cdot (f_{t} \rho_{t}) + \frac{1}{2} g_{t}^{2} \Delta \rho_{t}
\end{equation}
where $\rho: \mathbb{R}^{d} \times [0, T] \to [0,\infty)$ is the probability density of $X_{t}$ for every $t \in [0,T]$, and $\rho_{0} = \pdata$ and $\rho_{T} = \pinit$.
On the other hand, we also have the Fokker--Planck equation of the SDE \eqref{eq:dm-gen-sde}:
\begin{equation}
\label{eq:reverse-fpe}
\partial_{t} \pi_{t} = - \nabla \cdot (\mu_{t} \pi_{t}) + \frac{1}{2} \sigma_{t}^{2} \Delta \pi_{t}
\end{equation}
defined in $\mathbb{R}^{d} \times [0,T]$, such that $Z_{t}$ solving \eqref{eq:dm-gen-sde} satisfies $Z_{t} \sim \pi_{t}$ for all $t$.

By comparing the two Fokker--Planck equations \eqref{eq:forward-fpe} and \eqref{eq:reverse-fpe}, we can show that if $\pi_{0} = \rho_{T}$ and we choose $\mu_{t}$ and $\sigma_{t}$ according to $f_{t}$ and $g_{t}$ properly, then $\pi_{t} = \rho_{T-t}$ for all $t \in [0, T]$.
This result is given in the following theorem.

\begin{theorem}
\label{thm:fb-fpe}
Let $\rho$ and $\pi$ follow the Fokker--Planck equations \eqref{eq:forward-fpe} and \eqref{eq:reverse-fpe}, respectively, $\rho_{t}$ and $\pi_{t}$ are positive everywhere, and $\pi_{0} = \rho_{T}$.
Suppose $f_{t} : \mathbb{R}^{d} \to \mathbb{R}^{d}$ is $L_{f}$-Lipschitz continuous at every $t \in [0,T]$ for some constant $L_{f}>0$. If
\begin{equation}
\label{eq:reverse-fpe-mu}
\mu_{t}(x) = - f_{T- t}(x) + \frac{1}{2} g_{T-t}^{2}(x) \frac{\nabla ( \rho_{T-t}(x) + \pi_{t}(x))}{\pi_{t}(x)}
\end{equation}
and 
\begin{equation}
\label{eq:reverse-fpe-sigma}
\sigma_{t} = g_{T-t}
\end{equation}
for all $ x \in \mathbb{R}^{d}$ and $t \in [0,T]$, then 
\begin{equation}
\pi_{t} = \rho_{T-t}
\end{equation} 
almost everywhere in $\mathbb{R}^{d}$ for all $t \in [0, T]$.
In particular, $\pi_{T} = \rho_{0} = \pdata$.
\end{theorem}

\begin{proof}
We denote the halved squared $L^{2}$ error between $\rho_{T-t}$ and $\pi_{t}$ by 
\begin{equation*}
\delta(t) := \frac{1}{2} \| \rho_{T-t} - \pi_{t} \|_{L^{2}(\mathbb{R}^{d})}^{2} = \frac{1}{2} \int_{\mathbb{R}^{d}} |\rho_{T-t}(x) - \pi_{t}(x) |^{2} \, dx
\end{equation*}
for every $t \in [0, T]$. We omit $\mathbb{R}^{d}$ and $(x)$ for notation simplicity in the remainder of this proof. Now we notice that 
\begin{align*}
\delta'(t)
& = \int ( \rho_{T-t} - \pi_{t} )  (- \partial_{t} \rho_{T-t} - \partial_{t} \pi_{t}) \,dx \\
& = \int ( \rho_{T-t} - \pi_{t} )  \Big[ \nabla \cdot (f_{T-t} \rho_{T-t} ) - \frac{1}{2} g_{T-t}^{2} \Delta \rho_{T-t} + \nabla \cdot (\mu_{t} \pi_{t}) - \frac{1}{2} \sigma_{t}^{2} \Delta \pi_{t} \Big] \,dx \\
& = \int ( \rho_{T-t} - \pi_{t} )  \Big[ \nabla \cdot \Big( f_{T-t} \rho_{T-t}  - \frac{1}{2} g_{T-t}^{2} \nabla \rho_{T-t} + \mu_{t} \pi_{t} - \frac{1}{2} \sigma_{t}^{2} \nabla \pi_{t} \Big) \Big] \,dx \\
& = \int ( \rho_{T-t} - \pi_{t} ) [ \nabla \cdot (f_{T-t} ( \rho_{T-t} - \pi_{t} ) ) ] \,dx \\
& = - \frac{1}{2} \int f_{T-t}  \nabla | \rho_{T-t} - \pi_{t}  |^{2} \,dx \\
& = \frac{1}{2} \int (\nabla \cdot f_{T-t}) | \rho_{T-t} - \pi_{t}  |^{2} \,dx \\
& \le \| \nabla \cdot f_{T-t} \|_{L^{\infty}} \delta(t) \\
& \le d L_{f} \delta(t) ,
\end{align*}
where the second equality is due to \eqref{eq:forward-fpe} and \eqref{eq:reverse-fpe}, the third equality is due to $\Delta \rho = \nabla \cdot (\nabla \rho)$ (and the same for $\pi$), the fourth equality is due to \eqref{eq:reverse-fpe-mu} and \eqref{eq:reverse-fpe-sigma}, the fifth and the sixth equalities are by the divergence theorem in calculus, and the last two inequalities are due to 
\begin{equation*}
|\nabla \cdot f_{T-t}| \le \| \nabla \cdot f_{T-t} \|_{L^{\infty}(\mathbb{R}^{d})} \le d L_{f}
\end{equation*}
as a result of the $L_{f}$-Lipschitz continuity of $f_{t}$ at every $t$.

By Gr\"{o}nwall inequality, we know that
\begin{equation*}
\delta(t) \le \delta(0) e^{d L_{f} t} \le \delta(0) e^{d L_{f} T} = 0
\end{equation*}
for all $t$, since $\delta(0) = \frac{1}{2} \| \rho_{T} - \pi_{0} \|_{L^{2}(\mathbb{R}^{d})}^{2} = 0$. This implies that 
\[
\rho_{T-t} = \pi_{t}
\]
almost everywhere in $\mathbb{R}^{d}$ for all $t \in [0,T]$, which completes the proof.
\end{proof}

Theorem \ref{thm:fb-fpe} suggests us to set $\mu$ in \eqref{eq:reverse-fpe-mu} as follows: since $\pi_{t} = \rho_{T-t}$ for all $t$, \eqref{eq:reverse-fpe-mu} becomes
\begin{equation}
\label{eq:dm-mu-final}
\mu_{t} = - f_{T-t} + g_{T-t}^{2} \frac{\nabla \rho_{T-t}}{\rho_{T-t}} = - f_{T-t} + g_{T-t}^{2} \nabla \log \rho_{T-t} .
\end{equation}
This implies that the Fokker--Planck equation \eqref{eq:reverse-fpe} is
\begin{equation}
\label{eq:reverse-fpe-simplified}
\partial_{t} \pi_{t} = - \nabla \cdot \Big( (- f_{T-t} + g_{T-t}^{2} \nabla \log \rho_{T-t})\, \pi_{t} \Big) + \frac{1}{2} g_{T-t}^{2} \Delta \pi_{t} .
\end{equation}
%
% If $\pi_{0} = \rho_{T} = \pinit = N(0,I)$, then $\pi_{T} = \rho_{0} = \pdata$, as desired.
%
As a consequence, we know the SDE \eqref{eq:dm-gen-sde} associated with the Fokker--Planck equation \eqref{eq:reverse-fpe-simplified} becomes
\begin{align}
d Z_{t} 
= ( -f_{T-t}(Z_{t}) + g_{T-t}^{2} \nabla \log \rho_{T-t} (Z_{t}) ) dt + g_{T-t} d W_{t} \label{eq:dm-gen-sde-explicit} .
\end{align}
Moreover, if we start from an initial $Z_{0} \sim \pi_{0} = \Ncal(0,I_{d})$, then we will end up with a new sample $Z_{T} \sim \pdata$ at time $T$ by solving \eqref{eq:dm-gen-sde-explicit}.
The relations described above are shown in Figure \ref{fig:dm}.

\begin{figure}
\centering
\begin{tikzpicture}[scale=.95, transform shape, node distance=1cm, >=Latex]

\node at (-0.5,1) [draw, align=left] {$X_{0} \sim \rho_{0} = \pdata$};

\draw[->, line width=.6pt] (1.1, 1) -- (6.2,1) node[above left] {$dX_{t} = f_{t}(X_{t}) dt + g_{t} dW_{t}\quad $ } node[below left] {$\partial_{t} \rho_{t} = - \nabla \cdot (f_{t} \rho_{t}) + \frac{1}{2} g_{t}^{2} \Delta \rho_{t}\ $ };

\node at (8.7,1) [draw, align=left] {$X_{T} \sim \rho_{T} \approx \pinit = \Ncal(0,I_{d})$};

\node at (-0.5,-1) [draw, align=left] {$Z_{T} \sim \pi_{T} = \pdata$};

\draw[->, line width=.6pt] (6.2,-1) -- (1.1, -1) node[above right] {$\quad dZ_{t} = \mu_{t}(Z_{t}) dt + \sigma_{t} dW_{t}$ } node[below right] {$\ \partial_{t} \pi_{t} = - \nabla \cdot (\mu_{t} \pi_{t}) + \frac{1}{2} \sigma_{t}^{2} \Delta \pi_{t}$ };

\node at (8.7,-1) [draw, align=left] {$Z_{0} \sim \pi_{0} \approx \pinit = \Ncal(0,I_{d})$};

\end{tikzpicture}
\caption{The forward process (top) and reverse process (bottom) of a diffusion model. The time $t$ increases (from $0$ to $T$) from left to right in the top plot, and from right to left in the bottom plot. The forward process is described by the SDE of $X_{t}$ or equivalently its Fokker--Planck equation of $\rho_{t}$ with $X_{t} \sim \rho_{t}$ at every $t$. The reverse process is described by the SDE of $Z_{t}$ or equivalently its Fokker--Planck equation of $\pi_{t}$ with $Z_{t} \sim \pi_{t}$ at every $t$. The relations between $f_{t}$, $g_{t}$, $\mu_{t}$ and $\sigma_{t}$ are given in \eqref{eq:reverse-fpe-mu} and \eqref{eq:reverse-fpe-sigma}, which yield $\pi_{t} = \rho_{T-t}$ at every $t \in [0,T]$. Notice that \eqref{eq:reverse-fpe-mu} can be written as \eqref{eq:dm-mu-final}, giving the expression of $\mu_{t}$ in $f_{t}$, $g_{t}$ and $\nabla \log \rho_{t}$.}
\label{fig:dm}
\end{figure}

Since we already know $f$ and $g$ as given in, for example, \eqref{eq:ou-f-g}, the only term to be determined in \eqref{eq:dm-gen-sde-explicit} is the function $\nabla \log \rho_{t}$ for every $t$. 
To this end, we need a method called \emph{score matching}\index{score matching}.
The purpose of score matching is to approximate $\nabla \log \rho: \mathbb{R}^{d} \times [0, T] \to \mathbb{R}^{d}$, called the \emph{Stein score function}\index{Score function} (we will just call it score function for short in this section), where $X_{t} \sim \rho_{t}$ for all $t \in [0,T]$, $X_{t}$ is the process following the SDE \eqref{eq:forward-time-sde}, and $\rho_{t}$ is the solution to the corresponding Fokker--Planck equation \eqref{eq:forward-fpe}.

The intuition of approximating $\nabla \log \rho$ is to parameterize a mapping $s: \mathbb{R}^{d} \times [0,T] \to \mathbb{R}^{d}$ as a deep neural network, train $s$ by minimizing $|s_{t}(x) - \nabla \log \rho_{t}(x)|^{2}$ for all $x$ and $t$, and use $s_{t}$ as the approximation of the score function $\nabla \log \rho_{t}$. 
However, this does not work in practice since we do not know the true value of $\nabla \log \rho_{t}$.

We consider score matching as a practical alternative to approximate the score function $\nabla \log \rho_{t}$. To instantiate this procedure, we use the OU process \eqref{eq:ou} as the SDE \eqref{eq:forward-time-sde}. 
Then, as we mentioned above, the conditional probability of $X_{t} = x_{t}$ given its initial $X_{0} = x_{0}$, is 
%
%If we rewrite the forward SDE of $q_{t}$ as a backward SDE
%\begin{equation}
%d X_{t} = - f_{t}(X_{t}) + g_{t}^{2} \nabla \log p_{t}(X_{t}) ) dt + g(t) d \bar{W}_{t}
%\end{equation}
%where $X_{T} \sim \pinit$, where $\bar{W}_{t}$ is the time-reversal Wiener process ($\bar{W}_{t}$ has a rigorous definition in Anderson. It is a function $W_{t}$ but it is not important for now.
%
%The corresponding evolution equation is
%\begin{equation*}
%\partial_{t} p_{t} = - \nabla \cdot \Big( (- f_{t} + g_{t}^{2} \nabla \log p_{t}) p_{t} \Big) + \frac{1}{2} g_{t}^{2} \Delta p_{t}
%\end{equation*}
%with $p_{T} \approx \pinit$ for $0 \le t \le T$. The key is to find $\nabla \log p_{t}$ since $\pinit \ne N(0,I)$ we cannot use this trick as in neural ODE.
%%
%In this case, we cannot obtain
%\begin{equation*}
%p_{t}(x_{t}) = \int p_{t}(x_{t}, x_{0}) \, dx_{0} = \int p_{t}(x_{t} | x_{0}) \pdata(x_{0}) \, dx_{0}
%\end{equation*}
%for any $x_{t}$ because we do not know which $x_{0}$ corresponds to this $x_{t}$.
%%
%So we need to find an alternative.
%
\begin{equation}
\label{eq:ou-cond}
\rho_{t}(x_{t} | x_{0}) = \Ncal (x_{t}; \alpha_{t} x_{0}, \beta_{t}^{2} I_{d}) , 
\end{equation}
where
\begin{equation*}
\alpha_{t} = e^{-at} \quad \text{and} \quad \beta_{t}^{2} = 1 - e^{-2at} .
\end{equation*}
Therefore, we can compute the \emph{conditional} score function\index{Score function!conditional} given the simplicity of Gaussian distributions:
\begin{equation}
\label{eq:ou-cond-score}
\nabla_{x_{t}} \log \rho_{t}(x_{t} | x_{0}) = - \frac{x_{t} - \alpha_{t} x_{0}}{\beta_{t}^{2}} .
\end{equation}
In \eqref{eq:ou-cond-score}, we specified that the gradient is with respect to $x_{t}$, since $\rho_{t}(x_{t}|x_{0})$ is a function of both $x_{t}$ and $x_{0}$.

Notice that \eqref{eq:ou-cond-score} is not the marginal score function $\nabla_{x_{t}} \log \rho_{t}(x_{t})$ that we want the network $s$ to approximate.
However, we can find a way to represent $\nabla_{x_{t}} \log \rho_{t}(x_{t})$ using the conditional score\index{Score function!conditional} function $\nabla_{x_{t}} \log \rho_{t}(x_{t} | x_{0})$. 
To this end, we notice that
\begin{align}
\nabla_{x_{t}} \log \rho_{t}(x_{t}) 
& = \frac{\nabla_{x_{t}} \rho_{t}(x_{t})}{\rho_{t}(x_{t})}  \nonumber \\
& = \frac{1}{\rho_{t}(x_{t})} \nabla_{x_{t}} \Big( \int \rho_{t}(x_{t}, x_{0}) \, dx_{0} \Big) \nonumber \\
& = \frac{1}{\rho_{t}(x_{t})} \nabla_{x_{t}} \Big( \int \rho_{t}(x_{t}|x_{0}) \pdata(x_{0}) \, dx_{0} \Big) \nonumber \\
& = \frac{1}{\rho_{t}(x_{t})} \int \nabla_{x_{t}} \rho_{t}(x_{t}|x_{0}) \pdata(x_{0}) \, dx_{0} \label{eq:score-convert} \\
& = \frac{1}{\rho_{t}(x_{t})} \int [\nabla_{x_{t}} \log \rho_{t}(x_{t}|x_{0})] \rho_{t}(x_{t}|x_{0})\pdata(x_{0}) \, dx_{0} \nonumber \\
& = \int [\nabla_{x_{t}} \log \rho_{t}(x_{t}|x_{0})] \frac{\rho_{t}(x_{t}|x_{0})\pdata(x_{0})}{\rho_{t}(x_{t})} \, dx_{0} \nonumber \\
& = \int [\nabla_{x_{t}} \log \rho_{t}(x_{t}|x_{0})] \rho_{t}(x_{0}|x_{t}) \, dx_{0}  \nonumber \\
& = \mathbb{E}_{X_{0} \sim \rho_{t}( \cdot |X_{t} = x_{t})}[\nabla_{x_{t}} \log \rho_{t}(X_{t}|X_{0}) ] \nonumber 
\end{align}
where $\rho_{t}(x_{t}, x_{0})$ is the joint probability distribution of $(X_{t},X_{0})$ at time $t$ in the second equality, $\rho_{t}(x_{t},x_{0}) = \rho_{t}(x_{t}|x_{0}) \rho_{0}(x_{0}) = \rho_{t}(x_{t}|x_{0}) \pdata(x_{0})$ in the third equality. We exchanged the gradient with respect to $x_{t}$ and the integral with respect to $x_{0}$ in the fourth equality above by using the Fubini's theorem, moved the fraction into the integral since $\rho_{t}(x_{t})$ is a constant regarding the integrand in the fifth equality, and the Bayes' rule in the sixth equality.

The relation given in \eqref{eq:score-convert} shows a useful connection between the score function $\nabla_{x_{t}} \log \rho_{t}(x_{t})$ and the conditional score function\index{Score function!conditional} $\nabla_{x_{t}} \log \rho_{t}(x_{t}|x_{0})$. 
However, we still do not know the posterior probability $\rho_{t}(x_{0}|x_{t})$ nor how to sample from it in order to compute the expectation at the end of \eqref{eq:score-convert}.
To write this expectation as something computable, we need a simple fact from probability, as shown in the following lemma.\index{Mean squared error}

\begin{lemma}[Mean squared error]
\label{lem:mse}
Suppose $X$ and $Y$ are two random variables in $\mathbb{R}^{d}$ and $\mathbb{R}^{m}$, respectively. Let $w: \mathbb{R}^{d} \times \mathbb{R}^{m} \to \mathbb{R}^{k}$ be a function. Then for any $ y \in \mathbb{R}^{m}$, let $\phi_{y}: \mathbb{R}^{k} \to \mathbb{R}$ be defined by
\begin{equation*}
\phi_{y}(c): = \mathbb{E}_{X \sim p_{X|Y}(\cdot|Y = y)} [ | c - w(X, Y) |^{2} ] .
\end{equation*}
Then $\phi_{y}$ has a unique minimizer $c_{y} \in \mathbb{R}^{k}$ given by
\begin{equation*}
c_{y} := \mathbb{E}_{X \sim p_{X|Y}(\cdot|Y = y)} [ w(X, Y) ] .
\end{equation*}
\end{lemma}

\begin{proof}
Notice that
\begin{align*}
\phi_{y}(c) 
& = \mathbb{E}_{X \sim p_{X|Y}(\cdot|Y = y)} [ | c - w(X, Y) |^{2} ] \\
& = \int |c - w(x,y) |^{2} p_{X|Y}(x|y) \, dx .
\end{align*}
Since $p_{X|Y}(x|y) \ge 0$ for every $x$, we know $\phi_{y}$ is a convex function of $c$.
Therefore, every critical point of $\phi_{y}$ is its global minimizer\index{Minimizer!global}. 

Now we take gradient of $\phi_{y}$ with respect to $c$, set it to $0$, and solve for the critical point(s). To this end, we find that
\begin{equation*}
\nabla_{c} \phi_{y}(c) = 2 \int (c - w(x,y) ) p_{X|Y}(x|y) \, dx = 0 ,
\end{equation*}
which implies a unique solution 
\begin{equation*}
c = \int c \, p_{X|Y}(x,y) \, dx = \int w(x,y) p_{X|Y}(x,y) \, dx = \mathbb{E}_{X \sim p_{X|Y}(\cdot|Y = y)} [ w(X, Y) ] .
\end{equation*}
This completes the proof.
\end{proof}

Lemma \ref{lem:mse} implies that the function $u: y \mapsto c_{y}$ is well-defined since $c_{y}$ exists and is unique. This is summarized in the following theorem.

\begin{theorem}
\label{thm:mse}
Suppose the conditions in Lemma \ref{lem:mse} hold. Define $u:\mathbb{R}^{m} \to \mathbb{R}^{k}$ that maps any $y \in \mathbb{R}^{m}$ to
\begin{equation*}
u(y) = \argmin_{c \in \mathbb{R}^{m}} \mathbb{E}_{X \sim p_{X|Y}(\cdot|Y = y)} [ | c - w(X, Y) |^{2} ].
\end{equation*}
Then there is
\begin{equation*}
u(y) = \mathbb{E}_{X \sim p_{X|Y}(\cdot|Y = y)} [ w(X, Y) ]
\end{equation*}
for every $y \in \mathbb{R}^{m}$.
\end{theorem}
\begin{proof}
The claim immediately follows Lemma \ref{lem:mse}, and hence the proof is omitted.
\end{proof}

Theorem \ref{thm:mse} allows us to rewrite a conditional expectation of interest as a least squares minimization problem. In particular, we realize that \eqref{eq:score-convert} is effectively 
\begin{align*}
\nabla_{x_{t}} \log \rho_{t}(x_{t}) & = \mathbb{E}_{X_{0} \sim \rho_{t}(\cdot|X_{t} = x_{t})}  [\nabla_{x_{t}} \log \rho_{t}(X_{t}|X_{0}) ] \\
& 
= \argmin_{c \in \mathbb{R}^{d}} \mathbb{E}_{X_{0} \sim \rho_{t}(\cdot|X_{t} = x_{t})}[|c - \nabla_{x_{t}} \log \rho_{t}(X_{t}|X_{0}) |^{2}] .
\end{align*}
This can be seen by employing Theorem \ref{thm:mse}, where $X$, $Y$, and $w(X,Y)$ are substituted by $X_{0}$, $X_{t}$, and $\nabla_{x_{t}} \log \rho_{t}(X_{t}|X_{0})$ (which can be regarded as a function of $X_{0}$ and $X_{t}$), respectively.

At this point, we have constructed a loss function that can be potentially used in training a deep neural network $s$ such that $s_{t} = \nabla \log \rho_{t}$. 
More specifically, since we want $s_{t}(x) = \nabla \log \rho_{t}(x)$ for all $x$ and $t$, we can use $x_{t}$ from the sample trajectories obtained by solving the SDE \eqref{eq:forward-time-sde} (i.e., the OU process in \eqref{eq:ou} in our current case) numerically. As such, we know for every $t \in [0,T]$, $s_{t}$ can be obtained by minimizing
\begin{align}
& \ \ \int \mathbb{E}_{X_{0} \sim \rho_{t}(\cdot|X_{t} = x_{t})}[|s_{t}(x_{t}) - \nabla_{x_{t}} \log \rho_{t}(X_{t}|X_{0}) |^{2}] \, d x_{t} \nonumber \\ 
& = \iint |s_{t}(x_{t}) - \nabla_{x_{t}} \log \rho_{t}(x_{t}|x_{0}) |^{2}] \rho_{t}(x_{0}|x_{t}) \rho_{t}(x_{t})\, d x_{0} d x_{t}  \label{eq:dm-loss-no-time}\\
& = \iint |s_{t}(x_{t}) - \nabla_{x_{t}} \log \rho_{t}(x_{t}|x_{0}) |^{2}] \rho_{t}(x_{t}|x_{0}) \pdata(x_{0})\, d x_{t} d x_{0} \nonumber 
\end{align}
where we used the fact 
\[
\rho_{t}(x_{0}|x_{t}) \rho_{t}(x_{t}) = \rho_{t}(x_{t}, x_{0}) = \rho_{t}(x_{t}|x_{0}) \rho_{0}(x_{0})
\]
and $\rho_{0}(x_{0}) = \pdata(x_{0})$ in the second equality of \eqref{eq:dm-loss-no-time}.

As we usually want to train $s_{t}$ over the time interval $[0,T]$ all together, we can manually set a probability density $\xi(t)$ on $[0,T]$ with $\xi(t) > 0$ for all $t$ and $\int_{0}^{T} \xi(t) \, dt = 1$, and train the network $s:\mathbb{R}^{d} \times [0,T] \to \mathbb{R}^{d}$ by solving
\begin{equation}
\label{eq:score-loss}
\min_{s} \int_{0}^{T}\iint \Big| s_{t}(x_{t}) - \nabla_{x_{t}} \log \rho_{t}(x_{t}|x_{0}) \Big|^{2} \rho_{t}(x_{t}|x_{0}) \pdata(x_{0})\, d x_{t} d x_{0} dt ,
\end{equation}
where the minimization is with respect to the network parameters of $s$.

We can further simplify the loss function in \eqref{eq:score-loss} for practical implementations.
When we use the OU process \eqref{eq:ou} to obtain the sample trajectories $\{x_{t}\}$ and $\rho_{t}$, we know the conditional probability $\rho_{t}(x_{t}|x_{0})$ is given by \eqref{eq:ou-cond} and the conditional score function $\nabla_{x_{t}} \log \rho_{t}(x_{t}|x_{0})$ is given by \eqref{eq:ou-cond-score}. 
In fact, since $\rho_{t}(x_{t}|x_{0}) = \Ncal (x_{t}; \alpha_{t} x_{0}, \beta_{t}^{2} I_{d})$, we can simply sample $\epsilon \sim \Ncal (0, I_{d})$ independent from any $X_{t}$, and rewrite $x_{t}$ as a function of $x_{0}$ and $\epsilon$ as follows:
\begin{equation}
\label{eq:dm-xt-x0-convert}
x_{t} = \alpha_{t} x_{0} + \beta_{t} \epsilon .
\end{equation} 
In this case, the conditional probability $\rho_{t}(x_{t}|x_{0})$ becomes much simplier:
\begin{equation}
\label{eq:cond-prob-simplified}
X_{t} \sim \rho_{t}(\cdot|x_{0}) \quad \text{if} \quad X_{t} = a_{t}x_{0} + \beta_{t} \epsilon \quad \text{where}\quad \epsilon \sim \Ncal (0, I_{d}) ,
\end{equation}
and the conditional score function reduces to
\begin{equation}
\label{eq:cond-score-simplified}
\nabla_{x_{t}} \log \rho_{t}(x_{t}|x_{0}) = - \frac{\epsilon}{\beta_{t}} .
\end{equation}

To offset the effect of $\beta_{t} \to 0$ as $t \to 0$ in the OU process, we can choose $\xi(t) \propto \beta_{t}^{2}$. We can also set the network to be trained in practice as $\varepsilon: \mathbb{R}^{d} \times [0,T] \to \mathbb{R}^{d}$ instead of $s$ we used in \eqref{eq:score-loss}, such that
\begin{equation}
\label{eq:score-err-convert}
s_{t}(x) = -  \frac{\varepsilon_{t}(x)}{\beta_{t}}
\end{equation}
for all $x$ and $t$. 
Substituting \eqref{eq:dm-xt-x0-convert}, \eqref{eq:cond-prob-simplified}, and \eqref{eq:cond-score-simplified} into \eqref{eq:score-loss}, we obtain a minimization problem of training the network $\varepsilon$ as simple as 
\begin{equation}
\label{eq:score-err-loss}
\min_{\varepsilon} \int_{0}^{T} \iint |\varepsilon_{t}(\alpha_{t} x_{0} + \beta_{t} \epsilon ) - \epsilon |^{2} \Ncal (\epsilon; 0, I_{d})\pdata(x_{0})\, d x_{0} d \epsilon dt .
\end{equation}

In numerical implementation, the loss function in \eqref{eq:score-err-loss} can be approximated by Monte Carlo integration (see Appendix \ref{appsec:mc}) given by
\begin{equation}
\label{eq:dm-mc-loss}
\frac{1}{NM}\sum_{t} \sum_{(x_{0}, \epsilon)} |\varepsilon_{t}(\alpha_{t} x_{0} + \beta_{t} \epsilon ) - \epsilon |^{2} .
\end{equation}
More precisely, we first sample $N$ time points from the probability density $\xi$ on $(0,1)$. Then for every sampled time point $t$, we use one sample $x_{0}$ given in the original dataset $\Dcal$ consisting of $M$ samples drawn from $\pdata$, and draw an $\epsilon \sim \Ncal (0, I_{d})$ independently.
These samples are put in \eqref{eq:dm-mc-loss}, which is the loss function we use to train the network $\varepsilon$ in practice.
Once the network $\varepsilon$ is trained, we can set $s_{t}$ as in \eqref{eq:score-err-convert} and know that it is an approximation of the score function $\nabla \log \rho_{t}$.

\paragraph{Sample generation using trained score function}

Once we have the approximate score function $s_{t}$, we can put it in the place of the true score function $\nabla \log \rho_{t}$ in \eqref{eq:dm-gen-sde-explicit}.
Notice that for the OU process, we have $f$ and $g$ given in the simple forms in \eqref{eq:ou-f-g}, i.e., $f_{t} = -ax$ and $g_{t} = \sqrt{2a}$. Then the SDE \eqref{eq:dm-gen-sde-explicit} for generation is
\begin{equation}
\label{eq:dm-ou-gen}
d Z_{t} = (-aZ_{t} + 2a  s_{t} (Z_{t}) ) dt + \sqrt{2a} \, d {W}_{t} .
\end{equation}
The actual sample generation can be proceeded by following the Euler-Maruyama scheme: Partition $[0,T]$ into intervals of equal length $h>0$, and compute
\begin{equation}
\label{eq:dm-euler-gen}
Z_{t + h} = Z_{t} + (-aZ_{t} + 2a  s_{t} (Z_{t})) h +  \sqrt{2 a h} \zeta_{t} 
\end{equation}
for $t = 0, h, 2h, \dots, T$, where the initial $Z_{0}$ and the perturbations $\zeta_{t}$'s are sampled from $\Ncal (0, I_{d})$ independently. Then $Z_{T}$ is a newly generated sample and follows the probability $\pdata$.

There is an obvious drawback of \eqref{eq:dm-euler-gen}: At each time point $t$, we need to sample one $\zeta_{t}$ from the standard Gaussian. This can be very expensive when there are many time steps and we need to simulate many Gaussian samples.

We can overcome this drawback by converting the SDE \eqref{eq:dm-ou-gen} into an ODE, such that the solution follow the same probability distribution $\pi_{t}$ at every $t$.
To this end, we notice that $\Delta \pi_{t} = \nabla \cdot (\pi_{t} \nabla \log \pi_{t})$ in \eqref{eq:reverse-fpe-simplified}. Furthermore, we know $\pi_{t} = \rho_{T-t}$ for all $t \in [0, T]$ from Theorem \ref{thm:fb-fpe}. Therefore, the Fokker--Planck equation \eqref{eq:reverse-fpe-simplified} can be rewritten as a continuity equation\index{Continuity equation} of $\pi$:
\begin{equation}
\label{eq:reverse-fpe-simplified-ode}
\partial_{t} \pi_{t} = - \nabla \cdot \Big( (- f_{T-t} + \frac{1}{2} g_{T-t}^{2} \nabla \log \rho_{T-t})\, \pi_{t} \Big) ,
\end{equation}
and the corresponding ODE is
\begin{equation}
\label{eq:dm-ode}
\zdot_{t} = - f_{T-t}(z_{t}) + \frac{1}{2} g_{T-t}^{2} \nabla \log \rho_{T-t}(z_{t}) .
\end{equation}
where $\zdot_{t}$ is the time derivative of $z_{t}$.
In practice, we can substitute $\nabla \log \rho_{T-t}$ with the trained score function $s_{T-t}$, and solve this ODE \eqref{eq:dm-ode} for $z_{T}$ from any initial $z_{0}$ drawn from $\Ncal (0,I_{d})$. 
The generated $z_{T}$ also follows the distribution $\pi_{T} = \rho_{0} = \pdata$ as desired.

There are many numerical ODE solvers, such as the midpoint method and the fourth-order Runge-Kutta method \cite{burden2011numerical}, that can solve the ODE \eqref{eq:dm-ode} much faster than solving the SDE \eqref{eq:dm-ou-gen}.
The numerical error caused by these solvers is usually very small compared to the approximation error of the learned score function $s_{t}$ to the true score function $\nabla \log \rho_{t}$.

\paragraph{Other SDEs to generate sample trajectories for training}

There are actually many alternatives of the OU process \eqref{eq:ou} that we can use to design \eqref{eq:forward-time-sde}.
For example, the following class of variance-preserving (VP) SDEs is a slight generalization of the OU process and often used in practical implementations:
\begin{equation}
\label{eq:vp-sde}
d X_{t} = - \gamma_{t} dt + \sqrt{2 \gamma_{t}} d W_{t},
\end{equation}
where $\gamma_{t}$ is chosen between $[\gamma_{\min}, \gamma_{\max}]$ for every $t$.
Here the bounds $\gamma_{\min}$ and $\gamma_{\max}$ are some user-chosen positive numbers, and $\gamma_{\min}$ is often set as a small positive number, such as 0.01, to prevent $\gamma_{t}$ from being too close to $0$ and cause numerical instability in computation. On the other hand, the choice of $\gamma_{\max}$ is rather flexible, such as 20.

The VP SDE \eqref{eq:vp-sde} also has solution following Gaussian conditioned on the initial value $X_{0} = x_{0}$ like the OU process. 
More precisely, the conditional probability distribution of $X_{t}$ given $X_{0}=x_{0}$ is 
\[
\rho_{t}(x_{t} | x_{0}) = \Ncal (x_{t}; \alpha_{t} x_{0}, \beta_{t}^{2} I_{d} )
\]
where
\begin{equation}
\label{eq:vp-sde-alpha}
\alpha_{t} = e^{- \int_{0}^{t} \gamma_{\tau} d\tau}
\end{equation}
and 
\begin{equation}
\label{eq:vp-sde-beta}
\beta_{t}^{2} = 1 - e^{- 2\int_{0}^{t} \gamma_{\tau}\, d\tau} .
\end{equation}
One simple choice of $\gamma_{t}$ for \eqref{eq:vp-sde} is 
\[
\gamma_{t} = \gamma_{\min} + t (\gamma_{\max} - \gamma_{\min})
\]
for $t \in [0,1]$. Then 
\begin{equation*}
\int_{0}^{t} \gamma_{\tau} \, d \tau = t \gamma_{\min} + \frac{t^{2}}{2} (\gamma_{\max} - \gamma_{\min} ) ,
\end{equation*}
and $\alpha_{t}$ and $\beta_{t}$ in \eqref{eq:vp-sde-alpha} and \eqref{eq:vp-sde-beta} have closed form expressions.
We note that VP SDE \eqref{eq:vp-sde} does not really preserve the variance all the time as indicated in its name, but the variance does tend to a constant as $t$ becomes larger.

\paragraph{Discussions on the choice of the SDE \eqref{eq:forward-time-sde}}

As the sample data distribution $\pdata$ can vary in different applications, we would hope to find some ``universal'' drift $f$ and diffusion coefficient $g$ such that the forward-time SDE \eqref{eq:forward-time-sde} can drive $X_{0} \sim \rho_{0} = \pdata$ to $X_{T} \sim \rho_{T} = \pinit = \Ncal (0,I_{d})$. 
By universal we meant that $f$ and $g$ are chosen such that the SDE \eqref{eq:forward-time-sde}, or equivalently the Fokker--Planck equation \eqref{eq:forward-fpe}, can transfer \emph{any given} $\pdata$ to $\Ncal(0,I_{d})$ in finite time $T$.
In other words, the choice of $f$ and $g$ should not depend on $\pdata$.
However, this is not always easy because $\pdata$ is an unknown and usually complicated probability distribution in practice. 
In fact, we can claim that such universal $f$ and $g$ do not exist, as shown in the next proposition.

\begin{proposition}
\label{prop:f-g-nonexist}
If there exist $f$ and $g$ such that the Fokker--Planck equation \eqref{eq:forward-fpe} transfers $\rho_{0} = \pdata$ to $\rho_{T} = \Ncal(0,I_{d})$, then $f$ and $g$ must depend on $\pdata$.
\end{proposition}

\begin{proof}
We use a proof by contradiction. Assume that $f$ and $g$ do not depend on $\pdata$. We know that there exists $\pi$ as described in Proposition \ref{thm:fb-fpe} such that $\pi_{t} = \rho_{T-t}$ almost everywhere in $\mathbb{R}^{d}$ for every $t$, where $\rho_{t}$ satisfies \eqref{eq:forward-fpe} with such $f$ and $g$ and terminal values $\rho_{0} = \pdata$ and $\rho_{T} = \Ncal (0,I_{d})$. 
This implies that the Fokker--Planck equation \eqref{eq:reverse-fpe-simplified} of $\pi$ can be written as
\begin{equation*}
\partial_{t} \pi_{t} = - \nabla \cdot \Big( (- f_{T-t} + g_{T-t}^{2} \nabla \log \pi_{t}) \pi_{t} \Big) + \frac{1}{2} g_{T-t}^{2} \Delta \pi_{t} .
\end{equation*}
with initial $\pi_{0} = \rho_{T} = \Ncal (0, I_{d})$. However, none of $\pi_{0}$, $f$, and $g$ includes any information about $\pdata$. This is a contradiction to $\pi_{T} = \rho_{0} = \pdata$. 
\end{proof}

As we see now, transferring from $\pdata$ to $\pinit$ in a finite time interval $[0,T]$ is itself a non-trivial problem, perhaps not much simpler than transferring from $\pinit$ to $\pdata$. 
While we used the OU process \eqref{eq:ou} such that the probability distribution $\rho_{T}$ is approximately $\pinit$, we know it is not possible to reach $\pinit = \Ncal(0, I_{d})$ exactly within a finite time $T$, as justified in Proposition \ref{prop:f-g-nonexist}.
Theoretically speaking, we should use the actual $\rho_{T}$ solved from the Fokker--Planck equation \eqref{eq:forward-fpe} or the SDE \eqref{eq:forward-time-sde} as the initial distribution $\pi_{0}$ in sample generation.
However, this is not practically feasible because it is also difficult to evaluate or sample from $\rho_{T}$.

It is natural to think whether there exists some proper drift $f$ (and diffusion coefficient $g$) such that the solution $\rho$ of the Fokker--Planck equation \eqref{eq:forward-fpe} satisfies $\rho_{0} = \pdata$ and $\rho_{T} = \Ncal(0,I_{d})$ (or the other way around, which is the original target generation problem). The answer to this question is positive provided certain mild conditions on $\pdata$.

In fact, this is about the feasibility of the so-called Schr\"{o}dinger bridge problem (SBP): given two probability distributions $p_{0}$ and $p_{1}$ in $\mathbb{R}^{d}$, the SBP can be written as
\begin{align}
\min_{f} \quad & \frac{1}{2} \int_{0}^{T} \int |f_{t}(x)|^{2} \rho_{t}(x) \, dx dt , \nonumber \\
\text{s.t.} \quad & \partial_{t} \rho_{t} = - \nabla \cdot (f_{t} \rho_{t}) + \frac{\sigma^{2}}{2} \Delta \rho_{t} , \quad 0 \le t \le T , \label{eq:sbp} \\
& \rho_{0} = p_{0}, \quad \rho_{T} = p_{1} , \nonumber 
\end{align}
where $\sigma > 0$ is a given constant diffusion coefficient. 
If we set $p_{0} = \pdata$ and $p_{1} = \Ncal(0,I_{d})$, then the SBP does not only require transferring from $\pdata$ to $\Ncal(0,I_{d})$, which immediately answers the feasibility of $f$ (or $\mu$) we had earlier, but also attain the optimality regarding the objective function in \eqref{eq:sbp}.
Nevertheless, finding either a feasible solution or an optimal solution to \eqref{eq:sbp} is difficult in general.

Despite the approximation issue mentioned above, diffusion models still perform well in practice. The discussions above indicate the limitation of diffusion models and suggest the necessity of learning the score function $\nabla \log \rho_{t}$ to build the generating drift $\mu_{t}$ in \eqref{eq:dm-gen-sde}.

\section{Probability Density Control}
\label{sec:probability-density-control}

In the derivation of diffusion models in Section \ref{sec:diffusion-models}, we have two state-independent scalar functions $g_{t}$ and $\sigma_{t}$.
They are the diffusion coefficients in the Fokker--Planck equations\index{Fokker--Planck equation} \eqref{eq:forward-fpe} and \eqref{eq:reverse-fpe-simplified}, respectively, and satisfy $\sigma_{t} = g_{T-t}$ for all $t$, 
However, they do not play any critical role in the learning and generation process, but only determine the magnitude of fluctuations of the corresponding SDEs \eqref{eq:forward-time-sde} and \eqref{eq:dm-gen-sde-explicit}.

In light of Theorem \ref{thm:fb-fpe}, we can simply choose $g_{t} = 0$ and $\sigma_{t} = 0$ for all $t$, then the Fokker--Planck equations \eqref{eq:forward-fpe} and \eqref{eq:reverse-fpe-simplified} reduces to the following two continuity equations\index{Continuity equation} 
\begin{equation}
\label{eq:forward-ct}
\partial_{t} \rho_{t} = - \nabla \cdot (f_{t} \rho_{t})
\end{equation}
and
\begin{equation}
\label{eq:reverse-ct}
\partial_{t} \pi_{t} = - \nabla \cdot (-f_{T-t} \pi_{t})
\end{equation}
respectively, defined on $\mathbb{R}^{d} \times [0,T]$.
Moreover, $\rho$ and $\pi$ still satisfy 
\begin{equation*}
\pi_{t} = \rho_{T-t}
\end{equation*}
for all $t \in [0,T]$. In particular, $\pi_{0} = \rho_{T}$ and $\pi_{T} = \rho_{0}$.

We also note that, in this case, the corresponding SDEs \eqref{eq:forward-time-sde} and \eqref{eq:dm-gen-sde-explicit} reduce to the following two ODEs
\begin{equation}
\label{eq:forward-ode}
\xdot_{t} = f_{t}(x_{t})
\end{equation}
and
\begin{equation}
\label{eq:reverse-ode}
\zdot_{t} = -f_{T-t}(z_{t}) ,
\end{equation}
respectively.

Therefore, if we can find a drift $f: \mathbb{R}^{d} \times [0,T] \to \mathbb{R}^{d}$, such that the solution $\rho_{t}$ of \eqref{eq:forward-ct} satisfies $\rho_{0} = \pdata$ and $\rho_{T} = \Ncal (0, I_{d})$, then we have $\pi_{0} = \Ncal (0, I_{d})$ and $\pi_{T} = \pdata$. 
As a result, we can simply draw a sample $z_{0}$ from the standard Gaussian $\Ncal(0, I_{d})$, solve the ODE \eqref{eq:reverse-ode} to time $T$, and obtain $z_{T}$ which is a new sample following the probability distribution $\pi_{T} = \pdata$. 
This relation is illustrated in Figure \ref{fig:pdc}.

\begin{figure}
\centering
\begin{tikzpicture}[scale=1, transform shape, node distance=1cm, >=Latex]

\node at (-0.5,1) [draw, align=left] {$x_{0} \sim \rho_{0} = \pdata$};

\draw[->, line width=.6pt] (1.1, 1) -- (4.6,1) node[above left] {$\xdot_{t} = f_{t}(x_{t})\quad \ $ } node[below left] {$\partial_{t} \rho_{t} = - \nabla \cdot (f_{t} \rho_{t}) $ };

\node at (7,1) [draw, align=left] {$x_{T} \sim \rho_{T} \approx  \pinit = \Ncal(0,I_{d})$};

\node at (-0.5,-.5) [draw, align=left] {$z_{T} \sim \pi_{T} = \pdata$};

\draw[->, line width=.6pt] (4.6,-.5) -- (1.1, -.5) node[above right] {$\qquad \zdot_{t} = \mu_{t}(z_{t})$ } node[below right] {$\ \partial_{t} \pi_{t} = - \nabla \cdot (\mu_{t} \pi_{t}) $ };

\node at (7,-.5) [draw, align=left] {$z_{0} \sim \pi_{0} \approx  \pinit = \Ncal(0,I_{d})$};

\end{tikzpicture}
\caption{The forward process (top) and reverse process (bottom) of probability control. The time $t$ increases (from $0$ to $T$) from left to right in the top plot, and from right to left in the bottom plot. The forward process is described by the ODE of $x_{t}$ or equivalently its continuity equation of $\rho_{t}$ with $x_{t} \sim \rho_{t}$ at every $t$. The reverse process is described by the ODE of $z_{t}$ or equivalently its continuity equation of $\pi_{t}$ with $z_{t} \sim \pi_{t}$ at every $t$. There is $\mu_{t} = - f_{T-t}$ which yields $\pi_{t} = \rho_{T-t}$ at every $t \in [0,T]$. The approximation $\rho_{T} \approx \Ncal(0,I_{d})$ is attained by finding $f$ that minimizes $\kl(\rho_{T}, \Ncal(0,I_{d}))$, as given in \eqref{eq:dc}.}
\label{fig:pdc}
\end{figure}

Now we can see that the key is to find this drift $f$. 
Due to Proposition \ref{prop:f-g-nonexist}, we know that $f$ must depend on $\pdata$ for $\rho_{T} = \Ncal (0,I_{d})$ in finite time $T$. 
Therefore, manually designed $f$ are excluded from our consideration. 
Instead, we must learn this $f$ from the given samples of $\pdata$.

We can consider the problem of finding $f$ as a probability density control problem, i.e., learning a control vector field $f: \mathbb{R}^{d} \times [0,T] \to \mathbb{R}^{d}$ which can steer the probability density $\rho_{t}$ in time from $\rho_{0}=\pdata$ to $\rho_{T} = \Ncal (0, I_{d})$ at time $T$.
More precisely, this probability density control problem can be written as
\begin{subequations}
\label{eq:dc}
\begin{align}
\min_{f} \quad & \kl( \rho_{T}, \, \Ncal(0,I_{d})) ,  \label{subeq:dc-obj} \\
\text{s.t.} \quad & \partial_{t} \rho_{t} = - \nabla \cdot (f_{t} \rho_{t}), \quad 0 \le t \le T , \label{subeq:dc-ct} \\
& \rho_{0} = \pdata . \label{subeq:dc-initial}
\end{align}
\end{subequations}
Once the solution $f$ to \eqref{eq:dc} is found, we know $\rho_{T} = \Ncal(0,I_{d})$ approximately, and then we can use $f$ in the ODE \eqref{eq:reverse-ode} for sample generation.

To solve \eqref{eq:dc}, we can use the neural ODE method introduced in Section \ref{sec:node}. Let $\Dcal = \{x_{i} \in \mathbb{R}^{d}: i \in [M] \}$ be the dataset consisting of $M$ samples independently drawn from $\pdata$, we can set a group of initial values $x^{(i)}_{0} = x_{i}$ for all $i$, and for each $x^{(i)}_{0}$, we solve the ODE \eqref{eq:forward-ode}.
Then we know $x^{(i)}_{t} \sim \rho_{t}$ for all $t \in [0,T]$.

The remaining question is how to represent the objective function in \eqref{subeq:dc-obj} using the points $\{x^{(i)}_{T}\}$. We notice that the objective function can be decomposed into two parts:
\begin{equation}
\label{eq:dc-obj-AB}
\kl(\rho_{T}, \Ncal(0,I_{d}) )
= \underbrace{\int \rho_{T} (x) \log \rho_{T}(x) \, dx}_{=:\, A}
- \underbrace{\int \rho_{T} (x) \log \Ncal (x; 0, I_{d}) \, dx}_{=:\, B}.
\end{equation}
It is easy to see that the term $B$ in \eqref{eq:dc-obj-AB} is
\begin{align*}
B
& = \int \rho_{T}(x) \log \Ncal (x; 0, I_{d}) \, dx \\
& = - \int \rho_T(x) \Big(\frac{1}{2}|x|^2 + \frac{d}{2}\log(2\pi) \Big) \, dx \\
& = - \frac{1}{2}\int |x|^2 \rho_T(x)\, dx  - \frac{d}{2}\log(2\pi) .
\end{align*}
%
% We can ignore the constant term $\frac{d}{2}\log(2\pi)$ as it does not affect the minimization in \eqref{subeq:dc-obj}.
%
Then we can use Monte Carlo integration and approximate $B$ by 
\begin{equation}
\label{eq:dc-B-approx}
B \approx - \frac{1}{2}\sum_{i = 1}^{M} |x^{(i)}_{T}|^{2} - \frac{d}{2}\log(2\pi)
\end{equation}
since $x^{(i)}_{T} \sim \rho_{T}$.

For the term $A$ in \eqref{eq:dc-obj-AB}, we notice that it is the negative entropy of $\rho_{T}$. We need to find an approach that makes $A$ computable in practice.
To this end, we let $h_{t}$ denote the negative entropy of $\rho_{t}$ at every $t$, namely,
\begin{equation}
\label{eq:dc-ht}
h_{t} := \int \rho_{t}(x) \log \rho_{t}(x) \, dx .
\end{equation}
Notice that $h_{t} \in \mathbb{R}$ for every $t$. In particular, $A = h_{T}$, and $h_{0}$ is the negative entropy of $\pdata$ which is an unknown constant.
Furthermore, we see that
\begin{align}
\dot{h}_{t}
& = \frac{d}{dt} \Big( \int \rho_{t}(x) \log \rho_{t}(x) \, dx \Big) \nonumber \\
& = \int ( \log \rho_{t}(x) + 1) \partial_{t} \rho_{t}(x) \, dx \nonumber \\
& = - \int ( \log \rho_{t}(x) + 1) \nabla \cdot (\rho_{t}(x) f_{t}(x)) \, dx \label{eq:dc-dhdt}\\
& = \int \nabla (\log \rho_{t}(x)+1)  \cdot (\rho_{t}(x) f_{t}(x)) \, dx \nonumber \\
& = \int \nabla \rho_{t}(x)  \cdot  f_{t}(x) \, dx \nonumber \\
& = -\int \rho_{t}(x)  \nabla  \cdot  f_{t}(x) \, dx , \nonumber 
\end{align}
where the third equality is due to the continuity equation \eqref{eq:forward-ct}. 
In practice, we can use Monte Carlo integration to approximate the integral on the right-hand side of \eqref{eq:dc-dhdt} since $x_{t}^{(i)} \sim \rho_{t}$:
\begin{equation}
\label{eq:dc-approx-dhdt}
\dot{h}_{t} = -\int \rho_{t}(x)  \nabla  \cdot  f_{t}(x) \, dx
\approx - \frac{1}{M}\sum_{i=1}^{M} \nabla \cdot f_{t}(x_{t}^{(i)}) .
\end{equation}
Therefore, in practical implementation, we can add an auxiliary variable $h_{t}$ and solve the ODE \eqref{eq:dc-approx-dhdt} together with the ODEs \eqref{eq:forward-ode} of $\{x_{t}^{(i)}: i \in [M]\}$, and obtain $A = h_{T} - h_{0}$. 
Since $h_{0}$ is the negative entropy of $\pdata$, which is a constant independent of $f$, we can simply assume it is zero in computation as it will not change the solution to \eqref{eq:dc}.

To conclude, we have formulated the problem \eqref{eq:dc} of finding $f: \mathbb{R}^{d} \times [0,T] \to \mathbb{R}^{d}$ as the following practically implementable problem:
\begin{subequations}
\label{eq:dc-implement}
\begin{align}
\min_{f} \quad & h_{T} + \frac{1}{2} \sum_{i=1}^{M}|x_{T}^{(i)}|^{2},  \label{subeq:dc-implement-obj} \\
\text{s.t.} \quad & \xdot_{t}^{(i)} = f_{t} (x_{t}^{(i)}), \quad i \in [M], \label{subeq:dc-implement-x} \\
& \dot{h}_{t} = - \frac{1}{M}\sum_{i=1}^{M} \nabla \cdot f_{t}(x_{t}^{(i)}), \quad 0 \le t \le T , \label{subeq:dc-implement-h} \\
& x_{0}^{(i)} = x_{i}, \quad h_{0} = 0 , \label{subeq:dc-implement-initial}
\end{align}
\end{subequations}
where $\Dcal = \{x_{i} \in \mathbb{R}^{d}: i\in [M]\}$ is the set of $M$ given samples independently drawn from $\pdata$. We have omitted all irrelevant constants from \eqref{eq:dc-implement}.
As usual, the minimization is with respect to the network parameters of $f$.

We can employ the neural ODE method to solve \eqref{eq:dc-implement}, where the gradient of the loss function in \eqref{subeq:dc-implement-obj} with respect to the parameters of the network $f$ can be computed by the adjoint ODE part in this method.

As we mentioned before, once $f$ is trained by solving \eqref{eq:dc-implement}, we can draw initial $z_{0}$ from $\Ncal(0,I_{d})$, and solve the ODE \eqref{eq:reverse-ode} with the trained $f$ up to time $T$. Then we obtain $z_{T}$, which is a new sample following the probability distribution $\pdata$.

\section{Flow Matching}
\label{sec:flow-matching}

Flow matching\index{Flow matching} is an alternative generative model framework that aims at directly learning the vector field\index{Vector field!generative} to transfer $\Ncal(0,I_{d})$ to $\pdata$.
Same as before, we are only given a dataset $\Dcal = \{x_{i} \in \mathbb{R}^{d} : i \in [M]\}$ consisting of $M$ samples independently drawn from $\pdata$.
In flow matching, our goal is to learn a vector field $u: \mathbb{R}^{d} \times [0,T] \to \mathbb{R}^{d}$ using these samples, such that the solution $x_{T}$ to the ODE
\begin{equation}
\label{eq:fm-ode}
\xdot_{t} = u_{t}(x_{t})
\end{equation}
with any initial $x_{0} \sim \Ncal (0, I_{d})$ satisfies $x_{T} \sim \pdata$. 
As the value of $T$ does not matter in the flow matching method, we set $T=1$ in the remainder of this section.

We again parameterize the target time-dependent vector field $u$ as a deep neural network. Then the key of flow matching is to design a practically implementable training problem of $u$.
To this end, we adopt a similar approach used in diffusion models shown in Section \ref{sec:diffusion-models} and consider some ``conditional'' vector fields\index{Vector field!conditional generative} first.

Notice that transferring $\Ncal (0,I_{d})$ to $\pdata$ is difficult, but transferring $\Ncal (0,I_{d})$ to any specific sample $z$ from $\pdata$ is easy. 
For example, for any fixed $z \sim \pdata$ (i.e., $z$ is a given data sample following $\pdata$), the easiest way of moving from any point $\epsilon \in \mathbb{R}^{d}$ to $z$ in the time window $[0,1]$ is to follow the straight line from $\epsilon$ to $z$ with a constant speed.
Namely, we can move by following $x_{t}$ defined by
\begin{equation}
\label{eq:fm-cond-xt}
x_{t} = (1-t) \epsilon + t z .
\end{equation}
It is clear that $x_{0} = \epsilon$ and $x_{1} = z$.
We also notice two consequences of \eqref{eq:fm-cond-xt}:
\begin{equation}
\label{eq:fm-cond-dxtdt}
\xdot_{t} = z - \epsilon
\end{equation}
and 
\begin{equation}
\label{eq:fm-cond-epsilon}
\epsilon = \frac{x_{t} - tz}{1-t}
\end{equation}
where we assume the right-hand side of \eqref{eq:fm-cond-epsilon} (and alike later in this section) is well defined in the limit sense of $t\to 1^{-}$, and will ensure that this will not cause actual mathematical issue in the upcoming derivations.

Combining \eqref{eq:fm-cond-dxtdt} and \eqref{eq:fm-cond-epsilon}, we obtain 
\begin{equation}
\xdot_{t} = z - \epsilon = \frac{z-x_{t}}{1-t} .
\end{equation}

Now we can define this so-called \emph{conditional vector field} $v^{z}: \mathbb{R}^{d} \times [0,1] \to \mathbb{R}^{d}$ as follows: fix any $z \in \mathbb{R}^{d}$, define 
\begin{equation}
\label{eq:fm-v-t-z}
v_{t}^{z}(x) := \frac{z-x}{1-t}
\end{equation}
for any $x \in \mathbb{R}^{d}$ and $t \in [0,1]$.
The significance is that, from any initial $\epsilon \in \mathbb{R}^{d}$, by following this conditional vector field $v_{t}^{z}$, we will always end up at $z$ at time $t=1$.

We can construct a continuity equation of $\rho^{z}$ using this conditional as well.
More precisely, we let $\rho^{z}_{0} = \Ncal(0,I_{d})$ (can be thought of as a massive amount of particles in $\mathbb{R}^{d}$ following the standard Gaussian distribution) for any $z$, define this continuity equation by
\begin{equation}
\label{eq:fm-cond-ct}
\partial_{t} \rho_{t}^{z} = - \nabla \cdot (v_{t}^{z}\rho_{t}^{z}) .
\end{equation}
Due to the property of $v^{z}$, we know $\rho_{t}^{z} \to \delta_{z}$ ($\delta_{z}$ is the Dirac delta function with center at point $z$) as $t \to 1^{-}$.
Nevertheless, we still have $\rho^{z}_{t}(x) \ge 0$ for all $x$ and $\int \rho_{t}^{z}(x) \, dx = 1$, since $\rho^{z}_{t}$ describes the probability density of these particles (which are driven by $v^{z}$ since time $0$) at time $t$.

A particularly important fact is that, if the initial point $\epsilon$ is drawn from $\Ncal(0, I_{d})$, then we know 
\begin{equation*}
x_{t} = (1-t) \epsilon + t z \sim \Ncal (\,\cdot\,; tz, (1-t)^{2} I_{d}) .
\end{equation*}
Noticing also that $x_{t} \sim \rho_{t}^{z}$, we get 
\begin{equation}
\label{eq:fm-rho-t-z}
\rho_{t}^{z} = \Ncal (tz, (1-t)^{2} I_{d}) .
\end{equation}

Now, with $\rho_{t}^{z}$, we can define the following function of $t$, $x$ and $z$:
\begin{equation}
\label{eq:fm-joint-p}
\rho_{t}(x, z) := \rho_{t}^{z}(x) \pdata(z) .
\end{equation}
We notice that $\rho_{t}$ has several properties that make it a joint probability distribution: $\rho_{t}(x,z) \ge 0$ for all $x,z \in \mathbb{R}^{d}$ and $t \in [0,1]$, and there is
\begin{align*}
\iint \rho_{t}(x,z) \, dz dx
& = \iint \rho_{t}^{z}(x) \pdata(z) \, dz dx \\
& = \int \pdata(z) \Big(\int \rho_{t}^{z}(x) \, dx \Big) dz \\
& = \int \pdata(z) \, dz \\
& = 1
\end{align*}
where we used the fact $\int \rho_{t}^{z}(x) \, dx = 1$ in the third equality.
These confirm that $\rho_{t}(x,z)$ is a joint probability distribution of $(x,z)$ at every $t$.

Without risk of confusion, we also denote the marginal probability of $\rho_{t}(x,z)$ in $x$ at time $t$ as
\begin{equation}
\label{eq:fm-px}
\rho_{t}(x) := \int \rho_{t}(x,z) \, dz  .
\end{equation}
Again, $\rho_{t}(x) \ge 0$ for all $x$ and $t$ and $\int \rho_{t}(x) \, dx = \iint \rho_{t}(x,z) \, dz dx = 1$. 
Moreover, we find that this marginal probability distribution satisfies
\begin{equation*}
\rho_{0}(x) = \int \rho_{0}^{z}(x) \pdata(z) \, dz = \int \Ncal (x;0,I_{d}) \pdata(z) \, dz = \Ncal (x;0,I_{d}) ,
\end{equation*}
where we used \eqref{eq:fm-joint-p} in the first equality, and $\rho_{0}^{z}(x) = \Ncal (x;0,I_{d})$ which is a function of $x$ but not $z$ in the last two equalities, 
and
\begin{align}
\rho_{1}(x) 
& = \lim_{t \to 1^{-}} \rho_{t}(x) \nonumber \\
& = \lim_{t \to 1^{-}} \int \rho_{t}^{z}(x) \pdata(z) \, dz \nonumber \\
& = \int \Big(\lim_{t\to 1^{-}} \rho_{t}^{z}(x) \Big) \pdata(z) \, dz \label{eq:fm-rho1} \\
& = \int \delta_{z}(x) \pdata(z) \, dz  \nonumber \\
& = \pdata(x). \nonumber 
\end{align}
(Rigorously speaking, the exchange of limit and integral in \eqref{eq:fm-rho1} does not hold in general; However, we usually treat $\delta_{z}$ as a Gaussian probability density function with mean $z$ and very tiny variance, and the derivations in \eqref{eq:fm-rho1} still hold true assuming $\pdata$ is a continuous function.)
These two marginal distributions exactly match our initial probability distribution $\Ncal (0,I_{d})$ and target probability distribution $\pdata$, and we want to find out the vector field $u: \mathbb{R}^{d} \times [0,1] \to \mathbb{R}^{d}$, such that the continuity equation
\begin{equation}
\label{eq:fm-ct}
\partial_{t} \rho_{t} = - \nabla \cdot (u_{t} \rho_{t})
\end{equation}
indeed transfers $\rho_{0}= \Ncal(0,I_{d})$ to $\rho_{1} = \pdata$.

To this end, we notice that
\begin{align}
\partial_{t} \rho_{t}(x) 
& = \frac{\partial}{\partial t} \int \rho_{t}(x,z) \, dz \nonumber \\
& = \frac{\partial}{\partial t} \int \rho_{t}^{z}(x) \pdata(z) \, dz \nonumber \\
& = \int \Big( \partial_{t} \rho_{t}^{z}(x) \Big) \pdata(z) \, dz \label{eq:fm-ct-derive} \\
& = - \int \nabla \cdot ( v_{t}^{z}(x) \rho_{t}^{z}(x) ) \pdata(z) \, dz \nonumber \\
& = - \nabla \cdot \Big( \int v_{t}^{z}(x) \rho_{t}^{z}(x)\pdata(z) \, dz \Big) \nonumber 
\end{align}
where we used \eqref{eq:fm-cond-ct} in the fourth equality, and exchanged the order of the divergence $\nabla \cdot$, which is with respect to $x$, and the integral which is in $z$, in the last equality.
Notice that the integral in the last equality is applied to each component of the integrand, as this integrand is a vector.

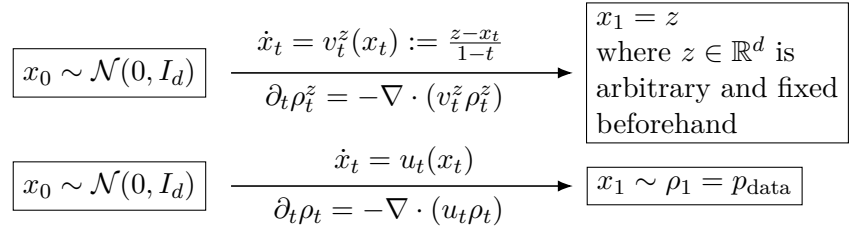
\begin{figure}
\centering
\begin{tikzpicture}[scale=1, transform shape, node distance=1cm, >=Latex]

\node at (-0.5,1) [draw, align=left] {$x_{0} \sim \Ncal(0,I_{d})$};

\draw[->, line width=.6pt] (1.1, 1) -- (5.6,1) node[above left] {$\xdot_{t} = v_{t}^{z}(x_{t}):= \frac{z-x_{t}}{1-t}\quad \quad $ } node[below left] {$\partial_{t} \rho_{t}^{z} = - \nabla \cdot (v_{t}^{z} \rho_{t}^{z})\qquad $ };

\node at (7.5,1) [draw, align=left] {$x_{1} = z$ \\ where $z \in \mathbb{R}^{d}$ is \\ arbitrary and fixed \\ beforehand};

\node at (-0.5,-.5) [draw, align=left] {$x_{0} \sim \Ncal(0,I_{d})$};

\draw[->, line width=.6pt] (1.1, -.5) -- (5.6,-.5) node[above left] {$\xdot_{t} = u_{t}(x_{t})\qquad \quad $ } node[below left] {$\ \partial_{t} \rho_{t} = - \nabla \cdot (u_{t} \rho_{t}) \qquad $ };

\node at (7.2,-.5) [draw, align=left] {$x_{1} \sim \rho_{1} = \pdata$};

\end{tikzpicture}
\caption{The conditional trajectory to an arbitrary and fixed point $z \in \mathbb{R}^{d}$ (top) and the averaged trajectory (bottom) in the flow matching framework. The time $t$ increases from $0$ to $T=1$ in both top and bottom plots. The conditional trajectory is described by the ODE with the conditional vector field $v_{t}^{z}$ or equivalently its continuity equation of $\rho_{t}^{z}$ with $x_{t} \sim \rho_{t}^{z}$ at every $t$, and $x_{t} \to z$ as $t \to 1$ regardless of what $x_{0}$ is. The averaged trajectory is described by the ODE with the target vector field $u_{t}$ or equivalently its continuity equation of $\rho_{t}$ with $\rho_{1} = \pdata$. The target vector field $u_{t}$ is learned by using $v_{t}^{z}$ and $z \sim \pdata$ jointly as in \eqref{eq:fm-loss}.}
\label{fig:fm}
\end{figure}

Comparing \eqref{eq:fm-ct} and \eqref{eq:fm-ct-derive}, we can see that
\begin{equation*}
- \nabla \cdot ( u_{t}(x) \rho_{t}(x) ) = - \nabla \cdot \Big( \int v_{t}^{z}(x) \rho_{t}^{z}(x) \pdata(z) \, dz \Big)
\end{equation*}
In other words, there exists a vector field $w: \mathbb{R}^{d} \times [0,1] \to \mathbb{R}^{d}$, such that $\nabla_{x} \cdot (w_{t}(x) \rho_{t}(x)) = 0$ for all $x$ and $t$, and
\begin{equation}
u_{t}(x) \rho_{t}(x) = \int v_{t}^{z}(x) \rho_{t}^{z}(x) \pdata(z) \, dz + w_{t}(x) \rho_{t}(x).
\end{equation}
Dividing $\rho_{t}(x)$ on both sides, we obtain
\begin{align*}
u_{t}(x)
& = \int v_{t}^{z}(x) \frac{\rho_{t}^{z}(x) \pdata(z)}{\rho_{t}(x)} \, dz + w_{t}(x) \\
& = \int v_{t}^{z}(x) \frac{\rho_{t}(x,z)}{\rho_{t}(x)} \, dz + w_{t}(x) \\
& = \int v_{t}^{z}(x) \rho_{t}(z | x) \, dz + w_{t}(x) ,
\end{align*}
where $\rho_{t}(z | x)$ is the conditional distribution of $z$ given $x$ at time $t$ (with their joint distribution at time $t$ being $\rho_{t}(x,z)$).
We can safely omit $w_{t}$ (because the solution $\rho_{t}$ to the continuity equation \eqref{eq:fm-ct} does not change when $u_{t}$ is replaced with $u_{t} - w_{t}$ provided that $\nabla \cdot (w_{t}\rho_{t}) = 0$) and set 
\begin{equation}
\label{eq:fm-u}
u_{t}(x) = \int v_{t}^{z}(x) \rho_{t}(z | x) \, dz .
\end{equation}
The comparison between the conditional vector field $v_{t}^{z}$ with any prescribed $z \in \mathbb{R}^{d}$ and the target vector field $u_{t}$ is shown in Figure \ref{fig:fm}.

By Theorem \ref{thm:mse}, we know that for any $x$ and $t$, there is
\begin{align*}
u_{t}(x) 
& = \int v_{t}^{z}(x) \rho_{t}(z | x) \, dz  \\
& = \mathbb{E}_{Z \sim \rho_{t}(\cdot|x)} [ v_{t}^{Z}(x) ] \\ 
& = \argmin_{c \in \mathbb{R}^{d}} \mathbb{E}_{Z \sim \rho_{t}(\cdot|x)} [ | c- v_{t}^{Z}(x) |^{2} ] .
\end{align*}
Therefore, when we parameterize the vector field $u: \mathbb{R}^{d} \times [0,1] \to \mathbb{R}^{d}$ as a deep neural network, it can be trained by solving the total loss over $\rho_{t}(x)$ (which is positive at every $x$) and the uniform distribution of $t$ in $[0,1]$:
\begin{equation*}
\min_{u}\ \int_{0}^{1} \mathbb{E}_{X\sim \rho_{t}}\mathbb{E}_{Z \sim \rho_{t}(\cdot|X)} [ | u_{t}(X) - v_{t}^{Z}(X) |^{2} ] \, dt ,
\end{equation*}
which can be written in a more readable form
\begin{equation}
\label{eq:fm-loss-readable}
\min_{u}\ \int_{0}^{1} \iint \rho_{t}(x) \rho_{t}(z|x) | u_{t}(x) - v_{t}^{z}(x) |^{2}  \, dx dz dt .
\end{equation}
Notice that
\begin{equation*}
\rho_{t}(x) \rho_{t}(z|x) = \rho_{t}(x,z) = \rho_{t}^{z}(x) \pdata(z) = \Ncal (x; tz, (1-t)^{2}I_{d}) \pdata(z) ,
\end{equation*}
where the second equality is due to \eqref{eq:fm-joint-p}, and the third equality is due to \eqref{eq:fm-rho-t-z}. 
Substituting this back to \eqref{eq:fm-loss-readable}, we obtain an equivalent problem
\begin{equation}
\label{eq:fm-loss}
\min_{u}\ \int_{0}^{1} \iint \pdata(z) \Ncal (x; tz, (1-t)^{2}I_{d}) | u_{t}(x) - v_{t}^{z}(x) |^{2}  \, dx dz dt .
\end{equation}

The objective function in \eqref{eq:fm-loss} can be further simplified. To see this, we notice that, to obtain a sample $x \sim \Ncal (tz, (1-t)^{2}I_{d})$, we can set
\begin{equation}
\label{eq:fm-loss-x}
x = tz + (1-t)\epsilon
\end{equation}
where $\epsilon \sim \Ncal (0, I_{d})$. 
In light of \eqref{eq:fm-v-t-z}, we know $v_{t}^{z}(x)$ in \eqref{eq:fm-loss} becomes
\begin{equation}
v_{t}^{z}(x) = \frac{z-x}{1-t} = \frac{z - (tz+(1-t)\epsilon)}{1-t} = z - \epsilon .
\end{equation}
Therefore, we can rewrite the minimization problem \eqref{eq:fm-loss} as
\begin{equation}
\label{eq:fm-loss-final}
\min_{u}\ \int_{0}^{1} \iint \pdata(z) \Ncal (\epsilon; 0, I_{d}) | u_{t}(tz + (1-t)\epsilon) - (z-\epsilon) |^{2}  \, d\epsilon dz dt .
\end{equation}
We can use Monte Carlo integration to approximate the loss function:
\begin{equation}
\label{eq:fm-eloss}
\frac{1}{NM}\sum_{t} \sum_{(z,\epsilon)} | u_{t}(tz + (1-t)\epsilon) - (z-\epsilon) |^{2}
\end{equation}
In practical implementation, we draw $N$ time points from the uniform distribution on $[0,1]$. For each time point $t$, we simulate an $\epsilon \sim \Ncal(0,I_{d})$ for each sample $z$ from $\Dcal$ which consists of $M$ samples independently drawn from $\pdata$. We plug all of them into the empirical loss \eqref{eq:fm-eloss} to train $u$. This trained $u$ is an approximation to the desired vector field in \eqref{eq:fm-ode}.

\section{Remarks and References}

\subsection{Sampling Noises in Generative Models}

In the variational autoencoder framework, diffusion model and flow matching methods discussed in Sections \ref{sec:vae}, \ref{sec:diffusion-models} and \ref{sec:flow-matching}, we often end up with a loss function in the form of
\begin{equation}
\label{eq:gm-loss}
\iint \phi(x, \epsilon) \pdata(x) \Ncal(\epsilon; 0, I_{d}) \, d\epsilon dx 
\end{equation}
with some scalar-valued function $\phi$ of $(x,\epsilon)$.
Examples include \eqref{eq:elbo-A}, \eqref{eq:score-err-loss} and \eqref{eq:fm-loss-final}. The time variable $t$ does not need to be taken into consideration here: \eqref{eq:gm-loss} is supposed to be minimized at every $t$, and we just artificially sample $t$ on $[0,T]$ for numerical implementation in practice. Therefore $t$ has a nature different from $(x,\epsilon)$.

We always use Monte Carlo integration to approximate the loss function\index{Loss function} of form \eqref{eq:gm-loss}.
More specifically, we have a set of $m$ (we used $M$ in the previous sections of this chapter, but change to $m$ here as we will consider sampling $n$ Gaussian noises for each given data sample) given samples $\{ x_{i} \in \mathbb{R}^{d}: i\in [m]\}$, which are assumed to be samples independently drawn from an unknown data distribution $\pdata$.
Meanwhile, we need to sample $\epsilon \sim \Ncal (0,I_{d})$, the $d$-dimensional standard Gaussian distribution, independently.

There are two ways to sample $\epsilon$. The first way is to only sample one $\epsilon_{i} \sim \Ncal (0,I_{d})$ for each $x_{i}$, and the second way is to sample many independent $\epsilon$'s for each $x_{i}$. 
In theory, the first way is correct in Monte Carlo integration because it uses $\pdata(x) \Ncal (\epsilon; 0, I_{d})$ as the joint probability distribution of $(x,\epsilon)$ in \eqref{eq:gm-loss}. 
However, the second way is tempting since we can easily draw as many samples as we want from $\Ncal (0,I_{d})$, unlike the samples $x_{i}$'s which are the only ones available. But is there any issue if we do the second way?

For ease of discussion, we consider the case where we sample $n$ independent $\epsilon$'s for each $x_{i}$. More precisely,
suppose the samples we used to form the Monte Carlo integration for \eqref{eq:gm-loss} are given by
\begin{equation}
(x_{i}, \epsilon_{ij}), \quad i \in [m], \ j \in [n] ,
\end{equation}
where $x_{i}$'s and $\epsilon_{ij}$'s are all independent. 
Then we approximate \eqref{eq:gm-loss} by
\begin{equation}
\label{eq:gm-eloss}
\frac{1}{mn} \sum_{i=1}^{m} \sum_{j=1}^{n} \phi(x_{i}, \epsilon_{ij}) .
\end{equation}
Therefore, the difference between the two ways we mentioned above is just that $n=1$ in the first way and $n>1$ in the second way.

It is straightforward to see that the expectation of \eqref{eq:gm-eloss} is \eqref{eq:gm-loss} regardless of the value of $n$.
In other words, \eqref{eq:gm-eloss} is always an unbiased estimate of \eqref{eq:gm-loss}.

Now we want to check the variances of \eqref{eq:gm-eloss}, and see if the variance can be smaller when $n>1$.
To this end, we first derive a general result for similar cases.

\begin{proposition}
\label{prop:gm-noise}
Let $m, n \in \mathbb{N}$ and $\{A_{ij}: i \in [m], \ j\in [n]\}$ be $mn$ scalar-valued random variables satisfying
\begin{enumerate}
\item
$\mathbb{E}[A_{ij}] = 0$ and for all $i\in [m]$ and $j \in [n]$.

\item
$\var[A_{ij}] = v$ and for all $i\in [m]$ and $j \in [n]$ for some $v > 0$.

\item
For every $i \in [m]$, $A_{ij}$ is independent of $\{A_{i'j'}: i'\in [m]\setminus\{i\},\ j \in [n]\}$.

\item
For every $i \in [n]$, $\cov[A_{ij}, A_{ij'}] = c$ for all $j,j'\in [n]$ and $j\ne j'$ for some $c > 0$.
\end{enumerate}
Then 
\begin{align*}
\mathbb{E} \Big[ \frac{1}{mn} \sum_{i=1}^{m} \sum_{j=1}^{n} A_{ij} \Big] = 0 
\end{align*}
and
\begin{align*}
\var \Big[ \frac{1}{mn} \sum_{i=1}^{m} \sum_{j=1}^{n} A_{ij} \Big] = \frac{v + (n-1)c}{mn} .
\end{align*}
In particular, if $n = 1$, then
\begin{equation*}
\var \Big[ \frac{1}{m} \sum_{i=1}^{m} A_{i1} \Big] = \frac{v}{m} .
\end{equation*}
\end{proposition}

\begin{proof}
We first notice that
\begin{equation*}
\mathbb{E} \Big[ \frac{1}{mn} \sum_{i=1}^{m} \sum_{j=1}^{n} A_{ij} \Big] = \frac{1}{mn} \sum_{i=1}^{m} \sum_{j=1}^{n} \mathbb{E} [A_{ij}] = 0 .
\end{equation*}
For every $i \in [m]$, denote
\begin{equation*}
B_{i} := \sum_{j=1}^{n} A_{ij} .
\end{equation*}
Then we know $B_{1}, \dots, B_{m}$ are independent. Furthermore, there are
\begin{equation*}
\mathbb{E}[B_{i}] = \mathbb{E}\Big[\sum_{j=1}^{n} A_{ij} \Big ] = \sum_{j=1}^{n} \mathbb{E}[A_{ij}] = 0 
\end{equation*}
and
\begin{align*}
\var [B_{i}] 
& = \var \Big[ \sum_{j=1}^{n} A_{ij} \Big] \\
& = \sum_{j=1}^{n} \var [A_{ij}] + 2 \sum_{1 \le j < j' \le n} \cov [A_{ij}, A_{ij'}] \\
& = nv + n(n-1)c .
\end{align*}
Therefore,
\begin{align*}
\var \Big[ \frac{1}{mn} \sum_{i=1}^{m} \sum_{j=1}^{n} A_{ij} \Big]
& = \frac{1}{(mn)^{2}}\var \Big[ \sum_{i=1}^{m} \sum_{j=1}^{n} A_{ij} \Big] \\
& = \frac{1}{(mn)^{2}}\var \Big[ \sum_{i=1}^{m} B_{i} \Big] \\
& = \frac{1}{(mn)^{2}}\sum_{i=1}^{m} \var [B_{i}] \\
& = \frac{v + (n-1)c}{mn} .
\end{align*}
Setting $n=1$, we obtain $\var [ \frac{1}{m} \sum_{i=1}^{m} A_{i1} ] = \frac{v}{m}$.
\end{proof}

We can compare the variances of $\frac{1}{m} \sum_{i=1}^{m} A_{i1}$ and $\frac{1}{mn} \sum_{i=1}^{m} \sum_{j=1}^{n} A_{ij}$ by taking their difference:
\begin{align}
\var \Big[ \frac{1}{m} \sum_{i=1}^{m} A_{i1} \Big]
- \var \Big[ \frac{1}{mn} \sum_{i=1}^{m} \sum_{j=1}^{n} A_{ij} \Big]
& = \frac{v}{m} - \frac{v + (n-1)c}{mn} \nonumber \\
& = \frac{(n-1)(v-c)}{mn} \label{eq:gm-eloss-gap}.
\end{align}
Therefore, using $n>1$ can be beneficial when $v>c$ since the variance of the unbiased estimate $\frac{1}{mn} \sum_{i=1}^{m} \sum_{j=1}^{n} A_{ij}$ has smaller variance.

To adapt the result of Proposition \ref{prop:gm-noise} to our problem, we assume \eqref{eq:gm-loss}, i.e., the expectation of \eqref{eq:gm-eloss}, is zero for notation simplicity.
This is because the value of expectation does not affect the value of variance.

Now we set $A_{ij}$ in Proposition \ref{prop:gm-noise} to $\phi(X_{i},E_{ij})$ where $X_{i} \sim \pdata$ and $E_{ij} \sim \Ncal (0,I_{d})$ for all $i\in[m]$ and $j\in[n]$ are independent, then we have the values of $v$ and $c$ in Proposition \ref{prop:gm-noise} given by
\begin{equation*}
v = \var_{(X_{i},E_{ij})} [\phi(X_{i}, E_{ij})] 
\end{equation*}
and
\begin{equation*}
c = \cov_{(X_{i},E_{ij},E_{ij'})} [\phi(X_{i},E_{ij}), \phi(X_{i}, E_{ij'})] .
\end{equation*}
Therefore, we need to check which of $v$ and $c$ is larger.

In general, $\phi(x,\epsilon)$ is a complicated function, such as \eqref{eq:score-err-loss} and \eqref{eq:fm-loss-final}. 
We can see that $\phi$ contains some deep neural networks to be trained in these cases.
Therefore, we cannot make a conclusion on which of $v$ and $c$ is larger in the general case.
However, we notice that the benefit (the value of the gap \eqref{eq:gm-eloss-gap}) may not be significant even if $v-c$ is positive (and fixed): (i) Increasing $n$ makes $\frac{n-1}{n} \to 1$ and therefore the gap \eqref{eq:gm-eloss-gap} becomes saturated for larger $n$ but only the computational cost of \eqref{eq:gm-eloss} increases; and (ii) The larger sample size $m$ (this is determined by the data we have and cannot be changed in the computation of \eqref{eq:gm-eloss}), the smaller the gap \eqref{eq:gm-eloss-gap} is, which suggests that it may not be worth using large $n$ if we already have a large amount of samples.

Nevertheless, we can still consider a simple case where $\phi(x,\epsilon) = x+\epsilon$ where $x$ and $\epsilon$ are both scalar-valued. Then we have $X_{i} \sim \pdata$ in $\mathbb{R}$ and $E_{ij}, E_{ij'} \sim \Ncal (0,1)$ which are all independent.
In this case, we have
\begin{equation*}
v = \var_{(X_{i},E_{ij})} [X_{i}+E_{ij}] = \var[X_{i}] + 1
\end{equation*}
and
\begin{equation*}
c = \cov_{(X_{i},E_{ij}, E_{ij'})} [X_{i}+E_{ij}, X_{i}+E_{ij'}] = \var[X_{i}] .
\end{equation*}
Therefore, we indeed find $v > c$ in this simple case. Moreover, we notice that $X_{i}+E_{ij}$ and $X_{i}+E_{ij'}$ are dependent since their covariance is positive.

This simple example also gives a warning that we need to be cautious when trying to create more ``training samples'' by adding noises or alike to the given ones, as these artificially created samples are statistically dependent (there covariances are nonzero).
As a consequence, some assumptions in the theoretical derivations of the methods we discussed may be invalid.

\subsection{Conditional Generations}

The generative models discussed in this chapter aim at generating new samples that resemble observed data following an unknown probability distribution $p_{\text{data}}$. 
In real-world scenarios, we often want to generate data conditional on some input information $y \in \mathbb{R}^{k}$ (represented as a vector here for ease of discussions below, while it can be a tensor of some specified shape in different methods)  provided by users, such as class labels, attributes, text descriptions, or other modalities. This leads to the problem of \emph{conditional generation}, the goal of which is to generate samples from conditional distribution $\pdata(x|y)$, enabling controllable, targeted, and interpretable generation.

Existing conditional generation frameworks differ mainly in how they represent, learn, and generate samples following $\pdata(x | y)$, but they all strongly based on their corresponding unconditional generation prototypes like those given earlier in this chapter. 
Nevertheless, the common principle they share is injecting the conditioning variable $y$ into the generative process, either explicitly through probabilistic modeling or implicitly through neural network architectures. Below we briefly introduce several existing approaches of conditional generation.
Notice that we also call the process \emph{guided generation}\index{Generation!guided} and $y$ a \emph{guidance} instead of condition which are common terms in the generative model community, as they sometimes help to avoid confusions with the various conditional probability distributions in derivations.

The conditional GANs is an approach that extends GANs by conditioning both the generator $g_{\theta}$ and discriminator $a_{\eta}$ on $y$. The generator $g_{\theta}$ is parameterized as a neural network with parameter $\theta$ and learns the mapping from a naive sample $z$ (often drawn from $\Ncal(0,I_{d})$ on $\mathbb{R}^{d}$ as in GANs) and a guidance $y \in \mathbb{R}^{k}$ to data space. 
As a result, the generator $g_{\theta}$ can be used to generate samples following the induced conditional probability $\ptheta(x|y)$. Meanwhile, the discriminator is parameterized as an adversarial network $a_{\eta}(\cdot|y)$ that distinguishes real from fake samples given $y$. The training of parameters $(\theta,\eta)$ is implemented by solving the saddle point problem\index{Saddle point problem}
\begin{align}
\min_{\theta} \max_{\eta} \; & \Big\{\iint \pdata(x,y) \log a_{\eta}(x|y) \, dx dy  \nonumber \\
& \qquad \qquad + \iint p(y)\ptheta (x|y) \log (1- a_{\eta}(x|y)) \, dx dy \Big\}, \label{eq:cgan}
\end{align}
where in practical implementations the first integral in \eqref{eq:cgan} is approximated by Monte Carlo integration on the given dataset consisting of independent sample pairs $(x,y)$ drawn from $\pdata$, and $y$ is the guidance corresponding to $x$. 
The probability $p(y)$ in \eqref{eq:cgan} refers to the probability distribution of classes of guidances and is implemented by sampling all guidance of interest during the training process.
Conditional GANs are particularly effective for sharp sample generation and are widely used in class-conditional synthesis and image-to-image translation.

We can also apply the idea of sampling the conditional probability $\pdata(\cdot|y)$ in the flow matching framework. 
More specifically, we can replace the flow probability $\rho_{t}(x)$ in the flow matching method (Section \ref{sec:flow-matching}) with the conditional counterpart $\rho_{t}(x|y)$ with guidance $y$. 
To this end, we first notice that
\begin{equation}
\label{eq:cfm-p}
\rho_{t}(x|y) = \int \rho_{t}(x|y,z) \pdata(z | y) \, dz = \int \rho_{t}^{z}(x) \pdata(z | y) \, dz  
\end{equation}
where $\rho_{t}^{z}$ is the probability flow with a given target $z$ in \eqref{eq:fm-rho-t-z}, and $\pdata(z|y)$ will again be replaced with given samples $z$ labeled by guidance variable $y$ in the approximation of \eqref{eq:cfm-p} by Monte Carlo integration.
Notice that we assumed in \eqref{eq:cfm-p} that $\rho_{t}(x|y,z) = \rho_{t}(x|z) =\rho_{t}^{z}(x)$ as the path from $x$ to any specified $z \sim \pdata$ is determined by $x$ and $z$, not $y$, because the information about $y$ is fully contained in $z$.
Similar to the derivations in Section \ref{sec:flow-matching}, we can show that the marginal distributions at the start time $t=0$ and end time $t=1$ are
\begin{equation*}
\rho_{0}(\cdot | y) = \Ncal(0, I_{d}) \quad \text{and} \quad \rho_{1}(\cdot | y ) = \pdata(\cdot | y) ,
\end{equation*}
and the guided vector field $u_{t}(\cdot|\cdot): \mathbb{R}^{d} \times \mathbb{R}^{k} \to \mathbb{R}^{d}$ is
\begin{equation*}
u_{t}(x | y) = \int v_{t}^{z}(x) \rho_{t}(z | x, y) \, dz = \argmin_{c \in \mathbb{R}^{d}}\mathbb{E}_{Z \sim \rho_{t}(\cdot |x,y)} \Bigl[ | c - v_{t}^{Z}(x)|^{2} \Bigr] ,
\end{equation*}
which means that $u_{t}$, when parameterized as a deep neural network, can be trained by solving
\begin{equation}
\label{eq:cfm-loss}
\min_{u} \ \int_{0}^{1} \mathbb{E}_{(X,Y) \sim \pdata} \mathbb{E}_{Z \sim \rho_{t}(\cdot|X,Y) } \Bigl[|u_{t}(X|Y) - v_{t}^{Z}(X)|^{2} \Bigr] \, dt.
\end{equation}
Notice that the minimization is again with respect to network parameters of $u$.
The literature shows that such conditional generation is most effective in applications where each guidance $y$ has a large amount of samples $x$ in the given dataset, such as in class guidance \cite{nichol2021improved}. However, this can be a challenge in general since many guidance signal $y$'s are non-repeating and complex, such as image captions \cite{lipman2024flow}.

Diffusion models can also be modified into conditional variants. There have been classifier guidance \cite{song2021score-based,dhariwal2021diffusion} and classifier-free guidance \cite{ho2021classifier-free} methods developed to construct such variants. 
Specifically, by Bayes rule, we know
\begin{equation*}
\rho_{t}(x|y) = \frac{\rho_{t}(y|x) \rho_{t}(x)}{p(y)}
\end{equation*}
where $p(y)$ is the distribution of guidance as above, and $\rho_{t}(x)$ and $\rho_{t}(x|y)$ are the unconditional and conditional probability distributions as above. 
Taking logarithm and then gradient with respect to $x$ on both sides, we obtain
\begin{equation}
\label{eq:cdm}
\underbrace{\nabla_{x} \log \rho_{t}(x|y)}_{\text{conditional score}} = \underbrace{\nabla_{x} \log \rho_{t}(y|x)}_{\text{classifier score}} + \underbrace{\nabla_{x} \log \rho_{t}(x)}_{\text{unconditional score}} .
\end{equation}
Therefore, as long as we have the classifier score, we know the desired conditional score and can train network $s_{t}(x|y)$ to approximate it following the derivations in diffusion models.

Based on this relation, classifier guidance methods train a time-dependent function $\phi_{t}(y|x)$, parameterized as a deep neural network, to approximate the classifier score $\nabla_{x} \rho_{t}(y|x)$.
We remark that this function $\phi$ is trained separately from the unconditional score in diffusion models, and more importantly, it is time-dependent and capable of approximating the classifier score when $x \sim \rho_{t}(\cdot)$ is a sample in the intermediate of the diffusion process.
It is shown in \cite{lipman2024flow} that the guided vector field is
\begin{equation}
\label{eq:gvf-fm}
u_{t}(x|y) = u_{t}(x) + \frac{1-t}{t} \phi_{t}(y|x) 
\end{equation}
when using the line interpolation as in \eqref{eq:fm-cond-xt} in the diffusion model, while providing a formula table of customizable interpolation coefficients.
This approach appears to approximate the guided vector field better than \eqref{eq:cfm-loss} for both class and text conditioning \cite{nichol2022glide:}.

Since the classifier score and the unconditional score are learned separately, it is often necessary to calibrate the weight on the classifier score with a scalar $\gamma \in \mathbb{R}$ on $\phi_{t}$ in the second term on the right-hand side of \eqref{eq:gvf-fm}. This scalar is typically chosen to be $\gamma>1$ \cite{dhariwal2021diffusion}.

An alternative approach called classifier-free guidance (CFG) method, which does not need to train a separate classifier score function, was proposed in \cite{ho2021classifier-free}.
From \eqref{eq:cdm}, we see that the classifier score is the difference between the conditional score and the unconditional score. 
Therefore, the CFG method only trains a single network to approximate the conditional score $\nabla_{x} \rho_{t}(x|y)$, and takes a fixed vector $y_{\mathrm{null}}$ as the place-holder when the input sample $z$ does not have a guidance signal (i.e., $z \sim \pdata(\cdot)$). This is shown to result in the following guided vector field \cite{lipman2024flow}:
\begin{equation}
u_{t}(x|y) = (1- \gamma) u_{t}(x| y_{\mathrm{null}}) + \gamma u_{t}(x|y) .
\end{equation}
The loss function for training $u$ includes another layer of data selection: one first draw either $y_{\mathrm{null}}$ or a real guidance $y$ based on Bernoulli($p$) with a user-chosen $p \in (0,1)$, and then sample $z$ without guidance if $y_{\mathrm{null}}$ is drawn, or with the drawn guidance $y$ otherwise. The following steps are the same as in diffusion models \cite{ho2021classifier-free}. The exact distribution which the CFG method samples from is unknown. Some followup works proposed different intuitive or theoretical justifications for CFG sampling, see, e.g., \cite{chidambaram2024what,guo2024gradient,bradley2024classifier}.

\subsection{References of Generative Models}

Early work on generative modeling was dominated by classical probabilistic approaches, such as Gaussian mixture models, hidden Markov models, and other latent variable models with manually designed structures. While these methods are mathematically well-founded and interpretable, they often struggle to scale to high-dimensional data such as images, audio signals, and natural language. The integration of deep neural networks as flexible function approximators marked a major shift, enabling generative models to capture complex, multimodal data distributions more effectively.

The concept of autoencoder was presented in early works \cite{rumelhart1988learning,hinton2006reducing}, whereas the goal was to establish a low-dimensional representations using latent variables to capture the essence of given high-dimensional data. 
VAE was first introduced in \cite{kingma2014auto-encoding}. A comprehensive introduction of VAE is given in \cite{kingma2019introduction}. VAE employs an encoder-decoder structure that maps data into a continuous latent space and reconstructs samples through a generative decoder. This formulation enables efficient sampling, structured representation learning, and stable optimization. However, VAEs are often observed to produce overly smooth or blurry samples, reflecting a trade-off between likelihood-based objectives and perceptual fidelity.

A substantial amount of improvements and extensions of VAE have been proposed. For example, conditional VAE was developed and showed that input conditions can be represented in different manners in latent spaces after encoding \cite{sohn2015learning}.
In \cite{sonderby2016ladder}, ladder VAEs are constructed to recursively correct the generative distribution with a data-dependent approximate likelihood term.
An autoregressive model over the output variables was proposed in \cite{gulrajani2017pixelvae}.
The quality of generated samples is shown to improve by using an adversarial loss in \cite{larsen2016autoencoding}.
There are also many different forms of the variational approximation to the posterior investigated in the past years, including normalizing flows \cite{rezende2015variational,kingma2016improved}, graphical models \cite{maaloe2016auxiliary}, and models for temporal data \cite{gregor2019temporal,gregor2016towards}.

Generative adversarial networks (GANs) was originally introduced in \cite{goodfellow2014generative}. Reviews of GANs and variants are provided in \cite{gui2021review,wang2021generative,mao2007stochastic}. The original GANs are known to be unstable in training \cite{farnia2020gans,jin2020local}, and a large amount of variants and improvements were proposed since then \cite{metz2017unrolled,gulrajani2017improved,arjovsky2017wasserstein,qi2020loss,mao2017least,zhao2017energy}. In particular, Wasserstein GANs was developed in \cite{arjovsky2017wasserstein} and several variants were proposed \cite{wu2018wasserstein,adler2018banach}.

%Normalizing flow models provide another likelihood-based approach by constructing invertible transformations between simple base distributions and complex data distributions. By design, these models allow exact likelihood computation and efficient sampling. Representative examples include RealNVP and related flow-based architectures \cite{rezende2015flows,dinh2017realnvp}. While normalizing flows offer attractive theoretical properties, the requirement of invertibility imposes architectural constraints that may limit expressiveness, particularly in very high-dimensional settings.

Early work illustrating the fundamental idea of diffusion models was presented in \cite{sohl2015deep}, which introduced the concept of progressively diffusing data into noise via a Markov chain and learning a reverse-time process to recover the data distribution. The practical success of diffusion models was significantly advanced by the work of Denoising Diffusion Probabilistic Models (DDPMs) developed in \cite{ho2020denoising}. In \cite{nichol2021improved}, DDPMs were improved through better noise schedules and architectural refinements, leading to substantial gains in sample fidelity. Furthermore, the work \cite{karras2022elucidating} analyzed design choices in diffusion models and proposed principled parameterizations that improve training stability and efficiency. These advances have contributed to diffusion models surpassing GANs in many benchmark tasks, especially in terms of mode coverage and training robustness. The connection between DDPMs and score-based generative modeling represented by SDEs was established in \cite{song2021score-based}. Deterministic versions represented by ODEs were given \cite{song2020denoising}. 
Beyond image generation, diffusion models have been successfully extended to conditional generation, super-resolution, text-to-image synthesis, and scientific applications. Notable examples include classifier-guided diffusion \cite{dhariwal2021diffusion} and large-scale text-conditioned models such as latent diffusion models \cite{rombach2022high}, which reduce computational cost by operating in a learned latent space.

Flow matching is a simplified and improved framework for generative modeling \cite{liu2023flow,lipman2023flow,albergo2023building} compared to diffusion models and has seen various applications of large-scale generation in speech \cite{le2023voicebox:}, images \cite{esser2024scaling}, videos \cite{polyak2024movie}, robotics \cite{black2024vision-language-action}, and proteins \cite{huguet2024sequence-augmented}. It was shown that the linear version of Flow Matching can be seen as a certain limiting case of bridge matching in \cite{liu2023l2sb,shi2023diffusion}. The idea of using vector field in flow matching is related to probability density matching in \cite{chen2018neural,grathwohl2019ffjord:} which are based on likelihood. Recent improvements and extensions are presented in \cite{heitz2023iterative,tong2023improving,neklyudov2023action}. Flow matching has also been considered in the discrete-state scenarios in \cite{gat2024discrete,campbell2024generative} and on Riemann manifolds \cite{chen2024flow}. The flow matching framework was shown to be applicable to general continuous-time Markov processes \cite{holderrieth2025generator}.

Overall, the generative models including all the aforementioned variants represent a flexible and theoretically grounded paradigm for generative modeling, with ongoing research aimed at accelerating sampling, improving scalability, and extending their applicability to broader data modalities.

\newpage

\appendix

\chapter{Supplementary Materials}

\section{Monte Carlo Integration}

\label{appsec:mc}

Monte Carlo integration is a computationally feasible approach to approximate integrals in subsets of high-dimensional Euclidean spaces.
For such integrals, classical numerical integration methods do not apply due to the curse of dimensionality issue arising form spatial discretization of the integral domains.
Monte Carlo integration, on the other hand, does not need explicit spatial discretization.
Instead, it draws sample points from the integration domain by following certain probability distribution, and evaluate the function to be integrated at these sample points. 
Then a proper averaging of the function values at these points often yields reasonable approximation to the desired integral.
In this section, we provide some details about the Monte Carlo integration methods and related mathematical properties.

The basic idea of Monte Carlo integration\index{Monte Carlo integration} is as follows. If $\Omega \subset \mathbb{R}^d$ is bounded and $h: \Omega \to \mathbb{R}$ is integrable on $\Omega$, then we can approximate its integral by
\begin{equation}
\label{eq:def-Ih-JhN-simple}
\int_{\Omega} h(x) \, dx \approx \frac{|\Omega|}{N} \sum_{i=1}^{N} h(x^{(i)}) ,
\end{equation}
where $\{x^{(i)} \in \Omega: i \in [N]\}$ are $N$ samples drawn independently from the uniform distribution in $\Omega$, and $|\Omega|$ is the volume of $\Omega$ in $\mathbb{R}^{d}$.
Next, we consider a slight generalization of this approach, and show the bias and variance of such Monte Carlo integral approximations.

The generalization is known as \emph{importance sampling}\index{Importance sampling}. Let $\Omega$ be a possibly unbounded subset of $\mathbb{R}^{d}$ and $p$ a probability density function defined on $\Omega$ with $p(x)>0$ for all $x\in\Omega$.
Let $h: \Omega \to \mathbb{R}$ be square integrable in $\Omega$, and $X^{(1)},\dots,X^{(N)}$ be $N$ i.i.d.\ random variables drawn from the probability $p$.
%
% Consider the following estimator $J_{N}$ of the integral $\int_{\Omega } h(x) dx$:
%
Denote 
\begin{equation}
\label{eq:def-Ih-JhN}
I_{h}: = \int_{\Omega} h(x) \, dx  \quad \text{and} \quad J_{h}^{N} := \frac{1}{N}\sum_{i=1}^N\frac{h(X^{(i)})}{p(X^{(i)})}.
\end{equation}
Then $J_{h}^{N}$ is a statistical estimate of the desired integral $J_{h}$.
Note that, $J_{h}^{N}$ is a statistic with sampling complexity $N$, and $I_{h}$ is a fixed unknown number to be approximated by the former.

We see that \eqref{eq:def-Ih-JhN-simple} is a special case of \eqref{eq:def-Ih-JhN} with uniform distribution $p(x)=1/|\Omega|$ for all $x \in \Omega$. However, \eqref{eq:def-Ih-JhN-simple} requires $\Omega$ to be a bounded subset of $\mathbb{R}^{d}$, whereas \eqref{eq:def-Ih-JhN} allows applications for unbounded $\Omega$.

Now let us check the approximation error of $J_{h}^{N}$ to $I_{h}$.
First, we notice that 
\begin{align*}
\mathbb{E}[J_{h}^{N}] 
& = \mathbb{E} \Big[ \frac{1}{N} \sum_{i=1}^{N} \frac{h(X^{(i)})}{p(X^{(i)})} \Big]
= \frac{1}{N} \sum_{i=1}^{N}\mathbb{E}_{X^{(i)} \sim p} \Big[  \frac{h(X^{(i)})}{p(X^{(i)})} \Big] \\
& = \frac{1}{N} \sum_{i=1}^{N} \int_{\Omega} \frac{h(x)}{p(x)} p(x) \, dx 
= J_{h} ,
\end{align*}
which means that $J_{h}^{N}$ is an unbiased estimator of $I_{h}$.

Furthermore, we have for each $X^{(i)}$ that
\begin{equation*}
\var\Big[ \frac{h(X^{(i)})}{p(X^{(i)})} \Big] 
= \mathbb{E}\Big[ \Big| \frac{h(X^{(i)})}{p(X^{(i)})} \Big|^{2} \Big] - \Big| \mathbb{E}\Big[  \frac{h(X^{(i)})}{p(X^{(i)})}  \Big] \Big|^{2} = \int_{\Omega} \frac{h(x)^{2}}{p(x)}\,dx - I_{h}^{2} .
\end{equation*}
Hence, there is 
\begin{align}
\var[J_{h}^{N}] 
& = \var \Big[ \frac{1}{N} \sum_{i=1}^{N} \frac{h(X^{(i)})}{p(X^{(i)})} \Big]
= \frac{1}{N} \var \Big[ \frac{h(X^{(i)})}{p(X^{(i)})} \Big] \nonumber \\
& = \frac{1}{N} \Big( \int_{\Omega} \frac{h(x)^{2}}{p(x)} \, dx  - I_{h}^{2} \Big) = O \Big( \frac{1}{N} \Big) , \label{eq:var-JhN}
\end{align}
which means the variance of $J_{h}^{N}$ reduces at the order of $1/N$, where $N$ is the number of sampled points.
Meanwhile, we have
\begin{equation*}
I_{h}^{2} = \Big( \int_{\Omega} h(x) dx \Big)^{2} = \Big( \int_{\Omega }\frac{h(x)}{p(x)^{1/2}} p(x)^{1/2} \, dx \Big)^{2} \le  \int_{\Omega} \frac{h(x)^{2}}{p(x)} \, dx ,
\end{equation*}
and the equality holds only if $p \propto h$, in which case the constant term of the variance $\var[J_{h}^{N}]$ in \eqref{eq:var-JhN} is minimized.
However, it is in general not possible to have $p \propto h$, and the idea of importance sampling is to use a relatively simple probability distribution, such as the mixture of Gaussians, as $p$ and make it as close to $h$ as possible.
The reason of setting $p$ as a mixture of Gaussians is that one can easily sample points $X^{(i)}$'s from $p$ and evaluate $h(X^{(i)})/p(X^{(i)})$ to compute $J_{h}^{N}$ in \eqref{eq:def-Ih-JhN}.
This can be done by setting the means of these Gaussians to the points where $h$ attains peak values, the standard deviations according to the landscape of $h$, and the mixing weights such that $p$ is near proportional to $h$.

Overall, we can see that by increasing the sample number $N$ and setting $p$ properly, the approximation error of $J_{h}^{N}$ to the true value $I_{h}$ can be made smaller.
Meanwhile, the rate $O(1/N)$ does not depend on the dimension $d$ of the integral and hence such approximated integrals are potential to mitigate the curse of dimensionality issue in the cases where spatial discretization is utilized.

We remark that the real-world situations of using Monte Carlo integrations can be more complicated. 
Specifically, we often need to use Monte Carlo integration to approximate the gradient of some objective functions of form
\begin{equation*}
I_{h}(\theta) := \frac{1}{2} \int_{\Omega} |h(x, \theta)|^{2} dx ,
\end{equation*}
for some $h: \mathbb{R}^{d} \times \mathbb{R}^{m} \to \mathbb{R}^{n}$, where $\theta \in \mathbb{R}^{m}$ represents the neural network parameter to be optimized.
As most network training algorithms are first-order optimization methods, we need to approximate
\begin{equation*}
\nabla_{\theta} I_{h}(\theta) = \int_{\Omega} \nabla_{\theta}  h(x,\theta) h(x,\theta) \, dx ,
\end{equation*}
which is a vector of dimension $m$.
When we use Monte Carlo integration, the variances at different components of $\nabla_{\theta} I_{h}(\theta)$ may vary and we need to be careful about choosing a suitable probability distribution $p$ for importance sampling.
The variances of the stochastic gradient approximating $\nabla_{\theta} I_{h}(\theta)$ play important roles in the convergence of the stochastic gradient method and alike, as shown in Chapter \ref{chpt:opt}.

\section{Banach Space and Fixed Point Theory}
\label{appsec:banach}

In this section, we present a crush review of Banach spaces and fixed point theory. The results given in this section are fundamental to the theory of Bellman operators and Markov decision processes in Section \ref{sec:rl-fundamentals}.

We first provide the definition of Banach spaces, which are complete normed linear spaces. We give the definitions of these terms and then the formal definition of Banach spaces afterwards.
Then we present the Banach fixed point theorem. Finally, we show that $B(\Xcal)$ considered in \ref{sec:rl-fundamentals} is a Banach space, on which Banach fixed point theorem can be applied.

\begin{definition}
[Linear space]
Let $X$ be a set. We call $X$ a \emph{linear space}\index{Linear space} (also called \emph{vector space}) if it has summation and scalar-multiplication defined (i.e., $x+y \in X$ and $ax\in X$ for all $x,y\in X$ and $a \in \mathbb{R}$), such that for any $x,y,z \in X$ and $a,b\in \mathbb{R}$, the following statements hold.
\begin{enumerate}
\item
$x+y = y+x$.
\item
$(x+y) + z = x + (y+z)$.
\item
There exists $ 0 \in X$ such that $x+0 = 0+x$ for all $x \in X$.
\item
For any $x\in X$, there exists $x' \in X$, called $-x$, such that $x + x' = 0$.
\item
$a(b x) = (ab) x$.
\item
$1\cdot x = x$ where $1 \in \mathbb{R}$.
\item
$(a+b) x = a x + b x$.
\item
$a(x+y) = a x + a y$.
\end{enumerate}
\end{definition}

\begin{definition}
[Norm]
The function $\| \cdot \|: X \to \mathbb{R}$ is called a \emph{norm}\index{Norm} if for any $x,y,z \in X$ and $a \in \mathbb{R}$ the following statements hold:
\begin{enumerate}
\item
(Positive definite) $\|x \| \ge 0$; and $\| x \| =0$ iff $x =0$.
\item
(Homogeneous) $\| ax\| = |a|\|x\|$.
\item
(Triangle inequality) $\|x+y\| \le \|x\| + \|y\|$.
\end{enumerate}
\end{definition}
If the linear space $X$ has a norm, then we call $(X, \|\cdot\|)$ a \emph{normed linear space}\index{Linear space!normed}.

\begin{definition}
[Distance]
\label{def:distance}
Let $X$ be a set. We call $d: X \times X \to \mathbb{R}$ a \emph{distance} (or \emph{metric}) if for any $x,y,z \in X$, there are
\begin{enumerate}
\item $d(x,y) \ge 0$; and $d(x,y) = 0$ iff $x=y$;
\item $d(x,y) = d(y,x)$; and
\item $d(x,z) \le d(x,y) + d(y,z)$ .
\end{enumerate}
\end{definition}

If $X$ is a linear space with norm $\| \cdot \|$, then the function $d: X \times X \to \mathbb{R}$ defined by $d(x,y) := \| x - y \|$ is a distance\index{Distance} (this can be easily verified by checking the conditions in Definition \ref{def:distance}). In what follows, we always assume the distance $d$ on a normed linear space $(X, \|\cdot\|)$ is defined this way. 

\begin{definition}
[Cauchy sequence]
A sequence $\{x_{k}\}$ in a normed linear space $(X, \|\cdot\|)$ is called a \emph{Cauchy sequence}\index{Cauchy sequence} if for any $\epsilon>0$, there exists $K \in \mathbb{N}$, such that $\|x_{k}-x_{j}\| < \epsilon$ for all $k,j \ge K$. 
\end{definition}

\begin{definition}
[Banach space]
\label{def:banach-space}
We call a normed linear space $(X,\|\cdot\|)$ \emph{complete} if every Cauchy sequence in $X$ converge to some point in $X$. A complete normed linear space $(X, \|\cdot\|)$ is called a \emph{Banach space}\index{Banach space}.
\end{definition}

We will simply write $(X, \| \cdot \|)$ as $X$ provided the norm $\|\cdot\|$ is clear from the context. Now we consider operations on Banach spaces.

\begin{definition}
[Lipschitz, Contractive, and Fixed Point]
Let $X$ be a Banach space. A mapping (operation) $T: X \to X$ is called \emph{Lipschitz} with constant $\gamma>0$ on $X$ if 
\begin{equation*}
\| Tx - T\hat{x} \| \le \gamma \| x - \hat{x}\|
\end{equation*}
for any $x, \hat{x} \in X$. If $\gamma = 1$, we call $T$ \emph{non-expansive}. If $\gamma \in (0,1)$, we call $T$ a \emph{$\gamma$-contraction} (or $T$ is \emph{$\gamma$-contractive}). If $x \in X$ satisfies $Tx = x$, then $x$ is called a \emph{fixed point of $T$}.
\end{definition}

\begin{theorem}
[Banach fixed point theorem]
\label{thm:banach-fixed-pt}
Let $X$ be a Banach space and $T: X \to X$ a $\gamma$-contraction with some $\gamma \in (0,1)$. Then the following statements hold true:\index{Banach fixed point theorem}
\begin{itemize}
\item[(i)]
$T$ has a unique fixed point $x^{*}$ in $X$.

\item[(ii)]
For any $x \in X$, define sequentially $x_{k+1} = Tx_{k}$ with $x_{0} = x$ (this scheme of generating a sequence is called the \emph{fixed point iteration}). Then for any $k \in \mathbb{N}$, there are
\begin{equation}
\label{eq:fpi-linear-rate}
\|x_{k+1} - x_{k} \| \le \gamma \| x_{k} - x_{k-1} \| ,
\end{equation}
and 
\begin{equation}
\label{eq:fpi-linear-rate-x-star}
\| x_{k} - x^{*} \| \le \gamma^{k} \| x - x^{*} \| .
\end{equation}
We call $x_{k}$ converges to $x^{*}$ at a linear rate\index{Convergence!linear} of $\gamma$ given \eqref{eq:fpi-linear-rate-x-star}.
\end{itemize}
\end{theorem}

\begin{proof}
We prove the two statements together. Let $x \in X$ be arbitrary and $x_{0} = x$. Let $\{x_{k}\}$ be the sequence generated by the fixed point iteration using the $\gamma$-contractive $T$. 
Then \eqref{eq:fpi-linear-rate} is obvious due to $T$. Furthermore, for any $k \in \mathbb{N}$, there is
\begin{align*}
\|x_{k+1} - x_{k} \| 
& = \| Tx_{k} - Tx_{k-1} \| \\
& \le \gamma \| x_{k} - x_{k-1} \| \\
& \le \cdots \\
& \le \gamma^{k} \| x_{1} - x_{0} \| .
\end{align*}
Then, for any $j \in \mathbb{N}$, we have
\begin{align*}
\| x_{k+j} - x_{k} \| 
& \le \|x_{k+j} - x_{k+j-1} \| + \dots + \|x_{k+1} - x_{k} \| \\
& \le (\gamma^{k+j-1}+\dots+\gamma^{k}) \| x_{1} - x_{0} \| \\
& \le \frac{\gamma^{k}}{1-\gamma} \|x_{1} - x_{0} \|. 
\end{align*}
Hence, $\| x_{k+j} - x_{k} \| $ can be arbitrarily small as long as $k$ is sufficiently large despite of $j$. Therefore $\{x_{k}\}$ is a Cauchy sequence in $X$. Since $X$ is complete, we know $x_{k}$ converges to some $x^{*} \in X$.  
Moreover, we know $\lim_{k\to \infty}\|Tx_{k} -Tx^{*}\| = \lim_{k\to \infty}\|x_{k+1} -x^{*}\| = 0$, and hence $\|Tx^{*} - x^{*} \| = \lim_{k\to \infty} \|Tx_{k} - x_{k}\| = 0$, which implies that $x^{*} = Tx^{*}$, i.e., $x^{*}$ is a fixed point of $T$.

Now we have 
\begin{align*}
\|x_{k} - x^{*} \| 
& = \| Tx_{k-1} - Tx^{*} \| \\
& \le \gamma \| x_{k-1} - x^{*} \| \\
& \le \cdots \\
& \le \gamma^{k} \| x_{0} - x^{*} \| \\
& = \gamma^{k} \| x - x^{*} \| ,
\end{align*}
since $x_{0} = x$, which proves \eqref{eq:fpi-linear-rate-x-star}.

If $x^{*}$ and $\hat{x}$ are both fixed points of $T$ but different, then $\|x^{*} - \hat{x}\|>0$. However
\[
\|x^{*} - \hat{x}\| = \|Tx^{*} - T\hat{x}\| \le \gamma \|x^{*} - \hat{x}\| < \|x^{*} - \hat{x}\|,
\]
which is a contradiction. Hence $x^{*}$ is the unique fixed point of $T$.
\end{proof}

Now we consider the space $B(\Xcal)$ in Section \ref{sec:rl-fundamentals}. Recall that $\Xcal$ is a set, which can be a set of finitely many or countably many elements, or a subset of some Euclidean space. Then we define
\begin{equation}
B(\Xcal) := \Big\{ v : \Xcal \to \mathbb{R}: \sup_{x \in \Xcal} |v(x)|  < \infty \Big\} .
\end{equation}
Namely, $B(\Xcal)$ is the set of all uniformly bounded real-valued functions defined on $\Xcal$. We also define the $\infty$-norm on $B(\Xcal)$:
\begin{equation}
\| v\|_{\infty} := \sup_{x \in \Xcal} |v(x)| .
\end{equation}
Then it is clear that $(B(\Xcal), \|\cdot\|_{\infty})$ is a normed linear space. Next, we want to show that $(B(\Xcal), \| \dot \|_{\infty})$ is complete.

To this end, let $\{v_{k}\}$ be a Cauchy sequence in $(B(\Xcal), \|\cdot\|_{\infty})$. That is, for any $\epsilon > 0$, there exists $K \in \mathbb{N}$, such that $\| v_{k} - v_{j} \|_{\infty} < \epsilon$ for all $k, j \ge K$. Therefore, for any $x \in \Xcal$, we have
\begin{equation}
\label{eq:BX-cauchy-ineq}
|v_{k}(x) - v_{j}(x) | \le \| v_{k} - v_{j} \|_{\infty} < \epsilon
\end{equation}
for all $k,j \ge K$. This implies that $\{v_{k}(x)\}$ is a Cauchy sequence of numbers in $\mathbb{R}$. 
Hence $\{v_{k}(x)\}$ converges and has a limit, denoted by $v(x)$, in $\mathbb{R}$. 
Define $v: \Xcal \to \mathbb{R}$ as such limit at every $x \in \Xcal$. Then, for every $x\in \Xcal$, we notice that by letting $j \to \infty$ in \eqref{eq:BX-cauchy-ineq}, we have
\begin{equation*}
|v_{k}(x) - v(x) | \le \epsilon
\end{equation*}
as long as $k \ge K$. 
Since $x$ is arbitrary, we obtain
\begin{equation}
\label{eq:BX-cauchy-limit}
\| v_{k} - v \|_{\infty} \le \epsilon 
\end{equation}
for all $k \ge K$. As $\epsilon$ is arbitrary, we know $\|v_{k} - v\|_{\infty} \to 0$ as $k \to \infty$. 
Moreover, we see that from \eqref{eq:BX-cauchy-limit}, that for any $k \ge K$, there is $\|v\|_{\infty} \le \epsilon + \|v_{k} \|_{\infty} < \infty$, which implies that $v \in B(\Xcal)$. 
To conclude, we have that $\{v_{k}\}$ is convergent in $\infty$-norm and the limit $v$ is in $B(\Xcal)$. Therefore, $B(\Xcal)$ is complete. By Definition \ref{def:banach-space}, $B(\Xcal)$ is a Banach space.

\section{Basics of Information Theory}
\label{appsec:info-theory}

In this section, we provide a crush review of the basics of information theory and its relations to probability.

\begin{definition}
[Entropy]
\label{def:entropy}
Let $X$ be a random variable following a probability $p$ on $\mathbb{R}^{n}$. Then the \emph{entropy}\index{Entropy} of $X$ is given by
\begin{equation}
\label{eq:entropy}
H(X) := \mathbb{E}_{X \sim p}[ - \log p(X) ] = - \int p(x) \log p(x) \, dx .
\end{equation}
If $X$ is a discrete random variable, i.e., the set $\Xcal$ of all possible values of $X$ is finite or countably infinite, then the integral in \eqref{eq:entropy} is replaced with the sum over $\Xcal$. We use integral throughout this section unless a formula only applies to summations. 
Notice that we always use the convention $0\log 0 = 0$ given that $\lim_{z \to 0^{+}} z \log z = 0$.
\end{definition}

The entropy of a random variable (or equivalently its probability distribution) quantifies the average level of uncertainty. The entropy is invariant to the mean (expectation) of the random variable, namely, if the mean of the random variable shifts, the entropy does not change. The variance and dimensionality of the random variables are main factors to determine the entropy, but there are also other factors that affect the value of entropy for general random variables.

If $X$ is a discrete random variable, then $H(X) \ge 0$. To see this, we notice that $p(x) \log p(x) \in [-e^{-1},0]$ at every $x \in \Xcal$ since $p(x) \in [0,1]$.

If $X$ is a continuous random variable, then $H(X)$ can be negative. A special and important example is the entropy of $d$-dimensional Gaussian random variable $X \sim \Ncal(\mu, \Sigma)$. One can check that the entropy of $X$ is 
\begin{equation*}
H(X) = \frac{d}{2} ( 1 + \log 2\pi ) + \frac{1}{2} \log \det(\Sigma) .
\end{equation*}
As we can see, $H(X) \to -\infty$ if $\Sigma = \sigma^{2}I_{n}$ and $\sigma \to 0$.

\begin{definition}
[Conditional entropy]
\label{def:cond-entropy}
Let $X$ and $Y$ be two random variables with joint probability $p$ on $\mathbb{R}^{n} \times \mathbb{R}^{m}$. Then the \emph{conditional entropy}\index{Entropy!conditional} of $Y$ given $X$ is
\begin{equation}
\label{eq:conditional-entropy}
H(Y|X) := - \iint p(x,y) \log \frac{p(x,y)}{p_{X}(x)} \, dx dy ,
\end{equation}
where $p_{X}$ is the marginal probability of $X$. 
\end{definition}

Let $H(X,Y)$ be the joint entropy of $(X,Y)$, namely,
\begin{equation*}
H(X,Y) = - \iint p(x,y) \log p(x,y) dx dy .
\end{equation*}
Then there is
\begin{equation*}
H(Y|X) = H(X,Y) - H(X) .
\end{equation*}
It is easy to verify that $H(Y|X) = 0$ if and only if $Y$ is completely determined by $X$ (i.e., $Y$ is a function of $X$). We also have $H(Y|X) = H(Y)$ if and only if $X$ and $Y$ are independent.

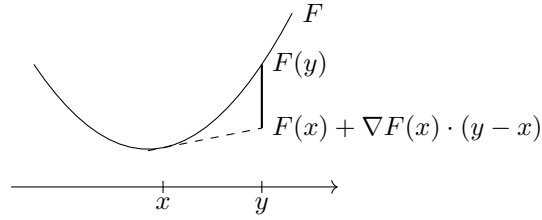
\begin{figure}
\centering
\begin{tikzpicture}[domain=-0.5:4.5, scale=1, transform shape]
	
	\draw[->] (-1.8, 0) -- (2.5, 0);

	\draw[thin, color=black] plot[domain=-1.5:1.9]({\x}, {\x*\x/2+1/2}) node[right]{\small{$F$}};
	\draw[thin, dashed, color=black] plot[domain=0:1.5]({\x}, {0.2*\x+0.48});
	
	\draw (1.5, 1.625) node[right]{\small{$F(y)$}};
	% \draw (1.5, 1.2) node[right]{\small{$D_{F}(y,x)$}};
	\draw (1.5, 0.78) node[right]{\small{$F(x) + \nabla F(x) \cdot (y-x)$}};
	
	\draw[thick, color=black] (1.5,0.78) -- (1.5,1.625);
			
	\draw[black] (0.2, -0.08) -- (0.2, 0.08);
	\draw[black] (1.5, -0.08) -- (1.5, 0.08);
	
	\draw[black] (0.2, 0) node[below]{\small{$x$}};
	\draw[black] (1.5, 0) node[below]{\small{$y$}};
\end{tikzpicture}
\caption{An illustration of the Bregman divergence associated with $F$ (the convex curve). The tangent line with gradient $\nabla F(x)$ passing through $(x,F(x))$ is shown by the dashed line. At $y$, the function $F$ and the tangent line take values $F(y)$ and $F(x)+ \nabla F(x) \cdot (y-x)$, respectively. Their difference (the length of the vertical solid line above the point $y$) is the Bregman divergence $D_{F}(y,x) = F(y) - F(x) - \nabla F(x) \cdot (y-x)$. Due to convexity, $F$ is on or above the tangent line everywhere, and hence $D_{F}(y,x) \ge 0$ for all $x,y$.}
\label{fig:bregman}
\end{figure}

\begin{definition}
[Bregman divergence]
\label{def:bregman}
Let $F$ be a convex functional with first variation $\delta F$. Then the \emph{Bregman divergence associated with $F$}\index{Bregman divergence} is defined by
\begin{equation*}
D_{F}(g, f) := F(g) - F(f) - \int \delta F(f) \cdot (g - f) ,
\end{equation*}
for any two functions $f$ and $g$.

Bregman divergences are mostly used in finite dimensional cases in practice. In these cases, $F: \Omega \to \mathbb{R}$ is a convex differentiable function and $\Omega \subset \mathbb{R}^{n}$ is a convex set. 
Then the \emph{Bregman divergence associated with $F$} is defined by
\begin{equation*}
D_{F}(y, x) := F(y) - F(x) - \nabla F(x) \cdot (y-x) ,
\end{equation*}
for any $x, y \in \Omega$. An illustration of the Bregman divergence $D_{F}(y,x)$ in the finite dimensional case is given in Figure \ref{fig:bregman}. 
\end{definition}

Bregman divergences can be defined when $F$ is not differentiable, in which case $\nabla F(x)$ can be replaced with any specified subgradient of $F$ at $x$. This subgradient should be clearly specified in the definition of the Bregman divergence. We will not consider this extension here.

A simple example of Bregman divergence is $F(x) = \frac{1}{2} |x|^{2}$ for $x \in \mathbb{R}^{n}$. In this case, one can verify that 
\begin{equation*}
D_{F}(y,x) = \frac{1}{2}|y-x|^{2} .
\end{equation*}

It is easy to verify the following properties of Bregman divergence associated with $F$, where $F$ is either a functional or function. We use the notations with $F$ being a function below.
\begin{enumerate}
\item
Non-negativity: $D_{F}(y,x) \ge 0$ for all $x,y$ because $F$ is convex. If $F$ is strictly convex, then $D_{F}(y,x) = 0$ if and only if $y = x$.

\item
Convexity in its first argument: $D_{F}(\cdot, x)$ is (strictly) convex for every $x$ if $F$ is (strictly) convex.

\item
Convexity in the first argument: for any $x,y,z$ and $\lambda \in [0,1]$, there is
\begin{equation*}
D_{F}((1-\lambda)y + \lambda z , x) \le (1-\lambda) D_{F}(y,x) + \lambda D_{F}(z,x) .
\end{equation*}
%In addition, we have the affine property:
%\begin{equation*}
%\nabla_{x} D_{F}((1-\lambda)y + \lambda z , x) = (1-\lambda) \nabla_{x} D_{F}(y,x) + \lambda \nabla_{x} D_{F}(z,x) .
%\end{equation*}
\end{enumerate}
However, Bregman divergences are not standard distance: They are not symmetric and do not satisfy the triangle inequality.

\begin{definition}
[Kullback--Leibler divergence]
\label{def:kl}
The \emph{Kullback--Leibler (KL) divergence}\index{Kullback--Leibler divergence} between two probability distributions $p$ and $q$ is defined as the Bregman divergence associated with the entropy functional 
\begin{equation*}
F(\rho):= - \int \rho(x) \log \rho(x) dx
\end{equation*}
for any $\rho \in P:= \{ \rho: \mathbb{R}^{n} \to \mathbb{R}: \rho \ge 0, \int \rho = 1 \}$. More precisely,
\begin{equation*}
\kl(p,q) := D_{F}(p,q) = \int p(x) \log \frac{p(x)}{q(x)} \, dx
\end{equation*}
If $F$ is the entropy of discrete random variables defined over $\Xcal$, then 
\begin{equation*}
\kl(p,q) := D_{F}(p,q) = \sum_{x \in \Xcal} p(x) \log \frac{p(x)}{q(x)} 
\end{equation*}
for any probability distributions $p$ and $q$ on $\Xcal$.
\end{definition}

Since the KL divergence is a special case of the Bregman divergence, it follows all properties of the latter.
In particular, we should be always aware that the KL divergence is non-negative and $\kl(p,q) = 0$ if and only if $p = q$ everywhere (because when $F$ is chosen as the entropy of $p$, we know $F$ is strictly convex).
Moreover, the KL divergence is \emph{not} a distance as it is not symmetric, i.e., $\kl (p,q)$ is not equal to $\kl (q,p)$ even if they both exist, and the KL divergence does not satisfy the triangle inequality.
Next, we take a closer look of the KL divergence and check more of its properties, as the KL divergence is prevalent in deep learning research and applications due to its ease of use when measuring the discrepancy between two probability distributions.

Assume that $p$ and $q$ are two probability distributions on $\mathbb{R}^{n}$. The KL divergence between them does not always exist due to the presence of logarithm, particularly when $p$ and $q$ do not have the same support.

If $p(x) > 0 $ and $q(x)=0$ at some $x \in \mathbb{R}^{n}$, then
\begin{equation*}
p(x) \log \frac{p(x)}{q(x)} = + \infty .
\end{equation*}
If the set of such points has positive measure in $\mathbb{R}^{n}$, then $\kl(p,q)$ is not defined because by its definition the integral does not exist. In this case, we cannot use the KL divergence to measure the difference between $p$ and $q$, nor take operations such as gradient of $\kl(p,q)$ either. 
This happens when $p(x) > 0$ occurs outside of the support of $q$.

If $p(x) = 0$ and $q(x) > 0$ at any $x \in \mathbb{R}^{n}$, then there is no such evaluation issue since
\begin{equation*}
p(x) \log \frac{p(x)}{q(x)} = 0 .
\end{equation*}
In this case, the KL divergence is defined because the integral in the KL is finite.

In summary, the use of $\kl(p,q)$ is restricted to the cases where the support of $p$ is contained in the support of $q$. Otherwise, we may use $\kl(q,p)$ or other distances or divergences between $p$ and $q$ to measure their difference.

We need to be particularly cautious about the aforementioned matter when we use KL divergence in deep learning models. 
More specificially, recall that in several generative models in Chapter \ref{chpt:gm}, we use KL divergence to measure the distance between a given data probability distribution $\pdata$ (usually in the form of a dataset $\Dcal = \{x_{i} \in \mathbb{R}^{d}: i \in [M]\}$, where $x_{i}$'s are samples drawn independently from $\pdata$) and a generated or pushforward probability $\ptheta$, where $\theta$ refers to the parameter of some deep neural network to be trained and $\ptheta$ is the probability induced by this network.
In this case, we often want to find the optimal parameter $\theta$ by solving
\begin{align*}
\min_{\theta} \ \Big\{ \kl (\pdata, \ptheta) = - H(\pdata) - \int \pdata(x) \log \ptheta(x) \, dx \Big \} .
\end{align*}
Since $\pdata$ is unknown and fixed, $H(\pdata)$ is independent of $\theta$, and the problem above reduces to 
\begin{align*}
\max_{\theta} \ \int \pdata(x) \log \ptheta(x) \, dx \quad \approx \quad \max_{\theta}\ \frac{1}{M} \sum_{i=1}^{M} \log \ptheta(x_{i}) ,
\end{align*}
where the approximation is obtained by Monte Carlo integration using the fact $x_{i} \sim \pdata$.
We can see that this requires $\ptheta(x_{i})>0$ for all $x_{i} \in \Dcal$. Meanwhile, $\ptheta$ often presents some 
continuity or smoothness property due to the use of neural networks. Therefore, it is very likely to have $\ptheta(x) > 0$ at $x$ outside of the support of $\pdata$. This will not cause computation issue, however, it may infer or generate samples that are not from $\pdata$.
This can cause inaccurate estimation of $\theta$ and even incorrect solution to the target problem. 
In such cases, proper modifications of the method or alternative approaches may be needed.

When the involved probabilities are of special types, KL divergence can be used without danger and often have closed forms. We present two examples of such cases.

\begin{example}
[KL divergence between two Gaussians]
Let $p_{1}$ and $p_{2}$ be two Gaussian probability distributions on $\mathbb{R}^{n}$. Specifically,
\[
p_1 = \mathcal{N}(\mu_1, \Sigma_1),
\qquad
p_2 = \mathcal{N}(\mu_2, \Sigma_2),
\]
where $\mu_1,\mu_2 \in \mathbb{R}^n$, and $\Sigma_1,\Sigma_2 \in \mathbb{R}^{n\times n}$ are symmetric positive definite. The KL divergence from $p_1$ to $p_2$ is defined as
\begin{equation}
\label{eq:kl-gauss}
\kl (p_1 , p_2)
= \int p_1(x)\log\frac{p_1(x)}{p_2(x)}\, dx 
= \mathbb{E}_{X \sim p_{1}} \Big[ \log \frac{p_{1}(X)}{p_{2}(X)} \Big] .
\end{equation}
Notice that the logarithm of the Gaussian probability density $\Ncal(x; \mu, \Sigma)$ is
\[
\log \Ncal(x; \mu, \Sigma)
= -\frac{n}{2}\log(2\pi)
-\frac{1}{2}\log\det(\Sigma)
-\frac{1}{2}(x-\mu)^\top\Sigma^{-1}(x-\mu).
\]
Substituting the expressions for $p_1$ and $p_2$ into \eqref{eq:kl-gauss} yields
\begin{align}
\log\frac{p_1(X)}{p_2(X)}
& = \frac{1}{2}\log\frac{\det(\Sigma_2)}{\det(\Sigma_1)} \label{eq:kl-gauss-ratio} \\
& \qquad -\frac{1}{2}\Big[
(X-\mu_1)^\top\Sigma_1^{-1}(X-\mu_1)
-
(X-\mu_2)^\top\Sigma_2^{-1}(X-\mu_2)
\Big]. \nonumber
\end{align}
Taking expectation of both sides of \eqref{eq:kl-gauss-ratio} with respect to $X \sim p_1 = \Ncal(\mu_{1}, \Sigma_{1})$, we evaluate each quadratic term above separately. Specifically, we have
\begin{align}
\mathbb{E}_{X \sim p_1}\!\left[(X-\mu_1)^\top\Sigma_1^{-1}(X-\mu_1)\right] 
& = \mathbb{E}_{X \sim p_1}\!\Big[\mathrm{tr}\Big((X-\mu_1)^\top\Sigma_1^{-1}(X-\mu_1)\Big)\Big] \nonumber \\
& = \mathbb{E}_{X \sim p_1}\!\Big[\mathrm{tr}\Big(\Sigma_1^{-1}(X-\mu_1)(X-\mu_1)^\top\Big)\Big] \nonumber \\
& = \mathrm{tr}\Big( \Sigma_1^{-1} \mathbb{E}_{X \sim p_1}\!\Big[(X-\mu_1)(X-\mu_1)^\top\Big] \Big) \nonumber \\
& = \mathrm{tr}(\Sigma_1^{-1}\Sigma_1) \nonumber \\
& = n. \label{eq:kl-gauss-q1}
\end{align}
Furthermore, we have
\[
X-\mu_2 = (X-\mu_1) + (\mu_1-\mu_2),
\]
which implies
\begin{equation}
\label{eq:kl-gauss-q2}
\mathbb{E}_{X \sim p_1}\!\left[(X-\mu_2)^\top\Sigma_2^{-1}(X-\mu_2)\right]
= \mathrm{tr}(\Sigma_2^{-1}\Sigma_1)
+ (\mu_1-\mu_2)^\top\Sigma_2^{-1}(\mu_1-\mu_2), 
\end{equation}
where the cross term vanishes due to $\mathbb{E}_{X \sim p_1}[X-\mu_1]=0$.

Combining \eqref{eq:kl-gauss}, \eqref{eq:kl-gauss-ratio}, \eqref{eq:kl-gauss-q1} and \eqref{eq:kl-gauss-q2}, the KL divergence admits the closed form
\begin{equation}
\label{eq:kl-gauss-final}
\mathrm{KL}(p_1 , p_2)
=
\frac{1}{2}\Big[
\log\frac{\det(\Sigma_2)}{\det(\Sigma_1)}
- n
+ \mathrm{tr}(\Sigma_2^{-1}\Sigma_1)
+ (\mu_2-\mu_1)^\top\Sigma_2^{-1}(\mu_2-\mu_1)
\Big].
\end{equation}
If $\Sigma_{1} = \Sigma_{2} = \Sigma$, then \eqref{eq:kl-gauss-final} reduces to 
\begin{equation*}
% \label{eq:kl-gauss-final-equal-var}
\mathrm{KL}(p_1 , p_2)
=
\frac{1}{2}(\mu_2-\mu_1)^\top\Sigma^{-1}(\mu_2-\mu_1).
\end{equation*}
In particular, $\mathrm{KL}(p_1, p_2)=0$ if and only if $\mu_1=\mu_2$ and $\Sigma_1=\Sigma_2$.
\end{example}

\begin{example}
[KL divergence between two Poissons]
Let $p_1 = \mathrm{Poisson}(\lambda_1)$ and $p_2 = \mathrm{Poisson}(\lambda_2)$ where $\lambda_1, \lambda_2 > 0$. The KL divergence from $p_1$ to $p_2$ is defined as
\begin{equation}
\label{eq:kl-poisson}
\mathrm{KL}(p_1 , p_2) = \sum_{x=0}^{\infty} p_1(x)\log\frac{p_1(x)}{p_2(x)}.
\end{equation}
Recall that the probability mass function of the Poisson distribution with parameter $\lambda$ is given by
\[
\text{Pr}(X=x)
= \frac{\lambda^x e^{-\lambda}}{x!}, \quad \text{for } x=0,1,2,\dots .
\]
Substituting this into \eqref{eq:kl-poisson} yields
\[
\kl(p_{1}, p_{2}) = \sum_{x=0}^{\infty} p_{1}(x)\log\frac{p_1(x)}{p_2(x)}
= \sum_{x=0}^{\infty} \frac{\lambda^x e^{-\lambda}}{x!} \Big(x\log\frac{\lambda_1}{\lambda_2}
+ \lambda_2 - \lambda_1 \Big) .
\]

Noticing that $\mathbb{E}[X] = \sum_{x=0}^{\infty} x \cdot \frac{\lambda^x e^{-\lambda}}{x!} = \lambda_{1}$ for $X \sim p_{1}$, we obtain the closed form expression
\[
\mathrm{KL}(p_{1}, p_{2}) = \lambda_1 \log\frac{\lambda_1}{\lambda_2}
+ \lambda_2 - \lambda_1.
\]
It is easy to verify that $\kl(p_{1}, p_{2}) = 0$ if and only if $\lambda_1 = \lambda_2$.
\end{example}

There are several variants of KL divergences that have symmetry and become well defined when $p$ and $q$ do not share the same support. The most notable one is the Jensen--Shannon divergence given as follows.

\begin{definition}
[Jensen--Shannon divergence]
\label{def:jsd}
The \emph{Jensen--Shannon (JS) divergence}\index{Jensen--Shannon divergence} between two probability distributions $p$ and $q$ is
\begin{equation}
\label{eq:jsd}
\js(p,q) := \frac{1}{2} \Big( \kl(p, r) + \kl(q, r) \Big)
\end{equation}
where $r = (p+q)/2$ is the pointwise average of $p$ and $q$. 
\end{definition}

Note that $r$ is also a probability distribution and its support contains both supports of $p$ and $q$. Therefore, $\js(p,q)$ is always well defined and takes a finite value. Indeed, one can check that $\js(p,q) \in [0, \log 2]$ for any $p$ and $q$. The maximum value $2\log 2$ occurs when the supports of $p$ and $q$ do not overlap. Note that, however, this is not necessarily an advantage because the generalized gradient of $\js(p,q)$ with respect to $p$ or $q$ becomes zero and we cannot update any of them (more precisely, the parameters of the neural network used to generate or pushforward them) using gradient-based optimization algorithms in practice.

Nevertheless, it is easy to check that $\js(p,q) \ge 0$ and $\js(p,q) = 0$ if and only if $p = q$. Moreover, unlike the KL divergence, the JS divergence is symmetric, namely, $\js(p,q) = \js(q,p)$ for any probability distributions $p$ and $q$ on $\mathbb{R}^{n}$.

Despite the simple variation of the JS divergence from the KL divergence and the few new properties, JS divergence is more difficult to use as it lacks a closed form for many typical probability distributions.
For example, there is no closed form of the JS divergence between two Gaussian distributions like the KL divergence in \eqref{eq:kl-gauss-final}, because $r$ in \eqref{eq:jsd} becomes a mixture of two Gaussian distributions and the KL divergence from $p$ (or $q$) to $r$ does not have a closed form.

There exists a distance to measure the difference between two probability distributions with mathematical elegance and overcome the issues of KL and JS divergences above. It is called the Wasserstein distance, which we define below.  
 
\begin{definition}
[Wasserstein distance]
\label{def:wasserstein-dist}
Let $P:=\{p: \mathbb{R}^{n} \to \mathbb{R}: p(x)\ge0, \forall\, x \in \mathbb{R}^{n}, \, \int p(x) dx = 1\}$ be the space of probability distributions on $\mathbb{R}^{n}$. For $s \in [1,\infty)$, the \emph{Wasserstein-$s$ distance}\index{Wasserstein distance} between $p, q \in P$ is defined by
\begin{equation}
\label{eq:wasserstein-p-dist}
W_{s}(p, q) := \inf_{\pi \in \Pi(p,q)} \Big( \int |x-y|^{s} \pi (x,y) \, dx \Big)^{1/s}, 
\end{equation}
where $\Pi(p,q)$ is the set of joint probability distributions on $\mathbb{R}^{n} \times \mathbb{R}^{n}$ with $p$ and $q$ as the marginal distributions, namely,
\begin{equation*}
\Pi(p, q) := \Big\{ \pi: \mathbb{R}^{d} \times \mathbb{R}^{d} \to [0,\infty) : 
\begin{array}{c}
p(x) = \int \pi(x,y) \, dy,\\
q(y) = \int \pi(x,y)\, dx,
\end{array}\ \forall \, x, y \in \mathbb{R}^{d} \Big\} .
\end{equation*}
For $s = \infty$, the Wasserstein-$\infty$ distance between $p, q \in P$ is defined by 
\begin{equation*}
W_{\infty} (p, q) := \lim_{s \to \infty} W_{s}(p,q) .
\end{equation*}
The optimal solution $\pi^{*}$ to \eqref{eq:wasserstein-p-dist} is called an \emph{optimal transport}\index{Optimal transport} from $p$ to $q$.
\end{definition}

As an example, the Wasserstein-2 distance between two Gaussian distributions $p_{1} = \Ncal(\mu_{1}, \Sigma_{1})$ and $p_{2}= \Ncal(\mu_{2}, \Sigma_{2})$ on $\mathbb{R}^{d}$ is shown to be 
\begin{equation*}
W_{2}(p_{1} , p_{2}) = \Big( |\mu_{1} - \mu_{2} |^{2} + \mathrm{tr}\Big( \Sigma_{1} + \Sigma_{2} - 2 (\Sigma_{1}^{1/2} \Sigma_{2} \Sigma_{1}^{1/2})^{1/2} \Big) \Big)^{1/2} .
\end{equation*}

Wasserstein distance exhibits many important mathematical properties. A basic one is that $W_{s}$ is a distance (see Definition \ref{def:distance}) in rigorous mathematical sense. Namely, $W_{s}$ satisfies the following properties:
\begin{itemize}
\item Positive definiteness: $W_{s}(p,q) \ge 0$ for all $p, q \in P$; and $W_{s}(p,q) = 0$ if and only if $p=q$ almost everywhere on $\mathbb{R}^{n}$. 

\item Symmetry: $W_{s}(p,q) = W_{s}(q,p)$ for any $p, q \in P$.

\item Triangle inequality: $W_{s}(p,q) \le W_{s}(p,r) + W_{s}(r,q)$ for any $p, q, r \in P$. 
\end{itemize}
In addition, $W_{s}(p,q)$ is well defined even if the supports of $p$ and $q$ are different. This is a significant advantage over KL divergence. Moreover, Wasserstein distances represent many useful physical interpretations. For example, the Wasserstein-1 distance, which is also called the \emph{earth mover's distance}, measures the optimal cost of moving probability mass $p$ to $q$ where the cost of moving a unit mass from location $x$ to $y$ is proportional to their distance $|x-y|$.

Wasserstein distances is a fundamental concept in the field of optimal transport, the theory of which has been extensively studied in the past decades. Optimal transport\index{Optimal transport} theory connects to many branches of mathematics, including geometric analysis, partial differential equations, probability theory, and optimal control. Many important results are presented in \cite{villani2008optimal,santambrogio2015optimal,ambrosio2021lectures}.

However, the main issue with Wasserstein distances is that they are difficult to compute numerically. This is due to the constraints involved in the definition of Wasserstein distances and high dimensionality of spaces they are used in practice. In simple cases where $p$ and $q$ are probability distributions on small finite set $\Xcal$, they are effectively normalized histograms, and \eqref{eq:wasserstein-p-dist} reduces to a linear programming problem that can be solved by existing optimization algorithms. The computation can be significantly accelerated by adding the entropy regularization of the joint probability distribution to the objective function in \eqref{eq:wasserstein-p-dist}, and the solution can be obtained solving the dual form of the problem using matrix re-scaling method called the Sinkhorn algorithm. However, this technique does not apply to high-dimensional problems or problems defined in contnuous spaces such as $\mathbb{R}^{n}$ due to computation infeasibility. In the past years, there are also deep learning methods developed to compute Wasserstein distances. For example, the problem \eqref{eq:wasserstein-p-dist} can be converted to its dual form, then the dual variables are parameterized as deep neural networks and trained to solve the dual problem, yielding the solution to \eqref{eq:wasserstein-p-dist}. Some of these results are included in \cite{peyre2019computational}. However, there is yet a computationally effective means to approximate Wasserstein distances accurately for deep learning models in general. This has hindered the extensive use of Wasserstein distances in deep learning.

The last concept to be introduced in this section is a classic one called mutual information. Unlike any of the divergences and distances above, mutual information exploits the joint probability distribution between random variables to measure their dependency. The formal definition of mutual information\index{Mutual information} is given as follows.

\begin{definition}
[Mutual information]
Let $X$ and $Y$ be two random variables with joint probability distribution $p(x,y)$.
The \emph{muutal information} between $X$ and $Y$ is defined by
\begin{equation}
\label{eq:mi}
\text{MI}(X,Y) = \kl ( p, p_{X}p_{Y} ) = \iint p(x,y) \frac{p(x,y)}{p_{X}(x)p_{Y}(y)} \, dx dy ,
\end{equation}
where $p_{X}$ and $p_{Y}$ are the marginal probability distributions of $X$ and $Y$, respectively.
\end{definition}
Notice that, the mutual information is non-negative as it is defined as the KL divergence from the joint probability distribution $p$ to the product of the marginal probability distributions $p_{X}p_{Y}$. 
Moreover, we can see $\text{MI}(X,Y) = 0$ if and only if $p = p_{X} p_{Y}$, namely, $X$ and $Y$ are independent.
Mutual information is also symmetric. However, we remark that mutual information is not a distance in any sense. 
Nevertheless, large $\text{MI}(X,Y)$ is an indication that $X$ and $Y$ are strongly dependent.
Indeed, we have the following relation between the mutual information of two random variables and their (conditional) entropy $H(X|Y)$:
\begin{align*}
\text{MI}(X, Y)
& = \int p(x, y) \log \frac{p(x, y)}{p_{X}(x) p_{Y}(y)} \, dx dy \\
& = \int p(x, y) \log \frac{p_{X|Y}(x | y)}{p_{X}(x)} \, dx dy \\
& = \int p(x, y) \log p_{X|Y} (x | y) \, dx dy - \int p_{X}(x) \log p_{X}(x) \, dx \\
& = H(X) - H(X | Y) .
\end{align*} 
This again implies that $\text{MI}(X,Y)=0$ if $X$ and $Y$ are independent in which case $H(X|Y) = H(X)$. The other extreme is that $X$ is completely determined by $Y$ (namely, $X$ is a function of $Y$), which implies $H(X|Y) = 0$ and $\text{MI}(X,Y) = H(X)$.

\section{Basics of Stochastic Differential Equations}
\label{appsec:sde}

Stochastic differential equations (SDEs)\index{Differential equation!stochastic} have become more prevalent in problem formulation, modeling and derivations in modern deep learning research and applications.
For example, we have seen some extensive uses of SDEs in diffusion models in Section \ref{sec:diffusion-models}.

In this section, we provide a brief review of some basics about SDEs. In particular, we cover the definitions of Wiener processes, stochastic integrals and differentials, the It\^{o} formula, and the Fokker--Planck equations.

Consider a particle (as a random walker) moving on the real line $\mathbb{R}$. At time $t=0$, it is at the origin. Then it moves with step size $\Delta x$ in every unit time $\Delta t$. Namely, it moves to its left (location changes by $-\Delta x$) or its right (location changes by $\Delta x$), with probability $\frac{1}{2}$ in either case, in every $\Delta t$. 

Let $X(t)$ denote the location of the particle at time $t = n \Delta t$ after moving $n$ steps. To study the probability distribution of $X(t)$, we consider $X_{i} \sim \text{Bernoulli}(\frac{1}{2})$ which are independent for $i=1,2,\dots$, and hence $X_{i}=1$ or $0$ with probability $\frac{1}{2}$ in either case. It is clear that $\mathbb{E}[X_{i}] = \frac{1}{2}$ and $\var[X_{i}] = \frac{1}{4}$. 

Let $S_{n}: = \sum_{i=1}^{n} X_{i}$, then $S_{n}$ is the number of times the particle moved right in the first $n$ steps, and $n- S_{n}$ is the number of times it moved left. 
We see that $\mathbb{E}[S_{n}] = \frac{n}{2}$ and $\var[S_{n}] = \frac{n}{4}$.
Furthermore, we have 
\begin{equation*}
X(t) = S_{n}\Delta x - (n-S_{n}) \Delta x = (2 S_{n} - n) \Delta x .
\end{equation*}
Therefore, $\mathbb{E}[X(t)] = 0$ and 
\begin{align*}
\var[X(t)] 
& = (\Delta x)^{2} \var[2S_{n} - n ] = (\Delta x)^{2} 4 \var[S_{n}] = (\Delta x)^{2} n = \frac{(\Delta x)^{2}}{\Delta t} t .
\end{align*}
Without loss of generality, we assume $\frac{(\Delta x)^{2}}{\Delta t} = 1$, and hence it is easy to verify that
\begin{equation*}
X(t) = (2S_{n} - n) \Delta x = \frac{S_{n} - \frac{n}{2}}{\sqrt{\frac{n}{4}}} \sqrt{t} .
\end{equation*}
Now for every fixed $t$, we let $n \to \infty$ while keeping $\frac{(\Delta x)^{2}}{\Delta t} = 1$, then obtain by Laplace-De Moivre theorem that
\begin{align*}
\lim_{n \to \infty} \text{Pr}(a \le X(t) \le b) 
& = \lim_{n \to \infty} \text{Pr} \Big( \frac{a}{\sqrt{t}} \le \frac{S_{n} - \frac{n}{2}}{\sqrt{\frac{n}{4}}} \le \frac{b}{\sqrt{t}} \Big) \\
& = \frac{1}{\sqrt{2\pi}} \int_{a/\sqrt{t}}^{b/\sqrt{t}} e^{-\frac{x^{2}}{2}} \, dx \\ 
& = \frac{1}{\sqrt{2\pi t}} \int_{a}^{b} e^{-\frac{x^{2}}{2t}} \, dx ,
\end{align*}
which implies that $X(t) \sim \Ncal(0, t)$.
In other words, if the particle is moving in the sense of continuous time, then its location $X(t) \sim \Ncal(0,t)$. This yields the definition of the \emph{Wiener process}\index{Weiner process}, also commonly called the \emph{Brownian motion}\index{Brownian motion}, as follows.

\begin{definition}
[Wiener process]
A real-valued stochastic process $W: [0,\infty) \to \mathbb{R}$ is called a \emph{Wiener process} if
\begin{enumerate}
\item
Start from the origin: $W(0) = 0$ almost surely;

\item
Independent increment: for any $0 < t_{1} < t_{2} < \dots < t_{k}$, the random variables
\begin{equation*}
W(t_{1}),\ W(t_{2}) - W(t_{1}),\ \dots,\ W(t_{k}) - W(t_{k-1})
\end{equation*}
are independent;

\item Normal distribution of increments: for any $t > s \ge 0$, there is $W(t)-W(s) \sim \Ncal(0, t-s)$; and

\item Continuous path: $t \mapsto W(t)$ is continuous in $t$.
\end{enumerate}
\end{definition}
Notice that $\mathbb{E}[W(t)] = 0$ and $\var[W(t)] = \mathbb{E}[W(t)^{2}] = t$ for any $t$ as $W(t) \sim \Ncal (0, t)$.

To rigorously define SDEs, we need a series of mathematical concepts from measure theory. We only go through them in brief below and refer interested to \cite{oksendal2010stochastic,evans2012introduction,mao2007stochastic} for more details.

Let $(\Omega, \Gamma, \mu)$ be a probability space, where $\Omega$ is taken as $\mathbb{R}^{d}$ here, $\Gamma$ is a $\sigma$-algebra on $\Omega$, and $\mu$ is a probability measure such that all elements of $\Gamma$ are measurable under $\mu$. Here a $\sigma$-algebra of $\Omega$ is defined as a set of subsets of $\Omega$, and it contains the empty set $\emptyset$ and is closed under complement and countable union of its elements in $\Omega$. Let $A$ be a collection of subsets of $\Omega$, we call $\Gamma(A)$ the $\sigma$-algebra generated by $A$ if $\Gamma(A)$ is the intersection of all $\sigma$-algebras of $\Omega$ containing $A$. It can be shown that $\Gamma(A)$ is indeed a $\sigma$-algebra as well. As random variables can be considered equivalent to $\mu$-measurable functions, we can also define $\Gamma(\{X_{\alpha}: \alpha \in \Acal\})$ as the $\sigma$-algebra generated by the random variables $\{X_{\alpha}: \alpha \in \Acal\}$ in $\Omega$, where $\Acal$ is an index set which can be infinite, and $X_{\alpha}$ here refers to the set of all pre-images of Lebesgue measurable sets of $\mathbb{R}$ under the mapping $X_{\alpha}$. 

We denote $\Wcal(t):= \Gamma(\{W(s): 0 \le s \le t\})$ the $\sigma$-algebra generated by the Wiener process up to time $t$. We also call $\Wcal(t)$ the history of the Wiener process $W$ up to time $t$. We let the generated $\sigma$-algebra $\Wcal^{+}(t):=\Gamma(\{W(s)-W(t): s \ge t\})$ denote the future of the Wiener process $W$ beyond time $t$.
We call the family of $\sigma$-algebras, denoted by $\{\Fcal(t) \subset \Gamma: t \ge 0\}$, non-anticipating (also called adapted) with respect to $W$ if (i) $\Fcal(s) \subset \Fcal(t)$ for all $0 \le s < t$; (ii) $\Wcal(t) \subset \Fcal(t)$ for all $t \ge 0$; and (iii) $\Fcal(t)$ is independent of $\Wcal^{+}(t)$ for all $t \ge 0$. We also call $\Fcal(t)$ a filtration up to time $t$.
We can informally interpret $\Fcal(t)$ as the $\sigma$-algebra that ``contains all information up to time $t$'', in the sense that all random variables generated by the Wiener process up to time $t$ and some initial random variables independent of the Wiener process are measurable under $\Fcal(t)$ for any $t$.

We call a real-valued stochastic process $G$ non-anticipating with respect to $\Fcal$ if $G(t)$ is $\Fcal(t)$-measurable for any $t \ge 0$. 
In addition, we denote by $L^{2}(0,T)$ the space of all real-valued non-anticipating stochastic process $G$ with $\mathbb{E}[ \int_{0}^{T} |G(t)|^{2} dt ] < \infty$.
Similarly, we denote by $L^{1}(0,T)$ the space of all real-valued non-anticipating stochastic process $F$ with $\mathbb{E}[ \int_{0}^{T} |F(t)| dt ] < \infty$.
A special type of stochastic processes in $L^{2}(0,T)$ is called step processes: let $0 = t_{0} < t_{1} < \dots < t_{K} =T$ be a partition of $[0,T]$ and $G(t) = G_{k}$ for every $t \in [t_{k}, t_{k+1})$ for $t = 0,\dots, K-1$ where $G_{k}$ is an $\Fcal(t_{k})$-measurable random variable, then $G$ is a step process.
For a step process $G \in L^{2}(0,T)$ given as such, we define its It\^{o} stochastic integral\index{It\^{o} stochastic integral} on $(0,T)$ by
\begin{equation*}
\int_{0}^{T} G(t) \, d W(t) := \sum_{k=0}^{K-1} G_{k} (W(t_{k+1}) - W(t_{k})) ,
\end{equation*}
which is a random variable.
It can be shown that, for any $G \in L^{2}(0,T)$, it can be approximated by a sequence of step processes $G_{n} \in L^{2}(0,T)$ in the sense of
\begin{equation*}
\mathbb{E}\Big[ \int_{0}^{T} |G - G_{n}|^{2} \, dt \Big] \to 0
\end{equation*}
as $n \to \infty$. Here $G_{n}$'s may have different partitions on $[0,T]$ and take different random variables over the partition subintervals. 
Then the following properties of It\^{o} stochastic integrals can be proved for step processes and then extended to all stochastic processes in $L^{2}(0,T)$: for any $a,b \in \mathbb{R}$ and $G, H \in L^{2}(0,T)$,
\begin{align*}
\int_{0}^{T} (a G(t) + b H(t)) \, dW(t) & = a \int_{0}^{T} G(t) \, dW(t) + b \int_{0}^{T} H(t) \, dW(t) , \\ 
\mathbb{E} \Big[ \int_{0}^{T} G(t)  \, dW(t) \Big] & = 0 , \\
\mathbb{E} \Big[ \int_{0}^{T} G(t) \, dW(t) \cdot \int_{0}^{T} H(t) \, dW(t) \Big] & = \int_{0}^{T} G(t) \cdot H(t) \, d t .
\end{align*}

Given the definition of It\^{o} stochastic integrals above, we can define real-valued stochastic process $X$ as follows:
\begin{equation*}
X(t) = X(0) + \int_{0}^{t} f(s,X(s)) \, ds + \int_{0}^{t} g(s,X(s)) \, dW(s)
\end{equation*}
for any $t \in [0,T]$, where $f , g: [0, T] \times \mathbb{R} \to \mathbb{R}$ are such that $f(\cdot,X(\cdot)) \in L^{1}(0,T)$ and $g(\cdot,X(\cdot)) \in L^{2}(0,T)$. In this case, we also say that $X$ has the stochastic differential
\begin{equation}
\label{appeq:sde}
dX(t) = f(t,X(t)) \, dt + g(t, X(t))\, dW(t) 
\end{equation}
for all $t \ge 0$ with initial value $X(0)$. We also call that $X$ follows the SDE \eqref{appeq:sde}.
The SDE can be extended to multi-dimensional cases where $X(t) \in \mathbb{R}^{d}$, $W(t) = (W_{1}(t), \dots, W_{m}(t))^{\top} \in \mathbb{R}^{m}$ where $W_{i}(t)$ is a one-dimensional Wiener process and the $m$ Wiener processes in $W$ are independent (we call $W$ the standard $m$-dimensional Wiener process), $f: [0,T] \times \mathbb{R}^{d} \to \mathbb{R}^{d}$, and $g: [0,T] \times \mathbb{R}^{d} \to \mathbb{R}^{d \times m}$ which also satisfy the integrability property. We often call $f$ the drift and $g$ the diffusion coefficient.

One of the most important properties of stochastic calculus is known as the It\^{o} formula\index{It\^{o} formula} (also called the It\^{o} chain rule\index{It\^{o} chain rule}), given as follows, and the proof of which can be found in, e.g., \cite{evans2012introduction,oksendal2010stochastic}.

\begin{lemma}
[It\^{o} formula]
Suppose $X$ follows the SDE \eqref{appeq:sde} and $u: [0,T] \times \mathbb{R} \to \mathbb{R}$ is continuous (and $\partial_{t}u$, $\partial_{x} u$ and $\partial_{xx}^{2} u$ are all continuous). Let $Y(t):= u(t, X(t))$, then $Y$ follows
\begin{equation}
\label{appeq:Ito-formula}
dY(t) = \Big( \partial_{t}u + \partial_{x}u \cdot f + \frac{1}{2} g^{2} \partial_{xx} u \Big) dt + (\partial_{x} u \cdot g) dW(t) 
\end{equation}
where we use $u=u(t,X(t))$, $f=f(t,X(t))$ and $g = g(t,X(t))$ for notation simplification hereafter.

The multi-dimensional version of the formula \eqref{appeq:Ito-formula} is given by 
\begin{equation}
\label{appeq:Ito-formula-gen}
dY(t) = \Big( \partial_{t}u + \partial_{x}u \cdot f + \frac{1}{2} \langle \nabla^{2}u, gg^{\top} \rangle \Big) dt + \partial_{x} u \cdot g \cdot dW(t) ,
\end{equation}
where $X(t) \in \mathbb{R}^{d}$, $W(t) \in \mathbb{R}^{m}$ is the standard $m$-dimensional Wiener process, $f: [0,T] \times \mathbb{R}^{d} \to \mathbb{R}^{d}$, $g: [0,T] \times \mathbb{R}^{d} \to \mathbb{R}^{d \times m}$, $u: [0,T] \times \mathbb{R}^{d} \to \mathbb{R}$, $\nabla^{2} u$ is the Hessian of $u$, and $\langle A, B \rangle := \mathrm{tr}(A^{\top} B) = \sum_{i,j=1}^{d} A_{ij} B_{ij}$ for matrices $A=[A_{ij}]_{i,j=1}^{d}$ and $B=[B_{ij}]_{i,j=1}^{d}$.
\end{lemma}

We can take the following example to see how the It\^{o} formula is used to find stochastic integrals. However, note that it is usually difficult to evaluate stochastic integrals analytically in general.

\begin{example}
Find
\begin{equation*}
\int_{0}^{T} W(t) \, dW(t) .
\end{equation*}
Suppose $X(t) = W(t) \in \mathbb{R}$ and $u(x) :=  x^{2}$. Then $\nabla u(x) = 2x$ and $\nabla^{2} u(x) = 2$. By It\^{o} formula, we have
\begin{align*}
d (W(t)^{2}) 
& = d \, u(X(t)) \\
& = \Big( \nabla u(X(t)) \cdot 0 + \frac{1}{2} \nabla^{2} u(X(t)) \cdot 1 \Big)\, dt + \nabla u(X(t)) \cdot 1 \, dW(t) \\
& =  dt + 2W(t)\, dW(t) .
\end{align*}
Taking integral of both sides of on $(0,T)$, we obtain
\begin{equation}
\label{appeq:Ito-ex}
\int_{0}^{T} d (W(t)^{2}) = \int_{0}^{T}  dt + 2 \int_{0}^{T} W(t) \, dW(t) .
\end{equation}
Noticing that the left-hand side of \eqref{appeq:Ito-ex} is
\begin{equation*}
\int_{0}^{T} d (W(t)^{2}) = W(T)^{2} - W(0)^{2} = W(T)^{2} ,
\end{equation*}
and the right-hand side of \eqref{appeq:Ito-ex} is
\begin{equation*}
T + 2\int_{0}^{T} W(t) \, dW(t) .
\end{equation*}
Therefore, rearranging \eqref{appeq:Ito-ex} yields
\begin{equation*}
\int_{0}^{T} W(t) \, dW(t) = \frac{W(T)^{2}}{2} - \frac{T}{2} .
\end{equation*}
\end{example}

Now we consider from the probability perspective of $X(t)$. Suppose $X(t)$ in SDE \eqref{appeq:sde} follows the probability $\rho(\cdot, t)$ for any $t$. We would like to find the property of $\rho(\cdot, t)$.

To this end, we take a heuristic but less rigorous derivation. The conclusion with the Fokker--Planck equation remains correct. For a more rigorous derivation, we refer interested readers to \cite{risken1985fokker-planck}.

Let $\phi : \mathbb{R}^{d} \to \mathbb{R}$ be any smooth test function. Suppose we have $N$ independent particles $X_{i}(t)$ following $\rho(\cdot, t)$ for any $t$. Then we have the following approximation by Monte Carlo integration 
\begin{equation}
\label{appeq:fpe-integral}
\int \rho(x,t) \phi(x) \, dx  \approx \frac{1}{N} \sum_{i=1}^{N} \phi(X_{i}(t)) .
\end{equation}
We take the time derivative on both sides, then the left-hand side of \eqref{appeq:fpe-integral} becomes
\begin{equation}
\label{appeq:fpe-integral-left}
\frac{d}{dt} \int \rho(x,t) \phi(x) \, dx = \int \partial_{t} \rho(x,t) \phi(x) \, dx .
\end{equation}
By It\^{o} formula, we have
\begin{align}
d\, \phi(X_{i}(t)) 
& = \Big( \nabla \phi(X_{i}(t)) f_{t}(X_{i}(t)) + \frac{1}{2} \langle \nabla^{2} \phi(X_{i}(t)),  g_{t}(X_{i}(t)) g_{t}(X_{i}(t))^{\top} \rangle \Big) dt \nonumber \\
& \qquad + \nabla \phi(X_{i}(t)) f_{t}(X_{i}(t)) dW(t) , \label{appeq:fpe-d-rhs}
\end{align}
where we write $f(t,\cdot)$ and $g(t,\cdot)$ as $f_{t}(\cdot)$ and $g_{t}(\cdot)$, respectively, for short.
We will also write $\rho(\cdot,t)$ as $\rho_{t}(\cdot)$ below.
Plugging \eqref{appeq:fpe-d-rhs} to the time derivative of the right-hand side of \eqref{appeq:fpe-integral}, and taking the continuum limit by letting $N\to\infty$, we obtain
\begin{align*}
\int \rho_{t}(x) \Big( \nabla \phi(x) \cdot f_{t}(x) + \frac{1}{2} \langle \nabla^{2} \phi(x), g_{t}(x) g_{t}(x)^{\top} \rangle \Big) \, dx .
\end{align*}
Taking integration by parts, we obtain 
\begin{equation}
\label{appeq:fpe-integral-right}
\int \Big( - \nabla \cdot(\rho_{t}(x) f_{t}(x)) + \frac{1}{2} \langle \nabla^{2} , g_{t}(x)g_{t}(x)^{\top} \rho(x,t) \rangle \Big) \phi(x) \, dx ,
\end{equation}
where 
\begin{equation*}
\langle \nabla^{2}, A(x) \rangle := \sum_{i,j=1}^{d} \partial_{x_{i}x_{j}}^{2} A_{ij}(x) 
\end{equation*}
for any matrix-valued function $A(x) = [A_{ij}(x)]_{i,j=1}^{d}$ of $x \in \mathbb{R}^{d}$.
Now we know \eqref{appeq:fpe-integral-left} and \eqref{appeq:fpe-integral-right} are equal for any $\phi$, which implies that
\begin{equation}
\label{appeq:fpe}
\partial_{t} \rho_{t}(x) = - \nabla \cdot(\rho_{t}(x) f_{t}(x)) + \frac{1}{2} \langle \nabla^{2}, g_{t}(x)g_{t}(x)^{\top} \rho_{t}(x) \rangle .
\end{equation}
We call \eqref{appeq:fpe} the \emph{Fokker--Planck equation}\index{Fokker--Planck equation} of $\rho$.

Notice that in many manually designed SDEs, such as those in diffusion models in Section \ref{sec:diffusion-models}, the diffusion coefficient is set to be state-independent function $g_{t} = \sigma_{t} I_{d}$ where $\sigma_{t} \in \mathbb{R}$ may depend on time $t$ but not state $x$, and then the Fokker--Planck equation simplifies to
\begin{equation*}
\partial_{t} \rho_{t}(x) = - \nabla \cdot(\rho_{t}(x) f_{t}(x)) + \frac{\sigma_{t}^{2}}{2} \Delta \rho_{t}(x) .
\end{equation*}

In the case where no diffusion presents, i.e., $g=0$, we obtain the so-called \emph{continuity equation}\index{Continuity equation}:
\begin{equation}
\label{appeq:ce}
\partial_{t} \rho_{t}(x) = - \nabla \cdot(\rho_{t}(x) f_{t}(x) ) ,
\end{equation}
and the corresponding SDE \eqref{appeq:sde} reduces to the ODE
\begin{equation*}
\xdot(t) = f(t,x(t)) .
\end{equation*}
The equation \eqref{appeq:ce} has appeared several times in this book, such as in the Neural ODE method in Section \ref{sec:node} and the flow matching method in Section \ref{sec:flow-matching}.
As a side note, continuity equations generally refer to the class of equations that follow certain conservation laws, and \eqref{appeq:ce} is a particular example that conserves probability.

\bibliographystyle{abbrv}
\bibliography{library_mfdl}

\printindex

\end{document}